\documentclass{article}

\PassOptionsToPackage{numbers, compress}{natbib}

\usepackage[preprint]{neurips_2026}
\makeatletter
\renewcommand{\@noticestring}{Accepted at NeurIPS 2026.}
\makeatother

\usepackage[utf8]{inputenc} %
\usepackage[T1]{fontenc}    %
\usepackage{hyperref}       %
\hypersetup{
  bookmarksdepth=2,
  pdftitle={Full Fine-Tuning is not the Problem: Why Adam Fails and SGD Succeeds in Few-Shot CLIP Adaptation},
  pdfauthor={Nairouz Mrabah, Ismail Ben Ayed, Eric Granger}
}
\usepackage{url}            %
\usepackage{booktabs}       %
\usepackage{amsfonts}       %
\usepackage{nicefrac}       %
\usepackage{microtype}      %
\usepackage{xcolor}         %
\usepackage[most]{tcolorbox} %
\usepackage{graphicx}
\usepackage{pifont}
\usepackage{tabularx}
\usepackage{subcaption}   %
\usepackage{algorithm}
\usepackage{algpseudocode}
\usepackage{adjustbox}
\usepackage{array}
\usepackage{multirow}
\usepackage{amsmath}
\usepackage{makecell}
\usepackage{colortbl}
\usepackage{dsfont}
\usepackage{amssymb}
\usepackage{amsthm}
\usepackage{siunitx}
\usepackage{longtable}

\definecolor{findingblueframe}{RGB}{41,98,180}
\definecolor{findingblueback}{RGB}{236,245,255}
\newtcolorbox{findingbox}[1][Finding]{%
  enhanced,
  breakable,
  colback=findingblueback,
  colframe=findingblueframe,
  colbacktitle=findingblueframe,
  coltitle=white,
  boxrule=0.7pt,
  arc=1.5pt,
  left=6pt,
  right=6pt,
  top=5pt,
  bottom=5pt,
  before skip=7pt,
  after skip=7pt,
  fonttitle=\bfseries\small,
  fontupper=\small,
  title={#1}}

\title{Full Fine-Tuning is not the Problem: Why Adam Fails and SGD Succeeds in Few-Shot CLIP Adaptation}

\author{%
  Nairouz Mrabah \qquad Ismail Ben Ayed \qquad Eric Granger \\
  LIVIA, ILLS, Department of Systems Engineering, ETS Montreal, Canada
}

\begin{document}

\maketitle
\addtocontents{toc}{\protect\setcounter{tocdepth}{-1}}
\begin{abstract}
Few-shot CLIP adaptation is often treated as a parameterization problem. Full fine-tuning is assumed to overfit, so parameter-efficient fine-tuning (PEFT) methods are preferred. In this paper, we show that this conclusion conflates parameterization with optimization. Across 11 datasets, 3 CLIP backbones, and 5 shot levels, dense full fine-tuning with plain SGD outperforms representative PEFT methods under the reported evaluation protocol, whereas Adam/AdamW collapses. We trace this collapse to unreliable support-set preconditioning. The few-shot second moment systematically underestimates the corresponding full-data statistic, already at initialization. Adam’s inverse-square-root normalization converts this error into an oversized adaptive update. Matching Adam’s first-step update budget to SGD resolves most of the collapse, whereas numerator-side alignment does not explain the failure. Shampoo and SOAP also fail, so richer preconditioning alone does not resolve the problem. Theory formalizes how few-shot estimation error is amplified into unsafe updates. SGD succeeds because it avoids unreliable preconditioners and stays close to pretrained CLIP. Interpolation probes show low-loss connections along tested directions, and curvature probes show favorable directional curvature. These results identify optimizer-induced preconditioning, not dense full fine-tuning itself, as the main source of failure in few-shot CLIP adaptation. Code is available at \url{https://github.com/nairouz/few-shot-clip-sgd}.
\end{abstract}

\section{Introduction}
\label{sec:introcution}

Foundation vision-language models (VLMs) such as CLIP align images and text in a shared embedding space and support zero-shot recognition through natural-language prompts. Previous few-shot methods avoid updating the full backbone and train a small set of parameters. This design is often motivated by the common belief that updating all parameters leads to overfitting under scarce labels~\cite{mrabah2025so, silva2024closer, zhu2022prograd}. Prompt-learning methods replace hand-crafted templates with learnable or instance-conditional context vectors, as in CoOp~\cite{zhou2022coop}. Adapter methods~\cite{houlsby2019parameter} and low-rank reparameterization methods~\cite{hu2022lora} provide complementary lightweight modules. These parameter-efficient fine-tuning (PEFT) methods are now the standard way to adapt CLIP under limited supervision.

A dominant narrative in the literature is that full fine-tuning of CLIP is impractical in few-shot regimes and is consistently outperformed by PEFT approaches due to overfitting. For instance, the authors of CLIP-Adapter~\cite{gao2024clipadapter} claim that “\textit{the previous widely-adopted pretrain-fine-tuning paradigm would fail in fine-tuning the whole CLIP backbone under the few-shot setting due to the enormous amount of parameters and the shortage of training examples.}” The authors of~\cite{farina2025rethinking} argue that “\textit{All of the methods above consistently outperform full fine-tuning and linear probing.}” In~\cite{hu2025lora}, the authors reiterate that “\textit{full fine-tuning performs the worst, as it tends to overfit when tuning large models with extremely few examples.}” Similarly, the authors of CoOp~\cite{zhou2022coop} conclude that “\textit{Obviously, fine-tuning the image encoder does not work well in low-shot regimes}”. These statements, among several similar claims~\cite{khattak2023maple,nguyen2024saft,silva2024closer}, have strongly shaped the prevailing belief that PEFT is not only more efficient but also more effective than full fine-tuning in few-shot CLIP adaptation.

\begin{figure*}[!t]
  \centering
  \includegraphics[width=\linewidth]{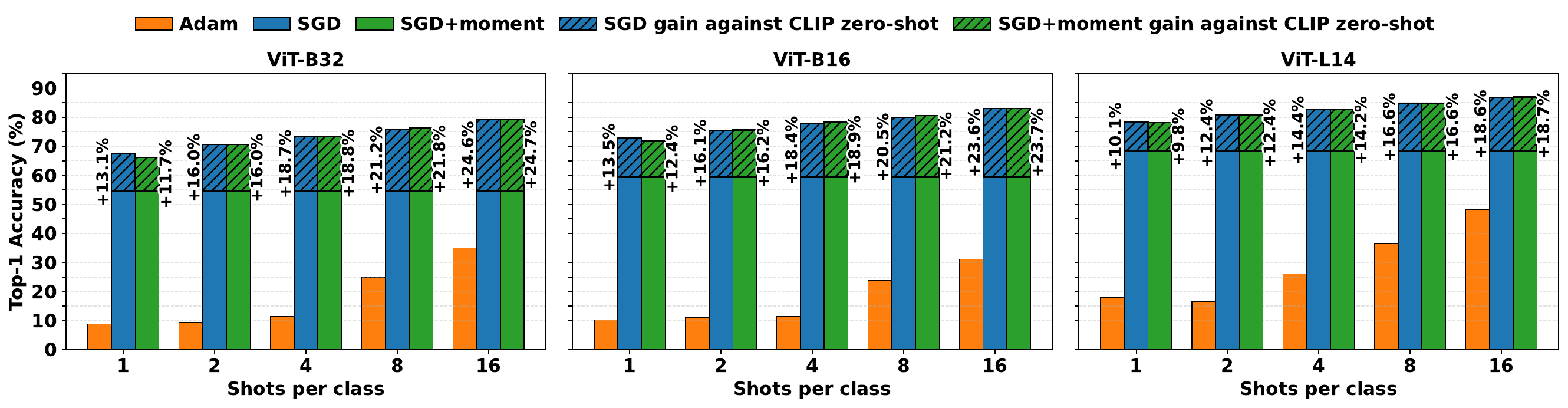}
  \caption{Average test-set accuracy using SGD, SGD with momentum, and Adam for few-shot CLIP adaptation over 11 datasets. Hatched labels report relative accuracy gains over zero-shot CLIP.}
  \label{fig:fewshot_sgd_vs_adam}
\end{figure*}

\begin{figure*}
  \centering
  \includegraphics[width=\linewidth]{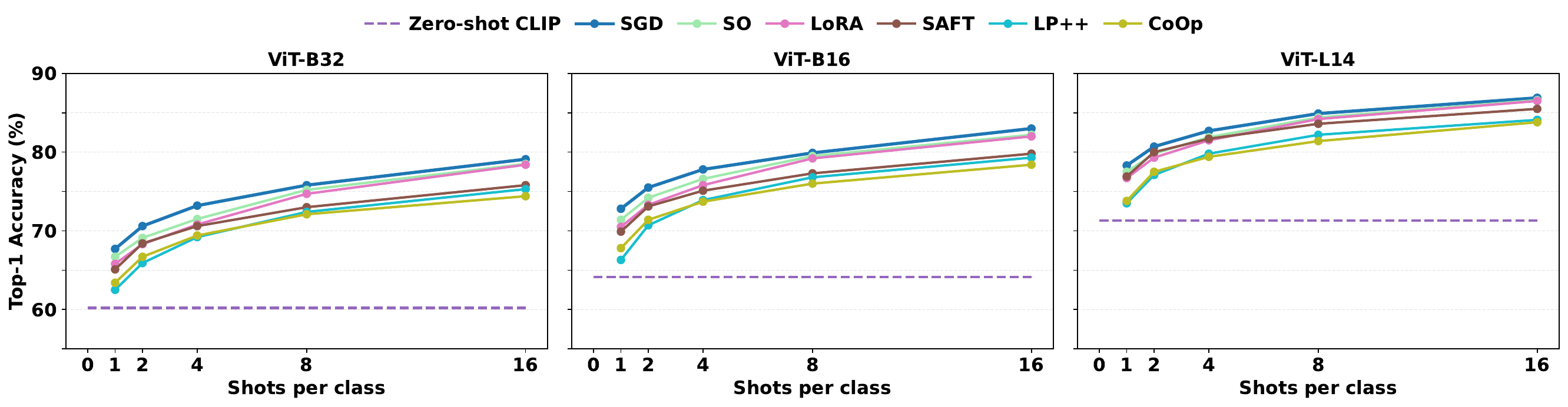}
  \caption{Average test-set accuracy using SGD compared with PEFT methods for few-shot CLIP adaptation over 11 datasets.}
  \label{fig:fewshot_sgd_vs_peft}
\end{figure*}

A closer look, however, reveals that this conclusion may stem from experiments run with adaptive preconditioned optimizers such as Adam and AdamW. Fig.~\ref{fig:fewshot_sgd_vs_adam} summarizes few-shot results averaged over 11 datasets and 3 random seeds for SGD, SGD with momentum (SMD), and Adam across three CLIP backbones (ViT-B/32, ViT-B/16, and ViT-L/14) and five shot levels (1, 2, 4, 8, and 16). Under a fixed set of hyperparameters (30 epochs, a batch-size of 64, and a learning rate of $10^{-4}$) shared across datasets, backbones, and shot levels, SGD consistently and substantially outperforms Adam at every shot level and on every backbone. The gap is particularly large in the lowest-shot settings, where Adam often performs far below the zero-shot baseline, while SGD achieves significant gains. As illustrated in Fig.~\ref{fig:fewshot_sgd_vs_adam}~(\textcolor{blue}{\hyperref[sec:sgd_smd_adam]{Appendix~\ref*{sec:sgd_smd_adam}}} for details), this behavior is traced to Adam’s second-moment state. Replacing Adam with SMD restores performance, indicating that the second-moment term is the primary driver of performance collapse. SMD also matches plain SGD performance, indicating that the first-moment term provides little to no benefit while introducing additional memory overhead. With fixed method-specific training configurations, full fine-tuning with SGD outperforms representative PEFT methods (e.g., SO~\cite{mrabah2025so}, LoRA~\cite{hu2022lora}, SAFT~\cite{nguyen2024saft}, LP++~\cite{huang2024lpplusplus}, and CoOp~\cite{zhou2022coop}) across all shot levels and backbones, as shown in Fig.~\ref{fig:fewshot_sgd_vs_peft} (\textcolor{blue}{\hyperref[app:sgd_vs_peft]{Appendix~\ref*{app:sgd_vs_peft}}} has further results and details).

\begin{figure*}
  \centering
  \includegraphics[width=\linewidth]{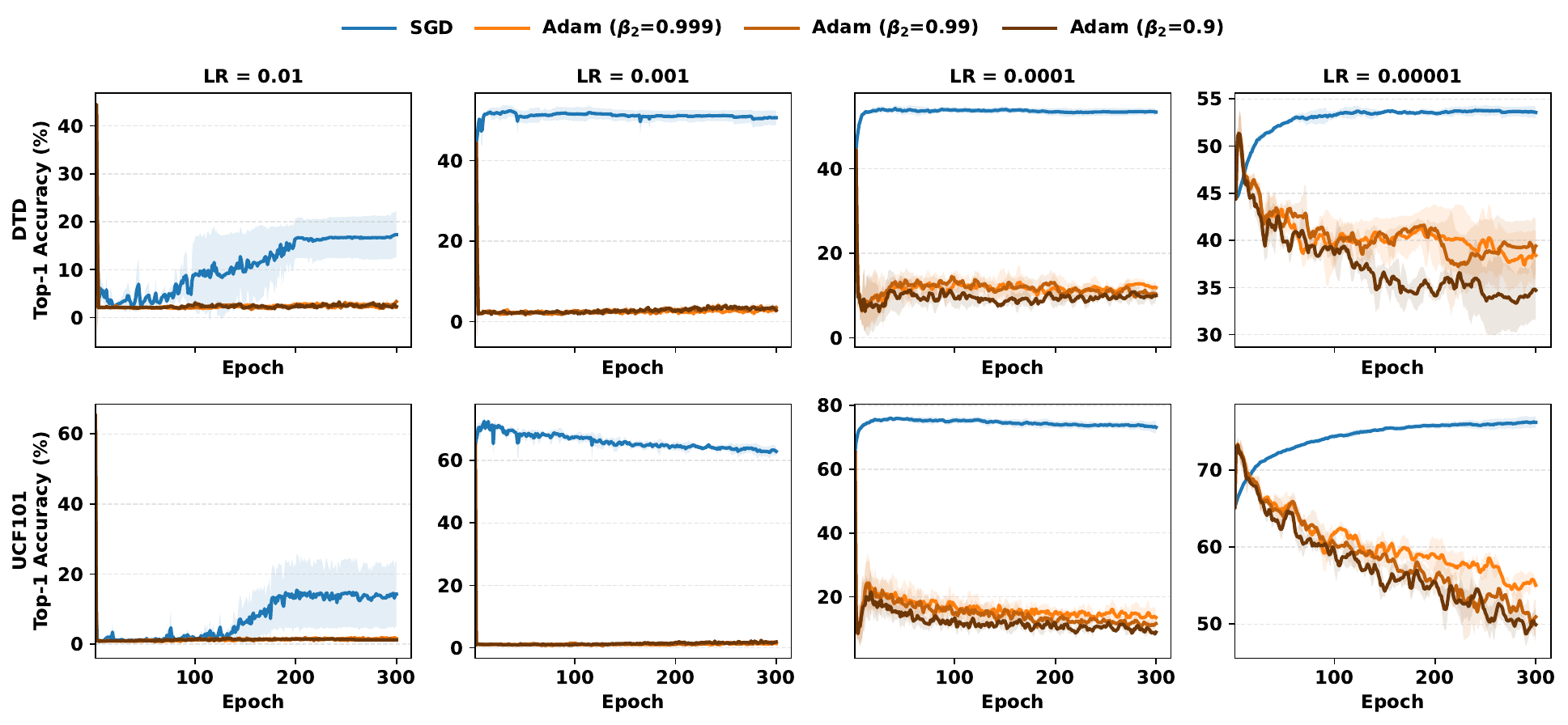}
  \caption{Average test-set accuracy (mean $\pm$ std with 3 fixed random seeds) over training epochs for SGD, Adam ($\beta_2=0.999$), Adam ($\beta_2=0.99$), and Adam ($\beta_2=0.9$) in a 1-shot setting using the DTD and UCF101 datasets and CLIP (ViT-B/16).}
  \label{fig:fewshot_curves_shots1_dtd_ucf}
\end{figure*}

Fig.~\ref{fig:fewshot_curves_shots1_dtd_ucf} (\textcolor{blue}{\hyperref[app:stability_analysis]{Appendix~\ref*{app:stability_analysis}}} for further results) compares dense SGD and dense Adam with three $\beta_2$ values on two datasets (DTD and UCF101) and four learning rates ($10^{-2}$, $10^{-3}$, $10^{-4}$, and $10^{-5}$). Across all learning rates and over the training epochs, SGD exceeds the accuracy of Adam while exhibiting lower sensitivity to the learning rate and training epochs. These results suggest that changing the hyperparameter $\beta_2$ cannot address the problem of the second-moment.  

We first explain why Adam fails. The few-shot support set is too small to calibrate a dense adaptive preconditioner for all CLIP parameters. The second moment of Adam's support set is systematically smaller than the corresponding full-data statistic. This error is already present at the pretrained checkpoint and grows as the support loss is optimized. Adaptive-norm diagnostics show the update-scale mismatch. Matching Adam's first-step update budget to SGD removes most of the collapse, whereas numerator-side alignment does not account for the observed behavior. After Adam has moved CLIP too far from its pretrained solution, denominator repair becomes only partially effective.

The failure is not limited to Adam. We also evaluate richer preconditioned optimizers such as Shampoo~\cite{gupta2018shampoo} and SOAP~\cite{vyas2025soap}. Like Adam, Shampoo~\citep{gupta2018shampoo} and SOAP~\citep{vyas2025soap} either collapse or provide a test-set accuracy that remains far below that of SGD (see \textcolor{blue}{\hyperref[app:stability_analysis]{Appendix~\ref*{app:stability_analysis}}}). Our analysis reveals that Shampoo preserves some salient second-order structure but develops large-scale distortion and poor full-data alignment. SOAP avoids Adam's diagonal scale-inflation signature in several cases, but its repaired preconditioned direction is still almost orthogonal to the full-data reference. Both directions use the same full-data preconditioner, so this cosine measures residual numerator mismatch under that map. The observed failures show that richer preconditioning alone does not resolve the problem.

Unlike adaptive and structured preconditioned methods, SGD avoids unreliable few-shot preconditioners and makes small structured updates near the pretrained model. Interpolation probes show broad low-loss spans along the tested initialization-to-final and seed-to-seed segments. Directional curvature probes show lower curvature along the SGD direction than along the full-data gradient direction. Comparisons with the dominant full-loss Hessian direction depend on the dataset and split. Blockwise probes further show that the useful update budget is structured by encoder, depth, and training time. Early in training, many blocks are safe, and SGD is conservative. Later, some directions may become support-specific.

\textbf{Contributions.}
\noindent \textbf{(1)} We separate the effect of parameterization from the effect of optimization in few-shot CLIP adaptation. Extensive experiments on 11 datasets, 3 CLIP backbones, and 5 shot levels show that dense full fine-tuning with SGD outperforms Adam/AdamW and representative PEFT methods under the reported evaluation protocol. This challenges the common view that full fine-tuning is intrinsically unsuitable for few-shot CLIP adaptation.
\noindent \textbf{(2)} We identify unreliable support-set preconditioning as the main cause of Adam collapse. The few-shot second moment is already miscalibrated at initialization. Its norm ratio declines more than the gradient norm ratio during training. Adaptive-norm diagnostics show oversized updates. Denominator counterfactuals, update-budget matching, structured-preconditioner diagnostics, and theoretical analysis all support this claim.
\noindent \textbf{(3)} A detailed analysis is provided to explain why plain SGD remains effective for few-shot CLIP adaptation. It avoids the unstable preconditioner and exhibits low-loss connections along the tested interpolation segments. Our blockwise and curvature analyses show that SGD's success is structured rather than accidental. They point to potential practical improvements based on update-budget control, early stopping, and block-aware schedules.

\section{Related Work}
\label{sec:related_work}

\textbf{Few-Shot CLIP Adaptation and PEFT.}
Few-shot CLIP adaptation improves the zero-shot model using a small labeled support set. Prompt-learning methods replace hand-crafted templates with learned context tokens~\citep{zhou2022coop} or conditional prompts~\citep{zhou2022conditional}. Regularized and multimodal prompt variants further constrain or expand the prompt space~\citep{yao2023kgcoop,zhu2022prograd,khattak2023maple,khattak2023promptsrc}. Adapter and cache methods keep the CLIP backbone frozen while learning lightweight feature transformations or support-set caches~\citep{gao2024clipadapter,cheng2024metaadapter,yang2024mma,zhang2022tipadapter}. Other approaches use low-rank modules~\citep{hu2022lora,zanella2024low}, sparse parameter subsets~\citep{nguyen2024saft,mrabah2025so}, or classifier-side objectives~\citep{huang2024lpplusplus}. Most of this literature treats parameterization as the central design choice. As a result, full fine-tuning is often considered a baseline that is assumed to overfit in low-shot regimes~\citep{zhou2022coop,gao2024clipadapter,silva2024closer,farina2025rethinking}. Our work separates this conclusion from the optimizer used to obtain it. We show that dense full fine-tuning with SGD can outperform representative PEFT methods under the same few-shot protocol. This indicates that the failure of full fine-tuning is not intrinsic to updating all CLIP parameters.

\textbf{Adaptive and Second-Order Preconditioning.}
Adaptive optimizers rescale gradients using statistics accumulated from the training data. Adam~\citep{kingma2015adam} and AdamW~\citep{loshchilov2019decoupled} combine first-moment averaging with a diagonal second-moment denominator. Later work studies convergence problems of Adam-style updates~\citep{reddi2018convergence} or reduces optimizer-state memory~\citep{shazeer2018adafactor,luo2023came,zhang2024adammini,modoranu2024microadam,luo2024badam}. These methods still rely on gradient statistics from the available samples. In few-shot CLIP adaptation, those samples are only the support set. Our analysis shows that this support-set geometry is not representative of the full-data objective. The few-shot second moment is already too small at initialization. Its norm ratio then declines more than the gradient norm ratio during support-set fitting. Adaptive-norm diagnostics show oversized updates. Matching Adam's first-step update budget to SGD resolves most of the collapse, while numerator-side alignment does not explain the failure. Structured preconditioners use richer geometry than diagonal adaptive methods. K-FAC~\citep{martens2015optimizing}, Shampoo~\citep{gupta2018shampoo}, and SOAP~\citep{vyas2025soap} are natural candidates for improving Adam. We find that Shampoo and SOAP also fail in this regime. Richer preconditioning alone does not resolve the failure in these experiments.

\textbf{Loss-Landscape Geometry and SGD Implicit Bias.}
Loss-landscape analyses study how optimization trajectories relate to generalization. Linear interpolation and low-dimensional visualization methods reveal barriers, sharp regions, and flat regions around trained networks~\citep{goodfellow2015qualitatively,li2018visualizing}. Large-batch studies connect sharp minima with poorer generalization~\citep{keskar2017largebatch}. Mode-connectivity work shows that independently trained models can be joined by low-loss paths~\citep{garipov2018loss,draxler2018essentially}. Weight averaging~\citep{izmailov2018averaging} and robust fine-tuning~\citep{wortsman2022robust} further link broad basins to transfer and generalization. SGD also has implicit regularization effects that depend on stochasticity, step size, and training dynamics~\citep{hardt2016train,mandt2017sgd}. We use these ideas to explain why dense SGD succeeds when adaptive preconditioning fails. Our probes show that SGD stays near pretrained CLIP. Interpolation reveals broad low-loss spans along the tested initialization-to-final and seed-to-seed segments. Directional curvature analysis shows that the SGD step often avoids directions of large positive full-data curvature. Blockwise probes further show that safe update budgets depend on encoder, depth, and training time. Thus, SGD is effective not because it freezes the model, but because it avoids unreliable support-set preconditioners while following a low-drift path. The support and full-data gradients can become poorly aligned later in training.

\section{Why Does Adam Fail?}
\label{sec:why_does_adam_fail_short}

\hyperref[sec:why_does_adam_fail]{Appendix~\ref*{sec:why_does_adam_fail}} provides the full diagnostic study. We train CLIP(ViT-B/16) for 30 epochs with $K=1$ shot per class on three datasets (EuroSAT, Oxford-Pets, and DTD). Both encoders are trainable. All Transformer blocks are trainable. The support set is processed as one batch, so one optimizer step equals one pass over the support set.

Let $\Theta_t\in\mathbb{R}^{d}$ be the flattened trainable CLIP parameter vector at step $t$. Here $d$ is the number of trainable scalar parameters. Let $\mathcal{E}=\{\mathrm{vision},\mathrm{text}\}$ be the set of trainable encoders. For $e\in\mathcal{E}$, let $\Theta_t^{(e)}\in\mathbb{R}^{d_e}$ be the parameters restricted to encoder $e$, where $d_e$ is the number of trainable scalar parameters in that encoder. Let $\mathcal{D}_{\mathrm{few}}$ be the support set with $n=KC$ examples, where $C$ is the number of classes. Let $\mathcal{D}_{\mathrm{full}}$ be the full training split with $N$ examples. The empirical losses are $L_{\mathrm{few}},L_{\mathrm{full}},L_{\mathrm{test}}:\mathbb{R}^{d}\to\mathbb{R}$.

For diagnostic mini-batch $\mathcal{B}\subset\mathcal{D}$, the encoder gradient is
\begin{equation}
    g^{(e)}_{\mathcal{B},t}
    = \nabla_{\Theta^{(e)}} L_{\mathcal{B}}(\Theta_t)
    \in \mathbb{R}^{d_e}.
\end{equation}
For $\mathcal{D}\in\{\mathcal{D}_{\mathrm{few}},\mathcal{D}_{\mathrm{full}}\}$, the encoder-restricted mean gradient and diagonal second moment are
\begin{equation}
    m^{(e)}_{\mathcal{D},t}
    = \nabla_{\Theta^{(e)}}L_{\mathcal{D}}(\Theta_t)
    \in\mathbb{R}^{d_e},
    \qquad
    v^{(e)}_{\mathcal{D},t}
    = \mathbb{E}_{\mathcal{B}\subset\mathcal{D}}
    \left[g^{(e)}_{\mathcal{B},t}\odot g^{(e)}_{\mathcal{B},t}\right]
    \in\mathbb{R}_{\ge 0}^{d_e}.
\end{equation}
The operator $\odot$ denotes elementwise multiplication. We denote the support-set quantities by $m^{(e)}_{\mathrm{few},t}$ and $v^{(e)}_{\mathrm{few},t}$, and the full-train reference quantities by 
$m^{(e)}_{\mathrm{full},t}$ and $v^{(e)}_{\mathrm{full},t}$. 
With numerical stabilizer $\varepsilon>0$, the three Adam-like directions are
\begin{equation}
    u^{(e)}_{ff,t}
    = m^{(e)}_{\mathrm{few},t}\oslash
    \sqrt{v^{(e)}_{\mathrm{few},t}+\varepsilon},
    \quad
    u^{(e)}_{fr,t}
    = m^{(e)}_{\mathrm{few},t}\oslash
    \sqrt{v^{(e)}_{\mathrm{full},t}+\varepsilon},
    \quad
    u^{(e)}_{rr,t}
    = m^{(e)}_{\mathrm{full},t}\oslash
    \sqrt{v^{(e)}_{\mathrm{full},t}+\varepsilon},
    \label{eq:adam_short_directions}
\end{equation}
where $\oslash$ and the square root are elementwise. All three vectors lie in $\mathbb{R}^{d_e}$. The direction $u^{(e)}_{ff,t}$ is the few-shot Adam-like direction. The direction $u^{(e)}_{fr,t}$ replaces only the denominator by the full-train denominator. The direction $u^{(e)}_{rr,t}$ is the full-train reference direction.

\begin{figure}[t]
    \centering
    \includegraphics[width=0.98\linewidth]{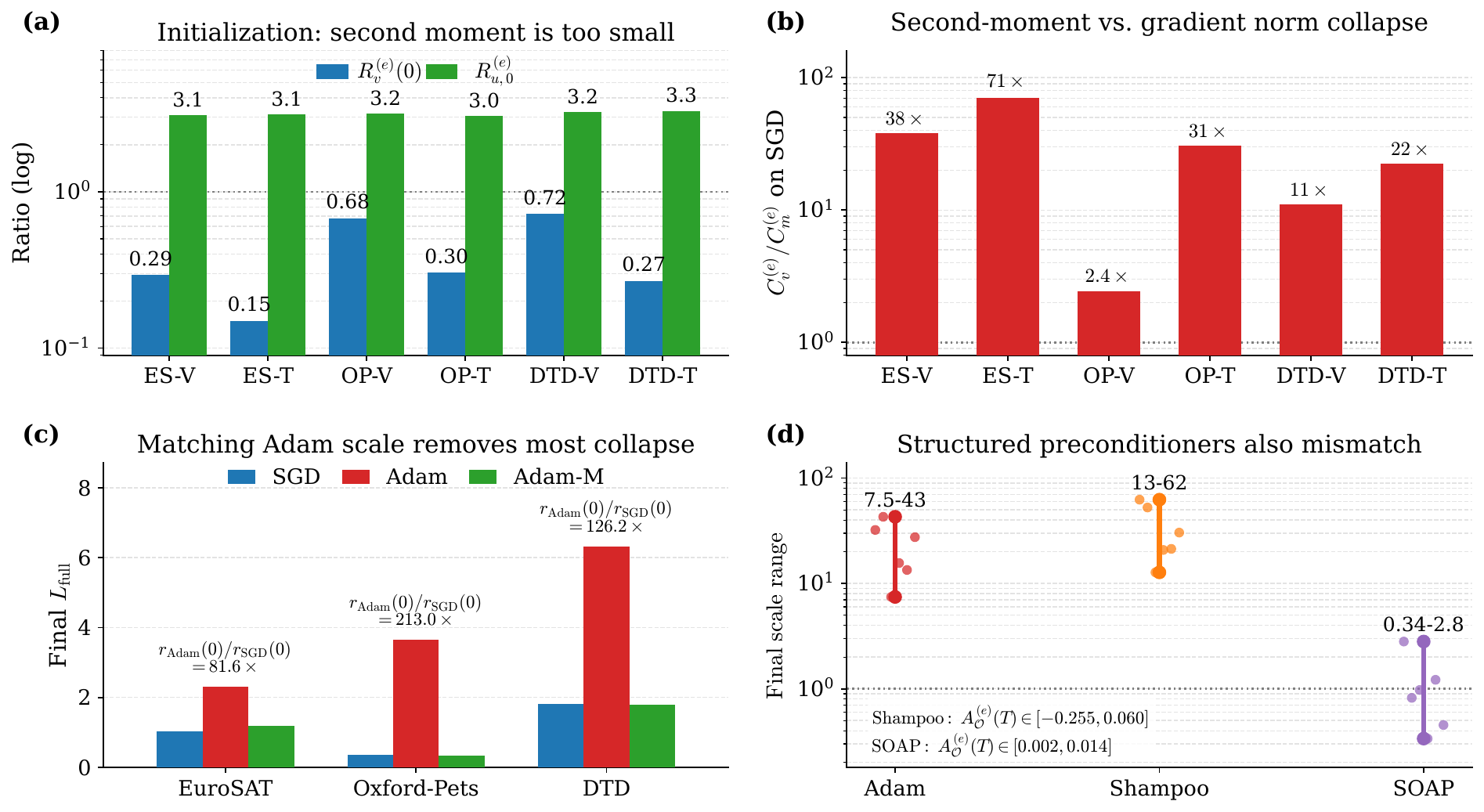}
    \caption{Summary of Adam failure diagnostics. Panel (a) shows the initialization second-moment norm ratio $R^{(e)}_v(0)$ and the induced adaptive norm ratio $R^{(e)}_{u,0}$. Panel (b) compares the relative decline of the second-moment and gradient norm ratios on the SGD trajectory. Panel (c) shows that matching Adam's first-step update budget to SGD removes most of the full-train loss collapse. Panel (d) separates structured-preconditioner scale differences from residual numerator misalignment under the full-train map. ES, OP, and DTD denote EuroSAT, Oxford-Pets, and DTD. V and T denote the vision and text encoders.}
    \label{fig:adam_failure_core_short}
\end{figure}

\textbf{Few-shot second moments are miscalibrated.}
Adam does not start from a calibrated denominator. At the pretrained checkpoint $\Theta_0$, the scale ratios are
\begin{equation}
    R^{(e)}_v(0)
    = \frac{\|v^{(e)}_{\mathrm{few},0}\|_2}
           {\|v^{(e)}_{\mathrm{full},0}\|_2},
    \qquad
    R^{(e)}_{u,0}
    = \frac{\|u^{(e)}_{ff,0}\|_2}
           {\|u^{(e)}_{rr,0}\|_2}.
\end{equation}
Both ratios are dimensionless. Figure~\ref{fig:adam_failure_core_short}(a) shows that $R^{(e)}_v(0)<1$ in all six dataset--encoder pairs. Thus, the support-set second moment is already smaller than the full-train second moment before any update. The same panel shows that $R^{(e)}_{u,0}$ is about $3\times$ in all cases. Therefore, Adam begins with an adaptive direction whose norm is already too large. This effect is strongest in the text encoder. In \hyperref[sec:why_does_adam_fail]{Appendix~\ref*{sec:why_does_adam_fail}}, the text gradients remain well aligned with the full-train gradients at initialization, while the denominator is too small.

\paragraph{Second-moment and gradient norm collapse.}
The initialization error also grows as the support loss is fitted. The following ratios compare gradient and second-moment norms:
\begin{equation}
    R^{(e)}_m(t)
    = \frac{\|m^{(e)}_{\mathrm{few},t}\|_2}
           {\|m^{(e)}_{\mathrm{full},t}\|_2},
    \qquad
    R^{(e)}_v(t)
    = \frac{\|v^{(e)}_{\mathrm{few},t}\|_2}
           {\|v^{(e)}_{\mathrm{full},t}\|_2}.
\end{equation}
These ratios are dimensionless. Let $T$ be the final logged checkpoint. The collapse factors are
\begin{equation}
    C^{(e)}_m=\frac{R^{(e)}_m(0)}{R^{(e)}_m(T)},
    \qquad
    C^{(e)}_v=\frac{R^{(e)}_v(0)}{R^{(e)}_v(T)}.
\end{equation}
Figure~\ref{fig:adam_failure_core_short}(b) reports $C^{(e)}_v/C^{(e)}_m$ on the stable SGD trajectory. The ratio is larger than one in every dataset--encoder pair. It reaches $38\times$ and $71\times$ on EuroSAT vision and text. It also reaches $30.6\times$ on Oxford-Pets text and $22\times$ on DTD text. The second-moment norm ratio decreases by a larger relative factor than the gradient norm ratio. These factors do not determine adaptive update norms.

\textbf{Adam collapse is scale-dominated.}
Adaptive-direction norms and update-budget interventions assess the effect on update scale. For an optimizer $o\in\{\mathrm{SGD},\mathrm{Adam},\text{Adam-M}\}$, let $\Theta_t^{(o)}\in\mathbb{R}^{d}$ be its parameter vector and let $\Delta\Theta_t^{(o)}=\Theta_{t+1}^{(o)}-\Theta_t^{(o)}\in\mathbb{R}^{d}$ be its update. The global relative update budget is
\begin{equation}
    r_o(t)=\frac{\|\Delta\Theta_t^{(o)}\|_2}{\|\Theta_t^{(o)}\|_2}.
\end{equation}
This scalar is dimensionless. Adam-M is Adam with a one-time learning-rate multiplier chosen so that $r_{\text{Adam-M}}(0)=r_{\mathrm{SGD}}(0)$. Figure~\ref{fig:adam_failure_core_short}(c) shows that raw Adam starts with a much larger budget than SGD. The ratios $r_{\mathrm{Adam}}(0)/r_{\mathrm{SGD}}(0)$ are $81.6\times$, $213.0\times$, and $126.2\times$ on EuroSAT, Oxford-Pets, and DTD. Matching this first-step budget removes most of the full-train loss collapse. \hyperref[sec:why_does_adam_fail]{Appendix~\ref*{sec:why_does_adam_fail}} shows that numerator-side alignment can look reasonable while raw Adam still fails. The main causal lever is the scale induced by the few-shot denominator.

\textbf{Accuracy recovery across shots and model families.}
\hyperref[app:additional_results]{Appendix~\ref*{app:additional_results}} extends the update-budget intervention to six CLIP datasets and all five shot levels. First-step budget matching raises Adam's mean final accuracy from $14.51\%$ to $76.03\%$, compared with $74.93\%$ for SGD. On SigLIP, the corresponding accuracies are $38.88\%$, $76.60\%$, and $76.28\%$ across four datasets and three shot levels. Reduced learning rates and warmup also restore Adam's 30-epoch accuracy on the common six-dataset subset. These controls support the role of early update scale. The supplementary results distinguish fixed-horizon accuracy from long-horizon trajectories.

\textbf{Richer preconditioners retain scale and numerator mismatch.}
The failure is not only a diagonal Adam artifact. Let $\mathcal{O}\in\{\mathrm{Shampoo},\mathrm{SOAP}\}$ denote a structured preconditioner. For encoder $e$, split $\mathcal{D}$, and step $t$, let
\begin{equation}
    B^{(e)}_{\mathcal{O},\mathcal{D},t}:\mathbb{R}^{d_e}\to\mathbb{R}^{d_e}
\end{equation}
be the effective structured preconditioning map. In the SOAP diagnostic, both splits use a shared eigenbasis estimated from the full training split at each checkpoint. Only the second-moment statistics differ between the two maps. The corresponding directions are
\begin{equation}
    q^{(\mathcal{O},e)}_{ff,t}=B^{(e)}_{\mathcal{O},\mathrm{few},t}m^{(e)}_{\mathrm{few},t},
    \quad
    q^{(\mathcal{O},e)}_{fr,t}=B^{(e)}_{\mathcal{O},\mathrm{full},t}m^{(e)}_{\mathrm{few},t},
    \quad
    q^{(\mathcal{O},e)}_{rr,t}=B^{(e)}_{\mathcal{O},\mathrm{full},t}m^{(e)}_{\mathrm{full},t}.
\end{equation}
All three vectors lie in $\mathbb{R}^{d_e}$. We define the scale ratio $S^{(e)}_{\mathcal{O}}(t)$ and the repaired reference alignment $A^{(e)}_{\mathcal{O}}(t)$ as
\begin{equation}
    S^{(e)}_{\mathcal{O}}(t)=\frac{\|q^{(\mathcal{O},e)}_{ff,t}\|_2}{\|q^{(\mathcal{O},e)}_{fr,t}\|_2},
    \qquad
    A^{(e)}_{\mathcal{O}}(t)=\cos\!\left(q^{(\mathcal{O},e)}_{fr,t},q^{(\mathcal{O},e)}_{rr,t}\right).
\end{equation}
Figure~\ref{fig:adam_failure_core_short}(d) summarizes the final structured-preconditioner diagnostics. Shampoo develops large-scale distortion and poor full-train alignment. SOAP avoids the largest Adam-like scale inflation in several cases, but its repaired direction is almost orthogonal to the full-train reference. This cosine measures residual numerator mismatch because both directions use the same full-train preconditioner. The scale ratio separately measures the effect of changing the preconditioning statistic.

\begin{tcolorbox}[
  enhanced,
  unbreakable,
  colback=findingblueback,
  colframe=findingblueframe,
  colbacktitle=findingblueframe,
  coltitle=white,
  boxrule=0.7pt,
  arc=1.5pt,
  left=6pt,
  right=6pt,
  top=5pt,
  bottom=5pt,
  before skip=7pt,
  after skip=7pt,
  fonttitle=\bfseries\small,
  fontupper=\small,
  title={Main Conclusion}]
Adam fails because the few-shot support set cannot calibrate a dense adaptive preconditioner. The support-set second moment is too small at $\Theta_0$. During training, its norm ratio declines more than the gradient norm ratio. Adaptive-norm and update-budget diagnostics show oversized updates. Matching the update budget removes most of the collapse. Shampoo and SOAP also exhibit preconditioning-statistic mismatch and residual numerator mismatch. Dense full fine-tuning itself remains effective with SGD.
\end{tcolorbox}

\section{Why Does SGD Perform Well?}
\label{sec:why_does_sgd_succeed_short}

This section summarizes the SGD landscape analysis from \hyperref[sec:why_does_sgd_succeed]{Appendix~\ref*{sec:why_does_sgd_succeed}}. %
We use the same protocol as in Section~\ref{sec:why_does_adam_fail_short}. Let
\begin{equation}
    \mathcal{M}=\{\mathrm{few},\mathrm{full},\mathrm{test}\}.
\end{equation}
For each $S\in\{\mathrm{few},\mathrm{full},\mathrm{test}\}$, let $L_S:\mathbb{R}^d\to\mathbb{R}$ be the empirical cross-entropy loss on the support set, the full training split, or the test split. The SGD gradient is
\begin{equation}
    g_t=\nabla_{\Theta}L_{\mathrm{few}}(\Theta_t)\in\mathbb{R}^d,
    \qquad
    \Theta_{t+1}=\Theta_t-\eta g_t,
\end{equation}
where $\eta>0$ is the scalar learning rate. The key finding is that SGD is effective because its update geometry remains locally benign for $L_{\mathrm{full}}$ and $L_{\mathrm{test}}$. This differs from Adam, Shampoo, and SOAP, which rely on support-set preconditioners.

\begin{figure}[t]
    \centering
    \begin{subfigure}[t]{0.98\linewidth}
        \centering
        \includegraphics[width=0.93\linewidth]{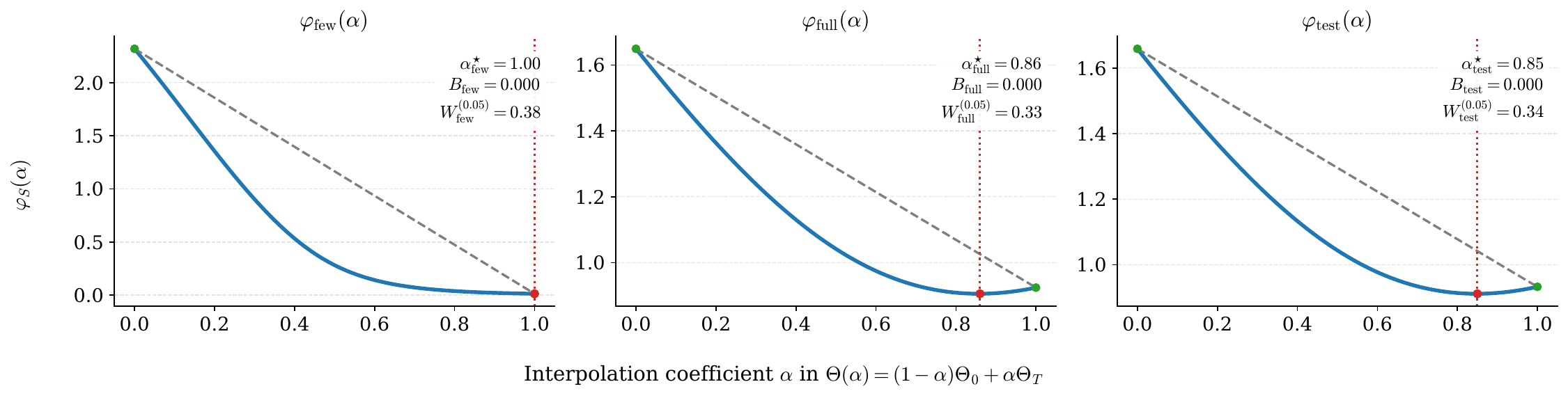}
        \caption{Init-to-final interpolation on EuroSAT.}
        \label{fig:sgd_short_basin:a}
    \end{subfigure}

    \vspace{0.25em}

    \begin{subfigure}[t]{0.98\linewidth}
        \centering
        \includegraphics[width=0.93\linewidth]{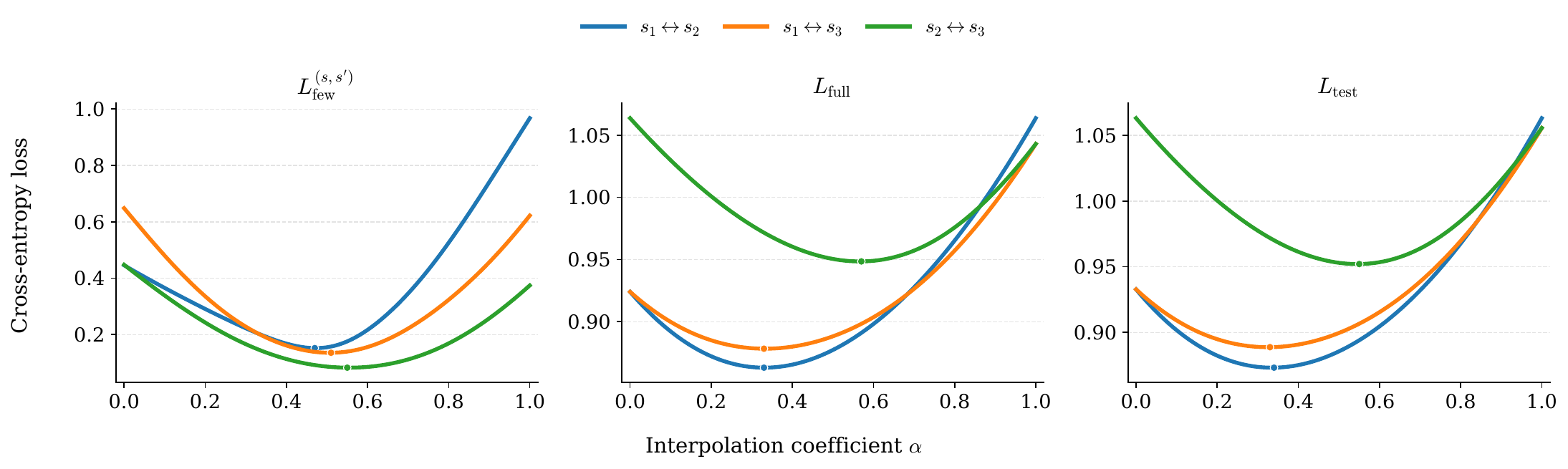}
        \caption{Seed-to-seed interpolation on EuroSAT.}
        \label{fig:sgd_short_basin:b}
    \end{subfigure}
    \caption{\textbf{Broadness and connectivity of SGD solutions.}
    Panel (a) shows a broad near-minimum span along the line from $\Theta_0$ to $\Theta_T$. Panel (b) shows zero measured interpolation barriers between final SGD solutions from the tested seeds. \hyperref[sec:why_does_sgd_succeed]{Appendix~\ref*{sec:why_does_sgd_succeed}} gives the corresponding Oxford-Pets and DTD plots.}
    \label{fig:sgd_short_basin}
\end{figure}

\textbf{SGD reaches broad low-loss corridors.}
Let $T$ be the final epoch. We first interpolate from the pretrained checkpoint $\Theta_0\in\mathbb{R}^d$ to the final SGD solution $\Theta_T\in\mathbb{R}^d$:
\begin{equation}
    \Theta(\alpha)=(1-\alpha)\Theta_0+\alpha\Theta_T,
    \qquad \alpha\in[0,1].
\end{equation}
For $S\in\mathcal{M}$, we define the scalar line loss
\begin{equation}
    \varphi_S(\alpha)=L_S(\Theta(\alpha)),
    \qquad \varphi_S:[0,1]\to\mathbb{R}.
\end{equation}
Let $\alpha_S^\star\in[0,1]$ be a minimizer of $\varphi_S$, and let $\varphi_S^\star=\varphi_S(\alpha_S^\star)$. Let $\varphi_S^{\mathrm{lin}}$ be the endpoint chord between $\varphi_S(0)$ and $\varphi_S(1)$. We measure the interpolation barrier
\begin{equation}
    B_S=\max_{\alpha\in[0,1]}
    \left(\varphi_S(\alpha)-\varphi_S^{\mathrm{lin}}(\alpha)\right)\in\mathbb{R}.
\end{equation}
We also measure the normalized near-minimum width
\begin{equation}
    W_S^{(\delta_{\mathrm{w}})}
    =\max \mathcal{S}_{S}^{(\delta_{\mathrm{w}})}-\min \mathcal{S}_{S}^{(\delta_{\mathrm{w}})}\in[0,1],
\end{equation}
where $\delta_{\mathrm{w}}=0.05$ and
$\mathcal{S}_{S}^{(\delta_{\mathrm{w}})}= \left\{\alpha\in[0,1]:\varphi_S(\alpha)\leq\varphi_S^\star+\delta_{\mathrm{w}}\left(\varphi_S(0)-\varphi_S^\star\right)\right\}.$
Figure~\ref{fig:sgd_short_basin}(a) illustrates the init-to-final loss profiles on EuroSAT. On EuroSAT and Oxford-Pets, the full-train and test barriers are zero. Their minima lie close to the final SGD point, with $\alpha_{\mathrm{full}}^\star$ and $\alpha_{\mathrm{test}}^\star$ in the range $0.85$--$0.87$. On DTD, the full-train and test optima occur earlier, at $\alpha_{\mathrm{full}}^\star=0.76$ and $\alpha_{\mathrm{test}}^\star=0.70$. The corresponding near-minimum spans remain broad along the sampled direction. DTD shows a late support-specialized tail on this line.

\begin{figure}[t]
    \centering
    \begin{subfigure}[t]{0.98\linewidth}
        \centering
        \includegraphics[width=0.93\linewidth]{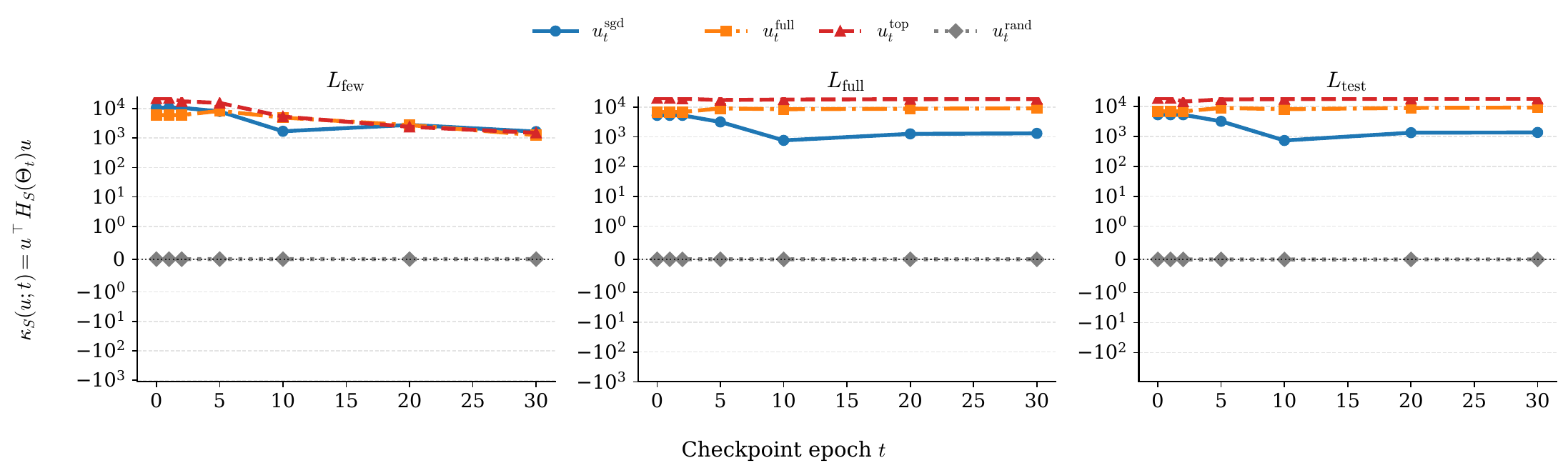}
        \caption{Directional curvature on EuroSAT.}
        \label{fig:sgd_short_curvature_blocks:a}
    \end{subfigure}

    \vspace{0.25em}

    \begin{subfigure}[t]{0.98\linewidth}
        \centering
        \includegraphics[width=0.93\linewidth]{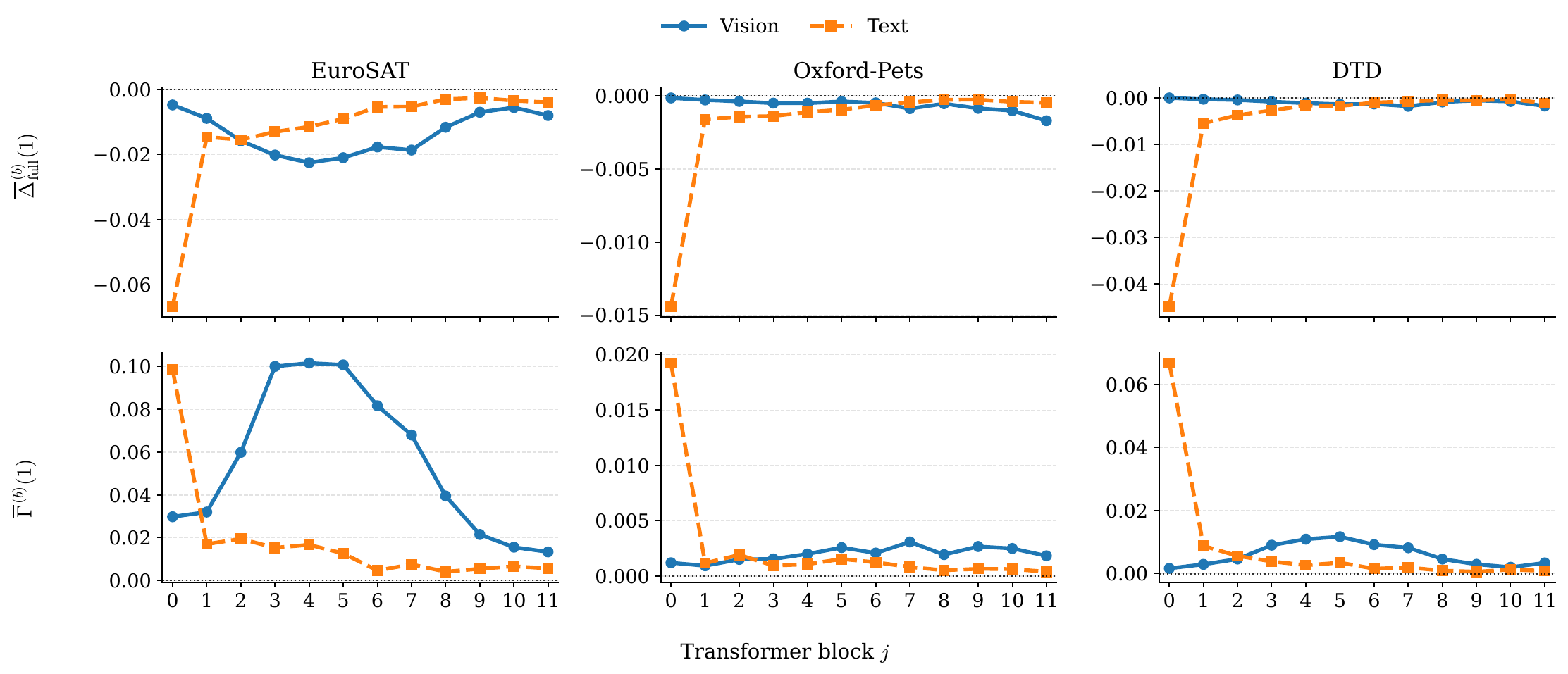}
        \caption{Checkpoint-averaged block safety on EuroSAT, Oxford-Pets, and DTD.}
        \label{fig:sgd_short_curvature_blocks:b}
    \end{subfigure}
    \caption{\textbf{Directional curvature and blockwise update budgets.}
    Panel (a) compares directional curvatures along $u_t^{\mathrm{sgd}}$, $u_t^{\mathrm{full}}$, $u_t^{\mathrm{top}}$, and $u_t^{\mathrm{rand}}$. Panel (b) summarizes the full-loss change $\overline{\Delta}_{\mathrm{full}}^{(b)}(1)$ and the support--full mismatch $\overline{\Gamma}^{(b)}(1)$ by block. Full dataset-level trajectories are reported in \hyperref[sec:why_does_sgd_succeed]{Appendix~\ref*{sec:why_does_sgd_succeed}}.}
    \label{fig:sgd_short_curvature_blocks}
\end{figure}

\textbf{SGD follows directions of lower positive curvature.}
A broad interpolation path does not imply low curvature in all local directions. We therefore measure directional curvature. For $S\in\mathcal{M}$, let
\begin{equation}
    H_S(\Theta_t)=\nabla^2 L_S(\Theta_t)\in\mathbb{R}^{d\times d}
\end{equation}
be the Hessian at checkpoint $\Theta_t$. For a unit vector $u\in\mathbb{R}^d$ with $\lVert u\rVert_2=1$, we define
\begin{equation}
    \kappa_S(u;t)=u^\top H_S(\Theta_t)u\in\mathbb{R}.
\end{equation}
Large positive values indicate scale-sensitive directions. Large negative values indicate negative-curvature directions. They do not alone establish a saddle point. We compare the actual SGD direction
\begin{equation}
    u_t^{\mathrm{sgd} }
    =\frac{-\eta g_t}{\lVert -\eta g_t\rVert_2}\in\mathbb{R}^d
\end{equation}
with three controls. These are the normalized full-train gradient $u_t^{\mathrm{full}}\in\mathbb{R}^d$, the largest-magnitude eigendirection $u_t^{\mathrm{top}}\in\mathbb{R}^d$ of $H_{\mathrm{full}}(\Theta_t)$, and a random unit direction $u_t^{\mathrm{rand}}\in\mathbb{R}^d$. Figure~\ref{fig:sgd_short_curvature_blocks}(a) compares these directions on EuroSAT. On EuroSAT, $u_t^{\mathrm{sgd}}$ has smaller positive curvature than the full-gradient and dominant-Hessian controls on the full-train and test losses. At $t=30$ on EuroSAT, the full-loss curvatures are $1317$, $9158$, and $18634$ along $u_t^{\mathrm{sgd}}$, $u_t^{\mathrm{full}}$, and $u_t^{\mathrm{top}}$. The test-loss values have the same order. On DTD, $u_t^{\mathrm{top}}$ has large negative curvature on $L_{\mathrm{full}}$, but it is nearly irrelevant on $L_{\mathrm{test}}$. The full-gradient direction has larger positive test curvature than the SGD direction. Finite-step safety also depends on gradient alignment and step size.

\textbf{Safe update budget is block- and time-dependent.}
The safe update geometry is not uniform across CLIP. Let $b=(e,j)$ denote a trainable encoder-Transformer block, where $e\in\{\mathrm{vision},\mathrm{text}\}$ and $j\in\{0,\ldots,11\}$. Let $P_b\in\mathbb{R}^{d\times d}$ be the diagonal projector that keeps the coordinates of block $b$ and zeros all other coordinates. The block-restricted SGD step is
\begin{equation}
    a_t^{(b)}=-\eta P_b g_t\in\mathbb{R}^d.
\end{equation}
For $S\in\mathcal{M}$ and a dimensionless multiplier $\alpha\in[-2,2]$, we define the blockwise line loss
\begin{equation}
    \psi_S^{(b)}(\alpha;t)
    =L_S(\Theta_t+\alpha a_t^{(b)})\in\mathbb{R}
\end{equation}
and the loss change
\begin{equation}
    \Delta_S^{(b)}(t;\alpha)
    =\psi_S^{(b)}(\alpha;t)-\psi_S^{(b)}(0;t)\in\mathbb{R}.
\end{equation}
The actual block step is locally useful for the full objective when $\Delta_{\mathrm{full}}^{(b)}(t;1)<0$. The support--full mismatch is
\begin{equation}
    \Gamma^{(b)}(t;\alpha)
    =\Delta_{\mathrm{full}}^{(b)}(t;\alpha)
    -\Delta_{\mathrm{few}}^{(b)}(t;\alpha).
    \label{eq:sgd_short_block_mismatch}
\end{equation}
Over the logged checkpoints $\mathcal{T}=\{0,1,2,5,10,20,30\}$, Figure~\ref{fig:sgd_short_curvature_blocks}(b) reports
\begin{equation}
    \overline{\Delta}_{\mathrm{full}}^{(b)}(1)
    =\frac{1}{|\mathcal{T}|}\sum_{t\in\mathcal{T}}\Delta_{\mathrm{full}}^{(b)}(t;1),
    \qquad
    \overline{\Gamma}^{(b)}(1)
    =\frac{1}{|\mathcal{T}|}\sum_{t\in\mathcal{T}}\Gamma^{(b)}(t;1).
    \label{eq:sgd_short_block_averages}
\end{equation}
The top row measures the mean full-loss change. Negative values indicate beneficial block updates. The bottom row measures the mean mismatch. Positive values indicate that the support loss overstates the benefit to the full-data objective. The vision and text curves show how these quantities vary with Transformer depth. Let
\begin{equation}
    \alpha_{\mathrm{full}}^{\star,(b)}(t)
    \in
    \arg\min_{\alpha\in[-2,2]}
    \psi_{\mathrm{full}}^{(b)}(\alpha;t)
\end{equation}
be the locally preferred full-loss scale. Values larger than one indicate a conservative block step. Values between zero and one favor damping; negative values favor the reverse direction. Early in training, many blocks have negative $\Delta_{\mathrm{full}}^{(b)}(t;1)$ and preferred scales near the upper end of the probe interval. Later, the gains shrink and some blocks become harmful. EuroSAT keeps persistent safe motion in middle vision blocks, especially blocks $3$--$5$. Oxford-Pets becomes nearly flat after the early phase. DTD has the strongest phase change. Early text directions are useful, but several become fragile late in training.

\textbf{SGD solutions remain connected across seeds.}
The previous diagnostics study one SGD trajectory. We also probe straight segments between final solutions from different few-shot seeds. Let $s$ and $s'$ be two independent support-set seeds. Let $\Theta_T^{(s)},\Theta_T^{(s')}\in\mathbb{R}^d$ be the corresponding final SGD states. We define
\begin{equation}
    \Theta^{(s,s')}(\alpha)
    =(1-\alpha)\Theta_T^{(s)}+
    \alpha\Theta_T^{(s')},
    \qquad \alpha\in[0,1].
\end{equation}
For $S\in\{\mathrm{few},\mathrm{full},\mathrm{test}\}$, we define the seed-pair line loss
\begin{equation}
    \varphi_S^{(s,s')}(\alpha)
    =L_S^{(s,s')}(\Theta^{(s,s')}(\alpha))\in\mathbb{R}.
\end{equation}
$L_{\mathrm{few}}^{(s,s')}$ is the average of the two support losses, while $L_{\mathrm{full}}^{(s,s')}=L_{\mathrm{full}}$ and $L_{\mathrm{test}}^{(s,s')}=L_{\mathrm{test}}$. We compute the same barrier and width diagnostics as above. We also record the relative endpoint separation
\begin{equation}
    D_{\mathrm{rel}}^{(s,s')}
    =
    \frac{\lVert\Theta_T^{(s)}-\Theta_T^{(s')}\rVert_2}
    {\frac{1}{2}
    \left(
    \lVert\Theta_T^{(s)}-\Theta_0\rVert_2+
    \lVert\Theta_T^{(s')}-\Theta_0\rVert_2
    \right)}
    \in\mathbb{R}_{\geq0}.
\end{equation}
Figure~\ref{fig:sgd_short_basin}(b) shows the seed-to-seed loss profiles on EuroSAT. Across all tested seed pairs and all three datasets, the measured full-train and test barriers are zero on the interpolation grid. The minima of $L_{\mathrm{full}}$ and $L_{\mathrm{test}}$ lie inside the interpolation segment. The endpoints are not nearly identical, since $D_{\mathrm{rel}}^{(s,s')}$ is between $1.05$ and $1.20$. The measured straight segments are consistent with connected low-loss regions across the tested seeds. Broadness is assessed only along the probed interpolation directions.

\begin{findingbox}[Conclusion]
SGD avoids unreliable support-set preconditioners. The init-to-final probes show broad near-minimum spans along the sampled direction. Its update direction has lower positive curvature than the full-gradient direction in the reported comparisons. The dominant full-Hessian direction need not have high test curvature. Its useful budget is block- and time-dependent. The seed-to-seed probes are consistent with low-loss connectivity across the tested seeds. The main limitation is late support specialization on harder datasets such as DTD.
\end{findingbox}

\section{Conclusion}
\label{conclusion}

This work shows that full fine-tuning is not intrinsically unsuitable for few-shot CLIP adaptation. The failure of dense adaptation comes mainly from optimization. Adam builds its preconditioner from a small support set, which miscalibrates the second moment and produces unsafe update scales. The failures of Shampoo and SOAP show that this issue extends beyond Adam’s diagonal form. In contrast, SGD avoids this unreliable preconditioning. Interpolation probes show broad low-loss spans along tested directions, although some blocks become support-specific late in training. These findings suggest that few-shot CLIP adaptation should be studied through the representativeness of optimizer statistics as well as the choice of trainable parameters. Future work should focus on budget-controlled adaptive methods and protected or regularized preconditioners.

\noindent\textbf{Acknowledgments.}
The experiments used computing resources provided by the Digital Research Alliance of Canada. N. Mrabah acknowledges a B3 postdoctoral fellowship from the Fonds de recherche du Qu\'ebec (FRQ), Nature et technologies (award~2011136; \href{https://doi.org/10.69777/2011136}{doi:10.69777/2011136}).

\clearpage
\appendix
\renewcommand{\contentsname}{Supplementary contents}
\tableofcontents
\clearpage
\addtocontents{toc}{\protect\setcounter{tocdepth}{1}}

\section{Notation}

\label{app:notations}

Table~\ref{tab:notation_common_revised} collects the common definitions. Complete symbols, including their subscripts and superscripts, identify the corresponding quantities. Table~\ref{tab:notation_local_reuse_revised} distinguishes related quantities used in different analyses. The remaining tables collect the notation for each section.

\begingroup

\small

\setlength{\tabcolsep}{3pt}

\renewcommand{\arraystretch}{1.12}

\begin{longtable}{@{}>{\raggedright\arraybackslash}p{0.23\textwidth}>{\raggedright\arraybackslash}p{0.22\textwidth}>{\raggedright\arraybackslash}p{0.49\textwidth}@{}}
\caption{Primary notation with a consistent meaning throughout the paper.}
\label{tab:notation_common_revised}\\
\toprule
Term & Dimension & Definition \\
\midrule
\endfirsthead
\toprule
Term & Dimension & Definition \\
\midrule
\endhead
\bottomrule
\endfoot
$\mathcal{E}$ & finite set & Trainable CLIP encoder set, $\mathcal{E}=\{\mathrm{vision},\mathrm{text}\}$. \\
$e$ & element of $\mathcal{E}$ & Encoder index. The values $\mathrm{vision}$ and $\mathrm{text}$ denote the image and text encoders. \\
$t$ & integer & Optimizer step, epoch, or logged checkpoint. In the full-support diagnostics, one step is one pass over the support set. \\
$\Theta_t$ & $\mathbb{R}^{d}$ & Flattened vector of all trainable CLIP parameters at step $t$. \\
$\Theta_0$ & $\mathbb{R}^{d}$ & Pretrained CLIP checkpoint before few-shot adaptation. \\
$\Theta_T$ & $\mathbb{R}^{d}$ & Final adapted checkpoint when $T$ denotes the terminal training step in a local experiment. \\
$\Theta_t^{(e)}$ & $\mathbb{R}^{d_e}$ & Trainable parameter vector restricted to encoder $e$ at step $t$. \\
$d$ & positive integer & Total number of trainable scalar parameters after flattening all trainable tensors. \\
$d_e$ & positive integer & Number of trainable scalar parameters in encoder $e$. \\
$d_{\mathrm{emb}}$ & positive integer & Dimension of normalized CLIP image and text embeddings. \\
$K$ & positive integer & Number of labeled examples per class in a few-shot split. \\
$C$ & positive integer & Number of task classes. \\
$n$ & positive integer & Few-shot support-set size. In balanced $K$-shot classification, $n=KC$. \\
$N$ & positive integer & Number of examples in the full training split. \\
$x$ & input object & Image input. \\
$y$ & element of $\{1,\ldots,C\}$ & Ground-truth class label. \\
$c$ & element of $\{1,\ldots,C\}$ & Class index when used in the CLIP classifier. \\
$\mathcal{D}_{\mathrm{few}}$ & set of $n$ labeled examples & Few-shot support set used for adaptation. \\
$\mathcal{D}_{\mathrm{full}}$ & set of $N$ labeled examples & Full training split. It is used as a diagnostic reference and is not used for the few-shot update. \\
$\mathcal{D}_{\mathrm{test}}$ & test set & Held-out test split used for evaluation and test-loss probes. \\
$\mathcal{D}_S$ & dataset indexed by split $S$ & Dataset for split $S\in\{\mathrm{few},\mathrm{full},\mathrm{test}\}$ when $S$ is used only as a split label. \\
$L_{\mathrm{few}}$ & $\mathbb{R}^{d}\to\mathbb{R}$ & Empirical cross-entropy loss on the support set. \\
$L_{\mathrm{full}}$ & $\mathbb{R}^{d}\to\mathbb{R}$ & Empirical cross-entropy loss on the full training split. \\
$L_{\mathrm{test}}$ & $\mathbb{R}^{d}\to\mathbb{R}$ & Empirical cross-entropy loss on the test split. \\
$L_{\mathcal{D}}$, $L_S$ & scalar-valued loss & Empirical loss on dataset $\mathcal{D}$ or its named split $S$. \\
$\ell((x,y);\Theta_t)$ & scalar & Per-example cross-entropy loss at parameters $\Theta_t$. \\
$g_t$ & $\mathbb{R}^{d}$ & Dense gradient used by the optimizer at step $t$. In full-support diagnostics, $g_t=\nabla_{\Theta}L_{\mathrm{few}}(\Theta_t)$. \\
$g_{\mathcal{B},t}^{(e)}$ & $\mathbb{R}^{d_e}$ & Encoder-restricted gradient on diagnostic mini-batch $\mathcal{B}$ at checkpoint $t$. \\
$m_{\mathcal{D},t}^{(e)}$ & $\mathbb{R}^{d_e}$ & Encoder-restricted empirical gradient, or first-moment numerator, on dataset $\mathcal{D}$ at checkpoint $t$. \\
$v_{\mathcal{D},t}^{(e)}$ & $\mathbb{R}_{\ge0}^{d_e}$ & Encoder-restricted diagonal second moment estimated on dataset $\mathcal{D}$ at checkpoint $t$. \\
$m_{\mathrm{few},t}^{(e)}$, $m_{\mathrm{full},t}^{(e)}$ & $\mathbb{R}^{d_e}$ & Support and full-train encoder-restricted first moments. \\
$v_{\mathrm{few},t}^{(e)}$, $v_{\mathrm{full},t}^{(e)}$ & $\mathbb{R}_{\ge0}^{d_e}$ & Support and full-train encoder-restricted diagonal second moments. \\
$u_{ff,t}^{(e)}$ & $\mathbb{R}^{d_e}$ & Adam-like direction with few-shot numerator and few-shot denominator. \\
$u_{fr,t}^{(e)}$ & $\mathbb{R}^{d_e}$ & Adam-like direction with few-shot numerator and full-data denominator. \\
$u_{rr,t}^{(e)}$ & $\mathbb{R}^{d_e}$ & Full-data reference Adam-like direction with full-data numerator and denominator. \\
$\eta$ & positive scalar & Learning rate. \\
$\beta_1$ & scalar in $(0,1)$ & First-moment or momentum coefficient. \\
$\beta_2$ & scalar in $(0,1)$ & Second-moment coefficient. \\
$\varepsilon$ & positive scalar & Numerical stabilizer in adaptive denominators and matrix roots. \\
$\odot$ & elementwise operator & Elementwise multiplication. \\
$\oslash$ & elementwise operator & Elementwise division. \\
$\sqrt{\cdot}$ & elementwise map on vectors & Elementwise square root when applied to a vector or tensor of nonnegative entries. \\
$\|\cdot\|_2$ & norm & Euclidean norm for vectors and flattened Frobenius norm for tensors unless stated otherwise. \\
$\|\cdot\|_1$ & norm & Entrywise $\ell_1$ norm. \\
$\langle\cdot,\cdot\rangle$ & scalar & Euclidean inner product. \\
$\cos(a,b)$ & scalar in $[-1,1]$ & Cosine similarity $\langle a,b\rangle/(\|a\|_2\|b\|_2)$ when both vectors are nonzero. \\
$\nabla L(\Theta)$ & gradient & Gradient of loss $L$ with respect to the trainable parameter vector. \\
$\nabla^2 L(\Theta)$ & Hessian & Second derivative matrix of loss $L$ with respect to trainable parameters. \\
$\mathbb{E}[\cdot]$ & expectation & Expectation over the explicitly stated random draw, diagnostic batches, or support sampling model. \\
$\operatorname{Var}(\cdot)$ & variance & Variance under the explicitly stated sampling model. \\
$\operatorname{top}_k(a)$ & set of indices & Indices of the $k$ largest entries of a nonnegative vector $a$. \\
$\log_{10}$ & scalar map & Base-10 logarithm used for coordinate mismatch diagnostics. \\
\end{longtable}

\begin{longtable}{@{}>{\raggedright\arraybackslash}p{0.18\textwidth}>{\raggedright\arraybackslash}p{0.28\textwidth}>{\raggedright\arraybackslash}p{0.50\textwidth}@{}}
\caption{Distinct notation for related quantities.}
\label{tab:notation_local_reuse_revised}\\
\toprule
Term & Dimension & Definition \\
\midrule
\endfirsthead
\toprule
Term & Dimension & Definition \\
\midrule
\endhead
\bottomrule
\endfoot
$\ell$, $S$ & loss function; split label & $\ell$ is the per-example loss. $S\in\{\mathrm{few},\mathrm{full},\mathrm{test}\}$ indexes empirical objectives. \\
$\mathcal{D}_{\mathrm{few}}$, $\mathcal{D}_{\mathrm{full}}$, $\mathcal{D}_{\mathrm{test}}$ & datasets & Support, full-training, and test datasets, respectively. \\
$m_t$, $m_{\mathcal{D},t}^{(e)}$ & vectors & Optimizer first-moment state and encoder-restricted dataset mean gradient, respectively. \\
$v_t$, $v_{\mathcal{D},t}^{(e)}$ & nonnegative vectors & Optimizer second-moment state and encoder-restricted dataset second moment, respectively. \\
$\Theta_t$ & parameter vector & Trainable parameters at step $t$; the same symbol is used in drift analyses. \\
$u_t^{\mathrm{sgd}}$, $a_t^{(p)}$, $a_t^{(b)}$ & vectors & Unit SGD direction, norm-matched probe displacement, and block-restricted displacement, respectively. \\
$u_t^{\mathrm{top}}$, $u_t^{\mathrm{top\text{-}few}}$ & unit vectors & Dominant full-training and support Hessian directions, respectively. \\
$p_t(c\mid x)$, $p_t^{\mathrm{step}}$ & probability; unit vector & Class probability and unit direction in the step-length analysis, respectively. \\
$\zeta_{\mathrm{err}}$, $\zeta_{\mathrm{loc}}$ & positive scalars & Uniform-error tolerance and local Taylor-expansion tolerance in the theory. \\
$\mathcal{L}_t$, $L_{\mathrm{sm},t}$ & loss function; positive scalar & Mini-batch loss and local smoothness bound, respectively. \\
$q_t^{\mathrm{sig}}$, $q_t^{\mathrm{err}}$ & vectors & Full-data signal and numerator-error term in the theoretical proof. \\
$\xi_t$, $\xi_t^{\mathrm{rand}}$ & scalar; random vector & Scale-inflation factor and raw random control vector, respectively. \\
\end{longtable}

\subsection*{Location-specific notation}

The following tables collect the quantities used in each section. Their definitions are consistent with the common notation above.

\begin{longtable}{@{}>{\raggedright\arraybackslash}p{0.23\textwidth}>{\raggedright\arraybackslash}p{0.23\textwidth}>{\raggedright\arraybackslash}p{0.50\textwidth}@{}}
\caption{Secondary notation for main Section~3: Why Does Adam Fail?}
\label{tab:not_sec3}\\
\toprule
Term & Dimension & Definition \\
\midrule
\endfirsthead
\toprule
Term & Dimension & Definition \\
\midrule
\endhead
\bottomrule
\endfoot
$R_v^{(e)}(0)$ & nonnegative scalar & Initialization ratio $\|v_{\mathrm{few},0}^{(e)}\|_2/\|v_{\mathrm{full},0}^{(e)}\|_2$. \\
$R_{u,0}^{(e)}$ & nonnegative scalar & Initialization adaptive norm ratio $\|u_{ff,0}^{(e)}\|_2/\|u_{rr,0}^{(e)}\|_2$. \\
$R_m^{(e)}(t)$ & nonnegative scalar & Numerator ratio $\|m_{\mathrm{few},t}^{(e)}\|_2/\|m_{\mathrm{full},t}^{(e)}\|_2$. \\
$R_v^{(e)}(t)$ & nonnegative scalar & Second-moment ratio $\|v_{\mathrm{few},t}^{(e)}\|_2/\|v_{\mathrm{full},t}^{(e)}\|_2$. \\
$C_m^{(e)}$, $C_v^{(e)}$ & nonnegative scalars & Collapse factors for the numerator ratio and second-moment ratio between initialization and the final logged checkpoint. \\
$o$ & optimizer label & Trajectory index in $\{\mathrm{SGD},\mathrm{Adam},\mathrm{Adam\mbox{-}M}\}$. \\
$\Theta_t^{(o)}$ & $\mathbb{R}^d$ & Parameter vector on optimizer trajectory $o$. \\
$\Delta\Theta_t^{(o)}$ & $\mathbb{R}^d$ & Update vector $\Theta_{t+1}^{(o)}-\Theta_t^{(o)}$. \\
$r_o(t)$ & nonnegative scalar & Global relative update budget $\|\Delta\Theta_t^{(o)}\|_2/\|\Theta_t^{(o)}\|_2$. \\
$\mathcal{O}$ & optimizer family & Structured preconditioner family, usually $\mathcal{O}\in\{\mathrm{Shampoo},\mathrm{SOAP}\}$. \\
$B_{\mathcal{O},\mathcal{D},t}^{(e)}$ & linear map $\mathbb{R}^{d_e}\to\mathbb{R}^{d_e}$ & Effective structured preconditioning map for family $\mathcal{O}$, split $\mathcal{D}$, and encoder $e$. \\
$q_{ff,t}^{(\mathcal{O},e)}$, $q_{fr,t}^{(\mathcal{O},e)}$, $q_{rr,t}^{(\mathcal{O},e)}$ & $\mathbb{R}^{d_e}$ & Structured preconditioned few/few, few/full, and full/full diagnostic directions. \\
$S_{\mathcal{O}}^{(e)}(t)$ & nonnegative scalar & Structured scale ratio $\|q_{ff,t}^{(\mathcal{O},e)}\|_2/\|q_{fr,t}^{(\mathcal{O},e)}\|_2$. \\
$A_{\mathcal{O}}^{(e)}(t)$ & scalar in $[-1,1]$ & Alignment between repaired structured direction and full-data structured reference. \\
\end{longtable}

\begin{longtable}{@{}>{\raggedright\arraybackslash}p{0.23\textwidth}>{\raggedright\arraybackslash}p{0.23\textwidth}>{\raggedright\arraybackslash}p{0.50\textwidth}@{}}
\caption{Secondary notation for main Section~4: Why Does SGD Perform Well?}
\label{tab:not_sec4}\\
\toprule
Term & Dimension & Definition \\
\midrule
\endfirsthead
\toprule
Term & Dimension & Definition \\
\midrule
\endhead
\bottomrule
\endfoot
$S$ & split label & Common objective index in $\{\mathrm{few},\mathrm{full},\mathrm{test}\}$. \\
$\Theta(\alpha)$ & $\mathbb{R}^{d}$ & Linear interpolation between pretrained and final SGD checkpoints. \\
$\varphi_S(\alpha)$ & scalar & Line loss $L_S(\Theta(\alpha))$ for $S\in\{\mathrm{few},\mathrm{full},\mathrm{test}\}$. \\
$\alpha_S^\star$, $\varphi_S^\star$ & scalar & Minimizer of the line loss and the corresponding minimum value. \\
$\varphi_S^{\mathrm{lin}}$ & scalar function & Endpoint chord used to compute the interpolation barrier. \\
$B_S$ & scalar & Interpolation barrier above the endpoint chord. \\
$W_S^{(\delta_{\mathrm w})}$ & scalar in $[0,1]$ & Normalized near-minimum width at tolerance $\delta_{\mathrm w}$. \\
$H_S(\Theta_t)$ & $\mathbb{R}^{d\times d}$ & Hessian of loss $L_S$ at checkpoint $\Theta_t$. \\
$\kappa_S(u;t)$ & scalar & Directional curvature $u^\top H_S(\Theta_t)u$. \\
$u_t^{\mathrm{sgd}}$, $u_t^{\mathrm{full}}$, $u_t^{\mathrm{top}}$, $u_t^{\mathrm{rand}}$ & unit vectors in $\mathbb{R}^{d}$ & Actual SGD direction, full-train gradient direction, dominant full-Hessian direction, and random control direction. \\
$b=(e,j)$ & block index & Encoder--Transformer-block pair. \\
$P_b$ & $\mathbb{R}^{d\times d}$ & Diagonal projector that keeps the coordinates of block $b$. \\
$a_t^{(b)}$ & $\mathbb{R}^{d}$ & Block-restricted SGD step $-\eta P_b g_t$. \\
$\psi_S^{(b)}(\alpha;t)$ & scalar & Blockwise line loss $L_S(\Theta_t+\alpha a_t^{(b)})$. \\
$\Delta_S^{(b)}(t;\alpha)$ & scalar & Loss change along the blockwise line. \\
$\alpha_{\mathrm{full}}^{\star,(b)}(t)$ & scalar in $[-2,2]$ & Locally preferred full-loss scale for block $b$. \\
$\Theta_T^{(s)}$, $\Theta_T^{(s')}$ & $\mathbb{R}^{d}$ & Final SGD states from two independent support-set seeds. \\
$\Theta^{(s,s')}(\alpha)$ & $\mathbb{R}^{d}$ & Linear interpolation between two final SGD seed solutions. \\
$D_{\mathrm{rel}}^{(s,s')}$ & nonnegative scalar & Relative endpoint separation between two final SGD solutions. \\
\end{longtable}

\begin{longtable}{@{}>{\raggedright\arraybackslash}p{0.22\textwidth}>{\raggedright\arraybackslash}p{0.23\textwidth}>{\raggedright\arraybackslash}p{0.51\textwidth}@{}}
\caption{Secondary notation for Appendix~B: Few-shot Adaptation of CLIP.}
\label{tab:not_appB}\\
\toprule
Term & Dimension & Definition \\
\midrule
\endfirsthead
\toprule
Term & Dimension & Definition \\
\midrule
\endhead
\bottomrule
\endfoot
$F_{\Theta_t^{(\mathrm{vision})}}$ & map to $\mathbb{R}^{d_{\mathrm{emb}}}$ & Unnormalized CLIP image encoder at step $t$. \\
$G_{\Theta_t^{(\mathrm{text})}}$ & map to $\mathbb{R}^{d_{\mathrm{emb}}}$ & Unnormalized CLIP text encoder at step $t$. \\
$h_{\mathrm{img},t}(x)$ & $\mathbb{R}^{d_{\mathrm{emb}}}$ & Normalized image feature for image $x$. \\
$a$ & text prompt & Prompt string passed to the text encoder. \\
$h_{\mathrm{prompt},t}(a)$ & $\mathbb{R}^{d_{\mathrm{emb}}}$ & Normalized text feature for prompt $a$. \\
$\mathcal{A}_c$ & set of $P$ prompts & Prompt set for class $c$, $\mathcal{A}_c=\{a_{j,c}\}_{j=1}^{P}$. \\
$P$ & positive integer & Number of prompts per class in the zero-shot CLIP classifier. \\
$\widetilde h_{\mathrm{text},t}(c)$ & $\mathbb{R}^{d_{\mathrm{emb}}}$ & Average unrenormalized class-text prototype. \\
$h_{\mathrm{text},t}(c)$ & $\mathbb{R}^{d_{\mathrm{emb}}}$ & Normalized class-text prototype. \\
$z_t(x)$ & $\mathbb{R}^{C}$ & CLIP class-logit vector for image $x$. \\
$\tau$ & positive scalar & CLIP temperature used in the logit definition. \\
$p_t(c\mid x)$ & scalar in $[0,1]$ & Predicted class probability for class $c$ and image $x$. \\
$\Delta^{C-1}$ & set & Probability simplex over $C$ classes. \\
\end{longtable}

\begin{longtable}{@{}>{\raggedright\arraybackslash}p{0.22\textwidth}>{\raggedright\arraybackslash}p{0.23\textwidth}>{\raggedright\arraybackslash}p{0.51\textwidth}@{}}
\caption{Secondary notation for Appendix~C: Algorithms.}
\label{tab:not_appC}\\
\toprule
Term & Dimension & Definition \\
\midrule
\endfirsthead
\toprule
Term & Dimension & Definition \\
\midrule
\endhead
\bottomrule
\endfoot
$\mathcal{B}_t$ & subset of $\mathcal{D}_{\mathrm{few}}$ & Support mini-batch at optimizer iteration $t$. \\
$\mathcal{D}_{\mathrm{few}}$ & support set & Few-shot support set in the optimizer algorithms. \\
$\mathcal{L}_t$ & $\mathbb{R}^{d}\to\mathbb{R}$ & Cross-entropy loss on $\mathcal{B}_t$. \\
$m_t$ & $\mathbb{R}^{d}$ & SMD or Adam first-moment state in Appendix~C. \\
$v_t$ & $\mathbb{R}^{d}_{\ge0}$ & Adam diagonal second-moment state in Appendix~C. \\
$\widehat m_t$, $\widehat v_t$ & vectors in $\mathbb{R}^{d}$ & Bias-corrected Adam first and second moments. \\
$\mathbf{1}_d$ & $\mathbb{R}^{d}$ & All-ones vector used in the Adam stabilizer term. \\
$R$ & positive integer & Number of trainable parameter tensors in structured optimizers. \\
$J$ & tensor index & Index of a trainable tensor, $J=1,\ldots,R$. \\
$p_J$ & positive integer & Order of tensor $J$. \\
$n_{J,j}$ & positive integer & Size of tensor $J$ along mode $j$. \\
$\Theta_t^{(J)}$ & tensor & Trainable parameter tensor $J$ at step $t$. \\
$N_J$ & positive integer & Number of scalar entries in tensor $J$. \\
$G_t^{(J)}$ & same shape as $\Theta_t^{(J)}$ & Gradient tensor for parameter tensor $J$. \\
$G_{t,(j)}^{(J)}$ & matrix & Mode-$j$ unfolding of $G_t^{(J)}$. \\
$\times_j$ & tensor operation & Mode-$j$ tensor-matrix product. \\
$I_n$ & $\mathbb{R}^{n\times n}$ & Identity matrix. \\
$A_{t,j}^{(J)}$ & $\mathbb{R}^{n_{J,j}\times n_{J,j}}$ & Shampoo/SOAP mode-wise second-order factor. \\
$P_{t,j}^{(J)}$ & $\mathbb{R}^{n_{J,j}\times n_{J,j}}$ & Shampoo inverse-root preconditioning factor. \\
$F$ & positive integer & Refresh frequency for structured preconditioners or SOAP bases. \\
$\widetilde G_t^{(J)}$ & same shape as $\Theta_t^{(J)}$ & Preconditioned tensor update before the parameter step. \\
$Q_{t,j}^{(J)}$ & orthogonal matrix & SOAP basis for mode $j$ of tensor $J$, initialized by eigendecomposition and refreshed by QR. \\
$Z_t^{(J)}$ & tensor & Gradient projected into the SOAP basis. \\
$M_t^{(J)}$, $V_t^{(J)}$ & tensors & SOAP first moment in the parameter basis and diagonal second-moment estimate associated with the SOAP basis, respectively. \\
$Y_t^{(J)}$ & tensor & SOAP bias-corrected Adam-like tensor direction in the projected basis. \\
\end{longtable}

\begin{longtable}{@{}>{\raggedright\arraybackslash}p{0.22\textwidth}>{\raggedright\arraybackslash}p{0.23\textwidth}>{\raggedright\arraybackslash}p{0.51\textwidth}@{}}
\caption{Secondary notation for Appendix~D: Datasets.}
\label{tab:not_appD}\\
\toprule
Term & Dimension & Definition \\
\midrule
\endfirsthead
\toprule
Term & Dimension & Definition \\
\midrule
\endhead
\bottomrule
\endfoot
$\mathrm{ES}$, $\mathrm{OP}$, $\mathrm{DTD}$ & dataset abbreviations & EuroSAT, Oxford-Pets, and Describable Textures Dataset. These are the three main diagnostic datasets. \\
$\mathrm{V}$, $\mathrm{T}$ & encoder abbreviations & Vision and text encoders in diagnostic figures. \\
$K$ & positive integer & Shot level used to sample the support set. \\
$\mathcal{D}_{\mathrm{train}}$ & training split & Training split from which few-shot support sets and full-train references are formed. \\
$\mathcal{D}_{\mathrm{test}}$ & test split & Held-out evaluation split for each dataset. \\
\end{longtable}

\begin{longtable}{@{}>{\raggedright\arraybackslash}p{0.22\textwidth}>{\raggedright\arraybackslash}p{0.23\textwidth}>{\raggedright\arraybackslash}p{0.51\textwidth}@{}}
\caption{Secondary notation for Appendix~E: Hyperparameters.}
\label{tab:not_appE}\\
\toprule
Term & Dimension & Definition \\
\midrule
\endfirsthead
\toprule
Term & Dimension & Definition \\
\midrule
\endhead
\bottomrule
\endfoot
$B$ & positive integer & Training batch size in benchmark and stability experiments. \\
$\eta$ & positive scalar & Learning rate in optimizer sweeps and method configurations. \\
$\omega_{\mathrm{wd}}$ & nonnegative scalar & Weight decay coefficient in full-fine-tuning optimizer hyperparameters. \\
$\beta_1$, $\beta_2$ & scalars in $(0,1)$ & Adam, SMD, Shampoo, or SOAP moment coefficients. \\
$\varepsilon$ & positive scalar & Numerical stabilizer in AdamW and related adaptive optimizers. \\
$\tau_{\mathrm{ProGrad}}$ & positive scalar & ProGrad softmax temperature in the hyperparameter table. \\
$r$ & positive integer & Low-rank dimension for LoRA, PiSSA, or GaLoRE. \\
$\rho$ & scalar in $(0,1]$ & Sparse active-coordinate density for SO and SAFT. \\
$\alpha_1$, $\alpha_2$ & positive scalars & Adapter mixing coefficients in the PEFT hyperparameter tables. \\
$\alpha_{\mathrm{LoRA}}$, $\alpha_{\mathrm{PiSSA}}$ & positive scalars & LoRA and PiSSA scaling coefficients, respectively. \\
$T_{\mathrm{update}}$ & positive integer & Refresh period for dynamic sparse or low-rank supports. \\
$F$ & positive integer & Preconditioning or basis refresh frequency for Shampoo and SOAP. \\
\end{longtable}

\begin{longtable}{@{}>{\raggedright\arraybackslash}p{0.22\textwidth}>{\raggedright\arraybackslash}p{0.23\textwidth}>{\raggedright\arraybackslash}p{0.51\textwidth}@{}}
\caption{Secondary notation for Appendix~F: Training Time \& Memory Analysis.}
\label{tab:not_appF}\\
\toprule
Term & Dimension & Definition \\
\midrule
\endfirsthead
\toprule
Term & Dimension & Definition \\
\midrule
\endhead
\bottomrule
\endfoot
$W$ & $\mathbb{R}^{d_{\mathrm{out}}\times d_{\mathrm{in}}}$ & Weight matrix of an idealized fully connected layer in the memory model. \\
$d_{\mathrm{out}}$ & positive integer & Output dimension of the layer. \\
$d_{\mathrm{in}}$ & positive integer & Input dimension of the layer in the memory model. \\
$r$ & positive integer & Low-rank dimension for LoRA, PiSSA, and GaLoRE memory counts. \\
$\rho$ & scalar in $(0,1]$ & Sparse update density. The sparsity ratio is $1-\rho$. \\
$d_{\mathrm{out}}d_{\mathrm{in}}$ & scalar count & Number of dense weight scalars in one fully connected layer. \\
$d_{\mathrm{out}}r+d_{\mathrm{in}}r$ & scalar count & Number of trainable low-rank factor scalars for one layer. \\
M & million scalar count & Unit used in memory tables for number of scalar variables. \\
MiB & memory unit & Memory in mebibytes ($2^{20}$ bytes) assuming 32-bit floating-point storage. \\
\end{longtable}

\begin{longtable}{@{}>{\raggedright\arraybackslash}p{0.22\textwidth}>{\raggedright\arraybackslash}p{0.23\textwidth}>{\raggedright\arraybackslash}p{0.51\textwidth}@{}}
\caption{Secondary notation for Appendix~G: SGD vs SMD vs Adam.}
\label{tab:not_appG}\\
\toprule
Term & Dimension & Definition \\
\midrule
\endfirsthead
\toprule
Term & Dimension & Definition \\
\midrule
\endhead
\bottomrule
\endfoot
$\mathrm{Acc}$ & percentage & Top-1 test accuracy in the SGD/SMD/Adam benchmark. \\
$K$ & positive integer & Shot level on the horizontal axis of few-shot results. \\
$\mathrm{SMD}$ & method label & SGD with momentum. \\
$\mu\pm\sigma$ & mean and standard deviation & Reported accuracy summary over random seeds. \\
\end{longtable}

\begin{longtable}{@{}>{\raggedright\arraybackslash}p{0.22\textwidth}>{\raggedright\arraybackslash}p{0.23\textwidth}>{\raggedright\arraybackslash}p{0.51\textwidth}@{}}
\caption{Secondary notation for Appendix~H: SGD vs PEFT.}
\label{tab:not_appH}\\
\toprule
Term & Dimension & Definition \\
\midrule
\endfirsthead
\toprule
Term & Dimension & Definition \\
\midrule
\endhead
\bottomrule
\endfoot
$\mathrm{PEFT}$ & method family label & Parameter-efficient fine-tuning family used for comparisons against dense SGD. \\
$\mathrm{SO}$, $\mathrm{LoRA}$, $\mathrm{SAFT}$, $\mathrm{LP{++}}$, $\mathrm{CoOp}$ & method labels & Representative PEFT methods in the benchmark. \\
$K$ & positive integer & Shot level in the PEFT comparison tables. \\
$\mathrm{Acc}$ & percentage & Top-1 test accuracy. \\
\end{longtable}

\begin{longtable}{@{}>{\raggedright\arraybackslash}p{0.22\textwidth}>{\raggedright\arraybackslash}p{0.23\textwidth}>{\raggedright\arraybackslash}p{0.51\textwidth}@{}}
\caption{Secondary notation for Appendix~I.1: SGD vs PEFT stability.}
\label{tab:not_appI1}\\
\toprule
Term & Dimension & Definition \\
\midrule
\endfirsthead
\toprule
Term & Dimension & Definition \\
\midrule
\endhead
\bottomrule
\endfoot
$\eta$ & positive scalar & Learning rate varied in the PEFT stability sweep. \\
$t$ & epoch & Training epoch used as the horizontal axis in stability curves. \\
$\mathrm{LoRA}$, $\mathrm{GaLoRE}$, $\mathrm{SAFT}$ & method labels & PEFT methods compared with SGD in the first stability subsection. \\
$\mu\pm\sigma$ & mean and standard deviation & Accuracy summary over three fixed seeds. \\
\end{longtable}

\begin{longtable}{@{}>{\raggedright\arraybackslash}p{0.22\textwidth}>{\raggedright\arraybackslash}p{0.23\textwidth}>{\raggedright\arraybackslash}p{0.51\textwidth}@{}}
\caption{Secondary notation for Appendix~I.2: SGD vs Adam vs Shampoo vs SOAP stability.}
\label{tab:not_appI2}\\
\toprule
Term & Dimension & Definition \\
\midrule
\endfirsthead
\toprule
Term & Dimension & Definition \\
\midrule
\endhead
\bottomrule
\endfoot
$\beta_2$ & scalar in $(0,1)$ & Adam second-moment coefficient used in the stability sweep. \\
$F$ & positive integer & Preconditioning frequency for Shampoo or SOAP. \\
$\eta$ & positive scalar & Learning rate varied in the second-order stability sweep. \\
$t$ & epoch & Training epoch used as the horizontal axis. \\
$\mathrm{Shampoo}$, $\mathrm{SOAP}$ & method labels & Structured preconditioned optimizers compared with SGD and Adam. \\
\end{longtable}

\begin{longtable}{@{}>{\raggedright\arraybackslash}p{0.22\textwidth}>{\raggedright\arraybackslash}p{0.23\textwidth}>{\raggedright\arraybackslash}p{0.51\textwidth}@{}}
\caption{Secondary notation for Appendix~J: SGD vs Adam for PEFT.}
\label{tab:not_appJ}\\
\toprule
Term & Dimension & Definition \\
\midrule
\endfirsthead
\toprule
Term & Dimension & Definition \\
\midrule
\endhead
\bottomrule
\endfoot
$\mathrm{Adam}$, $\mathrm{SGD}$ & optimizer labels & Optimizers used inside each PEFT family and dense full fine-tuning. \\
$\mathrm{Adapter}$, $\mathrm{TIP}$, $\mathrm{LP{++}}$ & method labels & Adapter-style PEFT methods. \\
$\mathrm{CoOp}$, $\mathrm{KgCoOp}$, $\mathrm{ProGrad}$ & method labels & Prompt-learning PEFT methods. \\
$\mathrm{LoRA}$, $\mathrm{PiSSA}$, $\mathrm{GaLoRE}$ & method labels & Low-rank or projected-gradient PEFT methods. \\
$\mathrm{LayerNorm}$, $\mathrm{BitFit}$, $\mathrm{SAFT}$, $\mathrm{SO}$ & method labels & Sparse or subset-based PEFT methods. \\
$K$ & positive integer & Shot level in each ablation table. \\
\end{longtable}

\begin{longtable}{@{}>{\raggedright\arraybackslash}p{0.22\textwidth}>{\raggedright\arraybackslash}p{0.23\textwidth}>{\raggedright\arraybackslash}p{0.51\textwidth}@{}}
\caption{Secondary notation for Appendix~K.1: Second-moment miscalibration.}
\label{tab:not_K1}\\
\toprule
Term & Dimension & Definition \\
\midrule
\endfirsthead
\toprule
Term & Dimension & Definition \\
\midrule
\endhead
\bottomrule
\endfoot
$\mathcal{B}$ & mini-batch subset of $\mathcal{D}$ & Diagnostic mini-batch used to estimate empirical first and second moments. \\
$i$ & coordinate index & Coordinate index in encoder $e$, $i=1,\ldots,d_e$. \\
$\zeta_{t,i}^{(e)}$ & scalar & Base-10 coordinate log-ratio of support and full-data diagonal second moments. \\
$M_t^{(e)}$ & nonnegative scalar & Mean absolute coordinate log-ratio in encoder $e$. \\
$k$ & positive integer & Number of salient coordinates selected by top-$k$. \\
$\rho_{\mathrm{top}}$ & scalar in $(0,1]$ & Density used to set $k=\lceil\rho_{\mathrm{top}} d_e\rceil$. \\
$O_t^{(e)}(\rho_{\mathrm{top}})$ & scalar in $[0,1]$ & Overlap between top-$k$ coordinates of $v_{\mathrm{few},t}^{(e)}$ and $v_{\mathrm{full},t}^{(e)}$. \\
\end{longtable}

\begin{longtable}{@{}>{\raggedright\arraybackslash}p{0.22\textwidth}>{\raggedright\arraybackslash}p{0.23\textwidth}>{\raggedright\arraybackslash}p{0.51\textwidth}@{}}
\caption{Secondary notation for Appendix~K.2: Initialization audit at $t=0$.}
\label{tab:not_K2}\\
\toprule
Term & Dimension & Definition \\
\midrule
\endfirsthead
\toprule
Term & Dimension & Definition \\
\midrule
\endhead
\bottomrule
\endfoot
$A_g^{(e)}(0)$ & scalar in $[-1,1]$ & Initialization cosine between $m_{\mathrm{few},0}^{(e)}$ and $m_{\mathrm{full},0}^{(e)}$. \\
$R_v^{(e)}(0)$ & nonnegative scalar & Initialization denominator norm ratio. \\
$M_0^{(e)}$ & nonnegative scalar & Coordinatewise denominator mismatch at initialization. \\
$R_{u,0}^{(e)}$ & nonnegative scalar & Initialization adaptive-direction norm ratio. \\
$M_{0,j}^{(e)}$ & nonnegative scalar & Mean absolute coordinate log-ratio restricted to Transformer block $j$ of encoder $e$. \\
$u_{fr,0}^{(e,j)}$, $u_{rr,0}^{(e,j)}$ & $\mathbb{R}^{d_{e,j}}$ & Block-restricted oracle-denominator and full-data reference directions. \\
\end{longtable}

\begin{longtable}{@{}>{\raggedright\arraybackslash}p{0.22\textwidth}>{\raggedright\arraybackslash}p{0.23\textwidth}>{\raggedright\arraybackslash}p{0.51\textwidth}@{}}
\caption{Secondary notation for Appendix~K.3: Gradient shrinkage versus second-moment collapse.}
\label{tab:not_K3}\\
\toprule
Term & Dimension & Definition \\
\midrule
\endfirsthead
\toprule
Term & Dimension & Definition \\
\midrule
\endhead
\bottomrule
\endfoot
$R_m^{(e)}(t)$ & nonnegative scalar & Numerator ratio between support and full-train gradients. \\
$R_v^{(e)}(t)$ & nonnegative scalar & Denominator-statistic ratio between support and full-train second moments. \\
$T$ & final logged checkpoint & Endpoint used to compute collapse factors. \\
$C_m^{(e)}$ & nonnegative scalar & Collapse factor $R_m^{(e)}(0)/R_m^{(e)}(T)$. \\
$C_v^{(e)}$ & nonnegative scalar & Collapse factor $R_v^{(e)}(0)/R_v^{(e)}(T)$. \\
\end{longtable}

\begin{longtable}{@{}>{\raggedright\arraybackslash}p{0.22\textwidth}>{\raggedright\arraybackslash}p{0.23\textwidth}>{\raggedright\arraybackslash}p{0.51\textwidth}@{}}
\caption{Secondary notation for Appendix~K.4: Instantaneous, batch, and EMA denominators. The source-specific diagnostics are shown for the instantaneous source. The batch and EMA variants replace the literal subscript $\mathrm{inst}$ by $\mathrm{batch}$ and $\mathrm{ema}$, respectively.}
\label{tab:not_K4}\\
\toprule
Term & Dimension & Definition \\
\midrule
\endfirsthead
\toprule
Term & Dimension & Definition \\
\midrule
\endhead
\bottomrule
\endfoot
$v_{\mathrm{inst},t}^{(e)}$ & $\mathbb{R}_{\ge0}^{d_e}$ & Squared full-support gradient $m_{\mathrm{few},t}^{(e)}\odot m_{\mathrm{few},t}^{(e)}$. \\
$v_{\mathrm{batch},t}^{(e)}$ & $\mathbb{R}_{\ge0}^{d_e}$ & Support-batch diagonal second moment. \\
$v_t^{(e)}$ & $\mathbb{R}_{\ge0}^{d_e}$ & Encoder restriction of Adam's uncorrected second-moment state $v_t$. \\
$v_{\mathrm{ema},t}^{(e)}$ & $\mathbb{R}_{\ge0}^{d_e}$ & Bias-corrected Adam running second moment. \\
$u_{\mathrm{inst},t}^{(e)}$ & $\mathbb{R}^{d_e}$ & Adaptive direction using the instantaneous second moment. \\
$R_{v,\mathrm{inst}}^{(e)}(t)$ & nonnegative scalar & Instantaneous second-moment norm ratio. \\
$M_{\mathrm{inst}}^{(e)}(t)$ & nonnegative scalar & Coordinatewise mismatch of the instantaneous second moment. \\
$R_{u,\mathrm{inst}}^{(e)}(t)$ & nonnegative scalar & Adaptive norm ratio for the instantaneous second moment. \\
$A_{u,\mathrm{inst}}^{(e)}(t)$ & scalar in $[-1,1]$ & Cosine between $u_{\mathrm{inst},t}^{(e)}$ and $u_{rr,t}^{(e)}$. \\
\end{longtable}

\begin{longtable}{@{}>{\raggedright\arraybackslash}p{0.22\textwidth}>{\raggedright\arraybackslash}p{0.23\textwidth}>{\raggedright\arraybackslash}p{0.51\textwidth}@{}}
\caption{Secondary notation for Appendix~K.5: Denominator-only counterfactuals.}
\label{tab:not_K5}\\
\toprule
Term & Dimension & Definition \\
\midrule
\endfirsthead
\toprule
Term & Dimension & Definition \\
\midrule
\endhead
\bottomrule
\endfoot
$h$ & positive integer & Lag length for the short-lag denominator, set to $5$ in the experiment. \\
$u_{f0,t}$ & $\mathbb{R}^{d}$ & Direction using current few-shot numerator and initialization denominator $v_{\mathrm{few},0}$. \\
$u_{f\mathrm{lag},t}$ & $\mathbb{R}^{d}$ & Direction using current few-shot numerator and lagged denominator $v_{\mathrm{few},t-h}$. \\
$q$ & direction identifier & Element of $\{ff,fr,f0,f\mathrm{lag},rr\}$. \\
$A_q^{(e)}(t)$ & scalar in $[-1,1]$ & Alignment between $u_{q,t}^{(e)}$ and $u_{rr,t}^{(e)}$. \\
$S_q^{(e)}(t)$ & nonnegative scalar & Scale ratio $\|u_{q,t}^{(e)}\|_2/\|u_{rr,t}^{(e)}\|_2$. \\
$\widetilde u_{q,t}$ & $\mathbb{R}^{d}$ & Norm-matched version of $u_{q,t}$ with global norm $\|u_{ff,t}\|_2$. \\
$\alpha_{\mathrm p}$ & positive scalar & Probe multiplier in the temporary counterfactual update. \\
$\Theta_t^{[q]}$ & $\mathbb{R}^{d}$ & Temporary parameter vector after applying probe direction $q$. \\
$\Delta L_{S,q}(t)$ & scalar & Loss change on $S\in\{\mathrm{few},\mathrm{full},\mathrm{test}\}$ after a norm-matched probe. \\
\end{longtable}

\begin{longtable}{@{}>{\raggedright\arraybackslash}p{0.22\textwidth}>{\raggedright\arraybackslash}p{0.23\textwidth}>{\raggedright\arraybackslash}p{0.51\textwidth}@{}}
\caption{Secondary notation for Appendix~K.6: Scale-versus-geometry experiment.}
\label{tab:not_K6}\\
\toprule
Term & Dimension & Definition \\
\midrule
\endfirsthead
\toprule
Term & Dimension & Definition \\
\midrule
\endhead
\bottomrule
\endfoot
$o$ & optimizer label & Trajectory index in $\{\mathrm{SGD},\mathrm{Adam},\mathrm{Adam\mbox{-}M}\}$. \\
$\Delta\Theta_t^{(o)}$ & $\mathbb{R}^{d}$ & Update vector on trajectory $o$. \\
$r_o(t)$ & nonnegative scalar & Global relative update budget. \\
$\gamma$ & positive scalar & One-time learning-rate multiplier that matches Adam's first-step budget to SGD. \\
$\eta_{\mathrm{Adam\mbox{-}M}}$ & positive scalar & Learning rate of step-0 matched Adam. \\
$u_{fr,t}^{(o,e)}$, $u_{rr,t}^{(o,e)}$ & $\mathbb{R}^{d_e}$ & Trajectory-specific numerator-side geometry directions with full-data denominator. \\
$A_o^{(e)}(t)$ & scalar in $[-1,1]$ & Numerator-side geometry diagnostic for encoder $e$ on trajectory $o$. \\
$\overline A_o(t)$ & scalar in $[-1,1]$ & Average of the vision and text values of $A_o^{(e)}(t)$. \\
\end{longtable}

\begin{longtable}{@{}>{\raggedright\arraybackslash}p{0.22\textwidth}>{\raggedright\arraybackslash}p{0.23\textwidth}>{\raggedright\arraybackslash}p{0.51\textwidth}@{}}
\caption{Secondary notation for Appendix~K.7: Parameter, feature, logit, and margin drift.}
\label{tab:not_K7}\\
\toprule
Term & Dimension & Definition \\
\midrule
\endfirsthead
\toprule
Term & Dimension & Definition \\
\midrule
\endhead
\bottomrule
\endfoot
$G$ & coordinate group & Subset of trainable parameters, such as all parameters, one encoder, or one block. \\
$\Theta_t^{(o,G)}$ & $\mathbb{R}^{|G|}$ & Parameters of group $G$ on trajectory $o$. \\
$R_\Theta^{(o,G)}(t)$ & nonnegative scalar & Relative parameter drift from the pretrained state. \\
$h_{\mathrm{img},t}^{(o)}(x)$ & $\mathbb{R}^{d_{\mathrm{emb}}}$ & Normalized image feature on trajectory $o$. \\
$h_{\mathrm{text},t}^{(o)}(c)$ & $\mathbb{R}^{d_{\mathrm{emb}}}$ & Normalized class-text feature on trajectory $o$. \\
$R_{\mathrm{img},S}^{(o)}(t)$ & nonnegative scalar & Mean angular image-feature drift on split $S$. \\
$R_{\mathrm{text}}^{(o)}(t)$ & nonnegative scalar & Mean angular text-feature drift over classes. \\
$R_{\mathrm{logit},S}^{(o)}(t)$ & nonnegative scalar & Mean absolute logit drift on split $S$. \\
$\mu_t^{(o)}(x,y)$ & scalar & Ground-truth class margin. \\
$R_{\mathrm{mar},S}^{(o)}(t)$ & nonnegative scalar & Mean absolute margin drift on split $S$. \\
\end{longtable}

\begin{longtable}{@{}>{\raggedright\arraybackslash}p{0.22\textwidth}>{\raggedright\arraybackslash}p{0.23\textwidth}>{\raggedright\arraybackslash}p{0.51\textwidth}@{}}
\caption{Secondary notation for Appendix~K.8: Temporal few-shot gradient staleness.}
\label{tab:not_K8}\\
\toprule
Term & Dimension & Definition \\
\midrule
\endfirsthead
\toprule
Term & Dimension & Definition \\
\midrule
\endhead
\bottomrule
\endfoot
$\Delta$ & positive integer & Temporal lag in optimizer steps or epochs. \\
$m_{\mathrm{few},t}^{(o,e)}$ & $\mathbb{R}^{d_e}$ & Support gradient at step $t$ on trajectory $o$, restricted to encoder $e$. \\
$m_{\mathrm{full},t}^{(o,e)}$ & $\mathbb{R}^{d_e}$ & Full-train gradient at step $t$ on trajectory $o$, restricted to encoder $e$. \\
$A_{\mathrm{self}}^{(o,e)}(t,\Delta)$ & scalar in $[-1,1]$ & Cosine between support gradients at steps $t$ and $t+\Delta$. \\
$A_{\mathrm{full}}^{(o,e)}(t,\Delta)$ & scalar in $[-1,1]$ & Cosine between support gradient at $t$ and full-train gradient at $t+\Delta$. \\
$d_{e,j}$ & positive integer & Number of trainable scalar parameters in block $j$ of encoder $e$. \\
$m_{\mathrm{few},t}^{(o,e,j)}$, $m_{\mathrm{full},t}^{(o,e,j)}$ & $\mathbb{R}^{d_{e,j}}$ & Block-restricted support and full-train gradients. \\
$A_{\mathrm{full}}^{(o,e,j)}(t,\Delta)$ & scalar in $[-1,1]$ & Blockwise future full-data alignment. \\
\end{longtable}

\begin{longtable}{@{}>{\raggedright\arraybackslash}p{0.22\textwidth}>{\raggedright\arraybackslash}p{0.23\textwidth}>{\raggedright\arraybackslash}p{0.51\textwidth}@{}}
\caption{Secondary notation for Appendix~K.9: Structured preconditioners.}
\label{tab:not_K9}\\
\toprule
Term & Dimension & Definition \\
\midrule
\endfirsthead
\toprule
Term & Dimension & Definition \\
\midrule
\endhead
\bottomrule
\endfoot
$\mathcal{O}$ & optimizer family & Structured preconditioner family, usually Shampoo or SOAP. \\
$B_{\mathcal{O},\mathcal{D},t}^{(e)}$ & linear map $\mathbb{R}^{d_e}\to\mathbb{R}^{d_e}$ & Effective structured preconditioning map. \\
$q_{ff,t}^{(\mathcal{O},e)}$, $q_{fr,t}^{(\mathcal{O},e)}$, $q_{rr,t}^{(\mathcal{O},e)}$ & $\mathbb{R}^{d_e}$ & Structured preconditioned directions. \\
$S_{\mathcal{O}}^{(e)}(t)$ & nonnegative scalar & Scale ratio of few-shot and full-data structured preconditioners with the same numerator. \\
$A_{\mathcal{O}}^{(e)}(t)$ & scalar in $[-1,1]$ & Repaired reference alignment for the structured preconditioner. \\
$z_{\mathcal{O},\mathcal{D},t}^{(e)}$ & $\mathbb{R}_{\ge0}^{p_{\mathcal{O},e}}$ & Flattened statistic used to build the structured preconditioner. \\
$p_{\mathcal{O},e}$ & positive integer & Number of entries in the flattened structured statistic. \\
$M_{\mathcal{O}}^{(e)}(t)$ & nonnegative scalar & Coordinate mismatch of the structured statistic. \\
$O_{\mathcal{O}}^{(e)}(\rho_{\mathrm{top}};t)$ & scalar in $[0,1]$ & Top-coordinate overlap between support and full-data structured statistics. \\
$P_{e,j}$ & projection map & Projection onto Transformer block $j$ of encoder $e$. \\
$q_{fr,t}^{(\mathcal{O},e,j)}$, $q_{rr,t}^{(\mathcal{O},e,j)}$ & $\mathbb{R}^{d_{e,j}}$ & Block-restricted structured directions. \\
\end{longtable}

\begin{longtable}{@{}>{\raggedright\arraybackslash}p{0.22\textwidth}>{\raggedright\arraybackslash}p{0.23\textwidth}>{\raggedright\arraybackslash}p{0.51\textwidth}@{}}
\caption{Secondary notation for Appendix~L.1: Finite-sample error in diagonal second-moment estimation.}
\label{tab:not_L1}\\
\toprule
Term & Dimension & Definition \\
\midrule
\endfirsthead
\toprule
Term & Dimension & Definition \\
\midrule
\endhead
\bottomrule
\endfoot
$z_i=(x_i,y_i)$ & labeled example & Example $i$ in the full training split. \\
$I_{\mathrm{few}}$ & subset of $\{1,\ldots,N\}$ & Indices selected into the few-shot support set. \\
$a_{i,t}^{(e)}$ & $\mathbb{R}^{d_e}$ & Per-example encoder gradient of the cross-entropy loss. \\
$X_{i,t,j}^{(e)}$ & nonnegative scalar & Squared per-example gradient coordinate $(a_{i,t,j}^{(e)})^2$. \\
$\widehat v_{\mathrm{few},t,j}^{(e)}$ & nonnegative scalar & Finite-population estimator of the few-shot second-moment coordinate. \\
$\sigma_{t,j}^{2,(e)}$ & nonnegative scalar & Finite-population variance of $X_{i,t,j}^{(e)}$. \\
$\mathcal{U}_t^{(e)}$ & nonnegative scalar & Dimensionless effective variability of the full-data diagonal statistic. \\
$S_i$ & Bernoulli indicator & Indicator that example $i$ is included in $I_{\mathrm{few}}$. \\
$\widehat v_j$, $v_j$, $\sigma_j^2$ & scalars & Short proof notation for the few-shot estimate, its full-data target, and the finite-population squared-gradient variance. \\
$B_j$ & positive scalar & Nonzero squared-gradient amplitude in the finite-population observability example. \\
$r_j$ & integer & Number of examples with squared-gradient value $B_j$. \\
$\zeta_{\mathrm{err}}$ & positive scalar & Relative-error tolerance in the high-probability diagnostic. \\
\end{longtable}

\begin{longtable}{@{}>{\raggedright\arraybackslash}p{0.22\textwidth}>{\raggedright\arraybackslash}p{0.23\textwidth}>{\raggedright\arraybackslash}p{0.51\textwidth}@{}}
\caption{Secondary notation for Appendix~L.2: Inverse-square-root amplification.}
\label{tab:not_L2}\\
\toprule
Term & Dimension & Definition \\
\midrule
\endfirsthead
\toprule
Term & Dimension & Definition \\
\midrule
\endhead
\bottomrule
\endfoot
$\phi_\varepsilon(x)$ & scalar map & Stabilized inverse-square-root map $(x+\varepsilon)^{-1/2}$. \\
$\zeta_{\mathrm{loc}}$ & scalar in $(0,1)$ & Local relative-error bound used only in Proposition~2. \\
$R_{t,j}^{(e)}$ & scalar & Taylor remainder in the inverse-square-root bias expansion. \\
$\widetilde R_{t,j}^{(e)}$ & scalar & Remainder normalized by $\phi_\varepsilon(v_{\mathrm{full},t,j}^{(e)})$. \\
$\xi_j$ & scalar & Intermediate point in Taylor's theorem between $\widehat v_j$ and $v_j$. \\
$\delta_{t,j}^{(e)}$ & scalar in $(-\infty,1)$ & Stabilized relative denominator error. Positive values mean support underestimation. \\
$w_{t,j}^{(e)}$ & nonnegative scalar & Active-coordinate weight induced by the few-shot numerator under the full-data denominator. \\
$\bar\delta_t^{(e)}$ & scalar & Weighted mean relative denominator error. \\
\end{longtable}

\begin{longtable}{@{}>{\raggedright\arraybackslash}p{0.22\textwidth}>{\raggedright\arraybackslash}p{0.23\textwidth}>{\raggedright\arraybackslash}p{0.51\textwidth}@{}}
\caption{Secondary notation for Appendix~L.3: Scale inflation and local step-length thresholds.}
\label{tab:not_L3}\\
\toprule
Term & Dimension & Definition \\
\midrule
\endfirsthead
\toprule
Term & Dimension & Definition \\
\midrule
\endhead
\bottomrule
\endfoot
$p_{ff,t}$ & unit vector in $\mathbb{R}^{d}$ & Unit direction of the Adam-like update. \\
$\omega_{ff,t}$ & positive scalar & Parameter-space length of the Adam-like step, $\eta\|u_{ff,t}\|_2$. \\
$p_t^{\mathrm{step}}$ & unit vector in $\mathbb{R}^{d}$ & Generic update direction used in the step-length theory. \\
$a_{\mathrm{full}}(p_t^{\mathrm{step}};t)$ & scalar in $[-1,1]$ & Alignment between $p_t^{\mathrm{step}}$ and the full-train gradient. \\
$\mathcal{S}_{t,\omega}(p_t^{\mathrm{step}})$ & line segment & Segment $\{\Theta_t-sp_t^{\mathrm{step}}:0\le s\le\omega\}$. \\
$\kappa_{\mathrm{full}}(s;p_t^{\mathrm{step}},t)$ & scalar & Directional curvature of $L_{\mathrm{full}}$ at $\Theta_t-sp_t^{\mathrm{step}}$. \\
$\lambda_{\mathrm{up}}$, $\lambda_{\mathrm{low}}$ & positive scalars & Upper and lower curvature bounds along the update segment. \\
$\psi(s)$ & scalar function & One-dimensional restriction $L_{\mathrm{full}}(\Theta_t-sp_t^{\mathrm{step}})$. \\
$\omega_{\mathrm{ref}}$, $\omega_{\mathrm{few}}$ & positive scalars & Reference and inflated step lengths in Proposition~4. \\
$\xi_t$ & scalar greater than $1$ & Scale inflation factor with $\omega_{\mathrm{few}}=\xi_t\omega_{\mathrm{ref}}$. \\
\end{longtable}

\begin{longtable}{@{}>{\raggedright\arraybackslash}p{0.22\textwidth}>{\raggedright\arraybackslash}p{0.23\textwidth}>{\raggedright\arraybackslash}p{0.51\textwidth}@{}}
\caption{Secondary notation for Appendix~L.4: Denominator repair and numerator staleness.}
\label{tab:not_L4}\\
\toprule
Term & Dimension & Definition \\
\midrule
\endfirsthead
\toprule
Term & Dimension & Definition \\
\midrule
\endhead
\bottomrule
\endfoot
$\Pi_{\mathrm{few},t}$ & $\mathbb{R}^{d\times d}$ & Diagonal preconditioner formed from $v_{\mathrm{few},t}$. \\
$\Pi_{\mathrm{full},t}$ & $\mathbb{R}^{d\times d}$ & Diagonal preconditioner formed from $v_{\mathrm{full},t}$. \\
$e_{\mathrm{den},t}$ & $\mathbb{R}^{d}$ & Adaptive-direction error caused by the denominator difference. \\
$e_{\mathrm{num},t}$ & $\mathbb{R}^{d}$ & Residual error caused by using the few-shot numerator after denominator repair. \\
$A_{\mathrm{full},t}$ & $\mathbb{R}^{d\times d}$ & Square root of the full-data diagonal preconditioner, $\Pi_{\mathrm{full},t}^{1/2}$. \\
$\rho_{\mathrm{num},t}$ & nonnegative scalar & Numerator-staleness ratio in the repaired-denominator metric. \\
$q_t^{\mathrm{sig}}$ & $\mathbb{R}^{d}$ & Proof shorthand $A_{\mathrm{full},t}m_{\mathrm{full},t}$. \\
$q_t^{\mathrm{err}}$ & $\mathbb{R}^{d}$ & Proof shorthand $A_{\mathrm{full},t}(m_{\mathrm{few},t}-m_{\mathrm{full},t})$. \\
$L_{\mathrm{sm},t}$ & positive scalar & Local smoothness constant for $L_{\mathrm{full}}$ along the repaired update segment. \\
\end{longtable}

\begin{longtable}{@{}>{\raggedright\arraybackslash}p{0.22\textwidth}>{\raggedright\arraybackslash}p{0.23\textwidth}>{\raggedright\arraybackslash}p{0.51\textwidth}@{}}
\caption{Secondary notation for Appendix~L.5: Blockwise scale mismatch after global rescaling.}
\label{tab:not_L5}\\
\toprule
Term & Dimension & Definition \\
\midrule
\endfirsthead
\toprule
Term & Dimension & Definition \\
\midrule
\endhead
\bottomrule
\endfoot
$\mathcal{B}_{\rm blk}$ & finite set & Set of trainable encoder--Transformer-block pairs. \\
$b$ & element of $\mathcal{B}_{\rm blk}$ & Block index, often $b=(e,j)$. \\
$d_b$ & positive integer & Number of trainable scalar parameters in block $b$. \\
$P_b$ & $\mathbb{R}^{d\times d}$ & Diagonal projector for block $b$. \\
$r_b(\Delta_t;t)$ & nonnegative scalar & Block relative update budget for update vector $\Delta_t$. \\
$\Delta_t^{\rm ad}$ & $\mathbb{R}^{d}$ & Raw Adam-like update. \\
$\Delta_t^{\rm ref}$ & $\mathbb{R}^{d}$ & Reference update, such as SGD or oracle-denominator update. \\
$\mathcal{A}_t$ & subset of $\mathcal{B}_{\rm blk}$ & Active reference blocks with nonzero reference update norm. \\
$a_b(t)$ & nonnegative scalar & Raw block scale ratio between Adam-like and reference updates. \\
$\gamma_t$ & positive scalar & Global coefficient that matches total update norm. \\
$\Delta_t^{\rm gm}$ & $\mathbb{R}^{d}$ & Globally matched update $\gamma_t\Delta_t^{\rm ad}$. \\
$\pi_b(t)$ & scalar in $[0,1]$ & Reference energy weight of block $b$. \\
$q_{b,t}$ & unit vector in $\mathbb{R}^{d}$ & Direction supported only on block $b$. \\
$c_{b,t}$ & positive scalar & Local full-loss descent coefficient along $q_{b,t}$. \\
$s_{b,t}^{\rm ref}$, $s_{b,t}^{\rm gm}$ & positive scalars & Reference and globally matched block step lengths. \\
$\lambda_{b,t}$, $\Lambda_{b,t}$ & positive scalars & Lower and upper curvature bounds along the block line. \\
$h_b(s)$ & scalar function & One-dimensional full-loss restriction along block direction $q_{b,t}$. \\
\end{longtable}

\begin{longtable}{@{}>{\raggedright\arraybackslash}p{0.22\textwidth}>{\raggedright\arraybackslash}p{0.23\textwidth}>{\raggedright\arraybackslash}p{0.51\textwidth}@{}}
\caption{Secondary notation for Appendix~M.1: One-dimensional basin interpolation.}
\label{tab:not_M1}\\
\toprule
Term & Dimension & Definition \\
\midrule
\endfirsthead
\toprule
Term & Dimension & Definition \\
\midrule
\endhead
\bottomrule
\endfoot
$\Theta(\alpha)$ & $\mathbb{R}^{d}$ & Init-to-final interpolation between $\Theta_0$ and $\Theta_T$. \\
$\varphi_S(\alpha)$ & scalar & Line loss for $S\in\{\mathrm{few},\mathrm{full},\mathrm{test}\}$. \\
$\alpha_S^\star$, $\varphi_S^\star$ & scalar & Line-loss minimizer and minimum value. \\
$\varphi_S^{\mathrm{lin}}$ & scalar function & Endpoint chord between $\varphi_S(0)$ and $\varphi_S(1)$. \\
$B_S$ & scalar & Interpolation barrier above the endpoint chord. \\
$\delta_{\mathrm w}$ & positive scalar & Tolerance used to define the near-minimum set. \\
$\mathcal{S}_S^{(\delta_{\mathrm w})}$ & subset of $[0,1]$ & Near-minimum set for line loss $\varphi_S$. \\
$W_S^{(\delta_{\mathrm w})}$ & scalar in $[0,1]$ & Span of the near-minimum set. \\
\end{longtable}

\begin{longtable}{@{}>{\raggedright\arraybackslash}p{0.22\textwidth}>{\raggedright\arraybackslash}p{0.23\textwidth}>{\raggedright\arraybackslash}p{0.51\textwidth}@{}}
\caption{Secondary notation for Appendix~M.2: Local safe step radius.}
\label{tab:not_M2}\\
\toprule
Term & Dimension & Definition \\
\midrule
\endfirsthead
\toprule
Term & Dimension & Definition \\
\midrule
\endhead
\bottomrule
\endfoot
$\Delta_t$ & $\mathbb{R}^{d}$ & SGD update $-\eta g_t$. \\
$a_t^{(p)}$ & $\mathbb{R}^{d}$ & Norm-matched probe displacement for $p\in\{\mathrm{sgd},\mathrm{rand},\mathrm{top\text{-}few}\}$. \\
$\alpha$ & scalar in $\mathcal{C}_{B}\subset[0,3]$ & Sampled nonnegative multiplier for a local line probe. \\
$\phi_{S,p}(\alpha;t)$ & scalar & Line loss $L_S(\Theta_t+\alpha a_t^{(p)})$. \\
$\alpha_{S,p}^\star(t)$ & scalar & Minimizer of $\phi_{S,p}$ over the sampled nonnegative grid. \\
$\rho_{S,p}(t)$ & nonnegative scalar & Largest sampled nonnegative multiplier whose loss does not exceed the unperturbed loss. It does not certify the intervening interval. \\
\end{longtable}

\begin{longtable}{@{}>{\raggedright\arraybackslash}p{0.22\textwidth}>{\raggedright\arraybackslash}p{0.23\textwidth}>{\raggedright\arraybackslash}p{0.51\textwidth}@{}}
\caption{Secondary notation for Appendix~M.3: Two-dimensional local surfaces around SGD checkpoints.}
\label{tab:not_M3}\\
\toprule
Term & Dimension & Definition \\
\midrule
\endfirsthead
\toprule
Term & Dimension & Definition \\
\midrule
\endhead
\bottomrule
\endfoot
$u_t^{\mathrm{sgd}}$ & unit vector in $\mathbb{R}^{d}$ & Normalized SGD update direction. \\
$p$ & direction family label & One of random, Adam-like, full-gradient, or top-Hessian directions. \\
$q_t^{(p)}$ & $\mathbb{R}^{d}$ & Raw comparison direction before orthogonalization. \\
$\xi_t^{\mathrm{rand}}$ & random vector & Raw random vector before orthogonalization in the 2D surface control. \\
$u_t^{\mathrm{top\text{-}few}}$ & unit vector & Largest-magnitude eigendirection of the support-loss Hessian. \\
$u_{\perp,t}^{(p)}$ & unit vector & Direction $q_t^{(p)}$ orthogonalized against $u_t^{\mathrm{sgd}}$. \\
$s_t$ & positive scalar & Scale used to set the 2D probe radius. \\
$\Psi_S^{(p)}(\alpha,\beta;t)$ & scalar & Two-dimensional loss surface around checkpoint $\Theta_t$. \\
$\Delta_{S,\beta}^{(p,+)}$, $\Delta_{S,\beta}^{(p,-)}$ & scalars & Loss changes along positive and negative off-SGD directions. \\
$R_S^{(p)}(t)$ & scalar & Ratio of off-SGD curvature to SGD-axis curvature. \\
\end{longtable}

\begin{longtable}{@{}>{\raggedright\arraybackslash}p{0.22\textwidth}>{\raggedright\arraybackslash}p{0.23\textwidth}>{\raggedright\arraybackslash}p{0.51\textwidth}@{}}
\caption{Secondary notation for Appendix~M.4: Directional curvature comparisons.}
\label{tab:not_M4}\\
\toprule
Term & Dimension & Definition \\
\midrule
\endfirsthead
\toprule
Term & Dimension & Definition \\
\midrule
\endhead
\bottomrule
\endfoot
$H_S(\Theta_t)$ & $\mathbb{R}^{d\times d}$ & Hessian of loss $L_S$ at checkpoint $\Theta_t$. \\
$\kappa_S(u;t)$ & scalar & Directional curvature $u^\top H_S(\Theta_t)u$. \\
$u_t^{\mathrm{sgd}}$ & unit vector & Actual normalized SGD update direction. \\
$u_t^{\mathrm{full}}$ & unit vector & Normalized full-train gradient direction. \\
$u_t^{\mathrm{top}}$ & unit vector & Largest-magnitude eigendirection of $H_{\mathrm{full}}(\Theta_t)$. \\
$u_t^{\mathrm{rand}}$ & unit vector & Random-direction control. \\
$\mathcal{I}_b$ & coordinate set & Coordinates belonging to block $b$ in curvature decomposition. \\
$\kappa_{\mathrm{full}}^{(b)}(u;t)$ & scalar & Block contribution to full-loss directional curvature. \\
\end{longtable}

\begin{longtable}{@{}>{\raggedright\arraybackslash}p{0.22\textwidth}>{\raggedright\arraybackslash}p{0.23\textwidth}>{\raggedright\arraybackslash}p{0.51\textwidth}@{}}
\caption{Secondary notation for Appendix~M.5: Blockwise update safety.}
\label{tab:not_M5}\\
\toprule
Term & Dimension & Definition \\
\midrule
\endfirsthead
\toprule
Term & Dimension & Definition \\
\midrule
\endhead
\bottomrule
\endfoot
$P_b$ & $\mathbb{R}^{d\times d}$ & Block projector used in the blockwise SGD step. Here $b=(e,j)$ indexes an encoder and its Transformer depth. \\
$a_t^{(b)}$ & $\mathbb{R}^{d}$ & Block-restricted SGD step $-\eta P_b g_t$. \\
$\psi_S^{(b)}(\alpha;t)$ & scalar & Blockwise line loss for objective $S$. \\
$\Delta_S^{(b)}(t;\alpha)$ & scalar & Blockwise loss change from scale $\alpha$. \\
$\alpha_{\mathrm{full}}^{\star,(b)}(t)$ & scalar in $[-2,2]$ & Full-loss preferred scale for block $b$. \\
$\Gamma^{(b)}(t;\alpha)$ & scalar & Support--full mismatch in loss change. \\
$\mathcal{T}$ & finite set of checkpoints & Checkpoint set used for averaging block diagnostics. \\
$\overline{\Delta}_{\mathrm{full}}^{(b)}(1)$ & scalar & Checkpoint-averaged full-loss change at actual block scale. \\
$\overline{\Gamma}^{(b)}(1)$ & scalar & Checkpoint-averaged support--full mismatch. \\
\end{longtable}

\begin{longtable}{@{}>{\raggedright\arraybackslash}p{0.22\textwidth}>{\raggedright\arraybackslash}p{0.23\textwidth}>{\raggedright\arraybackslash}p{0.51\textwidth}@{}}
\caption{Secondary notation for Appendix~M.6: Mode connectivity across SGD seeds.}
\label{tab:not_M6}\\
\toprule
Term & Dimension & Definition \\
\midrule
\endfirsthead
\toprule
Term & Dimension & Definition \\
\midrule
\endhead
\bottomrule
\endfoot
$s$, $s'$ & seed labels & Independent few-shot support-set seeds. \\
$\Theta_T^{(s)}$, $\Theta_T^{(s')}$ & $\mathbb{R}^{d}$ & Final SGD states from two seeds. \\
$\Theta^{(s,s')}(\alpha)$ & $\mathbb{R}^{d}$ & Seed-to-seed interpolation. \\
$L_{\mathrm{few}}^{(s,s')}$ & scalar-valued loss & Average of the two support losses from seeds $s$ and $s'$. \\
$L_S^{(s,s')}$ & scalar-valued loss & Seed-pair loss for $S\in\{\mathrm{few},\mathrm{full},\mathrm{test}\}$. \\
$\varphi_S^{(s,s')}(\alpha)$ & scalar & Line loss along the seed-to-seed interpolation. \\
$D_{\mathrm{rel}}^{(s,s')}$ & nonnegative scalar & Relative endpoint separation between the two final states. \\
\end{longtable}

\begin{longtable}{@{}>{\raggedright\arraybackslash}p{0.22\textwidth}>{\raggedright\arraybackslash}p{0.23\textwidth}>{\raggedright\arraybackslash}p{0.51\textwidth}@{}}
\caption{Secondary notation for Appendix~M.7: Support/full geometric alignment along SGD.}
\label{tab:not_M7}\\
\toprule
Term & Dimension & Definition \\
\midrule
\endfirsthead
\toprule
Term & Dimension & Definition \\
\midrule
\endhead
\bottomrule
\endfoot
$m_{\mathrm{few},t}^{(e)}$, $m_{\mathrm{full},t}^{(e)}$ & $\mathbb{R}^{d_e}$ & Support and full-train encoder-restricted gradients. \\
$A_g^{(e)}(t)$ & scalar in $[-1,1]$ & Cosine between support and full-train gradients. \\
$R_m^{(e)}(t)$ & nonnegative scalar & Gradient norm ratio. \\
$H_{\mathrm{few},t}^{(e)}$, $H_{\mathrm{full},t}^{(e)}$ & $\mathbb{R}^{d_e\times d_e}$ & Encoder-restricted support and full-train Hessians. \\
$u_{1,\mathrm{few},t}^{(e)}$, $u_{1,\mathrm{full},t}^{(e)}$ & unit vectors & Top Hessian eigendirections for support and full-train losses. \\
$\lambda_{1,\mathrm{few},t}^{(e)}$, $\lambda_{1,\mathrm{full},t}^{(e)}$ & scalars & Top Hessian eigenvalues. \\
$A_H^{(e)}(t)$ & scalar in $[0,1]$ & Absolute alignment of top support and full-train Hessian directions. \\
$R_\lambda^{(e)}(t)$ & scalar & Base-10 log-ratio of full and support Hessian eigenvalue magnitudes. \\
$\widehat u_{\mathrm{few},t}^{(e)}$, $\widehat u_{\mathrm{full},t}^{(e)}$ & unit vectors & Normalized support and full-train gradients. \\
$\kappa_S^{(e)}(u;t)$ & scalar & Encoder-restricted directional curvature. \\
$Q_\kappa^{(e)}(t)$ & nonnegative scalar & Curvature-ratio diagnostic with stabilizer $\varepsilon_\kappa$. \\
$\iota_e$ & embedding map & Injection from encoder-restricted coordinates into full parameter space. \\
$u_{\perp,t}^{(e)}$ & unit vector & Direction orthogonal to the support-gradient direction in encoder $e$. \\
$\mathcal{G}$ & grid & Two-dimensional probe grid for support/full surface correlation. \\
$C_{\mathrm{surf}}^{(e)}(t)$ & correlation coefficient & Correlation between support and full 2D surfaces on the grid. \\
\end{longtable}

\begin{longtable}{@{}>{\raggedright\arraybackslash}p{0.22\textwidth}>{\raggedright\arraybackslash}p{0.23\textwidth}>{\raggedright\arraybackslash}p{0.51\textwidth}@{}}
\caption{Secondary notation for Appendix~M.8: Low-drift CLIP feature region.}
\label{tab:not_M8}\\
\toprule
Term & Dimension & Definition \\
\midrule
\endfirsthead
\toprule
Term & Dimension & Definition \\
\midrule
\endhead
\bottomrule
\endfoot
$\Theta_t^{(\mathrm{SGD},G)}$ & $\mathbb{R}^{|G|}$ & Trainable parameters of group $G$ at checkpoint $t$. \\
$R_{\Theta}^{(\mathrm{SGD},G)}(t)$ & nonnegative scalar & Relative parameter displacement of group $G$. \\
$h_{\mathrm{img},t}^{(\mathrm{SGD})}(x)$ & $\mathbb{R}^{d_{\mathrm{emb}}}$ & Image feature used in SGD feature-drift diagnostics. \\
$h_{\mathrm{text},t}^{(\mathrm{SGD})}(c)$ & $\mathbb{R}^{d_{\mathrm{emb}}}$ & Class-text feature used in SGD feature-drift diagnostics. \\
$R_{\mathrm{img},S}^{(\mathrm{SGD})}(t)$ & nonnegative scalar & Mean angular image-feature drift on split $S$. \\
$R_{\mathrm{text}}^{(\mathrm{SGD})}(t)$ & nonnegative scalar & Mean angular text-feature drift over classes. \\
$\mu_t^{(\mathrm{SGD})}(x,y)$ & scalar & Ground-truth class margin. \\
$R_{\mathrm{mar},S}^{(\mathrm{SGD})}(t)$ & nonnegative scalar & Mean absolute margin drift. \\
$\Delta_{\mathrm{full\mbox{-}few}}(t)$ & scalar & Full-minus-support loss gap. \\
$\Delta_{\mathrm{test\mbox{-}few}}(t)$ & scalar & Test-minus-support loss gap. \\
\end{longtable}

\begin{longtable}{@{}>{\raggedright\arraybackslash}p{0.22\textwidth}>{\raggedright\arraybackslash}p{0.23\textwidth}>{\raggedright\arraybackslash}p{0.51\textwidth}@{}}
\caption{Secondary notation for Appendix~M.9: SGD stochasticity and flatness controls.}
\label{tab:not_M9}\\
\toprule
Term & Dimension & Definition \\
\midrule
\endfirsthead
\toprule
Term & Dimension & Definition \\
\midrule
\endhead
\bottomrule
\endfoot
$o$ & variant label & Full-batch, half-batch, quarter-batch, additive-noise, or momentum in I1--I3. In I4, half and quarter denote accumulation chunk sizes for full-support-sized bootstrap draws. \\
$\Theta_t^{(o)}$ & $\mathbb{R}^{d}$ & Parameter vector of variant $o$. \\
$g_{\mathcal B,t}^{(o)}$ & $\mathbb{R}^{d}$ & Micro-batch gradient for variant $o$. \\
$g_{\mathrm{few},t}^{(o)}$ & $\mathbb{R}^{d}$ & Full-support gradient for variant $o$. \\
$\mathfrak B_t$ & collection of micro-batches & Micro-batches used at checkpoint $t$ in the stochasticity diagnostic. \\
$B_t$ & positive integer & Number of micro-batches, $|\mathfrak B_t|$, at checkpoint $t$. \\
$\nu_{\mathrm{batch}}^{(o)}(t)$ & nonnegative scalar & Relative batch-noise level. \\
$\nu_{\mathrm{inj}}^{(o)}$ & nonnegative scalar & Relative injected-noise amplitude. \\
$\nu_{\mathrm{tot}}^{(o)}(t)$ & nonnegative scalar & Total stochasticity ratio. \\
$u_t^{(o,\mathrm{full})}$ & unit vector in $\mathbb{R}^{d}$ & Normalized full-train gradient. \\
$\delta_{\mathrm{fd}}$ & positive scalar & Finite-difference radius for directional flatness. \\
$\chi_{\mathrm{full}}^{(o)}(t)$ & scalar & Full-train directional flatness probe. \\
$\chi_{\mathrm{rand}}^{(o)}(t)$ & scalar & Random-direction flatness control. \\
$U_t^{(o)}$, $U_\star$ & nonnegative integers & Variant update count and full-batch baseline update count. \\
$\xi_{\mathrm{upd},t}^{(o)}$ & scalar in $[0,1]$ & Matched update-budget coordinate. \\
$E_t^{(o)}$, $E_\star$ & nonnegative integers & Processed support-example draws for variant $o$ and the full-batch baseline, including repeated draws in I4. \\
$\xi_{\mathrm{ex},t}^{(o)}$ & scalar in $[0,1]$ & Matched example-budget coordinate. \\
$\xi$ & scalar in $[0,1]$ & Common coordinate when both update and example budgets are matched. \\
\end{longtable}

\endgroup

\section{Few-shot Adaptation of CLIP}

CLIP~\cite{radford2021learning} uses an image encoder and a text encoder to represent images and class names in one shared embedding space. Let $e\in\{\mathrm{vision},\mathrm{text}\}$ index the two encoders. At adaptation step $t$, let
\begin{equation}
    \Theta_t=\big(\Theta_t^{(\mathrm{vision})},\Theta_t^{(\mathrm{text})}\big)\in\mathbb{R}^{d},
    \qquad
    \Theta_t^{(e)}\in\mathbb{R}^{d_e},
    \qquad
    d_{\mathrm{vision}}+d_{\mathrm{text}}=d .
    \label{eq:appendix_clip_parameters}
\end{equation}
Here, $d_e$ is the number of trainable scalar parameters in encoder $e$, and $d$ is the total number of trainable scalar parameters. The pretrained checkpoint is denoted by $\Theta_0$.

Let $d_{\mathrm{emb}}$ be the embedding dimension. For an image $x$, the normalized image feature at step $t$ is
\begin{equation}
    h_{\mathrm{img},t}(x)
    =\frac{F_{\Theta_t^{(\mathrm{vision})}}(x)}{
      \left\lVert F_{\Theta_t^{(\mathrm{vision})}}(x)\right\rVert_2}
    \in\mathbb{R}^{d_{\mathrm{emb}}} .
    \label{eq:appendix_clip_image_feature}
\end{equation}
For a text prompt $a$, the normalized text feature is
\begin{equation}
    h_{\mathrm{prompt},t}(a)
    =\frac{G_{\Theta_t^{(\mathrm{text})}}(a)}{
      \left\lVert G_{\Theta_t^{(\mathrm{text})}}(a)\right\rVert_2}
    \in\mathbb{R}^{d_{\mathrm{emb}}} .
    \label{eq:appendix_clip_text_feature}
\end{equation}
The maps $F_{\Theta_t^{(\mathrm{vision})}}$ and $G_{\Theta_t^{(\mathrm{text})}}$ are the unnormalized image and text encoders. Thus, $\lVert h_{\mathrm{img},t}(x)\rVert_2=1$ and $\lVert h_{\mathrm{prompt},t}(a)\rVert_2=1$.

\paragraph{Zero-shot CLIP classifier.}
The classification task has $C$ classes with labels $\{1,\ldots,C\}$. For each class $c$, let $\mathcal{A}_c=\{a_{j,c}\}_{j=1}^{P}$ be a set of $P$ text prompts, such as templates filled with the class name. The class-text prototype is
\begin{equation}
    \widetilde{h}_{\mathrm{text},t}(c)
    =\frac{1}{P}\sum_{j=1}^{P} h_{\mathrm{prompt},t}(a_{j,c})
    \in\mathbb{R}^{d_{\mathrm{emb}}},
    \qquad
    h_{\mathrm{text},t}(c)
    =\frac{\widetilde{h}_{\mathrm{text},t}(c)}{
      \left\lVert \widetilde{h}_{\mathrm{text},t}(c)\right\rVert_2}
    \in\mathbb{R}^{d_{\mathrm{emb}}} .
    \label{eq:appendix_clip_class_text_feature}
\end{equation}
For an image $x$, CLIP forms the logit vector $z_t(x)\in\mathbb{R}^{C}$ by
\begin{equation}
    z_t(x)_c
    =\frac{1}{\tau}\left\langle h_{\mathrm{img},t}(x),h_{\mathrm{text},t}(c)\right\rangle,
    \qquad c\in\{1,\ldots,C\},
    \label{eq:appendix_clip_logits}
\end{equation}
where $\tau>0$ is the CLIP temperature. The predicted class probabilities are
\begin{equation}
    p_t(c\mid x)
    =\frac{\exp\left(z_t(x)_c\right)}{
      \sum_{k=1}^{C}\exp\left(z_t(x)_k\right)},
    \qquad
    p_t(\cdot\mid x)\in\Delta^{C-1} .
    \label{eq:appendix_clip_probabilities}
\end{equation}
Here $\Delta^{C-1}$ is the probability simplex over $C$ classes. Zero-shot CLIP uses $\Theta_0$ in Eqs.~\eqref{eq:appendix_clip_image_feature}--\eqref{eq:appendix_clip_probabilities}. No task-specific parameters are updated in this case.

\paragraph{Few-shot objective.}
Let $\mathcal{D}_{\mathrm{full}}=\{(x_i,y_i)\}_{i=1}^{N}$ be the full training split, where $y_i\in\{1,\ldots,C\}$ and $N=|\mathcal{D}_{\mathrm{full}}|$. Let $\mathcal{D}_{\mathrm{few}}\subset \mathcal{D}_{\mathrm{full}}$ be the support set with $K$ labeled examples per class. Its size is $n=|\mathcal{D}_{\mathrm{few}}|=KC$. Let $\mathcal{D}_{\mathrm{test}}$ denote the test split. For a labeled example $(x,y)$, the CLIP cross-entropy loss is
\begin{equation}
    \ell\big((x,y);\Theta_t\big)
    =-\log p_t(y\mid x)
    \in\mathbb{R} .
    \label{eq:appendix_clip_example_loss}
\end{equation}
For a split $S\in\{\mathrm{few},\mathrm{full},\mathrm{test}\}$ with dataset $\mathcal{D}_S$, the empirical loss is
\begin{equation}
    L_S(\Theta_t)
    =\frac{1}{|\mathcal{D}_S|}\sum_{(x,y)\in \mathcal{D}_S}\ell\big((x,y);\Theta_t\big),
    \qquad
    L_S:\mathbb{R}^{d}\to\mathbb{R} .
    \label{eq:appendix_clip_empirical_loss}
\end{equation}
This notation gives the three losses used in the paper: $L_{\mathrm{few}}$, $L_{\mathrm{full}}$, and $L_{\mathrm{test}}$.

Few-shot adaptation starts from $\Theta_0$ and minimizes $L_{\mathrm{few}}$. Full fine-tuning means that both encoder parameter vectors, $\Theta_t^{(\mathrm{vision})}$ and $\Theta_t^{(\mathrm{text})}$, are trainable. At step $t$, the full-support few-shot gradient is
\begin{equation}
    g_t=\nabla_{\Theta}L_{\mathrm{few}}(\Theta_t)
    \in\mathbb{R}^{d} .
    \label{eq:appendix_clip_few_gradient}
\end{equation}
If a mini-batch $\mathcal{B}_t\subset \mathcal{D}_{\mathrm{few}}$ is used, $g_t$ is replaced by $\nabla_{\Theta}L_{\mathcal{B}_t}(\Theta_t)$. In the full-support protocol used in the main diagnostics, $\mathcal{B}_t=\mathcal{D}_{\mathrm{few}}$. Hence, one optimizer step corresponds to one pass over the support set. In the algorithms below, $\mathcal{L}$ denotes $L_{\mathrm{few}}$ or its mini-batch estimate.

\section{Algorithms}
\label{app:algorithms}

This appendix describes the optimizers used for full fine-tuning. Let \(\Theta_t\in\mathbb{R}^{d}\) denote the trainable CLIP parameters at iteration \(t\). This vector concatenates the encoder parameters \(\Theta_t^{(\mathrm{vision})}\) and \(\Theta_t^{(\mathrm{text})}\) defined in Eq.~\eqref{eq:appendix_clip_parameters}. Let \(\mathcal{B}_t\subseteq\mathcal{D}_{\mathrm{few}}\) be the support minibatch at iteration \(t\), and let \(\mathcal{L}_t:\mathbb{R}^{d}\to\mathbb{R}\) be the cross-entropy loss evaluated on \(\mathcal{B}_t\). The dense gradient is
\begin{equation}
    g_t = \nabla_{\Theta}\mathcal{L}_t(\Theta_t) \in \mathbb{R}^{d}.
\end{equation}
In the full-support experiments, \(\mathcal{B}_t=\mathcal{D}_{\mathrm{few}}\), and \(\mathcal{L}_t\) is the support loss \(L_{\mathrm{few}}\) used in the main text. The learning rate is \(\eta>0\). Elementwise multiplication and division are denoted by \(\odot\) and \(\oslash\). Square roots of vectors are also elementwise.

\paragraph{SGD.}
Stochastic gradient descent uses no optimizer state. It applies the same scalar learning rate to every coordinate of \(g_t\). Its update is
\begin{equation}
    \Theta_{t+1}=\Theta_t-\eta g_t .
\end{equation}

\begin{algorithm}
\caption{SGD}
\label{alg:sgd}
\small
\begin{algorithmic}[1]
\Require Initial parameters \(\Theta_0\in\mathbb{R}^{d}\), learning rate \(\eta>0\)
\For{\(t=0,1,2,\ldots\)}
    \State Compute \(g_t \leftarrow \nabla_{\Theta}\mathcal{L}_t(\Theta_t)\in\mathbb{R}^{d}\)
    \State \(\Theta_{t+1} \leftarrow \Theta_t-\eta g_t\)
\EndFor
\end{algorithmic}
\end{algorithm}

\paragraph{SMD (SGD with momentum).}
SGD with momentum, denoted by SMD in the paper, maintains a first-moment buffer \(m_t\in\mathbb{R}^{d}\). The coefficient \(\beta_1\in(0,1)\) controls the memory length. The method updates the parameters with the averaged direction \(m_{t+1}\).

\begin{algorithm}
\caption{SMD (SGD with momentum)}
\label{alg:smd}
\small
\begin{algorithmic}[1]
\Require Initial parameters \(\Theta_0\in\mathbb{R}^{d}\), learning rate \(\eta>0\), momentum \(\beta_1\in(0,1)\)
\State Initialize \(m_0\leftarrow 0\in\mathbb{R}^{d}\)
\For{\(t=0,1,2,\ldots\)}
    \State Compute \(g_t \leftarrow \nabla_{\Theta}\mathcal{L}_t(\Theta_t)\in\mathbb{R}^{d}\)
    \State \(m_{t+1} \leftarrow \beta_1m_t+(1-\beta_1)g_t\)
    \State \(\Theta_{t+1} \leftarrow \Theta_t-\eta m_{t+1}\)
\EndFor
\end{algorithmic}
\end{algorithm}

\paragraph{Adam.}
Adam maintains a first moment \(m_t\in\mathbb{R}^{d}\) and a diagonal second moment \(v_t\in\mathbb{R}^{d}_{\ge 0}\). The coefficients \(\beta_1,\beta_2\in(0,1)\) control the two exponential averages. The stabilizer \(\varepsilon>0\) prevents division by zero. Adam rescales each coordinate by the inverse square root of the bias-corrected second moment.

\begin{algorithm}
\caption{Adam}
\label{alg:adam}
\small
\begin{algorithmic}[1]
\Require Initial parameters \(\Theta_0\in\mathbb{R}^{d}\), learning rate \(\eta>0\), moments \(\beta_1,\beta_2\in(0,1)\), stabilizer \(\varepsilon>0\)
\State Initialize \(m_0\leftarrow 0\in\mathbb{R}^{d}\), \(v_0\leftarrow 0\in\mathbb{R}^{d}_{\ge 0}\)
\For{\(t=0,1,2,\ldots\)}
    \State Compute \(g_t \leftarrow \nabla_{\Theta}\mathcal{L}_t(\Theta_t)\in\mathbb{R}^{d}\)
    \State \(m_{t+1} \leftarrow \beta_1m_t+(1-\beta_1)g_t\)
    \State \(v_{t+1} \leftarrow \beta_2v_t+(1-\beta_2)(g_t\odot g_t)\)
    \State \(\widehat m_{t+1} \leftarrow m_{t+1}/(1-\beta_1^{t+1})\)
    \State \(\widehat v_{t+1} \leftarrow v_{t+1}/(1-\beta_2^{t+1})\)
    \State \(\Theta_{t+1} \leftarrow \Theta_t-\eta\,\widehat m_{t+1}\oslash(\sqrt{\widehat v_{t+1}}+\varepsilon\mathbf{1}_d)\)
\EndFor
\end{algorithmic}
\end{algorithm}

\begin{algorithm}[h]
\caption{Shampoo}
\label{alg:shampoo}
\small
\begin{algorithmic}[1]
\Require Initial tensors \(\{\Theta_0^{(J)}\}_{J=1}^{R}\), learning rate \(\eta>0\), moment coefficient \(\beta_2\in(0,1)\), stabilizer \(\varepsilon>0\), refresh frequency \(F\in\mathbb{N}\)
\ForAll{tensors \(J=1,\ldots,R\) and modes \(j=1,\ldots,p_J\)}
    \State Initialize \(A_{0,j}^{(J)}\leftarrow 0\in\mathbb{R}^{n_{J,j}\times n_{J,j}}\), \(P_{0,j}^{(J)}\leftarrow I_{n_{J,j}}\)
\EndFor
\For{\(t=0,1,2,\ldots\)}
    \State Compute tensor gradients \(G_t^{(J)}\leftarrow\nabla_{\Theta^{(J)}}\mathcal{L}_t(\Theta_t)\) for all \(J\)
    \ForAll{tensors \(J\) and modes \(j\)}
        \State \(A_{t+1,j}^{(J)} \leftarrow \beta_2A_{t,j}^{(J)}+(1-\beta_2)G_{t,(j)}^{(J)}(G_{t,(j)}^{(J)})^{\top}\)
        \If{\(t=0\) or \((t+1)\bmod F=0\)}
            \State \(P_{t+1,j}^{(J)} \leftarrow (A_{t+1,j}^{(J)}+\varepsilon I_{n_{J,j}})^{-1/(2p_J)}\)
        \Else
            \State \(P_{t+1,j}^{(J)} \leftarrow P_{t,j}^{(J)}\)
        \EndIf
    \EndFor
    \ForAll{tensors \(J=1,\ldots,R\)}
        \State \(\widetilde G_t^{(J)} \leftarrow G_t^{(J)}\times_1P_{t+1,1}^{(J)}\times_2\cdots\times_{p_J}P_{t+1,p_J}^{(J)}\)
        \State \(\Theta_{t+1}^{(J)} \leftarrow \Theta_t^{(J)}-\eta\widetilde G_t^{(J)}\)
    \EndFor
\EndFor
\end{algorithmic}
\end{algorithm}

\paragraph{Tensor notation for structured preconditioners.}
Shampoo and SOAP operate on parameter tensors rather than only on the flattened vector \(\Theta_t\). The trainable tensors are \(\{\Theta_t^{(J)}\}_{J=1}^{R}\), with \(\Theta_t=\operatorname{concat}_{J=1}^{R}\operatorname{vec}(\Theta_t^{(J)})\). Tensor \(J\) has order \(p_J\) and shape
\begin{equation}
    \Theta_t^{(J)}\in\mathbb{R}^{n_{J,1}\times\cdots\times n_{J,p_J}}.
\end{equation}
Let \(N_J=\prod_{j=1}^{p_J}n_{J,j}\), so \(\sum_{J=1}^{R}N_J=d\). The tensor gradient is
\begin{equation}
    G_t^{(J)} = \nabla_{\Theta^{(J)}}\mathcal{L}_t(\Theta_t)
    \in\mathbb{R}^{n_{J,1}\times\cdots\times n_{J,p_J}}.
\end{equation}
The vector \(g_t\) is obtained by concatenating \(\operatorname{vec}(G_t^{(J)})\) over tensors. For mode \(j\), let \(G_{t,(j)}^{(J)}\in\mathbb{R}^{n_{J,j}\times\prod_{k\ne j}n_{J,k}}\) be the mode-\(j\) unfolding of \(G_t^{(J)}\). Let \(\times_j\) denote the mode-\(j\) tensor-matrix product, and let \(I_n\in\mathbb{R}^{n\times n}\) be the identity matrix.

\paragraph{Shampoo.}
Shampoo~\citep{gupta2018shampoo} maintains one positive semidefinite factor for each tensor mode. For tensor \(J\) and mode \(j\), the factor is \(A_{t,j}^{(J)}\in\mathbb{R}^{n_{J,j}\times n_{J,j}}\). It estimates the second-order geometry of \(G_t^{(J)}\) along that mode. The preconditioner is formed from inverse matrix roots of these factors. For an order-\(p_J\) tensor, the exponent is \(-1/(2p_J)\), which gives the usual Shampoo scaling across modes.

\paragraph{SOAP.}
SOAP~\citep{vyas2025soap} uses Shampoo factors to define an orthogonal tensor basis. Its first moment is maintained in the parameter basis and projected when the adaptive update is formed. For tensor \(J\) and mode \(j\), \(Q_{t,j}^{(J)}\in\mathbb{R}^{n_{J,j}\times n_{J,j}}\) is the current orthogonal basis. The projected gradient is
\begin{equation}
    Z_t^{(J)} = G_t^{(J)}\times_1(Q_{t,1}^{(J)})^{\top}\times_2\cdots\times_{p_J}(Q_{t,p_J}^{(J)})^{\top}.
\end{equation}
The tensor \(M_t^{(J)}\in\mathbb{R}^{n_{J,1}\times\cdots\times n_{J,p_J}}\) stores the first moment in the parameter basis. The tensor \(V_t^{(J)}\in\mathbb{R}^{n_{J,1}\times\cdots\times n_{J,p_J}}_{\ge0}\) stores the diagonal second-moment estimate associated with the current SOAP basis. The preconditioned tensor is mapped back to the parameter basis before the update. This combined operation is abstracted as the structured map \(B_{\mathcal{O},\mathcal{D},t}^{(e)}\).

Algorithm~\ref{alg:soap} describes one parameter-update step. Before the first step, a gradient evaluation initializes the factors and their eigenbases without changing \(\Theta_0\), while \(M_0^{(J)}=V_0^{(J)}=0\). The index \(t\) counts subsequent parameter updates. Basis refreshes follow the parameter update. At a refresh, the current basis columns are sorted by their estimated eigenvalues and one power iteration followed by QR updates the basis. The same ordering is applied to the corresponding mode of \(V_{t+1}^{(J)}\). This reordering retains the diagonal approximation; it does not reconstruct the second moment under a general rotation.

\begin{algorithm}
\caption{SOAP: one parameter-update step}
\label{alg:soap}
\small
\begin{algorithmic}[1]
\Require Current tensors \(\{\Theta_t^{(J)},M_t^{(J)},V_t^{(J)}\}_{J=1}^{R}\), factors \(A_{t,j}^{(J)}\), bases \(Q_{t,j}^{(J)}\), learning rate \(\eta>0\), moments \(\beta_1,\beta_2\in(0,1)\), stabilizer \(\varepsilon>0\), refresh frequency \(F\in\mathbb{N}\)
\ForAll{tensors \(J=1,\ldots,R\)}
    \State \(G_t^{(J)}\leftarrow\nabla_{\Theta^{(J)}}\mathcal{L}_t(\Theta_t)\)
    \State \(Z_t^{(J)}\leftarrow G_t^{(J)}\times_1(Q_{t,1}^{(J)})^{\top}\times_2\cdots\times_{p_J}(Q_{t,p_J}^{(J)})^{\top}\)
    \State \(M_{t+1}^{(J)}\leftarrow\beta_1M_t^{(J)}+(1-\beta_1)G_t^{(J)}\)
    \State \(V_{t+1}^{(J)}\leftarrow\beta_2V_t^{(J)}+(1-\beta_2)(Z_t^{(J)}\odot Z_t^{(J)})\)
    \State \(\widehat M_{t+1}^{(J)}\leftarrow M_{t+1}^{(J)}/(1-\beta_1^{t+1})\), \(\widehat V_{t+1}^{(J)}\leftarrow V_{t+1}^{(J)}/(1-\beta_2^{t+1})\)
    \State \(Y_{t+1}^{(J)}\leftarrow\left(\widehat M_{t+1}^{(J)}\times_1(Q_{t,1}^{(J)})^{\top}\times_2\cdots\times_{p_J}(Q_{t,p_J}^{(J)})^{\top}\right)\oslash(\sqrt{\widehat V_{t+1}^{(J)}}+\varepsilon)\)
    \State \(\widetilde G_t^{(J)}\leftarrow Y_{t+1}^{(J)}\times_1Q_{t,1}^{(J)}\times_2\cdots\times_{p_J}Q_{t,p_J}^{(J)}\)
    \State \(\Theta_{t+1}^{(J)}\leftarrow\Theta_t^{(J)}-\eta\widetilde G_t^{(J)}\)
    \ForAll{modes \(j=1,\ldots,p_J\)}
        \State \(A_{t+1,j}^{(J)}\leftarrow\beta_2A_{t,j}^{(J)}+(1-\beta_2)G_{t,(j)}^{(J)}(G_{t,(j)}^{(J)})^{\top}\)
        \State \(Q_{t+1,j}^{(J)}\leftarrow Q_{t,j}^{(J)}\)
        \If{\((t+1)\bmod F=0\)}
            \State Columns of \(Q_{t+1,j}^{(J)}\) are sorted by descending entries of
            \Statex \hspace{\algorithmicindent}\(\operatorname{diag}\!\left((Q_{t,j}^{(J)})^{\top}A_{t+1,j}^{(J)}Q_{t,j}^{(J)}\right)\)
            \State The same permutation is applied to mode \(j\) of \(V_{t+1}^{(J)}\)
            \State \(Q_{t+1,j}^{(J)}\) is replaced by the orthogonal QR factor of \(A_{t+1,j}^{(J)}Q_{t+1,j}^{(J)}\)
        \EndIf
    \EndFor
\EndFor
\end{algorithmic}
\end{algorithm}

\section{Datasets}
\label{app:datasets}

\noindent\textbf{Benchmark.}
Our experiments use eleven datasets that are widely used for few-shot adaptation of CLIP. The benchmark covers general object, scene, fine-grained object, remote-sensing, food, species, texture, and action recognition. It includes ImageNet~\cite{imagenet} for large-scale object recognition and SUN397~\cite{sun397} for scene recognition. Fine-grained recognition is represented by FGVC-Aircraft~\cite{aircraft} and Stanford-Cars~\cite{stanford_cars}. EuroSAT~\cite{eurosat} evaluates land-use and land-cover recognition from satellite imagery. Food101~\cite{food101} evaluates food recognition. Oxford-Pets~\cite{oxford_pets} and Flowers102~\cite{flowers102} evaluate animal and plant species recognition. Caltech101~\cite{caltech101} provides diverse object categories. DTD~\cite{dtd} focuses on describable texture recognition. UCF101~\cite{ucf101} evaluates human action recognition.

\noindent\textbf{Few-shot splits.}
For each dataset, we use the standard class set and train/test split used in few-shot CLIP evaluation. For a shot level $K$, the support set $\mathcal{D}_{\mathrm{few}}$ contains $K$ labeled examples per class sampled from the training split. The main benchmark uses $K\in\{1,2,4,8,16\}$ and averages results over fixed random seeds. The model is optimized only on $\mathcal{D}_{\mathrm{few}}$. Test accuracy is measured on the held-out test split.

\noindent\textbf{Full-data reference for diagnostics.}
Several diagnostic experiments compare the support objective with a full-data reference. In these experiments, $\mathcal{D}_{\mathrm{full}}$ denotes the full training split of the same dataset. It is used only to measure losses, gradients, second moments, curvature, and probe quantities. It is not used to update the model during few-shot optimization. This convention separates the optimization signal available to the method from the reference signal used to analyze generalization.

\noindent\textbf{Diagnostic subset.}
Most mechanistic studies in Sections~\ref{sec:why_does_adam_fail} and~\ref{sec:why_does_sgd_succeed} use EuroSAT, Oxford-Pets, and DTD. This subset covers a distribution-shifted remote-sensing task, a fine-grained recognition task, and a texture-recognition task. It also includes both stable and overfitting-prone adaptation regimes. The full eleven-dataset benchmark is used for the main accuracy comparisons.

\section{Hyperparameters}
\label{app:hyperparameters}

This appendix reports the hyperparameters used in the main few-shot benchmark. All values are fixed across datasets, backbones, shot levels, and random seeds unless stated otherwise.

Let $K\in\mathbb{N}$ be the number of shots per class, and let $C\in\mathbb{N}$ be the number of classes. The support set is $\mathcal{D}_{\mathrm{few}}\subset \mathcal{D}_{\mathrm{full}}$ and has size $n=|\mathcal{D}_{\mathrm{few}}|=KC$. At iteration $t$, the trainable CLIP parameters are $\Theta_t\in\mathbb{R}^{d}$, where $d$ is the number of trainable scalar parameters. A support mini-batch is denoted by $\mathcal{B}_t\subseteq \mathcal{D}_{\mathrm{few}}$. Its nominal size is $B\in\mathbb{N}$. The mini-batch loss is $\mathcal{L}_t:\mathbb{R}^{d}\to\mathbb{R}$, and the gradient is
\begin{equation}
    g_t=\nabla_{\Theta}\mathcal{L}_t(\Theta_t)\in\mathbb{R}^{d} .
    \label{eq:appendix_hyperparameter_gradient}
\end{equation}
We use $\eta\in\mathbb{R}_{>0}$ for the learning rate, $T\in\mathbb{N}$ for the number of training epochs, $\beta_1,\beta_2\in(0,1)$ for moment coefficients, and $\varepsilon\in\mathbb{R}_{>0}$ for the numerical stabilizer. For low-rank methods, $r\in\mathbb{N}$ denotes the rank. For sparse methods, $\rho\in(0,1]$ denotes the fraction of trainable coordinates that remain active. For structured preconditioners, $F\in\mathbb{N}$ denotes the refresh frequency in optimizer iterations.

\paragraph{Image preprocessing.}
Table~\ref{tab:aug-pipeline} gives the shared training-time image preprocessing pipeline used in the main benchmark. The same normalization as CLIP is used after these augmentations.

\begin{table}[h]
\centering
\caption{Training-time image preprocessing used in the main benchmark.}
\label{tab:aug-pipeline}
\begin{tabular}{ll}
\toprule
Operation & Setting \\
\midrule
RandomResizedCrop & size $224$, scale $(0.08,1.0)$, bicubic interpolation \\
RandomHorizontalFlip & probability $0.5$ \\
\bottomrule
\end{tabular}
\end{table}

\paragraph{Prompt, adapter, and linear-probe baselines.}
Table~\ref{tab:hp-prompt-adapter} lists the hyperparameters for CoOp, KgCoOp, ProGrad, CLIP-Adapter, TIP-Adapter, and LP++. The symbol $M\in\mathbb{N}$ denotes the number of learnable context tokens per class. The scalar $\lambda_1\in\mathbb{R}_{>0}$ is the KgCoOp knowledge regularization weight. The scalar $\lambda_2\in\mathbb{R}_{>0}$ is the ProGrad projection weight. The scalar $\tau_{\mathrm{ProGrad}}\in\mathbb{R}_{>0}$ is the softmax temperature used by ProGrad. The CLIP-Adapter bottleneck reduction ratio is $r_{\mathrm{red}}\in\mathbb{N}$. The adapter mixing weights $\alpha_1,\alpha_2\in\mathbb{R}_{>0}$ are scalar coefficients.

\begin{table*}
\centering
\caption{Hyperparameters for prompt, adapter, and linear-probe baselines.}
\label{tab:hp-prompt-adapter}
\begin{tabularx}{\textwidth}{>{\raggedright\arraybackslash}p{0.18\textwidth} >{\raggedright\arraybackslash}p{0.20\textwidth} >{\raggedright\arraybackslash}p{0.25\textwidth} X}
\toprule
Method & Method-specific settings & Optimization settings & Trainable component \\
\midrule
CoOp & $M=16$ & $\eta=10^{-4}$, $T=200$, $B=64$, Adam & Class-specific context tokens \\
KgCoOp & $M=16$, $\lambda_1=8.0$ & $\eta=10^{-4}$, $T=200$, $B=64$, SGD & Class-specific context tokens with knowledge regularization \\
ProGrad & $M=16$, $\lambda_2=1.0$, $\tau_{\mathrm{ProGrad}}=1.0$ & $\eta=10^{-4}$, $T=200$, $B=64$, Adam & Class-specific context tokens with gradient projection \\
CLIP-Adapter & $\alpha_1=0.2$, $r_{\mathrm{red}}=4$ & $\eta=10^{-3}$, $T=30$, $B=64$, Adam & Adapter module \\
TIP-Adapter & $\alpha_1=0.27$, $\alpha_2=20.0$ & $\eta=5\times10^{-4}$, $T=10$, $B=64$, Adam & Cache-based adapter parameters \\
LP++ & -- & $\eta=10^{-3}$, $T=300$, $B=64$, SGD & Class-wise linear-probe coefficients \\
\bottomrule
\end{tabularx}
\end{table*}

\begin{table*}
\centering
\caption{Hyperparameters for low-rank and sparse baselines.}
\label{tab:hp-lowrank-sparse}
\begin{tabularx}{\textwidth}{>{\raggedright\arraybackslash}p{0.16\textwidth} >{\raggedright\arraybackslash}p{0.36\textwidth} X}
\toprule
Method & Method-specific settings & Optimization settings \\
\midrule
LoRA & rank $r=2$, scaling $\alpha_{\mathrm{LoRA}}=1$ & $\eta=10^{-4}$, $T=30$, $B=64$, Adam \\
PiSSA & rank $r=2$, scaling $\alpha_{\mathrm{PiSSA}}=1$ & $\eta=10^{-4}$, $T=30$, $B=64$, Adam \\
GaLoRE & rank $r=2$, refresh period $T_{\mathrm{update}}=10$ & $\eta=10^{-4}$, $T=30$, $B=64$, Adam \\
BitFit & bias parameters only & $\eta=10^{-4}$, $T=30$, $B=64$, Adam \\
LayerNorm & affine scale and shift parameters only & $\eta=10^{-4}$, $T=30$, $B=64$, Adam \\
SO & density $\rho=0.001$, support refresh period $T_{\mathrm{update}}=10$, $(\beta_1,\beta_2)=(0.9,0.999)$, $\varepsilon=10^{-8}$ & $\eta=10^{-4}$, $T=30$, $B=64$ \\
SAFT & density $\rho=0.001$, static top-$k$ support from $|g_0|$, $(\beta_1,\beta_2)=(0.9,0.999)$, $\varepsilon=10^{-8}$ & $\eta=10^{-4}$, $T=30$, $B=64$ \\
\bottomrule
\end{tabularx}
\end{table*}

\begin{table*}
\centering
\caption{Hyperparameters for full-fine-tuning optimizers.}
\label{tab:hp-full-fine-tuning}
\begin{tabularx}{\textwidth}{>{\raggedright\arraybackslash}p{0.16\textwidth} >{\raggedright\arraybackslash}p{0.38\textwidth} X}
\toprule
Optimizer & Optimizer-specific settings & Shared training settings \\
\midrule
SGD & no optimizer state & $\eta=10^{-4}$, $T=30$, $B=64$ \\
SMD & first-moment coefficient $\beta_1=0.9$ & $\eta=10^{-4}$, $T=30$, $B=64$ \\
Adam & $(\beta_1,\beta_2)=(0.9,0.999)$, $\varepsilon=10^{-8}$ & $\eta=10^{-4}$, $T=30$, $B=64$ \\
Shampoo & default Shampoo settings, preconditioning frequency $F=1$ & $\eta=10^{-4}$, $T=30$, $B=64$ \\
SOAP & default SOAP settings, preconditioning frequency $F=10$ & $\eta=10^{-4}$, $T=30$, $B=64$ \\
\bottomrule
\end{tabularx}
\end{table*}

\paragraph{Low-rank and sparse baselines.}
Table~\ref{tab:hp-lowrank-sparse} gives the hyperparameters for LoRA, PiSSA, GaLoRE, BitFit, LayerNorm tuning, SO, and SAFT. LoRA and PiSSA use rank $r=2$ and scaling coefficients $\alpha_{\mathrm{LoRA}}=1$ and $\alpha_{\mathrm{PiSSA}}=1$, respectively. GaLoRE uses rank $r=2$ and refresh period $T_{\mathrm{update}}=10$ iterations. SO and SAFT use density $\rho=0.001$, so $0.1\%$ of trainable coordinates are active. In SO, the active support is refreshed every $T_{\mathrm{update}}=10$ iterations. In SAFT, the active support is fixed from the initial gradient $g_0\in\mathbb{R}^{d}$.

For LoRA, PiSSA, GaLoRE, SO, and SAFT, the adaptation modules or sparse supports are applied to the Transformer blocks of both encoders. BitFit and LayerNorm tuning update only their corresponding parameter subsets.

\paragraph{Full-fine-tuning optimizers.}
Table~\ref{tab:hp-full-fine-tuning} lists the hyperparameters for these methods. All entries in this table update the trainable parameters in both CLIP encoders. In the default full-fine-tuning benchmark, all Transformer blocks are trainable. SGD has no optimizer state. SMD stores one first-moment vector $m_t\in\mathbb{R}^{d}$. Adam stores $m_t\in\mathbb{R}^{d}$ and a diagonal second-moment vector $v_t\in\mathbb{R}_{\ge0}^{d}$.

Shampoo and SOAP use the tensor notation from Appendix~\ref{app:algorithms}. For tensor $J\in\{1,\ldots,R\}$ with order $p_J\in\mathbb{N}$ and shape $n_{J,1}\times\cdots\times n_{J,p_J}$, Shampoo stores mode factors $A_{t,j}^{(J)}\in\mathbb{R}^{n_{J,j}\times n_{J,j}}$ for $j\in\{1,\ldots,p_J\}$. SOAP stores the same Shampoo factors and additionally stores $M_t^{(J)},V_t^{(J)}\in\mathbb{R}^{n_{J,1}\times\cdots\times n_{J,p_J}}$. The first moment is maintained in the parameter basis, while the diagonal second-moment estimate is associated with the SOAP basis. We use the default implementation settings for both methods. The preconditioning frequencies are reported in Table~\ref{tab:hp-full-fine-tuning}: $F=1$ for Shampoo and $F=10$ for SOAP.

\section{Training Time \& Memory Analysis}
\label{sec:memory_time}

\paragraph{Computational cost.}
Table~\ref{tab:time_sun397} reports wall-clock training time on SUN397 with CLIP (ViT-B/16) in the 16-shot setting. The fastest methods are those that avoid full-backbone adaptation. LP++, TIP-Adapter, and Adapter require 2, 4, and 9 minutes, respectively. Their cost is dominated by optimizing a small head or a cache over frozen CLIP features.

Methods that backpropagate through the CLIP backbone have a different profile. LayerNorm tuning, BitFit, LoRA, PiSSA, SAFT, dense SGD, SMD, and Adam all lie between 17 and 21 minutes. Thus, reducing the number of trainable parameters does not necessarily reduce wall-clock time once the full forward and backward passes through CLIP are required. The backbone computation is the dominant cost. SO is slower, at 26 minutes, because it maintains sparse moment states and refreshes the active support. GaLoRE is slower still, at 38 minutes, due to its low-rank gradient projections.

Prompt-learning methods have the largest wall-clock cost in this comparison. CoOp and KgCoOp each require 1 h 18 min, and ProGrad requires 2 h 08 min. This overhead mainly comes from their longer optimization schedules. In contrast, dense SGD takes 20 minutes, which is essentially the same runtime as Adam and SMD. Hence, the practical advantage of SGD in our main experiments is not obtained by using a more expensive optimizer.

\begin{table}[h]
  \centering
  \small
  \caption{Training time comparison on SUN397 for 16-shot classification with CLIP (ViT-B/16).}
  \label{tab:time_sun397}
  \begin{tabular}{l c}
    \toprule
    \textbf{Method} & \textbf{Training time} \\
    \midrule
    Adapter         & 9 min \\
    TIP-Adapter     & 4 min \\
    LP++            & 2 min \\
    \midrule
    CoOp            & 1 h 18 min \\
    KgCoOp          & 1 h 18 min \\
    ProGrad         & 2 h 08 min \\
    \midrule
    LoRA            & 19 min \\
    PiSSA           & 19 min \\
    GaLoRE          & 38 min \\
    \midrule
    BitFit          & 18 min \\
    LayerNorm       & 17 min \\
    SAFT            & 21 min \\
    SO              & 26 min \\
    \midrule
    Adam            & 21 min \\
    SMD             & 21 min \\
    SGD             & 20 min \\
    \bottomrule
  \end{tabular}
\end{table}

\paragraph{Optimization-memory model.}
\begin{table}[h]
\centering
\caption{Comparison of SGD, Adam, SAFT, SO, LoRA, PiSSA, and GaLoRE in memory requirements for a single fully-connected layer. The weight matrix is $W \in \mathbb{R}^{d_{\mathrm{out}} \times d_{\mathrm{in}}}$, with input dimension $d_{\mathrm{in}}$, output dimension $d_{\mathrm{out}}$, rank $r$, and density ratio $\rho$ (so the sparsity ratio is $1-\rho$). We count the number of stored scalars for the weight parameters, the gradients of the trainable parameters, and persistent optimizer states, excluding activations and temporary backpropagation buffers.}
\label{tab:memory_single_layer}
\resizebox{\textwidth}{!}{%
\begin{tabular}{lccccccc}
\toprule
& SGD & Adam  & SAFT  &  SO  & LoRA  & PiSSA & GaLoRE  \\
\midrule
Weight & $d_{\mathrm{out}}d_{\mathrm{in}}$  & $d_{\mathrm{out}}d_{\mathrm{in}}$  & $d_{\mathrm{out}}d_{\mathrm{in}}$  & $d_{\mathrm{out}}d_{\mathrm{in}}$ & $d_{\mathrm{out}}d_{\mathrm{in}} + d_{\mathrm{out}}r + d_{\mathrm{in}}r$  & $d_{\mathrm{out}}d_{\mathrm{in}} + d_{\mathrm{out}}r + d_{\mathrm{in}}r$ & $d_{\mathrm{out}}d_{\mathrm{in}}$ \\
Gradient & $d_{\mathrm{out}}d_{\mathrm{in}}$  & $d_{\mathrm{out}}d_{\mathrm{in}}$ & $d_{\mathrm{out}}d_{\mathrm{in}}\rho$ & $2d_{\mathrm{out}}d_{\mathrm{in}}\rho$ & $d_{\mathrm{out}}r + d_{\mathrm{in}}r$ & $d_{\mathrm{out}}r + d_{\mathrm{in}}r$ & $d_{\mathrm{out}}d_{\mathrm{in}}$ \\
Optimizer states & $0$  & $2d_{\mathrm{out}}d_{\mathrm{in}}$  & $2d_{\mathrm{out}}d_{\mathrm{in}}\rho$ & $3d_{\mathrm{out}}d_{\mathrm{in}}\rho$ & $2d_{\mathrm{out}}r + 2d_{\mathrm{in}}r$ & $2d_{\mathrm{out}}r + 2d_{\mathrm{in}}r$ & $r\min(d_{\mathrm{out}},d_{\mathrm{in}})+2r\max(d_{\mathrm{out}},d_{\mathrm{in}})$ \\
\bottomrule
\end{tabular}}
\end{table}

\begin{table*}[h]
\centering
\caption{Theoretical memory consumption for CLIP (ViT-B/16) when adapting all vision and text transformer blocks. We report the number of scalars (in millions) and the corresponding memory in MiB, assuming 32-bit floats. The table counts the stated weight matrices and their gradient and optimizer-state terms. Other parameters, activations, and temporary buffers are excluded. These values are not peak device memory.}
\scalebox{0.65}{\begin{tabular}{lccccc}
\toprule
Method & Weight (\#Vars, MiB) & Gradient (\#Vars, MiB) & Opt.\ States (\#Vars, MiB) & \#Trainable & Total Mem.\ (MiB) \\
\midrule
LoRA ($r=2$) & $123.175$M, $469.9$ & $0.492$M, $1.9$ & $0.983$M, $3.8$ & $0.492$M & $475.5$ \\
LoRA ($r=4$) & $123.666$M, $471.8$ & $0.983$M, $3.8$ & $1.966$M, $7.5$ & $0.983$M & $483.0$ \\
LoRA ($r=8$) & $124.649$M, $475.5$ & $1.966$M, $7.5$ & $3.932$M, $15.0$ & $1.966$M & $498.0$ \\
LoRA ($r=16$) & $126.616$M, $483.0$ & $3.932$M, $15.0$ & $7.864$M, $30.0$ & $3.932$M & $528.0$ \\
\midrule
PiSSA ($r=2$) & $123.175$M, $469.9$ & $0.492$M, $1.9$ & $0.983$M, $3.8$ & $0.492$M & $475.5$ \\
PiSSA ($r=4$) & $123.666$M, $471.8$ & $0.983$M, $3.8$ & $1.966$M, $7.5$ & $0.983$M & $483.0$ \\
PiSSA ($r=8$) & $124.649$M, $475.5$ & $1.966$M, $7.5$ & $3.932$M, $15.0$ & $1.966$M & $498.0$ \\
PiSSA ($r=16$) & $126.616$M, $483.0$ & $3.932$M, $15.0$ & $7.864$M, $30.0$ & $3.932$M & $528.0$ \\
\midrule
GaLoRE ($r=2$) & $122.683$M, $468.0$ & $122.683$M, $468.0$ & $0.860$M, $3.3$ & $122.683$M & $939.3$ \\
GaLoRE ($r=4$) & $122.683$M, $468.0$ & $122.683$M, $468.0$ & $1.720$M, $6.6$ & $122.683$M & $942.6$ \\
GaLoRE ($r=8$) & $122.683$M, $468.0$ & $122.683$M, $468.0$ & $3.441$M, $13.1$ & $122.683$M & $949.1$ \\
GaLoRE ($r=16$) & $122.683$M, $468.0$ & $122.683$M, $468.0$ & $6.881$M, $26.2$ & $122.683$M & $962.2$ \\
\midrule
SO ($\rho=0.01\%$) & $122.683$M, $468.0$ & $0.025$M, $0.1$ & $0.037$M, $0.1$ & $0.012$M & $468.2$ \\
SO ($\rho=0.1\%$) & $122.683$M, $468.0$ & $0.245$M, $0.9$ & $0.368$M, $1.4$ & $0.123$M & $470.3$ \\
SO ($\rho=1\%$) & $122.683$M, $468.0$ & $2.454$M, $9.4$ & $3.681$M, $14.0$ & $1.227$M & $491.4$ \\
SO ($\rho=10\%$) & $122.683$M, $468.0$ & $24.537$M, $93.6$ & $36.805$M, $140.4$ & $12.268$M & $702.0$ \\
\midrule
Adam (Full FT) & $122.683$M, $468.0$ & $122.683$M, $468.0$ & $245.367$M, $936.0$ & $122.683$M & $1872.0$ \\
\midrule
SGD (Full FT) & $122.683$M, $468.0$ & $122.683$M, $468.0$ & $0.000$M, $0.0$ & $122.683$M & $936.0$ \\
\bottomrule
\end{tabular}}
\label{tab:memory_clip}
\end{table*}

We next analyze the idealized optimization memory. We count only weights, gradients, and persistent optimizer states. Activations, temporary backpropagation buffers, parameters outside the counted matrices, sparse-index storage, and implementation-dependent workspaces are excluded. These terms can differ across methods. The estimates therefore do not measure peak device memory.

Table~\ref{tab:memory_single_layer} gives the scalar counts for a fully-connected layer with weight matrix $W\in\mathbb{R}^{d_{\mathrm{out}}\times d_{\mathrm{in}}}$. Here $d_{\mathrm{in}}$ is the input dimension, $d_{\mathrm{out}}$ is the output dimension, $r$ is the low-rank dimension, and $\rho\in(0,1]$ is the sparse-update density. SGD stores the dense weight and dense gradient, but it has no persistent optimizer state. Adam stores the same weight and gradient, and adds two dense buffers, $m_t$ and $v_t$. Thus, Adam stores $4d_{\mathrm{out}}d_{\mathrm{in}}$ counted scalars in total, while SGD stores $2d_{\mathrm{out}}d_{\mathrm{in}}$. Equivalently, Adam triples the non-weight optimization footprint because it stores the gradient plus two moment buffers. SMD is not shown in Table~\ref{tab:memory_single_layer}. By Algorithm~\ref{alg:smd}, it adds one dense first-moment buffer and would store $3d_{\mathrm{out}}d_{\mathrm{in}}$ counted scalars under the same accounting.

The idealized SAFT and SO counts retain support-restricted gradients and states. Dense temporary gradients used for support selection are excluded. SAFT uses $d_{\mathrm{out}}d_{\mathrm{in}}\rho$ gradient scalars and $2d_{\mathrm{out}}d_{\mathrm{in}}\rho$ moment scalars. SO uses $2d_{\mathrm{out}}d_{\mathrm{in}}\rho$ gradient-related scalars and $3d_{\mathrm{out}}d_{\mathrm{in}}\rho$ optimizer-state scalars because its dynamic support requires additional sparse bookkeeping. LoRA and PiSSA replace the dense trainable update with two rank-$r$ factors. Their trainable gradients and Adam states scale with $d_{\mathrm{out}}r+d_{\mathrm{in}}r$. GaLoRE keeps the dense gradient. Its retained state contains a projection with $r\min(d_{\mathrm{out}},d_{\mathrm{in}})$ scalars and two moment arrays with $r\max(d_{\mathrm{out}},d_{\mathrm{in}})$ scalars each. Therefore, its memory is close to dense SGD even at small $r$.

\paragraph{Memory analysis.}
Table~\ref{tab:memory_clip} applies the same accounting to CLIP (ViT-B/16) when all vision and text Transformer blocks are adapted. The counted matrices contain $122{,}683{,}392$ scalars and occupy $468.0$ MiB in 32-bit precision. The model counts 12 blocks of width 768 and 12 blocks of width 512. For block width $d_{\mathrm{width}}$, the matrix shapes are $3d_{\mathrm{width}}\times d_{\mathrm{width}}$, $d_{\mathrm{width}}\times d_{\mathrm{width}}$, $4d_{\mathrm{width}}\times d_{\mathrm{width}}$, and $d_{\mathrm{width}}\times4d_{\mathrm{width}}$. The low-rank rows apply the single-layer formulas to the same matrices. Dense SGD stores these weights and their dense gradients. Its counted optimization memory is 936.0 MiB. Adam stores the same weights and gradients, and also stores two dense moment vectors. Its counted memory is 1872.0 MiB. Adam therefore doubles the total counted memory relative to SGD and adds 936.0 MiB of persistent optimizer states. SMD would add one dense first-moment buffer and would require 1404.0 MiB under the same accounting.

LoRA and PiSSA are substantially smaller at low rank because only the rank-$r$ factors are trainable. At $r=2$, each requires 475.5 MiB. This is close to the frozen-backbone weight memory because the added gradients and moment states are small. GaLoRE remains near dense SGD memory because it keeps dense gradients. At $r=2$, it requires 939.3 MiB. SO ranges from 468.2 MiB at $\rho=0.01\%$ to 702.0 MiB at $\rho=10\%$. Its footprint is controlled by $\rho$, but it still uses sparse moment states.

These results separate runtime from optimizer-state memory. Parameter-efficient methods can reduce the persistent optimization footprint, but they do not always reduce training time when gradients still traverse the full CLIP backbone. Dense SGD has the same wall-clock cost as Adam in Table~\ref{tab:time_sun397}, but it removes Adam's two dense moment buffers. It is therefore the most memory-efficient dense full-fine-tuning optimizer among the dense optimizers considered here.

\section{SGD vs SMD vs Adam}
\label{sec:sgd_smd_adam}

In the dataset columns of the benchmark tables in Appendices~G, H, and J, values are means and standard deviations over three seeds. The Average column is the arithmetic mean of the eleven dataset means; its $\pm$ term is the arithmetic mean of the dataset standard deviations. This term is a descriptive variability summary, not the standard deviation of the aggregate mean.

\begin{table*}[h]
\caption{Few-shot classification with CLIP (ViT-B/32) comparing AdamW, Adam, SGD, and SMD. AdamW denotes Adam with weight-decay coefficients 0.1, 0.01, and 0.001; SMD corresponds to SGD with momentum. Top-1 accuracy (\%) averaged over 3 fixed random seeds.}
\centering
\resizebox{1.\textwidth}{!}{
\begin{tabular}{llcccccccccccc}
\toprule
Shots & Method & ImageNet & SUN & Aircraft & EuroSAT & Cars & Food & Pets & Flowers & Caltech & DTD & UCF & Average \\ \midrule
\multirow{1}{*}{0} & CLIP & 62.0 & 62.1 & 18.1 & 35.8 & 60.1 & 78.9 & 85.1 & 64.0 & 91.4 & 43.0 & 61.6 & 60.2 \\
\midrule
\multirow{6}{*}{1} & AdamW (0.1) & 7.1 $\pm$ 0.8 & 13.4 $\pm$ 0.6 & 1.0 $\pm$ 0.0 & 8.6 $\pm$ 0.8 & 0.5 $\pm$ 0.1 & 1.8 $\pm$ 1.0 & 6.6 $\pm$ 3.5 & 1.1 $\pm$ 0.4 & 22.6 $\pm$ 11.4 & 13.7 $\pm$ 3.6 & 1.8 $\pm$ 0.8 & 7.1 $\pm$ 2.1 \\
& AdamW (0.01) & 6.6 $\pm$ 1.1 & 12.9 $\pm$ 1.4 & 1.0 $\pm$ 0.1 & 9.8 $\pm$ 1.0 & 0.5 $\pm$ 0.1 & 2.2 $\pm$ 1.4 & 6.1 $\pm$ 3.7 & 1.2 $\pm$ 0.4 & 20.4 $\pm$ 10.8 & 14.1 $\pm$ 3.4 & 1.5 $\pm$ 0.4 & 6.9 $\pm$ 2.2 \\
& AdamW (0.001) & 6.8 $\pm$ 1.0 & 14.0 $\pm$ 0.3 & 1.0 $\pm$ 0.0 & 11.0 $\pm$ 3.0 & 0.5 $\pm$ 0.1 & 2.1 $\pm$ 1.1 & 5.4 $\pm$ 2.3 & 1.1 $\pm$ 0.4 & 20.2 $\pm$ 15.4 & 12.3 $\pm$ 2.6 & 2.2 $\pm$ 0.8 & 7.0 $\pm$ 2.5 \\
\cmidrule(lr){2-14}
& Adam & 12.4 $\pm$ 2.6 & 16.6 $\pm$ 1.7 & 1.1 $\pm$ 0.2 & 9.1 $\pm$ 2.0 & 0.5 $\pm$ 0.1 & 2.1 $\pm$ 0.8 & 5.7 $\pm$ 3.7 & 3.6 $\pm$ 2.5 & 30.9 $\pm$ 21.9 & 13.7 $\pm$ 2.6 & 1.4 $\pm$ 0.5 & 8.8 $\pm$ 3.5 \\
& SGD & 64.6 $\pm$ 0.3 & 67.4 $\pm$ 0.1 & 22.3 $\pm$ 1.2 & 65.1 $\pm$ 2.4 & 63.5 $\pm$ 0.7 & 80.5 $\pm$ 0.4 & 89.0 $\pm$ 0.1 & 77.4 $\pm$ 1.0 & 93.1 $\pm$ 0.5 & 50.0 $\pm$ 0.1 & 71.7 $\pm$ 0.3 & 67.7 $\pm$ 0.6 \\
& SMD & 64.9 $\pm$ 0.3 & 67.5 $\pm$ 0.1 & 21.8 $\pm$ 0.3 & 59.3 $\pm$ 5.4 & 59.3 $\pm$ 2.5 & 79.9 $\pm$ 0.2 & 86.4 $\pm$ 1.4 & 76.2 $\pm$ 0.5 & 93.1 $\pm$ 0.4 & 48.5 $\pm$ 0.7 & 71.5 $\pm$ 0.2 & 66.2 $\pm$ 1.1 \\
\midrule
\multirow{6}{*}{2} & AdamW (0.1) & 9.7 $\pm$ 0.9 & 20.5 $\pm$ 0.8 & 1.0 $\pm$ 0.0 & 11.5 $\pm$ 2.1 & 0.8 $\pm$ 0.2 & 1.3 $\pm$ 0.3 & 2.8 $\pm$ 0.1 & 7.1 $\pm$ 4.7 & 3.0 $\pm$ 3.0 & 15.3 $\pm$ 3.1 & 1.8 $\pm$ 0.6 & 6.8 $\pm$ 1.4 \\
& AdamW (0.01) & 9.8 $\pm$ 0.8 & 20.5 $\pm$ 1.3 & 1.0 $\pm$ 0.0 & 11.8 $\pm$ 2.4 & 0.5 $\pm$ 0.1 & 1.4 $\pm$ 0.3 & 2.8 $\pm$ 0.1 & 7.0 $\pm$ 5.7 & 3.5 $\pm$ 3.9 & 17.0 $\pm$ 2.2 & 1.9 $\pm$ 1.0 & 7.0 $\pm$ 1.6 \\
& AdamW (0.001) & 10.0 $\pm$ 0.8 & 20.6 $\pm$ 1.3 & 1.0 $\pm$ 0.0 & 11.7 $\pm$ 2.3 & 0.5 $\pm$ 0.1 & 1.2 $\pm$ 0.3 & 2.7 $\pm$ 0.0 & 6.5 $\pm$ 4.2 & 5.5 $\pm$ 6.6 & 16.4 $\pm$ 0.9 & 1.3 $\pm$ 0.2 & 7.0 $\pm$ 1.5 \\
\cmidrule(lr){2-14}
& Adam & 18.3 $\pm$ 0.6 & 22.5 $\pm$ 1.4 & 1.3 $\pm$ 0.2 & 11.8 $\pm$ 2.4 & 0.5 $\pm$ 0.1 & 2.1 $\pm$ 0.8 & 3.4 $\pm$ 0.7 & 8.0 $\pm$ 8.6 & 19.5 $\pm$ 25.7 & 15.0 $\pm$ 1.6 & 1.5 $\pm$ 1.0 & 9.4 $\pm$ 3.9 \\
& SGD & 65.2 $\pm$ 0.1 & 68.8 $\pm$ 0.3 & 22.6 $\pm$ 2.6 & 80.7 $\pm$ 3.6 & 61.6 $\pm$ 3.0 & 80.2 $\pm$ 0.3 & 89.6 $\pm$ 0.6 & 83.5 $\pm$ 0.7 & 94.2 $\pm$ 0.5 & 55.6 $\pm$ 1.1 & 74.5 $\pm$ 0.6 & 70.6 $\pm$ 1.2 \\
& SMD & 65.5 $\pm$ 0.1 & 69.0 $\pm$ 0.3 & 25.2 $\pm$ 0.6 & 77.1 $\pm$ 6.9 & 63.3 $\pm$ 0.9 & 80.1 $\pm$ 0.1 & 88.7 $\pm$ 1.2 & 82.7 $\pm$ 0.9 & 94.3 $\pm$ 0.4 & 55.7 $\pm$ 1.6 & 74.6 $\pm$ 0.2 & 70.5 $\pm$ 1.2 \\
\midrule
\multirow{6}{*}{4} & AdamW (0.1) & 13.1 $\pm$ 0.2 & 25.6 $\pm$ 0.9 & 1.6 $\pm$ 0.4 & 16.9 $\pm$ 1.8 & 0.8 $\pm$ 0.1 & 2.8 $\pm$ 2.2 & 4.1 $\pm$ 0.9 & 18.1 $\pm$ 17.6 & 30.9 $\pm$ 13.0 & 24.8 $\pm$ 1.9 & 8.8 $\pm$ 3.7 & 13.4 $\pm$ 3.9 \\
& AdamW (0.01) & 14.2 $\pm$ 0.2 & 25.3 $\pm$ 0.5 & 1.6 $\pm$ 0.4 & 17.0 $\pm$ 2.2 & 0.8 $\pm$ 0.2 & 2.5 $\pm$ 2.1 & 4.0 $\pm$ 0.8 & 16.4 $\pm$ 17.4 & 32.1 $\pm$ 12.8 & 25.2 $\pm$ 2.4 & 7.5 $\pm$ 4.2 & 13.3 $\pm$ 3.9 \\
& AdamW (0.001) & 13.9 $\pm$ 0.3 & 25.1 $\pm$ 0.8 & 1.5 $\pm$ 0.2 & 17.1 $\pm$ 1.3 & 0.9 $\pm$ 0.2 & 2.4 $\pm$ 1.7 & 4.1 $\pm$ 0.9 & 17.4 $\pm$ 15.8 & 32.3 $\pm$ 11.7 & 24.1 $\pm$ 1.0 & 7.2 $\pm$ 5.0 & 13.3 $\pm$ 3.5 \\
\cmidrule(lr){2-14}
& Adam & 22.4 $\pm$ 0.3 & 31.1 $\pm$ 0.7 & 1.0 $\pm$ 0.0 & 18.9 $\pm$ 1.0 & 0.7 $\pm$ 0.1 & 1.6 $\pm$ 0.7 & 3.9 $\pm$ 0.7 & 10.9 $\pm$ 7.7 & 5.8 $\pm$ 3.8 & 23.1 $\pm$ 1.0 & 6.4 $\pm$ 5.4 & 11.4 $\pm$ 1.9 \\
& SGD & 66.3 $\pm$ 0.1 & 71.1 $\pm$ 0.2 & 25.3 $\pm$ 3.3 & 84.5 $\pm$ 0.7 & 67.4 $\pm$ 0.8 & 80.5 $\pm$ 0.3 & 90.5 $\pm$ 0.1 & 87.9 $\pm$ 1.0 & 94.7 $\pm$ 0.2 & 60.0 $\pm$ 0.3 & 77.5 $\pm$ 0.7 & 73.2 $\pm$ 0.7 \\
& SMD & 66.5 $\pm$ 0.2 & 71.1 $\pm$ 0.2 & 26.3 $\pm$ 1.1 & 81.7 $\pm$ 1.7 & 69.2 $\pm$ 0.2 & 80.5 $\pm$ 0.2 & 90.2 $\pm$ 0.2 & 88.6 $\pm$ 1.0 & 95.0 $\pm$ 0.2 & 60.3 $\pm$ 0.6 & 77.8 $\pm$ 0.5 & 73.4 $\pm$ 0.5 \\
\midrule
\multirow{6}{*}{8} & AdamW (0.1) & 17.4 $\pm$ 0.4 & 30.3 $\pm$ 0.7 & 1.8 $\pm$ 0.1 & 18.9 $\pm$ 3.3 & 2.7 $\pm$ 1.9 & 7.9 $\pm$ 4.0 & 22.7 $\pm$ 17.8 & 54.8 $\pm$ 4.8 & 67.9 $\pm$ 2.5 & 31.1 $\pm$ 2.3 & 13.6 $\pm$ 9.0 & 24.5 $\pm$ 4.2 \\
& AdamW (0.01) & 17.9 $\pm$ 0.2 & 29.5 $\pm$ 0.2 & 2.2 $\pm$ 0.3 & 19.3 $\pm$ 3.8 & 2.8 $\pm$ 1.7 & 7.3 $\pm$ 2.8 & 22.0 $\pm$ 16.0 & 52.9 $\pm$ 7.5 & 69.9 $\pm$ 1.7 & 30.2 $\pm$ 1.7 & 14.5 $\pm$ 2.8 & 24.4 $\pm$ 3.5 \\
& AdamW (0.001) & 17.7 $\pm$ 0.5 & 31.2 $\pm$ 0.4 & 1.9 $\pm$ 0.1 & 18.9 $\pm$ 3.1 & 3.5 $\pm$ 1.7 & 10.3 $\pm$ 3.7 & 20.6 $\pm$ 14.4 & 50.9 $\pm$ 8.5 & 68.3 $\pm$ 2.9 & 30.6 $\pm$ 0.5 & 18.9 $\pm$ 4.6 & 24.8 $\pm$ 3.7 \\
\cmidrule(lr){2-14}
& Adam & 27.3 $\pm$ 0.6 & 37.2 $\pm$ 0.2 & 1.6 $\pm$ 0.1 & 21.0 $\pm$ 8.8 & 1.2 $\pm$ 0.1 & 5.7 $\pm$ 4.4 & 10.2 $\pm$ 4.9 & 44.9 $\pm$ 4.0 & 74.7 $\pm$ 3.1 & 28.9 $\pm$ 3.2 & 19.5 $\pm$ 7.2 & 24.8 $\pm$ 3.3 \\
& SGD & 67.3 $\pm$ 0.1 & 72.9 $\pm$ 0.1 & 29.1 $\pm$ 2.4 & 86.2 $\pm$ 3.8 & 74.4 $\pm$ 0.4 & 80.8 $\pm$ 0.2 & 91.1 $\pm$ 0.2 & 91.9 $\pm$ 0.6 & 95.3 $\pm$ 0.3 & 65.1 $\pm$ 0.3 & 79.4 $\pm$ 0.3 & 75.8 $\pm$ 0.8 \\
& SMD & 67.5 $\pm$ 0.1 & 72.8 $\pm$ 0.1 & 33.5 $\pm$ 0.1 & 88.6 $\pm$ 1.1 & 75.4 $\pm$ 0.2 & 80.9 $\pm$ 0.3 & 91.1 $\pm$ 0.1 & 91.7 $\pm$ 0.6 & 95.2 $\pm$ 0.2 & 63.9 $\pm$ 1.4 & 79.9 $\pm$ 0.4 & 76.4 $\pm$ 0.4 \\
\midrule
\multirow{6}{*}{16} & AdamW (0.1) & 22.8 $\pm$ 0.4 & 35.9 $\pm$ 0.6 & 2.8 $\pm$ 0.4 & 38.3 $\pm$ 6.5 & 4.5 $\pm$ 2.3 & 8.9 $\pm$ 5.4 & 22.5 $\pm$ 17.3 & 63.5 $\pm$ 8.9 & 75.5 $\pm$ 1.8 & 39.9 $\pm$ 2.4 & 23.9 $\pm$ 8.8 & 30.8 $\pm$ 5.0 \\
& AdamW (0.01) & 23.2 $\pm$ 0.4 & 36.7 $\pm$ 0.4 & 2.9 $\pm$ 0.4 & 37.1 $\pm$ 1.2 & 5.5 $\pm$ 3.0 & 6.2 $\pm$ 5.5 & 20.5 $\pm$ 14.3 & 65.0 $\pm$ 9.2 & 73.4 $\pm$ 2.0 & 40.8 $\pm$ 0.4 & 33.5 $\pm$ 2.0 & 31.3 $\pm$ 3.5 \\
& AdamW (0.001) & 22.7 $\pm$ 0.3 & 37.0 $\pm$ 0.9 & 3.6 $\pm$ 0.3 & 40.4 $\pm$ 2.6 & 4.8 $\pm$ 2.7 & 8.7 $\pm$ 3.3 & 25.5 $\pm$ 16.8 & 63.0 $\pm$ 9.0 & 75.5 $\pm$ 1.2 & 40.5 $\pm$ 0.9 & 33.7 $\pm$ 0.9 & 32.3 $\pm$ 3.5 \\
\cmidrule(lr){2-14}
& Adam & 31.9 $\pm$ 0.6 & 42.3 $\pm$ 0.9 & 2.1 $\pm$ 0.1 & 31.7 $\pm$ 1.6 & 5.6 $\pm$ 2.2 & 18.2 $\pm$ 10.9 & 25.2 $\pm$ 7.5 & 76.9 $\pm$ 2.6 & 80.1 $\pm$ 0.7 & 39.3 $\pm$ 2.2 & 31.8 $\pm$ 4.2 & 35.0 $\pm$ 3.0 \\
& SGD & 68.8 $\pm$ 0.1 & 74.4 $\pm$ 0.1 & 39.5 $\pm$ 0.3 & 92.2 $\pm$ 0.6 & 78.8 $\pm$ 0.7 & 81.8 $\pm$ 0.0 & 91.9 $\pm$ 0.2 & 95.1 $\pm$ 0.5 & 96.2 $\pm$ 0.1 & 68.7 $\pm$ 0.2 & 82.7 $\pm$ 0.1 & 79.1 $\pm$ 0.3 \\
& SMD & 68.9 $\pm$ 0.1 & 74.4 $\pm$ 0.0 & 40.1 $\pm$ 0.9 & 91.5 $\pm$ 0.8 & 80.4 $\pm$ 0.5 & 81.8 $\pm$ 0.1 & 91.9 $\pm$ 0.1 & 94.9 $\pm$ 0.6 & 96.0 $\pm$ 0.1 & 68.6 $\pm$ 0.4 & 83.1 $\pm$ 0.1 & 79.2 $\pm$ 0.3 \\
\bottomrule
\end{tabular}}

\label{tab:ablation_adamw_ViTB32_with_aug}
\end{table*}
 
\begin{table*}[h]
\caption{Few-shot classification with CLIP (ViT-B/16) comparing AdamW, Adam, SGD, and SMD. AdamW denotes Adam with weight-decay coefficients 0.1, 0.01, and 0.001; SMD corresponds to SGD with momentum. Top-1 accuracy (\%) averaged over 3 fixed random seeds.}
\centering
\resizebox{1.\textwidth}{!}{
\begin{tabular}{llcccccccccccc}
\toprule
Shots & Method & ImageNet & SUN & Aircraft & EuroSAT & Cars & Food & Pets & Flowers & Caltech & DTD & UCF & Average \\ \midrule
\multirow{1}{*}{0} & CLIP & 66.7 & 62.6 & 23.9 & 42.2 & 65.5 & 85.2 & 88.3 & 67.4 & 93.3 & 44.4 & 65.2 & 64.1 \\
\midrule
\multirow{6}{*}{1} & AdamW (0.1) & 8.7 $\pm$ 0.6 & 5.0 $\pm$ 2.1 & 1.3 $\pm$ 0.4 & 15.3 $\pm$ 1.5 & 0.5 $\pm$ 0.1 & 1.8 $\pm$ 0.6 & 4.9 $\pm$ 1.6 & 7.2 $\pm$ 3.7 & 19.7 $\pm$ 11.0 & 13.1 $\pm$ 3.2 & 1.5 $\pm$ 0.7 & 7.2 $\pm$ 2.3 \\
& AdamW (0.01) & 8.7 $\pm$ 0.5 & 3.8 $\pm$ 2.8 & 1.0 $\pm$ 0.0 & 15.3 $\pm$ 1.4 & 0.5 $\pm$ 0.1 & 1.3 $\pm$ 0.2 & 3.9 $\pm$ 1.3 & 7.8 $\pm$ 2.9 & 22.7 $\pm$ 14.0 & 13.1 $\pm$ 2.7 & 1.6 $\pm$ 0.7 & 7.3 $\pm$ 2.4 \\
& AdamW (0.001) & 8.6 $\pm$ 0.4 & 6.3 $\pm$ 0.7 & 1.1 $\pm$ 0.5 & 16.3 $\pm$ 2.2 & 0.4 $\pm$ 0.1 & 1.4 $\pm$ 0.6 & 3.9 $\pm$ 1.0 & 7.2 $\pm$ 3.7 & 24.8 $\pm$ 12.9 & 13.3 $\pm$ 4.3 & 2.0 $\pm$ 1.1 & 7.8 $\pm$ 2.5 \\
\cmidrule(lr){2-14}
& Adam & 18.3 $\pm$ 0.9 & 19.5 $\pm$ 2.2 & 1.1 $\pm$ 0.1 & 11.7 $\pm$ 3.1 & 0.7 $\pm$ 0.1 & 1.4 $\pm$ 0.1 & 3.3 $\pm$ 0.6 & 1.4 $\pm$ 0.7 & 40.3 $\pm$ 6.1 & 13.5 $\pm$ 1.8 & 2.3 $\pm$ 1.8 & 10.3 $\pm$ 1.6 \\
& SGD & 70.1 $\pm$ 0.2 & 70.3 $\pm$ 0.2 & 30.5 $\pm$ 0.9 & 73.8 $\pm$ 2.7 & 70.8 $\pm$ 0.6 & 85.8 $\pm$ 0.2 & 92.7 $\pm$ 0.2 & 83.6 $\pm$ 0.6 & 94.0 $\pm$ 0.2 & 53.7 $\pm$ 0.8 & 76.1 $\pm$ 0.5 & 72.8 $\pm$ 0.6 \\
& SMD & 70.2 $\pm$ 0.2 & 70.5 $\pm$ 0.2 & 30.1 $\pm$ 0.3 & 70.4 $\pm$ 7.5 & 66.9 $\pm$ 2.2 & 85.3 $\pm$ 0.1 & 90.5 $\pm$ 0.1 & 82.4 $\pm$ 0.3 & 94.0 $\pm$ 0.0 & 52.6 $\pm$ 1.1 & 76.3 $\pm$ 0.7 & 71.7 $\pm$ 1.1 \\
\midrule
\multirow{6}{*}{2} & AdamW (0.1) & 8.9 $\pm$ 5.2 & 22.7 $\pm$ 0.9 & 1.1 $\pm$ 0.2 & 20.3 $\pm$ 4.1 & 0.6 $\pm$ 0.2 & 1.0 $\pm$ 0.0 & 2.7 $\pm$ 0.0 & 6.0 $\pm$ 4.7 & 5.7 $\pm$ 6.0 & 15.9 $\pm$ 6.4 & 6.6 $\pm$ 7.9 & 8.3 $\pm$ 3.2 \\
& AdamW (0.01) & 9.0 $\pm$ 4.4 & 23.0 $\pm$ 1.1 & 1.1 $\pm$ 0.1 & 18.4 $\pm$ 4.2 & 0.6 $\pm$ 0.2 & 1.2 $\pm$ 0.2 & 2.7 $\pm$ 0.0 & 5.6 $\pm$ 3.5 & 9.7 $\pm$ 12.5 & 15.7 $\pm$ 8.1 & 7.1 $\pm$ 8.5 & 8.6 $\pm$ 3.9 \\
& AdamW (0.001) & 8.4 $\pm$ 5.7 & 22.1 $\pm$ 1.8 & 1.2 $\pm$ 0.2 & 18.5 $\pm$ 3.7 & 0.5 $\pm$ 0.1 & 1.3 $\pm$ 0.4 & 2.7 $\pm$ 0.0 & 4.7 $\pm$ 3.6 & 4.1 $\pm$ 4.4 & 17.1 $\pm$ 9.3 & 6.8 $\pm$ 8.2 & 7.9 $\pm$ 3.4 \\
\cmidrule(lr){2-14}
& Adam & 23.0 $\pm$ 1.0 & 23.4 $\pm$ 2.7 & 1.0 $\pm$ 0.2 & 21.6 $\pm$ 5.3 & 0.6 $\pm$ 0.1 & 1.7 $\pm$ 0.4 & 3.3 $\pm$ 0.2 & 13.3 $\pm$ 8.7 & 12.3 $\pm$ 15.8 & 15.0 $\pm$ 6.8 & 6.4 $\pm$ 7.0 & 11.0 $\pm$ 4.4 \\
& SGD & 70.5 $\pm$ 0.2 & 71.8 $\pm$ 0.4 & 33.5 $\pm$ 1.4 & 82.4 $\pm$ 2.3 & 71.4 $\pm$ 0.6 & 86.0 $\pm$ 0.2 & 92.5 $\pm$ 0.6 & 89.7 $\pm$ 0.3 & 94.4 $\pm$ 0.2 & 59.4 $\pm$ 1.0 & 79.0 $\pm$ 1.0 & 75.5 $\pm$ 0.7 \\
& SMD & 70.7 $\pm$ 0.1 & 71.9 $\pm$ 0.3 & 33.8 $\pm$ 2.0 & 81.5 $\pm$ 3.6 & 73.6 $\pm$ 0.9 & 85.6 $\pm$ 0.3 & 91.4 $\pm$ 0.4 & 89.3 $\pm$ 0.6 & 94.4 $\pm$ 0.5 & 60.1 $\pm$ 1.1 & 79.3 $\pm$ 0.8 & 75.6 $\pm$ 1.0 \\
\midrule
\multirow{6}{*}{4} & AdamW (0.1) & 16.9 $\pm$ 0.9 & 27.7 $\pm$ 1.8 & 1.1 $\pm$ 0.2 & 16.3 $\pm$ 4.4 & 0.7 $\pm$ 0.2 & 2.7 $\pm$ 2.1 & 3.0 $\pm$ 0.5 & 12.0 $\pm$ 2.1 & 40.0 $\pm$ 27.9 & 18.1 $\pm$ 8.8 & 1.5 $\pm$ 0.7 & 12.7 $\pm$ 4.5 \\
& AdamW (0.01) & 17.0 $\pm$ 0.4 & 27.7 $\pm$ 1.0 & 1.3 $\pm$ 0.3 & 16.1 $\pm$ 4.1 & 0.6 $\pm$ 0.1 & 2.3 $\pm$ 1.5 & 2.5 $\pm$ 0.5 & 11.2 $\pm$ 1.4 & 43.4 $\pm$ 30.0 & 18.7 $\pm$ 9.5 & 1.8 $\pm$ 1.1 & 13.0 $\pm$ 4.5 \\
& AdamW (0.001) & 17.1 $\pm$ 0.2 & 28.3 $\pm$ 1.1 & 1.6 $\pm$ 0.3 & 17.3 $\pm$ 4.4 & 0.6 $\pm$ 0.1 & 2.7 $\pm$ 1.7 & 2.5 $\pm$ 0.5 & 11.4 $\pm$ 2.4 & 38.7 $\pm$ 26.8 & 19.4 $\pm$ 9.5 & 2.6 $\pm$ 2.3 & 12.9 $\pm$ 4.5 \\
\cmidrule(lr){2-14}
& Adam & 28.8 $\pm$ 0.7 & 31.0 $\pm$ 6.0 & 1.1 $\pm$ 0.1 & 15.8 $\pm$ 3.5 & 0.7 $\pm$ 0.2 & 0.9 $\pm$ 0.1 & 2.7 $\pm$ 0.1 & 8.3 $\pm$ 2.8 & 18.5 $\pm$ 16.5 & 16.0 $\pm$ 7.7 & 2.1 $\pm$ 1.6 & 11.4 $\pm$ 3.6 \\
& SGD & 71.4 $\pm$ 0.2 & 73.7 $\pm$ 0.2 & 35.5 $\pm$ 2.0 & 84.4 $\pm$ 1.5 & 77.2 $\pm$ 0.1 & 86.1 $\pm$ 0.0 & 93.0 $\pm$ 0.4 & 93.3 $\pm$ 0.5 & 95.2 $\pm$ 0.4 & 64.4 $\pm$ 0.9 & 81.8 $\pm$ 1.0 & 77.8 $\pm$ 0.7 \\
& SMD & 71.7 $\pm$ 0.1 & 73.8 $\pm$ 0.3 & 37.7 $\pm$ 0.5 & 85.2 $\pm$ 1.2 & 78.2 $\pm$ 0.3 & 86.2 $\pm$ 0.2 & 92.8 $\pm$ 0.3 & 92.9 $\pm$ 0.6 & 95.2 $\pm$ 0.4 & 64.8 $\pm$ 0.6 & 82.0 $\pm$ 0.5 & 78.2 $\pm$ 0.5 \\
\midrule
\multirow{6}{*}{8} & AdamW (0.1) & 22.3 $\pm$ 0.3 & 36.7 $\pm$ 0.2 & 2.3 $\pm$ 0.2 & 24.5 $\pm$ 11.0 & 1.2 $\pm$ 0.3 & 2.7 $\pm$ 2.2 & 4.4 $\pm$ 1.2 & 30.6 $\pm$ 6.2 & 58.4 $\pm$ 4.5 & 35.2 $\pm$ 1.8 & 33.0 $\pm$ 22.8 & 22.8 $\pm$ 4.6 \\
& AdamW (0.01) & 23.2 $\pm$ 0.6 & 36.2 $\pm$ 1.0 & 2.1 $\pm$ 0.3 & 20.4 $\pm$ 8.1 & 1.1 $\pm$ 0.2 & 2.2 $\pm$ 1.5 & 3.4 $\pm$ 0.6 & 27.4 $\pm$ 7.2 & 60.1 $\pm$ 5.2 & 29.6 $\pm$ 3.4 & 32.4 $\pm$ 20.2 & 21.6 $\pm$ 4.4 \\
& AdamW (0.001) & 23.0 $\pm$ 0.1 & 36.1 $\pm$ 0.2 & 2.2 $\pm$ 0.1 & 23.9 $\pm$ 5.1 & 1.0 $\pm$ 0.2 & 3.1 $\pm$ 1.6 & 5.2 $\pm$ 2.6 & 33.1 $\pm$ 4.0 & 61.3 $\pm$ 6.7 & 29.0 $\pm$ 2.6 & 31.2 $\pm$ 19.7 & 22.6 $\pm$ 3.9 \\
\cmidrule(lr){2-14}
& Adam & 34.2 $\pm$ 0.9 & 41.8 $\pm$ 0.8 & 1.6 $\pm$ 0.1 & 24.7 $\pm$ 4.6 & 13.3 $\pm$ 15.4 & 5.8 $\pm$ 5.1 & 3.2 $\pm$ 0.7 & 23.3 $\pm$ 23.9 & 55.4 $\pm$ 27.7 & 31.6 $\pm$ 1.5 & 26.8 $\pm$ 13.4 & 23.8 $\pm$ 8.5 \\
& SGD & 72.4 $\pm$ 0.3 & 75.2 $\pm$ 0.3 & 42.0 $\pm$ 0.8 & 85.3 $\pm$ 3.5 & 81.2 $\pm$ 1.2 & 86.5 $\pm$ 0.1 & 93.3 $\pm$ 0.2 & 95.2 $\pm$ 0.4 & 95.7 $\pm$ 0.1 & 68.1 $\pm$ 0.9 & 84.2 $\pm$ 0.2 & 79.9 $\pm$ 0.7 \\
& SMD & 72.6 $\pm$ 0.2 & 75.2 $\pm$ 0.4 & 43.8 $\pm$ 0.7 & 89.4 $\pm$ 2.0 & 82.9 $\pm$ 0.3 & 86.6 $\pm$ 0.2 & 93.4 $\pm$ 0.3 & 95.3 $\pm$ 0.3 & 95.8 $\pm$ 0.1 & 67.4 $\pm$ 0.5 & 84.1 $\pm$ 0.1 & 80.6 $\pm$ 0.4 \\
\midrule
\multirow{6}{*}{16} & AdamW (0.1) & 28.7 $\pm$ 0.3 & 40.6 $\pm$ 0.4 & 3.0 $\pm$ 0.5 & 18.1 $\pm$ 6.9 & 3.5 $\pm$ 1.3 & 4.8 $\pm$ 3.3 & 14.0 $\pm$ 11.9 & 64.4 $\pm$ 7.7 & 44.0 $\pm$ 31.2 & 40.1 $\pm$ 2.4 & 22.0 $\pm$ 13.0 & 25.8 $\pm$ 7.2 \\
& AdamW (0.01) & 29.5 $\pm$ 0.6 & 42.1 $\pm$ 0.4 & 3.4 $\pm$ 0.4 & 21.8 $\pm$ 11.7 & 3.6 $\pm$ 1.2 & 3.4 $\pm$ 2.4 & 19.9 $\pm$ 19.5 & 63.8 $\pm$ 5.5 & 48.1 $\pm$ 31.7 & 35.8 $\pm$ 6.7 & 18.5 $\pm$ 7.3 & 26.3 $\pm$ 8.0 \\
& AdamW (0.001) & 29.2 $\pm$ 0.1 & 41.4 $\pm$ 0.3 & 3.4 $\pm$ 0.6 & 16.7 $\pm$ 8.3 & 3.7 $\pm$ 1.2 & 7.3 $\pm$ 5.7 & 18.3 $\pm$ 15.3 & 64.3 $\pm$ 12.1 & 49.0 $\pm$ 31.0 & 34.3 $\pm$ 10.3 & 22.6 $\pm$ 10.6 & 26.4 $\pm$ 8.7 \\
\cmidrule(lr){2-14}
& Adam & 39.7 $\pm$ 0.2 & 48.5 $\pm$ 0.3 & 4.5 $\pm$ 1.0 & 34.8 $\pm$ 16.1 & 7.6 $\pm$ 5.5 & 12.8 $\pm$ 8.5 & 8.7 $\pm$ 8.4 & 66.0 $\pm$ 8.0 & 59.0 $\pm$ 36.5 & 40.4 $\pm$ 1.3 & 20.4 $\pm$ 10.6 & 31.1 $\pm$ 8.8 \\
& SGD & 73.9 $\pm$ 0.1 & 76.7 $\pm$ 0.1 & 51.2 $\pm$ 0.9 & 92.2 $\pm$ 0.7 & 85.4 $\pm$ 0.1 & 87.2 $\pm$ 0.1 & 94.4 $\pm$ 0.2 & 97.4 $\pm$ 0.2 & 96.2 $\pm$ 0.1 & 72.0 $\pm$ 0.8 & 86.6 $\pm$ 0.2 & 83.0 $\pm$ 0.3 \\
& SMD & 74.0 $\pm$ 0.1 & 76.8 $\pm$ 0.1 & 52.3 $\pm$ 0.8 & 92.1 $\pm$ 0.4 & 85.8 $\pm$ 0.3 & 87.2 $\pm$ 0.1 & 94.4 $\pm$ 0.1 & 97.4 $\pm$ 0.2 & 96.2 $\pm$ 0.2 & 71.4 $\pm$ 0.4 & 86.3 $\pm$ 0.1 & 83.1 $\pm$ 0.2 \\
\bottomrule
\end{tabular}}

\label{tab:ablation_adamw_ViTB16_with_aug}
\end{table*}
 
\begin{table*}[h]
\caption{Few-shot classification with CLIP (ViT-L/14) comparing AdamW, Adam, SGD, and SMD. AdamW denotes Adam with weight-decay coefficients 0.1, 0.01, and 0.001; SMD corresponds to SGD with momentum. Top-1 accuracy (\%) averaged over 3 fixed random seeds.}
\centering
\resizebox{1.\textwidth}{!}{
\begin{tabular}{llcccccccccccc}
\toprule
Shots & Method & ImageNet & SUN & Aircraft & EuroSAT & Cars & Food & Pets & Flowers & Caltech & DTD & UCF & Average \\ \midrule
\multirow{1}{*}{0} & CLIP & 73.5 & 67.7 & 30.0 & 55.1 & 76.8 & 90.3 & 93.1 & 76.2 & 95.2 & 52.4 & 73.7 & 71.3 \\
\midrule
\multirow{6}{*}{1} & AdamW (0.1) & 14.3 $\pm$ 0.3 & 22.0 $\pm$ 2.0 & 1.2 $\pm$ 0.2 & 19.9 $\pm$ 6.0 & 0.5 $\pm$ 0.0 & 7.8 $\pm$ 6.3 & 22.7 $\pm$ 9.6 & 23.5 $\pm$ 12.4 & 52.6 $\pm$ 3.5 & 21.0 $\pm$ 2.1 & 33.1 $\pm$ 5.3 & 19.9 $\pm$ 4.3 \\
& AdamW (0.01) & 14.2 $\pm$ 1.1 & 22.9 $\pm$ 1.9 & 1.4 $\pm$ 0.2 & 16.7 $\pm$ 6.6 & 0.5 $\pm$ 0.0 & 6.0 $\pm$ 4.7 & 23.6 $\pm$ 9.5 & 26.0 $\pm$ 12.8 & 56.2 $\pm$ 3.2 & 22.5 $\pm$ 3.1 & 33.6 $\pm$ 1.5 & 20.3 $\pm$ 4.0 \\
& AdamW (0.001) & 14.5 $\pm$ 0.9 & 22.2 $\pm$ 2.0 & 1.2 $\pm$ 0.3 & 14.4 $\pm$ 2.0 & 0.5 $\pm$ 0.0 & 5.7 $\pm$ 5.3 & 25.2 $\pm$ 10.5 & 24.4 $\pm$ 11.0 & 58.8 $\pm$ 1.5 & 21.4 $\pm$ 1.7 & 35.4 $\pm$ 2.9 & 20.3 $\pm$ 3.5 \\
\cmidrule(lr){2-14}
& Adam & 14.8 $\pm$ 0.7 & 23.5 $\pm$ 1.2 & 1.2 $\pm$ 0.0 & 18.3 $\pm$ 8.2 & 0.5 $\pm$ 0.0 & 5.2 $\pm$ 4.0 & 17.6 $\pm$ 11.2 & 18.8 $\pm$ 4.7 & 44.0 $\pm$ 17.8 & 21.5 $\pm$ 3.2 & 33.8 $\pm$ 4.3 & 18.1 $\pm$ 5.0 \\
& SGD & 76.6 $\pm$ 0.2 & 74.3 $\pm$ 0.1 & 41.1 $\pm$ 0.9 & 75.3 $\pm$ 4.1 & 80.7 $\pm$ 0.4 & 91.0 $\pm$ 0.1 & 94.6 $\pm$ 0.2 & 90.2 $\pm$ 0.8 & 95.6 $\pm$ 0.5 & 60.5 $\pm$ 0.4 & 81.9 $\pm$ 0.5 & 78.3 $\pm$ 0.7 \\
& SMD & 76.6 $\pm$ 0.2 & 74.4 $\pm$ 0.1 & 41.9 $\pm$ 0.7 & 72.8 $\pm$ 2.3 & 80.9 $\pm$ 0.1 & 91.0 $\pm$ 0.2 & 94.1 $\pm$ 0.4 & 89.9 $\pm$ 1.2 & 95.6 $\pm$ 0.5 & 60.1 $\pm$ 1.1 & 81.8 $\pm$ 0.4 & 78.1 $\pm$ 0.6 \\
\midrule
\multirow{6}{*}{2} & AdamW (0.1) & 18.6 $\pm$ 0.9 & 29.5 $\pm$ 1.7 & 1.3 $\pm$ 0.3 & 14.5 $\pm$ 3.6 & 0.6 $\pm$ 0.1 & 6.7 $\pm$ 2.6 & 3.2 $\pm$ 0.5 & 15.7 $\pm$ 2.6 & 58.4 $\pm$ 8.7 & 24.7 $\pm$ 2.6 & 21.1 $\pm$ 14.0 & 17.7 $\pm$ 3.4 \\
& AdamW (0.01) & 20.1 $\pm$ 0.6 & 31.0 $\pm$ 1.3 & 1.0 $\pm$ 0.0 & 20.4 $\pm$ 5.1 & 0.5 $\pm$ 0.0 & 6.9 $\pm$ 3.8 & 7.9 $\pm$ 7.2 & 10.6 $\pm$ 5.2 & 43.5 $\pm$ 30.1 & 23.1 $\pm$ 2.9 & 26.6 $\pm$ 8.8 & 17.4 $\pm$ 5.9 \\
& AdamW (0.001) & 19.3 $\pm$ 0.7 & 29.5 $\pm$ 0.9 & 1.1 $\pm$ 0.1 & 20.3 $\pm$ 1.3 & 0.6 $\pm$ 0.1 & 6.1 $\pm$ 4.5 & 9.8 $\pm$ 9.8 & 16.0 $\pm$ 6.2 & 60.0 $\pm$ 9.5 & 24.7 $\pm$ 1.6 & 17.6 $\pm$ 13.7 & 18.6 $\pm$ 4.4 \\
\cmidrule(lr){2-14}
& Adam & 18.2 $\pm$ 0.3 & 31.3 $\pm$ 1.1 & 1.1 $\pm$ 0.1 & 20.4 $\pm$ 4.9 & 0.6 $\pm$ 0.1 & 3.1 $\pm$ 2.2 & 3.1 $\pm$ 0.5 & 14.6 $\pm$ 6.0 & 57.7 $\pm$ 6.0 & 10.2 $\pm$ 7.6 & 20.9 $\pm$ 12.2 & 16.5 $\pm$ 3.7 \\
& SGD & 76.8 $\pm$ 0.2 & 76.0 $\pm$ 0.1 & 43.3 $\pm$ 1.6 & 82.0 $\pm$ 1.5 & 82.9 $\pm$ 0.2 & 91.2 $\pm$ 0.2 & 94.4 $\pm$ 0.2 & 94.1 $\pm$ 0.4 & 96.6 $\pm$ 0.6 & 66.8 $\pm$ 1.2 & 83.7 $\pm$ 0.5 & 80.7 $\pm$ 0.6 \\
& SMD & 76.8 $\pm$ 0.3 & 76.1 $\pm$ 0.2 & 43.5 $\pm$ 1.6 & 81.2 $\pm$ 0.5 & 82.8 $\pm$ 0.2 & 91.2 $\pm$ 0.2 & 94.4 $\pm$ 0.5 & 94.3 $\pm$ 0.2 & 96.6 $\pm$ 0.6 & 66.8 $\pm$ 1.0 & 84.0 $\pm$ 0.4 & 80.7 $\pm$ 0.5 \\
\midrule
\multirow{6}{*}{4} & AdamW (0.1) & 23.2 $\pm$ 1.3 & 35.8 $\pm$ 0.4 & 2.9 $\pm$ 1.3 & 23.6 $\pm$ 7.5 & 1.9 $\pm$ 0.8 & 1.6 $\pm$ 0.6 & 20.9 $\pm$ 19.6 & 52.2 $\pm$ 0.9 & 68.1 $\pm$ 10.3 & 35.9 $\pm$ 1.9 & 36.5 $\pm$ 20.7 & 27.5 $\pm$ 5.9 \\
& AdamW (0.01) & 23.5 $\pm$ 0.6 & 36.6 $\pm$ 0.2 & 2.7 $\pm$ 1.1 & 17.8 $\pm$ 3.7 & 4.6 $\pm$ 3.4 & 1.8 $\pm$ 1.2 & 21.5 $\pm$ 18.0 & 50.4 $\pm$ 3.0 & 62.1 $\pm$ 13.4 & 37.1 $\pm$ 2.1 & 34.5 $\pm$ 20.9 & 26.6 $\pm$ 6.1 \\
& AdamW (0.001) & 24.0 $\pm$ 0.2 & 36.7 $\pm$ 0.3 & 2.2 $\pm$ 0.8 & 23.0 $\pm$ 13.0 & 1.6 $\pm$ 0.8 & 1.7 $\pm$ 0.5 & 19.6 $\pm$ 17.3 & 49.0 $\pm$ 2.2 & 65.1 $\pm$ 10.9 & 37.4 $\pm$ 1.3 & 35.5 $\pm$ 23.6 & 26.9 $\pm$ 6.4 \\
\cmidrule(lr){2-14}
& Adam & 24.1 $\pm$ 0.6 & 36.1 $\pm$ 1.4 & 1.7 $\pm$ 0.5 & 21.2 $\pm$ 5.3 & 5.9 $\pm$ 2.7 & 3.3 $\pm$ 3.3 & 20.6 $\pm$ 17.6 & 47.5 $\pm$ 10.6 & 60.6 $\pm$ 22.5 & 38.6 $\pm$ 2.4 & 27.2 $\pm$ 19.0 & 26.1 $\pm$ 7.8 \\
& SGD & 77.7 $\pm$ 0.1 & 77.6 $\pm$ 0.2 & 48.1 $\pm$ 0.5 & 83.4 $\pm$ 0.7 & 85.8 $\pm$ 0.3 & 91.5 $\pm$ 0.1 & 95.4 $\pm$ 0.1 & 96.8 $\pm$ 0.3 & 97.2 $\pm$ 0.4 & 69.6 $\pm$ 1.1 & 86.3 $\pm$ 0.3 & 82.7 $\pm$ 0.4 \\
& SMD & 77.7 $\pm$ 0.1 & 77.6 $\pm$ 0.2 & 49.9 $\pm$ 0.5 & 80.1 $\pm$ 1.9 & 85.6 $\pm$ 0.3 & 91.5 $\pm$ 0.1 & 95.5 $\pm$ 0.1 & 96.9 $\pm$ 0.3 & 97.2 $\pm$ 0.5 & 69.3 $\pm$ 1.2 & 86.7 $\pm$ 0.6 & 82.5 $\pm$ 0.5 \\
\midrule
\multirow{6}{*}{8} & AdamW (0.1) & 28.8 $\pm$ 0.3 & 41.4 $\pm$ 1.0 & 2.2 $\pm$ 0.2 & 14.6 $\pm$ 4.6 & 1.8 $\pm$ 0.4 & 32.9 $\pm$ 11.4 & 45.3 $\pm$ 29.2 & 67.1 $\pm$ 8.9 & 82.2 $\pm$ 1.7 & 43.3 $\pm$ 2.1 & 59.4 $\pm$ 0.8 & 38.1 $\pm$ 5.5 \\
& AdamW (0.01) & 29.0 $\pm$ 0.5 & 42.2 $\pm$ 0.8 & 3.2 $\pm$ 0.1 & 20.3 $\pm$ 6.6 & 1.8 $\pm$ 0.9 & 33.9 $\pm$ 11.5 & 41.1 $\pm$ 24.6 & 64.9 $\pm$ 6.2 & 80.7 $\pm$ 0.5 & 40.2 $\pm$ 2.0 & 57.9 $\pm$ 2.0 & 37.7 $\pm$ 5.0 \\
& AdamW (0.001) & 28.1 $\pm$ 1.1 & 40.4 $\pm$ 1.3 & 2.8 $\pm$ 0.6 & 20.0 $\pm$ 6.3 & 1.8 $\pm$ 0.8 & 32.5 $\pm$ 11.1 & 47.2 $\pm$ 23.6 & 63.1 $\pm$ 7.8 & 83.5 $\pm$ 1.0 & 43.2 $\pm$ 1.7 & 59.4 $\pm$ 1.1 & 38.4 $\pm$ 5.1 \\
\cmidrule(lr){2-14}
& Adam & 28.1 $\pm$ 1.0 & 42.1 $\pm$ 1.0 & 2.8 $\pm$ 0.6 & 28.8 $\pm$ 16.0 & 3.0 $\pm$ 1.5 & 5.6 $\pm$ 4.3 & 42.6 $\pm$ 26.3 & 62.4 $\pm$ 13.7 & 83.8 $\pm$ 1.2 & 42.7 $\pm$ 3.4 & 60.7 $\pm$ 1.5 & 36.6 $\pm$ 6.4 \\
& SGD & 78.4 $\pm$ 0.2 & 79.1 $\pm$ 0.2 & 56.5 $\pm$ 0.9 & 88.2 $\pm$ 1.3 & 88.5 $\pm$ 0.2 & 91.8 $\pm$ 0.1 & 95.3 $\pm$ 0.4 & 97.7 $\pm$ 0.1 & 97.5 $\pm$ 0.5 & 72.2 $\pm$ 0.4 & 88.3 $\pm$ 0.1 & 84.9 $\pm$ 0.4 \\
& SMD & 78.5 $\pm$ 0.2 & 79.2 $\pm$ 0.2 & 56.1 $\pm$ 0.5 & 88.3 $\pm$ 1.2 & 88.5 $\pm$ 0.3 & 91.8 $\pm$ 0.1 & 95.4 $\pm$ 0.5 & 97.8 $\pm$ 0.2 & 97.5 $\pm$ 0.5 & 72.3 $\pm$ 0.4 & 88.4 $\pm$ 0.2 & 84.9 $\pm$ 0.4 \\
\midrule
\multirow{6}{*}{16} & AdamW (0.1) & 34.0 $\pm$ 0.5 & 46.1 $\pm$ 1.0 & 8.9 $\pm$ 2.6 & 43.7 $\pm$ 16.9 & 40.2 $\pm$ 27.1 & 21.2 $\pm$ 11.6 & 62.1 $\pm$ 9.0 & 87.6 $\pm$ 3.9 & 83.4 $\pm$ 3.2 & 50.3 $\pm$ 0.9 & 63.5 $\pm$ 0.8 & 49.2 $\pm$ 7.0 \\
& AdamW (0.01) & 34.7 $\pm$ 1.1 & 46.7 $\pm$ 1.2 & 5.7 $\pm$ 1.9 & 34.7 $\pm$ 8.3 & 39.3 $\pm$ 23.5 & 21.4 $\pm$ 15.5 & 67.2 $\pm$ 8.1 & 88.4 $\pm$ 2.2 & 86.5 $\pm$ 1.4 & 51.6 $\pm$ 0.7 & 63.0 $\pm$ 2.2 & 49.0 $\pm$ 6.0 \\
& AdamW (0.001) & 35.0 $\pm$ 0.6 & 47.0 $\pm$ 0.9 & 10.9 $\pm$ 5.1 & 43.6 $\pm$ 8.9 & 26.3 $\pm$ 32.0 & 28.6 $\pm$ 7.8 & 64.0 $\pm$ 10.8 & 86.3 $\pm$ 2.4 & 82.1 $\pm$ 5.5 & 50.0 $\pm$ 0.6 & 63.5 $\pm$ 1.7 & 48.8 $\pm$ 6.9 \\
\cmidrule(lr){2-14}
& Adam & 34.2 $\pm$ 0.5 & 47.0 $\pm$ 0.5 & 11.2 $\pm$ 6.7 & 45.3 $\pm$ 14.5 & 42.6 $\pm$ 30.1 & 19.0 $\pm$ 10.9 & 56.1 $\pm$ 8.8 & 86.0 $\pm$ 2.2 & 74.6 $\pm$ 16.8 & 49.3 $\pm$ 0.7 & 63.9 $\pm$ 3.2 & 48.1 $\pm$ 8.6 \\
& SGD & 79.5 $\pm$ 0.2 & 80.5 $\pm$ 0.2 & 63.2 $\pm$ 0.5 & 92.0 $\pm$ 0.4 & 90.4 $\pm$ 0.1 & 92.0 $\pm$ 0.0 & 95.7 $\pm$ 0.1 & 98.7 $\pm$ 0.2 & 97.6 $\pm$ 0.2 & 76.8 $\pm$ 0.4 & 90.0 $\pm$ 0.2 & 86.9 $\pm$ 0.2 \\
& SMD & 79.5 $\pm$ 0.1 & 80.5 $\pm$ 0.1 & 63.5 $\pm$ 0.4 & 92.0 $\pm$ 0.1 & 90.4 $\pm$ 0.2 & 92.0 $\pm$ 0.1 & 95.9 $\pm$ 0.0 & 98.7 $\pm$ 0.1 & 97.6 $\pm$ 0.2 & 76.5 $\pm$ 0.6 & 90.0 $\pm$ 0.2 & 87.0 $\pm$ 0.2 \\
\bottomrule
\end{tabular}}

\label{tab:ablation_adamw_ViTL14_with_aug}
\end{table*}

Tables~\ref{tab:ablation_adamw_ViTB32_with_aug}, \ref{tab:ablation_adamw_ViTB16_with_aug}, and \ref{tab:ablation_adamw_ViTL14_with_aug} compare full fine-tuning of CLIP with Adam, AdamW, SGD, and SGD with momentum (SMD) for ViT-B/32, ViT-B/16, and ViT-L/14 under the shared training configuration used in the main paper (30 epochs, batch size $64$, learning rate $10^{-4}$).  
AdamW is the decoupled weight-decay variant of Adam. Instead of adding an $\ell_2$ penalty to the loss, AdamW shrinks the parameters at every step as:
\begin{equation}
    \Theta_{t+1} \leftarrow (1 - \eta \,\omega_{\mathrm{wd}})\,\Theta_t - \eta\,\widehat m_{t+1}\oslash(\sqrt{\widehat v_{t+1}}+\varepsilon\mathbf{1}_d),
\end{equation}

\noindent where $\eta$ is the learning rate, $\omega_{\mathrm{wd}}$ is the weight-decay coefficient, and $(\widehat m_{t+1},\widehat v_{t+1})$ are the bias-corrected first- and second-moment estimates. We sweep $\omega_{\mathrm{wd}} \in \{0.1, 0.01, 0.001\}$. Larger values correspond to stronger parameter shrinkage and thus stronger regularization, while $\omega_{\mathrm{wd}} = 0.001$ is a comparatively mild decay.

Across all backbones and shot levels, Adam and AdamW collapse in the few-shot regime, and varying the weight-decay coefficient cannot remedy this behavior. For ViT-B/32 (Table~\ref{tab:ablation_adamw_ViTB32_with_aug}), the zero-shot CLIP baseline averages $60.2\%$ accuracy, whereas 1-shot fine-tuning with AdamW achieves only $6.9$--$7.1\%$ and Adam reaches $8.8\%$ on average. All Adam/AdamW variants remain far below the zero-shot baseline, even at 16 shots, where the best AdamW configuration obtains $32.3\%$, and Adam reaches $35.0\%$, i.e., more than $25$ percentage points below CLIP. A similar pattern holds for ViT-B/16 (Table~\ref{tab:ablation_adamw_ViTB16_with_aug}): CLIP starts at $64.1\%$, but AdamW yields only $7.2$--$7.8\%$ at 1 shot and at most $26.4\%$ at 16 shots, while Adam remains between $10.3\%$ (1 shot) and $31.1\%$ (16 shots). For the stronger ViT-L/14 backbone (Table~\ref{tab:ablation_adamw_ViTL14_with_aug}), AdamW performs slightly better in absolute terms (around $20.3\%$ at 1 shot and up to $49.2\%$ at 16 shots), yet still underperforms the zero-shot CLIP baseline of $71.3\%$ by more than $20$ percentage points at the highest shot level. Within each backbone and shot level, the three AdamW configurations span $0.1$--$1.5$ percentage points, and there is no consistent preference for $\omega_{\mathrm{wd}} = 0.1$, $0.01$, or $0.001$. Changing weight decay has little effect relative to the gaps to CLIP and SGD. This result is consistent with the second-moment mechanism examined in the diagnostic experiments.

In contrast, SGD consistently yields strong gains over the zero-shot baseline and substantially outperforms all adaptive variants. For ViT-B/32, SGD achieves $67.7\%$ at 1 shot and $79.1\%$ at 16 shots, improving over CLIP by $+7.5$ and $+18.9$ percentage points, respectively. The corresponding gaps to Adam are $+58.9$ (1 shot) and $+44.1$ (16 shots), and to the best AdamW configuration, they are $+60.6$ and $+46.8$ percentage points. For ViT-B/16, SGD reaches $72.8\%$ at 1 shot and $83.0\%$ at 16 shots, i.e., $+8.7$ and $+18.9$ over CLIP. In this case, the margins over Adam range from $+62.5$ (1 shot) to $+51.9$ (16 shots), and over the best AdamW variant from $+65.0$ to $+56.6$ points. For ViT-L/14, SGD attains $78.3\%$ at 1 shot and $86.9\%$ at 16 shots. It improves over CLIP by $+7.0$ and $+15.6$ points, while exceeding Adam by $+60.2$ to $+38.8$ points and the best AdamW configuration by $+58.0$ to $+37.7$ points as the number of shots increases from 1 to 16. Averaged across all three backbones and five shot levels, SGD improves over CLIP by roughly $+12.7$ percentage points, whereas Adam and the best AdamW configurations reduce accuracy by approximately $-44$ points on average. This indicates that plain SGD is a much more reliable optimizer for few-shot CLIP fine-tuning than adaptive Adam-type variants.

Finally, SMD (SGD with momentum) closely tracks the behavior of plain SGD across all settings. For ViT-B/32, SMD ranges from $66.2\%$ (1 shot) to $79.2\%$ (16 shots). For ViT-B/16, it ranges from $71.7\%$ to $83.1\%$. For ViT-L/14, it ranges from $78.1\%$ to $87.0\%$. These results are always above the zero-shot baseline and nearly indistinguishable from SGD. The absolute difference between SGD and SMD is at most $1.5$ percentage points and often below $1$ point, with SMD sometimes slightly better and sometimes slightly worse, and without any systematic trend in favor of momentum. These results corroborate the main paper observation that the first-moment accumulator provides, at best, marginal benefits in this regime while incurring additional memory overhead. The primary factor behind the collapse of Adam and AdamW is the second-moment term, and once this is removed, both SGD and SMD converge to similarly good solutions.  
Overall, the tables confirm that, under a unified and simple training configuration, full fine-tuning with SGD (or SMD) is consistently superior to Adam and AdamW across all backbones and shot levels, and that tuning the weight-decay coefficient of AdamW alone cannot rescue the performance of adaptive optimizers in few-shot CLIP adaptation.

\section{SGD vs PEFT}
\label{app:sgd_vs_peft}

\begin{figure*}[h]
  \centering
  \resizebox{0.75\textwidth}{!}{\includegraphics[width=\linewidth]{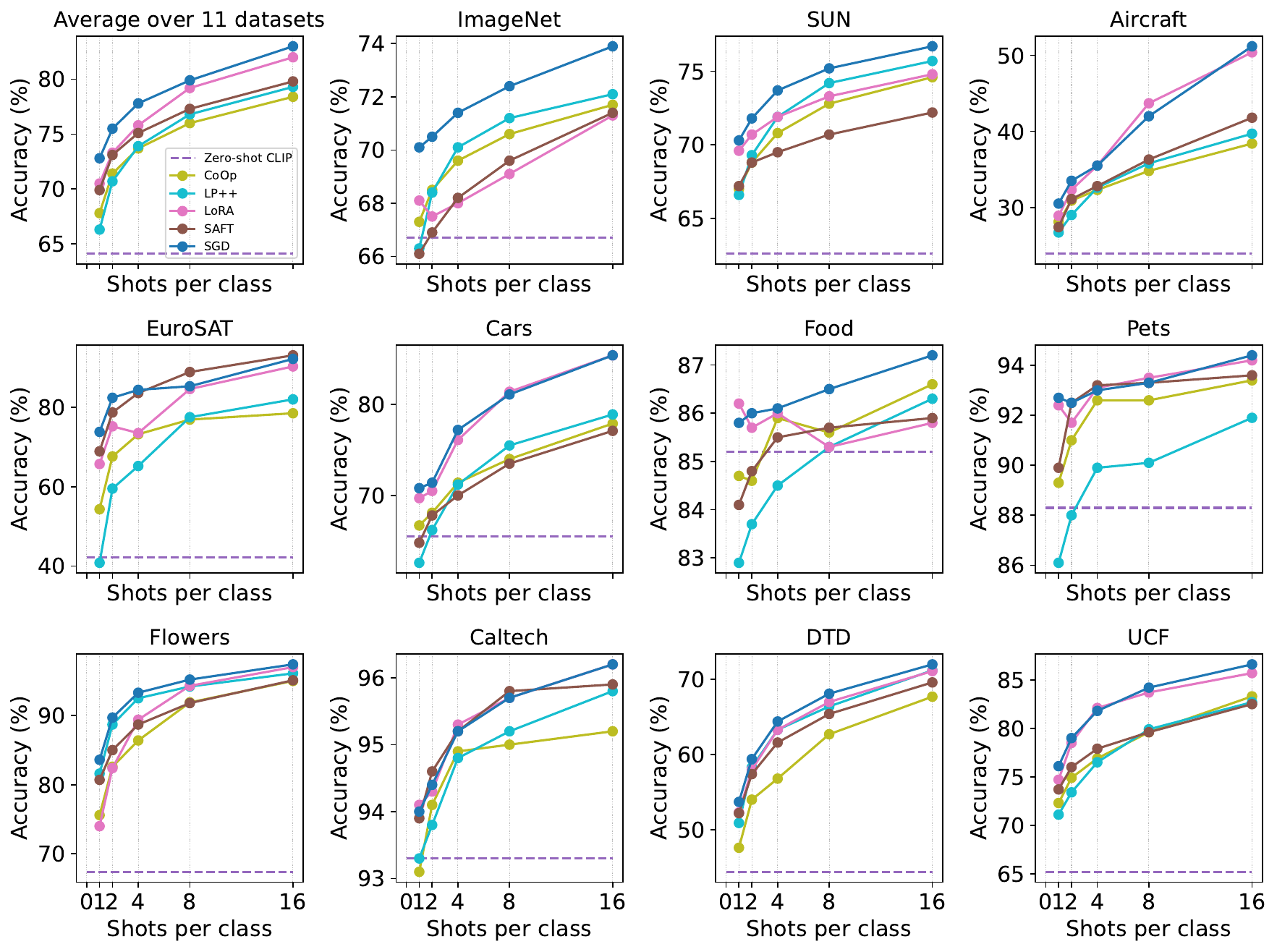}}
  \caption{Few-shot performance of SGD compared with PEFT methods on 11 downstream datasets using CLIP (ViT-B/16) backbone.}
  \label{fig:fewshot_sgd_vs_peft_ViT-B16}
\end{figure*}

\begin{figure*}
  \centering
  \resizebox{0.75\textwidth}{!}{\includegraphics[width=\linewidth]{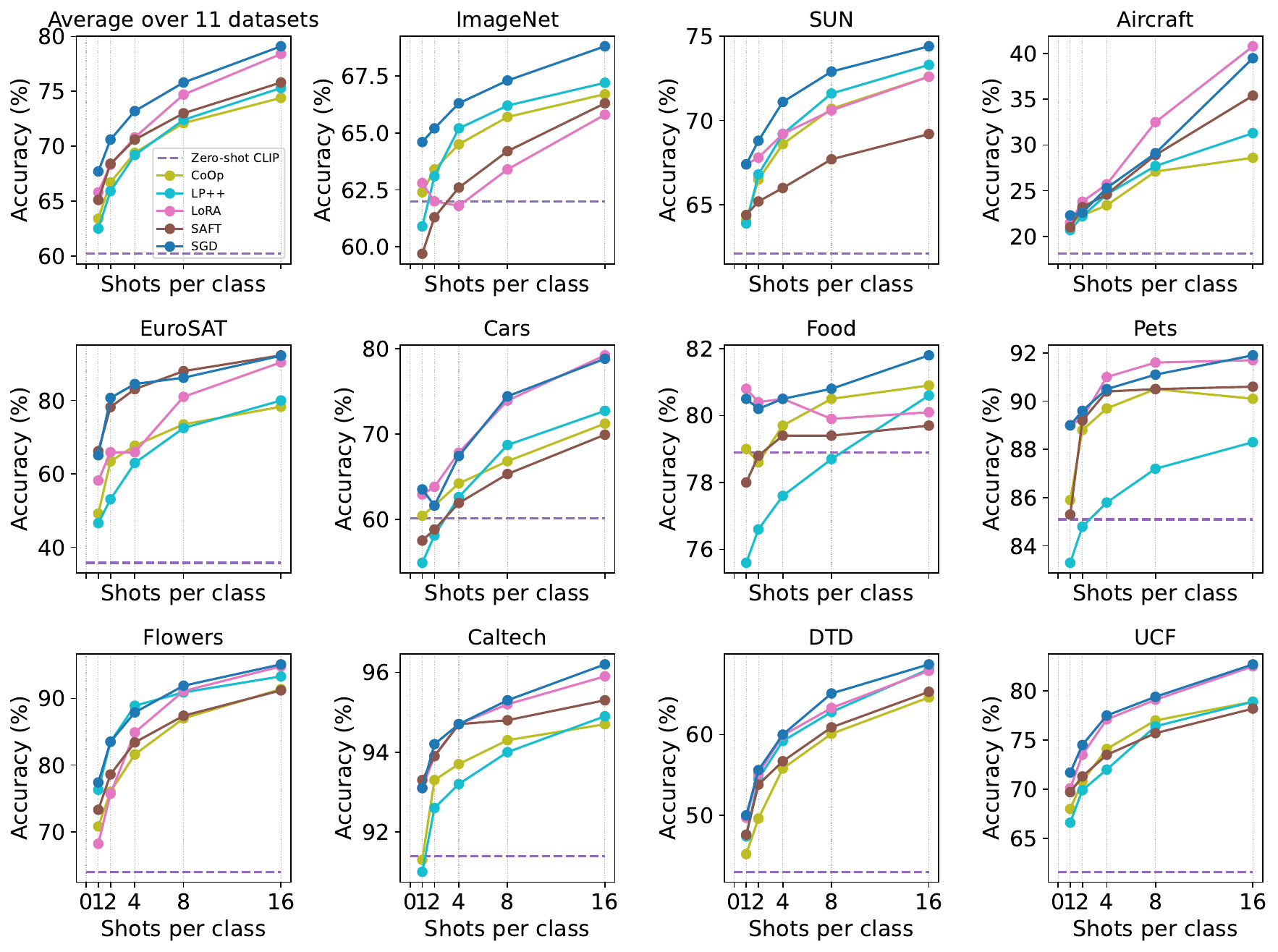}}
  \caption{Few-shot performance of SGD compared with PEFT methods on 11 downstream datasets using CLIP (ViT-B/32) backbone.}
  \label{fig:fewshot_sgd_vs_peft_ViT-B32}
\end{figure*}

\begin{figure*}
  \centering
  \resizebox{0.75\textwidth}{!}{\includegraphics[width=\linewidth]{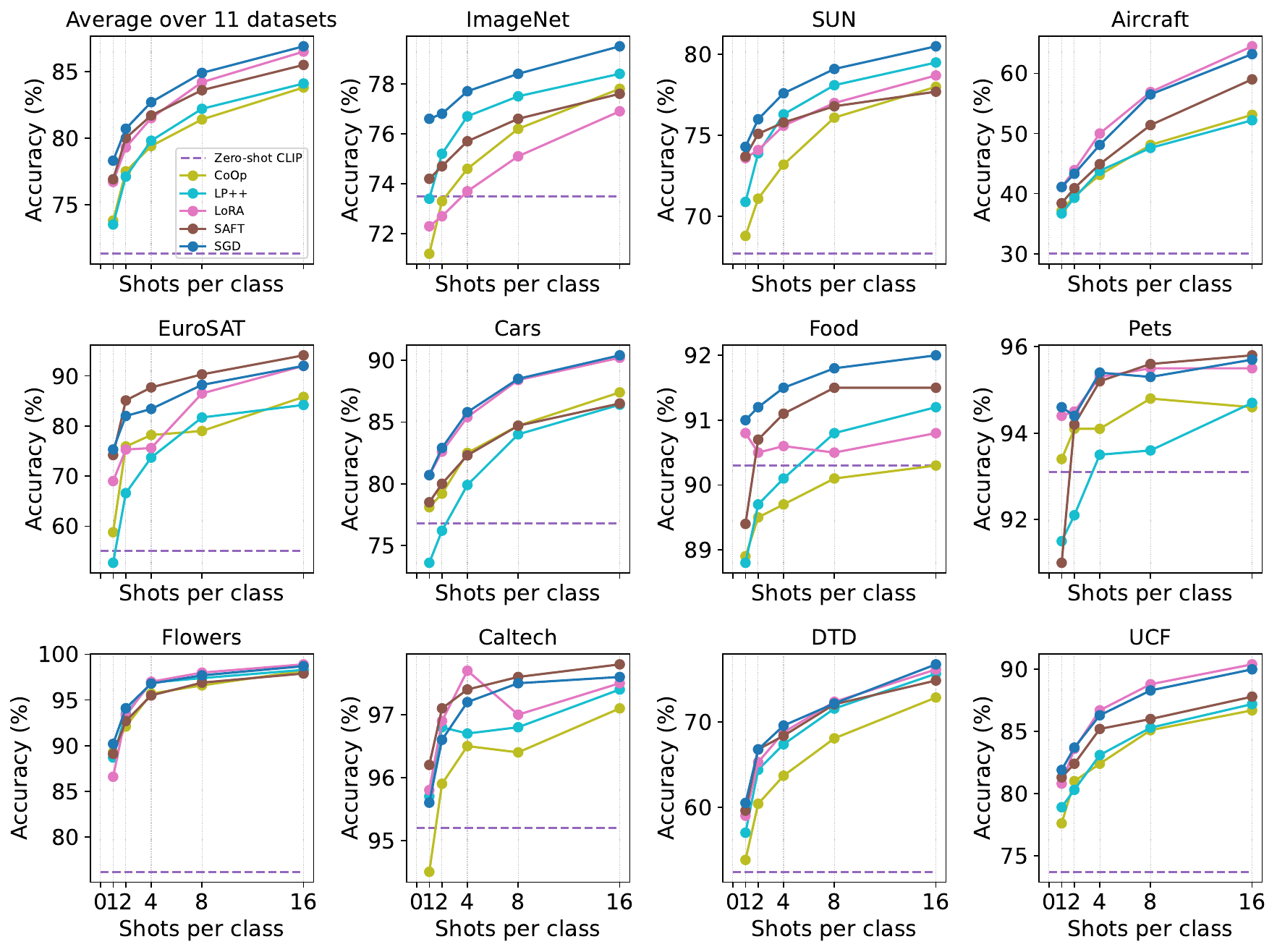}}
  \caption{Few-shot performance of SGD compared with PEFT methods on 11 downstream datasets using CLIP (ViT-L/14) backbone.}
  \label{fig:fewshot_sgd_vs_peft_ViT-L14}
\end{figure*}

This appendix provides a complete comparison between dense full fine-tuning with SGD and parameter-efficient fine-tuning (PEFT). Figures~\ref{fig:fewshot_sgd_vs_peft_ViT-B16}--\ref{fig:fewshot_sgd_vs_peft_ViT-L14} show the averages across 11 datasets. Tables~\ref{tab:few_shot_ViTB32_with_aug}, \ref{tab:few_shot_ViTB16_with_aug}, and \ref{tab:few_shot_ViTL14_with_aug} report the dataset-level results. The tables serve as the primary evidence. They display both the average ordering and the dataset-specific exceptions.

\begin{table*}
\caption{Few‑shot classification performance on 11 datasets with CLIP(ViT-B/32) backbone. Top‑1 accuracy (\%) averaged over 3 random seeds.}
\label{tab:few_shot_ViTB32_with_aug}
\centering
\footnotesize
\setlength{\tabcolsep}{1.8pt}
\renewcommand{\arraystretch}{1.00}
\begin{adjustbox}{max totalsize={\textwidth}{0.95\textheight},center}
\begin{tabular}{@{}c l *{12}{c}@{}}
\toprule
Shots & Method & ImageNet & SUN & Aircr. & EuroSAT & Cars & Food & Pets & Flow. & Caltech & DTD & UCF & Avg. \\ \midrule
\multirow{1}{*}{0} & CLIP & 62.0 & 62.1 & 18.1 & 35.8 & 60.1 & 78.9 & 85.1 & 64.0 & 91.4 & 43.0 & 61.6 & 60.2 \\
\midrule
\multirow{15}{*}{1} 
& CoOp & 62.4 $\pm$ 0.5 & 64.1 $\pm$ 0.2 & 21.4 $\pm$ 1.1 & 49.2 $\pm$ 4.4 & 60.4 $\pm$ 0.8 & 79.0 $\pm$ 1.2 & 85.9 $\pm$ 0.3 & 70.8 $\pm$ 0.9 & 91.3 $\pm$ 0.1 & 45.2 $\pm$ 1.4 & 68.0 $\pm$ 1.7 & 63.4 $\pm$ 1.1 \\
& KgCoOp & 63.5 $\pm$ 0.2 & 65.4 $\pm$ 0.4 & 20.6 $\pm$ 0.6 & 51.9 $\pm$ 0.9 & 60.4 $\pm$ 0.2 & 80.5 $\pm$ 0.0 & 88.5 $\pm$ 0.5 & 65.5 $\pm$ 0.6 & 92.2 $\pm$ 0.3 & 47.4 $\pm$ 1.8 & 66.6 $\pm$ 0.2 & 63.9 $\pm$ 0.5 \\
& ProGrad & 54.2 $\pm$ 4.9 & 61.9 $\pm$ 2.0 & 16.8 $\pm$ 1.8 & 50.7 $\pm$ 2.6 & 59.1 $\pm$ 1.7 & 77.0 $\pm$ 1.6 & 82.8 $\pm$ 4.5 & 69.0 $\pm$ 1.0 & 90.3 $\pm$ 0.8 & 45.1 $\pm$ 0.8 & 66.8 $\pm$ 1.5 & 61.2 $\pm$ 2.1 \\
\cmidrule(lr){2-14}
& Adapter & 62.8 $\pm$ 0.1 & 63.0 $\pm$ 0.0 & 18.5 $\pm$ 0.2 & 39.0 $\pm$ 1.1 & 60.7 $\pm$ 0.1 & 79.3 $\pm$ 0.1 & 86.1 $\pm$ 0.1 & 64.7 $\pm$ 0.3 & 91.6 $\pm$ 0.0 & 43.7 $\pm$ 0.2 & 62.6 $\pm$ 0.3 & 61.1 $\pm$ 0.2 \\
& LP++ & 60.9 $\pm$ 0.2 & 63.9 $\pm$ 0.6 & 20.7 $\pm$ 0.5 & 46.6 $\pm$ 1.2 & 54.9 $\pm$ 0.6 & 75.6 $\pm$ 0.6 & 83.3 $\pm$ 1.5 & 76.3 $\pm$ 1.1 & 91.0 $\pm$ 0.7 & 47.4 $\pm$ 1.4 & 66.6 $\pm$ 0.2 & 62.5 $\pm$ 0.8 \\
& TIP-Adapter & 62.1 $\pm$ 0.0 & 62.1 $\pm$ 0.0 & 18.2 $\pm$ 0.1 & 36.5 $\pm$ 0.4 & 60.2 $\pm$ 0.0 & 78.9 $\pm$ 0.0 & 85.1 $\pm$ 0.0 & 64.3 $\pm$ 0.0 & 91.4 $\pm$ 0.1 & 43.0 $\pm$ 0.0 & 61.7 $\pm$ 0.0 & 60.3 $\pm$ 0.1 \\
\cmidrule(lr){2-14}
& LoRA & 62.8 $\pm$ 0.6 & 67.4 $\pm$ 0.2 & 21.5 $\pm$ 0.2 & 58.2 $\pm$ 2.2 & 62.9 $\pm$ 0.2 & 80.8 $\pm$ 0.2 & 89.0 $\pm$ 0.3 & 68.2 $\pm$ 0.3 & 93.1 $\pm$ 0.1 & 49.7 $\pm$ 0.7 & 70.1 $\pm$ 0.2 & 65.8 $\pm$ 0.5 \\
& PiSSA & 60.7 $\pm$ 0.3 & 65.1 $\pm$ 0.2 & 22.2 $\pm$ 2.0 & 64.3 $\pm$ 6.5 & 62.1 $\pm$ 0.3 & 77.2 $\pm$ 0.7 & 86.4 $\pm$ 0.4 & 76.9 $\pm$ 1.5 & 92.9 $\pm$ 0.8 & 50.1 $\pm$ 1.2 & 70.6 $\pm$ 0.5 & 66.2 $\pm$ 1.3 \\
& GaLoRE & 55.9 $\pm$ 0.6 & 63.5 $\pm$ 0.3 & 20.2 $\pm$ 1.5 & 64.1 $\pm$ 8.1 & 54.4 $\pm$ 1.6 & 75.5 $\pm$ 0.7 & 85.2 $\pm$ 0.4 & 74.7 $\pm$ 0.9 & 91.1 $\pm$ 0.5 & 48.1 $\pm$ 2.1 & 70.5 $\pm$ 0.6 & 63.9 $\pm$ 1.6 \\
\cmidrule(lr){2-14}
& BitFit & 64.1 $\pm$ 0.3 & 66.8 $\pm$ 0.1 & 22.0 $\pm$ 0.4 & 63.0 $\pm$ 2.9 & 62.2 $\pm$ 0.3 & 80.3 $\pm$ 0.4 & 88.9 $\pm$ 0.2 & 70.2 $\pm$ 0.6 & 93.1 $\pm$ 0.2 & 49.9 $\pm$ 0.1 & 70.2 $\pm$ 0.1 & 66.4 $\pm$ 0.5 \\
& LayerNorm & 63.9 $\pm$ 0.1 & 64.3 $\pm$ 0.1 & 19.6 $\pm$ 0.0 & 40.3 $\pm$ 0.5 & 61.3 $\pm$ 0.2 & 79.8 $\pm$ 0.1 & 86.7 $\pm$ 0.1 & 65.8 $\pm$ 0.0 & 91.9 $\pm$ 0.1 & 44.4 $\pm$ 0.3 & 63.4 $\pm$ 0.1 & 62.0 $\pm$ 0.1 \\
& SAFT & 59.7 $\pm$ 0.7 & 64.4 $\pm$ 0.2 & 21.0 $\pm$ 1.0 & 66.2 $\pm$ 1.8 & 57.5 $\pm$ 1.0 & 78.0 $\pm$ 0.6 & 85.3 $\pm$ 1.1 & 73.3 $\pm$ 0.7 & 93.3 $\pm$ 0.7 & 47.6 $\pm$ 0.9 & 69.7 $\pm$ 0.6 & 65.1 $\pm$ 0.8 \\
& SO & 63.2 $\pm$ 0.5 & 67.4 $\pm$ 0.3 & 22.1 $\pm$ 0.2 & 62.5 $\pm$ 1.0 & 63.3 $\pm$ 0.2 & 80.5 $\pm$ 0.4 & 89.2 $\pm$ 0.5 & 70.8 $\pm$ 0.9 & 93.1 $\pm$ 0.3 & 51.0 $\pm$ 0.4 & 71.1 $\pm$ 0.2 & 66.7 $\pm$ 0.4 \\
\cmidrule(lr){2-14}
& Adam & 12.4 $\pm$ 2.6 & 16.6 $\pm$ 1.7 & 1.1 $\pm$ 0.2 & 9.1 $\pm$ 2.0 & 0.5 $\pm$ 0.1 & 2.1 $\pm$ 0.8 & 5.7 $\pm$ 3.7 & 3.6 $\pm$ 2.5 & 30.9 $\pm$ 21.9 & 13.7 $\pm$ 2.6 & 1.4 $\pm$ 0.5 & 8.8 $\pm$ 3.5 \\
& SGD & 64.6 $\pm$ 0.3 & 67.4 $\pm$ 0.1 & 22.3 $\pm$ 1.2 & 65.1 $\pm$ 2.4 & 63.5 $\pm$ 0.7 & 80.5 $\pm$ 0.4 & 89.0 $\pm$ 0.1 & 77.4 $\pm$ 1.0 & 93.1 $\pm$ 0.5 & 50.0 $\pm$ 0.1 & 71.7 $\pm$ 0.3 & 67.7 $\pm$ 0.6 \\
\midrule
\multirow{15}{*}{2} 
& CoOp & 63.4 $\pm$ 0.2 & 66.5 $\pm$ 0.1 & 22.3 $\pm$ 0.3 & 63.4 $\pm$ 3.4 & 61.6 $\pm$ 0.1 & 78.6 $\pm$ 1.4 & 88.8 $\pm$ 0.8 & 76.0 $\pm$ 0.9 & 93.3 $\pm$ 0.3 & 49.6 $\pm$ 1.4 & 70.8 $\pm$ 1.0 & 66.7 $\pm$ 0.9 \\
& KgCoOp & 63.8 $\pm$ 0.2 & 66.0 $\pm$ 0.3 & 20.9 $\pm$ 0.4 & 51.7 $\pm$ 1.0 & 60.8 $\pm$ 0.3 & 80.7 $\pm$ 0.2 & 88.3 $\pm$ 0.3 & 66.5 $\pm$ 0.9 & 92.7 $\pm$ 0.1 & 49.3 $\pm$ 1.0 & 68.1 $\pm$ 0.8 & 64.4 $\pm$ 0.5 \\
& ProGrad & 61.6 $\pm$ 1.2 & 65.2 $\pm$ 2.7 & 19.4 $\pm$ 3.3 & 61.9 $\pm$ 5.1 & 59.3 $\pm$ 2.2 & 74.7 $\pm$ 2.9 & 87.4 $\pm$ 1.2 & 75.0 $\pm$ 1.5 & 92.7 $\pm$ 0.6 & 48.1 $\pm$ 2.3 & 70.4 $\pm$ 0.7 & 65.1 $\pm$ 2.1 \\
\cmidrule(lr){2-14}
& Adapter & 63.4 $\pm$ 0.0 & 64.3 $\pm$ 0.1 & 19.1 $\pm$ 0.1 & 39.5 $\pm$ 1.4 & 61.0 $\pm$ 0.2 & 79.8 $\pm$ 0.2 & 86.3 $\pm$ 0.1 & 65.2 $\pm$ 0.4 & 91.9 $\pm$ 0.1 & 44.7 $\pm$ 0.2 & 63.5 $\pm$ 0.1 & 61.7 $\pm$ 0.3 \\
& LP++ & 63.1 $\pm$ 0.0 & 66.8 $\pm$ 0.2 & 22.2 $\pm$ 0.6 & 53.1 $\pm$ 2.2 & 58.1 $\pm$ 0.5 & 76.6 $\pm$ 0.7 & 84.8 $\pm$ 0.3 & 83.5 $\pm$ 1.0 & 92.6 $\pm$ 0.2 & 54.5 $\pm$ 0.2 & 69.9 $\pm$ 0.7 & 65.9 $\pm$ 0.6 \\
& TIP-Adapter & 62.1 $\pm$ 0.0 & 62.1 $\pm$ 0.0 & 18.2 $\pm$ 0.1 & 37.6 $\pm$ 0.2 & 60.3 $\pm$ 0.0 & 78.9 $\pm$ 0.0 & 85.1 $\pm$ 0.0 & 64.6 $\pm$ 0.1 & 91.4 $\pm$ 0.0 & 43.0 $\pm$ 0.1 & 61.8 $\pm$ 0.1 & 60.5 $\pm$ 0.1 \\
\cmidrule(lr){2-14}
& LoRA & 62.0 $\pm$ 0.2 & 67.8 $\pm$ 0.5 & 23.8 $\pm$ 0.4 & 65.9 $\pm$ 2.0 & 63.8 $\pm$ 0.6 & 80.4 $\pm$ 0.2 & 89.3 $\pm$ 0.9 & 75.7 $\pm$ 1.2 & 93.9 $\pm$ 0.5 & 55.1 $\pm$ 0.5 & 73.5 $\pm$ 0.8 & 68.3 $\pm$ 0.7 \\
& PiSSA & 61.1 $\pm$ 0.1 & 66.8 $\pm$ 0.3 & 24.9 $\pm$ 0.8 & 80.5 $\pm$ 1.2 & 63.7 $\pm$ 0.2 & 77.1 $\pm$ 0.5 & 88.1 $\pm$ 1.0 & 83.0 $\pm$ 1.3 & 94.0 $\pm$ 0.1 & 54.8 $\pm$ 1.0 & 74.0 $\pm$ 1.5 & 69.8 $\pm$ 0.7 \\
& GaLoRE & 57.8 $\pm$ 0.2 & 64.9 $\pm$ 0.2 & 22.6 $\pm$ 1.3 & 76.7 $\pm$ 0.8 & 60.6 $\pm$ 0.9 & 75.1 $\pm$ 0.5 & 85.8 $\pm$ 0.9 & 80.5 $\pm$ 0.7 & 92.3 $\pm$ 0.4 & 53.9 $\pm$ 0.8 & 73.7 $\pm$ 0.6 & 67.6 $\pm$ 0.7 \\
\cmidrule(lr){2-14}
& BitFit & 64.9 $\pm$ 0.2 & 68.2 $\pm$ 0.2 & 23.5 $\pm$ 0.4 & 68.6 $\pm$ 3.9 & 62.9 $\pm$ 0.2 & 80.6 $\pm$ 0.1 & 89.7 $\pm$ 0.6 & 74.0 $\pm$ 1.2 & 93.9 $\pm$ 0.3 & 53.7 $\pm$ 0.7 & 72.5 $\pm$ 0.7 & 68.4 $\pm$ 0.8 \\
& LayerNorm & 64.4 $\pm$ 0.0 & 65.4 $\pm$ 0.1 & 20.3 $\pm$ 0.2 & 41.7 $\pm$ 0.4 & 61.4 $\pm$ 0.1 & 80.5 $\pm$ 0.1 & 86.7 $\pm$ 0.1 & 66.6 $\pm$ 0.0 & 92.3 $\pm$ 0.1 & 46.4 $\pm$ 0.0 & 64.7 $\pm$ 0.1 & 62.8 $\pm$ 0.1 \\
& SAFT & 61.3 $\pm$ 0.1 & 65.2 $\pm$ 0.4 & 23.2 $\pm$ 0.6 & 78.2 $\pm$ 2.5 & 58.8 $\pm$ 0.0 & 78.8 $\pm$ 0.2 & 89.2 $\pm$ 0.2 & 78.6 $\pm$ 1.1 & 93.9 $\pm$ 0.3 & 53.8 $\pm$ 1.4 & 71.3 $\pm$ 0.2 & 68.4 $\pm$ 0.6 \\
& SO & 62.6 $\pm$ 0.2 & 67.6 $\pm$ 0.6 & 24.6 $\pm$ 0.3 & 69.3 $\pm$ 2.1 & 64.2 $\pm$ 0.4 & 80.1 $\pm$ 0.5 & 89.8 $\pm$ 1.0 & 78.7 $\pm$ 1.3 & 93.9 $\pm$ 0.7 & 55.0 $\pm$ 1.0 & 74.6 $\pm$ 0.6 & 69.1 $\pm$ 0.8 \\
\cmidrule(lr){2-14}
& Adam & 18.3 $\pm$ 0.6 & 22.5 $\pm$ 1.4 & 1.3 $\pm$ 0.2 & 11.8 $\pm$ 2.4 & 0.5 $\pm$ 0.1 & 2.1 $\pm$ 0.8 & 3.4 $\pm$ 0.7 & 8.0 $\pm$ 8.6 & 19.5 $\pm$ 25.7 & 15.0 $\pm$ 1.6 & 1.5 $\pm$ 1.0 & 9.4 $\pm$ 3.9 \\
& SGD & 65.2 $\pm$ 0.1 & 68.8 $\pm$ 0.3 & 22.6 $\pm$ 2.6 & 80.7 $\pm$ 3.6 & 61.6 $\pm$ 3.0 & 80.2 $\pm$ 0.3 & 89.6 $\pm$ 0.6 & 83.5 $\pm$ 0.7 & 94.2 $\pm$ 0.5 & 55.6 $\pm$ 1.1 & 74.5 $\pm$ 0.6 & 70.6 $\pm$ 1.2 \\
\midrule
\multirow{15}{*}{4} 
& CoOp & 64.5 $\pm$ 0.2 & 68.6 $\pm$ 0.2 & 23.4 $\pm$ 0.8 & 67.7 $\pm$ 1.2 & 64.2 $\pm$ 0.7 & 79.7 $\pm$ 0.4 & 89.7 $\pm$ 0.3 & 81.6 $\pm$ 1.3 & 93.7 $\pm$ 0.4 & 55.8 $\pm$ 0.3 & 74.1 $\pm$ 0.9 & 69.4 $\pm$ 0.6 \\
& KgCoOp & 64.2 $\pm$ 0.1 & 66.6 $\pm$ 0.3 & 21.1 $\pm$ 0.7 & 52.7 $\pm$ 2.6 & 61.1 $\pm$ 0.1 & 80.9 $\pm$ 0.1 & 88.8 $\pm$ 0.6 & 67.4 $\pm$ 0.7 & 92.9 $\pm$ 0.2 & 51.0 $\pm$ 0.4 & 68.9 $\pm$ 0.4 & 65.0 $\pm$ 0.6 \\
& ProGrad & 62.3 $\pm$ 1.4 & 68.9 $\pm$ 0.1 & 22.1 $\pm$ 1.0 & 67.1 $\pm$ 2.9 & 61.5 $\pm$ 1.3 & 79.1 $\pm$ 0.8 & 89.7 $\pm$ 0.6 & 77.0 $\pm$ 5.7 & 92.6 $\pm$ 0.7 & 54.3 $\pm$ 0.6 & 72.7 $\pm$ 0.5 & 67.9 $\pm$ 1.4 \\
\cmidrule(lr){2-14}
& Adapter & 63.9 $\pm$ 0.1 & 65.8 $\pm$ 0.1 & 19.8 $\pm$ 0.1 & 40.6 $\pm$ 1.2 & 61.6 $\pm$ 0.0 & 80.2 $\pm$ 0.1 & 86.5 $\pm$ 0.2 & 66.0 $\pm$ 0.5 & 92.0 $\pm$ 0.1 & 46.7 $\pm$ 0.4 & 66.8 $\pm$ 0.1 & 62.7 $\pm$ 0.3 \\
& LP++ & 65.2 $\pm$ 0.0 & 69.2 $\pm$ 0.4 & 24.6 $\pm$ 0.5 & 63.0 $\pm$ 2.5 & 62.6 $\pm$ 0.3 & 77.6 $\pm$ 0.2 & 85.8 $\pm$ 1.6 & 88.9 $\pm$ 0.4 & 93.2 $\pm$ 0.1 & 59.2 $\pm$ 1.0 & 72.0 $\pm$ 0.7 & 69.2 $\pm$ 0.7 \\
& TIP-Adapter & 62.2 $\pm$ 0.0 & 62.3 $\pm$ 0.0 & 18.6 $\pm$ 0.1 & 38.7 $\pm$ 0.3 & 60.4 $\pm$ 0.0 & 78.9 $\pm$ 0.0 & 85.2 $\pm$ 0.0 & 65.3 $\pm$ 0.1 & 91.5 $\pm$ 0.1 & 43.0 $\pm$ 0.0 & 61.9 $\pm$ 0.0 & 60.7 $\pm$ 0.1 \\
\cmidrule(lr){2-14}
& LoRA & 61.8 $\pm$ 0.5 & 69.2 $\pm$ 0.1 & 25.7 $\pm$ 0.4 & 65.9 $\pm$ 0.7 & 67.8 $\pm$ 0.9 & 80.5 $\pm$ 0.1 & 91.0 $\pm$ 0.2 & 84.9 $\pm$ 0.2 & 94.7 $\pm$ 0.2 & 59.9 $\pm$ 0.7 & 77.1 $\pm$ 0.2 & 70.8 $\pm$ 0.4 \\
& PiSSA & 61.9 $\pm$ 0.3 & 68.2 $\pm$ 0.3 & 28.0 $\pm$ 0.9 & 82.3 $\pm$ 0.7 & 67.9 $\pm$ 0.3 & 77.8 $\pm$ 0.0 & 89.0 $\pm$ 0.8 & 88.6 $\pm$ 1.0 & 94.4 $\pm$ 0.6 & 60.4 $\pm$ 0.4 & 76.6 $\pm$ 1.2 & 72.3 $\pm$ 0.6 \\
& GaLoRE & 58.7 $\pm$ 0.4 & 66.8 $\pm$ 0.4 & 26.5 $\pm$ 1.8 & 81.5 $\pm$ 0.4 & 64.8 $\pm$ 0.5 & 75.8 $\pm$ 0.9 & 87.9 $\pm$ 0.9 & 86.9 $\pm$ 0.1 & 93.9 $\pm$ 0.2 & 59.3 $\pm$ 1.2 & 76.1 $\pm$ 1.4 & 70.7 $\pm$ 0.7 \\
\cmidrule(lr){2-14}
& BitFit & 65.4 $\pm$ 0.1 & 69.9 $\pm$ 0.3 & 25.0 $\pm$ 0.1 & 72.6 $\pm$ 0.7 & 65.0 $\pm$ 0.4 & 80.9 $\pm$ 0.3 & 90.9 $\pm$ 0.1 & 78.6 $\pm$ 0.1 & 94.5 $\pm$ 0.2 & 57.1 $\pm$ 0.5 & 74.7 $\pm$ 0.1 & 70.4 $\pm$ 0.3 \\
& LayerNorm & 65.0 $\pm$ 0.0 & 66.7 $\pm$ 0.1 & 20.6 $\pm$ 0.1 & 42.4 $\pm$ 0.4 & 62.4 $\pm$ 0.2 & 80.9 $\pm$ 0.1 & 87.1 $\pm$ 0.1 & 67.2 $\pm$ 0.1 & 92.7 $\pm$ 0.1 & 47.7 $\pm$ 0.3 & 66.2 $\pm$ 0.1 & 63.5 $\pm$ 0.1 \\
& SAFT & 62.6 $\pm$ 0.4 & 66.0 $\pm$ 0.3 & 24.6 $\pm$ 0.5 & 83.1 $\pm$ 1.6 & 61.9 $\pm$ 0.4 & 79.4 $\pm$ 0.4 & 90.4 $\pm$ 0.1 & 83.4 $\pm$ 1.0 & 94.7 $\pm$ 0.2 & 56.7 $\pm$ 0.3 & 73.5 $\pm$ 0.3 & 70.6 $\pm$ 0.5 \\
& SO & 62.4 $\pm$ 0.3 & 69.0 $\pm$ 0.1 & 26.5 $\pm$ 0.1 & 70.7 $\pm$ 0.8 & 68.3 $\pm$ 0.8 & 80.2 $\pm$ 0.1 & 90.7 $\pm$ 0.2 & 86.7 $\pm$ 0.6 & 94.7 $\pm$ 0.3 & 60.1 $\pm$ 0.6 & 77.2 $\pm$ 0.4 & 71.5 $\pm$ 0.4 \\
\cmidrule(lr){2-14}
& Adam & 22.4 $\pm$ 0.3 & 31.1 $\pm$ 0.7 & 1.0 $\pm$ 0.0 & 18.9 $\pm$ 1.0 & 0.7 $\pm$ 0.1 & 1.6 $\pm$ 0.7 & 3.9 $\pm$ 0.7 & 10.9 $\pm$ 7.7 & 5.8 $\pm$ 3.8 & 23.1 $\pm$ 1.0 & 6.4 $\pm$ 5.4 & 11.4 $\pm$ 1.9 \\
& SGD & 66.3 $\pm$ 0.1 & 71.1 $\pm$ 0.2 & 25.3 $\pm$ 3.3 & 84.5 $\pm$ 0.7 & 67.4 $\pm$ 0.8 & 80.5 $\pm$ 0.3 & 90.5 $\pm$ 0.1 & 87.9 $\pm$ 1.0 & 94.7 $\pm$ 0.2 & 60.0 $\pm$ 0.3 & 77.5 $\pm$ 0.7 & 73.2 $\pm$ 0.7 \\
\midrule
\multirow{15}{*}{8} 
& CoOp & 65.7 $\pm$ 0.2 & 70.7 $\pm$ 0.0 & 27.1 $\pm$ 0.3 & 73.5 $\pm$ 1.9 & 66.8 $\pm$ 0.2 & 80.5 $\pm$ 0.4 & 90.5 $\pm$ 0.5 & 87.0 $\pm$ 1.1 & 94.3 $\pm$ 0.2 & 60.1 $\pm$ 1.2 & 77.0 $\pm$ 0.5 & 72.1 $\pm$ 0.6 \\
& KgCoOp & 64.5 $\pm$ 0.1 & 67.2 $\pm$ 0.1 & 22.2 $\pm$ 0.5 & 57.3 $\pm$ 3.0 & 61.6 $\pm$ 0.1 & 81.2 $\pm$ 0.1 & 89.3 $\pm$ 0.6 & 68.4 $\pm$ 0.7 & 93.1 $\pm$ 0.2 & 53.4 $\pm$ 1.5 & 70.2 $\pm$ 0.5 & 66.2 $\pm$ 0.7 \\
& ProGrad & 63.2 $\pm$ 1.8 & 70.8 $\pm$ 0.5 & 23.7 $\pm$ 1.5 & 73.6 $\pm$ 0.6 & 64.8 $\pm$ 3.0 & 80.7 $\pm$ 0.3 & 90.1 $\pm$ 0.7 & 84.3 $\pm$ 2.5 & 91.9 $\pm$ 2.5 & 59.1 $\pm$ 0.3 & 75.8 $\pm$ 1.2 & 70.7 $\pm$ 1.3 \\
\cmidrule(lr){2-14}
& Adapter & 64.5 $\pm$ 0.1 & 67.5 $\pm$ 0.0 & 21.2 $\pm$ 0.3 & 44.5 $\pm$ 0.8 & 63.2 $\pm$ 0.3 & 81.2 $\pm$ 0.1 & 86.9 $\pm$ 0.2 & 69.2 $\pm$ 0.6 & 92.7 $\pm$ 0.1 & 50.3 $\pm$ 0.5 & 69.9 $\pm$ 0.4 & 64.6 $\pm$ 0.3 \\
& LP++ & 66.2 $\pm$ 0.2 & 71.6 $\pm$ 0.1 & 27.7 $\pm$ 0.8 & 72.5 $\pm$ 2.7 & 68.7 $\pm$ 0.6 & 78.7 $\pm$ 0.4 & 87.2 $\pm$ 0.8 & 90.9 $\pm$ 0.4 & 94.0 $\pm$ 0.4 & 62.8 $\pm$ 0.6 & 76.4 $\pm$ 0.5 & 72.4 $\pm$ 0.7 \\
& TIP-Adapter & 62.5 $\pm$ 0.0 & 62.9 $\pm$ 0.1 & 19.2 $\pm$ 0.1 & 40.4 $\pm$ 0.4 & 60.9 $\pm$ 0.1 & 78.9 $\pm$ 0.0 & 85.4 $\pm$ 0.0 & 68.7 $\pm$ 0.2 & 91.7 $\pm$ 0.1 & 43.5 $\pm$ 0.1 & 62.1 $\pm$ 0.1 & 61.5 $\pm$ 0.1 \\
\cmidrule(lr){2-14}
& LoRA & 63.4 $\pm$ 0.1 & 70.6 $\pm$ 0.5 & 32.5 $\pm$ 1.4 & 81.0 $\pm$ 0.5 & 73.9 $\pm$ 0.8 & 79.9 $\pm$ 0.4 & 91.6 $\pm$ 0.5 & 91.1 $\pm$ 0.4 & 95.2 $\pm$ 0.3 & 63.3 $\pm$ 1.2 & 79.1 $\pm$ 0.4 & 74.7 $\pm$ 0.6 \\
& PiSSA & 63.3 $\pm$ 0.2 & 69.8 $\pm$ 0.4 & 34.8 $\pm$ 0.8 & 87.5 $\pm$ 2.2 & 74.1 $\pm$ 0.6 & 78.2 $\pm$ 0.2 & 90.0 $\pm$ 0.4 & 91.5 $\pm$ 0.5 & 94.9 $\pm$ 0.3 & 63.6 $\pm$ 1.2 & 78.5 $\pm$ 1.4 & 75.1 $\pm$ 0.7 \\
& GaLoRE & 60.3 $\pm$ 0.4 & 69.4 $\pm$ 0.2 & 31.8 $\pm$ 1.4 & 87.9 $\pm$ 1.1 & 71.8 $\pm$ 0.3 & 75.8 $\pm$ 0.3 & 89.8 $\pm$ 0.5 & 90.9 $\pm$ 0.8 & 94.3 $\pm$ 0.2 & 62.6 $\pm$ 0.7 & 79.1 $\pm$ 0.7 & 74.0 $\pm$ 0.6 \\
\cmidrule(lr){2-14}
& BitFit & 66.3 $\pm$ 0.1 & 71.2 $\pm$ 0.1 & 28.8 $\pm$ 0.1 & 81.1 $\pm$ 1.4 & 68.2 $\pm$ 0.1 & 81.0 $\pm$ 0.2 & 91.2 $\pm$ 0.2 & 84.8 $\pm$ 0.2 & 95.0 $\pm$ 0.1 & 61.2 $\pm$ 1.0 & 77.2 $\pm$ 0.8 & 73.3 $\pm$ 0.4 \\
& LayerNorm & 65.6 $\pm$ 0.0 & 68.1 $\pm$ 0.1 & 21.5 $\pm$ 0.1 & 47.7 $\pm$ 0.8 & 63.3 $\pm$ 0.3 & 81.8 $\pm$ 0.1 & 87.5 $\pm$ 0.2 & 69.2 $\pm$ 0.1 & 93.4 $\pm$ 0.1 & 51.0 $\pm$ 0.4 & 68.5 $\pm$ 0.1 & 65.2 $\pm$ 0.2 \\
& SAFT & 64.2 $\pm$ 0.2 & 67.7 $\pm$ 0.3 & 28.9 $\pm$ 0.2 & 88.0 $\pm$ 2.0 & 65.3 $\pm$ 0.4 & 79.4 $\pm$ 0.4 & 90.5 $\pm$ 0.3 & 87.4 $\pm$ 0.5 & 94.8 $\pm$ 0.5 & 60.9 $\pm$ 1.1 & 75.7 $\pm$ 0.5 & 73.0 $\pm$ 0.6 \\
& SO & 63.7 $\pm$ 0.1 & 70.6 $\pm$ 0.5 & 33.4 $\pm$ 1.1 & 85.0 $\pm$ 0.5 & 74.0 $\pm$ 0.8 & 79.6 $\pm$ 0.3 & 91.6 $\pm$ 0.5 & 91.6 $\pm$ 0.4 & 95.2 $\pm$ 0.2 & 63.2 $\pm$ 1.1 & 79.2 $\pm$ 0.3 & 75.2 $\pm$ 0.5 \\
\cmidrule(lr){2-14}
& Adam & 27.3 $\pm$ 0.6 & 37.2 $\pm$ 0.2 & 1.6 $\pm$ 0.1 & 21.0 $\pm$ 8.8 & 1.2 $\pm$ 0.1 & 5.7 $\pm$ 4.4 & 10.2 $\pm$ 4.9 & 44.9 $\pm$ 4.0 & 74.7 $\pm$ 3.1 & 28.9 $\pm$ 3.2 & 19.5 $\pm$ 7.2 & 24.8 $\pm$ 3.3 \\
& SGD & 67.3 $\pm$ 0.1 & 72.9 $\pm$ 0.1 & 29.1 $\pm$ 2.4 & 86.2 $\pm$ 3.8 & 74.4 $\pm$ 0.4 & 80.8 $\pm$ 0.2 & 91.1 $\pm$ 0.2 & 91.9 $\pm$ 0.6 & 95.3 $\pm$ 0.3 & 65.1 $\pm$ 0.3 & 79.4 $\pm$ 0.3 & 75.8 $\pm$ 0.8 \\
\midrule
\multirow{15}{*}{16} 
& CoOp & 66.7 $\pm$ 0.1 & 72.6 $\pm$ 0.1 & 28.6 $\pm$ 0.8 & 78.3 $\pm$ 2.2 & 71.2 $\pm$ 0.9 & 80.9 $\pm$ 0.0 & 90.1 $\pm$ 0.1 & 91.4 $\pm$ 0.5 & 94.7 $\pm$ 0.1 & 64.6 $\pm$ 0.3 & 78.9 $\pm$ 0.5 & 74.4 $\pm$ 0.5 \\
& KgCoOp & 64.8 $\pm$ 0.0 & 67.9 $\pm$ 0.0 & 22.7 $\pm$ 0.3 & 58.3 $\pm$ 1.0 & 62.3 $\pm$ 0.2 & 81.3 $\pm$ 0.2 & 89.6 $\pm$ 0.3 & 70.2 $\pm$ 0.9 & 93.5 $\pm$ 0.1 & 56.5 $\pm$ 1.1 & 71.2 $\pm$ 0.1 & 67.1 $\pm$ 0.4 \\
& ProGrad & 56.3 $\pm$ 4.2 & 72.5 $\pm$ 0.1 & 28.0 $\pm$ 1.0 & 80.6 $\pm$ 0.6 & 70.6 $\pm$ 0.2 & 80.9 $\pm$ 0.3 & 90.9 $\pm$ 0.3 & 88.8 $\pm$ 1.0 & 94.7 $\pm$ 0.3 & 63.3 $\pm$ 0.8 & 78.6 $\pm$ 0.3 & 73.2 $\pm$ 0.8 \\
\cmidrule(lr){2-14}
& Adapter & 65.4 $\pm$ 0.1 & 69.5 $\pm$ 0.1 & 22.9 $\pm$ 0.4 & 49.9 $\pm$ 0.6 & 65.3 $\pm$ 0.1 & 81.5 $\pm$ 0.1 & 88.8 $\pm$ 0.2 & 74.5 $\pm$ 0.3 & 94.2 $\pm$ 0.1 & 56.4 $\pm$ 0.5 & 73.7 $\pm$ 0.2 & 67.5 $\pm$ 0.2 \\
& LP++ & 67.2 $\pm$ 0.1 & 73.3 $\pm$ 0.1 & 31.3 $\pm$ 0.7 & 80.0 $\pm$ 2.4 & 72.7 $\pm$ 0.4 & 80.6 $\pm$ 0.1 & 88.3 $\pm$ 0.6 & 93.3 $\pm$ 0.7 & 94.9 $\pm$ 0.3 & 68.1 $\pm$ 0.3 & 78.9 $\pm$ 0.5 & 75.3 $\pm$ 0.6 \\
& TIP-Adapter & 63.7 $\pm$ 0.0 & 65.9 $\pm$ 0.1 & 21.4 $\pm$ 0.3 & 49.9 $\pm$ 0.7 & 64.3 $\pm$ 0.1 & 79.2 $\pm$ 0.1 & 85.8 $\pm$ 0.1 & 82.4 $\pm$ 0.2 & 92.0 $\pm$ 0.1 & 45.0 $\pm$ 0.3 & 64.5 $\pm$ 0.1 & 64.9 $\pm$ 0.2 \\
\cmidrule(lr){2-14}
& LoRA & 65.8 $\pm$ 0.3 & 72.6 $\pm$ 0.3 & 40.8 $\pm$ 0.6 & 90.4 $\pm$ 0.5 & 79.2 $\pm$ 0.1 & 80.1 $\pm$ 0.1 & 91.7 $\pm$ 0.1 & 94.8 $\pm$ 0.7 & 95.9 $\pm$ 0.3 & 67.9 $\pm$ 0.6 & 82.5 $\pm$ 0.5 & 78.4 $\pm$ 0.4 \\
& PiSSA & 65.5 $\pm$ 0.1 & 72.0 $\pm$ 0.2 & 40.7 $\pm$ 0.5 & 90.3 $\pm$ 0.5 & 79.0 $\pm$ 0.5 & 79.0 $\pm$ 0.3 & 89.9 $\pm$ 0.5 & 94.9 $\pm$ 0.7 & 95.5 $\pm$ 0.0 & 68.2 $\pm$ 0.0 & 81.4 $\pm$ 1.1 & 77.8 $\pm$ 0.4 \\
& GaLoRE & 63.0 $\pm$ 0.1 & 71.0 $\pm$ 0.1 & 40.7 $\pm$ 1.5 & 92.1 $\pm$ 0.6 & 76.6 $\pm$ 0.3 & 76.6 $\pm$ 0.6 & 89.5 $\pm$ 0.4 & 94.4 $\pm$ 0.5 & 95.0 $\pm$ 0.4 & 66.7 $\pm$ 1.0 & 80.6 $\pm$ 0.6 & 76.9 $\pm$ 0.6 \\
\cmidrule(lr){2-14}
& BitFit & 67.4 $\pm$ 0.1 & 73.0 $\pm$ 0.1 & 31.9 $\pm$ 0.3 & 87.6 $\pm$ 0.4 & 71.6 $\pm$ 0.3 & 81.9 $\pm$ 0.1 & 91.9 $\pm$ 0.1 & 90.2 $\pm$ 0.4 & 95.9 $\pm$ 0.1 & 66.4 $\pm$ 0.4 & 79.5 $\pm$ 0.4 & 76.1 $\pm$ 0.3 \\
& LayerNorm & 66.4 $\pm$ 0.1 & 69.8 $\pm$ 0.0 & 23.4 $\pm$ 0.1 & 55.9 $\pm$ 0.8 & 65.0 $\pm$ 0.1 & 82.2 $\pm$ 0.1 & 89.7 $\pm$ 0.1 & 73.0 $\pm$ 0.3 & 94.2 $\pm$ 0.1 & 55.2 $\pm$ 0.6 & 71.6 $\pm$ 0.2 & 67.9 $\pm$ 0.2 \\
& SAFT & 66.3 $\pm$ 0.2 & 69.2 $\pm$ 0.5 & 35.4 $\pm$ 0.3 & 92.3 $\pm$ 0.7 & 69.9 $\pm$ 0.2 & 79.7 $\pm$ 0.3 & 90.6 $\pm$ 0.1 & 91.2 $\pm$ 0.3 & 95.3 $\pm$ 0.1 & 65.3 $\pm$ 0.5 & 78.2 $\pm$ 0.4 & 75.8 $\pm$ 0.3 \\
& SO & 66.1 $\pm$ 0.4 & 72.6 $\pm$ 0.2 & 41.8 $\pm$ 0.6 & 90.6 $\pm$ 0.7 & 79.5 $\pm$ 0.2 & 80.0 $\pm$ 0.1 & 91.6 $\pm$ 0.1 & 94.8 $\pm$ 0.8 & 95.8 $\pm$ 0.2 & 68.4 $\pm$ 0.5 & 82.7 $\pm$ 0.3 & 78.5 $\pm$ 0.4 \\
\cmidrule(lr){2-14}
& Adam & 31.9 $\pm$ 0.6 & 42.3 $\pm$ 0.9 & 2.1 $\pm$ 0.1 & 31.7 $\pm$ 1.6 & 5.6 $\pm$ 2.2 & 18.2 $\pm$ 10.9 & 25.2 $\pm$ 7.5 & 76.9 $\pm$ 2.6 & 80.1 $\pm$ 0.7 & 39.3 $\pm$ 2.2 & 31.8 $\pm$ 4.2 & 35.0 $\pm$ 3.0 \\
& SGD & 68.8 $\pm$ 0.1 & 74.4 $\pm$ 0.1 & 39.5 $\pm$ 0.3 & 92.2 $\pm$ 0.6 & 78.8 $\pm$ 0.7 & 81.8 $\pm$ 0.0 & 91.9 $\pm$ 0.2 & 95.1 $\pm$ 0.5 & 96.2 $\pm$ 0.1 & 68.7 $\pm$ 0.2 & 82.7 $\pm$ 0.1 & 79.1 $\pm$ 0.3 \\
\bottomrule
\end{tabular}
\end{adjustbox}
\end{table*} 
\begin{table*}
\caption{Few‑shot classification performance on 11 datasets with CLIP(ViT-B/16) backbone. Top‑1 accuracy (\%) averaged over 3 random seeds.}
\label{tab:few_shot_ViTB16_with_aug}
\centering
\footnotesize
\setlength{\tabcolsep}{1.8pt}
\renewcommand{\arraystretch}{1.00}
\begin{adjustbox}{max totalsize={\textwidth}{0.95\textheight},center}
\begin{tabular}{@{}c l *{12}{c}@{}}
\toprule
Shots & Method & ImageNet & SUN & Aircr. & EuroSAT & Cars & Food & Pets & Flow. & Caltech & DTD & UCF & Avg. \\ \midrule
\multirow{1}{*}{0} & CLIP & 66.7 & 62.6 & 23.9 & 42.2 & 65.5 & 85.2 & 88.3 & 67.4 & 93.3 & 44.4 & 65.2 & 64.1 \\
\midrule
\multirow{15}{*}{1} 
& CoOp & 67.3 $\pm$ 0.3 & 67.0 $\pm$ 0.6 & 28.1 $\pm$ 0.4 & 54.3 $\pm$ 6.9 & 66.7 $\pm$ 0.6 & 84.7 $\pm$ 0.5 & 89.3 $\pm$ 0.6 & 75.6 $\pm$ 1.6 & 93.1 $\pm$ 0.4 & 47.6 $\pm$ 1.6 & 72.3 $\pm$ 0.7 & 67.8 $\pm$ 1.3 \\
& KgCoOp & 68.6 $\pm$ 0.1 & 67.5 $\pm$ 0.4 & 25.7 $\pm$ 0.3 & 57.4 $\pm$ 2.7 & 65.0 $\pm$ 0.5 & 86.1 $\pm$ 0.1 & 90.9 $\pm$ 0.4 & 69.9 $\pm$ 0.4 & 93.4 $\pm$ 0.5 & 47.4 $\pm$ 1.0 & 71.0 $\pm$ 0.6 & 67.5 $\pm$ 0.6 \\
& ProGrad & 65.4 $\pm$ 1.8 & 65.7 $\pm$ 1.0 & 24.0 $\pm$ 2.6 & 56.1 $\pm$ 4.3 & 63.6 $\pm$ 4.0 & 83.2 $\pm$ 1.2 & 89.2 $\pm$ 1.9 & 75.3 $\pm$ 1.3 & 92.0 $\pm$ 1.3 & 48.0 $\pm$ 1.4 & 71.5 $\pm$ 0.7 & 66.7 $\pm$ 2.0 \\
\cmidrule(lr){2-14}
& Adapter & 67.6 $\pm$ 0.1 & 64.3 $\pm$ 0.1 & 24.1 $\pm$ 0.2 & 45.6 $\pm$ 0.7 & 66.2 $\pm$ 0.2 & 85.4 $\pm$ 0.1 & 89.2 $\pm$ 0.1 & 68.6 $\pm$ 0.4 & 93.4 $\pm$ 0.1 & 45.0 $\pm$ 0.1 & 66.3 $\pm$ 0.2 & 65.1 $\pm$ 0.2 \\
& LP++ & 66.3 $\pm$ 0.1 & 66.6 $\pm$ 0.2 & 26.7 $\pm$ 0.8 & 40.8 $\pm$ 2.4 & 62.6 $\pm$ 0.4 & 82.9 $\pm$ 1.3 & 86.1 $\pm$ 0.4 & 81.6 $\pm$ 0.8 & 93.3 $\pm$ 0.7 & 50.9 $\pm$ 1.2 & 71.1 $\pm$ 0.9 & 66.3 $\pm$ 0.8 \\
& TIP-Adapter & 66.7 $\pm$ 0.0 & 62.6 $\pm$ 0.0 & 23.9 $\pm$ 0.0 & 43.1 $\pm$ 0.1 & 65.6 $\pm$ 0.1 & 85.2 $\pm$ 0.0 & 88.3 $\pm$ 0.0 & 67.6 $\pm$ 0.1 & 93.3 $\pm$ 0.0 & 44.4 $\pm$ 0.0 & 65.3 $\pm$ 0.0 & 64.2 $\pm$ 0.0 \\
\cmidrule(lr){2-14}
& LoRA & 68.1 $\pm$ 0.6 & 69.6 $\pm$ 0.4 & 28.9 $\pm$ 0.3 & 65.7 $\pm$ 2.9 & 69.7 $\pm$ 0.1 & 86.2 $\pm$ 0.1 & 92.4 $\pm$ 0.1 & 74.0 $\pm$ 0.5 & 94.1 $\pm$ 0.1 & 52.3 $\pm$ 0.5 & 74.7 $\pm$ 0.4 & 70.5 $\pm$ 0.5 \\
& PiSSA & 66.0 $\pm$ 0.0 & 67.7 $\pm$ 0.7 & 29.9 $\pm$ 1.1 & 72.3 $\pm$ 3.0 & 69.7 $\pm$ 0.9 & 83.2 $\pm$ 0.7 & 90.5 $\pm$ 0.1 & 83.5 $\pm$ 0.5 & 93.4 $\pm$ 0.6 & 51.4 $\pm$ 0.8 & 75.4 $\pm$ 1.6 & 71.2 $\pm$ 0.9 \\
& GaLoRE & 64.1 $\pm$ 0.2 & 67.0 $\pm$ 0.3 & 27.7 $\pm$ 0.6 & 68.6 $\pm$ 9.3 & 62.4 $\pm$ 2.9 & 81.6 $\pm$ 0.8 & 90.1 $\pm$ 0.7 & 82.7 $\pm$ 1.0 & 92.7 $\pm$ 0.4 & 52.1 $\pm$ 0.6 & 75.0 $\pm$ 0.7 & 69.5 $\pm$ 1.6 \\
\cmidrule(lr){2-14}
& BitFit & 69.4 $\pm$ 0.2 & 69.3 $\pm$ 0.2 & 29.5 $\pm$ 0.3 & 72.0 $\pm$ 1.9 & 69.1 $\pm$ 0.4 & 85.5 $\pm$ 0.2 & 92.3 $\pm$ 0.3 & 74.8 $\pm$ 0.3 & 93.9 $\pm$ 0.2 & 52.1 $\pm$ 0.6 & 74.6 $\pm$ 0.4 & 71.1 $\pm$ 0.5 \\
& LayerNorm & 68.7 $\pm$ 0.1 & 66.7 $\pm$ 0.1 & 25.5 $\pm$ 0.2 & 49.6 $\pm$ 2.0 & 66.7 $\pm$ 0.2 & 85.7 $\pm$ 0.1 & 89.4 $\pm$ 0.1 & 69.9 $\pm$ 0.1 & 93.5 $\pm$ 0.1 & 45.9 $\pm$ 0.2 & 67.8 $\pm$ 0.2 & 66.3 $\pm$ 0.3 \\
& SAFT & 66.1 $\pm$ 0.1 & 67.2 $\pm$ 0.3 & 27.4 $\pm$ 0.8 & 68.9 $\pm$ 3.5 & 64.8 $\pm$ 0.7 & 84.1 $\pm$ 0.4 & 89.9 $\pm$ 0.9 & 80.7 $\pm$ 0.9 & 93.9 $\pm$ 0.2 & 52.2 $\pm$ 1.0 & 73.7 $\pm$ 0.4 & 69.9 $\pm$ 0.8 \\
& SO & 68.5 $\pm$ 0.4 & 69.4 $\pm$ 0.4 & 30.0 $\pm$ 0.5 & 69.6 $\pm$ 2.6 & 70.3 $\pm$ 0.2 & 85.9 $\pm$ 0.2 & 92.5 $\pm$ 0.1 & 76.1 $\pm$ 0.5 & 93.8 $\pm$ 0.2 & 53.6 $\pm$ 1.3 & 75.2 $\pm$ 0.6 & 71.4 $\pm$ 0.6 \\
\cmidrule(lr){2-14}
& Adam & 18.3 $\pm$ 0.9 & 19.5 $\pm$ 2.2 & 1.1 $\pm$ 0.1 & 11.7 $\pm$ 3.1 & 0.7 $\pm$ 0.1 & 1.4 $\pm$ 0.1 & 3.3 $\pm$ 0.6 & 1.4 $\pm$ 0.7 & 40.3 $\pm$ 6.1 & 13.5 $\pm$ 1.8 & 2.3 $\pm$ 1.8 & 10.3 $\pm$ 1.6 \\
& SGD & 70.1 $\pm$ 0.2 & 70.3 $\pm$ 0.2 & 30.5 $\pm$ 0.9 & 73.8 $\pm$ 2.7 & 70.8 $\pm$ 0.6 & 85.8 $\pm$ 0.2 & 92.7 $\pm$ 0.2 & 83.6 $\pm$ 0.6 & 94.0 $\pm$ 0.2 & 53.7 $\pm$ 0.8 & 76.1 $\pm$ 0.5 & 72.8 $\pm$ 0.6 \\
\midrule
\multirow{15}{*}{2} 
& CoOp & 68.5 $\pm$ 0.2 & 68.8 $\pm$ 0.4 & 30.9 $\pm$ 0.3 & 67.6 $\pm$ 2.3 & 68.1 $\pm$ 0.5 & 84.6 $\pm$ 0.3 & 91.0 $\pm$ 0.8 & 82.6 $\pm$ 2.2 & 94.1 $\pm$ 0.3 & 54.0 $\pm$ 1.2 & 74.9 $\pm$ 0.7 & 71.4 $\pm$ 0.8 \\
& KgCoOp & 68.9 $\pm$ 0.1 & 68.0 $\pm$ 0.4 & 26.3 $\pm$ 0.4 & 61.6 $\pm$ 2.9 & 65.6 $\pm$ 0.3 & 86.1 $\pm$ 0.1 & 91.3 $\pm$ 0.5 & 70.9 $\pm$ 0.3 & 93.8 $\pm$ 0.4 & 49.8 $\pm$ 0.5 & 72.3 $\pm$ 0.5 & 68.6 $\pm$ 0.6 \\
& ProGrad & 67.4 $\pm$ 0.6 & 69.0 $\pm$ 0.1 & 29.2 $\pm$ 1.3 & 66.7 $\pm$ 3.6 & 65.9 $\pm$ 1.3 & 83.9 $\pm$ 0.2 & 90.8 $\pm$ 0.3 & 80.8 $\pm$ 0.8 & 93.9 $\pm$ 0.3 & 53.3 $\pm$ 1.7 & 74.6 $\pm$ 0.6 & 70.5 $\pm$ 1.0 \\
\cmidrule(lr){2-14}
& Adapter & 68.3 $\pm$ 0.0 & 66.6 $\pm$ 0.1 & 24.5 $\pm$ 0.1 & 46.5 $\pm$ 0.4 & 66.3 $\pm$ 0.1 & 85.6 $\pm$ 0.1 & 89.1 $\pm$ 0.1 & 70.1 $\pm$ 0.7 & 93.6 $\pm$ 0.1 & 46.0 $\pm$ 0.1 & 67.2 $\pm$ 0.2 & 65.8 $\pm$ 0.2 \\
& LP++ & 68.4 $\pm$ 0.1 & 69.3 $\pm$ 0.2 & 29.0 $\pm$ 2.0 & 59.5 $\pm$ 6.4 & 66.2 $\pm$ 0.5 & 83.7 $\pm$ 0.5 & 88.0 $\pm$ 0.8 & 88.7 $\pm$ 0.6 & 93.8 $\pm$ 0.2 & 58.3 $\pm$ 1.1 & 73.4 $\pm$ 0.6 & 70.7 $\pm$ 1.2 \\
& TIP-Adapter & 66.8 $\pm$ 0.0 & 62.7 $\pm$ 0.0 & 24.0 $\pm$ 0.1 & 44.6 $\pm$ 0.3 & 65.6 $\pm$ 0.1 & 85.3 $\pm$ 0.0 & 88.3 $\pm$ 0.0 & 68.0 $\pm$ 0.1 & 93.3 $\pm$ 0.0 & 44.4 $\pm$ 0.0 & 65.3 $\pm$ 0.0 & 64.4 $\pm$ 0.1 \\
\cmidrule(lr){2-14}
& LoRA & 67.5 $\pm$ 0.3 & 70.7 $\pm$ 0.3 & 32.3 $\pm$ 0.6 & 75.2 $\pm$ 0.3 & 70.5 $\pm$ 0.3 & 85.7 $\pm$ 0.1 & 91.7 $\pm$ 0.8 & 82.4 $\pm$ 0.4 & 94.3 $\pm$ 0.5 & 58.0 $\pm$ 1.5 & 78.5 $\pm$ 0.8 & 73.3 $\pm$ 0.5 \\
& PiSSA & 66.7 $\pm$ 0.3 & 69.4 $\pm$ 0.3 & 33.9 $\pm$ 1.2 & 80.8 $\pm$ 1.4 & 71.3 $\pm$ 1.1 & 83.0 $\pm$ 0.8 & 91.1 $\pm$ 0.1 & 89.4 $\pm$ 0.5 & 94.8 $\pm$ 0.4 & 59.4 $\pm$ 0.9 & 78.8 $\pm$ 0.4 & 74.4 $\pm$ 0.7 \\
& GaLoRE & 64.3 $\pm$ 0.9 & 68.7 $\pm$ 0.1 & 31.6 $\pm$ 0.8 & 80.3 $\pm$ 1.6 & 70.5 $\pm$ 0.4 & 82.4 $\pm$ 0.4 & 88.8 $\pm$ 1.3 & 87.8 $\pm$ 0.4 & 93.6 $\pm$ 0.5 & 58.2 $\pm$ 1.0 & 77.6 $\pm$ 0.9 & 73.1 $\pm$ 0.7 \\
\cmidrule(lr){2-14}
& BitFit & 70.0 $\pm$ 0.1 & 70.5 $\pm$ 0.3 & 31.3 $\pm$ 1.0 & 78.0 $\pm$ 2.1 & 69.4 $\pm$ 0.5 & 85.7 $\pm$ 0.3 & 92.0 $\pm$ 0.8 & 79.8 $\pm$ 0.4 & 94.5 $\pm$ 0.3 & 56.5 $\pm$ 0.9 & 77.0 $\pm$ 0.4 & 73.2 $\pm$ 0.6 \\
& LayerNorm & 69.4 $\pm$ 0.0 & 68.0 $\pm$ 0.0 & 26.3 $\pm$ 0.2 & 52.4 $\pm$ 0.7 & 67.1 $\pm$ 0.1 & 86.1 $\pm$ 0.1 & 89.4 $\pm$ 0.0 & 71.2 $\pm$ 0.2 & 93.8 $\pm$ 0.0 & 48.1 $\pm$ 0.4 & 69.5 $\pm$ 0.1 & 67.4 $\pm$ 0.2 \\
& SAFT & 66.9 $\pm$ 0.3 & 68.8 $\pm$ 0.3 & 31.1 $\pm$ 0.9 & 78.7 $\pm$ 2.7 & 67.8 $\pm$ 0.2 & 84.8 $\pm$ 0.2 & 92.5 $\pm$ 0.6 & 85.0 $\pm$ 0.5 & 94.6 $\pm$ 0.1 & 57.4 $\pm$ 2.1 & 76.0 $\pm$ 0.3 & 73.1 $\pm$ 0.7 \\
& SO & 67.9 $\pm$ 0.2 & 70.7 $\pm$ 0.3 & 33.3 $\pm$ 0.8 & 78.8 $\pm$ 1.9 & 71.1 $\pm$ 0.3 & 85.3 $\pm$ 0.1 & 91.9 $\pm$ 1.0 & 85.0 $\pm$ 0.5 & 94.3 $\pm$ 0.4 & 58.6 $\pm$ 0.6 & 79.0 $\pm$ 0.8 & 74.2 $\pm$ 0.6 \\
\cmidrule(lr){2-14}
& Adam & 23.0 $\pm$ 1.0 & 23.4 $\pm$ 2.7 & 1.0 $\pm$ 0.2 & 21.6 $\pm$ 5.3 & 0.6 $\pm$ 0.1 & 1.7 $\pm$ 0.4 & 3.3 $\pm$ 0.2 & 13.3 $\pm$ 8.7 & 12.3 $\pm$ 15.8 & 15.0 $\pm$ 6.8 & 6.4 $\pm$ 7.0 & 11.0 $\pm$ 4.4 \\
& SGD & 70.5 $\pm$ 0.2 & 71.8 $\pm$ 0.4 & 33.5 $\pm$ 1.4 & 82.4 $\pm$ 2.3 & 71.4 $\pm$ 0.6 & 86.0 $\pm$ 0.2 & 92.5 $\pm$ 0.6 & 89.7 $\pm$ 0.3 & 94.4 $\pm$ 0.2 & 59.4 $\pm$ 1.0 & 79.0 $\pm$ 1.0 & 75.5 $\pm$ 0.7 \\
\midrule
\multirow{15}{*}{4} 
& CoOp & 69.6 $\pm$ 0.2 & 70.8 $\pm$ 0.3 & 32.3 $\pm$ 0.6 & 73.2 $\pm$ 2.0 & 71.4 $\pm$ 0.3 & 85.9 $\pm$ 0.4 & 92.6 $\pm$ 0.5 & 86.4 $\pm$ 1.1 & 94.9 $\pm$ 0.1 & 56.8 $\pm$ 1.3 & 76.9 $\pm$ 0.5 & 73.7 $\pm$ 0.7 \\
& KgCoOp & 69.2 $\pm$ 0.1 & 68.7 $\pm$ 0.3 & 27.3 $\pm$ 0.3 & 60.5 $\pm$ 4.0 & 66.1 $\pm$ 0.1 & 86.3 $\pm$ 0.1 & 91.8 $\pm$ 0.4 & 71.9 $\pm$ 0.5 & 94.3 $\pm$ 0.0 & 52.0 $\pm$ 0.4 & 73.1 $\pm$ 0.7 & 69.2 $\pm$ 0.6 \\
& ProGrad & 68.6 $\pm$ 0.3 & 71.1 $\pm$ 0.6 & 30.0 $\pm$ 0.8 & 68.5 $\pm$ 1.5 & 68.7 $\pm$ 3.4 & 85.6 $\pm$ 0.3 & 92.0 $\pm$ 0.5 & 85.4 $\pm$ 0.9 & 94.3 $\pm$ 0.2 & 57.5 $\pm$ 0.1 & 76.2 $\pm$ 0.4 & 72.5 $\pm$ 0.8 \\
\cmidrule(lr){2-14}
& Adapter & 68.8 $\pm$ 0.1 & 68.2 $\pm$ 0.1 & 25.6 $\pm$ 0.1 & 47.8 $\pm$ 0.8 & 67.4 $\pm$ 0.1 & 85.8 $\pm$ 0.1 & 89.5 $\pm$ 0.1 & 71.1 $\pm$ 0.3 & 93.8 $\pm$ 0.0 & 47.6 $\pm$ 0.2 & 70.0 $\pm$ 0.3 & 66.9 $\pm$ 0.2 \\
& LP++ & 70.1 $\pm$ 0.1 & 71.9 $\pm$ 0.2 & 32.6 $\pm$ 0.4 & 65.2 $\pm$ 4.8 & 71.2 $\pm$ 0.5 & 84.5 $\pm$ 0.2 & 89.9 $\pm$ 0.3 & 92.5 $\pm$ 1.2 & 94.8 $\pm$ 0.3 & 63.3 $\pm$ 1.3 & 76.5 $\pm$ 1.0 & 73.9 $\pm$ 0.9 \\
& TIP-Adapter & 66.9 $\pm$ 0.0 & 62.9 $\pm$ 0.0 & 24.1 $\pm$ 0.1 & 47.5 $\pm$ 0.5 & 65.9 $\pm$ 0.0 & 85.3 $\pm$ 0.0 & 88.3 $\pm$ 0.0 & 69.0 $\pm$ 0.1 & 93.4 $\pm$ 0.1 & 44.5 $\pm$ 0.1 & 65.3 $\pm$ 0.0 & 64.8 $\pm$ 0.1 \\
\cmidrule(lr){2-14}
& LoRA & 68.0 $\pm$ 0.1 & 71.9 $\pm$ 0.2 & 35.5 $\pm$ 0.5 & 73.5 $\pm$ 1.4 & 76.1 $\pm$ 0.3 & 86.0 $\pm$ 0.4 & 93.1 $\pm$ 0.1 & 89.4 $\pm$ 0.3 & 95.3 $\pm$ 0.4 & 63.3 $\pm$ 0.4 & 82.1 $\pm$ 0.5 & 75.8 $\pm$ 0.4 \\
& PiSSA & 67.5 $\pm$ 0.4 & 70.9 $\pm$ 0.5 & 36.3 $\pm$ 0.7 & 85.1 $\pm$ 2.1 & 76.1 $\pm$ 0.7 & 83.5 $\pm$ 0.9 & 91.9 $\pm$ 0.3 & 91.9 $\pm$ 0.0 & 95.6 $\pm$ 0.2 & 65.0 $\pm$ 0.5 & 80.6 $\pm$ 0.6 & 76.8 $\pm$ 0.6 \\
& GaLoRE & 65.6 $\pm$ 0.4 & 70.1 $\pm$ 0.4 & 37.7 $\pm$ 0.4 & 86.6 $\pm$ 1.1 & 74.7 $\pm$ 0.6 & 82.7 $\pm$ 0.2 & 91.6 $\pm$ 0.3 & 91.2 $\pm$ 0.3 & 94.7 $\pm$ 0.5 & 62.8 $\pm$ 1.0 & 80.1 $\pm$ 1.0 & 76.2 $\pm$ 0.6 \\
\cmidrule(lr){2-14}
& BitFit & 70.8 $\pm$ 0.0 & 72.2 $\pm$ 0.2 & 33.5 $\pm$ 0.1 & 78.7 $\pm$ 0.6 & 72.5 $\pm$ 0.3 & 86.3 $\pm$ 0.2 & 93.0 $\pm$ 0.2 & 85.1 $\pm$ 0.5 & 95.3 $\pm$ 0.3 & 60.9 $\pm$ 0.3 & 79.4 $\pm$ 0.5 & 75.2 $\pm$ 0.3 \\
& LayerNorm & 70.1 $\pm$ 0.0 & 69.2 $\pm$ 0.1 & 27.3 $\pm$ 0.2 & 53.4 $\pm$ 0.5 & 68.3 $\pm$ 0.1 & 86.4 $\pm$ 0.0 & 89.7 $\pm$ 0.1 & 72.7 $\pm$ 0.1 & 93.9 $\pm$ 0.2 & 49.9 $\pm$ 0.4 & 71.3 $\pm$ 0.1 & 68.4 $\pm$ 0.2 \\
& SAFT & 68.2 $\pm$ 0.2 & 69.5 $\pm$ 0.2 & 32.8 $\pm$ 0.9 & 83.6 $\pm$ 1.3 & 70.0 $\pm$ 0.2 & 85.5 $\pm$ 0.2 & 93.2 $\pm$ 0.4 & 88.7 $\pm$ 0.3 & 95.2 $\pm$ 0.3 & 61.6 $\pm$ 0.8 & 77.9 $\pm$ 0.9 & 75.1 $\pm$ 0.5 \\
& SO & 68.2 $\pm$ 0.2 & 71.8 $\pm$ 0.3 & 36.2 $\pm$ 0.6 & 78.3 $\pm$ 0.9 & 76.8 $\pm$ 0.2 & 85.9 $\pm$ 0.4 & 93.2 $\pm$ 0.2 & 90.8 $\pm$ 0.7 & 95.3 $\pm$ 0.4 & 63.7 $\pm$ 0.6 & 82.1 $\pm$ 0.7 & 76.6 $\pm$ 0.5 \\
\cmidrule(lr){2-14}
& Adam & 28.8 $\pm$ 0.7 & 31.0 $\pm$ 6.0 & 1.1 $\pm$ 0.1 & 15.8 $\pm$ 3.5 & 0.7 $\pm$ 0.2 & 0.9 $\pm$ 0.1 & 2.7 $\pm$ 0.1 & 8.3 $\pm$ 2.8 & 18.5 $\pm$ 16.5 & 16.0 $\pm$ 7.7 & 2.1 $\pm$ 1.6 & 11.4 $\pm$ 3.6 \\
& SGD & 71.4 $\pm$ 0.2 & 73.7 $\pm$ 0.2 & 35.5 $\pm$ 2.0 & 84.4 $\pm$ 1.5 & 77.2 $\pm$ 0.1 & 86.1 $\pm$ 0.0 & 93.0 $\pm$ 0.4 & 93.3 $\pm$ 0.5 & 95.2 $\pm$ 0.4 & 64.4 $\pm$ 0.9 & 81.8 $\pm$ 1.0 & 77.8 $\pm$ 0.7 \\
\midrule
\multirow{15}{*}{8} 
& CoOp & 70.6 $\pm$ 0.1 & 72.8 $\pm$ 0.4 & 34.8 $\pm$ 0.6 & 76.9 $\pm$ 1.3 & 74.0 $\pm$ 0.5 & 85.6 $\pm$ 0.5 & 92.6 $\pm$ 0.3 & 91.9 $\pm$ 0.3 & 95.0 $\pm$ 0.3 & 62.7 $\pm$ 1.4 & 79.6 $\pm$ 0.4 & 76.0 $\pm$ 0.6 \\
& KgCoOp & 69.5 $\pm$ 0.0 & 69.4 $\pm$ 0.2 & 28.0 $\pm$ 0.3 & 64.3 $\pm$ 2.4 & 66.7 $\pm$ 0.3 & 86.5 $\pm$ 0.1 & 92.2 $\pm$ 0.4 & 73.5 $\pm$ 0.7 & 94.2 $\pm$ 0.0 & 55.6 $\pm$ 0.3 & 73.9 $\pm$ 0.5 & 70.3 $\pm$ 0.5 \\
& ProGrad & 69.6 $\pm$ 0.4 & 73.3 $\pm$ 0.2 & 34.8 $\pm$ 0.2 & 76.7 $\pm$ 0.9 & 73.5 $\pm$ 1.9 & 86.1 $\pm$ 0.2 & 93.2 $\pm$ 0.3 & 89.9 $\pm$ 0.8 & 95.0 $\pm$ 0.2 & 61.3 $\pm$ 1.2 & 78.6 $\pm$ 0.5 & 75.6 $\pm$ 0.6 \\
\cmidrule(lr){2-14}
& Adapter & 69.5 $\pm$ 0.1 & 69.7 $\pm$ 0.1 & 27.3 $\pm$ 0.2 & 54.0 $\pm$ 0.9 & 69.0 $\pm$ 0.1 & 86.4 $\pm$ 0.1 & 89.7 $\pm$ 0.2 & 72.7 $\pm$ 0.4 & 94.2 $\pm$ 0.2 & 50.5 $\pm$ 0.2 & 73.7 $\pm$ 0.2 & 68.8 $\pm$ 0.2 \\
& LP++ & 71.2 $\pm$ 0.1 & 74.2 $\pm$ 0.2 & 35.8 $\pm$ 0.9 & 77.5 $\pm$ 0.1 & 75.5 $\pm$ 0.3 & 85.3 $\pm$ 0.1 & 90.1 $\pm$ 0.7 & 94.2 $\pm$ 0.6 & 95.2 $\pm$ 0.3 & 66.4 $\pm$ 1.5 & 79.9 $\pm$ 0.7 & 76.8 $\pm$ 0.5 \\
& TIP-Adapter & 67.4 $\pm$ 0.0 & 63.6 $\pm$ 0.1 & 24.7 $\pm$ 0.2 & 51.2 $\pm$ 0.3 & 66.7 $\pm$ 0.0 & 85.3 $\pm$ 0.0 & 88.6 $\pm$ 0.0 & 73.3 $\pm$ 0.2 & 93.5 $\pm$ 0.1 & 44.9 $\pm$ 0.1 & 65.7 $\pm$ 0.2 & 65.9 $\pm$ 0.1 \\
\cmidrule(lr){2-14}
& LoRA & 69.1 $\pm$ 0.2 & 73.3 $\pm$ 0.4 & 43.7 $\pm$ 0.5 & 84.6 $\pm$ 1.2 & 81.4 $\pm$ 0.2 & 85.3 $\pm$ 0.2 & 93.5 $\pm$ 0.1 & 94.3 $\pm$ 0.1 & 95.7 $\pm$ 0.1 & 67.0 $\pm$ 1.2 & 83.7 $\pm$ 0.2 & 79.2 $\pm$ 0.4 \\
& PiSSA & 69.0 $\pm$ 0.2 & 72.5 $\pm$ 0.2 & 43.5 $\pm$ 0.3 & 87.7 $\pm$ 0.9 & 82.0 $\pm$ 0.1 & 84.0 $\pm$ 0.0 & 93.4 $\pm$ 0.2 & 94.7 $\pm$ 0.5 & 95.5 $\pm$ 0.2 & 66.6 $\pm$ 1.4 & 82.8 $\pm$ 0.3 & 79.2 $\pm$ 0.4 \\
& GaLoRE & 66.5 $\pm$ 0.2 & 72.0 $\pm$ 0.6 & 43.6 $\pm$ 1.8 & 88.5 $\pm$ 2.2 & 80.1 $\pm$ 0.4 & 82.9 $\pm$ 0.2 & 92.6 $\pm$ 0.3 & 93.5 $\pm$ 1.0 & 95.3 $\pm$ 0.1 & 65.3 $\pm$ 0.8 & 82.5 $\pm$ 0.6 & 78.4 $\pm$ 0.7 \\
\cmidrule(lr){2-14}
& BitFit & 71.6 $\pm$ 0.2 & 73.8 $\pm$ 0.1 & 37.1 $\pm$ 0.9 & 84.1 $\pm$ 1.8 & 76.1 $\pm$ 0.6 & 86.5 $\pm$ 0.1 & 93.4 $\pm$ 0.1 & 90.6 $\pm$ 0.6 & 95.4 $\pm$ 0.2 & 64.4 $\pm$ 1.5 & 81.8 $\pm$ 0.5 & 77.7 $\pm$ 0.6 \\
& LayerNorm & 70.7 $\pm$ 0.1 & 70.7 $\pm$ 0.0 & 28.8 $\pm$ 0.1 & 60.2 $\pm$ 0.6 & 69.7 $\pm$ 0.1 & 87.1 $\pm$ 0.1 & 90.5 $\pm$ 0.2 & 74.5 $\pm$ 0.2 & 94.6 $\pm$ 0.1 & 52.6 $\pm$ 0.2 & 73.7 $\pm$ 0.2 & 70.3 $\pm$ 0.2 \\
& SAFT & 69.6 $\pm$ 0.3 & 70.7 $\pm$ 0.2 & 36.3 $\pm$ 0.9 & 88.9 $\pm$ 0.9 & 73.5 $\pm$ 0.5 & 85.7 $\pm$ 0.3 & 93.3 $\pm$ 0.1 & 91.8 $\pm$ 0.3 & 95.8 $\pm$ 0.2 & 65.4 $\pm$ 1.4 & 79.6 $\pm$ 0.5 & 77.3 $\pm$ 0.5 \\
& SO & 69.4 $\pm$ 0.1 & 73.0 $\pm$ 0.3 & 44.0 $\pm$ 0.9 & 86.7 $\pm$ 1.7 & 81.7 $\pm$ 0.2 & 85.1 $\pm$ 0.1 & 93.7 $\pm$ 0.0 & 94.6 $\pm$ 0.2 & 95.8 $\pm$ 0.3 & 67.1 $\pm$ 1.1 & 83.8 $\pm$ 0.2 & 79.5 $\pm$ 0.5 \\
\cmidrule(lr){2-14}
& Adam & 34.2 $\pm$ 0.9 & 41.8 $\pm$ 0.8 & 1.6 $\pm$ 0.1 & 24.7 $\pm$ 4.6 & 13.3 $\pm$ 15.4 & 5.8 $\pm$ 5.1 & 3.2 $\pm$ 0.7 & 23.3 $\pm$ 23.9 & 55.4 $\pm$ 27.7 & 31.6 $\pm$ 1.5 & 26.8 $\pm$ 13.4 & 23.8 $\pm$ 8.5 \\
& SGD & 72.4 $\pm$ 0.3 & 75.2 $\pm$ 0.3 & 42.0 $\pm$ 0.8 & 85.3 $\pm$ 3.5 & 81.2 $\pm$ 1.2 & 86.5 $\pm$ 0.1 & 93.3 $\pm$ 0.2 & 95.2 $\pm$ 0.4 & 95.7 $\pm$ 0.1 & 68.1 $\pm$ 0.9 & 84.2 $\pm$ 0.2 & 79.9 $\pm$ 0.7 \\
\midrule
\multirow{15}{*}{16} 
& CoOp & 71.7 $\pm$ 0.1 & 74.6 $\pm$ 0.3 & 38.4 $\pm$ 0.5 & 78.5 $\pm$ 0.5 & 77.9 $\pm$ 0.4 & 86.6 $\pm$ 0.2 & 93.4 $\pm$ 0.4 & 95.0 $\pm$ 0.6 & 95.2 $\pm$ 0.6 & 67.7 $\pm$ 0.5 & 83.3 $\pm$ 0.2 & 78.4 $\pm$ 0.4 \\
& KgCoOp & 69.7 $\pm$ 0.0 & 70.1 $\pm$ 0.2 & 29.1 $\pm$ 0.5 & 67.6 $\pm$ 1.1 & 67.4 $\pm$ 0.1 & 86.6 $\pm$ 0.1 & 92.6 $\pm$ 0.3 & 77.0 $\pm$ 1.0 & 94.3 $\pm$ 0.0 & 59.7 $\pm$ 1.1 & 75.5 $\pm$ 0.3 & 71.8 $\pm$ 0.4 \\
& ProGrad & 70.2 $\pm$ 1.8 & 74.7 $\pm$ 0.4 & 38.0 $\pm$ 0.1 & 79.6 $\pm$ 1.7 & 76.1 $\pm$ 1.8 & 86.5 $\pm$ 0.3 & 93.2 $\pm$ 0.3 & 93.0 $\pm$ 0.9 & 95.2 $\pm$ 0.4 & 65.3 $\pm$ 0.5 & 81.5 $\pm$ 0.4 & 77.6 $\pm$ 0.8 \\
\cmidrule(lr){2-14}
& Adapter & 70.3 $\pm$ 0.1 & 71.8 $\pm$ 0.1 & 29.2 $\pm$ 0.5 & 60.2 $\pm$ 1.0 & 71.1 $\pm$ 0.1 & 86.8 $\pm$ 0.0 & 92.0 $\pm$ 0.0 & 80.3 $\pm$ 0.2 & 94.9 $\pm$ 0.1 & 57.7 $\pm$ 0.4 & 78.4 $\pm$ 0.3 & 72.1 $\pm$ 0.3 \\
& LP++ & 72.1 $\pm$ 0.1 & 75.7 $\pm$ 0.1 & 39.7 $\pm$ 0.2 & 82.0 $\pm$ 0.8 & 78.9 $\pm$ 0.1 & 86.3 $\pm$ 0.3 & 91.9 $\pm$ 0.2 & 96.1 $\pm$ 0.2 & 95.8 $\pm$ 0.5 & 71.2 $\pm$ 0.8 & 82.7 $\pm$ 0.3 & 79.3 $\pm$ 0.3 \\
& TIP-Adapter & 68.9 $\pm$ 0.1 & 67.3 $\pm$ 0.1 & 28.7 $\pm$ 0.1 & 57.3 $\pm$ 0.3 & 70.9 $\pm$ 0.1 & 85.5 $\pm$ 0.1 & 89.2 $\pm$ 0.1 & 87.0 $\pm$ 0.3 & 93.7 $\pm$ 0.1 & 46.8 $\pm$ 0.3 & 68.2 $\pm$ 0.4 & 69.4 $\pm$ 0.2 \\
\cmidrule(lr){2-14}
& LoRA & 71.3 $\pm$ 0.2 & 74.8 $\pm$ 0.4 & 50.4 $\pm$ 1.0 & 90.3 $\pm$ 0.5 & 85.4 $\pm$ 0.3 & 85.8 $\pm$ 0.1 & 94.2 $\pm$ 0.3 & 97.0 $\pm$ 0.4 & 96.2 $\pm$ 0.2 & 71.1 $\pm$ 1.5 & 85.7 $\pm$ 0.5 & 82.0 $\pm$ 0.5 \\
& PiSSA & 70.9 $\pm$ 0.2 & 74.4 $\pm$ 0.2 & 51.3 $\pm$ 0.9 & 92.1 $\pm$ 1.0 & 85.6 $\pm$ 0.5 & 84.9 $\pm$ 0.2 & 93.4 $\pm$ 0.3 & 97.1 $\pm$ 0.1 & 96.1 $\pm$ 0.1 & 71.6 $\pm$ 0.8 & 85.5 $\pm$ 0.4 & 82.1 $\pm$ 0.4 \\
& GaLoRE & 68.7 $\pm$ 0.1 & 73.5 $\pm$ 0.2 & 50.4 $\pm$ 0.8 & 91.0 $\pm$ 1.6 & 83.7 $\pm$ 0.3 & 83.3 $\pm$ 0.4 & 92.9 $\pm$ 0.1 & 96.5 $\pm$ 0.4 & 95.7 $\pm$ 0.2 & 69.7 $\pm$ 0.3 & 85.4 $\pm$ 0.3 & 81.0 $\pm$ 0.4 \\
\cmidrule(lr){2-14}
& BitFit & 72.6 $\pm$ 0.0 & 75.5 $\pm$ 0.1 & 42.6 $\pm$ 0.3 & 88.5 $\pm$ 0.8 & 79.6 $\pm$ 0.2 & 87.0 $\pm$ 0.0 & 94.0 $\pm$ 0.2 & 94.5 $\pm$ 0.4 & 95.6 $\pm$ 0.3 & 68.3 $\pm$ 1.5 & 84.0 $\pm$ 0.1 & 80.2 $\pm$ 0.4 \\
& LayerNorm & 71.4 $\pm$ 0.0 & 72.2 $\pm$ 0.1 & 31.4 $\pm$ 0.2 & 67.8 $\pm$ 0.5 & 71.8 $\pm$ 0.1 & 87.4 $\pm$ 0.0 & 92.9 $\pm$ 0.1 & 78.2 $\pm$ 0.2 & 94.9 $\pm$ 0.0 & 56.5 $\pm$ 0.2 & 76.9 $\pm$ 0.1 & 72.8 $\pm$ 0.1 \\
& SAFT & 71.4 $\pm$ 0.1 & 72.2 $\pm$ 0.1 & 41.8 $\pm$ 0.2 & 93.1 $\pm$ 0.5 & 77.1 $\pm$ 0.1 & 85.9 $\pm$ 0.3 & 93.6 $\pm$ 0.4 & 95.1 $\pm$ 0.1 & 95.9 $\pm$ 0.2 & 69.6 $\pm$ 0.7 & 82.5 $\pm$ 0.2 & 79.8 $\pm$ 0.3 \\
& SO & 71.3 $\pm$ 0.2 & 74.7 $\pm$ 0.4 & 51.5 $\pm$ 1.1 & 90.6 $\pm$ 1.2 & 85.5 $\pm$ 0.3 & 85.7 $\pm$ 0.1 & 94.2 $\pm$ 0.1 & 97.2 $\pm$ 0.4 & 96.2 $\pm$ 0.1 & 71.1 $\pm$ 1.4 & 85.7 $\pm$ 0.4 & 82.2 $\pm$ 0.5 \\
\cmidrule(lr){2-14}
& Adam & 39.7 $\pm$ 0.2 & 48.5 $\pm$ 0.3 & 4.5 $\pm$ 1.0 & 34.8 $\pm$ 16.1 & 7.6 $\pm$ 5.5 & 12.8 $\pm$ 8.5 & 8.7 $\pm$ 8.4 & 66.0 $\pm$ 8.0 & 59.0 $\pm$ 36.5 & 40.4 $\pm$ 1.3 & 20.4 $\pm$ 10.6 & 31.1 $\pm$ 8.8 \\
& SGD & 73.9 $\pm$ 0.1 & 76.7 $\pm$ 0.1 & 51.2 $\pm$ 0.9 & 92.2 $\pm$ 0.7 & 85.4 $\pm$ 0.1 & 87.2 $\pm$ 0.1 & 94.4 $\pm$ 0.2 & 97.4 $\pm$ 0.2 & 96.2 $\pm$ 0.1 & 72.0 $\pm$ 0.8 & 86.6 $\pm$ 0.2 & 83.0 $\pm$ 0.3 \\
\bottomrule
\end{tabular}
\end{adjustbox}
\end{table*} 
\begin{table*}
\caption{Few‑shot classification performance on 11 datasets with CLIP(ViT-L/14) backbone. Top‑1 accuracy (\%) averaged over 3 random seeds.}
\label{tab:few_shot_ViTL14_with_aug}
\centering
\footnotesize
\setlength{\tabcolsep}{1.8pt}
\renewcommand{\arraystretch}{1.00}
\begin{adjustbox}{max totalsize={\textwidth}{0.95\textheight},center}
\begin{tabular}{@{}c l *{12}{c}@{}}
\toprule
Shots & Method & ImageNet & SUN & Aircr. & EuroSAT & Cars & Food & Pets & Flow. & Caltech & DTD & UCF & Avg. \\ \midrule
\multirow{1}{*}{0} & CLIP & 73.5 & 67.7 & 30.0 & 55.1 & 76.8 & 90.3 & 93.1 & 76.2 & 95.2 & 52.4 & 73.7 & 71.3 \\
\midrule
\multirow{15}{*}{1} 
& CoOp & 71.2 $\pm$ 0.6 & 68.8 $\pm$ 0.6 & 37.2 $\pm$ 0.4 & 58.8 $\pm$ 3.3 & 78.1 $\pm$ 0.9 & 88.9 $\pm$ 0.7 & 93.4 $\pm$ 0.3 & 89.3 $\pm$ 1.0 & 94.5 $\pm$ 0.3 & 53.8 $\pm$ 1.6 & 77.6 $\pm$ 0.4 & 73.8 $\pm$ 0.9 \\
& KgCoOp & 75.6 $\pm$ 0.0 & 71.6 $\pm$ 0.2 & 34.1 $\pm$ 0.8 & 69.1 $\pm$ 2.0 & 77.7 $\pm$ 0.7 & 90.9 $\pm$ 0.2 & 93.8 $\pm$ 0.1 & 80.9 $\pm$ 0.5 & 95.3 $\pm$ 0.6 & 56.6 $\pm$ 0.4 & 76.8 $\pm$ 0.9 & 74.8 $\pm$ 0.6 \\
& ProGrad & 70.9 $\pm$ 0.6 & 69.2 $\pm$ 0.9 & 35.3 $\pm$ 1.0 & 59.3 $\pm$ 4.4 & 72.7 $\pm$ 6.5 & 88.6 $\pm$ 0.1 & 93.4 $\pm$ 0.2 & 86.2 $\pm$ 1.5 & 94.2 $\pm$ 1.2 & 54.8 $\pm$ 1.9 & 77.4 $\pm$ 0.7 & 72.9 $\pm$ 1.7 \\
\cmidrule(lr){2-14}
& Adapter & 75.1 $\pm$ 0.1 & 70.4 $\pm$ 0.1 & 32.0 $\pm$ 0.2 & 59.0 $\pm$ 1.0 & 78.0 $\pm$ 0.1 & 90.6 $\pm$ 0.1 & 93.4 $\pm$ 0.1 & 77.0 $\pm$ 0.3 & 95.2 $\pm$ 0.1 & 53.6 $\pm$ 0.1 & 75.5 $\pm$ 0.3 & 72.7 $\pm$ 0.2 \\
& LP++ & 73.4 $\pm$ 0.3 & 70.9 $\pm$ 0.4 & 36.7 $\pm$ 0.8 & 52.7 $\pm$ 6.2 & 73.6 $\pm$ 1.9 & 88.8 $\pm$ 0.6 & 91.5 $\pm$ 0.9 & 88.7 $\pm$ 1.7 & 95.7 $\pm$ 0.1 & 57.0 $\pm$ 0.8 & 78.9 $\pm$ 1.1 & 73.5 $\pm$ 1.3 \\
& TIP-Adapter & 73.5 $\pm$ 0.0 & 67.7 $\pm$ 0.0 & 30.1 $\pm$ 0.1 & 55.5 $\pm$ 0.2 & 76.9 $\pm$ 0.0 & 90.3 $\pm$ 0.0 & 93.1 $\pm$ 0.0 & 76.4 $\pm$ 0.0 & 95.2 $\pm$ 0.0 & 52.4 $\pm$ 0.0 & 73.7 $\pm$ 0.0 & 71.3 $\pm$ 0.0 \\
\cmidrule(lr){2-14}
& LoRA & 72.3 $\pm$ 0.5 & 73.6 $\pm$ 0.4 & 41.1 $\pm$ 1.2 & 69.0 $\pm$ 5.6 & 80.7 $\pm$ 0.2 & 90.8 $\pm$ 0.2 & 94.4 $\pm$ 0.1 & 86.6 $\pm$ 1.4 & 95.8 $\pm$ 0.5 & 59.0 $\pm$ 0.3 & 80.8 $\pm$ 0.2 & 76.7 $\pm$ 1.0 \\
& PiSSA & 72.2 $\pm$ 0.6 & 72.8 $\pm$ 0.5 & 39.2 $\pm$ 0.9 & 75.1 $\pm$ 1.1 & 78.4 $\pm$ 0.5 & 89.3 $\pm$ 0.5 & 93.4 $\pm$ 0.7 & 92.2 $\pm$ 0.8 & 95.7 $\pm$ 0.4 & 62.1 $\pm$ 0.6 & 82.9 $\pm$ 0.5 & 77.6 $\pm$ 0.6 \\
& GaLoRE & 72.5 $\pm$ 0.5 & 72.5 $\pm$ 0.5 & 39.6 $\pm$ 1.1 & 74.3 $\pm$ 2.8 & 74.2 $\pm$ 4.3 & 89.3 $\pm$ 0.4 & 92.9 $\pm$ 0.5 & 90.2 $\pm$ 2.0 & 95.5 $\pm$ 0.6 & 60.0 $\pm$ 0.9 & 82.1 $\pm$ 0.3 & 76.6 $\pm$ 1.2 \\
\cmidrule(lr){2-14}
& BitFit & 75.9 $\pm$ 0.4 & 73.6 $\pm$ 0.1 & 39.7 $\pm$ 0.9 & 71.4 $\pm$ 4.1 & 80.0 $\pm$ 0.1 & 90.9 $\pm$ 0.3 & 94.5 $\pm$ 0.1 & 83.2 $\pm$ 1.0 & 95.8 $\pm$ 0.2 & 59.5 $\pm$ 0.4 & 80.5 $\pm$ 0.3 & 76.8 $\pm$ 0.7 \\
& LayerNorm & 75.9 $\pm$ 0.1 & 71.2 $\pm$ 0.1 & 33.7 $\pm$ 0.1 & 61.9 $\pm$ 0.8 & 78.3 $\pm$ 0.2 & 90.9 $\pm$ 0.1 & 93.9 $\pm$ 0.1 & 78.7 $\pm$ 0.1 & 95.4 $\pm$ 0.2 & 54.1 $\pm$ 0.2 & 75.7 $\pm$ 0.3 & 73.6 $\pm$ 0.2 \\
& SAFT & 74.2 $\pm$ 0.6 & 73.7 $\pm$ 0.1 & 38.4 $\pm$ 1.6 & 74.2 $\pm$ 3.5 & 78.5 $\pm$ 0.0 & 89.4 $\pm$ 0.5 & 91.0 $\pm$ 1.9 & 89.1 $\pm$ 0.8 & 96.2 $\pm$ 0.8 & 59.6 $\pm$ 1.7 & 81.3 $\pm$ 0.8 & 76.9 $\pm$ 1.1 \\
& SO & 72.2 $\pm$ 0.5 & 73.5 $\pm$ 0.3 & 41.7 $\pm$ 1.4 & 72.9 $\pm$ 3.7 & 80.9 $\pm$ 0.3 & 90.7 $\pm$ 0.3 & 94.6 $\pm$ 0.1 & 88.7 $\pm$ 1.1 & 95.7 $\pm$ 0.4 & 60.2 $\pm$ 0.3 & 81.6 $\pm$ 0.2 & 77.5 $\pm$ 0.8 \\
\cmidrule(lr){2-14}
& Adam & 14.8 $\pm$ 0.7 & 23.5 $\pm$ 1.2 & 1.2 $\pm$ 0.0 & 18.3 $\pm$ 8.2 & 0.5 $\pm$ 0.0 & 5.2 $\pm$ 4.0 & 17.6 $\pm$ 11.2 & 18.8 $\pm$ 4.7 & 44.0 $\pm$ 17.8 & 21.5 $\pm$ 3.2 & 33.8 $\pm$ 4.3 & 18.1 $\pm$ 5.0 \\
& SGD & 76.6 $\pm$ 0.2 & 74.3 $\pm$ 0.1 & 41.1 $\pm$ 0.9 & 75.3 $\pm$ 4.1 & 80.7 $\pm$ 0.4 & 91.0 $\pm$ 0.1 & 94.6 $\pm$ 0.2 & 90.2 $\pm$ 0.8 & 95.6 $\pm$ 0.5 & 60.5 $\pm$ 0.4 & 81.9 $\pm$ 0.5 & 78.3 $\pm$ 0.7 \\
\midrule
\multirow{15}{*}{2} 
& CoOp & 73.3 $\pm$ 0.2 & 71.1 $\pm$ 0.3 & 39.6 $\pm$ 1.9 & 75.9 $\pm$ 1.0 & 79.2 $\pm$ 0.1 & 89.5 $\pm$ 0.8 & 94.1 $\pm$ 0.6 & 92.1 $\pm$ 0.5 & 95.9 $\pm$ 0.9 & 60.4 $\pm$ 1.1 & 81.0 $\pm$ 0.8 & 77.5 $\pm$ 0.7 \\
& KgCoOp & 76.0 $\pm$ 0.1 & 72.5 $\pm$ 0.2 & 35.2 $\pm$ 0.9 & 75.2 $\pm$ 1.8 & 78.1 $\pm$ 0.6 & 91.1 $\pm$ 0.1 & 94.4 $\pm$ 0.5 & 82.6 $\pm$ 0.3 & 95.8 $\pm$ 0.5 & 59.3 $\pm$ 0.5 & 78.0 $\pm$ 0.5 & 76.2 $\pm$ 0.5 \\
& ProGrad & 68.3 $\pm$ 7.6 & 71.9 $\pm$ 0.8 & 37.6 $\pm$ 1.8 & 75.4 $\pm$ 1.0 & 70.8 $\pm$ 11.9 & 89.5 $\pm$ 1.0 & 94.4 $\pm$ 0.6 & 90.1 $\pm$ 2.1 & 95.8 $\pm$ 0.6 & 58.4 $\pm$ 0.4 & 80.1 $\pm$ 0.3 & 75.6 $\pm$ 2.6 \\
\cmidrule(lr){2-14}
& Adapter & 75.6 $\pm$ 0.1 & 72.0 $\pm$ 0.1 & 32.9 $\pm$ 0.5 & 60.9 $\pm$ 1.5 & 78.7 $\pm$ 0.1 & 90.7 $\pm$ 0.2 & 93.8 $\pm$ 0.1 & 78.9 $\pm$ 0.3 & 95.5 $\pm$ 0.2 & 55.0 $\pm$ 0.4 & 76.6 $\pm$ 0.2 & 73.7 $\pm$ 0.3 \\
& LP++ & 75.2 $\pm$ 0.1 & 73.9 $\pm$ 0.1 & 39.3 $\pm$ 1.9 & 66.6 $\pm$ 4.8 & 76.2 $\pm$ 0.5 & 89.7 $\pm$ 0.4 & 92.1 $\pm$ 0.9 & 93.5 $\pm$ 0.8 & 96.8 $\pm$ 0.3 & 64.4 $\pm$ 0.8 & 80.3 $\pm$ 0.2 & 77.1 $\pm$ 1.0 \\
& TIP-Adapter & 73.5 $\pm$ 0.0 & 67.8 $\pm$ 0.0 & 30.0 $\pm$ 0.1 & 56.1 $\pm$ 0.4 & 77.1 $\pm$ 0.1 & 90.3 $\pm$ 0.0 & 93.1 $\pm$ 0.0 & 76.8 $\pm$ 0.1 & 95.2 $\pm$ 0.0 & 52.4 $\pm$ 0.0 & 73.8 $\pm$ 0.0 & 71.5 $\pm$ 0.1 \\
\cmidrule(lr){2-14}
& LoRA & 72.7 $\pm$ 0.1 & 74.1 $\pm$ 0.2 & 43.9 $\pm$ 1.0 & 75.3 $\pm$ 1.7 & 82.6 $\pm$ 0.4 & 90.5 $\pm$ 0.3 & 94.5 $\pm$ 0.8 & 93.3 $\pm$ 0.2 & 96.9 $\pm$ 0.5 & 65.3 $\pm$ 0.7 & 83.6 $\pm$ 0.7 & 79.3 $\pm$ 0.6 \\
& PiSSA & 72.2 $\pm$ 0.1 & 74.0 $\pm$ 0.2 & 41.6 $\pm$ 1.8 & 86.7 $\pm$ 2.7 & 81.6 $\pm$ 0.5 & 89.7 $\pm$ 0.4 & 92.9 $\pm$ 0.4 & 94.6 $\pm$ 0.9 & 96.6 $\pm$ 0.5 & 66.8 $\pm$ 0.4 & 84.0 $\pm$ 0.6 & 80.1 $\pm$ 0.8 \\
& GaLoRE & 73.0 $\pm$ 0.2 & 73.9 $\pm$ 0.3 & 40.7 $\pm$ 3.3 & 85.3 $\pm$ 2.2 & 80.7 $\pm$ 0.3 & 89.5 $\pm$ 0.3 & 93.5 $\pm$ 0.6 & 93.9 $\pm$ 0.5 & 96.1 $\pm$ 1.2 & 66.8 $\pm$ 0.9 & 83.9 $\pm$ 0.6 & 79.8 $\pm$ 0.9 \\
\cmidrule(lr){2-14}
& BitFit & 76.3 $\pm$ 0.2 & 75.0 $\pm$ 0.1 & 41.5 $\pm$ 1.1 & 76.5 $\pm$ 2.2 & 81.0 $\pm$ 0.2 & 91.1 $\pm$ 0.2 & 94.5 $\pm$ 0.3 & 87.8 $\pm$ 0.3 & 96.6 $\pm$ 0.7 & 64.0 $\pm$ 0.5 & 82.2 $\pm$ 0.6 & 78.8 $\pm$ 0.6 \\
& LayerNorm & 76.7 $\pm$ 0.1 & 72.6 $\pm$ 0.1 & 35.5 $\pm$ 0.3 & 63.5 $\pm$ 0.3 & 79.0 $\pm$ 0.1 & 91.1 $\pm$ 0.2 & 94.1 $\pm$ 0.1 & 80.7 $\pm$ 0.4 & 95.9 $\pm$ 0.0 & 55.7 $\pm$ 0.1 & 76.8 $\pm$ 0.2 & 74.7 $\pm$ 0.2 \\
& SAFT & 74.7 $\pm$ 0.3 & 75.1 $\pm$ 0.2 & 40.9 $\pm$ 1.4 & 85.1 $\pm$ 2.6 & 80.0 $\pm$ 0.2 & 90.7 $\pm$ 0.2 & 94.2 $\pm$ 0.2 & 92.7 $\pm$ 0.7 & 97.1 $\pm$ 0.3 & 66.8 $\pm$ 0.9 & 82.4 $\pm$ 0.8 & 80.0 $\pm$ 0.7 \\
& SO & 72.9 $\pm$ 0.2 & 73.9 $\pm$ 0.1 & 44.2 $\pm$ 1.2 & 78.4 $\pm$ 2.0 & 82.5 $\pm$ 0.5 & 90.5 $\pm$ 0.4 & 94.7 $\pm$ 0.6 & 93.5 $\pm$ 0.3 & 97.0 $\pm$ 0.5 & 66.5 $\pm$ 0.6 & 84.1 $\pm$ 0.6 & 79.8 $\pm$ 0.6 \\
\cmidrule(lr){2-14}
& Adam & 18.2 $\pm$ 0.3 & 31.3 $\pm$ 1.1 & 1.1 $\pm$ 0.1 & 20.4 $\pm$ 4.9 & 0.6 $\pm$ 0.1 & 3.1 $\pm$ 2.2 & 3.1 $\pm$ 0.5 & 14.6 $\pm$ 6.0 & 57.7 $\pm$ 6.0 & 10.2 $\pm$ 7.6 & 20.9 $\pm$ 12.2 & 16.5 $\pm$ 3.7 \\
& SGD & 76.8 $\pm$ 0.2 & 76.0 $\pm$ 0.1 & 43.3 $\pm$ 1.6 & 82.0 $\pm$ 1.5 & 82.9 $\pm$ 0.2 & 91.2 $\pm$ 0.2 & 94.4 $\pm$ 0.2 & 94.1 $\pm$ 0.4 & 96.6 $\pm$ 0.6 & 66.8 $\pm$ 1.2 & 83.7 $\pm$ 0.5 & 80.7 $\pm$ 0.6 \\
\midrule
\multirow{15}{*}{4} 
& CoOp & 74.6 $\pm$ 0.1 & 73.2 $\pm$ 0.3 & 43.1 $\pm$ 0.7 & 78.2 $\pm$ 3.8 & 82.5 $\pm$ 0.4 & 89.7 $\pm$ 0.2 & 94.1 $\pm$ 0.0 & 95.7 $\pm$ 0.2 & 96.5 $\pm$ 0.2 & 63.7 $\pm$ 0.9 & 82.4 $\pm$ 0.3 & 79.4 $\pm$ 0.6 \\
& KgCoOp & 76.2 $\pm$ 0.0 & 73.3 $\pm$ 0.0 & 36.6 $\pm$ 0.6 & 73.4 $\pm$ 2.0 & 78.4 $\pm$ 0.3 & 91.2 $\pm$ 0.1 & 94.9 $\pm$ 0.1 & 84.1 $\pm$ 0.6 & 96.5 $\pm$ 0.2 & 61.2 $\pm$ 1.2 & 79.0 $\pm$ 0.4 & 76.8 $\pm$ 0.5 \\
& ProGrad & 74.9 $\pm$ 0.3 & 74.2 $\pm$ 0.1 & 41.8 $\pm$ 0.5 & 78.0 $\pm$ 3.0 & 82.8 $\pm$ 0.2 & 90.2 $\pm$ 0.4 & 94.7 $\pm$ 0.4 & 93.2 $\pm$ 0.6 & 96.3 $\pm$ 0.2 & 63.6 $\pm$ 0.5 & 82.1 $\pm$ 0.7 & 79.2 $\pm$ 0.6 \\
\cmidrule(lr){2-14}
& Adapter & 76.2 $\pm$ 0.1 & 74.0 $\pm$ 0.1 & 35.9 $\pm$ 0.5 & 61.0 $\pm$ 0.8 & 80.3 $\pm$ 0.1 & 90.8 $\pm$ 0.2 & 94.4 $\pm$ 0.1 & 83.3 $\pm$ 0.5 & 96.1 $\pm$ 0.1 & 57.2 $\pm$ 0.4 & 79.7 $\pm$ 0.3 & 75.3 $\pm$ 0.3 \\
& LP++ & 76.7 $\pm$ 0.2 & 76.3 $\pm$ 0.2 & 43.8 $\pm$ 0.3 & 73.7 $\pm$ 1.0 & 79.9 $\pm$ 0.4 & 90.1 $\pm$ 0.2 & 93.5 $\pm$ 0.7 & 96.9 $\pm$ 0.2 & 96.7 $\pm$ 0.2 & 67.4 $\pm$ 0.3 & 83.1 $\pm$ 0.4 & 79.8 $\pm$ 0.4 \\
& TIP-Adapter & 73.8 $\pm$ 0.1 & 68.1 $\pm$ 0.1 & 30.6 $\pm$ 0.0 & 56.8 $\pm$ 0.3 & 77.4 $\pm$ 0.1 & 90.3 $\pm$ 0.0 & 93.2 $\pm$ 0.1 & 78.4 $\pm$ 0.0 & 95.2 $\pm$ 0.0 & 52.5 $\pm$ 0.0 & 73.9 $\pm$ 0.0 & 71.8 $\pm$ 0.1 \\
\cmidrule(lr){2-14}
& LoRA & 73.7 $\pm$ 0.3 & 75.6 $\pm$ 0.3 & 50.0 $\pm$ 0.6 & 75.6 $\pm$ 0.9 & 85.4 $\pm$ 0.7 & 90.6 $\pm$ 0.1 & 95.3 $\pm$ 0.2 & 97.0 $\pm$ 0.5 & 97.7 $\pm$ 0.3 & 68.8 $\pm$ 1.0 & 86.7 $\pm$ 0.2 & 81.5 $\pm$ 0.4 \\
& PiSSA & 73.2 $\pm$ 0.2 & 75.3 $\pm$ 0.0 & 49.0 $\pm$ 0.8 & 85.3 $\pm$ 0.7 & 84.3 $\pm$ 0.5 & 89.5 $\pm$ 0.3 & 94.7 $\pm$ 0.5 & 96.6 $\pm$ 0.1 & 96.8 $\pm$ 0.3 & 70.3 $\pm$ 0.3 & 87.0 $\pm$ 0.4 & 82.0 $\pm$ 0.4 \\
& GaLoRE & 73.7 $\pm$ 0.4 & 75.2 $\pm$ 0.2 & 48.3 $\pm$ 1.0 & 85.3 $\pm$ 0.8 & 83.0 $\pm$ 1.2 & 89.3 $\pm$ 0.4 & 94.0 $\pm$ 1.1 & 96.0 $\pm$ 1.5 & 97.2 $\pm$ 0.5 & 68.4 $\pm$ 0.9 & 87.0 $\pm$ 0.5 & 81.6 $\pm$ 0.8 \\
\cmidrule(lr){2-14}
& BitFit & 77.0 $\pm$ 0.1 & 76.5 $\pm$ 0.1 & 45.8 $\pm$ 0.5 & 78.8 $\pm$ 1.2 & 83.4 $\pm$ 0.2 & 91.2 $\pm$ 0.1 & 95.4 $\pm$ 0.3 & 93.8 $\pm$ 0.3 & 97.3 $\pm$ 0.3 & 66.0 $\pm$ 0.4 & 85.1 $\pm$ 0.6 & 80.9 $\pm$ 0.4 \\
& LayerNorm & 77.5 $\pm$ 0.1 & 74.4 $\pm$ 0.1 & 37.6 $\pm$ 0.2 & 64.2 $\pm$ 0.4 & 80.5 $\pm$ 0.0 & 91.4 $\pm$ 0.0 & 94.7 $\pm$ 0.2 & 82.4 $\pm$ 0.1 & 96.2 $\pm$ 0.1 & 57.9 $\pm$ 0.2 & 78.5 $\pm$ 0.2 & 75.9 $\pm$ 0.1 \\
& SAFT & 75.7 $\pm$ 0.3 & 75.8 $\pm$ 0.1 & 44.9 $\pm$ 0.7 & 87.7 $\pm$ 1.9 & 82.3 $\pm$ 0.5 & 91.1 $\pm$ 0.1 & 95.2 $\pm$ 0.2 & 95.5 $\pm$ 0.3 & 97.4 $\pm$ 0.3 & 68.4 $\pm$ 0.4 & 85.2 $\pm$ 0.5 & 81.7 $\pm$ 0.5 \\
& SO & 73.7 $\pm$ 0.3 & 75.3 $\pm$ 0.2 & 51.0 $\pm$ 0.4 & 79.9 $\pm$ 0.7 & 85.4 $\pm$ 0.2 & 90.4 $\pm$ 0.0 & 95.3 $\pm$ 0.2 & 97.1 $\pm$ 0.4 & 97.3 $\pm$ 0.3 & 68.8 $\pm$ 1.0 & 87.0 $\pm$ 0.6 & 81.9 $\pm$ 0.4 \\
\cmidrule(lr){2-14}
& Adam & 24.1 $\pm$ 0.6 & 36.1 $\pm$ 1.4 & 1.7 $\pm$ 0.5 & 21.2 $\pm$ 5.3 & 5.9 $\pm$ 2.7 & 3.3 $\pm$ 3.3 & 20.6 $\pm$ 17.6 & 47.5 $\pm$ 10.6 & 60.6 $\pm$ 22.5 & 38.6 $\pm$ 2.4 & 27.2 $\pm$ 19.0 & 26.1 $\pm$ 7.8 \\
& SGD & 77.7 $\pm$ 0.1 & 77.6 $\pm$ 0.2 & 48.1 $\pm$ 0.5 & 83.4 $\pm$ 0.7 & 85.8 $\pm$ 0.3 & 91.5 $\pm$ 0.1 & 95.4 $\pm$ 0.1 & 96.8 $\pm$ 0.3 & 97.2 $\pm$ 0.4 & 69.6 $\pm$ 1.1 & 86.3 $\pm$ 0.3 & 82.7 $\pm$ 0.4 \\
\midrule
\multirow{15}{*}{8} 
& CoOp & 76.2 $\pm$ 0.1 & 76.1 $\pm$ 0.2 & 48.1 $\pm$ 1.2 & 79.0 $\pm$ 1.0 & 84.7 $\pm$ 0.8 & 90.1 $\pm$ 0.3 & 94.8 $\pm$ 0.1 & 96.6 $\pm$ 0.5 & 96.4 $\pm$ 0.9 & 68.1 $\pm$ 0.5 & 85.1 $\pm$ 0.9 & 81.4 $\pm$ 0.6 \\
& KgCoOp & 76.4 $\pm$ 0.1 & 74.0 $\pm$ 0.3 & 37.9 $\pm$ 1.0 & 76.9 $\pm$ 1.7 & 79.0 $\pm$ 0.1 & 91.4 $\pm$ 0.1 & 94.7 $\pm$ 0.1 & 85.1 $\pm$ 0.8 & 96.6 $\pm$ 0.1 & 64.3 $\pm$ 0.6 & 80.6 $\pm$ 0.8 & 77.9 $\pm$ 0.5 \\
& ProGrad & 75.7 $\pm$ 0.8 & 76.5 $\pm$ 0.3 & 47.0 $\pm$ 0.8 & 82.0 $\pm$ 1.7 & 84.8 $\pm$ 0.2 & 90.1 $\pm$ 1.2 & 95.1 $\pm$ 0.3 & 95.0 $\pm$ 0.4 & 96.7 $\pm$ 0.5 & 67.9 $\pm$ 0.4 & 84.2 $\pm$ 0.4 & 81.4 $\pm$ 0.6 \\
\cmidrule(lr){2-14}
& Adapter & 76.9 $\pm$ 0.1 & 76.2 $\pm$ 0.1 & 39.6 $\pm$ 0.3 & 64.6 $\pm$ 0.5 & 82.4 $\pm$ 0.2 & 91.3 $\pm$ 0.0 & 94.1 $\pm$ 0.3 & 90.4 $\pm$ 0.4 & 97.1 $\pm$ 0.3 & 61.6 $\pm$ 0.4 & 82.7 $\pm$ 0.2 & 77.9 $\pm$ 0.3 \\
& LP++ & 77.5 $\pm$ 0.2 & 78.1 $\pm$ 0.3 & 47.6 $\pm$ 0.6 & 81.7 $\pm$ 1.2 & 84.0 $\pm$ 0.3 & 90.8 $\pm$ 0.1 & 93.6 $\pm$ 0.5 & 97.4 $\pm$ 0.2 & 96.8 $\pm$ 0.4 & 71.6 $\pm$ 0.5 & 85.3 $\pm$ 0.3 & 82.2 $\pm$ 0.4 \\
& TIP-Adapter & 74.6 $\pm$ 0.1 & 69.7 $\pm$ 0.1 & 32.9 $\pm$ 0.5 & 58.5 $\pm$ 0.2 & 80.0 $\pm$ 0.1 & 90.4 $\pm$ 0.0 & 93.2 $\pm$ 0.2 & 86.1 $\pm$ 0.3 & 95.4 $\pm$ 0.0 & 52.8 $\pm$ 0.1 & 74.2 $\pm$ 0.1 & 73.4 $\pm$ 0.1 \\
\cmidrule(lr){2-14}
& LoRA & 75.1 $\pm$ 0.1 & 77.0 $\pm$ 0.2 & 56.9 $\pm$ 0.8 & 86.5 $\pm$ 1.9 & 88.4 $\pm$ 0.2 & 90.5 $\pm$ 0.2 & 95.5 $\pm$ 0.2 & 98.0 $\pm$ 0.1 & 97.0 $\pm$ 0.5 & 72.4 $\pm$ 0.2 & 88.8 $\pm$ 0.5 & 84.2 $\pm$ 0.5 \\
& PiSSA & 74.1 $\pm$ 0.2 & 76.7 $\pm$ 0.4 & 55.7 $\pm$ 0.7 & 89.2 $\pm$ 1.8 & 88.1 $\pm$ 0.3 & 89.8 $\pm$ 0.3 & 94.7 $\pm$ 0.3 & 97.5 $\pm$ 0.1 & 96.9 $\pm$ 0.5 & 72.2 $\pm$ 0.5 & 88.6 $\pm$ 0.6 & 83.9 $\pm$ 0.5 \\
& GaLoRE & 74.7 $\pm$ 0.1 & 76.7 $\pm$ 0.2 & 54.5 $\pm$ 1.5 & 85.7 $\pm$ 4.6 & 86.8 $\pm$ 0.5 & 89.9 $\pm$ 0.2 & 94.9 $\pm$ 0.4 & 97.0 $\pm$ 0.6 & 96.9 $\pm$ 1.0 & 72.2 $\pm$ 0.7 & 88.3 $\pm$ 1.0 & 83.4 $\pm$ 1.0 \\
\cmidrule(lr){2-14}
& BitFit & 77.9 $\pm$ 0.1 & 78.1 $\pm$ 0.2 & 49.6 $\pm$ 0.6 & 84.0 $\pm$ 0.7 & 86.2 $\pm$ 0.4 & 91.4 $\pm$ 0.1 & 95.5 $\pm$ 0.4 & 96.6 $\pm$ 0.2 & 97.6 $\pm$ 0.4 & 70.7 $\pm$ 0.7 & 86.7 $\pm$ 0.3 & 83.1 $\pm$ 0.4 \\
& LayerNorm & 78.1 $\pm$ 0.1 & 76.1 $\pm$ 0.1 & 41.6 $\pm$ 0.2 & 68.2 $\pm$ 0.3 & 82.1 $\pm$ 0.1 & 91.9 $\pm$ 0.1 & 94.7 $\pm$ 0.1 & 85.9 $\pm$ 0.1 & 96.9 $\pm$ 0.2 & 61.4 $\pm$ 0.2 & 81.1 $\pm$ 0.4 & 78.0 $\pm$ 0.2 \\
& SAFT & 76.6 $\pm$ 0.1 & 76.8 $\pm$ 0.2 & 51.4 $\pm$ 0.6 & 90.3 $\pm$ 1.6 & 84.7 $\pm$ 0.4 & 91.5 $\pm$ 0.2 & 95.6 $\pm$ 0.2 & 96.9 $\pm$ 0.2 & 97.6 $\pm$ 0.2 & 72.1 $\pm$ 0.3 & 86.0 $\pm$ 0.3 & 83.6 $\pm$ 0.4 \\
& SO & 75.2 $\pm$ 0.3 & 77.0 $\pm$ 0.2 & 57.1 $\pm$ 0.6 & 88.2 $\pm$ 1.8 & 88.4 $\pm$ 0.4 & 90.6 $\pm$ 0.2 & 95.7 $\pm$ 0.1 & 98.1 $\pm$ 0.1 & 97.1 $\pm$ 0.6 & 72.5 $\pm$ 0.3 & 88.8 $\pm$ 0.2 & 84.4 $\pm$ 0.4 \\
\cmidrule(lr){2-14}
& Adam & 28.1 $\pm$ 1.0 & 42.1 $\pm$ 1.0 & 2.8 $\pm$ 0.6 & 28.8 $\pm$ 16.0 & 3.0 $\pm$ 1.5 & 5.6 $\pm$ 4.3 & 42.6 $\pm$ 26.3 & 62.4 $\pm$ 13.7 & 83.8 $\pm$ 1.2 & 42.7 $\pm$ 3.4 & 60.7 $\pm$ 1.5 & 36.6 $\pm$ 6.4 \\
& SGD & 78.4 $\pm$ 0.2 & 79.1 $\pm$ 0.2 & 56.5 $\pm$ 0.9 & 88.2 $\pm$ 1.3 & 88.5 $\pm$ 0.2 & 91.8 $\pm$ 0.1 & 95.3 $\pm$ 0.4 & 97.7 $\pm$ 0.1 & 97.5 $\pm$ 0.5 & 72.2 $\pm$ 0.4 & 88.3 $\pm$ 0.1 & 84.9 $\pm$ 0.4 \\
\midrule
\multirow{15}{*}{16} 
& CoOp & 77.8 $\pm$ 0.1 & 78.0 $\pm$ 0.1 & 53.1 $\pm$ 0.6 & 85.8 $\pm$ 0.8 & 87.4 $\pm$ 0.2 & 90.3 $\pm$ 0.4 & 94.6 $\pm$ 0.1 & 98.2 $\pm$ 0.4 & 97.1 $\pm$ 0.2 & 72.9 $\pm$ 1.2 & 86.7 $\pm$ 0.2 & 83.8 $\pm$ 0.4 \\
& KgCoOp & 76.7 $\pm$ 0.1 & 74.9 $\pm$ 0.2 & 39.9 $\pm$ 0.6 & 78.8 $\pm$ 2.1 & 79.4 $\pm$ 0.3 & 91.4 $\pm$ 0.0 & 94.8 $\pm$ 0.2 & 87.9 $\pm$ 0.6 & 96.9 $\pm$ 0.2 & 67.1 $\pm$ 0.1 & 81.9 $\pm$ 0.5 & 79.1 $\pm$ 0.4 \\
& ProGrad & 68.7 $\pm$ 11.0 & 78.1 $\pm$ 0.3 & 51.2 $\pm$ 0.1 & 85.4 $\pm$ 0.9 & 86.9 $\pm$ 0.4 & 90.8 $\pm$ 0.4 & 94.9 $\pm$ 0.2 & 96.7 $\pm$ 0.7 & 97.4 $\pm$ 0.1 & 71.5 $\pm$ 1.0 & 86.7 $\pm$ 0.4 & 82.6 $\pm$ 1.4 \\
\cmidrule(lr){2-14}
& Adapter & 77.8 $\pm$ 0.1 & 77.9 $\pm$ 0.2 & 43.3 $\pm$ 0.4 & 72.2 $\pm$ 0.8 & 84.4 $\pm$ 0.0 & 91.6 $\pm$ 0.1 & 95.0 $\pm$ 0.1 & 96.4 $\pm$ 0.3 & 97.6 $\pm$ 0.1 & 68.6 $\pm$ 0.4 & 85.6 $\pm$ 0.2 & 80.9 $\pm$ 0.2 \\
& LP++ & 78.4 $\pm$ 0.0 & 79.5 $\pm$ 0.1 & 52.2 $\pm$ 0.3 & 84.2 $\pm$ 0.8 & 86.4 $\pm$ 0.4 & 91.2 $\pm$ 0.1 & 94.7 $\pm$ 0.3 & 98.3 $\pm$ 0.1 & 97.4 $\pm$ 0.2 & 75.7 $\pm$ 0.7 & 87.2 $\pm$ 0.6 & 84.1 $\pm$ 0.3 \\
& TIP-Adapter & 76.0 $\pm$ 0.0 & 73.5 $\pm$ 0.1 & 41.6 $\pm$ 0.2 & 61.9 $\pm$ 0.3 & 84.2 $\pm$ 0.1 & 90.7 $\pm$ 0.0 & 93.7 $\pm$ 0.0 & 96.0 $\pm$ 0.1 & 95.5 $\pm$ 0.1 & 53.8 $\pm$ 0.1 & 75.7 $\pm$ 0.2 & 76.6 $\pm$ 0.1 \\
\cmidrule(lr){2-14}
& LoRA & 76.9 $\pm$ 0.2 & 78.7 $\pm$ 0.0 & 64.5 $\pm$ 1.0 & 92.0 $\pm$ 0.1 & 90.2 $\pm$ 0.3 & 90.8 $\pm$ 0.2 & 95.5 $\pm$ 0.0 & 98.9 $\pm$ 0.0 & 97.5 $\pm$ 0.3 & 76.2 $\pm$ 0.7 & 90.4 $\pm$ 0.1 & 86.5 $\pm$ 0.3 \\
& PiSSA & 75.5 $\pm$ 0.1 & 78.1 $\pm$ 0.2 & 63.4 $\pm$ 0.4 & 92.5 $\pm$ 0.5 & 89.9 $\pm$ 0.3 & 89.8 $\pm$ 0.1 & 95.0 $\pm$ 0.1 & 99.1 $\pm$ 0.2 & 97.2 $\pm$ 0.4 & 76.2 $\pm$ 0.7 & 90.3 $\pm$ 0.3 & 86.1 $\pm$ 0.3 \\
& GaLoRE & 75.9 $\pm$ 0.2 & 78.5 $\pm$ 0.2 & 61.4 $\pm$ 0.7 & 93.2 $\pm$ 1.0 & 89.0 $\pm$ 0.3 & 89.6 $\pm$ 0.3 & 94.9 $\pm$ 0.4 & 97.9 $\pm$ 0.6 & 96.8 $\pm$ 0.2 & 76.3 $\pm$ 0.4 & 89.0 $\pm$ 0.2 & 85.7 $\pm$ 0.4 \\
\cmidrule(lr){2-14}
& BitFit & 79.0 $\pm$ 0.1 & 79.5 $\pm$ 0.2 & 55.0 $\pm$ 0.4 & 90.4 $\pm$ 0.8 & 88.1 $\pm$ 0.2 & 91.7 $\pm$ 0.0 & 95.8 $\pm$ 0.2 & 97.9 $\pm$ 0.3 & 97.6 $\pm$ 0.2 & 75.0 $\pm$ 0.5 & 88.6 $\pm$ 0.5 & 85.3 $\pm$ 0.3 \\
& LayerNorm & 78.9 $\pm$ 0.0 & 77.8 $\pm$ 0.1 & 45.5 $\pm$ 0.2 & 72.9 $\pm$ 0.7 & 83.9 $\pm$ 0.2 & 92.0 $\pm$ 0.1 & 95.2 $\pm$ 0.2 & 91.7 $\pm$ 0.2 & 97.5 $\pm$ 0.1 & 66.3 $\pm$ 0.4 & 84.2 $\pm$ 0.4 & 80.5 $\pm$ 0.2 \\
& SAFT & 77.6 $\pm$ 0.2 & 77.7 $\pm$ 0.2 & 59.0 $\pm$ 0.4 & 94.1 $\pm$ 0.8 & 86.5 $\pm$ 0.2 & 91.5 $\pm$ 0.2 & 95.8 $\pm$ 0.1 & 97.9 $\pm$ 0.3 & 97.8 $\pm$ 0.2 & 74.9 $\pm$ 0.2 & 87.8 $\pm$ 0.2 & 85.5 $\pm$ 0.3 \\
& SO & 76.9 $\pm$ 0.1 & 78.5 $\pm$ 0.3 & 64.6 $\pm$ 0.5 & 92.5 $\pm$ 0.2 & 90.7 $\pm$ 0.1 & 90.8 $\pm$ 0.2 & 95.6 $\pm$ 0.1 & 98.9 $\pm$ 0.1 & 97.3 $\pm$ 0.1 & 76.6 $\pm$ 0.5 & 90.9 $\pm$ 0.2 & 86.6 $\pm$ 0.2 \\
\cmidrule(lr){2-14}
& Adam & 34.2 $\pm$ 0.5 & 47.0 $\pm$ 0.5 & 11.2 $\pm$ 6.7 & 45.3 $\pm$ 14.5 & 42.6 $\pm$ 30.1 & 19.0 $\pm$ 10.9 & 56.1 $\pm$ 8.8 & 86.0 $\pm$ 2.2 & 74.6 $\pm$ 16.8 & 49.3 $\pm$ 0.7 & 63.9 $\pm$ 3.2 & 48.1 $\pm$ 8.6 \\
& SGD & 79.5 $\pm$ 0.2 & 80.5 $\pm$ 0.2 & 63.2 $\pm$ 0.5 & 92.0 $\pm$ 0.4 & 90.4 $\pm$ 0.1 & 92.0 $\pm$ 0.0 & 95.7 $\pm$ 0.1 & 98.7 $\pm$ 0.2 & 97.6 $\pm$ 0.2 & 76.8 $\pm$ 0.4 & 90.0 $\pm$ 0.2 & 86.9 $\pm$ 0.2 \\
\bottomrule
\end{tabular}
\end{adjustbox}
\end{table*}

The PEFT baselines cover representative families. The prompt-learning group includes CoOp, KgCoOp, and ProGrad~\cite{zhou2022coop,yao2023kgcoop,zhu2022prograd}. The adapter, cache, and classifier-side group includes CLIP-Adapter, TIP-Adapter, and LP++~\cite{gao2024clipadapter,zhang2022tipadapter,huang2024lpplusplus}. The low-rank or projected-update group includes LoRA, PiSSA, and GaLoRE~\cite{hu2022lora,meng2024pissa,zhao2024galore}. The sparse or subset-based group includes BitFit, LayerNorm tuning, SAFT, and SO~\cite{ben-zaken2022bitfit,ba2016layer,nguyen2024saft,mrabah2025so}. All hyperparameters are fixed across datasets, shot levels, backbones, and seeds, as described in Appendix~\ref{app:hyperparameters}. This appendix compares test accuracy over 3 fixed random seeds on 11 datasets.

SGD is the best method in every averaged row of the comprehensive tables. For ViT-B/32, SGD exceeds the best PEFT average by $+1.0$, $+0.8$, $+0.9$, $+0.6$, and $+0.6$ points for \(K=1,2,4,8,16\). For ViT-B/16, the corresponding margins are $+1.4$, $+1.1$, $+1.0$, $+0.4$, and $+0.8$ points. For ViT-L/14, they are $+0.7$, $+0.6$, $+0.7$, $+0.5$, and $+0.3$ points. Across all $15$ backbone-shot settings, SGD averages $77.9\%$ accuracy. The best PEFT method selected separately for each row averages $77.2\%$. The mean SGD margin is +0.76 points.

The family-level comparison provides a larger separation. Against the best prompt-learning baseline in each averaged row, SGD is higher by 3.9 points on average. Against the best adapter, cache, or classifier-side baseline, the mean margin is 4.0 points. The strongest competitors are the low-rank and sparse families. Even for these families, SGD remains higher by 0.9 points on average. The best PEFT method is almost always PiSSA or SO. PiSSA is the most competitive in several low- and mid-shot rows. SO is the strongest in most high-shot rows. No PEFT family gives the best average across all backbones and shot levels.

The dataset-level tables also clarify the exceptions. SGD does not win every individual dataset cell. Some PEFT methods are locally better on Aircraft, EuroSAT, Food101, Oxford-Pets, Flowers102, Caltech101, DTD, or UCF101 for particular backbones and shot levels. These gains are method-specific. They do not form a stable dominance pattern, and they are offset by losses on other datasets. In contrast, SGD beats each individual PEFT method on the averaged row for all 15 backbone-shot settings. The aggregate advantage is therefore not caused by a weak PEFT baseline. It also holds against the strongest PEFT method selected separately for each row.

These results support the central claim of the paper. Full fine-tuning is not intrinsically worse than PEFT in few-shot CLIP adaptation. When dense full fine-tuning is optimized with plain SGD, it consistently outperforms state-of-the-art PEFT methods in average accuracy across backbones and shot levels. The empirical limitation is therefore not the dense parameterization itself. It is the optimizer and, in particular, the unreliable few-shot preconditioning used by adaptive methods.

\section{Stability Analysis}
\label{app:stability_analysis}

This appendix studies stability over long training. We use the 1-shot CLIP (ViT-B/16) setting with the same training protocol as in the main benchmark. We sweep the learning rate in $\{10^{-2},10^{-3},10^{-4},10^{-5}\}$ and train for 300 epochs. The curves show mean test accuracy across three fixed random seeds. The goal is not only to compare the best endpoint. It is also to test whether a method remains stable as training continues.

\subsection{SGD vs PEFT}
\label{app:stability_sgd_vs_peft}

\begin{figure*}
  \centering
  \includegraphics[width=\linewidth]{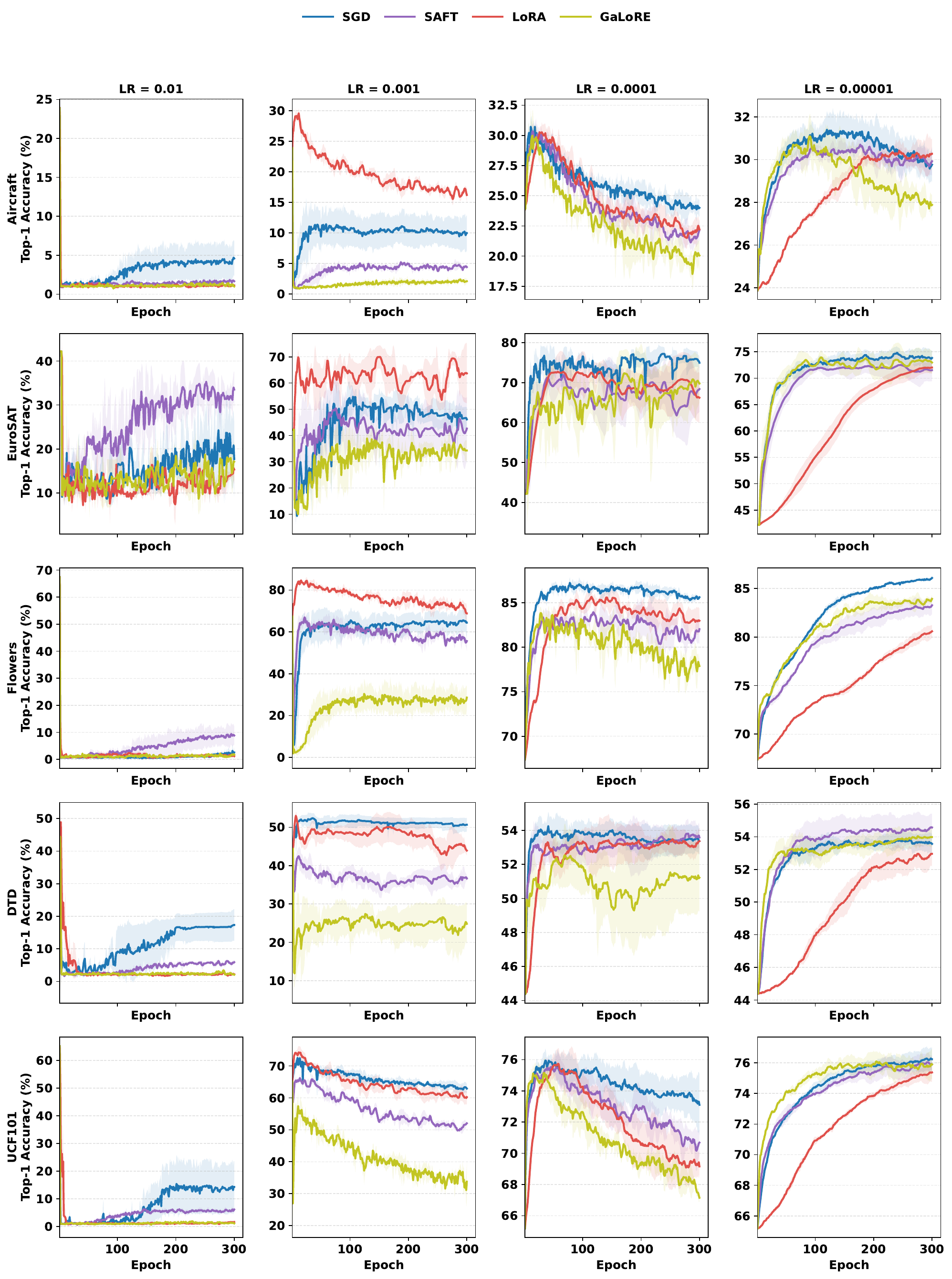}
  \caption{Test accuracy (mean $\pm$ std over 3 seeds) over training epochs for SGD, LoRA, GaLoRE, and SAFT in a 1-shot setting on five datasets (Aircraft, EuroSAT, Flowers, DTD, and UCF101) using CLIP (ViT-B/16). Each column corresponds to one learning rate: $10^{-2}$, $10^{-3}$, $10^{-4}$, and $10^{-5}$.}
  \label{fig:stability_group1_with_aug}
\end{figure*}

\begin{figure*}
  \centering
  \includegraphics[width=\linewidth]{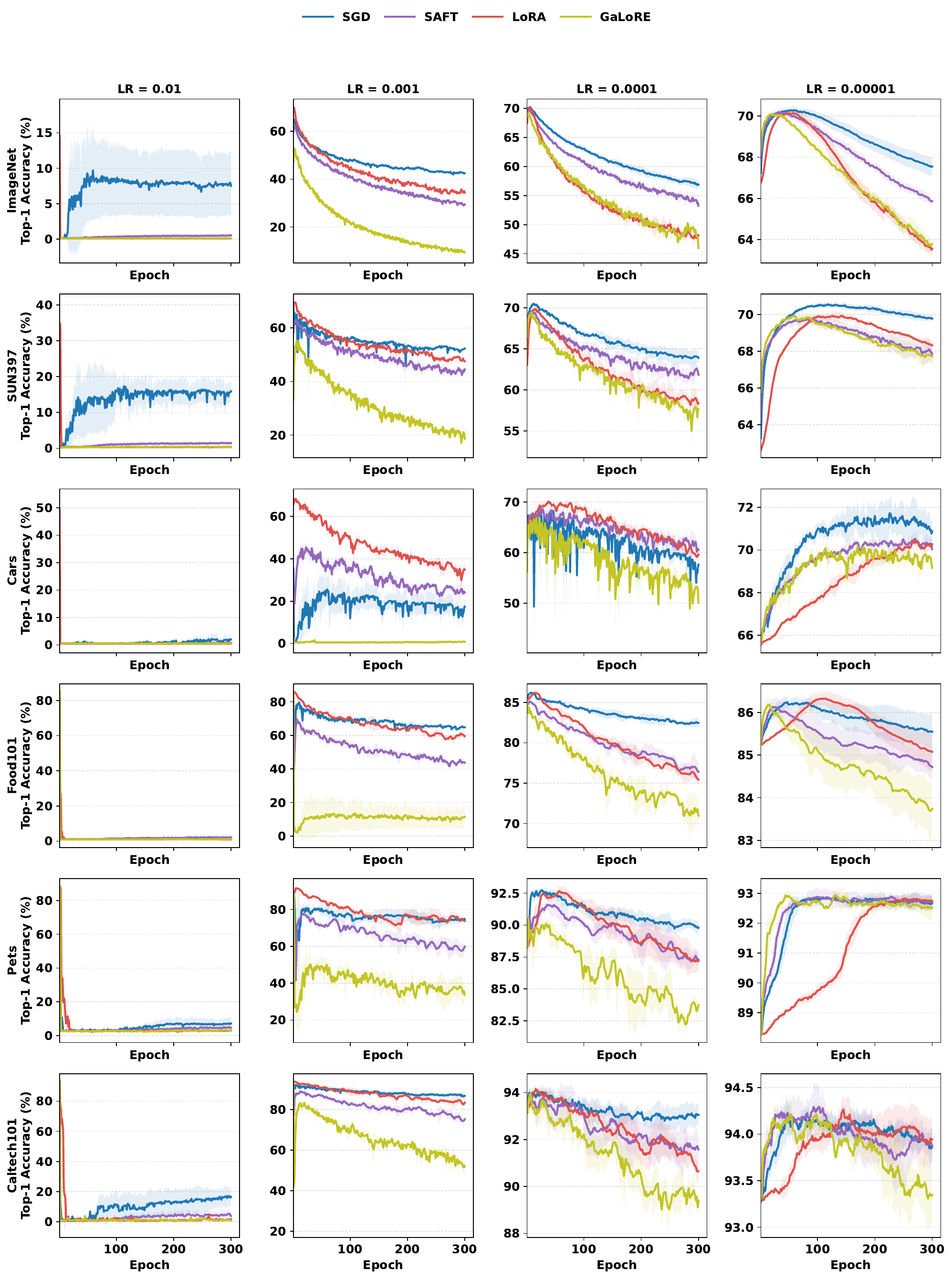}
  \caption{Test accuracy (mean $\pm$ std over 3 seeds) over training epochs for SGD, LoRA, GaLoRE, and SAFT in a 1-shot setting on six datasets (ImageNet, SUN397, Cars, Food101, Pets, and Caltech101) using CLIP (ViT-B/16). Each column corresponds to one learning rate: $10^{-2}$, $10^{-3}$, $10^{-4}$, and $10^{-5}$.}
  \label{fig:stability_group2_with_aug}
\end{figure*}

\begin{figure*}
  \centering
  \includegraphics[width=\linewidth]{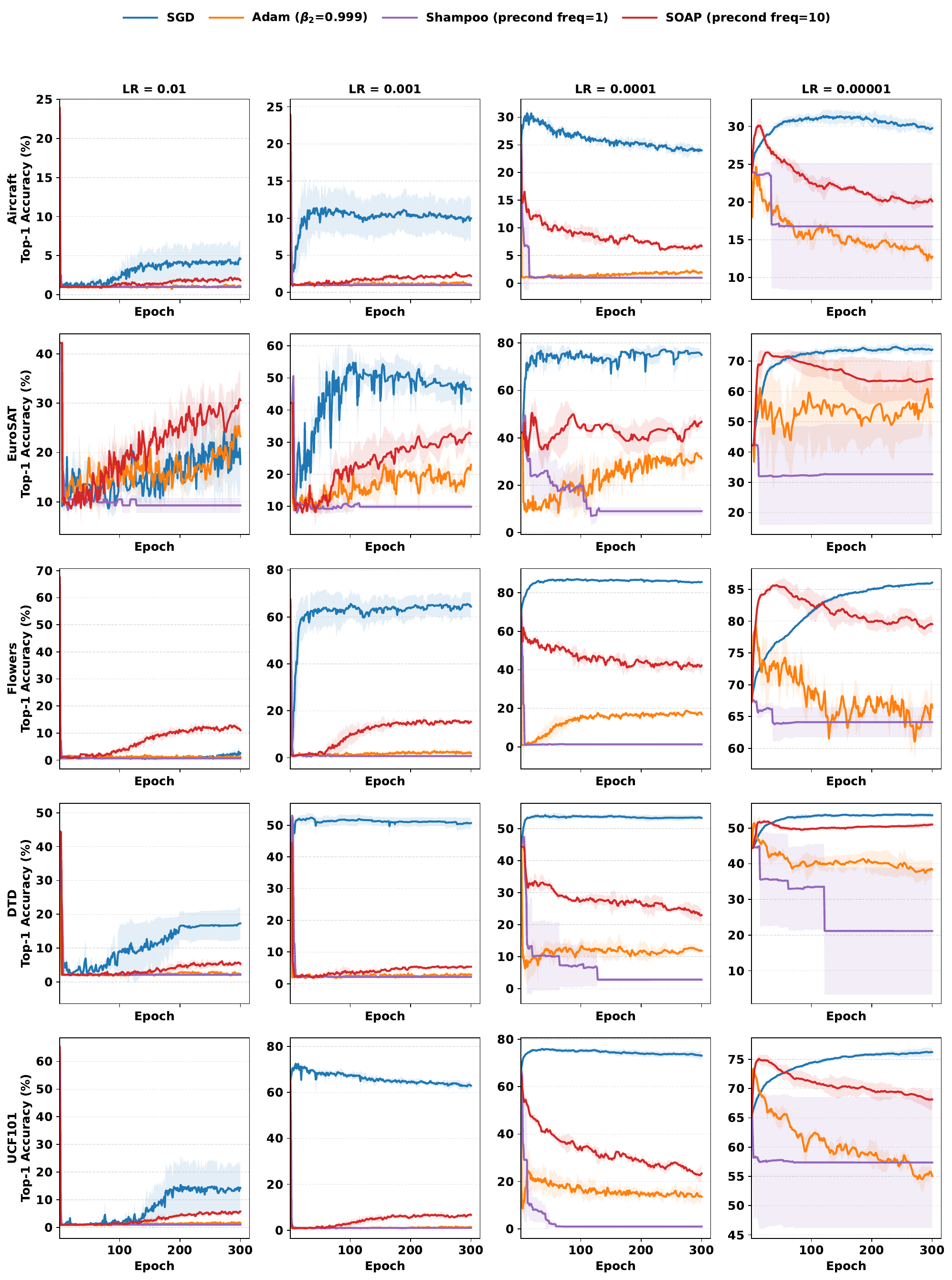}
  \caption{Test accuracy (mean $\pm$ std over 3 fixed seeds) over training epochs for SGD, Adam ($\beta_2=0.999$), Shampoo ($F=1$), and SOAP ($F=10$) in a 1-shot setting on five datasets (Aircraft, EuroSAT, Flowers, DTD, and UCF101) using CLIP (ViT-B/16). Each column corresponds to one learning rate: $10^{-2}$, $10^{-3}$, $10^{-4}$, and $10^{-5}$.}
  \label{fig:fewshot_curves_shots1_SGD_vs_Adam_vs_Shampoo_vs_SOAP}
\end{figure*}

Figures~\ref{fig:stability_group1_with_aug} and~\ref{fig:stability_group2_with_aug} compare dense SGD with three PEFT baselines over 300 epochs. SGD is the most stable method across the learning-rate grid. At $10^{-4}$, its mean final accuracy over the 11 datasets is $68.6\%$. The corresponding values for SAFT, LoRA, and GaLoRE are $66.2\%$, $64.9\%$, and $62.2\%$. At $10^{-5}$, all methods become more stable, but SGD still has the highest mean final accuracy, $72.7\%$.

The peak-to-final behavior gives a clearer stability signal. At $10^{-4}$, the average peak-to-final drop is $4.8$ points for SGD. It is $6.1$ points for SAFT, $8.0$ points for LoRA, and $9.6$ points for GaLoRE. At $10^{-3}$, the same drop is $12.1$ points for SGD, but $18.6$--$33.7$ points for the PEFT methods. Thus, several PEFT curves can reach a high early accuracy and then decay. SGD has a smaller late-epoch degradation.

The highest learning rate, $10^{-2}$, is a stress test. Most methods become unstable or collapse on many datasets. Even in this regime, SGD usually degrades more gracefully than the PEFT methods. The final accuracy is not competitive with smaller learning rates, so this setting should not be interpreted as a useful operating point.

There are dataset-specific exceptions. LoRA or SAFT can be best on a few dataset--learning-rate pairs, especially at $10^{-5}$. These gains are isolated. Across all 44 dataset--learning-rate cells in Figures~\ref{fig:stability_group1_with_aug} and~\ref{fig:stability_group2_with_aug}, SGD has the best final accuracy in 32 cells. The aggregate result is therefore not that SGD wins every curve. It is that SGD provides the most reliable long-horizon behavior across datasets and learning rates.

\subsection{SGD vs Adam vs Shampoo vs SOAP}
\label{app:stability_sgd_vs_second_order}

Figure~\ref{fig:fewshot_curves_shots1_SGD_vs_Adam_vs_Shampoo_vs_SOAP} compares SGD with Adam and two structured preconditioned optimizers. The contrast is stronger than in the PEFT comparison. At $10^{-4}$, SGD achieves a mean final accuracy of $62.2\%$ across the five datasets. SOAP reaches $28.4\%$, Adam reaches $15.1\%$, and Shampoo reaches $3.0\%$. At $10^{-5}$, the preconditioned methods are less unstable, but they remain below SGD. The mean final accuracies are $63.9\%$ for SGD, $56.6\%$ for SOAP, $45.5\%$ for Adam, and $38.4\%$ for Shampoo.

The long-horizon degradation is also much larger for the preconditioned optimizers. At $10^{-4}$, SGD loses only $2.9$ points from its peak to the final epoch. Adam loses $33.5$ points, SOAP loses $21.9$ points, and Shampoo loses $48.9$ points. At $10^{-5}$, SGD loses $0.6$ points, while Adam, SOAP, and Shampoo lose $12.5$, $6.6$, and $10.6$ points. Thus, lowering the learning rate can dampen the collapse of adaptive and structured preconditioned methods, but it does not recover the stability of SGD.

The structured optimizers are not rescued by richer preconditioning. Shampoo often collapses after an early high-accuracy phase. SOAP is more stable than Adam and Shampoo at the lowest learning rate, but it still trails behind SGD on all five datasets after learning rate tuning. Across the 20 dataset--learning-rate cells in Figure~\ref{fig:fewshot_curves_shots1_SGD_vs_Adam_vs_Shampoo_vs_SOAP}, SGD has the best final accuracy in 18 cells. The two exceptions occur at $10^{-2}$, where all methods are unstable, and the final accuracies are low.

These results support the main conclusion of this work. Full fine-tuning is stable in the few-shot regime when it is optimized by plain SGD. The unstable component is not the dense parameterization. It is the support-set preconditioning used by Adam, Shampoo, and SOAP. SGD avoids this unreliable preconditioner and makes smaller, more stable updates near the pretrained CLIP solution.

\section{SGD vs Adam for PEFT}
\label{app:sgd_vs_adam_peft}

\begin{table*}
\centering
\caption{Ablation on adapter-style PEFT optimizers compared with dense full fine-tuning using SGD for CLIP (ViT-B/32). We report Adapter, TIP-Adapter, and LP++ trained with Adam and SGD. Top-1 accuracy (\%) averaged over 3 fixed random seeds.}
\resizebox{1.\textwidth}{!}{
\begin{tabular}{llcccccccccccc}
\toprule
Shots & Method & ImageNet & SUN & Aircraft & EuroSAT & Cars & Food & Pets & Flowers & Caltech & DTD & UCF & Average \\ \midrule
\multirow{1}{*}{0} & CLIP & 62.0 & 62.1 & 18.1 & 35.8 & 60.1 & 78.9 & 85.1 & 64.0 & 91.4 & 43.0 & 61.6 & 60.2 \\
\midrule
\multirow{7}{*}{1} & Adapter (Adam) & 62.8 $\pm$ 0.1 & 63.0 $\pm$ 0.0 & 18.5 $\pm$ 0.2 & 39.0 $\pm$ 1.1 & 60.7 $\pm$ 0.1 & 79.3 $\pm$ 0.1 & 86.1 $\pm$ 0.1 & 64.7 $\pm$ 0.3 & 91.6 $\pm$ 0.0 & 43.7 $\pm$ 0.2 & 62.6 $\pm$ 0.3 & 61.1 $\pm$ 0.2 \\
& Adapter (SGD) & 62.0 $\pm$ 0.0 & 62.1 $\pm$ 0.0 & 18.1 $\pm$ 0.0 & 35.7 $\pm$ 0.6 & 60.1 $\pm$ 0.1 & 78.9 $\pm$ 0.2 & 85.3 $\pm$ 0.2 & 63.9 $\pm$ 0.2 & 91.5 $\pm$ 0.1 & 42.8 $\pm$ 0.2 & 61.6 $\pm$ 0.1 & 60.2 $\pm$ 0.2 \\
\cmidrule(lr){2-14}
& TIP-Adapter (Adam) & 62.1 $\pm$ 0.0 & 62.1 $\pm$ 0.0 & 18.2 $\pm$ 0.1 & 36.5 $\pm$ 0.4 & 60.2 $\pm$ 0.0 & 78.9 $\pm$ 0.0 & 85.1 $\pm$ 0.0 & 64.3 $\pm$ 0.0 & 91.4 $\pm$ 0.1 & 43.0 $\pm$ 0.0 & 61.7 $\pm$ 0.0 & 60.3 $\pm$ 0.1 \\
& TIP-Adapter (SGD) & 62.1 $\pm$ 0.0 & 62.1 $\pm$ 0.0 & 18.2 $\pm$ 0.1 & 36.3 $\pm$ 0.4 & 60.2 $\pm$ 0.0 & 78.9 $\pm$ 0.0 & 85.1 $\pm$ 0.0 & 64.2 $\pm$ 0.0 & 91.4 $\pm$ 0.1 & 43.0 $\pm$ 0.0 & 61.7 $\pm$ 0.0 & 60.3 $\pm$ 0.1 \\
\cmidrule(lr){2-14}
& LP++ (Adam) & 13.3 $\pm$ 0.8 & 17.6 $\pm$ 0.3 & 7.0 $\pm$ 1.2 & 27.8 $\pm$ 4.9 & 11.0 $\pm$ 0.7 & 18.7 $\pm$ 0.6 & 19.9 $\pm$ 4.9 & 27.7 $\pm$ 0.7 & 48.0 $\pm$ 4.8 & 21.4 $\pm$ 1.8 & 24.7 $\pm$ 1.2 & 21.5 $\pm$ 2.0 \\
& LP++ (SGD) & 60.9 $\pm$ 0.2 & 63.9 $\pm$ 0.6 & 20.7 $\pm$ 0.5 & 46.6 $\pm$ 1.2 & 54.9 $\pm$ 0.6 & 75.6 $\pm$ 0.6 & 83.3 $\pm$ 1.5 & 76.3 $\pm$ 1.1 & 91.0 $\pm$ 0.7 & 47.4 $\pm$ 1.4 & 66.6 $\pm$ 0.2 & 62.5 $\pm$ 0.8 \\
\bottomrule
\end{tabular}}

\label{tab:ablation_adapter_vs_sgd_ViTB32_with_aug}
\end{table*}
 
\begin{table*}
\centering
\caption{Ablation on adapter-style PEFT optimizers compared with dense full fine-tuning using SGD for CLIP (ViT-B/16). We report Adapter, TIP-Adapter, and LP++ trained with Adam and SGD. Top-1 accuracy (\%) averaged over 3 fixed random seeds.}
\resizebox{1.\textwidth}{!}{
\begin{tabular}{llcccccccccccc}
\toprule
Shots & Method & ImageNet & SUN & Aircraft & EuroSAT & Cars & Food & Pets & Flowers & Caltech & DTD & UCF & Average \\ \midrule
\multirow{1}{*}{0} & CLIP & 66.7 & 62.6 & 23.9 & 42.2 & 65.5 & 85.2 & 88.3 & 67.4 & 93.3 & 44.4 & 65.2 & 64.1 \\
\midrule
\multirow{7}{*}{1} & Adapter (Adam) & 67.6 $\pm$ 0.1 & 64.3 $\pm$ 0.1 & 24.1 $\pm$ 0.2 & 45.6 $\pm$ 0.7 & 66.2 $\pm$ 0.2 & 85.4 $\pm$ 0.1 & 89.2 $\pm$ 0.1 & 68.6 $\pm$ 0.4 & 93.4 $\pm$ 0.1 & 45.0 $\pm$ 0.1 & 66.3 $\pm$ 0.2 & 65.1 $\pm$ 0.2 \\
& Adapter (SGD) & 66.7 $\pm$ 0.1 & 62.6 $\pm$ 0.1 & 23.6 $\pm$ 0.2 & 41.2 $\pm$ 0.3 & 65.6 $\pm$ 0.1 & 85.2 $\pm$ 0.1 & 88.2 $\pm$ 0.2 & 67.4 $\pm$ 0.2 & 93.2 $\pm$ 0.1 & 44.2 $\pm$ 0.2 & 65.1 $\pm$ 0.1 & 63.9 $\pm$ 0.1 \\
\cmidrule(lr){2-14}
& TIP-Adapter (Adam) & 66.7 $\pm$ 0.0 & 62.6 $\pm$ 0.0 & 23.9 $\pm$ 0.0 & 43.1 $\pm$ 0.1 & 65.6 $\pm$ 0.1 & 85.2 $\pm$ 0.0 & 88.3 $\pm$ 0.0 & 67.6 $\pm$ 0.1 & 93.3 $\pm$ 0.0 & 44.4 $\pm$ 0.0 & 65.3 $\pm$ 0.0 & 64.2 $\pm$ 0.0 \\
& TIP-Adapter (SGD) & 66.7 $\pm$ 0.0 & 62.6 $\pm$ 0.0 & 24.0 $\pm$ 0.0 & 42.9 $\pm$ 0.1 & 65.5 $\pm$ 0.1 & 85.2 $\pm$ 0.0 & 88.3 $\pm$ 0.0 & 67.5 $\pm$ 0.1 & 93.3 $\pm$ 0.0 & 44.4 $\pm$ 0.0 & 65.3 $\pm$ 0.0 & 64.2 $\pm$ 0.0 \\
\cmidrule(lr){2-14}
& LP++ (Adam) & 15.4 $\pm$ 0.5 & 18.3 $\pm$ 1.2 & 9.0 $\pm$ 0.6 & 23.0 $\pm$ 2.6 & 15.2 $\pm$ 1.1 & 24.1 $\pm$ 2.6 & 23.2 $\pm$ 1.9 & 37.1 $\pm$ 1.9 & 45.9 $\pm$ 3.3 & 22.2 $\pm$ 2.1 & 25.1 $\pm$ 2.3 & 23.5 $\pm$ 1.8 \\
& LP++ (SGD) & 66.3 $\pm$ 0.1 & 66.6 $\pm$ 0.2 & 26.7 $\pm$ 0.8 & 40.8 $\pm$ 2.4 & 62.6 $\pm$ 0.4 & 82.9 $\pm$ 1.3 & 86.1 $\pm$ 0.4 & 81.6 $\pm$ 0.8 & 93.3 $\pm$ 0.7 & 50.9 $\pm$ 1.2 & 71.1 $\pm$ 0.9 & 66.3 $\pm$ 0.8 \\
\bottomrule
\end{tabular}}
\label{tab:ablation_adapter_vs_sgd_ViTB16_with_aug}
\end{table*}
 
\begin{table*}
\centering
\caption{Ablation on adapter-style PEFT optimizers compared with dense full fine-tuning using SGD for CLIP (ViT-L/14). We report Adapter, TIP-Adapter, and LP++ trained with Adam and SGD. Top-1 accuracy (\%) averaged over 3 fixed random seeds.}
\resizebox{1.\textwidth}{!}{
\begin{tabular}{llcccccccccccc}
\toprule
Shots & Method & ImageNet & SUN & Aircraft & EuroSAT & Cars & Food & Pets & Flowers & Caltech & DTD & UCF & Average \\ \midrule
\multirow{1}{*}{0} & CLIP & 73.5 & 67.7 & 30.0 & 55.1 & 76.8 & 90.3 & 93.1 & 76.2 & 95.2 & 52.4 & 73.7 & 71.3 \\
\midrule
\multirow{7}{*}{1} & Adapter (Adam) & 75.1 $\pm$ 0.1 & 70.4 $\pm$ 0.1 & 32.0 $\pm$ 0.2 & 59.0 $\pm$ 1.0 & 78.0 $\pm$ 0.1 & 90.6 $\pm$ 0.1 & 93.4 $\pm$ 0.1 & 77.0 $\pm$ 0.3 & 95.2 $\pm$ 0.1 & 53.6 $\pm$ 0.1 & 75.5 $\pm$ 0.3 & 72.7 $\pm$ 0.2 \\
& Adapter (SGD) & 73.4 $\pm$ 0.0 & 67.6 $\pm$ 0.1 & 30.2 $\pm$ 0.1 & 55.3 $\pm$ 0.2 & 76.9 $\pm$ 0.1 & 90.2 $\pm$ 0.0 & 93.2 $\pm$ 0.1 & 76.0 $\pm$ 0.1 & 95.2 $\pm$ 0.1 & 52.3 $\pm$ 0.1 & 73.8 $\pm$ 0.1 & 71.3 $\pm$ 0.1 \\
\cmidrule(lr){2-14}
& TIP-Adapter (Adam) & 73.5 $\pm$ 0.0 & 67.7 $\pm$ 0.0 & 30.1 $\pm$ 0.1 & 55.5 $\pm$ 0.2 & 76.9 $\pm$ 0.0 & 90.3 $\pm$ 0.0 & 93.1 $\pm$ 0.0 & 76.4 $\pm$ 0.0 & 95.2 $\pm$ 0.0 & 52.4 $\pm$ 0.0 & 73.7 $\pm$ 0.0 & 71.3 $\pm$ 0.0 \\
& TIP-Adapter (SGD) & 73.5 $\pm$ 0.0 & 67.7 $\pm$ 0.0 & 30.1 $\pm$ 0.1 & 55.4 $\pm$ 0.1 & 76.9 $\pm$ 0.0 & 90.3 $\pm$ 0.0 & 93.1 $\pm$ 0.0 & 76.3 $\pm$ 0.0 & 95.1 $\pm$ 0.0 & 52.4 $\pm$ 0.0 & 73.7 $\pm$ 0.0 & 71.3 $\pm$ 0.0 \\
\cmidrule(lr){2-14}
& LP++ (Adam) & 19.1 $\pm$ 0.4 & 23.3 $\pm$ 0.7 & 11.5 $\pm$ 1.3 & 31.2 $\pm$ 6.5 & 22.4 $\pm$ 0.8 & 29.7 $\pm$ 1.7 & 31.9 $\pm$ 0.9 & 42.8 $\pm$ 3.4 & 42.0 $\pm$ 3.1 & 25.9 $\pm$ 2.2 & 30.9 $\pm$ 1.4 & 28.2 $\pm$ 2.0 \\
& LP++ (SGD) & 73.4 $\pm$ 0.3 & 70.9 $\pm$ 0.4 & 36.7 $\pm$ 0.8 & 52.7 $\pm$ 6.2 & 73.6 $\pm$ 1.9 & 88.8 $\pm$ 0.6 & 91.5 $\pm$ 0.9 & 88.7 $\pm$ 1.7 & 95.7 $\pm$ 0.1 & 57.0 $\pm$ 0.8 & 78.9 $\pm$ 1.1 & 73.5 $\pm$ 1.3 \\
\bottomrule
\end{tabular}}

\label{tab:ablation_adapter_vs_sgd_ViTL14_with_aug}
\end{table*}

\begin{table*}
\centering
\caption{Ablation on prompt-tuning PEFT optimizers compared with dense full fine-tuning using SGD for CLIP (ViT-B/32). We report CoOp, KgCoOp, and ProGrad trained with Adam and SGD. Top-1 accuracy (\%) averaged over 3 fixed random seeds.}
\resizebox{1.\textwidth}{!}{
\begin{tabular}{llcccccccccccc}
\toprule
Shots & Method & ImageNet & SUN & Aircraft & EuroSAT & Cars & Food & Pets & Flowers & Caltech & DTD & UCF & Average \\ \midrule
\multirow{1}{*}{0} & CLIP & 62.0 & 62.1 & 18.1 & 35.8 & 60.1 & 78.9 & 85.1 & 64.0 & 91.4 & 43.0 & 61.6 & 60.2 \\
\midrule
\multirow{7}{*}{1} & CoOp (Adam) & 62.4 $\pm$ 0.5 & 64.1 $\pm$ 0.2 & 21.4 $\pm$ 1.1 & 49.2 $\pm$ 4.4 & 60.4 $\pm$ 0.8 & 79.0 $\pm$ 1.2 & 85.9 $\pm$ 0.3 & 70.8 $\pm$ 0.9 & 91.3 $\pm$ 0.1 & 45.2 $\pm$ 1.4 & 68.0 $\pm$ 1.7 & 63.4 $\pm$ 1.1 \\
& CoOp (SGD) & 63.4 $\pm$ 0.1 & 64.4 $\pm$ 0.2 & 14.8 $\pm$ 7.7 & 50.7 $\pm$ 0.9 & 59.4 $\pm$ 0.6 & 80.2 $\pm$ 0.5 & 88.2 $\pm$ 0.7 & 64.6 $\pm$ 1.7 & 90.5 $\pm$ 0.4 & 45.0 $\pm$ 2.4 & 65.0 $\pm$ 0.6 & 62.4 $\pm$ 1.4 \\
\cmidrule(lr){2-14}
& KgCoOp (Adam) & 64.3 $\pm$ 0.2 & 66.5 $\pm$ 0.0 & 21.4 $\pm$ 0.2 & 55.9 $\pm$ 3.1 & 61.5 $\pm$ 0.2 & 80.8 $\pm$ 0.2 & 88.5 $\pm$ 0.2 & 67.8 $\pm$ 0.2 & 93.0 $\pm$ 0.1 & 49.5 $\pm$ 1.9 & 69.3 $\pm$ 0.2 & 65.3 $\pm$ 0.6 \\
& KgCoOp (SGD) & 63.5 $\pm$ 0.2 & 65.4 $\pm$ 0.4 & 20.6 $\pm$ 0.6 & 51.9 $\pm$ 0.9 & 60.4 $\pm$ 0.2 & 80.5 $\pm$ 0.0 & 88.5 $\pm$ 0.5 & 65.5 $\pm$ 0.6 & 92.2 $\pm$ 0.3 & 47.4 $\pm$ 1.8 & 66.6 $\pm$ 0.2 & 63.9 $\pm$ 0.5 \\
\cmidrule(lr){2-14}
& ProGrad (Adam) & 54.2 $\pm$ 4.9 & 61.9 $\pm$ 2.0 & 16.8 $\pm$ 1.8 & 50.7 $\pm$ 2.6 & 59.1 $\pm$ 1.7 & 77.0 $\pm$ 1.6 & 82.8 $\pm$ 4.5 & 69.0 $\pm$ 1.0 & 90.3 $\pm$ 0.8 & 45.1 $\pm$ 0.8 & 66.8 $\pm$ 1.5 & 61.2 $\pm$ 2.1 \\
& ProGrad (SGD) & 54.1 $\pm$ 4.1 & 58.8 $\pm$ 1.9 & 11.6 $\pm$ 6.8 & 49.2 $\pm$ 1.4 & 53.2 $\pm$ 1.3 & 68.5 $\pm$ 7.6 & 70.0 $\pm$ 18.7 & 50.1 $\pm$ 12.1 & 71.7 $\pm$ 3.1 & 36.8 $\pm$ 5.5 & 60.8 $\pm$ 0.3 & 53.2 $\pm$ 5.7 \\
\bottomrule
\end{tabular}}

\label{tab:ablation_prompt_vs_sgd_ViTB32_with_aug}
\end{table*}

\begin{table*}
\centering
\caption{Ablation on prompt-tuning PEFT optimizers compared with dense full fine-tuning using SGD for CLIP (ViT-B/16). We report CoOp, KgCoOp, and ProGrad trained with Adam and SGD. Top-1 accuracy (\%) averaged over 3 fixed random seeds.}
\resizebox{1.\textwidth}{!}{
\begin{tabular}{llcccccccccccc}
\toprule
Shots & Method & ImageNet & SUN & Aircraft & EuroSAT & Cars & Food & Pets & Flowers & Caltech & DTD & UCF & Average \\ \midrule
\multirow{1}{*}{0} & CLIP & 66.7 & 62.6 & 23.9 & 42.2 & 65.5 & 85.2 & 88.3 & 67.4 & 93.3 & 44.4 & 65.2 & 64.1 \\
\midrule
\multirow{7}{*}{1} & CoOp (Adam) & 67.3 $\pm$ 0.3 & 67.0 $\pm$ 0.6 & 28.1 $\pm$ 0.4 & 54.3 $\pm$ 6.9 & 66.7 $\pm$ 0.6 & 84.7 $\pm$ 0.5 & 89.3 $\pm$ 0.6 & 75.6 $\pm$ 1.6 & 93.1 $\pm$ 0.4 & 47.6 $\pm$ 1.6 & 72.3 $\pm$ 0.7 & 67.8 $\pm$ 1.3 \\
& CoOp (SGD) & 67.7 $\pm$ 0.4 & 66.8 $\pm$ 0.5 & 26.7 $\pm$ 0.1 & 52.6 $\pm$ 2.1 & 64.6 $\pm$ 0.7 & 85.8 $\pm$ 0.2 & 89.6 $\pm$ 0.4 & 70.1 $\pm$ 0.5 & 91.8 $\pm$ 1.4 & 46.6 $\pm$ 1.2 & 70.0 $\pm$ 1.3 & 66.6 $\pm$ 0.8 \\
\cmidrule(lr){2-14}
& KgCoOp (Adam) & 69.3 $\pm$ 0.2 & 68.6 $\pm$ 0.3 & 26.8 $\pm$ 0.2 & 60.1 $\pm$ 2.7 & 66.5 $\pm$ 0.2 & 86.1 $\pm$ 0.2 & 92.3 $\pm$ 0.3 & 72.9 $\pm$ 0.7 & 94.1 $\pm$ 0.4 & 52.7 $\pm$ 0.6 & 72.6 $\pm$ 0.8 & 69.3 $\pm$ 0.6 \\
& KgCoOp (SGD) & 68.6 $\pm$ 0.1 & 67.5 $\pm$ 0.4 & 25.7 $\pm$ 0.3 & 57.4 $\pm$ 2.7 & 65.0 $\pm$ 0.5 & 86.1 $\pm$ 0.1 & 90.9 $\pm$ 0.4 & 69.9 $\pm$ 0.4 & 93.4 $\pm$ 0.5 & 47.4 $\pm$ 1.0 & 71.0 $\pm$ 0.6 & 67.5 $\pm$ 0.6 \\
\cmidrule(lr){2-14}
& ProGrad (Adam) & 65.4 $\pm$ 1.8 & 65.7 $\pm$ 1.0 & 24.0 $\pm$ 2.6 & 56.1 $\pm$ 4.3 & 63.6 $\pm$ 4.0 & 83.2 $\pm$ 1.2 & 89.2 $\pm$ 1.9 & 75.3 $\pm$ 1.3 & 92.0 $\pm$ 1.3 & 48.0 $\pm$ 1.4 & 71.5 $\pm$ 0.7 & 66.7 $\pm$ 2.0 \\
& ProGrad (SGD) & 65.1 $\pm$ 1.1 & 64.6 $\pm$ 2.0 & 15.0 $\pm$ 7.9 & 53.3 $\pm$ 4.1 & 54.0 $\pm$ 12.3 & 53.0 $\pm$ 19.8 & 64.4 $\pm$ 31.8 & 31.8 $\pm$ 22.8 & 75.3 $\pm$ 9.0 & 39.3 $\pm$ 6.3 & 67.9 $\pm$ 2.8 & 53.1 $\pm$ 10.9 \\
\bottomrule
\end{tabular}}

\label{tab:ablation_prompt_vs_sgd_ViTB16_with_aug}
\end{table*}
 
\begin{table*}
\centering
\caption{Ablation on prompt-tuning PEFT optimizers compared with dense full fine-tuning using SGD for CLIP (ViT-L/14). We report CoOp, KgCoOp, and ProGrad trained with Adam and SGD. Top-1 accuracy (\%) averaged over 3 fixed random seeds.}
\resizebox{1.\textwidth}{!}{
\begin{tabular}{llcccccccccccc}
\toprule
Shots & Method & ImageNet & SUN & Aircraft & EuroSAT & Cars & Food & Pets & Flowers & Caltech & DTD & UCF & Average \\ \midrule
\multirow{1}{*}{0} & CLIP & 73.5 & 67.7 & 30.0 & 55.1 & 76.8 & 90.3 & 93.1 & 76.2 & 95.2 & 52.4 & 73.7 & 71.3 \\
\midrule
\multirow{7}{*}{1} & CoOp (Adam) & 71.2 $\pm$ 0.6 & 68.8 $\pm$ 0.6 & 37.2 $\pm$ 0.4 & 58.8 $\pm$ 3.3 & 78.1 $\pm$ 0.9 & 88.9 $\pm$ 0.7 & 93.4 $\pm$ 0.3 & 89.3 $\pm$ 1.0 & 94.5 $\pm$ 0.3 & 53.8 $\pm$ 1.6 & 77.6 $\pm$ 0.4 & 73.8 $\pm$ 0.9 \\
& CoOp (SGD) & 74.7 $\pm$ 0.2 & 69.9 $\pm$ 0.5 & 35.8 $\pm$ 1.5 & 60.1 $\pm$ 3.4 & 77.6 $\pm$ 0.2 & 90.1 $\pm$ 0.0 & 93.9 $\pm$ 0.2 & 80.8 $\pm$ 1.7 & 93.3 $\pm$ 0.8 & 53.5 $\pm$ 1.7 & 76.4 $\pm$ 0.6 & 73.3 $\pm$ 1.0 \\
\cmidrule(lr){2-14}
& KgCoOp (Adam) & 75.9 $\pm$ 0.1 & 73.0 $\pm$ 0.0 & 36.9 $\pm$ 0.5 & 68.4 $\pm$ 4.3 & 78.7 $\pm$ 0.3 & 91.0 $\pm$ 0.1 & 94.8 $\pm$ 0.4 & 85.2 $\pm$ 0.4 & 96.0 $\pm$ 0.5 & 59.6 $\pm$ 0.3 & 80.0 $\pm$ 0.7 & 76.3 $\pm$ 0.7 \\
& KgCoOp (SGD) & 75.6 $\pm$ 0.0 & 71.6 $\pm$ 0.2 & 34.1 $\pm$ 0.8 & 69.1 $\pm$ 2.0 & 77.7 $\pm$ 0.7 & 90.9 $\pm$ 0.2 & 93.8 $\pm$ 0.1 & 80.9 $\pm$ 0.5 & 95.3 $\pm$ 0.6 & 56.6 $\pm$ 0.4 & 76.8 $\pm$ 0.9 & 74.8 $\pm$ 0.6 \\
\cmidrule(lr){2-14}
& ProGrad (Adam) & 70.9 $\pm$ 0.6 & 69.2 $\pm$ 0.9 & 35.3 $\pm$ 1.0 & 59.3 $\pm$ 4.4 & 72.7 $\pm$ 6.5 & 88.6 $\pm$ 0.1 & 93.4 $\pm$ 0.2 & 86.2 $\pm$ 1.5 & 94.2 $\pm$ 1.2 & 54.8 $\pm$ 1.9 & 77.4 $\pm$ 0.7 & 72.9 $\pm$ 1.7 \\
& ProGrad (SGD) & 69.8 $\pm$ 1.7 & 68.6 $\pm$ 0.9 & 32.8 $\pm$ 1.9 & 58.7 $\pm$ 2.0 & 74.1 $\pm$ 1.8 & 86.3 $\pm$ 2.1 & 89.8 $\pm$ 3.1 & 74.5 $\pm$ 2.7 & 84.3 $\pm$ 3.5 & 50.9 $\pm$ 1.5 & 74.1 $\pm$ 1.5 & 69.4 $\pm$ 2.1 \\
\bottomrule
\end{tabular}}

\label{tab:ablation_prompt_vs_sgd_ViTL14_with_aug}
\end{table*}

\begin{table*}
\centering
\caption{Ablation on low-rank PEFT optimizers compared with dense full fine-tuning using SGD for CLIP (ViT-B/32). We report LoRA and PiSSA trained with Adam and SGD. Top-1 accuracy (\%) averaged over 3 fixed random seeds.}
\resizebox{1.\textwidth}{!}{
\begin{tabular}{llcccccccccccc}
\toprule
Shots & Method & ImageNet & SUN & Aircraft & EuroSAT & Cars & Food & Pets & Flowers & Caltech & DTD & UCF & Average \\ \midrule
\multirow{1}{*}{0} & CLIP & 62.0 & 62.1 & 18.1 & 35.8 & 60.1 & 78.9 & 85.1 & 64.0 & 91.4 & 43.0 & 61.6 & 60.2 \\
\midrule
\multirow{5}{*}{1} & LoRA (Adam) & 62.8 $\pm$ 0.6 & 67.4 $\pm$ 0.2 & 21.5 $\pm$ 0.2 & 58.2 $\pm$ 2.2 & 62.9 $\pm$ 0.2 & 80.8 $\pm$ 0.2 & 89.0 $\pm$ 0.3 & 68.2 $\pm$ 0.3 & 93.1 $\pm$ 0.1 & 49.7 $\pm$ 0.7 & 70.1 $\pm$ 0.2 & 65.8 $\pm$ 0.5 \\
& LoRA (SGD) & 62.1 $\pm$ 0.0 & 62.1 $\pm$ 0.0 & 18.1 $\pm$ 0.0 & 36.0 $\pm$ 0.0 & 60.2 $\pm$ 0.0 & 78.9 $\pm$ 0.0 & 85.1 $\pm$ 0.0 & 64.0 $\pm$ 0.0 & 91.4 $\pm$ 0.0 & 42.9 $\pm$ 0.0 & 61.7 $\pm$ 0.0 & 60.2 $\pm$ 0.0 \\
\cmidrule(lr){2-14}
& PiSSA (Adam) & 60.7 $\pm$ 0.3 & 65.1 $\pm$ 0.2 & 22.2 $\pm$ 2.0 & 64.3 $\pm$ 6.5 & 62.1 $\pm$ 0.3 & 77.2 $\pm$ 0.7 & 86.4 $\pm$ 0.4 & 76.9 $\pm$ 1.5 & 92.9 $\pm$ 0.8 & 50.1 $\pm$ 1.2 & 70.6 $\pm$ 0.5 & 66.2 $\pm$ 1.3 \\
& PiSSA (SGD) & 64.4 $\pm$ 0.2 & 65.8 $\pm$ 0.1 & 20.6 $\pm$ 0.3 & 57.5 $\pm$ 2.5 & 62.0 $\pm$ 0.2 & 80.4 $\pm$ 0.2 & 87.6 $\pm$ 0.3 & 67.8 $\pm$ 0.0 & 92.3 $\pm$ 0.1 & 48.4 $\pm$ 0.4 & 66.8 $\pm$ 0.2 & 64.9 $\pm$ 0.4 \\
\cmidrule(lr){2-14}
& SGD & 64.6 $\pm$ 0.3 & 67.4 $\pm$ 0.1 & 22.3 $\pm$ 1.2 & 65.1 $\pm$ 2.4 & 63.5 $\pm$ 0.7 & 80.5 $\pm$ 0.4 & 89.0 $\pm$ 0.1 & 77.4 $\pm$ 1.0 & 93.1 $\pm$ 0.5 & 50.0 $\pm$ 0.1 & 71.7 $\pm$ 0.3 & 67.7 $\pm$ 0.6 \\
\midrule
\multirow{5}{*}{2} & LoRA (Adam) & 62.0 $\pm$ 0.2 & 67.8 $\pm$ 0.5 & 23.8 $\pm$ 0.4 & 65.9 $\pm$ 2.0 & 63.8 $\pm$ 0.6 & 80.4 $\pm$ 0.2 & 89.3 $\pm$ 0.9 & 75.7 $\pm$ 1.2 & 93.9 $\pm$ 0.5 & 55.1 $\pm$ 0.5 & 73.5 $\pm$ 0.8 & 68.3 $\pm$ 0.7 \\
\bottomrule
\end{tabular}}

\label{tab:ablation_lowrank_vs_sgd_ViTB32_with_aug}
\end{table*}

\begin{table*}
\centering
\caption{Ablation on low-rank PEFT optimizers compared with dense full fine-tuning using SGD for CLIP (ViT-B/16). We report LoRA and PiSSA trained with Adam and SGD. Top-1 accuracy (\%) averaged over 3 fixed random seeds.}
\resizebox{1.\textwidth}{!}{
\begin{tabular}{llcccccccccccc}
\toprule
Shots & Method & ImageNet & SUN & Aircraft & EuroSAT & Cars & Food & Pets & Flowers & Caltech & DTD & UCF & Average \\ \midrule
\multirow{1}{*}{0} & CLIP & 66.7 & 62.6 & 23.9 & 42.2 & 65.5 & 85.2 & 88.3 & 67.4 & 93.3 & 44.4 & 65.2 & 64.1 \\
\midrule
\multirow{5}{*}{1} & LoRA (Adam) & 68.1 $\pm$ 0.6 & 69.6 $\pm$ 0.4 & 28.9 $\pm$ 0.3 & 65.7 $\pm$ 2.9 & 69.7 $\pm$ 0.1 & 86.2 $\pm$ 0.1 & 92.4 $\pm$ 0.1 & 74.0 $\pm$ 0.5 & 94.1 $\pm$ 0.1 & 52.3 $\pm$ 0.5 & 74.7 $\pm$ 0.4 & 70.5 $\pm$ 0.5 \\
& LoRA (SGD) & 66.8 $\pm$ 0.0 & 62.6 $\pm$ 0.0 & 24.0 $\pm$ 0.1 & 42.4 $\pm$ 0.0 & 65.6 $\pm$ 0.0 & 85.2 $\pm$ 0.0 & 88.3 $\pm$ 0.0 & 67.5 $\pm$ 0.0 & 93.3 $\pm$ 0.0 & 44.4 $\pm$ 0.0 & 65.2 $\pm$ 0.0 & 64.1 $\pm$ 0.0 \\
\cmidrule(lr){2-14}
& PiSSA (Adam) & 66.0 $\pm$ 0.0 & 67.7 $\pm$ 0.7 & 29.9 $\pm$ 1.1 & 72.3 $\pm$ 3.0 & 69.7 $\pm$ 0.9 & 83.2 $\pm$ 0.7 & 90.5 $\pm$ 0.1 & 83.5 $\pm$ 0.5 & 93.4 $\pm$ 0.6 & 51.4 $\pm$ 0.8 & 75.4 $\pm$ 1.6 & 71.2 $\pm$ 0.9 \\
& PiSSA (SGD) & 69.3 $\pm$ 0.1 & 68.5 $\pm$ 0.1 & 27.9 $\pm$ 0.1 & 70.5 $\pm$ 2.7 & 67.5 $\pm$ 0.1 & 86.0 $\pm$ 0.1 & 90.7 $\pm$ 0.4 & 72.7 $\pm$ 0.1 & 93.7 $\pm$ 0.0 & 50.7 $\pm$ 0.7 & 71.3 $\pm$ 0.2 & 69.9 $\pm$ 0.4 \\
\cmidrule(lr){2-14}
& SGD & 70.1 $\pm$ 0.2 & 70.3 $\pm$ 0.2 & 30.5 $\pm$ 0.9 & 73.8 $\pm$ 2.7 & 70.8 $\pm$ 0.6 & 85.8 $\pm$ 0.2 & 92.7 $\pm$ 0.2 & 83.6 $\pm$ 0.6 & 94.0 $\pm$ 0.2 & 53.7 $\pm$ 0.8 & 76.1 $\pm$ 0.5 & 72.8 $\pm$ 0.6 \\
\midrule
\multirow{5}{*}{2} & LoRA (Adam) & 67.5 $\pm$ 0.3 & 70.7 $\pm$ 0.3 & 32.3 $\pm$ 0.6 & 75.2 $\pm$ 0.3 & 70.5 $\pm$ 0.3 & 85.7 $\pm$ 0.1 & 91.7 $\pm$ 0.8 & 82.4 $\pm$ 0.4 & 94.3 $\pm$ 0.5 & 58.0 $\pm$ 1.5 & 78.5 $\pm$ 0.8 & 73.3 $\pm$ 0.5 \\
\bottomrule
\end{tabular}}

\label{tab:ablation_lowrank_vs_sgd_ViTB16_with_aug}
\end{table*}
 
\begin{table*}
\centering
\caption{Ablation on low-rank PEFT optimizers compared with dense full fine-tuning using SGD for CLIP (ViT-L/14). We report LoRA and PiSSA trained with Adam and SGD. Top-1 accuracy (\%) averaged over 3 fixed random seeds.}
\resizebox{1.\textwidth}{!}{
\begin{tabular}{llcccccccccccc}
\toprule
Shots & Method & ImageNet & SUN & Aircraft & EuroSAT & Cars & Food & Pets & Flowers & Caltech & DTD & UCF & Average \\ \midrule
\multirow{1}{*}{0} & CLIP & 73.5 & 67.7 & 30.0 & 55.1 & 76.8 & 90.3 & 93.1 & 76.2 & 95.2 & 52.4 & 73.7 & 71.3 \\
\midrule
\multirow{5}{*}{1} & LoRA (Adam) & 72.3 $\pm$ 0.5 & 73.6 $\pm$ 0.4 & 41.1 $\pm$ 1.2 & 69.0 $\pm$ 5.6 & 80.7 $\pm$ 0.2 & 90.8 $\pm$ 0.2 & 94.4 $\pm$ 0.1 & 86.6 $\pm$ 1.4 & 95.8 $\pm$ 0.5 & 59.0 $\pm$ 0.3 & 80.8 $\pm$ 0.2 & 76.7 $\pm$ 1.0 \\
& LoRA (SGD) & 73.5 $\pm$ 0.0 & 67.7 $\pm$ 0.0 & 30.1 $\pm$ 0.0 & 55.3 $\pm$ 0.0 & 76.9 $\pm$ 0.0 & 90.3 $\pm$ 0.0 & 93.1 $\pm$ 0.0 & 76.2 $\pm$ 0.0 & 95.2 $\pm$ 0.0 & 52.4 $\pm$ 0.0 & 73.7 $\pm$ 0.0 & 71.3 $\pm$ 0.0 \\
\cmidrule(lr){2-14}
& PiSSA (Adam) & 72.2 $\pm$ 0.6 & 72.8 $\pm$ 0.5 & 39.2 $\pm$ 0.9 & 75.1 $\pm$ 1.1 & 78.4 $\pm$ 0.5 & 89.3 $\pm$ 0.5 & 93.4 $\pm$ 0.7 & 92.2 $\pm$ 0.8 & 95.7 $\pm$ 0.4 & 62.1 $\pm$ 0.6 & 82.9 $\pm$ 0.5 & 77.6 $\pm$ 0.6 \\
& PiSSA (SGD) & 76.4 $\pm$ 0.1 & 72.5 $\pm$ 0.1 & 37.3 $\pm$ 0.4 & 70.2 $\pm$ 2.3 & 79.3 $\pm$ 0.2 & 91.1 $\pm$ 0.1 & 94.3 $\pm$ 0.2 & 81.4 $\pm$ 0.4 & 95.6 $\pm$ 0.3 & 57.2 $\pm$ 0.5 & 77.8 $\pm$ 0.4 & 75.7 $\pm$ 0.4 \\
\cmidrule(lr){2-14}
& SGD & 76.6 $\pm$ 0.2 & 74.3 $\pm$ 0.1 & 41.1 $\pm$ 0.9 & 75.3 $\pm$ 4.1 & 80.7 $\pm$ 0.4 & 91.0 $\pm$ 0.1 & 94.6 $\pm$ 0.2 & 90.2 $\pm$ 0.8 & 95.6 $\pm$ 0.5 & 60.5 $\pm$ 0.4 & 81.9 $\pm$ 0.5 & 78.3 $\pm$ 0.7 \\
\midrule
\multirow{5}{*}{2} & LoRA (Adam) & 72.7 $\pm$ 0.1 & 74.1 $\pm$ 0.2 & 43.9 $\pm$ 1.0 & 75.3 $\pm$ 1.7 & 82.6 $\pm$ 0.4 & 90.5 $\pm$ 0.3 & 94.5 $\pm$ 0.8 & 93.3 $\pm$ 0.2 & 96.9 $\pm$ 0.5 & 65.3 $\pm$ 0.7 & 83.6 $\pm$ 0.7 & 79.3 $\pm$ 0.6 \\
\bottomrule
\end{tabular}}

\label{tab:ablation_lowrank_vs_sgd_ViTL14_with_aug}
\end{table*}

\begin{table*}
\centering
\caption{Ablation on sparse and subset-based PEFT optimizers compared with dense full fine-tuning using SGD for CLIP (ViT-B/32). We report LayerNorm, BitFit, and SAFT trained with Adam and SGD. Top-1 accuracy (\%) averaged over 3 fixed random seeds.}
\resizebox{1.\textwidth}{!}{
\begin{tabular}{llcccccccccccc}
\toprule
Shots & Method & ImageNet & SUN & Aircraft & EuroSAT & Cars & Food & Pets & Flowers & Caltech & DTD & UCF & Average \\ \midrule
\multirow{1}{*}{0} & CLIP & 62.0 & 62.1 & 18.1 & 35.8 & 60.1 & 78.9 & 85.1 & 64.0 & 91.4 & 43.0 & 61.6 & 60.2 \\
\midrule
\multirow{7}{*}{1} & LayerNorm (Adam) & 63.9 $\pm$ 0.1 & 64.3 $\pm$ 0.1 & 19.6 $\pm$ 0.0 & 40.3 $\pm$ 0.5 & 61.3 $\pm$ 0.2 & 79.8 $\pm$ 0.1 & 86.7 $\pm$ 0.1 & 65.8 $\pm$ 0.0 & 91.9 $\pm$ 0.1 & 44.4 $\pm$ 0.3 & 63.4 $\pm$ 0.1 & 62.0 $\pm$ 0.1 \\
& LayerNorm (SGD) & 62.1 $\pm$ 0.0 & 62.1 $\pm$ 0.0 & 18.1 $\pm$ 0.0 & 35.9 $\pm$ 0.0 & 60.2 $\pm$ 0.0 & 78.9 $\pm$ 0.0 & 85.1 $\pm$ 0.0 & 64.0 $\pm$ 0.0 & 91.4 $\pm$ 0.0 & 42.9 $\pm$ 0.1 & 61.7 $\pm$ 0.0 & 60.2 $\pm$ 0.0 \\
\cmidrule(lr){2-14}
& BitFit (Adam) & 64.1 $\pm$ 0.3 & 66.8 $\pm$ 0.1 & 22.0 $\pm$ 0.4 & 63.0 $\pm$ 2.9 & 62.2 $\pm$ 0.3 & 80.3 $\pm$ 0.4 & 88.9 $\pm$ 0.2 & 70.2 $\pm$ 0.6 & 93.1 $\pm$ 0.2 & 49.9 $\pm$ 0.1 & 70.2 $\pm$ 0.1 & 66.4 $\pm$ 0.5 \\
& BitFit (SGD) & 63.5 $\pm$ 0.1 & 64.2 $\pm$ 0.1 & 20.0 $\pm$ 0.0 & 42.0 $\pm$ 1.5 & 60.8 $\pm$ 0.1 & 79.5 $\pm$ 0.1 & 86.7 $\pm$ 0.1 & 65.6 $\pm$ 0.2 & 91.7 $\pm$ 0.1 & 44.5 $\pm$ 0.3 & 63.8 $\pm$ 0.2 & 62.0 $\pm$ 0.3 \\
\cmidrule(lr){2-14}
& SAFT (Adam) & 59.7 $\pm$ 0.7 & 64.4 $\pm$ 0.2 & 21.0 $\pm$ 1.0 & 66.2 $\pm$ 1.8 & 57.5 $\pm$ 1.0 & 78.0 $\pm$ 0.6 & 85.3 $\pm$ 1.1 & 73.3 $\pm$ 0.7 & 93.3 $\pm$ 0.7 & 47.6 $\pm$ 0.9 & 69.7 $\pm$ 0.6 & 65.1 $\pm$ 0.8 \\
& SAFT (SGD) & 64.3 $\pm$ 0.2 & 66.1 $\pm$ 0.1 & 21.7 $\pm$ 0.4 & 60.6 $\pm$ 5.3 & 61.1 $\pm$ 0.5 & 80.0 $\pm$ 0.3 & 88.5 $\pm$ 0.7 & 69.0 $\pm$ 0.8 & 93.0 $\pm$ 0.4 & 49.1 $\pm$ 0.3 & 67.7 $\pm$ 0.4 & 65.6 $\pm$ 0.9 \\
\bottomrule
\end{tabular}}

\label{tab:ablation_sparse_vs_sgd_ViTB32_with_aug}
\end{table*}
 
\begin{table*}
\centering
\caption{Ablation on sparse and subset-based PEFT optimizers compared with dense full fine-tuning using SGD for CLIP (ViT-B/16). We report LayerNorm, BitFit, and SAFT trained with Adam and SGD. Top-1 accuracy (\%) averaged over 3 fixed random seeds.}
\resizebox{1.\textwidth}{!}{
\begin{tabular}{llcccccccccccc}
\toprule
Shots & Method & ImageNet & SUN & Aircraft & EuroSAT & Cars & Food & Pets & Flowers & Caltech & DTD & UCF & Average \\ \midrule
\multirow{1}{*}{0} & CLIP & 66.7 & 62.6 & 23.9 & 42.2 & 65.5 & 85.2 & 88.3 & 67.4 & 93.3 & 44.4 & 65.2 & 64.1 \\
\midrule
\multirow{7}{*}{1} & LayerNorm (Adam) & 68.7 $\pm$ 0.1 & 66.7 $\pm$ 0.1 & 25.5 $\pm$ 0.2 & 49.6 $\pm$ 2.0 & 66.7 $\pm$ 0.2 & 85.7 $\pm$ 0.1 & 89.4 $\pm$ 0.1 & 69.9 $\pm$ 0.1 & 93.5 $\pm$ 0.1 & 45.9 $\pm$ 0.2 & 67.8 $\pm$ 0.2 & 66.3 $\pm$ 0.3 \\
& LayerNorm (SGD) & 66.7 $\pm$ 0.0 & 62.6 $\pm$ 0.0 & 23.9 $\pm$ 0.0 & 42.3 $\pm$ 0.1 & 65.6 $\pm$ 0.0 & 85.2 $\pm$ 0.0 & 88.3 $\pm$ 0.0 & 67.5 $\pm$ 0.0 & 93.3 $\pm$ 0.0 & 44.4 $\pm$ 0.1 & 65.3 $\pm$ 0.0 & 64.1 $\pm$ 0.0 \\
\cmidrule(lr){2-14}
& BitFit (Adam) & 69.4 $\pm$ 0.2 & 69.3 $\pm$ 0.2 & 29.5 $\pm$ 0.3 & 72.0 $\pm$ 1.9 & 69.1 $\pm$ 0.4 & 85.5 $\pm$ 0.2 & 92.3 $\pm$ 0.3 & 74.8 $\pm$ 0.3 & 93.9 $\pm$ 0.2 & 52.1 $\pm$ 0.6 & 74.6 $\pm$ 0.4 & 71.1 $\pm$ 0.5 \\
& BitFit (SGD) & 68.0 $\pm$ 0.1 & 66.9 $\pm$ 0.2 & 25.5 $\pm$ 0.1 & 52.0 $\pm$ 1.8 & 66.3 $\pm$ 0.1 & 85.5 $\pm$ 0.1 & 89.3 $\pm$ 0.1 & 69.5 $\pm$ 0.2 & 93.4 $\pm$ 0.0 & 46.4 $\pm$ 0.2 & 67.9 $\pm$ 0.3 & 66.4 $\pm$ 0.3 \\
\cmidrule(lr){2-14}
& SAFT (Adam) & 66.1 $\pm$ 0.1 & 67.2 $\pm$ 0.3 & 27.4 $\pm$ 0.8 & 68.9 $\pm$ 3.5 & 64.8 $\pm$ 0.7 & 84.1 $\pm$ 0.4 & 89.9 $\pm$ 0.9 & 80.7 $\pm$ 0.9 & 93.9 $\pm$ 0.2 & 52.2 $\pm$ 1.0 & 73.7 $\pm$ 0.4 & 69.9 $\pm$ 0.8 \\
& SAFT (SGD) & 70.0 $\pm$ 0.1 & 68.6 $\pm$ 0.2 & 28.9 $\pm$ 0.3 & 65.2 $\pm$ 1.0 & 68.0 $\pm$ 0.3 & 85.6 $\pm$ 0.5 & 91.3 $\pm$ 0.4 & 73.6 $\pm$ 0.2 & 94.3 $\pm$ 0.1 & 51.8 $\pm$ 0.6 & 71.3 $\pm$ 0.1 & 69.9 $\pm$ 0.4 \\
\bottomrule
\end{tabular}}

\label{tab:ablation_sparse_vs_sgd_ViTB16_with_aug}
\end{table*}
 
\begin{table*}
\centering
\caption{Ablation on sparse and subset-based PEFT optimizers compared with dense full fine-tuning using SGD for CLIP (ViT-L/14). We report LayerNorm, BitFit, and SAFT trained with Adam and SGD. Top-1 accuracy (\%) averaged over 3 fixed random seeds.}
\resizebox{1.\textwidth}{!}{
\begin{tabular}{llcccccccccccc}
\toprule
Shots & Method & ImageNet & SUN & Aircraft & EuroSAT & Cars & Food & Pets & Flowers & Caltech & DTD & UCF & Average \\ \midrule
\multirow{1}{*}{0} & CLIP & 73.5 & 67.7 & 30.0 & 55.1 & 76.8 & 90.3 & 93.1 & 76.2 & 95.2 & 52.4 & 73.7 & 71.3 \\
\midrule
\multirow{7}{*}{1} & LayerNorm (Adam) & 75.9 $\pm$ 0.1 & 71.2 $\pm$ 0.1 & 33.7 $\pm$ 0.1 & 61.9 $\pm$ 0.8 & 78.3 $\pm$ 0.2 & 90.9 $\pm$ 0.1 & 93.9 $\pm$ 0.1 & 78.7 $\pm$ 0.1 & 95.4 $\pm$ 0.2 & 54.1 $\pm$ 0.2 & 75.7 $\pm$ 0.3 & 73.6 $\pm$ 0.2 \\
& LayerNorm (SGD) & 73.5 $\pm$ 0.0 & 67.8 $\pm$ 0.0 & 30.1 $\pm$ 0.1 & 55.3 $\pm$ 0.0 & 76.8 $\pm$ 0.0 & 90.3 $\pm$ 0.0 & 93.1 $\pm$ 0.0 & 76.2 $\pm$ 0.0 & 95.2 $\pm$ 0.0 & 52.4 $\pm$ 0.0 & 73.7 $\pm$ 0.0 & 71.3 $\pm$ 0.0 \\
\cmidrule(lr){2-14}
& BitFit (Adam) & 75.9 $\pm$ 0.4 & 73.6 $\pm$ 0.1 & 39.7 $\pm$ 0.9 & 71.4 $\pm$ 4.1 & 80.0 $\pm$ 0.1 & 90.9 $\pm$ 0.3 & 94.5 $\pm$ 0.1 & 83.2 $\pm$ 1.0 & 95.8 $\pm$ 0.2 & 59.5 $\pm$ 0.4 & 80.5 $\pm$ 0.3 & 76.8 $\pm$ 0.7 \\
& BitFit (SGD) & 74.6 $\pm$ 0.1 & 69.1 $\pm$ 0.1 & 31.6 $\pm$ 0.1 & 59.6 $\pm$ 0.5 & 77.3 $\pm$ 0.1 & 90.4 $\pm$ 0.0 & 93.3 $\pm$ 0.0 & 76.7 $\pm$ 0.1 & 95.2 $\pm$ 0.0 & 52.6 $\pm$ 0.2 & 74.5 $\pm$ 0.1 & 72.3 $\pm$ 0.1 \\
\cmidrule(lr){2-14}
& SAFT (Adam) & 74.2 $\pm$ 0.6 & 73.7 $\pm$ 0.1 & 38.4 $\pm$ 1.6 & 74.2 $\pm$ 3.5 & 78.5 $\pm$ 0.0 & 89.4 $\pm$ 0.5 & 91.0 $\pm$ 1.9 & 89.1 $\pm$ 0.8 & 96.2 $\pm$ 0.8 & 59.6 $\pm$ 1.7 & 81.3 $\pm$ 0.8 & 76.9 $\pm$ 1.1 \\
& SAFT (SGD) & 75.9 $\pm$ 0.0 & 71.7 $\pm$ 0.1 & 36.7 $\pm$ 1.0 & 66.9 $\pm$ 2.2 & 78.5 $\pm$ 0.1 & 90.9 $\pm$ 0.1 & 94.2 $\pm$ 0.1 & 80.6 $\pm$ 0.2 & 95.5 $\pm$ 0.5 & 56.4 $\pm$ 1.0 & 77.7 $\pm$ 0.0 & 75.0 $\pm$ 0.5 \\
\bottomrule
\end{tabular}}

\label{tab:ablation_sparse_vs_sgd_ViTL14_with_aug}
\end{table*}

Tables~\ref{tab:ablation_adapter_vs_sgd_ViTB32_with_aug}--\ref{tab:ablation_sparse_vs_sgd_ViTL14_with_aug} test whether the advantage of dense SGD over PEFT is only an optimizer artifact. Each table compares one PEFT family with dense full fine-tuning. For each of the 11 PEFT methods included in this ablation, we report both Adam and SGD. The last row in each shot block reports the dense full-fine-tuning SGD baseline from Tables~\ref{tab:few_shot_ViTB32_with_aug}--\ref{tab:few_shot_ViTL14_with_aug}. This design separates the effect of the trainable parameterization from the effect of the optimizer.

Across the 12 tables and 5 shot levels, dense SGD gives the best average accuracy in all 60 backbone--family--shot comparisons. For each row, we select the strongest PEFT method after allowing either Adam or SGD. Dense SGD is still higher by 2.57 percentage points on average, with a minimum margin of 0.4 points. If we compare only with Adam-trained PEFT methods, the mean margin is 3.77 points. If we compare only with SGD-trained PEFT methods, the mean margin is 4.65 points. Thus, replacing Adam with SGD inside PEFT does not close the gap to dense full fine-tuning.

The adapter-style results show that the optimizer effect is method-dependent. Adapter and TIP-Adapter do not benefit from SGD. Their SGD rows are usually slightly below their Adam rows. LP++ is different. It is highly unstable with Adam in the one-shot regime, but it becomes much stronger when trained with SGD. Averaged over all adapter rows, switching LP++ from Adam to SGD improves its average accuracy by 25.7 points. Even after this correction, dense SGD remains ahead of the best adapter-style PEFT method by 3.99 points on average. The margin over the best Adam-trained adapter method is 8.81 points.

The prompt-tuning tables give the opposite optimizer pattern. CoOp, KgCoOp, and ProGrad are generally better with Adam than with SGD. The mean loss from switching prompt methods to SGD is 3.94 points. Dense SGD still outperforms the strongest prompt-tuning method by 3.49 points on average. Its margin over the strongest SGD-trained prompt method is 6.30 points. This shows that the dense-SGD advantage is not obtained by penalizing prompt methods through Adam. In this family, Adam is usually the stronger PEFT optimizer.

The low-rank tables provide the most competitive PEFT baselines. The low-rank optimizer tables include LoRA and PiSSA. Adam is better than SGD for these methods in the reported average rows. Their average SGD-minus-Adam difference is -7.22 points. Dense SGD, nevertheless, remains the best averaged method in every low-rank row. The mean margin over the strongest low-rank PEFT method is smaller than in the other families, at 0.87 points, but it is positive for all backbones and shot levels. The smallest gap is 0.4 points for ViT-L/14 at 16 shots.

The sparse and subset-based tables show a similar trend. Adam is usually the better optimizer for LayerNorm tuning, BitFit, and SAFT. The best Adam-trained sparse method is 2.13 points above the best SGD-trained sparse method on average. Dense SGD still exceeds the best sparse PEFT method by 1.93 points on average, with a minimum margin of 0.7 points. The gap over SGD-trained sparse PEFT methods is larger, at 4.06 points on average.

Optimizer choice within PEFT matters. Adam can severely hurt some PEFT methods, especially LP++, and SGD can rescue them. However, Adam is also the better optimizer for several PEFT families, including prompt, low-rank, and sparse methods. Dense full fine-tuning with SGD remains better after each PEFT baseline included in this ablation is given both optimizer choices. Therefore, the conclusion is not that PEFT loses only because it is commonly trained with Adam. Rather, dense SGD combines a larger trainable space with an optimizer that avoids unreliable few-shot preconditioning. This combination yields a stronger average few-shot generalization than the PEFT parameterizations tested here.

\section{Why Does Adam Fail?}
\label{sec:why_does_adam_fail}

We now isolate the failure mechanism of Adam in few-shot CLIP adaptation. Unless stated otherwise, this section uses CLIP (ViT-B/16) with $K{=}1$ shot per class on EuroSAT, Oxford-Pets, and DTD. We use no augmentation. Both encoders are trainable, and all transformer blocks are trainable. Updates use the full support set, so one optimizer step equals one epoch. Let $\Theta_t\in\mathbb{R}^{d}$ denote the trainable parameter vector at step $t$, where $d$ is the number of trainable scalar parameters. Let $L_{\text{few}}, L_{\text{full}}, L_{\text{test}}:\mathbb{R}^{d}\to\mathbb{R}$ denote the support, full-train, and test cross-entropy losses.

\subsection{Second-moment miscalibration}
\label{subsec:second_moment_miscalibration}

Experiment~1 audits Adam's diagonal denominator along a stable SGD trajectory. The trajectory is generated by SGD, not by Adam. At each checkpoint, we ask what Adam's diagonal second moment would be on the same one-shot support set. This separates the denominator estimation error from instability caused by Adam's own earlier updates.

Let $\mathcal{E}=\{\text{vision},\text{text}\}$ index the two CLIP encoders. For encoder $e\in\mathcal{E}$, let $\Theta_t^{(e)}\in\mathbb{R}^{d_e}$ be the encoder-restricted trainable parameter vector at checkpoint $\Theta_t$. Here $d_e$ is the number of trainable scalar parameters in encoder $e$, and $d_{\text{vision}}+d_{\text{text}}=d$. Let $\mathcal{D}_{\text{few}}$ denote the one-shot support set, and let $\mathcal{D}_{\text{full}}$ denote the full training split. For $\mathcal{D}\in\{\mathcal{D}_{\text{few}},\mathcal{D}_{\text{full}}\}$, let $L_{\mathcal{D}}$ be the empirical loss on $\mathcal{D}$. Thus $L_{\mathcal{D}_{\text{few}}}=L_{\text{few}}$ and $L_{\mathcal{D}_{\text{full}}}=L_{\text{full}}$.

For mini-batch $\mathcal{B}\subset\mathcal{D}$, the encoder-restricted batch gradient is
\begin{equation}
    g^{(e)}_{\mathcal{B},t}
    = \nabla_{\Theta^{(e)}} L_{\mathcal{B}}(\Theta_t)
    \in \mathbb{R}^{d_e}.
\end{equation}
The corresponding first moment and diagonal second moment are
\begin{equation}
    m^{(e)}_{\mathcal{D},t}
    = \nabla_{\Theta^{(e)}} L_{\mathcal{D}}(\Theta_t)
    \in \mathbb{R}^{d_e},
    \qquad
    v^{(e)}_{\mathcal{D},t}
    = \mathbb{E}_{\mathcal{B}\subset\mathcal{D}}
    \left[g^{(e)}_{\mathcal{B},t}\odot g^{(e)}_{\mathcal{B},t}\right]
    \in \mathbb{R}_{\ge 0}^{d_e}.
    \label{eq:adam_diagnostic_moments}
\end{equation}
The operator $\odot$ denotes elementwise multiplication. The expectation is over the diagnostic batches used to estimate the empirical diagonal statistic. This is the instantaneous statistic underlying Adam's second-moment denominator, before exponential averaging and bias correction.

The abbreviations are $m^{(e)}_{\text{few},t}=m^{(e)}_{\mathcal{D}_{\text{few}},t}$ and $v^{(e)}_{\text{few},t}=v^{(e)}_{\mathcal{D}_{\text{few}},t}$. The full-training abbreviations are analogous. With Adam stabilizer $\varepsilon>0$, we compare the encoder-wise Adam-like directions
\begin{equation}
    u^{(e)}_{ff,t}
    = \frac{m^{(e)}_{\text{few},t}}
           {\sqrt{v^{(e)}_{\text{few},t}+\varepsilon}},
    \qquad
    u^{(e)}_{fr,t}
    = \frac{m^{(e)}_{\text{few},t}}
           {\sqrt{v^{(e)}_{\text{full},t}+\varepsilon}},
    \qquad
    u^{(e)}_{rr,t}
    = \frac{m^{(e)}_{\text{full},t}}
           {\sqrt{v^{(e)}_{\text{full},t}+\varepsilon}}.
    \label{eq:adam_diagnostic_directions}
\end{equation}
All square roots and divisions are elementwise. The subscript $f$ denotes the few-shot statistic. The subscript $r$ denotes the full-data reference statistic. Global vectors $m_{\text{few},t}, m_{\text{full},t}, v_{\text{few},t}, v_{\text{full},t}, u_{ff,t}, u_{fr,t}, u_{rr,t}\in\mathbb{R}^{d}$ are obtained by concatenating the encoder-restricted vectors. When the encoder superscript is omitted, the notation refers to this global vector. The comparison between $u^{(e)}_{ff,t}$ and $u^{(e)}_{fr,t}$ isolates the denominator because the numerator is fixed. The comparison between $u^{(e)}_{fr,t}$ and $u^{(e)}_{rr,t}$ isolates the remaining numerator error because the denominator is fixed.

For coordinate $i\in\{1,\ldots,d_e\}$, the second-moment log-ratio is
\begin{equation}
    \zeta^{(e)}_{t,i}
    = \log_{10}\!\left(
    \frac{v^{(e)}_{\text{few},t,i}+\varepsilon}
         {v^{(e)}_{\text{full},t,i}+\varepsilon}
    \right).
\end{equation}
A negative value means that the one-shot support set underestimates the full-data second moment in that coordinate. We summarize the mismatch magnitude by
\begin{equation}
    M^{(e)}_t
    = \frac{1}{d_e}\sum_{i=1}^{d_e} |\zeta^{(e)}_{t,i}| .
    \label{eq:adam_diagnostic_mismatch}
\end{equation}
The statistic $M^{(e)}_t$ is measured in base-$10$ logarithmic units. A value of $M^{(e)}_t=1$ corresponds to an average one-order-of-magnitude discrepancy.

\begin{figure*}[t]
    \centering
    \includegraphics[width=0.98\linewidth]{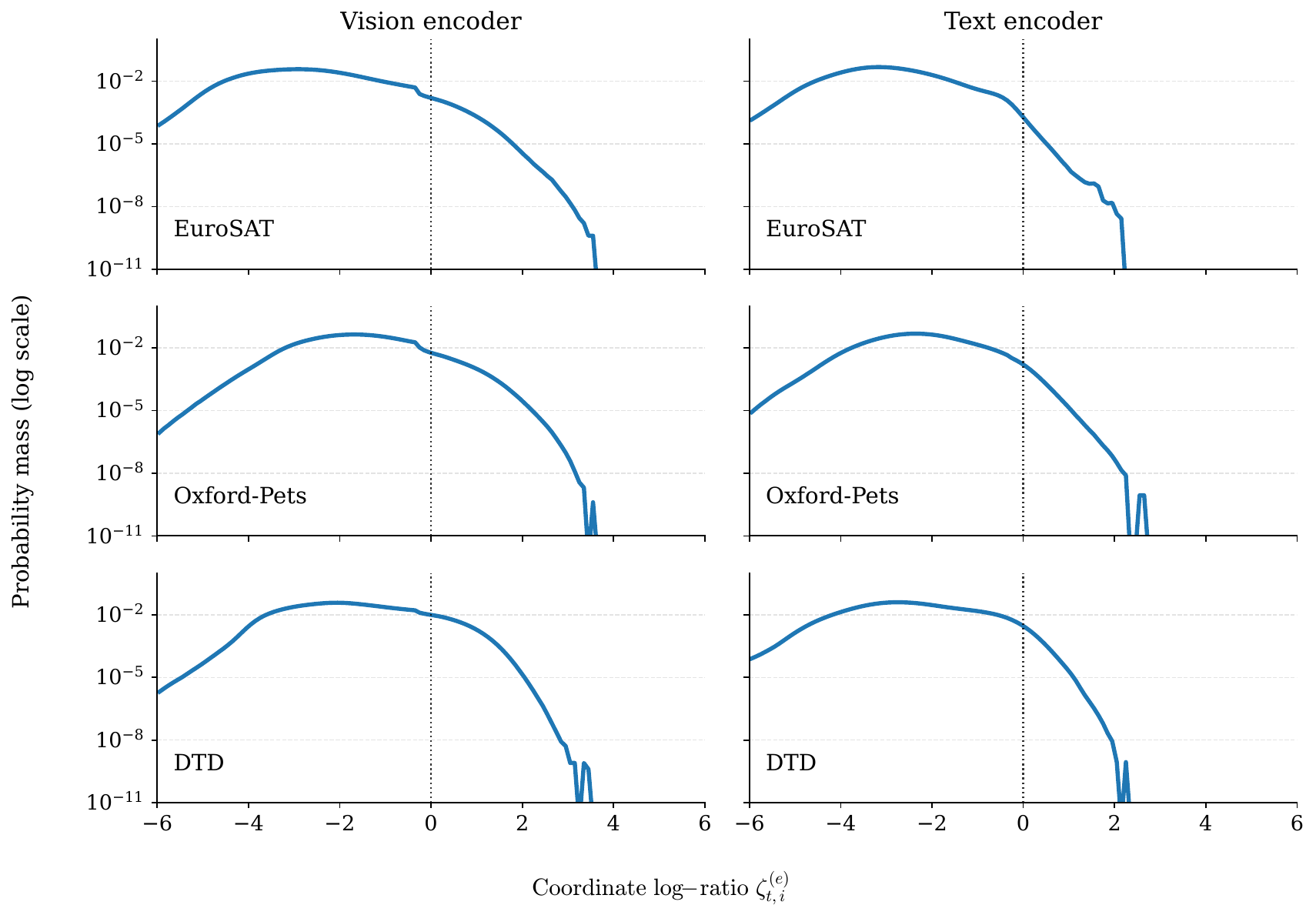}
    \caption{Experiment~1: coordinate log-ratio histograms for the SGD trajectory. Each panel shows $\zeta^{(e)}_{t,i}=\log_{10}((v^{(e)}_{\text{few},t,i}+\varepsilon)/(v^{(e)}_{\text{full},t,i}+\varepsilon))$, aggregated over diagnostic checkpoints. The dotted vertical line is zero. Mass to the left of zero means that the one-shot second moment is smaller than the full-data second moment.}
    \label{fig:adam_fail_exp0_histograms}
\end{figure*}

Figure~\ref{fig:adam_fail_exp0_histograms} shows that the denominator error is widespread. Across all dataset--encoder pairs, most of the probability mass lies to the left of zero. The one-shot statistic, therefore, underestimates the full-data denominator for most trainable coordinates. The effect is strongest in the text encoder. In the EuroSAT text, almost all coordinates lie on the underestimated side along the trajectory. DTD vision is the least extreme case, but even there, the mass left of zero is about $94\%$.

\begin{figure*}[t]
    \centering
    \includegraphics[width=0.98\linewidth]{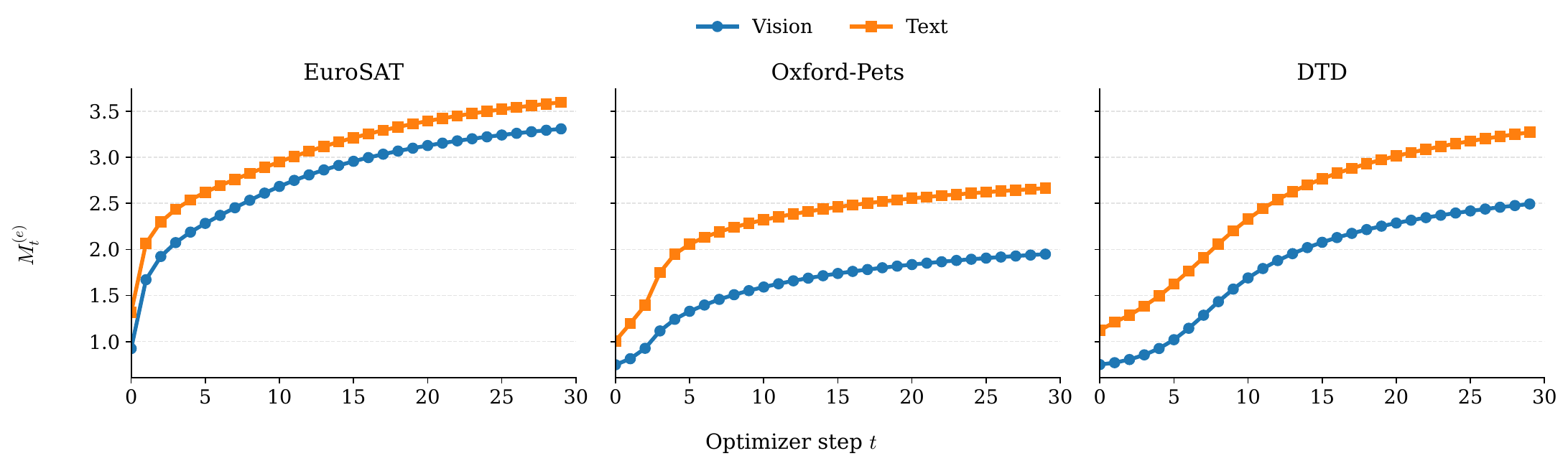}
    \caption{Experiment~1: second-moment miscalibration over time. The plotted statistic is $M^{(e)}_t=d_e^{-1}\sum_i |\log_{10}((v^{(e)}_{\text{few},t,i}+\varepsilon)/(v^{(e)}_{\text{full},t,i}+\varepsilon))|$. Larger values indicate a larger multiplicative mismatch between the one-shot and full-data diagonal second moments.}
    \label{fig:adam_fail_exp0_miscalibration}
\end{figure*}

Figure~\ref{fig:adam_fail_exp0_miscalibration} shows that the mismatch grows over the SGD trajectory. At initialization, $M^{(e)}_0$ already lies between $0.75$ and $1.32$ across the six dataset--encoder pairs. By the final diagnostic step, it increases to the range $1.95$--$3.60$. EuroSAT has the largest final mismatch, especially in the text encoder. Oxford-Pets has the mildest final mismatch, but it still grows throughout training. Thus, the denominator is not merely noisy at initialization. It becomes progressively less calibrated as the support loss is fitted.

\begin{figure*}[t]
    \centering
    \includegraphics[width=0.98\linewidth]{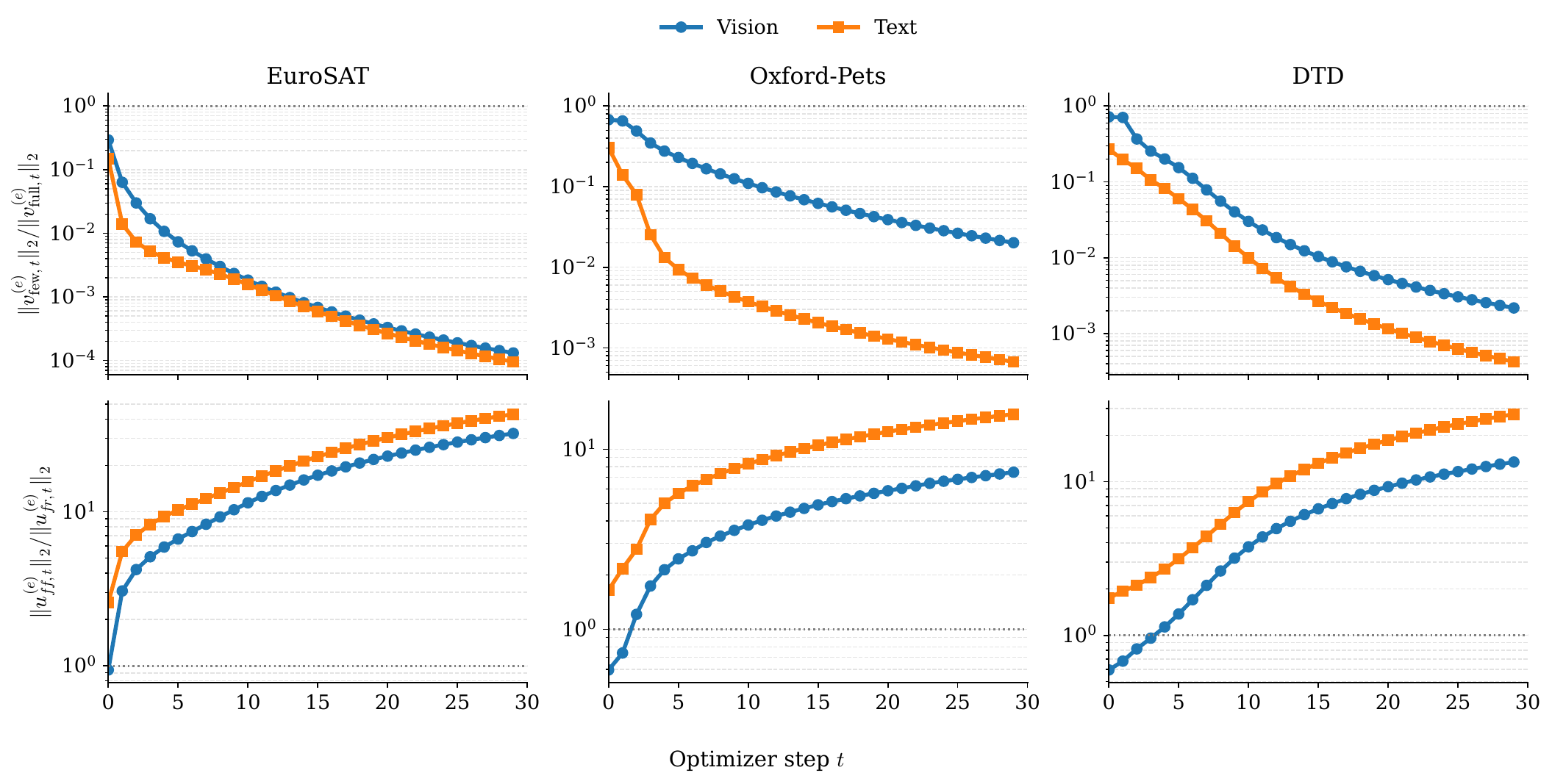}
    \caption{Experiment~1: scale consequences of the second-moment mismatch. Top row: encoder-wise norm ratio $\lVert v^{(e)}_{\text{few},t}\rVert_2/\lVert v^{(e)}_{\text{full},t}\rVert_2$. Bottom row: denominator-only adaptive norm inflation $\lVert u^{(e)}_{ff,t}\rVert_2/\lVert u^{(e)}_{fr,t}\rVert_2$. The dotted horizontal line marks perfect calibration.}
    \label{fig:adam_fail_exp0_scale}
\end{figure*}

The same phenomenon appears in the global second-moment norm. Figure~\ref{fig:adam_fail_exp0_scale} reports the ratio $\lVert v^{(e)}_{\text{few},t}\rVert_2/\lVert v^{(e)}_{\text{full},t}\rVert_2$. At initialization, this ratio is already below one for every dataset and encoder, with values from $0.15$ to $0.72$. By the final step, it falls between $9.5\times 10^{-5}$ and $2.0\times 10^{-2}$. The bottom row of Figure~\ref{fig:adam_fail_exp0_scale} isolates the induced scale error by comparing $u^{(e)}_{ff,t}$ with $u^{(e)}_{fr,t}$. These two directions use the same numerator and differ only in the denominator. The resulting norm ratio $\lVert u^{(e)}_{ff,t}\rVert_2/\lVert u^{(e)}_{fr,t}\rVert_2$ reaches about $7.5$--$43$ by the end of training. Hence, a few-shot Adam denominator would make the same numerator much larger than a full-data denominator would.

\begin{figure*}[t]
    \centering
    \includegraphics[width=0.98\linewidth]{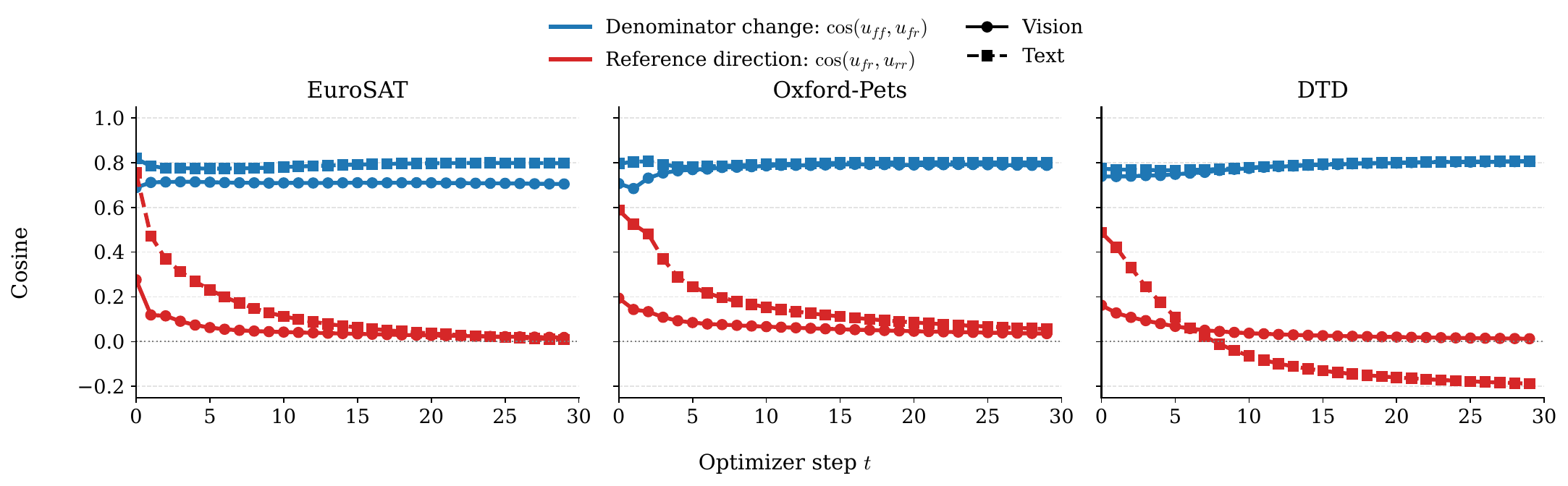}
    \caption{Experiment~1: adaptive-direction geometry. The blue curves show the encoder-wise cosine $\cos(u^{(e)}_{ff,t},u^{(e)}_{fr,t})$, which isolates the effect of changing only the denominator. The red curves show $\cos(u^{(e)}_{fr,t},u^{(e)}_{rr,t})$, which keeps the full-data denominator fixed and measures the remaining numerator mismatch.}
    \label{fig:adam_fail_exp0_geometry}
\end{figure*}

Figure~\ref{fig:adam_fail_exp0_geometry} separates scale from direction. The encoder-wise cosine $\cos(u^{(e)}_{ff,t},u^{(e)}_{fr,t})$ remains relatively high throughout training. It stays in the range $0.69$--$0.82$ at initialization and $0.70$--$0.81$ at the final diagnostic step. Replacing the few-shot denominator with the full-data denominator changes the magnitude more than it changes the direction. In contrast, $\cos(u^{(e)}_{fr,t},u^{(e)}_{rr,t})$ decays toward zero or becomes negative. This comparison uses the same full-data denominator in both directions, so it measures the remaining numerator error. The result shows a two-stage failure. The denominator creates a scale error immediately. Later, as SGD fits the support set, the few-shot numerator also drifts away from the full-data numerator.

\begin{figure*}[t]
    \centering
    \includegraphics[width=0.98\linewidth]{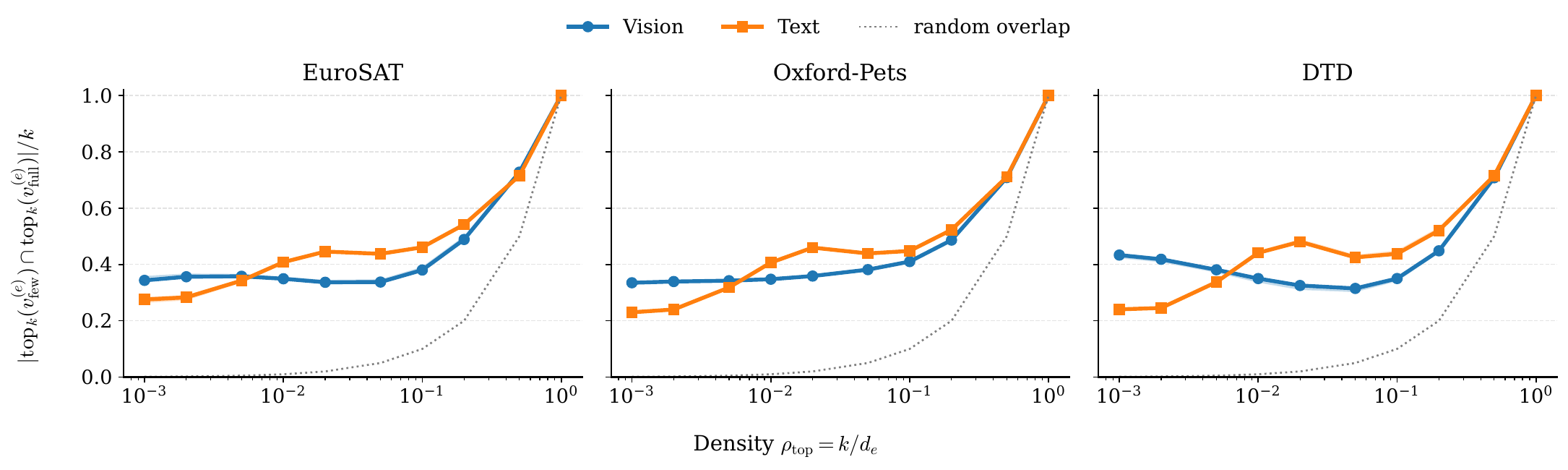}
    \caption{Experiment~1: overlap between salient coordinates of the one-shot and full-data second moments. The density is $\rho_{\mathrm{top}}=k/d_e$. The dotted line is the random-overlap baseline. The overlap is far above random at small densities, which indicates that the few-shot statistic preserves part of the support while miscalibrating its scale.}
    \label{fig:adam_fail_exp0_topk}
\end{figure*}

Finally, Figure~\ref{fig:adam_fail_exp0_topk} shows that the failure is not caused by a complete loss of coordinate support. Let $\operatorname{top}_k(a)$ be the set of the $k$ largest coordinates of a nonnegative vector $a\in\mathbb{R}_{\ge 0}^{d_e}$, and let $k=\lceil \rho_{\mathrm{top}} d_e\rceil$ for density $\rho_{\mathrm{top}}\in(0,1]$. We measure
\begin{equation}
    O^{(e)}_t(\rho_{\mathrm{top}})
    = \frac{|\operatorname{top}_k(v^{(e)}_{\text{few},t})\cap
              \operatorname{top}_k(v^{(e)}_{\text{full},t})|}{k}
    \in [0,1].
\end{equation}
At density $\rho_{\mathrm{top}}=0.01$, the overlap is about $0.34$--$0.54$ at initialization and $0.35$--$0.44$ at the final step. These values are far above the random baseline $\rho_{\mathrm{top}}=0.01$. Thus, the few-shot second moment often identifies many of the same coordinates with large second moments. Its main defect is calibration. It assigns them the wrong magnitude.

Taken together, Figures~\ref{fig:adam_fail_exp0_histograms}--\ref{fig:adam_fail_exp0_topk} identify the mechanism behind Adam's early failure. The one-shot second moment is systematically too small. The mismatch increases as the support set is fitted. This produces a large denominator-induced scale error while preserving part of the salient coordinate support. The denominator error appears before severe numerator drift, but the numerator becomes unreliable later in training.

\begin{findingbox}[Conclusion]
Adam's denominator is miscalibrated on a stable SGD trajectory. The few-shot second moment underestimates the full-data second moment for most coordinates, and the mismatch grows during training. This makes the Adam-like update too large when the few-shot numerator is held fixed. The high top-$k$ overlap shows that the support of the second moment is not completely wrong. The dominant error is scale calibration. Later checkpoints also show numerator drift, which explains why denominator repair alone may not be sufficient after the model has already moved away from the pretrained solution.
\end{findingbox}

\subsection{Is the problem already there at $t{=}0$, before any adaptation?}
\label{subsec:adam_init_audit}

Experiment~2 evaluates the diagnostics from Eqs.~\eqref{eq:adam_diagnostic_moments}--\eqref{eq:adam_diagnostic_mismatch} at the pretrained checkpoint $\Theta_0$, before any adaptation step. This audit is trajectory-independent because no optimizer update has been applied. For each encoder $e\in\{\text{vision},\text{text}\}$, the quantities $m^{(e)}_{\text{few},0},m^{(e)}_{\text{full},0}\in\mathbb{R}^{d_e}$ are the support and full-train gradients. The quantities $v^{(e)}_{\text{few},0},v^{(e)}_{\text{full},0}\in\mathbb{R}_{\ge 0}^{d_e}$ are the corresponding diagonal second moments. The Adam-like directions $u^{(e)}_{ff,0}$, $u^{(e)}_{fr,0}$, and $u^{(e)}_{rr,0}$ are defined as in Eq.~\eqref{eq:adam_diagnostic_directions}, and all belong to $\mathbb{R}^{d_e}$.

We use four scalar summaries. The first is the gradient alignment
\begin{equation}
    A_g^{(e)}(0) = \cos\!\left(m^{(e)}_{\text{few},0},m^{(e)}_{\text{full},0}\right) \in [-1,1].
\end{equation}
The second is the denominator scale ratio
\begin{equation}
    R^{(e)}_v(0)=\frac{\lVert v^{(e)}_{\text{few},0}\rVert_2}{\lVert v^{(e)}_{\text{full},0}\rVert_2}\in\mathbb{R}_{\ge 0}.
\end{equation}
Values $R^{(e)}_v(0)<1$ mean that the support set underestimates the full-data denominator in encoder $e$. The third is the coordinatewise mismatch $M^{(e)}_0$ from Eq.~\eqref{eq:adam_diagnostic_mismatch}. The fourth is the induced adaptive norm ratio
\begin{equation}
    R^{(e)}_{u,0}=\frac{\lVert u^{(e)}_{ff,0}\rVert_2}{\lVert u^{(e)}_{rr,0}\rVert_2}\in\mathbb{R}_{\ge 0}.
\end{equation}
This ratio compares the few-shot Adam-like direction with the full-data reference direction. Thus, $R^{(e)}_{u,0}>1$ means that Adam already proposes an oversized adaptive step at initialization.

\begin{figure*}
    \centering
    \includegraphics[width=0.98\linewidth]{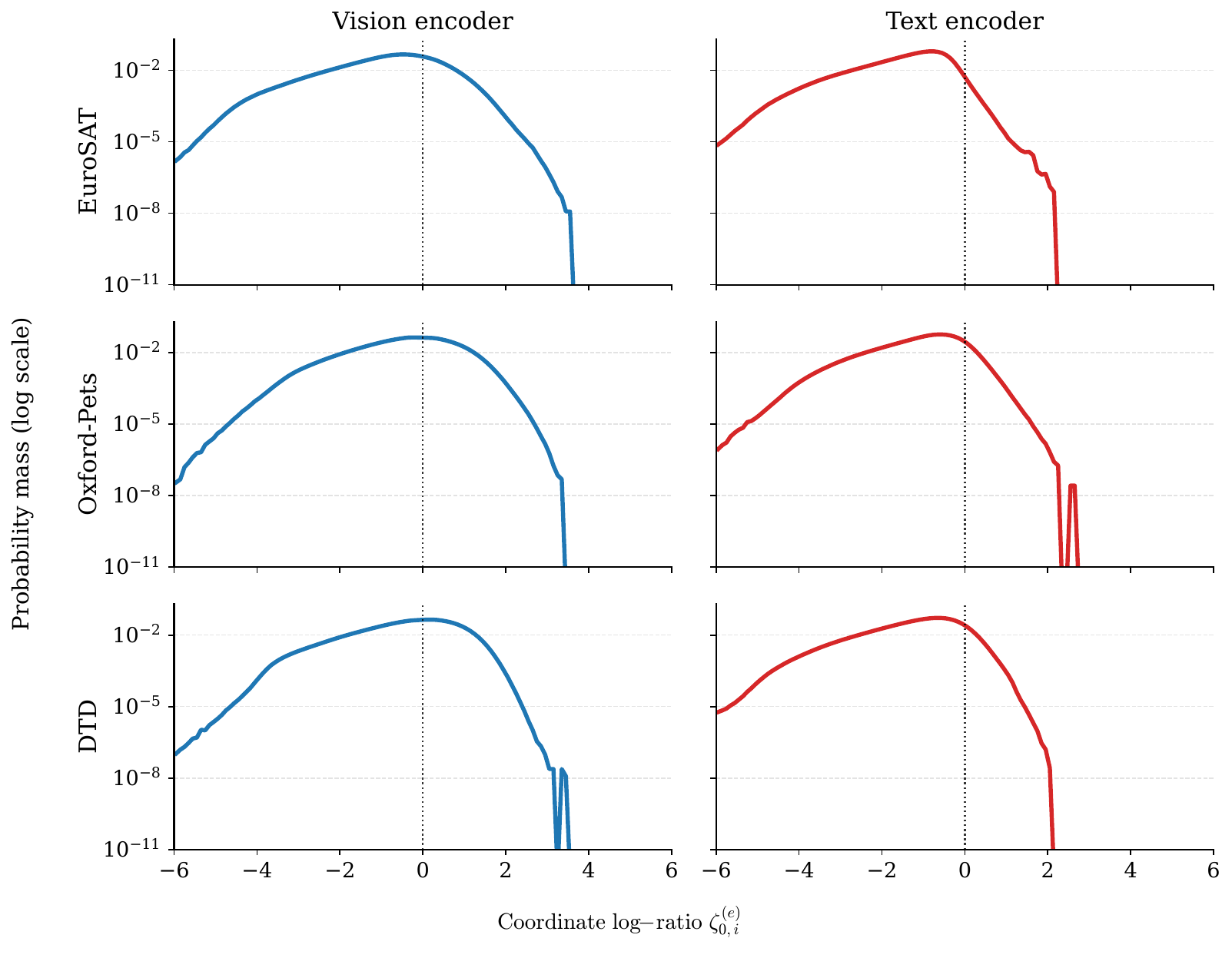}
    \caption{Experiment~2 : coordinate-level initialization audit at $t=0$. Each panel shows $\zeta^{(e)}_{0,i}=\log_{10}((v^{(e)}_{\mathrm{few},0,i}+\varepsilon)/(v^{(e)}_{\mathrm{full},0,i}+\varepsilon))$. The dotted line marks perfect calibration. Mass to the left of zero means that the support-set second moment is smaller than the full-train second moment. The underestimation is already present before any adaptation step, especially in the text encoder.}
    \label{fig:adam_fail_exp1_histograms}
\end{figure*}

Figure~\ref{fig:adam_fail_exp1_histograms} shows the coordinate log-ratio $\zeta^{(e)}_{0,i}$ for all datasets and encoders. Most coordinates lie to the left of zero before the first update. The bias is strongest in the text encoder. In the EuroSAT text, $99.1\%$ of coordinates satisfy $v^{(e)}_{\text{few},0,i}<v^{(e)}_{\text{full},0,i}$, $57.4\%$ are at least one order of magnitude too small, and $18.3\%$ are at least two orders of magnitude too small. The same pattern appears on Oxford-Pets text and DTD text, where the mass left of zero is $92.3\%$ and $92.9\%$, respectively. The vision encoder is less extreme, but it is not calibrated either. The mass left of zero is $77.7\%$ on EuroSAT, $60.6\%$ on Oxford-Pets, and $56.0\%$ on DTD.

\begin{figure*}
    \centering
    \includegraphics[width=0.98\linewidth]{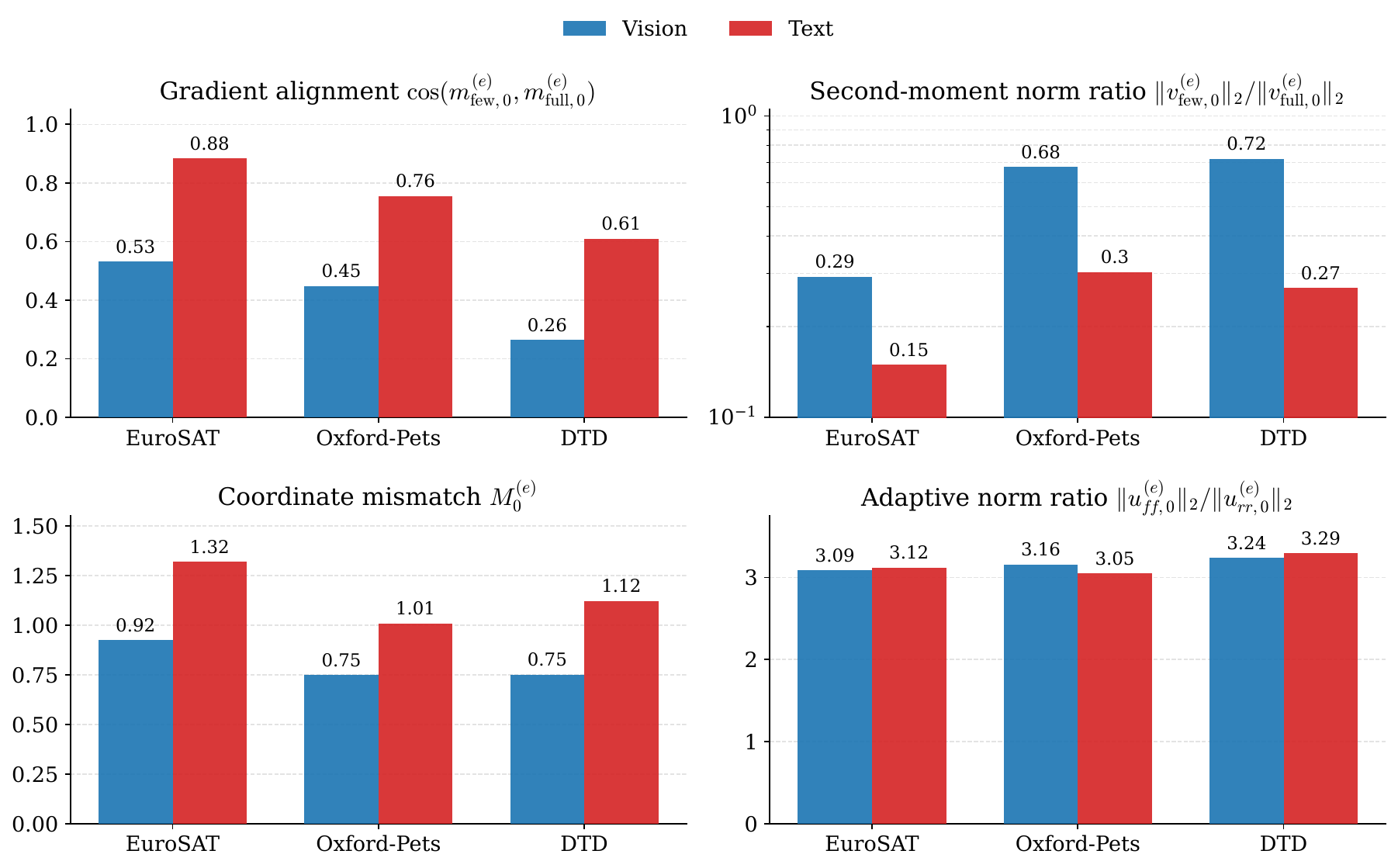}
    \caption{Experiment~2 : encoder-level initialization audit. The top-left panel reports the gradient alignment $A_g^{(e)}(0)$. The top-right panel reports the denominator scale ratio $R^{(e)}_v(0)$. The bottom-left panel reports $M^{(e)}_0$, the mean absolute coordinate log-ratio. The bottom-right panel reports the adaptive norm ratio $R^{(e)}_{u,0}$. The support-set second moment is too small in every dataset--encoder pair, and the resulting few-shot Adam-like direction is already about $3\times$ larger than the full-data reference direction.}
    \label{fig:adam_fail_exp1_encoder_summary}
\end{figure*}

Figure~\ref{fig:adam_fail_exp1_encoder_summary} summarizes the encoder-level effect. The second-moment norm ratio is below one in all six dataset--encoder pairs: $0.293$ on EuroSAT vision, $0.150$ on EuroSAT text, $0.677$ on Oxford-Pets vision, $0.303$ on Oxford-Pets text, $0.720$ on DTD vision, and $0.269$ on DTD text. The coordinate-wise mismatch is also large at initialization. The values of $M^{(e)}_0$ range from $0.749$ to $0.924$ in the vision encoder and from $1.008$ to $1.318$ in the text encoder. Thus, the typical coordinate-level error is about one order of magnitude in the text branch before Adam has taken any step.

The same figure shows that this denominator error immediately translates into update-scale error. The ratio $R^{(e)}_{u,0}$ is between $3.05$ and $3.29$ across all six cases. Thus, the few-shot Adam-like direction is already about three times larger than the full-data reference direction. This is the most direct evidence that the scale problem exists at initialization.

\begin{figure*}
    \centering
    \includegraphics[width=0.98\linewidth]{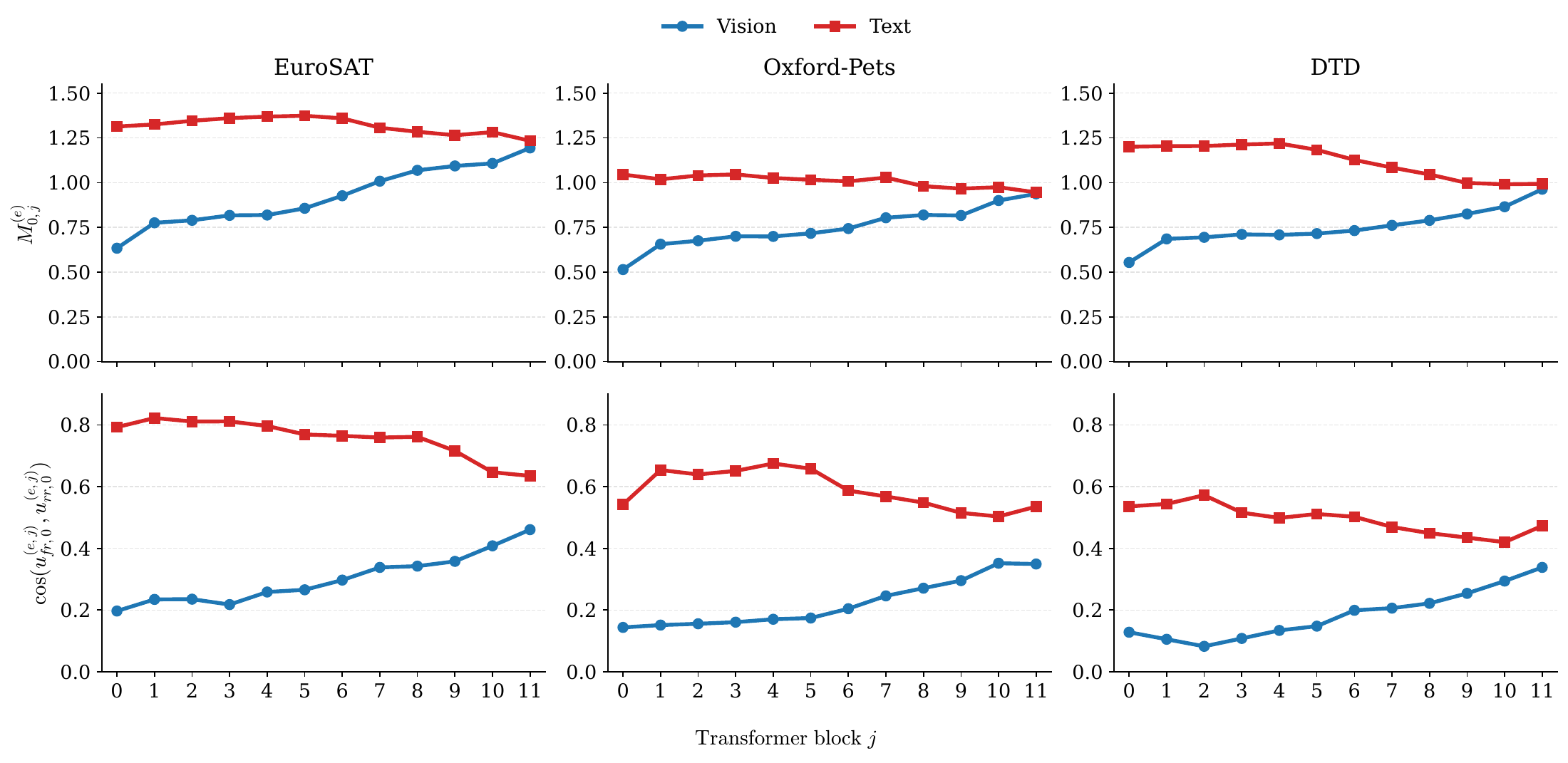}
    \caption{Experiment~2 : blockwise initialization audit. Top row: $M^{(e)}_{0,j}$, the mean absolute coordinate log-ratio in Transformer block $j$ of encoder $e$. Bottom row: $\cos(u^{(e,j)}_{fr,0},u^{(e,j)}_{rr,0})$, which measures how close the few-shot numerator becomes to the full-data reference direction when the denominator is replaced by the full-data denominator. The vision mismatch grows with depth, while the text mismatch is already large across nearly all blocks.}
    \label{fig:adam_fail_exp1_block_depth}
\end{figure*}

The text and vision encoders fail in different ways. The text branch is the cleaner denominator failure. Its gradients are already reasonably aligned with the full-data gradients: $A_g^{(e)}(0)=0.884$ on EuroSAT text, $0.755$ on Oxford-Pets text, and $0.609$ on DTD text. Yet its denominator is far too small. EuroSAT text is the clearest example. It has $A_g^{(e)}(0)=0.884$ but $R^{(e)}_v(0)=0.150$, $M^{(e)}_0=1.318$, and $R^{(e)}_{u,0}=3.121$. The numerator is therefore not the main problem in that case. The denominator already mis-scales an otherwise useful direction.

The vision branch is more mixed. The denominator is still miscalibrated, but the numerator is also less aligned. On DTD vision, $A_g^{(e)}(0)=0.264$ and $\lVert m^{(e)}_{\text{few},0}\rVert_2/\lVert m^{(e)}_{\text{full},0}\rVert_2=3.024$. Replacing only the denominator cannot fully recover the full-data reference direction. This is visible in the oracle-denominator cosine $\cos(u^{(e)}_{fr,0},u^{(e)}_{rr,0})$, which is only $0.163$ on DTD vision, $0.194$ on Oxford-Pets vision, and $0.276$ on EuroSAT vision. By contrast, the same cosine is $0.488$--$0.755$ in the text encoder. Thus, text is mostly a denominator-scale problem at $t=0$, while vision is both a denominator problem and a numerator-alignment problem.

Figure~\ref{fig:adam_fail_exp1_block_depth} shows that the initialization error is structured across depth. In the vision encoder, $M^{(e)}_{0,j}$ increases with block depth. On EuroSAT vision, it grows from $0.63$ at block $0$ to $1.19$ at block $11$. On DTD vision, it grows from $0.55$ to $0.96$. The text encoder behaves differently. Its denominator mismatch is high in all blocks and decreases mildly with depth. For example, EuroSAT text stays between $1.23$ and $1.38$ across blocks. The bottom row shows that the text-side oracle-denominator direction remains much closer to the full-data reference direction than the vision-side one. This supports a block-aware view of the failure. Adam's denominator is not merely wrong by one global scalar. The scale error depends on the encoder and depth.

Finally, the few-shot second moment is not random. At density $\rho_{\mathrm{top}}=0.01$, the top-$k$ overlap between $v^{(e)}_{\text{few},0}$ and $v^{(e)}_{\text{full},0}$ lies between $0.337$ and $0.542$ across the six dataset--encoder pairs. This is far above the random baseline $\rho_{\mathrm{top}}=0.01$. Therefore, the support statistic still identifies many coordinates with large second moments. Its main error is calibration. It assigns too little magnitude to a large fraction of the diagonal denominator.

\begin{findingbox}[Conclusion]
Adam does not start from a healthy denominator. At $\Theta_0$, the support-set second moment is already smaller than the full-train second moment in every dataset--encoder pair. This creates an Adam-like adaptive direction whose norm is already about three times larger. The text encoder is primarily a denominator-scale failure at initialization because its gradient remains well aligned with the full-data gradient. The vision encoder already has both denominator miscalibration and numerator mismatch. Therefore, training amplifies an existing scale problem. It does not create the pathology from scratch.
\end{findingbox}

\subsection{Does the gradient become small during training, or does the second moment collapse much faster?}
\label{subsec:grad_vs_second_moment_collapse}

\begin{figure*}
    \centering
    \includegraphics[width=0.98\linewidth]{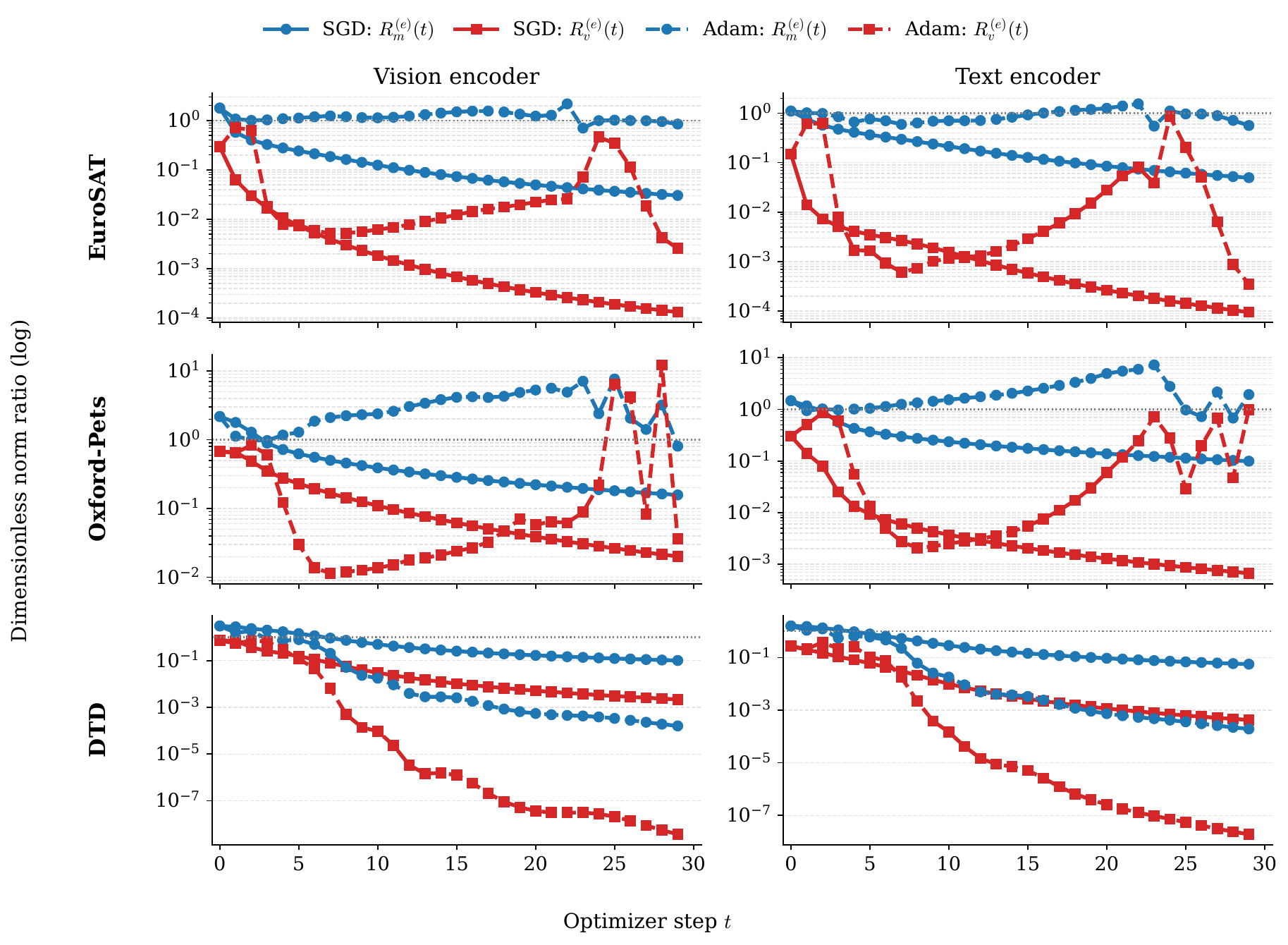}
    \caption{Experiment~3: gradient and second-moment norm ratios on SGD and Adam trajectories. Each panel reports $R^{(e)}_m(t)=\lVert m^{(e)}_{\mathrm{few},t}\rVert_2/\lVert m^{(e)}_{\mathrm{full},t}\rVert_2$ and $R^{(e)}_v(t)=\lVert v^{(e)}_{\mathrm{few},t}\rVert_2/\lVert v^{(e)}_{\mathrm{full},t}\rVert_2$. The second moment ratio collapses faster than the gradient ratio on the stable SGD trajectory. Adam can fail while the gradient ratio is still moderate, which rules out a purely vanishing gradient explanation.}
    \label{fig:adam_fail_exp2}
\end{figure*}

\begin{figure*}
    \centering
    \includegraphics[width=0.98\linewidth]{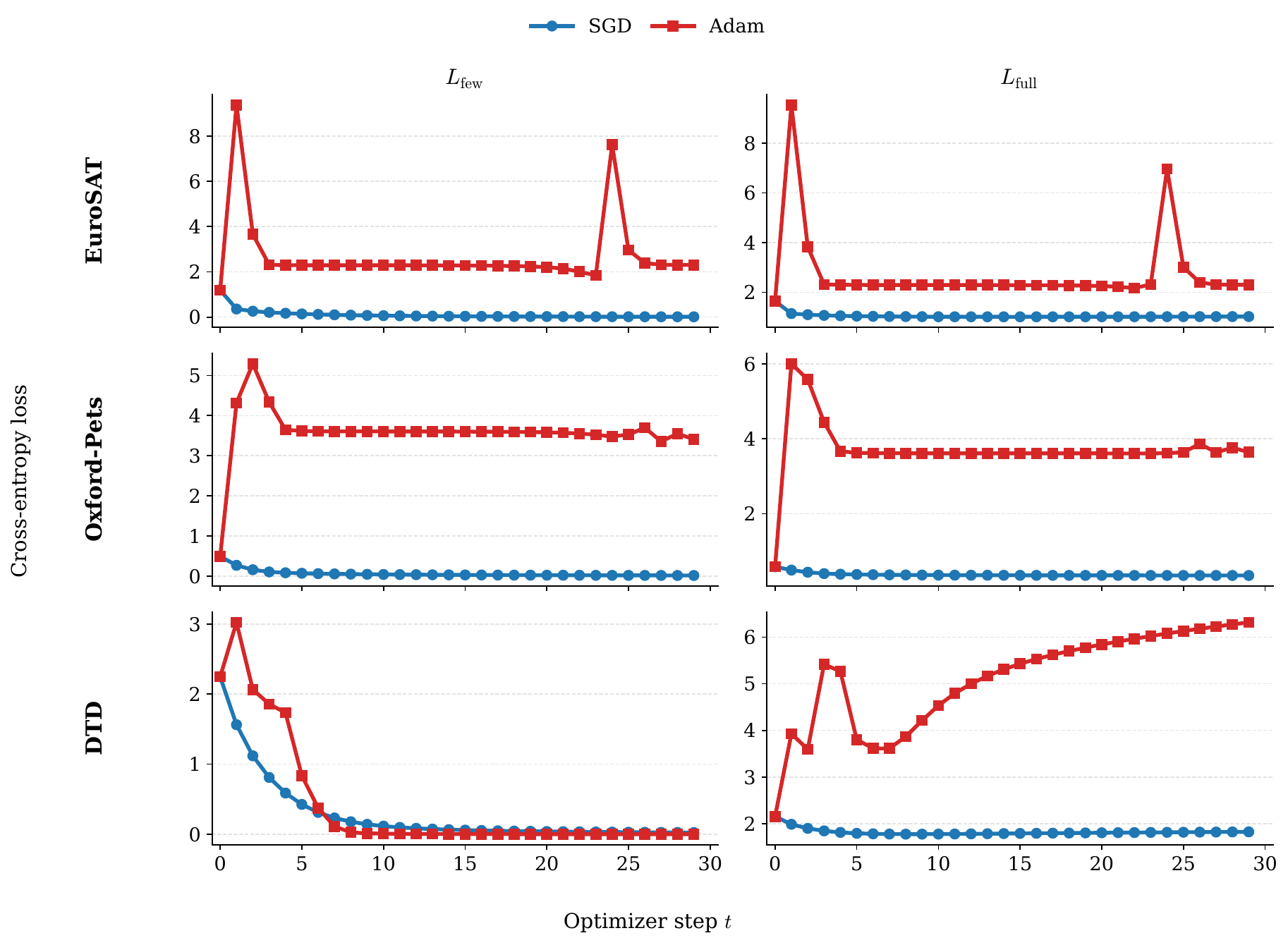}
    \caption{Experiment~3: support and full-train losses for the same SGD and Adam trajectories. SGD fits the support while keeping the full-train loss controlled. Adam fails in different ways across datasets. On EuroSAT and Oxford-Pets, it destabilizes the adaptation before the gradient ratio vanishes. On DTD, it memorizes the support set and destroys the full-train objective.}
    \label{fig:adam_fail_exp2_losses}
\end{figure*}

\begin{figure*}
    \centering
    \includegraphics[width=0.98\linewidth]{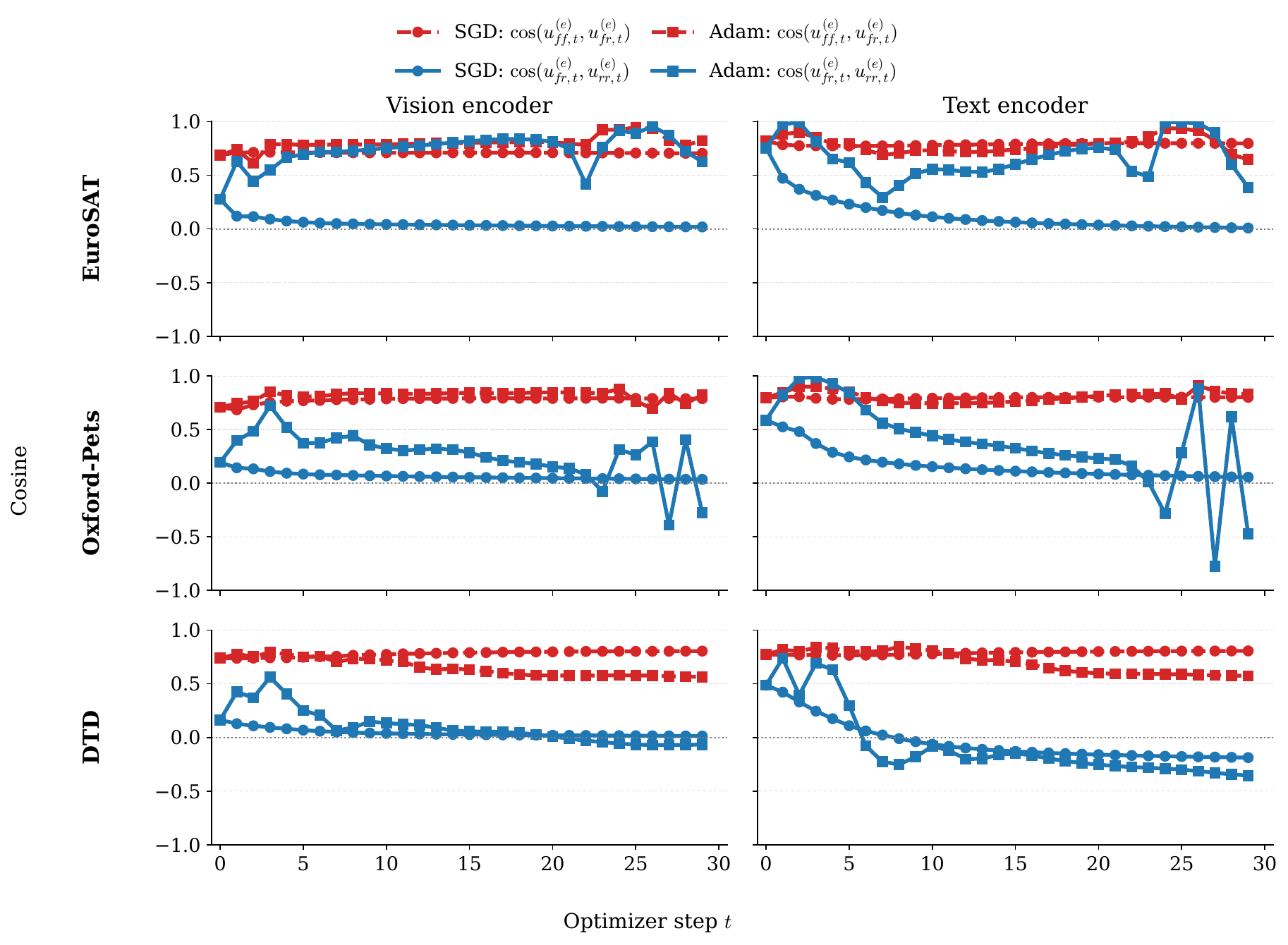}
    \caption{Experiment~3: adaptive-direction cosines. The red curves compare $u^{(e)}_{ff,t}$ with $u^{(e)}_{fr,t}$ and isolate the effect of changing only the denominator. The blue curves compare $u^{(e)}_{fr,t}$ with $u^{(e)}_{rr,t}$ and measure numerator error after the denominator has been replaced by the full-train reference. Denominator changes mostly preserve direction but change scale. Numerator drift becomes important later on the harder Adam trajectories.}
    \label{fig:adam_fail_exp2_precond}
\end{figure*}

Experiment~3 compares the relative decline of gradient and second-moment norms. The numerator is the few-shot gradient $m^{(e)}_{\text{few},t}\in\mathbb{R}^{d_e}$. The denominator statistic is the diagonal second moment $v^{(e)}_{\text{few},t}\in\mathbb{R}_{\ge 0}^{d_e}$. Both are compared with the corresponding full-train reference quantities $m^{(e)}_{\text{full},t}\in\mathbb{R}^{d_e}$ and $v^{(e)}_{\text{full},t}\in\mathbb{R}_{\ge 0}^{d_e}$. Here $e\in\{\text{vision},\text{text}\}$, and $d_e$ is the number of trainable scalar parameters in the encoder $e$. We evaluate these quantities on two trajectories. The first trajectory is generated by SGD. The second trajectory is generated by Adam under the same few-shot protocol.

At each checkpoint, the dimensionless gradient and second-moment norm ratios are
\begin{equation}
    R^{(e)}_m(t)
    = \frac{\lVert m^{(e)}_{\text{few},t}\rVert_2}
           {\lVert m^{(e)}_{\text{full},t}\rVert_2},
    \qquad
    R^{(e)}_v(t)
    = \frac{\lVert v^{(e)}_{\text{few},t}\rVert_2}
           {\lVert v^{(e)}_{\text{full},t}\rVert_2}.
    \label{eq:exp2_num_den_ratios}
\end{equation}
The ratio $R^{(e)}_m(t)$ measures whether the support gradient has the same scale as the full-train gradient. The ratio $R^{(e)}_v(t)$ measures whether the support second moment has the same scale as the full-train second moment. Adam divides by $\sqrt{v+\varepsilon}$ elementwise. The effect on update magnitude is measured separately using the preconditioned-direction norms.

Let $T$ denote the final logged checkpoint. To compare collapse rates, we also use
\begin{equation}
    C^{(e)}_m = \frac{R^{(e)}_m(0)}{R^{(e)}_m(T)},
    \qquad
    C^{(e)}_v = \frac{R^{(e)}_v(0)}{R^{(e)}_v(T)}.
    \label{eq:exp2_collapse_factors}
\end{equation}
The factors $C^{(e)}_m$ and $C^{(e)}_v$ are also dimensionless. If $C^{(e)}_v\gg C^{(e)}_m$, then the second-moment norm ratio declines by a larger relative factor than the gradient norm ratio. This comparison does not determine the norm of an elementwise preconditioned direction.

Figure~\ref{fig:adam_fail_exp2} gives the direct comparison. On the SGD trajectory, $R^{(e)}_v(t)$ decreases faster than $R^{(e)}_m(t)$ in all six dataset--encoder pairs. EuroSAT is the cleanest case. In the vision encoder, $R_m$ decreases from $1.79$ to $0.030$, while $R_v$ decreases from $0.293$ to $1.31\times 10^{-4}$. Hence $C_v/C_m\approx 38$. In the text encoder, $R_m$ decreases from $1.10$ to $0.050$, while $R_v$ decreases from $0.150$ to $9.53\times 10^{-5}$. Hence $C_v/C_m\approx 71$. The gradient norm ratio shrinks, but the second-moment norm ratio declines by a larger relative factor.

Oxford-Pets shows the same norm-ratio pattern with a milder vision branch. In vision, $C_v/C_m\approx 2.4$. In text, the second-moment norm ratio again declines by a larger relative factor, with $C_v/C_m\approx 31$. DTD is the strongest support-fitting case on the SGD trajectory. The vision branch gives $C_v/C_m\approx 11$, and the text branch gives $C_v/C_m\approx 22$. Thus, the second-moment norm ratio declines by a larger relative factor even when the support gradient becomes small.

The Adam trajectory rules out a simpler explanation. Figure~\ref{fig:adam_fail_exp2_losses} shows that Adam fails on EuroSAT and Oxford-Pets without first driving the support gradient to zero. On EuroSAT, Adam ends with both $L_{\text{few}}$ and $L_{\text{full}}$ near $2.30$. Yet its final numerator ratios are still moderate: $R_m=0.849$ in vision and $R_m=0.564$ in text. The corresponding second-moment ratios are much smaller: $R_v=2.57\times 10^{-3}$ in vision and $R_v=3.49\times 10^{-4}$ in text. Adam therefore fails while the gradient norm ratio remains moderate and the second-moment norm ratio is much smaller.

Oxford-Pets gives a related but more dynamic failure. Adam ends with $L_{\text{few}}\approx 3.41$ and $L_{\text{full}}\approx 3.64$. The final numerator ratios are not small. They are $0.805$ in vision and $1.93$ in text. Thus, the optimizer has not failed because the few-shot gradient norm vanished. Instead, Adam has moved to a region where the support and full objectives are no longer well represented by the same adaptive direction. DTD shows the late memorization limit. Adam drives $L_{\text{few}}$ to $7.6\times 10^{-5}$ while $L_{\text{full}}$ grows to $6.32$. In this case, both $R_m$ and $R_v$ eventually become tiny, but the second moment still collapses far faster. The final collapse factor ratio $C_v/C_m$ is about $1.0\times 10^4$ in vision and $1.7\times 10^3$ in text.

Figure~\ref{fig:adam_fail_exp2_precond} explains why denominator repair is easier early than late. The cosine $\cos(u^{(e)}_{ff,t},u^{(e)}_{fr,t})$ compares two adaptive directions with the same numerator and different denominators. This cosine remains high across most checkpoints. Therefore, replacing the few-shot denominator mainly changes the scale. The cosine $\cos(u^{(e)}_{fr,t},u^{(e)}_{rr,t})$ keeps the full denominator fixed and measures the remaining numerator error. On EuroSAT under Adam, this cosine stays positive for most of the training. Therefore, the failure is denominator-dominated. On Oxford-Pets and DTD, the same cosine decays and can become negative. This means that after Adam has moved the model, the numerator also becomes stale. The sequence is important: denominator collapse appears first, and trajectory-induced numerator drift appears later.

Experiment~3 sharpens the diagnosis from the previous subsections. The gradient can become small during successful support fitting. This happens on the stable SGD trajectory and in the late DTD Adam trajectory. However, this is not the primary failure mechanism of Adam. The second moment is already underestimated at initialization, and it collapses much faster during training. The collapse factors quantify this relative decline. The adaptive-norm diagnostics and update-budget interventions separately establish the update-scale mismatch.

\begin{findingbox}[Conclusion]
The second-moment norm ratio declines by a larger relative factor than the gradient norm ratio. On stable SGD trajectories, this holds for every dataset and encoder. Adam can fail on EuroSAT and Oxford-Pets while the few-shot gradient ratio remains moderate. On DTD, both ratios eventually collapse, with a much larger relative decline in the second-moment ratio. These factors describe norm shrinkage, not adaptive update growth. The separate adaptive-norm and update-budget diagnostics support a scale mismatch, followed on harder trajectories by numerator drift.
\end{findingbox}

\subsection{Is the problem specifically Adam's running second moment, or would even the instantaneous few-shot second moment already be harmful?}
\label{subsec:adam_ema_vs_instant}

Experiment~4 separates two possible explanations for the Adam denominator failure. The first is an \emph{EMA-state failure}: Adam's exponential moving average may be stale or badly biased even if the current few-shot second moment is useful. The second is a \emph{few-shot estimator failure}: the current few-shot second moment may already be harmful before temporal averaging is considered.

We evaluate this question along an Adam trajectory. The trainable state at iteration $t$ is still denoted by $\Theta_t\in\mathbb{R}^d$. For encoder $e\in\{\text{vision},\text{text}\}$, all denominator vectors below belong to $\mathbb{R}_{\ge0}^{d_e}$. We compare three sources:
\begin{align}
    v^{(e)}_{\mathrm{inst},t}
    &= m^{(e)}_{\mathrm{few},t}\odot m^{(e)}_{\mathrm{few},t},
    \\
    v^{(e)}_{\mathrm{batch},t}
    &= \mathbb{E}_{\mathcal{B}\subset\mathcal{D}_{\mathrm{few}}}
       \left[g^{(e)}_{\mathcal{B},t}\odot g^{(e)}_{\mathcal{B},t}\right],
    \\
    v^{(e)}_{\mathrm{ema},t}
    &= \frac{v_t^{(e)}}{1-\beta_2^t}, \qquad t\ge 1 .
\end{align}
Here $m^{(e)}_{\mathrm{few},t}\in\mathbb{R}^{d_e}$ is the support gradient from Eq.~\eqref{eq:adam_diagnostic_moments}, and $g^{(e)}_{\mathcal{B},t}\in\mathbb{R}^{d_e}$ is a support mini-batch gradient. The vector $v_t^{(e)}\in\mathbb{R}_{\ge0}^{d_e}$ is the restriction to encoder $e$ of the uncorrected Adam state $v_t$ defined in Algorithm~\ref{alg:adam}. The vector $v^{(e)}_{\mathrm{inst},t}$ is the squared full-support gradient. The vector $v^{(e)}_{\mathrm{batch},t}$ is the diagnostic support-batch second moment. The vector $v^{(e)}_{\mathrm{ema},t}$ is Adam's bias-corrected running second moment. At $t=0$, the EMA state is empty, so the EMA curve is omitted from the EMA-specific interpretation at initialization. The full-data reference remains $v^{(e)}_{\mathrm{full},t}$.

For the instantaneous second moment, the adaptive direction is
\begin{equation}
    u^{(e)}_{\mathrm{inst},t}
    = \frac{m^{(e)}_{\mathrm{few},t}}
           {\sqrt{v^{(e)}_{\mathrm{inst},t}+\varepsilon}}
    \in\mathbb{R}^{d_e} .
\end{equation}
We compare it with the reference direction $u^{(e)}_{rr,t}$ from Eq.~\eqref{eq:adam_diagnostic_directions}. We also use the previously defined oracle-denominator direction $u^{(e)}_{fr,t}$.
This direction fixes the denominator while keeping the few-shot numerator. It is useful because it separates denominator error from numerator drift.

The instantaneous source has four scalar diagnostics:
\begin{align}
    R^{(e)}_{v,\mathrm{inst}}(t)
    &= \frac{\lVert v^{(e)}_{\mathrm{inst},t}\rVert_2}
            {\lVert v^{(e)}_{\mathrm{full},t}\rVert_2},
    \\
    M^{(e)}_{\mathrm{inst}}(t)
    &= \frac{1}{d_e}\sum_{i=1}^{d_e}
       \left|\log_{10}\!\left(
       \frac{v^{(e)}_{\mathrm{inst},t,i}+\varepsilon}
            {v^{(e)}_{\mathrm{full},t,i}+\varepsilon}
       \right)\right|,
    \\
    R^{(e)}_{u,\mathrm{inst}}(t)
    &= \frac{\lVert u^{(e)}_{\mathrm{inst},t}\rVert_2}
            {\lVert u^{(e)}_{rr,t}\rVert_2},
    \\
    A^{(e)}_{u,\mathrm{inst}}(t)
    &= \cos\!\left(u^{(e)}_{\mathrm{inst},t},u^{(e)}_{rr,t}\right).
\end{align}
The batch and EMA quantities use the same definitions with their respective second moments and the literal subscripts $\mathrm{batch}$ and $\mathrm{ema}$.
The first two quantities measure denominator scale and coordinatewise denominator mismatch. The last two quantities measure the scale and alignment of the adaptive update induced by that denominator.

\begin{figure*}[t]
    \centering
    \includegraphics[width=0.98\linewidth]{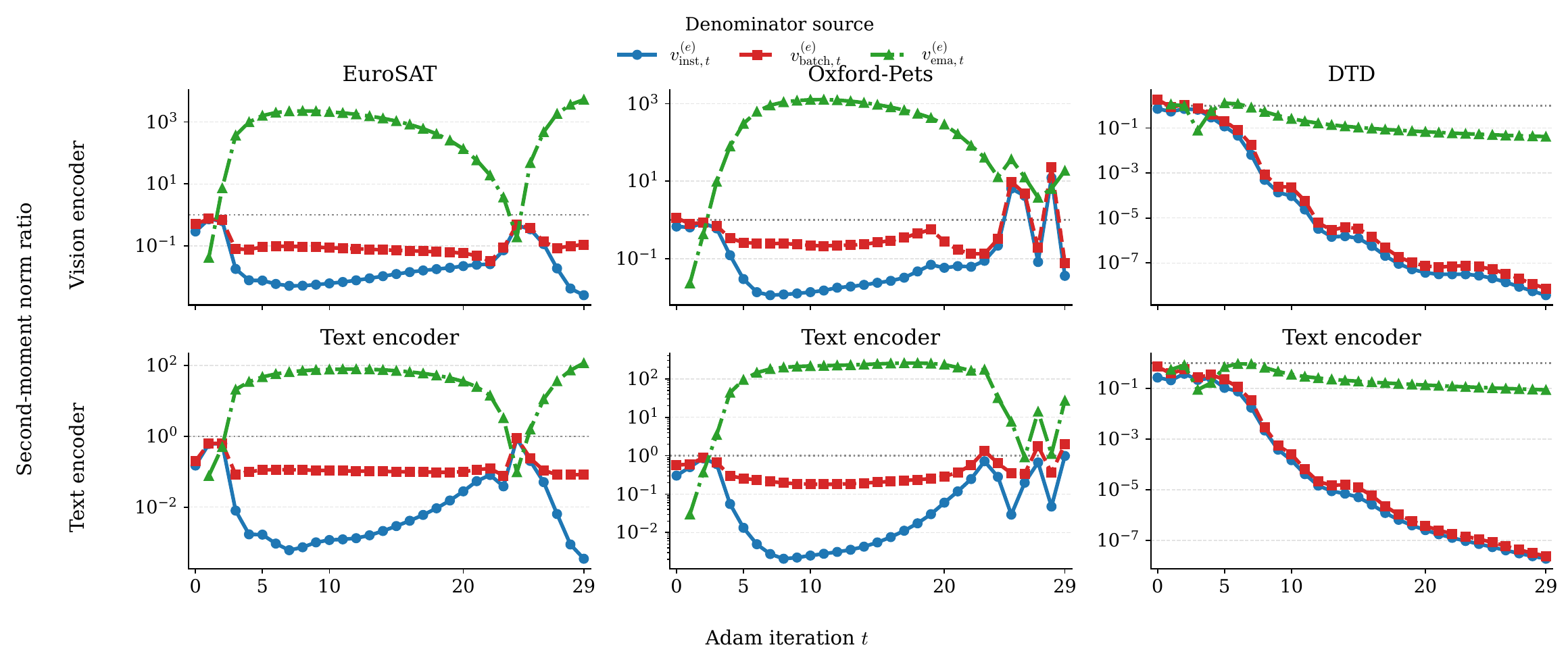}
    \caption{Experiment~4: denominator norm ratios for the instantaneous, support-batch, and EMA denominators on an Adam trajectory with $\beta_2=0.999$. The dotted horizontal line marks perfect scale calibration. The EMA curve starts at $t=1$ because no running second-moment state exists at $t=0$.}
    \label{fig:adam_fail_exp3_vnormratio}
\end{figure*}

\begin{figure*}[t]
    \centering
    \includegraphics[width=0.98\linewidth]{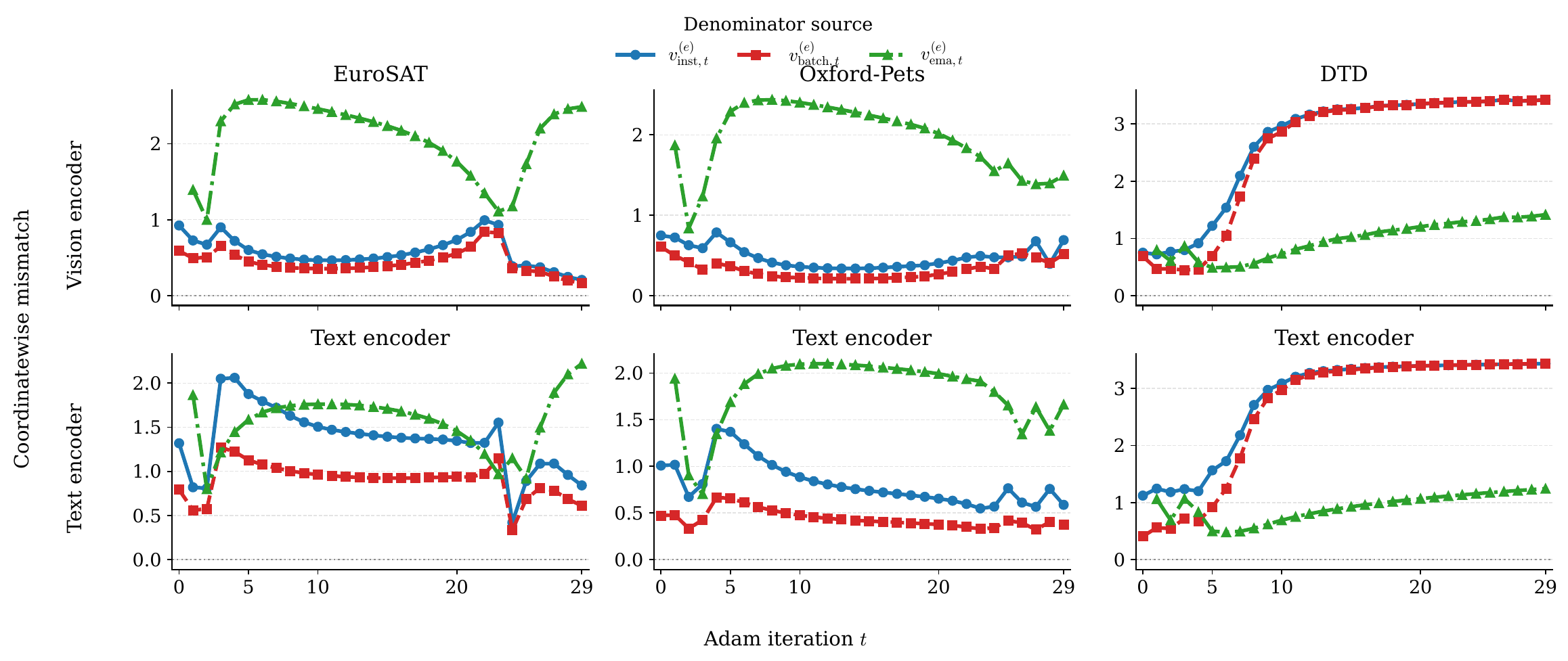}
    \caption{Experiment~4: coordinatewise denominator mismatch for the instantaneous, support-batch, and EMA sources. The support-batch estimator is usually closer to the full-data denominator than the instantaneous squared support gradient. The EMA state often increases the mismatch on EuroSAT and Oxford-Pets. On DTD, the late EMA denominator can have a smaller aggregate mismatch while still inducing a poor adaptive direction.}
    \label{fig:adam_fail_exp3_vmismatch}
\end{figure*}

\begin{figure*}[t]
    \centering
    \includegraphics[width=0.98\linewidth]{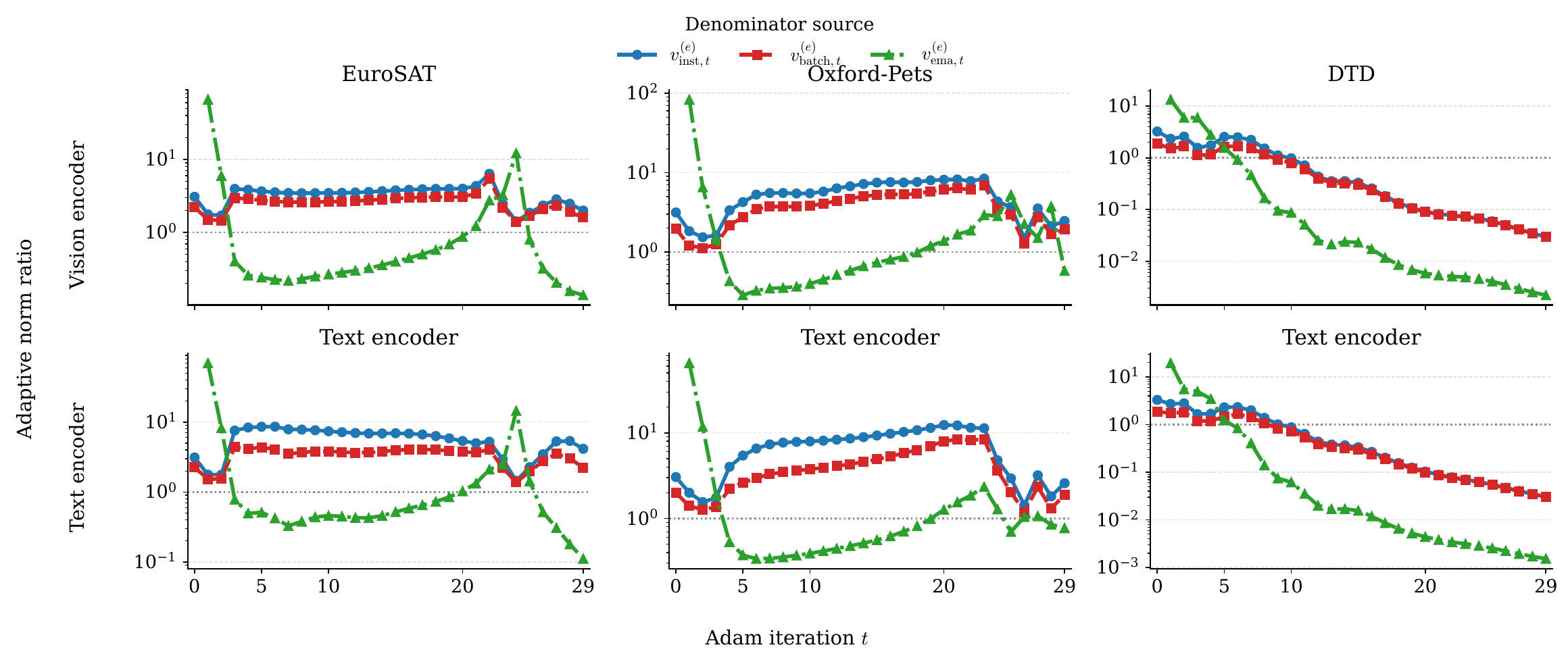}
    \caption{Experiment~4: adaptive norm ratios for the instantaneous, support-batch, and EMA sources. Values above one indicate an oversized adaptive direction relative to the full-data reference. Values below one indicate an overly damped direction.}
    \label{fig:adam_fail_exp3_precond_norm}
\end{figure*}

\begin{figure*}[t]
    \centering
    \includegraphics[width=0.98\linewidth]{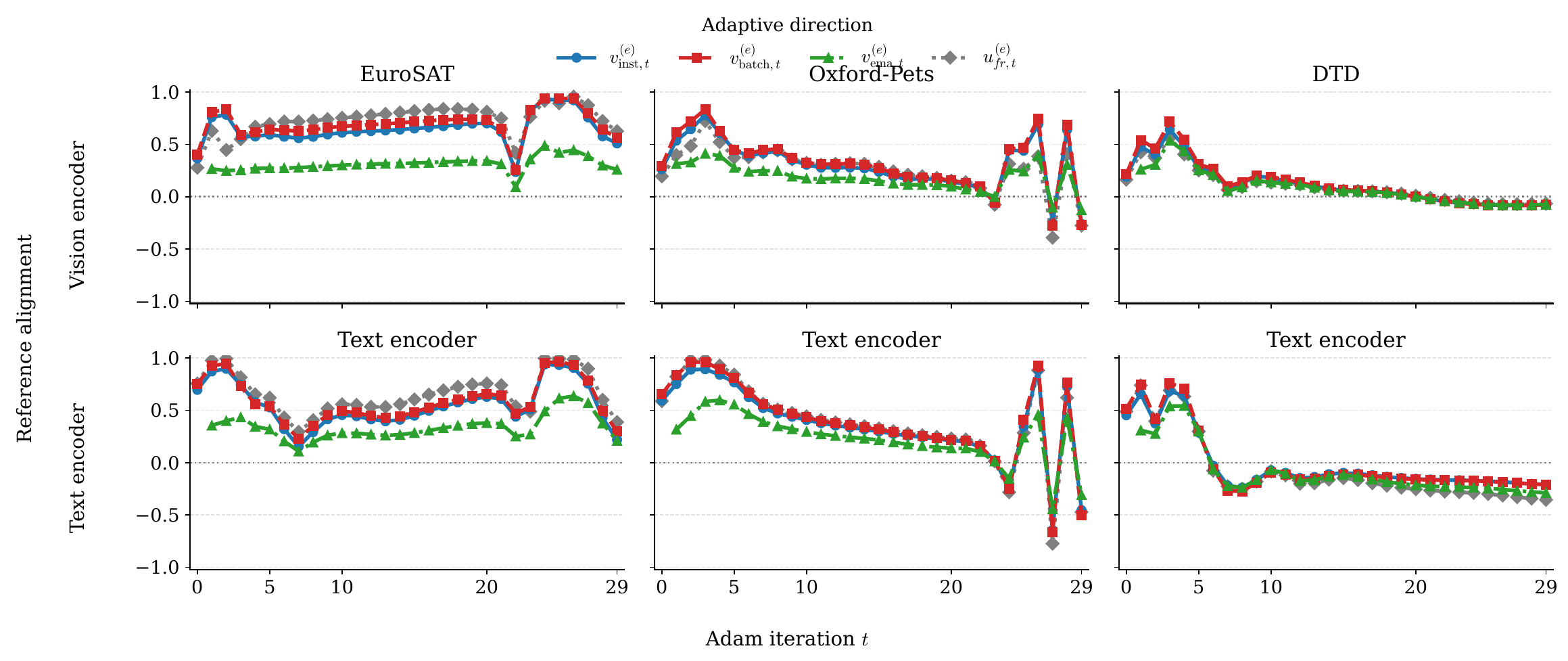}
    \caption{Experiment~4: adaptive-direction alignment with $u^{(e)}_{rr,t}$ for the instantaneous, support-batch, and EMA sources. The gray curve is the oracle-denominator direction $u^{(e)}_{fr,t}$, which keeps the few-shot numerator and replaces only the denominator by the full-data reference. A low oracle curve indicates numerator drift that denominator repair alone cannot fix.}
    \label{fig:adam_fail_exp3_precond_alignment}
\end{figure*}

\begin{figure*}[t]
    \centering
    \includegraphics[width=0.98\linewidth]{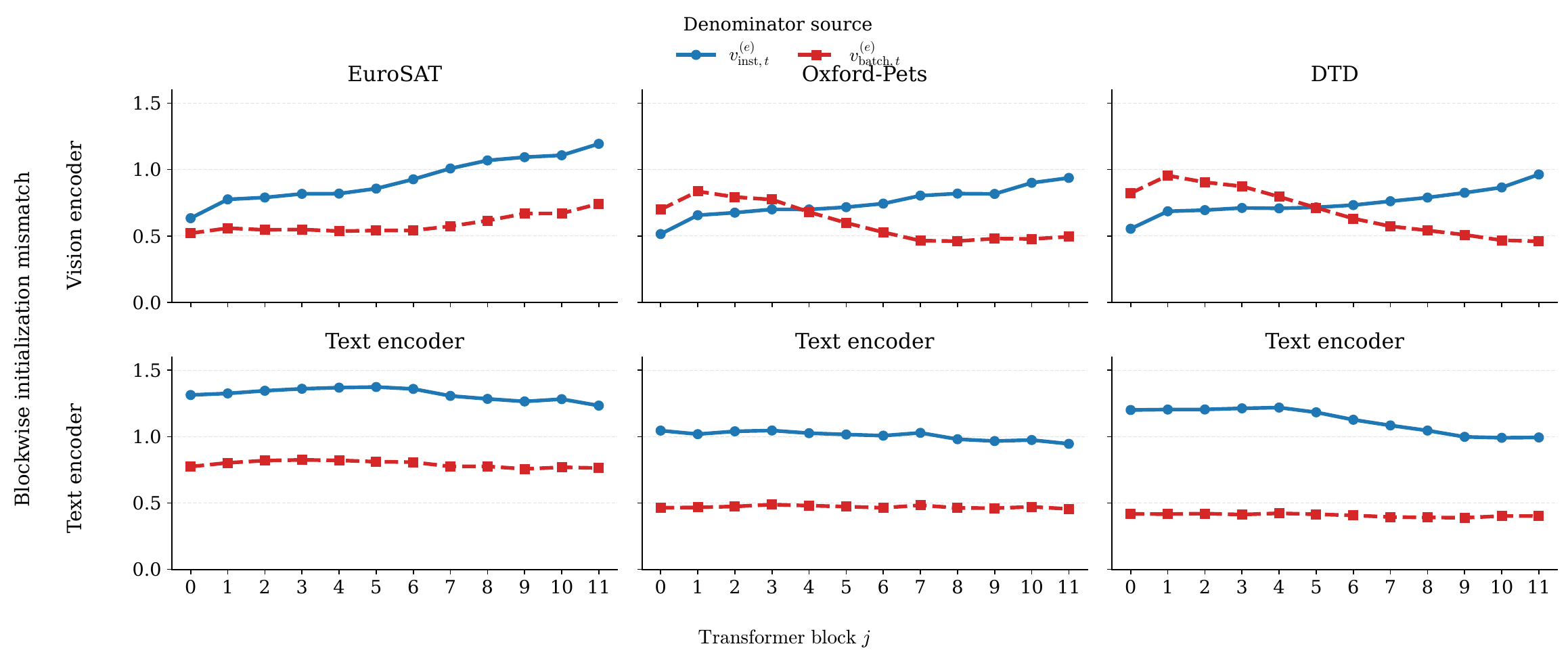}
    \caption{Experiment~4: blockwise initialization mismatches $M^{(e,j)}_{\mathrm{inst}}(0)$ and $M^{(e,j)}_{\mathrm{batch}}(0)$ for the instantaneous and support-batch denominators. The EMA state is not shown because it is empty at $t=0$. The mismatch is structured by encoder and depth. Vision mismatch generally increases with depth, whereas text mismatch is large across nearly all blocks.}
    \label{fig:adam_fail_exp3_block_mismatch}
\end{figure*}

Figures~\ref{fig:adam_fail_exp3_vnormratio} and~\ref{fig:adam_fail_exp3_vmismatch} show that the instantaneous denominator is already a bad estimator. At $t=0$, $R^{(e)}_{v,\mathrm{inst}}(0)$ is below one in all six dataset--encoder pairs. It equals $0.293$ and $0.150$ on EuroSAT vision and text, $0.677$ and $0.303$ on Oxford-Pets, and $0.720$ and $0.269$ on DTD. The corresponding mismatch values lie between $0.749$ and $1.318$ base-$10$ log units. Thus, before any EMA state can become stale, the instantaneous few-shot second moment already underestimates the full-data denominator by a large multiplicative factor.

The same initialization already produces harmful adaptive scales. Figure~\ref{fig:adam_fail_exp3_precond_norm} shows that $R^{(e)}_{u,\mathrm{inst}}(0)$ lies between $3.05$ and $3.29$ across the six cases. Thus, using $v_{\mathrm{inst}}$ would make the few-shot adaptive direction about three times larger than the full-data reference. This is the direct answer to the subsection question. Adam's running state is not required to create the scale pathology. The instantaneous squared support gradient is already harmful.

Averaging over support batches helps, but it does not solve the problem. In Figure~\ref{fig:adam_fail_exp3_vmismatch}, $v_{\mathrm{batch}}$ is usually below $v_{\mathrm{inst}}$ in coordinatewise mismatch. At initialization, the text-encoder mismatch decreases from $1.318$ to $0.792$ on EuroSAT, from $1.008$ to $0.470$ on Oxford-Pets, and from $1.119$ to $0.407$ on DTD. The adaptive norm inflation also drops from about $3\times$ to roughly $1.9$--$2.3\times$. The batch estimator therefore recovers useful second-moment information from the support set. However, it remains miscalibrated relative to $v_{\mathrm{full}}$.

The running EMA often amplifies this few-shot estimation error. Figure~\ref{fig:adam_fail_exp3_vnormratio} shows that $v_{\mathrm{ema}}$ becomes orders of magnitude larger than $v_{\mathrm{full}}$ through much of the EuroSAT and Oxford-Pets trajectories. Figure~\ref{fig:adam_fail_exp3_vmismatch} shows the same effect at the coordinate level. This behavior is not a harmless scalar rescaling. In Figure~\ref{fig:adam_fail_exp3_precond_alignment}, the EMA-induced direction usually has lower alignment with $u_{rr}$ than the instantaneous or batch directions. The support-batch denominator is the closest of the three few-shot denominators to the oracle direction in most early and middle checkpoints.

DTD shows why the aggregate denominator scale is not sufficient. In late DTD training, the instantaneous and batch denominators collapse almost to zero because Adam has nearly memorized the support set. The EMA denominator remains larger and has a smaller value of $M^{(e)}_{\mathrm{ema}}(t)$. Nevertheless, Figures~\ref{fig:adam_fail_exp3_precond_norm} and~\ref{fig:adam_fail_exp3_precond_alignment} show that the induced adaptive direction remains poor. The oracle curve in Figure~\ref{fig:adam_fail_exp3_precond_alignment} also drops below zero on the late DTD text trajectory. This means that the few-shot numerator has drifted as well. Once this happens, denominator repair alone cannot reconstruct the full-data reference direction.

Figure~\ref{fig:adam_fail_exp3_block_mismatch} shows that the initialization error is structured across blocks. In the vision encoder, the instantaneous mismatch increases with depth on all three datasets. In the text encoder, the instantaneous mismatch is already high across nearly every block. The support-batch estimator reduces this mismatch in most blocks, especially in text. The blockwise pattern is important because it shows that the few-shot second-moment problem is not a single scalar error. It is an encoder- and depth-dependent calibration error.

Changing $\beta_2$ reduces some EMA effects but does not change the source ordering. In the additional $\beta_2\in\{0.99,0.9\}$ runs, a shorter EMA half-life makes the running denominator less stale. It does not make the instantaneous or support-batch denominators fully calibrated. It also does not remove the late numerator drift on DTD. Therefore, EMA staleness is real, but it is not the root cause.

\begin{findingbox}[Conclusion]
The problem is not specific to Adam's running second moment. The instantaneous few-shot second moment is already miscalibrated at initialization and already induces an adaptive direction whose norm is about three times larger. The support-batch second moment is better, but it remains miscalibrated. Adam's EMA is best understood as an amplifier of this few-shot denominator error. On harder trajectories such as DTD, the failure later becomes denominator error plus numerator drift, so denominator repair alone is no longer sufficient.
\end{findingbox}

\subsection{Recovery under denominator-only interventions}

Experiment~5 tests whether Adam can be rescued by changing only its denominator. The test is performed on the Adam trajectory. Thus, the checkpoint $\Theta_t\in\mathbb{R}^d$ has already been produced by Adam itself. Let $m_{\mathrm{few},t},m_{\mathrm{full},t}\in\mathbb{R}^d$ and $v_{\mathrm{few},t},v_{\mathrm{full},t}\in\mathbb{R}_{\ge0}^{d}$ denote the global concatenations of the encoder-wise first- and second-moment statistics from Eq.~\eqref{eq:adam_diagnostic_moments}. The directions $u_{ff,t}$, $u_{fr,t}$, and $u_{rr,t}$ are defined in Eq.~\eqref{eq:adam_diagnostic_directions}. The direction $u_{ff,t}$ is the live few-shot Adam-like direction. The direction $u_{fr,t}$ keeps the same few-shot numerator but replaces the denominator with the full-train statistic. The direction $u_{rr,t}$ is the full-train reference direction. It is not a few-shot update. It is a full-data reference diagnostic.

We add two denominator-only counterfactuals. The first freezes the denominator at initialization. The second uses a short-lag denominator. For lag $h=5$, these directions are
\begin{equation}
    u_{f0,t}
    = \frac{m_{\mathrm{few},t}}
           {\sqrt{v_{\mathrm{few},0}+\varepsilon}}
    \in\mathbb{R}^d,
    \qquad
    u_{f\mathrm{lag},t}
    = \frac{m_{\mathrm{few},t}}
           {\sqrt{v_{\mathrm{few},t-h}+\varepsilon}}
    \in\mathbb{R}^d,
    \quad t\ge h .
    \label{eq:exp4_frozen_lagged}
\end{equation}
The square root and division are elementwise. The frozen direction $u_{f0,t}$ tests whether a protected initial diagonal scale is useful. The lagged direction $u_{f\mathrm{lag},t}$ tests whether a short-memory denominator is preferable to the live denominator. When $t<h$, the lagged diagnostic is omitted.

\begin{figure*}[t]
    \centering
    \includegraphics[width=0.98\linewidth]{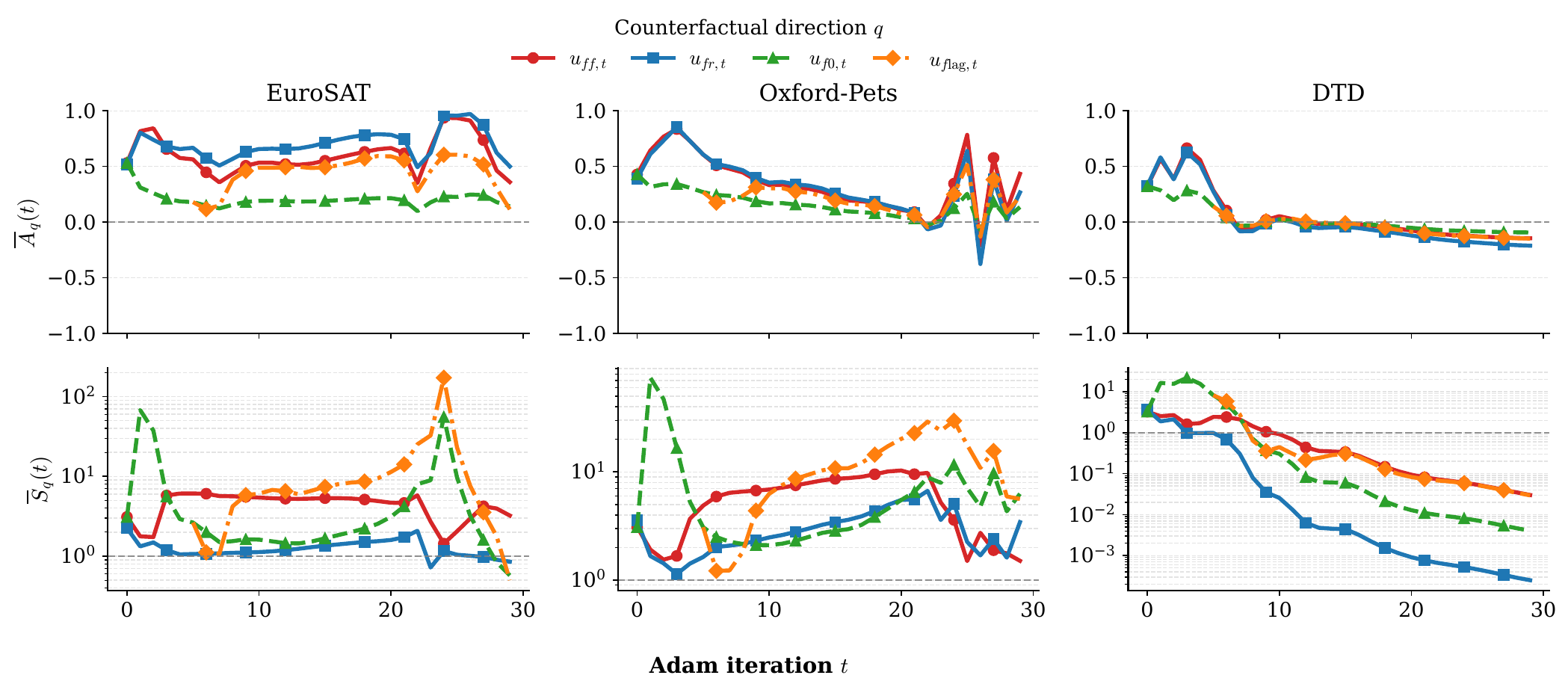}
    \caption{Experiment~5: encoder-averaged counterfactual geometry on the Adam trajectory. The top row shows $\overline A_q(t)$, the average of $A^{(e)}_q(t)=\cos(u^{(e)}_{q,t},u^{(e)}_{rr,t})$ over the vision and text encoders. The bottom row shows $\overline S_q(t)$, the corresponding average of $S^{(e)}_q(t)=\lVert u^{(e)}_{q,t}\rVert_2/\lVert u^{(e)}_{rr,t}\rVert_2$. The gray dashed line marks zero alignment in the top row and unit scale in the bottom row. The oracle denominator improves both alignment and scale on EuroSAT, improves mainly scale on Oxford-Pets, and does not rescue alignment on DTD.}
    \label{fig:adam_fail_exp4_geometry}
\end{figure*}

For encoder $e\in\{\mathrm{vision},\mathrm{text}\}$ and direction $q\in\{ff,fr,f0,f\mathrm{lag}\}$, let $u^{(e)}_{q,t}\in\mathbb{R}^{d_e}$ be the encoder-restricted vector. We measure alignment and scale relative to the full-train reference by
\begin{equation}
    A^{(e)}_q(t)
    = \cos\!\left(u^{(e)}_{q,t},u^{(e)}_{rr,t}\right),
    \qquad
    S^{(e)}_q(t)
    = \frac{\lVert u^{(e)}_{q,t}\rVert_2}
           {\lVert u^{(e)}_{rr,t}\rVert_2} .
    \label{eq:exp4_alignment_scale}
\end{equation}
The cosine $A^{(e)}_q(t)$ measures whether the counterfactual direction points toward the full-train reference. The ratio $S^{(e)}_q(t)$ is dimensionless. Values of $S^{(e)}_q(t)$ close to one indicate a matched update scale. Figure~\ref{fig:adam_fail_exp4_geometry} plots the encoder averages $\overline A_q(t)$ and $\overline S_q(t)$ over vision and text.

\begin{figure*}[t]
    \centering
    \includegraphics[width=0.98\linewidth]{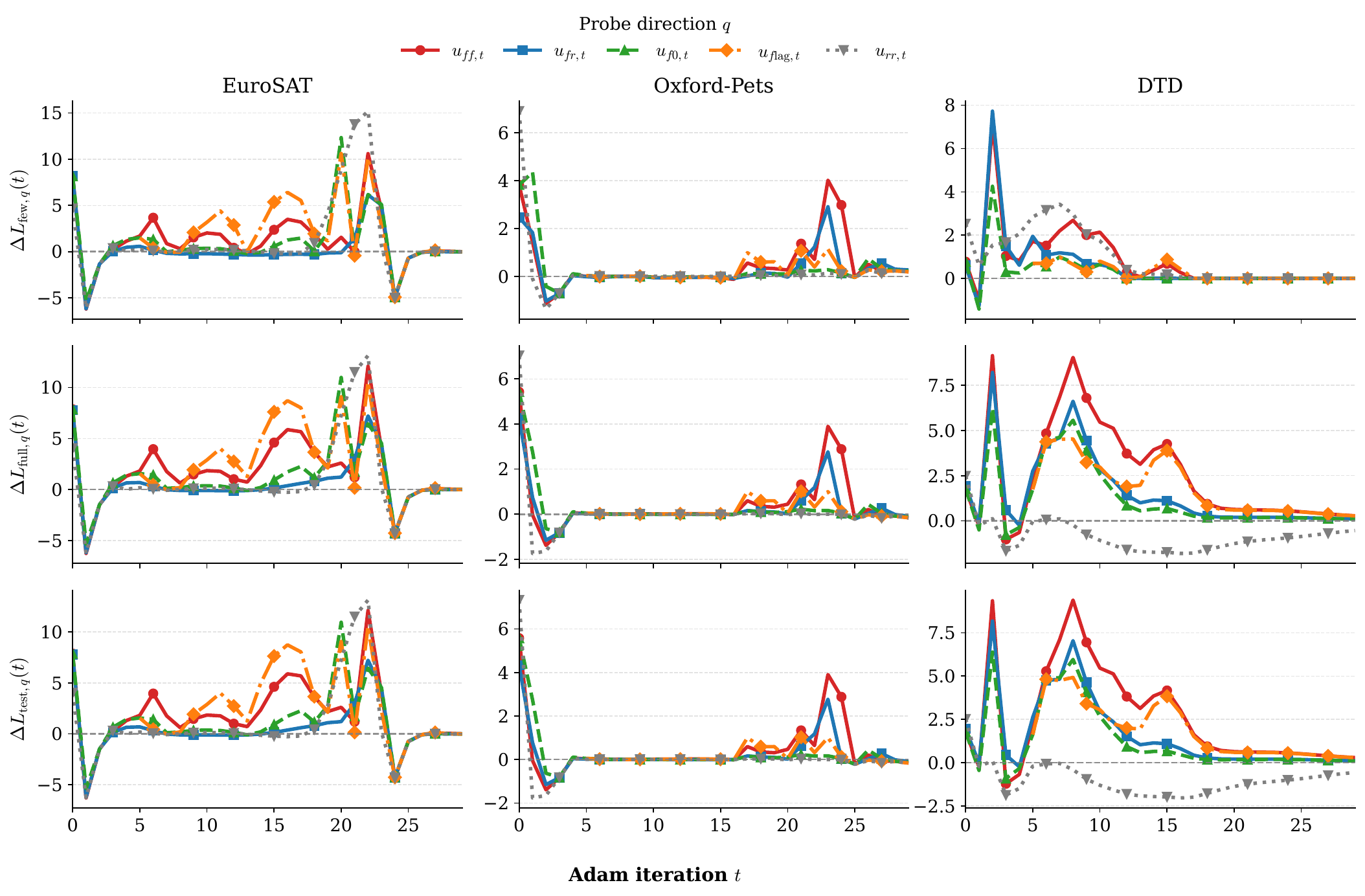}
    \caption{Experiment~5: norm-matched one-step probe losses on the Adam trajectory. Each direction $q$ is rescaled to satisfy $\lVert\widetilde u_{q,t}\rVert_2=\lVert u_{ff,t}\rVert_2$ before applying the temporary update in Eq.~\eqref{eq:exp4_probe_update}. Rows report $\Delta L_{\mathrm{few},q}(t)$, $\Delta L_{\mathrm{full},q}(t)$, and $\Delta L_{\mathrm{test},q}(t)$. Negative values are beneficial. The full-train and test rows are the decisive diagnostics for rescue.}
    \label{fig:adam_fail_exp4_probe}
\end{figure*}

The strongest diagnostic is a norm-matched one-step probe. For $q\in\{ff,fr,f0,f\mathrm{lag},rr\}$, the rescaled direction is
\begin{equation}
    \widetilde u_{q,t}
    = \frac{\lVert u_{ff,t}\rVert_2}
           {\lVert u_{q,t}\rVert_2}u_{q,t}
    \in\mathbb{R}^d .
    \label{eq:exp4_norm_matched_direction}
\end{equation}
Thus, all probes have the same global norm as the live Adam-like direction $u_{ff,t}$. With probe multiplier $\alpha_{\mathrm{p}}=1$, we apply the temporary update
\begin{equation}
    \Theta^{[q]}_t
    = \Theta_t - \eta\alpha_{\mathrm{p}}\widetilde u_{q,t} .
    \label{eq:exp4_probe_update}
\end{equation}
For $S\in\{\mathrm{few},\mathrm{full},\mathrm{test}\}$, the probe loss change is
\begin{equation}
    \Delta L_{S,q}(t)
    = L_{S}(\Theta^{[q]}_t)-L_{S}(\Theta_t) .
    \label{eq:exp4_probe_delta}
\end{equation}
Negative values are beneficial. Positive values are damaging. Because the probe norms are matched, differences in $\Delta L_{S,q}(t)$ are caused by the choice of the denominator and by any remaining numerator error, not by a different global update budget. Figure~\ref{fig:adam_fail_exp4_probe} reports the support, full-train, and test probes.

Unless stated otherwise, the probe means below use all 30 checkpoints $t=0,\ldots,29$. Comparisons involving the lagged denominator use the common 25 checkpoints $t=5,\ldots,29$.

On EuroSAT, denominator-only repair is strong. Figure~\ref{fig:adam_fail_exp4_geometry} shows that the oracle denominator improves the Adam direction in both encoders. Averaged over all iterations, $\cos(u^{(\mathrm{vision})}_{ff,t},u^{(\mathrm{vision})}_{rr,t})$ increases from $0.649$ to $0.734$ when $u_{ff,t}$ is replaced by $u_{fr,t}$. In the text encoder, it increases from $0.551$ to $0.663$. The scale correction is even clearer. The vision norm ratio drops from $3.307$ to $1.501$, and the text norm ratio drops from $5.740$ to $1.039$. Thus, the oracle denominator almost fixes the text-side update scale and substantially fixes the vision-side scale.

The EuroSAT probe losses in Figure~\ref{fig:adam_fail_exp4_probe} give the causal effect. Averaged over the 30 probes, the live direction $u_{ff,t}$ increases the test loss by $1.862$. The oracle-denominator direction $u_{fr,t}$ increases it by only $0.499$. This is a $73.2\%$ reduction in test-loss damage at the same norm. The full-train loss gives the same number, up to rounding: $1.861$ for $u_{ff,t}$ and $0.499$ for $u_{fr,t}$. The frozen direction $u_{f0,t}$ also helps, but less strongly. It reduces the mean test damage to $1.171$, a $37.1\%$ reduction. Over the common checkpoints $t=5,\ldots,29$, the lagged direction is worse than the live direction on EuroSAT, with mean test damage $2.687$ versus $2.168$. Therefore, EuroSAT is the cleanest denominator-dominated case. The few-shot numerator remains useful enough that replacing the denominator rescues most of the one-step failure.

Oxford-Pets is a partial rescue. The oracle denominator does not clearly improve alignment in the vision encoder. Its average cosine changes from $0.332$ for $u_{ff,t}$ to $0.272$ for $u_{fr,t}$. In the text encoder, the change is small, from $0.371$ to $0.378$. However, the scale still improves. The vision norm ratio decreases from $5.059$ to $3.652$, and the text norm ratio decreases from $6.555$ to $2.462$. Thus, the denominator fix repairs scale more than direction.

The probe losses confirm this intermediate behavior. On Oxford-Pets, $u_{ff,t}$ increases the mean test loss by $0.457$, while $u_{fr,t}$ increases it by $0.272$. This is a $40.6\%$ reduction. The frozen direction is almost identical to the oracle direction, with mean test damage $0.271$. The reference direction $u_{rr,t}$ gives $0.084$ over the same 30 checkpoints. Over the common checkpoints $t=5,\ldots,29$, the frozen denominator gives the smallest mean test-loss increase among denominator-only variants ($0.046$), followed by the lagged denominator ($0.171$) and the full-train denominator ($0.199$). Thus, denominator repair helps on Oxford-Pets, but it does not reconstruct the reference update. The remaining gap is explained by numerator error that is already present on the Adam trajectory.

DTD is the limiting case. In Figure~\ref{fig:adam_fail_exp4_geometry}, replacing the denominator does not improve alignment. The average vision cosine changes from $0.114$ for $u_{ff,t}$ to $0.099$ for $u_{fr,t}$. The average text cosine changes from $-0.017$ to $-0.074$. The oracle denominator also drives the norm ratio below one during much of the trajectory. This is not a useful rescue of scale. It indicates that the few-shot numerator has already become too small or misaligned relative to the full-train numerator.

Even on DTD, denominator repair is not useless. The mean test damage decreases from $2.558$ for $u_{ff,t}$ to $1.590$ for $u_{fr,t}$, a $37.8\%$ reduction. The frozen direction performs better, with mean test damage $1.258$, corresponding to a $50.8\%$ reduction. Over the common checkpoints $t=5,\ldots,29$, the lagged direction is weaker, with mean test damage $1.863$ versus $1.251$ for the frozen denominator and $1.496$ for the full-train denominator. However, all denominator-only directions still increase the test loss. In contrast, the full reference direction $u_{rr,t}$ decreases the mean test loss by $1.004$ over the 30 checkpoints. Therefore, DTD cannot be repaired by the denominator alone. By the time Adam reaches these checkpoints, the numerator is already stale.

The frozen and lagged denominators clarify the mechanism. The frozen denominator $u_{f0,t}$ is useful when the live few-shot denominator has been corrupted by the trajectory. It gives the smallest mean test-loss increase among denominator-only probes on Oxford-Pets and DTD over the common 25-checkpoint window. The lagged denominator $u_{f\mathrm{lag},t}$ is inconsistent. It is worse than the live denominator on EuroSAT and improves on it on Oxford-Pets and DTD, but remains worse than the frozen denominator on both datasets over the common window. Therefore, temporal lag alone is not the universal cause. The more stable pattern is denominator-scale corruption followed by trajectory-induced numerator drift.

Experiment~5 sharpens the diagnosis from Experiments~3 and~4. The denominator is causal. On EuroSAT, fixing it rescues most of Adam's one-step damage. On Oxford-Pets, it provides a meaningful but incomplete rescue. On DTD, it reduces the damage but does not change the sign of the full-train or test probe. The failure has already become a joint denominator-and-numerator problem.

\begin{tcolorbox}[
  enhanced,
  unbreakable,
  colback=findingblueback,
  colframe=findingblueframe,
  colbacktitle=findingblueframe,
  coltitle=white,
  boxrule=0.7pt,
  arc=1.5pt,
  left=6pt,
  right=6pt,
  top=5pt,
  bottom=5pt,
  before skip=7pt,
  after skip=7pt,
  fonttitle=\bfseries\small,
  fontupper=\small,
  title={Conclusion}]
Fixing only Adam's denominator rescues a large part of the failure on EuroSAT, a moderate part on Oxford-Pets, and only a limited part on DTD. The oracle denominator reduces the mean test-loss damage by $73.2\%$, $40.6\%$, and $37.8\%$ on the three datasets. A frozen initialization denominator is especially useful on DTD, where it reduces the damage by $50.8\%$. These results show that denominator miscalibration is a causal trigger of Adam collapse. They also show that denominator repair is not sufficient after Adam has induced numerator drift. %
\end{tcolorbox}

\subsection{Does Adam's performance collapse mainly because of scale, or because of geometry?}

Experiment~6 separates the update scale from the update geometry. We compare three trajectories: plain SGD, raw Adam, and a step-$0$ matched Adam trajectory. We denote the matched trajectory as Adam--M. For optimizer $o\in\{\mathrm{SGD},\mathrm{Adam},\mathrm{Adam\mbox{-}M}\}$, let $\Theta^{(o)}_t\in\mathbb{R}^d$ be the trainable parameter vector at optimizer step $t$, where $d$ is the total number of trainable scalar parameters. Let
\begin{equation}
    \Delta\Theta^{(o)}_t
    = \Theta^{(o)}_{t+1}-\Theta^{(o)}_t
    \in\mathbb{R}^d
\end{equation}
be the update applied at step $t$. We measure its global relative update budget by
\begin{equation}
    r_o(t)
    = \frac{\lVert\Delta\Theta^{(o)}_t\rVert_2}
           {\lVert\Theta^{(o)}_t\rVert_2}
    \in\mathbb{R}_{\ge 0} .
    \label{eq:exp5_global_budget}
\end{equation}
This quantity is dimensionless. Raw Adam and SGD both use the requested learning rate $10^{-4}$. Adam--M uses a single scalar calibration at step~$0$. If $r_{\mathrm{Adam}}(0)$ is the first raw Adam budget and $r_{\mathrm{SGD}}(0)$ is the first SGD budget, then Adam--M uses the learning rate
\begin{equation}
    \eta_{\mathrm{Adam\mbox{-}M}}
    = \gamma\eta_{\mathrm{Adam}},
    \qquad
    \gamma = \frac{r_{\mathrm{SGD}}(0)}{r_{\mathrm{Adam}}(0)} .
    \label{eq:exp5_lr_match}
\end{equation}
No later learning-rate correction is applied. Thus, Experiment~6 asks whether Adam still fails after its initial global update budget has been matched to SGD.

We also measure a geometry diagnostic that keeps the full-train denominator fixed. For encoder $e\in\{\mathrm{vision},\mathrm{text}\}$ and trajectory $o$, let $m^{(o,e)}_{\mathrm{few},t},m^{(o,e)}_{\mathrm{full},t}\in\mathbb{R}^{d_e}$ be the support and full-train gradients evaluated at $\Theta_t^{(o)}$. Let $v^{(o,e)}_{\mathrm{full},t}\in\mathbb{R}_{\ge0}^{d_e}$ be the full-train diagonal second moment at the same point. We define
\begin{equation}
    u^{(o,e)}_{fr,t}
    = \frac{m^{(o,e)}_{\mathrm{few},t}}
           {\sqrt{v^{(o,e)}_{\mathrm{full},t}+\varepsilon}},
    \qquad
    u^{(o,e)}_{rr,t}
    = \frac{m^{(o,e)}_{\mathrm{full},t}}
           {\sqrt{v^{(o,e)}_{\mathrm{full},t}+\varepsilon}}
    \in \mathbb{R}^{d_e} .
    \label{eq:exp5_geometry_directions}
\end{equation}
The square root and division are elementwise. The first direction uses the few-shot numerator and the full-train denominator. The second direction uses both the full-train numerator and the full-train denominator. Therefore, their cosine isolates the numerator-side geometry after the denominator scale has been fixed:
\begin{equation}
    A^{(e)}_o(t)
    = \cos\!\left(u^{(o,e)}_{fr,t},u^{(o,e)}_{rr,t}\right)
    \in[-1,1] .
    \label{eq:exp5_geometry}
\end{equation}
We plot the encoder average
\begin{equation}
    \overline{A}_o(t)
    = \frac{1}{2}\left(A^{(\mathrm{vision})}_o(t)+A^{(\mathrm{text})}_o(t)\right) .
    \label{eq:exp5_avg_geometry}
\end{equation}
A high value of $\overline{A}_o(t)$ means that the few-shot numerator is well aligned with the full-train reference direction under the same denominator. It does not measure the size of the update.

\begin{figure*}[t]
    \centering
    \includegraphics[width=0.98\linewidth]{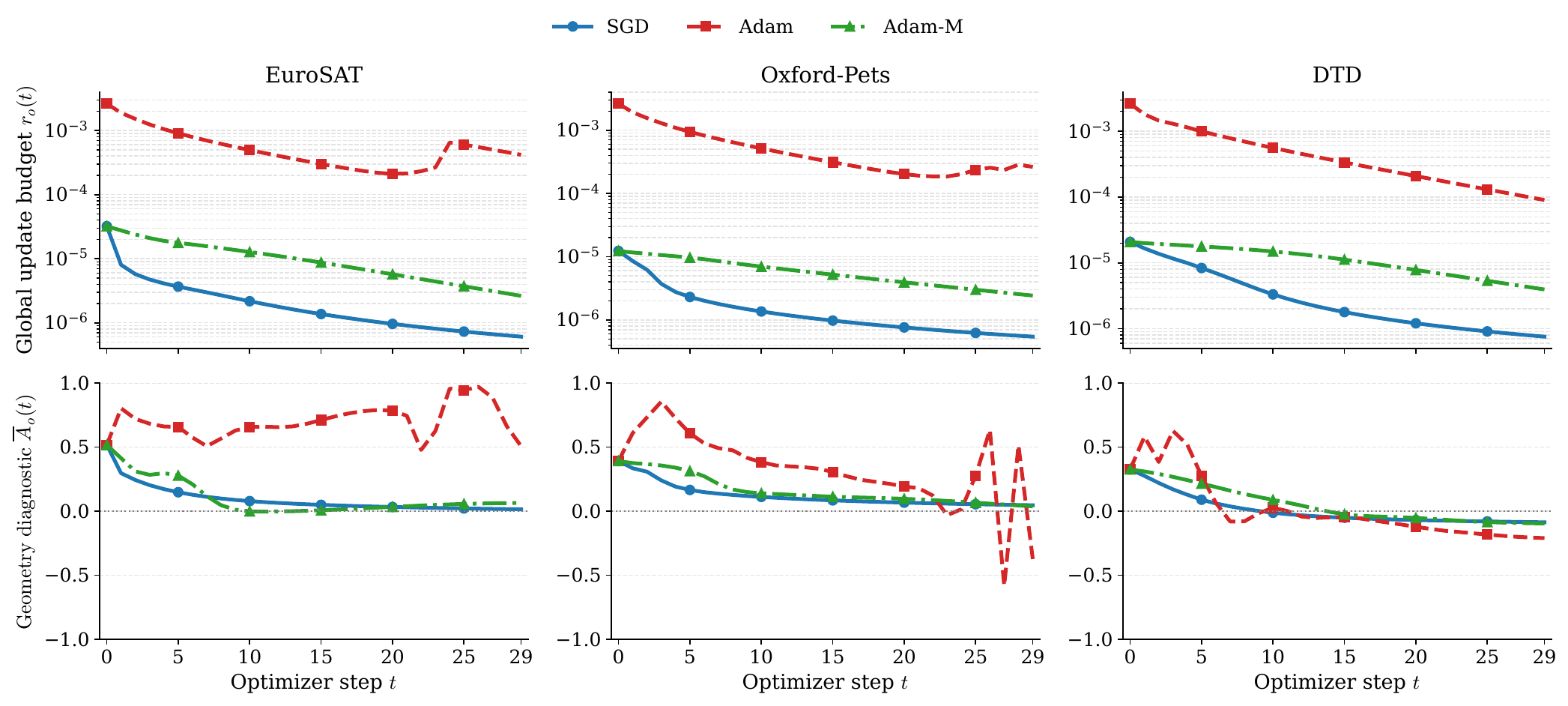}
    \caption{Experiment~6: scale versus geometry. Top row: global relative update budget $r_o(t)$ from Eq.~\eqref{eq:exp5_global_budget}. Bottom row: encoder-averaged geometry diagnostic $\overline{A}_o(t)$ from Eq.~\eqref{eq:exp5_avg_geometry}. Adam--M is calibrated only once at step~$0$. Plain Adam starts with a much larger budget than SGD. In contrast, its geometry diagnostic can be higher than SGD while the trajectory still fails.}
    \label{fig:adam_fail_exp5_scale_geometry}
\end{figure*}

\begin{figure*}[t]
    \centering
    \includegraphics[width=0.98\linewidth]{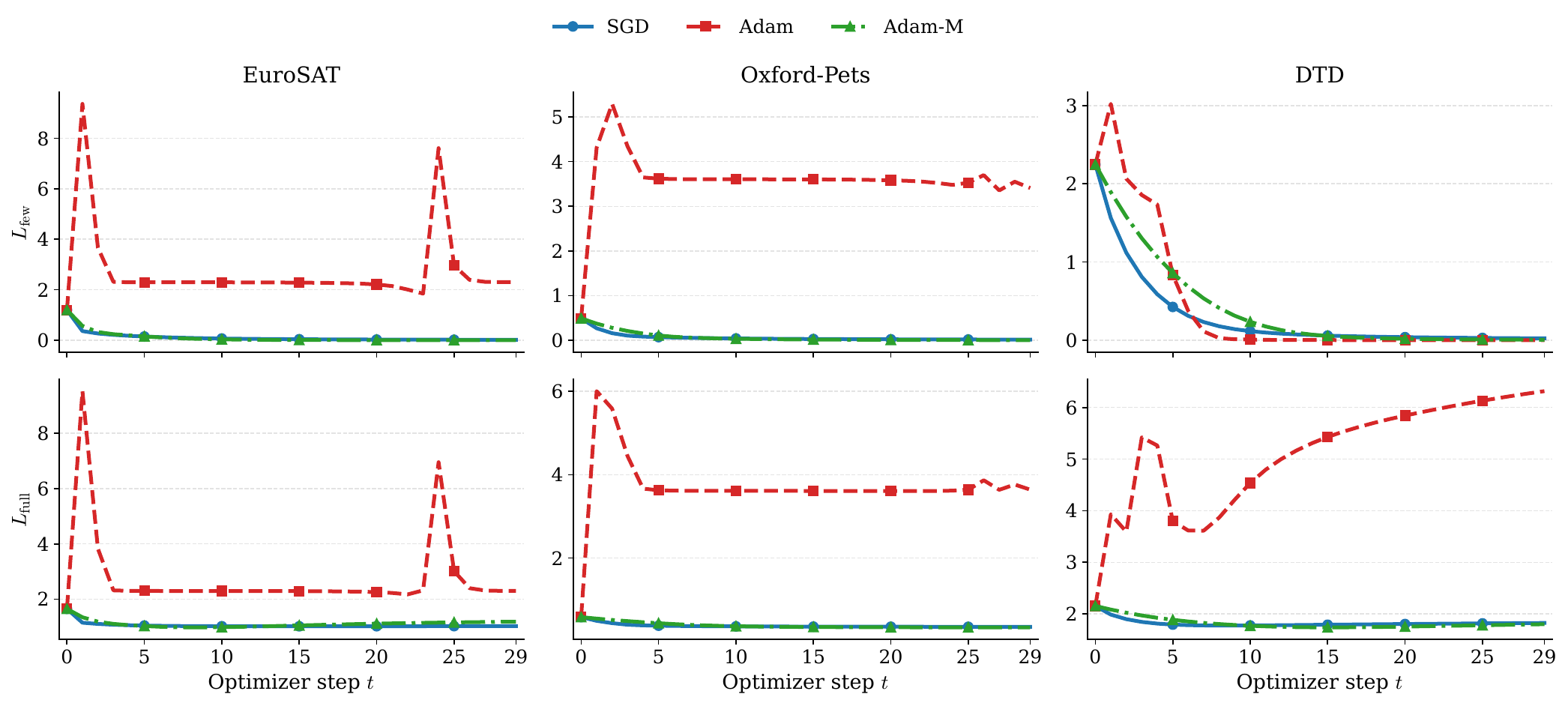}
    \caption{Experiment~6: support and full-train losses for SGD, plain Adam, and Adam--M. The full-train loss is $L_{\mathrm{full}}$ because the reference split is the full train split. Plain Adam either fails to make useful progress or destructively overfits the support set. Adam--M removes most of this collapse.}
    \label{fig:adam_fail_exp5_loss}
\end{figure*}

\begin{figure*}[t]
    \centering
    \includegraphics[width=0.98\linewidth]{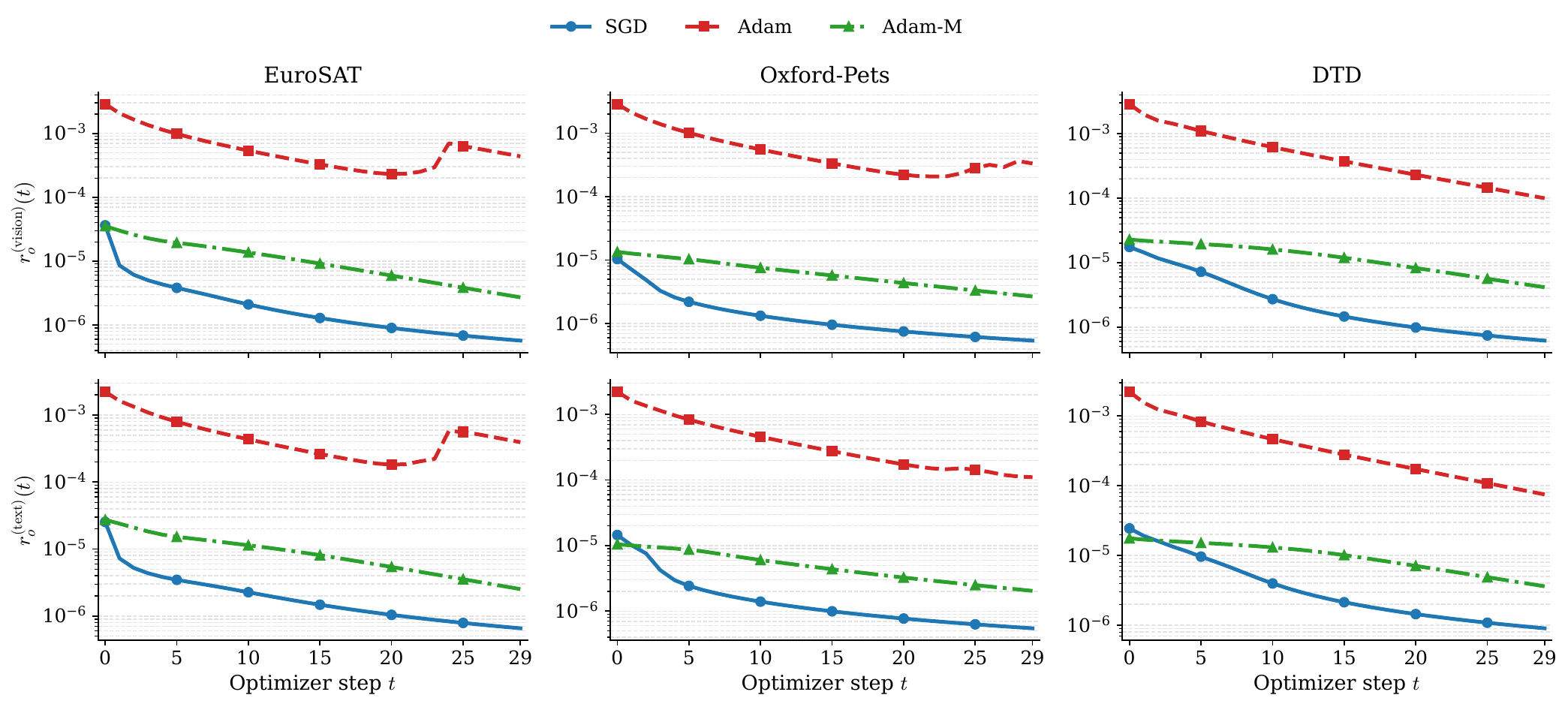}
    \caption{Experiment~6: encoder-level relative update budgets. For encoder $e$, the plotted quantity is $r_o^{(e)}(t)=\lVert\Delta\Theta^{(o,e)}_t\rVert_2/\lVert\Theta^{(o,e)}_t\rVert_2$. Global step-$0$ matching aligns the whole-model budget but does not keep the vision and text budgets aligned over time.}
    \label{fig:adam_fail_exp5_encoder_budget}
\end{figure*}

\begin{figure*}[t]
    \centering
    \includegraphics[width=0.98\linewidth]{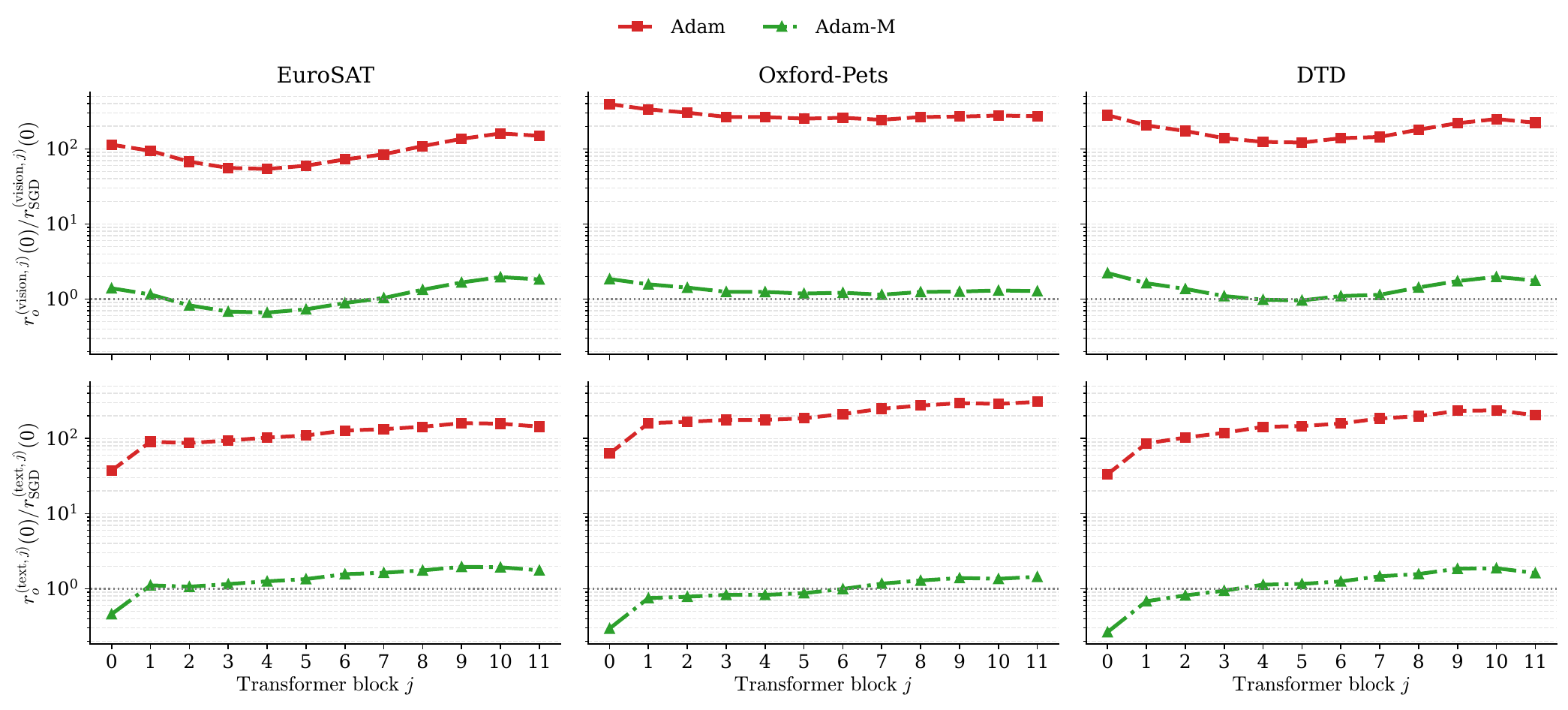}
    \caption{Experiment~6: blockwise step-$0$ update-budget ratios relative to SGD. For encoder $e$ and Transformer block $j$, the plotted quantity is $r_o^{(e,j)}(0)/r_{\mathrm{SGD}}^{(e,j)}(0)$, where $r_o^{(e,j)}(0)=\lVert\Delta\Theta^{(o,e,j)}_0\rVert_2/\lVert\Theta^{(o,e,j)}_0\rVert_2$. The dotted line marks perfect blockwise matching. Adam--M matches the global step-$0$ budget, but a substantial blockwise scale mismatch remains.}
    \label{fig:adam_fail_exp5_block_budget}
\end{figure*}

Figure~\ref{fig:adam_fail_exp5_scale_geometry} shows the central scale fact. At the first step, plain Adam has $r_{\mathrm{Adam}}(0)=2.62\times10^{-3}$ on all three datasets. The corresponding SGD budgets are much smaller: $3.21\times10^{-5}$ on EuroSAT, $1.23\times10^{-5}$ on Oxford-Pets, and $2.08\times10^{-5}$ on DTD. Therefore, the first plain Adam update is $81.6\times$, $213.0\times$, and $126.2\times$ larger than SGD. The matching coefficients in Eq.~\eqref{eq:exp5_lr_match} are $0.01225$, $0.00470$, and $0.00792$. These numbers show that plain Adam begins with an update budget that is about two orders of magnitude too large.

Figure~\ref{fig:adam_fail_exp5_loss} shows that this scale correction removes most of Adam's collapse in performance. Plain Adam fails badly on the full-train objective. Its final values of $L_{\mathrm{full}}$ are $2.300$ on EuroSAT, $3.640$ on Oxford-Pets, and $6.317$ on DTD. The DTD behavior is especially revealing: plain Adam drives the support loss close to zero, but the full-train loss grows. Thus, Adam is not merely slow. It can also move into a destructive support-fitting regime. Adam--M changes this behavior. Its final full-train losses are $1.185$ on EuroSAT, $0.336$ on Oxford-Pets, and $1.800$ on DTD. The corresponding SGD values are $1.027$, $0.350$, and $1.824$. Hence, a single step-$0$ budget match makes Adam essentially competitive with SGD on Oxford-Pets and DTD. It also results in a large rescue on EuroSAT, although SGD remains better there.

The geometry diagnostic in Figure~\ref{fig:adam_fail_exp5_scale_geometry} rules out a geometry-first explanation. Plain Adam often has a larger value of $\overline{A}_o(t)$ than SGD, especially on EuroSAT. Encoder-wise, the late mean of $A^{(e)}_o(t)$ on EuroSAT is about $0.773$ in the vision encoder and $0.738$ in the text encoder for plain Adam, but only about $0.023$ and $0.022$ for SGD. Plain Adam still fails. Conversely, Adam--M is stable without a dramatic improvement in $\overline{A}_o(t)$. Its geometry is close to SGD on EuroSAT and DTD, and only moderately better on Oxford-Pets. Therefore, favorable numerator-side geometry is not sufficient for Adam to work. The decisive intervention is the scale correction.

Figures~\ref{fig:adam_fail_exp5_encoder_budget} and~\ref{fig:adam_fail_exp5_block_budget} show residual scale differences after global matching. Global step-$0$ calibration does not imply encoder-wise or blockwise calibration. At step~$0$, the Adam--M to SGD encoder-budget ratio is $1.30$ in Oxford-Pets vision but $0.72$ in Oxford-Pets text. The same split appears on DTD, where the ratios are $1.30$ in vision and $0.72$ in text. At the final step, Adam--M again has a larger global update budget than SGD: $4.33\times$ on EuroSAT, $4.46\times$ on Oxford-Pets, and $5.22\times$ on DTD. The blockwise plot shows the same issue at a finer resolution. Even at step~$0$, after global matching, the matched-to-SGD block ratios range from $0.46$ to $1.96$ on EuroSAT text, from $0.30$ to $1.44$ on Oxford-Pets text, and from $0.26$ to $1.88$ on DTD text. Thus, Adam's scale problem is not a single global scalar. It is structured by encoder, block, and time.

Experiment~6, therefore, separates two notions of scale. The first is the global update budget $r_o(t)$. Plain Adam fails because this budget is about two orders of magnitude larger than the SGD budget at the first step. The step-$0$ match corrects this global error and removes most of the loss collapse. The second notion is the local update scale. A single scalar $\gamma$ does not keep the encoder budgets $r_o^{(e)}(t)$ aligned over time, and it does not match the block budgets $r_o^{(e,j)}(0)$ at initialization. These measurements show residual scale differences across time, encoders, and blocks. They do not isolate the cause of the remaining EuroSAT full-train loss gap between Adam--M and SGD.

\begin{findingbox}[Conclusion]
Adam's failure is scale-dominated. Plain Adam starts with a first-step relative update norm that is $81.6\times$, $213.0\times$, and $126.2\times$ larger than SGD on EuroSAT, Oxford-Pets, and DTD. A one-time match to SGD's step-$0$ global budget removes most of the collapse and makes Adam--M nearly as stable as SGD on Oxford-Pets and DTD. The geometry diagnostic $\overline{A}_o(t)$ does not explain plain Adam's failure because it measures numerator-side alignment but not update magnitude. Raw Adam can have a larger $\cos(u_{fr},u_{rr})$ than SGD while still failing. Global matching leaves substantial variation across blocks and later steps. These residual scale differences may contribute to the remaining EuroSAT full-train loss gap.
\end{findingbox}

\subsection{Does Adam fail because it pushes CLIP too far from the pretrained solution too fast?}

Experiment~7 tests whether the scale failure identified in Experiment~6 becomes a fast drift away from pretrained CLIP. We compare two trajectories, $o\in\{\mathrm{SGD},\mathrm{Adam}\}$. Let $\Theta^{(o)}_t\in\mathbb{R}^d$ be the trainable parameter vector at optimizer step $t$, where $d$ is the number of trainable scalar parameters. The checkpoint $t=0$ is the pretrained initialization, so $\Theta^{(o)}_0=\Theta_0$. The first post-update checkpoint is $t=1$. Let $T$ denote the final logged checkpoint.

For parameter group $G\subseteq\{1,\ldots,d\}$, the relative parameter drift is
\begin{equation}
    R_{\Theta}^{(o,G)}(t)
    = \frac{\lVert \Theta^{(o,G)}_t-\Theta^{(G)}_0\rVert_2}
           {\lVert \Theta^{(G)}_0\rVert_2}
    \in\mathbb{R}_{\ge0} .
    \label{eq:exp6_param_drift}
\end{equation}
We report this quantity for all trainable parameters, for the two encoders, and for each Transformer block. For representation drift, let $h^{(o)}_{\mathrm{img},t}(x)\in\mathbb{R}^{d_{\mathrm{emb}}}$ and $h^{(o)}_{\mathrm{text},t}(c)\in\mathbb{R}^{d_{\mathrm{emb}}}$ be the normalized CLIP image feature and class-text feature. Here $d_{\mathrm{emb}}$ is the CLIP embedding dimension, and $c\in\{1,\ldots,C\}$ is a class index. For a split $S\in\{\mathrm{few},\mathrm{full},\mathrm{test}\}$ with dataset $\mathcal{D}_S$, the feature drifts are
\begin{equation}
    R_{\mathrm{img},S}^{(o)}(t)
    = 1-\frac{1}{|\mathcal{D}_S|}\sum_{(x,y)\in\mathcal{D}_S}
      \left\langle h^{(o)}_{\mathrm{img},t}(x),h_{\mathrm{img},0}(x)\right\rangle,
    \label{eq:exp6_image_drift}
\end{equation}
\begin{equation}
    R_{\mathrm{text}}^{(o)}(t)
    = 1-\frac{1}{C}\sum_{c=1}^{C}
      \left\langle h^{(o)}_{\mathrm{text},t}(c),h_{\mathrm{text},0}(c)\right\rangle .
    \label{eq:exp6_text_drift}
\end{equation}
Both quantities are zero at initialization. Larger values mean larger angular drift from the pretrained CLIP features.

We also measure drift at the classifier-output level. Let $z^{(o)}_t(x)\in\mathbb{R}^{C}$ be the class-logit vector. The mean absolute logit drift on split $S$ is
\begin{equation}
    R_{\mathrm{logit},S}^{(o)}(t)
    = \frac{1}{|\mathcal{D}_S|C}\sum_{(x,y)\in\mathcal{D}_S}
      \left\lVert z^{(o)}_t(x)-z_0(x)\right\rVert_1 .
    \label{eq:exp6_logit_drift}
\end{equation}
For ground-truth class $y$, the margin is
\begin{equation}
    \mu^{(o)}_t(x,y)
    = z^{(o)}_t(x)_y - \max_{c\ne y} z^{(o)}_t(x)_c
    \in\mathbb{R} .
    \label{eq:exp6_margin}
\end{equation}
The corresponding margin drift is
\begin{equation}
    R_{\mathrm{mar},S}^{(o)}(t)
    = \frac{1}{|\mathcal{D}_S|}\sum_{(x,y)\in\mathcal{D}_S}
      \left|\mu^{(o)}_t(x,y)-\mu_0(x,y)\right| .
    \label{eq:exp6_margin_drift}
\end{equation}
These output-level diagnostics distinguish small feature changes from changes that actually perturb the classifier geometry.

\begin{figure*}[t]
    \centering
    \includegraphics[width=0.98\linewidth]{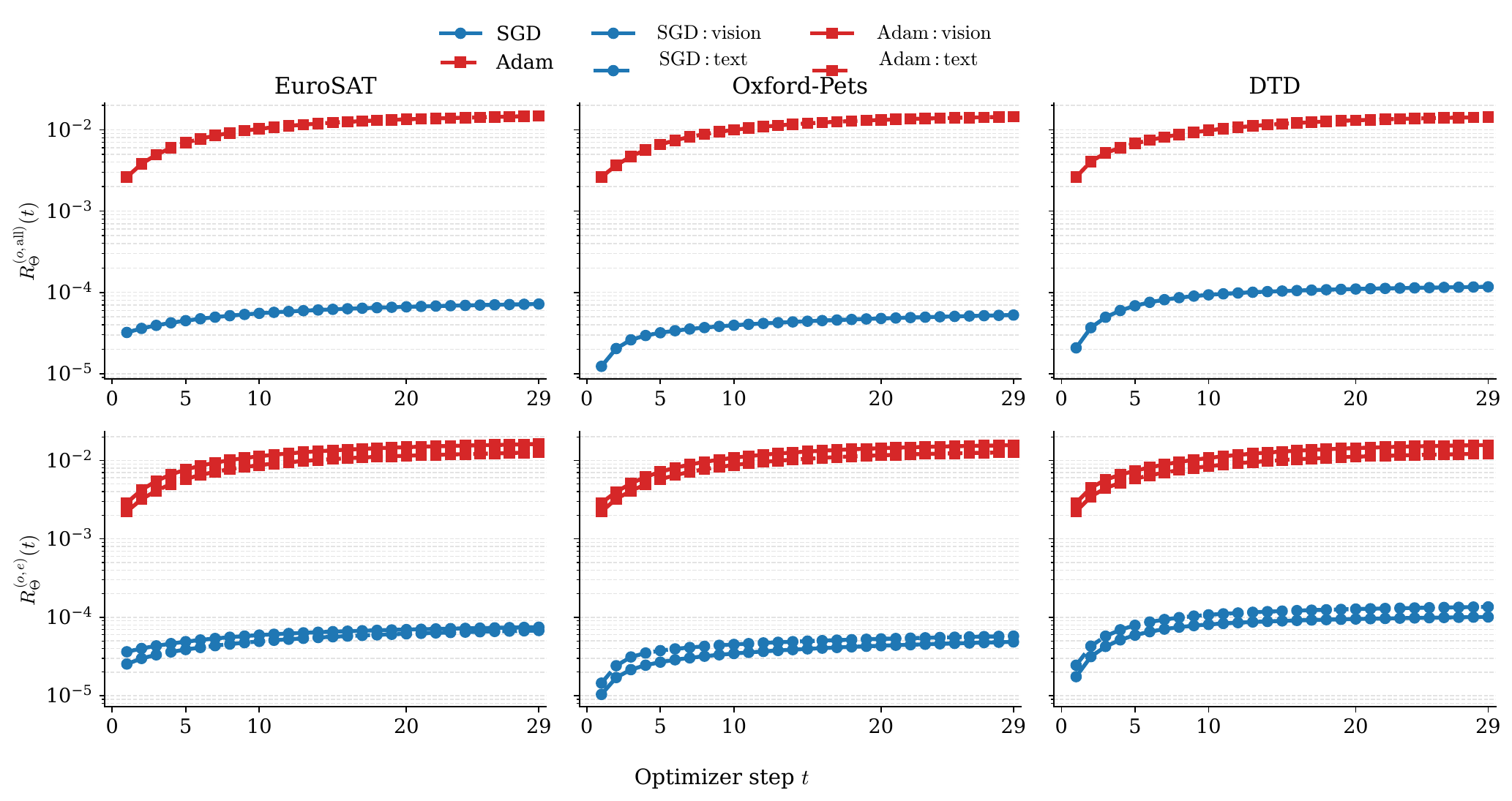}
    \caption{Experiment~7: parameter drift from the pretrained CLIP initialization. The top row reports $R_{\Theta}^{(\mathrm{all})}(t)$ from Eq.~\eqref{eq:exp6_param_drift}. The bottom row reports encoder-level drift $R_{\Theta}^{(e)}(t)$ for $e\in\{\mathrm{vision},\mathrm{text}\}$. The vertical axis is logarithmic. Adam moves two to three orders of magnitude farther from $\Theta_0$ than SGD.}
    \label{fig:adam_fail_exp6_parameter_drift}
\end{figure*}

\begin{figure*}
    \centering
    \includegraphics[width=0.98\linewidth]{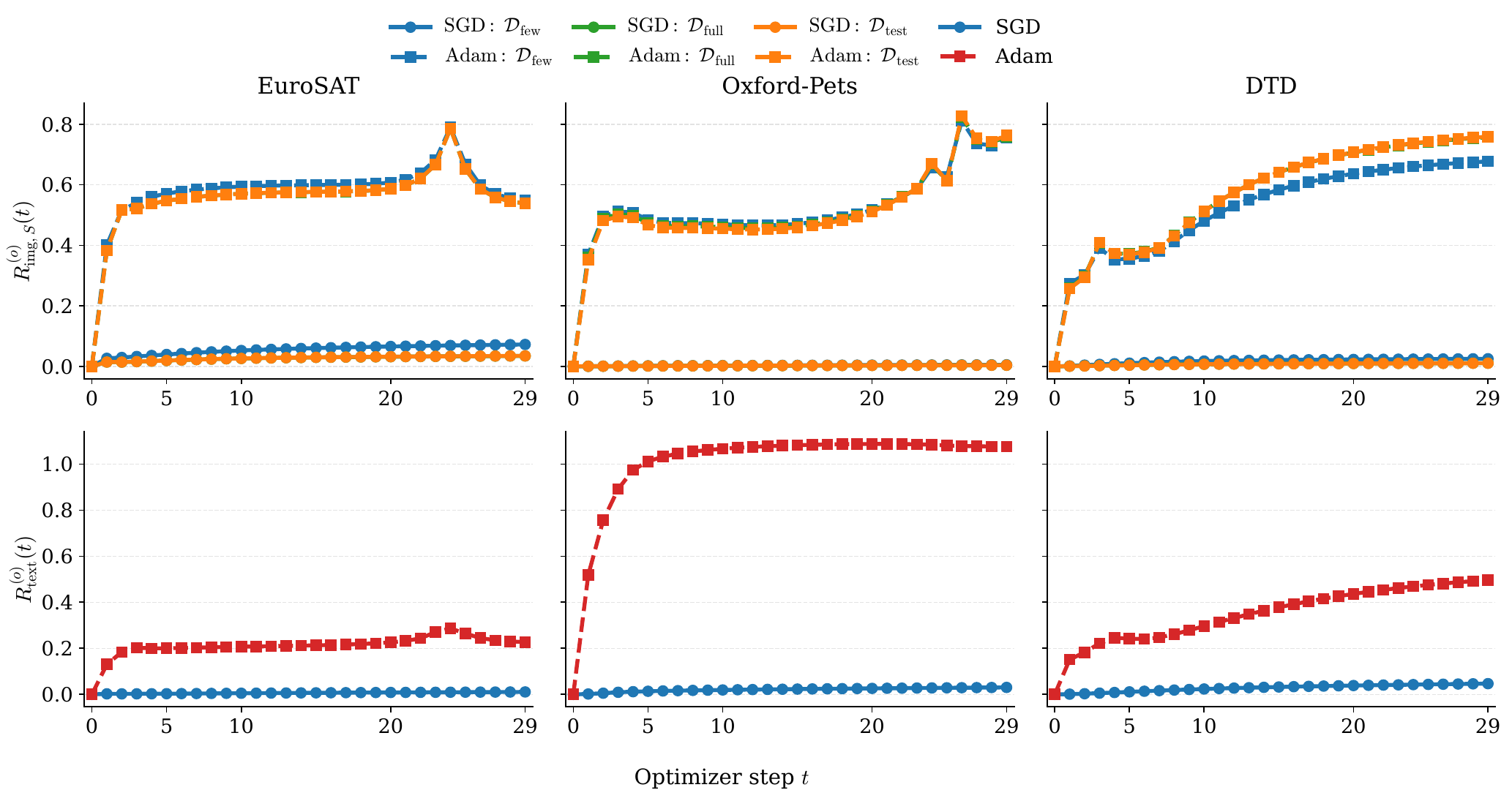}
    \caption{Experiment~7: image-feature and text-feature drift. The top row reports $R_{\mathrm{img},S}^{(o)}(t)$ from Eq.~\eqref{eq:exp6_image_drift} for $S\in\{\mathrm{few},\mathrm{full},\mathrm{test}\}$. The bottom row reports $R_{\mathrm{text}}^{(o)}(t)$ from Eq.~\eqref{eq:exp6_text_drift}. Adam drifts on the full-train and test splits as well as on the support set, so the damage is not confined to support memorization.}
    \label{fig:adam_fail_exp6_feature_drift}
\end{figure*}

\begin{figure*}
    \centering
    \includegraphics[width=0.98\linewidth]{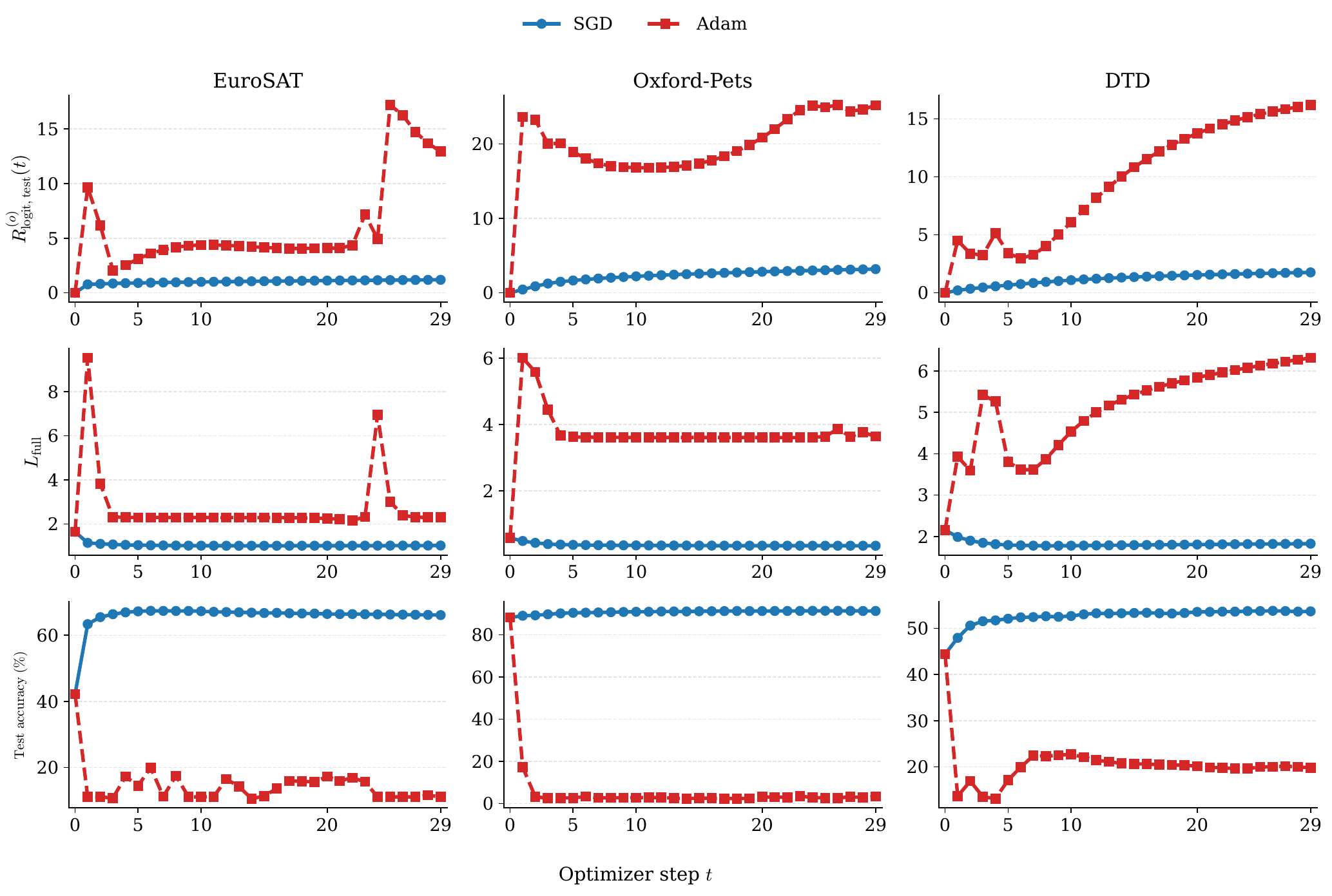}
    \caption{Experiment~7: output drift and performance collapse. The top row reports test-logit drift $R_{\mathrm{logit},\mathrm{test}}^{(o)}(t)$ from Eq.~\eqref{eq:exp6_logit_drift}. The middle row reports $L_{\mathrm{full}}$. The bottom row reports test accuracy. A collapse of the full-train and test objectives accompanies the same fast-drift event that perturbs the logits.}
    \label{fig:adam_fail_exp6_logit_performance}
\end{figure*}

\begin{figure*}
    \centering
    \includegraphics[width=0.98\linewidth]{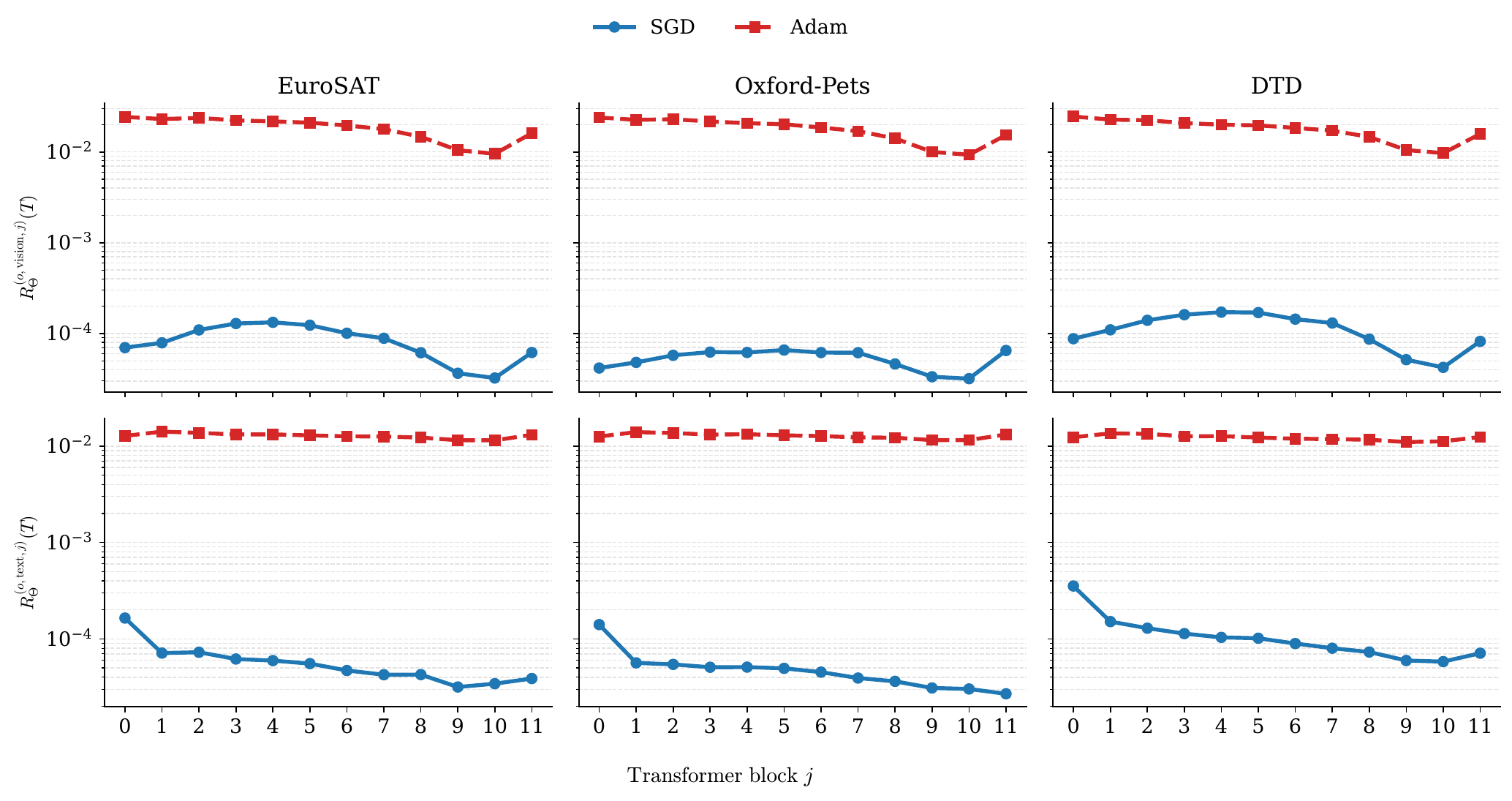}
    \caption{Experiment~7: final blockwise parameter drift. The plotted value is $R_{\Theta}^{(o,e,j)}(T)$ for encoder $e$ and Transformer block $j$. Adam's extra drift is structured. It is the largest in the earliest vision blocks and in the first text blocks, whereas SGD remains much closer to the pretrained solution across all blocks.}
    \label{fig:adam_fail_exp6_block_drift}
\end{figure*}

Figure~\ref{fig:adam_fail_exp6_parameter_drift} shows that Adam does not fail by slowly drifting into a bad region. It jumps away from the pretrained solution after one update. At $t=1$, $R_{\Theta}^{(\mathrm{Adam},\mathrm{all})}(1)\simeq2.62\times10^{-3}$ on all three datasets. The corresponding SGD values are $3.21\times10^{-5}$ on EuroSAT, $1.23\times10^{-5}$ on Oxford-Pets, and $2.08\times10^{-5}$ on DTD. Thus, the first Adam displacement is $81.6\times$, $213.0\times$, and $126.2\times$ larger than the first SGD displacement. These ratios match the update-budget ratios from Experiment~6. Thus, the scale problem appears directly as parameter drift.

The final checkpoint shows the same effect at a larger magnitude. Adam ends with $R_{\Theta}^{(\mathrm{all})}(T)=1.48\times10^{-2}$ on EuroSAT, $1.45\times10^{-2}$ on Oxford-Pets, and $1.43\times10^{-2}$ on DTD. SGD ends at $7.19\times10^{-5}$, $5.26\times10^{-5}$, and $1.17\times10^{-4}$. Adam is therefore $206\times$, $275\times$, and $122\times$ farther from $\Theta_0$ at the final checkpoint. The encoder-level curves in Figure~\ref{fig:adam_fail_exp6_parameter_drift} show that the drift occurs in both branches of CLIP. It is not a text-only or vision-only effect.

Figure~\ref{fig:adam_fail_exp6_feature_drift} shows that this parameter drift is representation drift. The final test image-feature drift under Adam is $0.538$ on EuroSAT, $0.764$ on Oxford-Pets, and $0.759$ on DTD. The corresponding SGD values are $0.0346$, $0.00381$, and $0.0103$. The text side also moves far more under Adam. The final text-feature drift is $0.227$ on EuroSAT, $1.076$ on Oxford-Pets, and $0.497$ on DTD, compared with $0.0106$, $0.0303$, and $0.0467$ for SGD. Oxford-Pets is especially revealing. Adam starts from a high zero-shot and SGD baseline, but it still drives the class-text features far from the pretrained prompt geometry.

The top row of Figure~\ref{fig:adam_fail_exp6_feature_drift} also rules out a narrow support-only interpretation. Under Adam, the support, full-train, and test image-feature drifts rise together. If Adam only memorized the support set locally, the support features would move much more than the full-train and test features. Instead, the full-train and test features are also strongly displaced. Therefore, Adam distorts the shared CLIP representation with respect to the broader data distribution.

Figure~\ref{fig:adam_fail_exp6_logit_performance} shows that feature drift reaches the classifier outputs. At the final checkpoint, the test-logit drift is $12.95$ on EuroSAT, $25.20$ on Oxford-Pets, and $16.18$ on DTD for Adam. The corresponding SGD values are $1.20$, $3.19$, and $1.76$. Hence, Adam changes the test logits by about $8\times$ to $11\times$ more than SGD. The same figure shows the performance consequence. EuroSAT Adam collapses to chance-level test accuracy, about $11.1\%$, while SGD ends at $66.1\%$. Oxford-Pets Adam falls to $3.38\%$, while SGD ends at $91.3\%$. DTD Adam ends at $19.8\%$, while SGD ends at $53.7\%$.

DTD clarifies the mechanism. Adam reaches $100\%$ support accuracy by the end of training, and its final support loss is $7.6\times10^{-5}$. Yet its full-train loss is $6.32$, and its test loss is $6.78$. Thus, Adam is not simply unable to optimize the support objective. It can fit the support while destroying the pretrained structure needed for full-train and test generalization. This is the behavior expected from an update that is too large for preserving CLIP, not from a lack of descent signal.

Figure~\ref{fig:adam_fail_exp6_block_drift} shows that Adam's drift is structured across depth. The largest final Adam drift in the vision encoder occurs in the earliest vision blocks, especially block $0$. In the text encoder, the largest drift occurs in the first blocks, especially block $1$. This pattern matters because early blocks affect all downstream features. SGD is also structured, but its absolute drift is much smaller. Therefore, Adam does not merely move more. It moves more in the sensitive early parts of the pretrained model.

Experiment~7 completes the causal chain from Experiments~1--6. The few-shot second moment is too small. Adam's adaptive update is therefore too large. The update that is too large produces immediate parameter drift. That drift changes image features, text features, logits, and margins. Once this occurs, the full-train and test objectives collapse. Therefore, feature drift is not a separate explanation for Adam's failure. It is the dynamical consequence of the denominator-scale problem.

\begin{findingbox}[Conclusion]
The miscalibrated few-shot denominator of Adam pushes CLIP too far from the pretrained solution too fast. After one update, Adam is already $81.6\times$, $213.0\times$, and $126.2\times$ farther from $\Theta_0$ than SGD on EuroSAT, Oxford-Pets, and DTD. By the final checkpoint, Adam is $122\times$--$275\times$ farther away in parameter space and much farther away in image, text, and logit space. The drift affects the full-train and test splits, not only the support set. Thus, Adam's collapse is a fast representation-drift consequence of the second-moment scale failure.
\end{findingbox}

\subsection{Does the few-shot gradient itself become stale over time?}

Experiment~8 tests whether Adam fails because the support gradient itself becomes temporally stale. This is a different question from the denominator pathologies studied above. The experiment compares two trajectories, $o\in\{\mathrm{SGD},\mathrm{Adam}\}$. For each trajectory, let $\Theta_t^{(o)}\in\mathbb{R}^d$ be the trainable parameter vector at step $t$. The full-support protocol uses one optimizer step per epoch. Hence, a temporal lag $\Delta\in\{1,2,5\}$ corresponds to $1$, $2$, or $5$ epochs.

For encoder $e\in\{\mathrm{vision},\mathrm{text}\}$, let $d_e$ be the number of trainable scalar parameters in that encoder. We define the encoder-restricted support and full-train gradients
\begin{equation}
    m_{\mathrm{few},t}^{(o,e)}
    = \nabla_{\Theta^{(e)}} L_{\mathrm{few}}\!\left(\Theta_t^{(o)}\right)
    \in\mathbb{R}^{d_e},
    \qquad
    m_{\mathrm{full},t}^{(o,e)}
    = \nabla_{\Theta^{(e)}} L_{\mathrm{full}}\!\left(\Theta_t^{(o)}\right)
    \in\mathbb{R}^{d_e} .
    \label{eq:exp7_temporal_gradients}
\end{equation}
The cosine is the inner product of two nonzero vectors divided by the product of their Euclidean norms. We measure two forms of temporal alignment:
\begin{align}
    A_{\mathrm{self}}^{(o,e)}(t,\Delta)
    &= \cos\!\left(m_{\mathrm{few},t}^{(o,e)},
                   m_{\mathrm{few},t+\Delta}^{(o,e)}\right),
    \label{eq:exp7_self_alignment}\\
    A_{\mathrm{full}}^{(o,e)}(t,\Delta)
    &= \cos\!\left(m_{\mathrm{few},t}^{(o,e)},
                   m_{\mathrm{full},t+\Delta}^{(o,e)}\right).
    \label{eq:exp7_full_alignment}
\end{align}
The first quantity measures \emph{directional self-stability}. It asks whether the few-shot gradient direction changes over time. As a cosine, it is invariant to positive rescaling and does not measure changes in gradient magnitude. The second quantity measures \emph{future full-data alignment}. It asks whether the current support gradient still predicts the future full-train objective. These two quantities should not be conflated. A support-gradient direction can be temporally stable and still be stale as a surrogate for $L_{\mathrm{full}}$.

\begin{figure*}[t]
    \centering
    \includegraphics[width=0.98\linewidth]{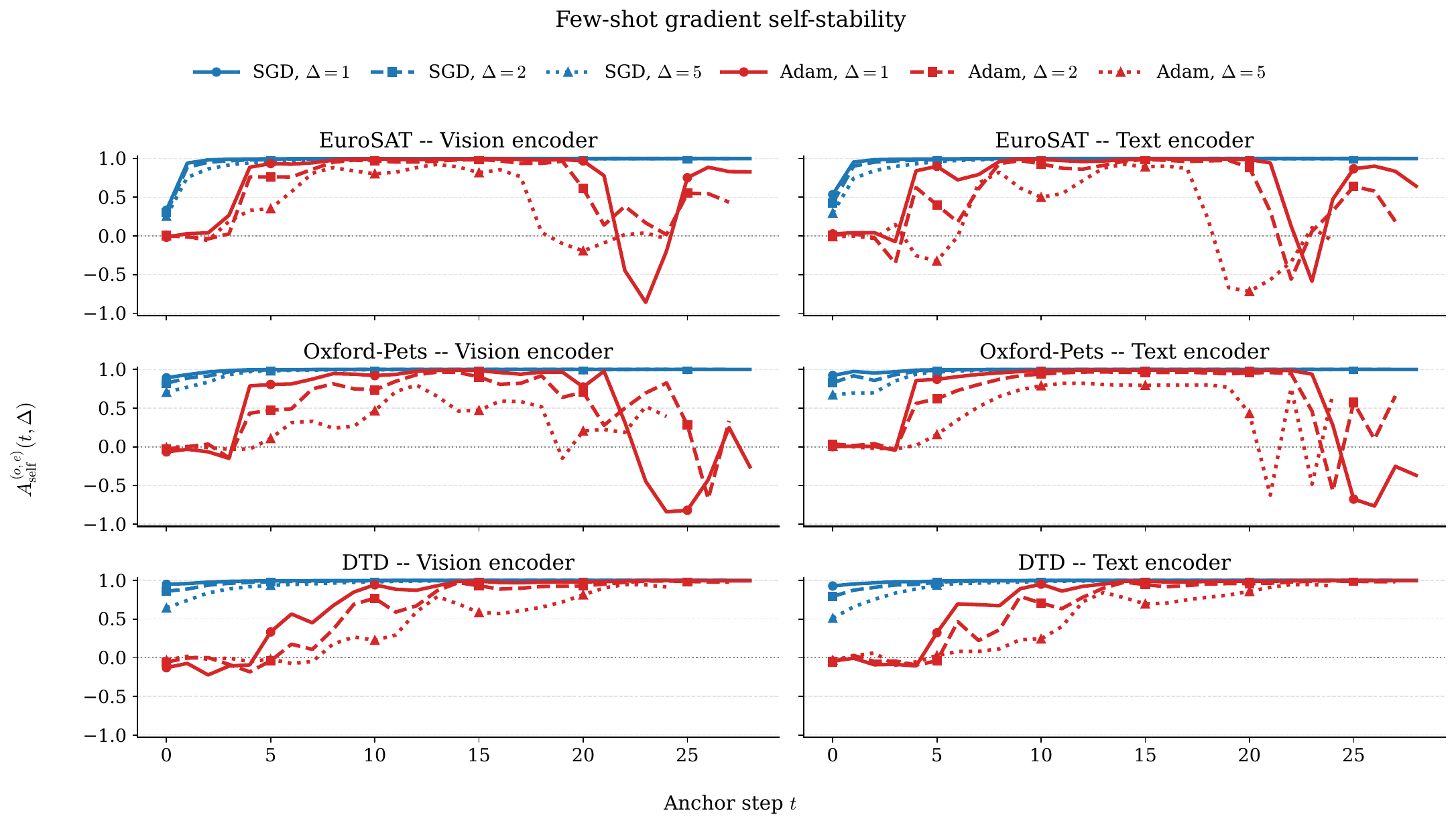}
    \caption{Experiment~8: few-shot gradient self-stability. The plotted quantity is $A_{\mathrm{self}}^{(o,e)}(t,\Delta)$ from Eq.~\eqref{eq:exp7_self_alignment}. Under SGD, the support-gradient direction rapidly becomes nearly time-invariant. Under Adam, the same quantity is lower and more oscillatory, especially on EuroSAT and Oxford-Pets.}
    \label{fig:adam_fail_exp7_self}
\end{figure*}

\begin{figure*}
    \centering
    \includegraphics[width=0.98\linewidth]{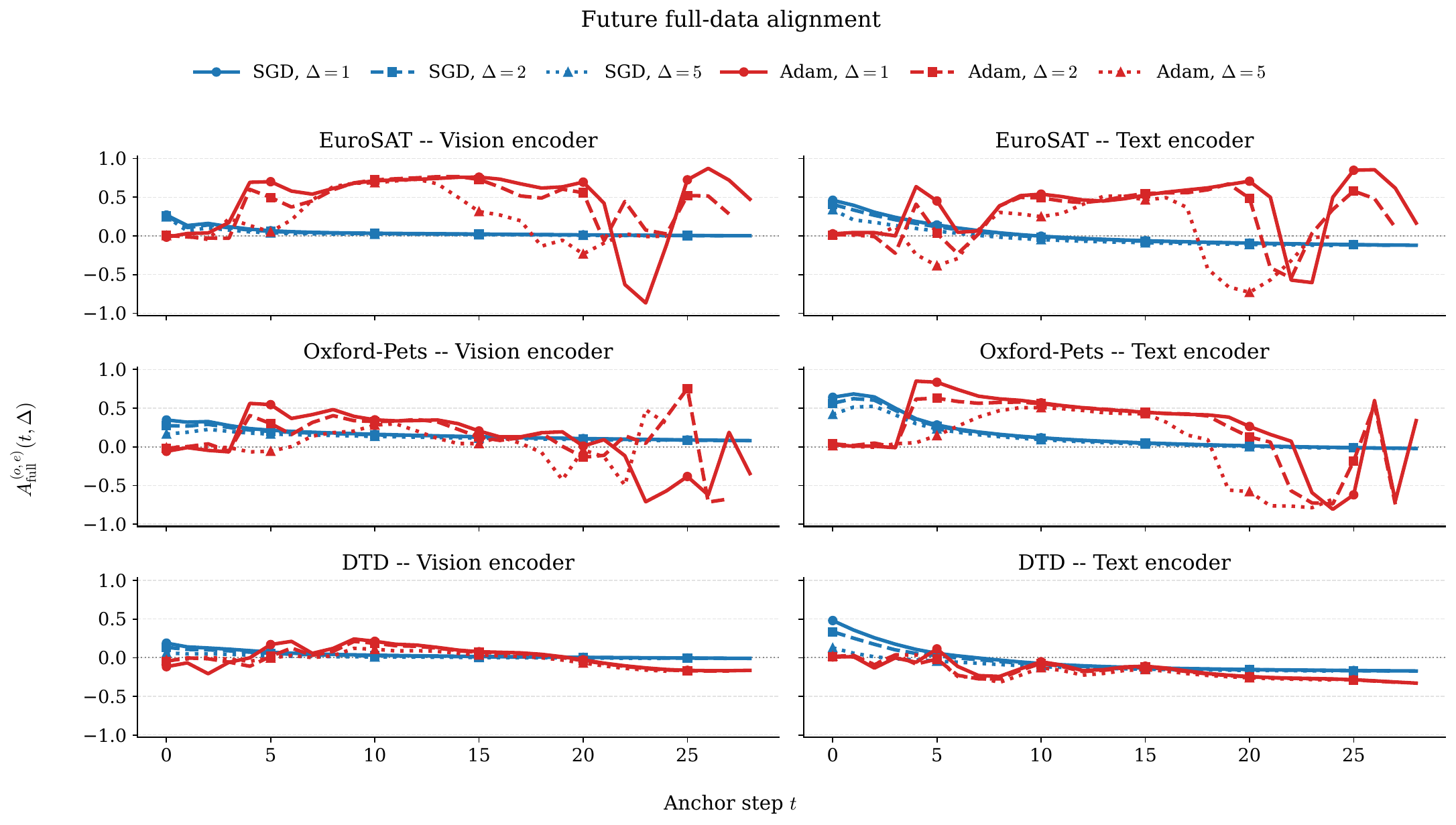}
    \caption{Experiment~8: future full-data alignment. The plotted quantity is $A_{\mathrm{full}}^{(o,e)}(t,\Delta)$ from Eq.~\eqref{eq:exp7_full_alignment}. Under SGD, the directionally stable support gradient often becomes nearly orthogonal to the future full-train gradient. Under Adam, the behavior is dataset dependent. EuroSAT keeps stronger short-lag full-data alignment, while DTD text becomes more negative.}
    \label{fig:adam_fail_exp7_futurefull}
\end{figure*}

\begin{figure*}
    \centering
    \includegraphics[width=0.98\linewidth]{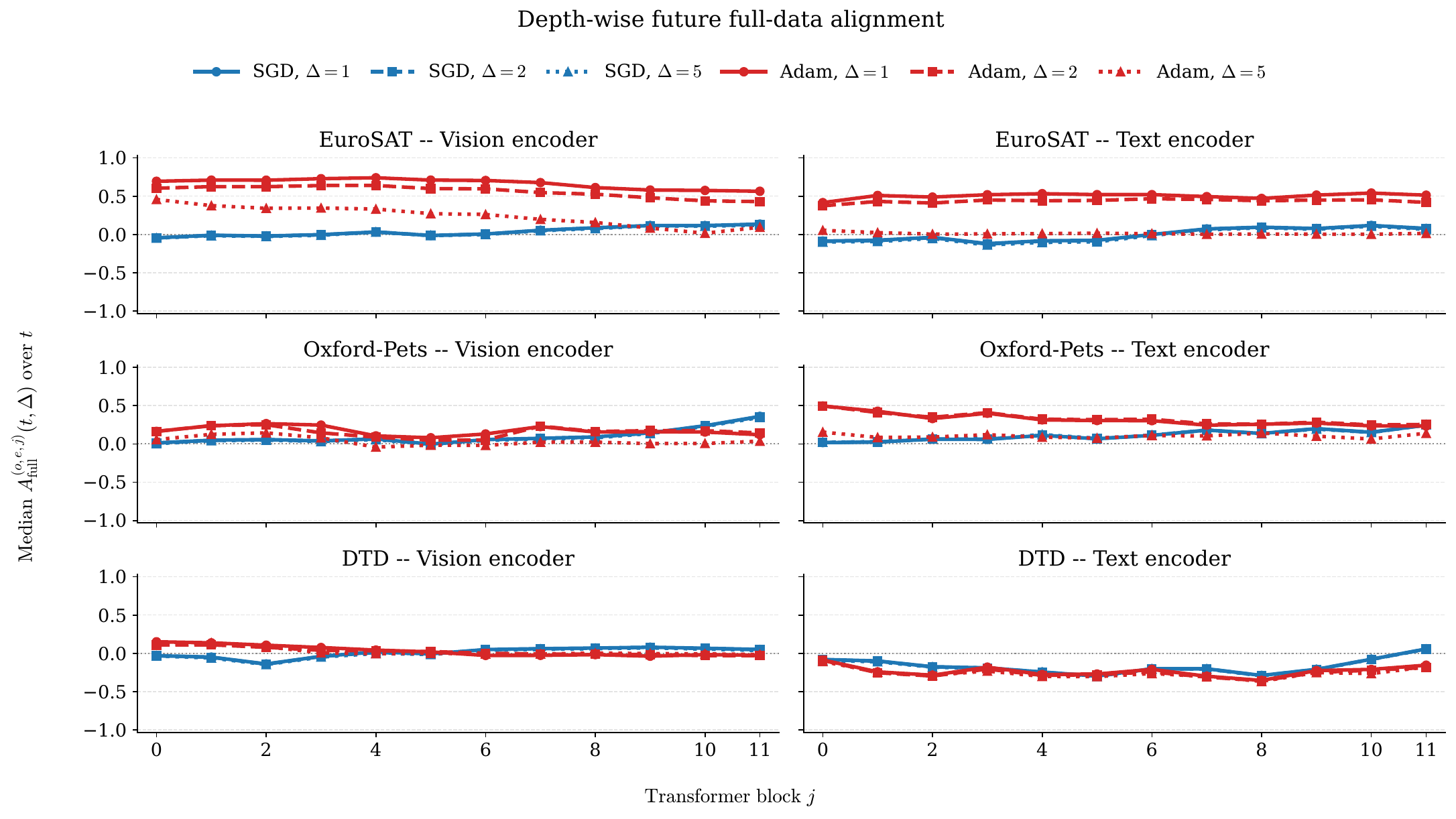}
    \caption{Experiment~8: blockwise future full-data alignment. The plotted quantity is the median over $t$ of $A_{\mathrm{full}}^{(o,e,j)}(t,\Delta)$ from Eq.~\eqref{eq:exp7_block_full_alignment}. EuroSAT shows strong positive Adam alignment across most blocks, while SGD remains close to zero. Oxford-Pets is mixed. DTD text is negative across many blocks, and Adam is usually more negative than SGD.}
    \label{fig:adam_fail_exp7_blocks}
\end{figure*}

\begin{figure*}
    \centering
    \includegraphics[width=0.98\linewidth]{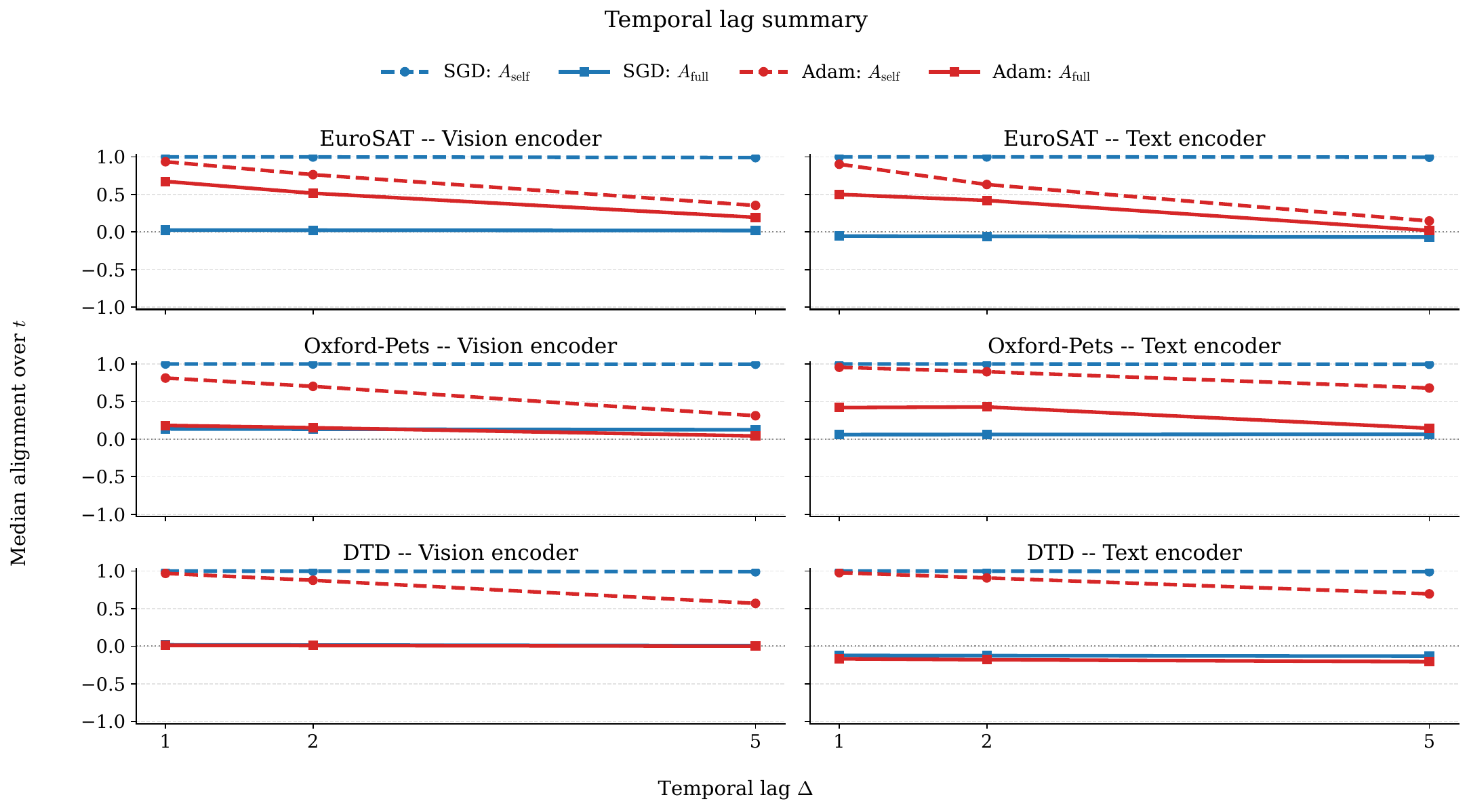}
    \caption{Experiment~8: median lag summary. Each point is the median over anchor steps $t$ of either $A_{\mathrm{self}}^{(o,e)}(t,\Delta)$ or $A_{\mathrm{full}}^{(o,e)}(t,\Delta)$. SGD has very high self-stability at all lags, but weak future full-data alignment. Adam has lower self-stability. Its future full-data alignment is better than SGD on EuroSAT, mixed on Oxford-Pets, and worse on DTD text.}
    \label{fig:adam_fail_exp7_lags}
\end{figure*}

For a block-level view, let $d_{e,j}$ be the number of trainable scalar parameters in Transformer block $j$ of encoder $e$. Let $m_{\mathrm{few},t}^{(o,e,j)}\in\mathbb{R}^{d_{e,j}}$ and $m_{\mathrm{full},t}^{(o,e,j)}\in\mathbb{R}^{d_{e,j}}$ be the corresponding block-restricted gradients. We define
\begin{equation}
    A_{\mathrm{full}}^{(o,e,j)}(t,\Delta)
    = \cos\!\left(m_{\mathrm{few},t}^{(o,e,j)},
                  m_{\mathrm{full},t+\Delta}^{(o,e,j)}\right)
    \in[-1,1].
    \label{eq:exp7_block_full_alignment}
\end{equation}
Figure~\ref{fig:adam_fail_exp7_blocks} reports the median of this quantity over $t$ for each block.

Figures~\ref{fig:adam_fail_exp7_self} and~\ref{fig:adam_fail_exp7_lags} show that the few-shot gradient direction under SGD quickly becomes nearly time-invariant. Averaged over anchor steps, SGD gives $A_{\mathrm{self}}^{(o,e)}(t,\Delta)$ values of $(0.973,0.966,0.941)$ on EuroSAT vision and $(0.981,0.971,0.943)$ on EuroSAT text for $\Delta\in\{1,2,5\}$. The same values are even higher on Oxford-Pets and DTD at short lags. At late anchors $t\ge20$, the SGD self-stability is essentially one for all datasets, encoders, and lags. Thus, SGD enters a regime where the support-gradient direction is almost fixed.

Figure~\ref{fig:adam_fail_exp7_futurefull} shows that this directionally stable gradient is not necessarily a good proxy for the future full-train objective. On EuroSAT under SGD, the all-time mean of $A_{\mathrm{full}}^{(o,e)}(t,\Delta)$ is only $(0.046,0.042,0.035)$ in vision and $(0.016,0.006,-0.016)$ in text. On DTD text, it is negative at every lag: $(-0.045,-0.064,-0.099)$. Oxford-Pets is milder, with values around $0.13$--$0.16$. Therefore, the support-gradient direction under SGD is temporally stable but often stale as a full-data surrogate.

Adam changes the self-stability picture. For $A_{\mathrm{self}}^{(o,e)}(t,\Delta)$, Adam is much lower than SGD. On EuroSAT, the all-time means are $(0.670,0.627,0.414)$ in vision and $(0.700,0.544,0.234)$ in text. On Oxford-Pets, they are $(0.483,0.538,0.313)$ in vision and $(0.603,0.651,0.446)$ in text. The time-series curves in Figure~\ref{fig:adam_fail_exp7_self} show oscillations and sign changes. In this directional sense, Adam does make the few-shot gradient temporally unstable.

However, Figures~\ref{fig:adam_fail_exp7_futurefull} and~\ref{fig:adam_fail_exp7_lags} show that this instability is not the universal cause of Adam collapse. If stale gradients were the primary explanation, Adam would consistently have worse $A_{\mathrm{full}}^{(o,e)}(t,\Delta)$ than SGD. It does not. On EuroSAT, Adam has much higher future full-data alignment than SGD. The Adam means are $(0.463,0.437,0.238)$ in vision and $(0.367,0.260,0.018)$ in text. Oxford-Pets is mixed. Adam is weaker than SGD in the vision encoder, but stronger in the text encoder at $\Delta=1$ and $\Delta=2$. DTD is the only dataset where Adam is clearly worse in the text encoder, with means $(-0.164,-0.172,-0.173)$.

The blockwise results in Figure~\ref{fig:adam_fail_exp7_blocks} refine this conclusion. On EuroSAT, Adam maintains a positive future full-data alignment across most vision and text blocks, especially for $\Delta=1$ and $\Delta=2$, whereas SGD remains close to zero. On Oxford-Pets, the Adam text blocks are positive for short lags but become weak at $\Delta=5$. The vision blocks are closer to SGD and are sometimes weaker. On DTD, the text blocks are the clearest failure case. Many blocks are negative for both optimizers, and Adam is usually more negative than SGD. Thus, temporal gradient staleness is block- and dataset-dependent.

The loss trajectories explain why positive future alignment under Adam should not be interpreted as successful optimization. At the final logged checkpoint, Adam has $L_{\mathrm{full}}\approx2.30$ on EuroSAT and $L_{\mathrm{full}}\approx3.64$ on Oxford-Pets, whereas the corresponding SGD losses are about $1.03$ and $0.35$. On DTD, Adam drives $L_{\mathrm{few}}$ to $7.6\times10^{-5}$, but $L_{\mathrm{full}}$ rises to $6.32$. These outcomes occur despite the fact that Adam can have better future full-data gradient alignment than SGD on some datasets. This rules out a gradient-staleness-only explanation.

\begin{tcolorbox}[
  enhanced,
  unbreakable,
  colback=findingblueback,
  colframe=findingblueframe,
  colbacktitle=findingblueframe,
  coltitle=white,
  boxrule=0.7pt,
  arc=1.5pt,
  left=6pt,
  right=6pt,
  top=5pt,
  bottom=5pt,
  before skip=7pt,
  after skip=7pt,
  fonttitle=\bfseries\small,
  fontupper=\small,
  title={Conclusion}]
Under SGD, the few-shot gradient direction becomes nearly time-invariant. This stable direction can still become stale relative to the future full-train objective, especially in the text encoder and on DTD. Under Adam, the few-shot gradient direction is much less temporally stable, but this does not consistently reduce future full-data alignment. On EuroSAT, Adam often has stronger future full-data alignment than SGD and still fails. Therefore, temporal gradient staleness is real, but it is not the primary universal cause of Adam collapse. It is a downstream, dataset-dependent effect that becomes important after the denominator-scale failure has already moved the model into a bad region.
\end{tcolorbox}

\subsection{Do richer preconditioning optimizers fail?}
\label{subsec:richer_second_order_optimizers_fail}

Experiment~9 studies whether Adam's failure is only a diagonal-preconditioning artifact. We compare SGD, Adam, Shampoo, and SOAP under the same one-shot CLIP adaptation protocol used in this section. Adam uses the hyperparameters $\beta_1=0.9$ and $\beta_2=0.999$, Shampoo uses a preconditioning frequency $1$, and SOAP uses a preconditioning frequency $10$. These are the default settings used in the optimizer implementations. The goal is to test whether richer preconditioned optimizers avoid the few-shot failure observed for Adam.

\begin{figure*}[t]
    \centering
    \includegraphics[width=0.98\linewidth]{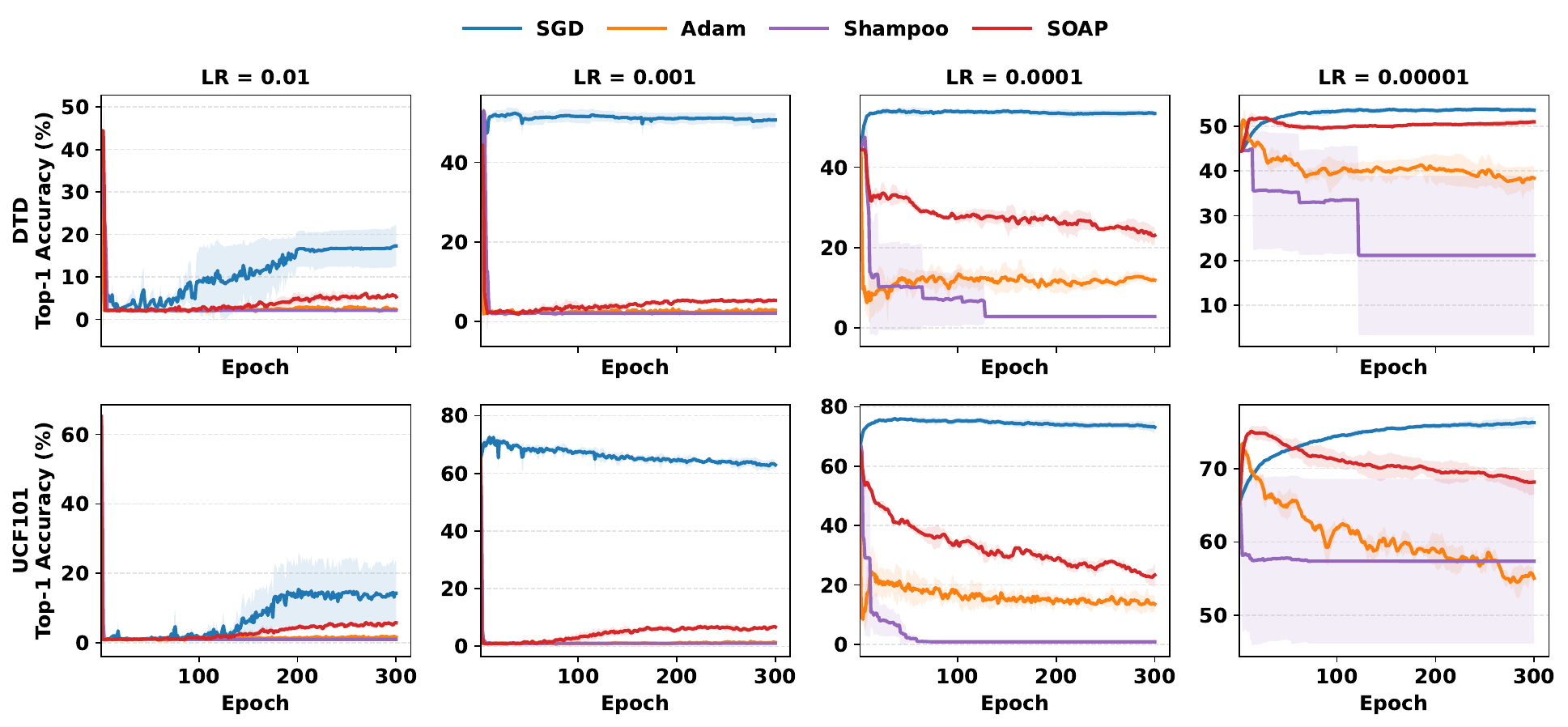}
    \caption{Experiment~9: test accuracy over training epochs for SGD, Adam ($\beta_2=0.999$), Shampoo (preconditioning frequency $1$), and SOAP (preconditioning frequency $10$) on DTD and UCF101.}
    \label{fig:adam_fail_exp9_training_curves}
\end{figure*}

Figure~\ref{fig:adam_fail_exp9_training_curves} shows that richer preconditioning does not rescue few-shot full fine-tuning. Across the learning-rate grid, SGD is the only method that remains consistently stable and competitive. Adam collapses or stays far below SGD. Shampoo and SOAP follow the same qualitative pattern. At large learning rates, their trajectories are unstable. At small learning rates, the collapse is damped, but neither method recovers the robust behavior of SGD. Thus, the failure is not specific to Adam's coordinatewise denominator. It also appears when the optimizer is replaced by other structured preconditioned methods. %

We next diagnose why this happens. Let $\mathcal{O}\in\{\mathrm{Shampoo},\mathrm{SOAP}\}$ denote a structured preconditioner family. For encoder $e\in\{\mathrm{vision},\mathrm{text}\}$, let $d_e$ be the number of trainable scalar parameters in that encoder. Let
\[
    B_{\mathcal{O},\mathcal{D},t}^{(e)}:\mathbb{R}^{d_e}\to\mathbb{R}^{d_e}
\]
be the effective linear preconditioning map using statistics from split $\mathcal{D}\in\{\mathcal{D}_{\mathrm{few}},\mathcal{D}_{\mathrm{full}}\}$ at checkpoint $\Theta_t$. For Shampoo, this map is induced by the tensor-factor inverse roots. For the SOAP diagnostic, both splits use a shared eigenbasis estimated from the full training split at each checkpoint. The maps differ only in their Adam-like second moments within that basis. Their comparison therefore isolates second-moment mismatch in a common basis, rather than mismatch between separately estimated eigenbases. We use this notation only as a diagnostic abstraction. The experiment applies the map to gradients without forming the full $d_e\times d_e$ matrix.

The two-letter source convention is unchanged. The first letter denotes the numerator source. The second letter denotes the source of the preconditioning statistic. The SOAP eigenbasis remains shared. Thus,
\begin{equation}
    q_{ff,t}^{(\mathcal{O},e)}
    = B_{\mathcal{O},\mathcal{D}_{\mathrm{few}},t}^{(e)}m_{\mathrm{few},t}^{(e)},
    \qquad
    q_{fr,t}^{(\mathcal{O},e)}
    = B_{\mathcal{O},\mathcal{D}_{\mathrm{full}},t}^{(e)}m_{\mathrm{few},t}^{(e)},
    \qquad
    q_{rr,t}^{(\mathcal{O},e)}
    = B_{\mathcal{O},\mathcal{D}_{\mathrm{full}},t}^{(e)}m_{\mathrm{full},t}^{(e)} .
    \label{eq:exp9_structured_directions}
\end{equation}
All three vectors belong to $\mathbb{R}^{d_e}$. The direction $q_{ff,t}^{(\mathcal{O},e)}$ is the support-preconditioned few-shot update. The direction $q_{fr,t}^{(\mathcal{O},e)}$ keeps the same few-shot numerator but replaces the preconditioner by the full-train diagnostic preconditioner. The direction $q_{rr,t}^{(\mathcal{O},e)}$ is the full-train reference direction under the same full-train preconditioner.

We summarize these directions with two encoder-level diagnostics:
\begin{equation}
    S_{\mathcal{O}}^{(e)}(t)
    = \frac{\left\lVert q_{ff,t}^{(\mathcal{O},e)}\right\rVert_2}
           {\left\lVert q_{fr,t}^{(\mathcal{O},e)}\right\rVert_2},
    \qquad
    A_{\mathcal{O}}^{(e)}(t)
    = \cos\!\left(q_{fr,t}^{(\mathcal{O},e)},q_{rr,t}^{(\mathcal{O},e)}\right).
    \label{eq:exp9_structured_diagnostics}
\end{equation}
The ratio $S_{\mathcal{O}}^{(e)}(t)\in\mathbb{R}_{\ge0}$ is dimensionless. It isolates preconditioner-scale error because the numerator is fixed. The cosine $A_{\mathcal{O}}^{(e)}(t)\in[-1,1]$ measures the remaining numerator mismatch under the same full-train diagnostic preconditioner. A value near zero means that the repaired few-shot direction is nearly orthogonal to the full-data reference.

\begin{figure*}[t]
    \centering
    \includegraphics[width=0.98\linewidth]{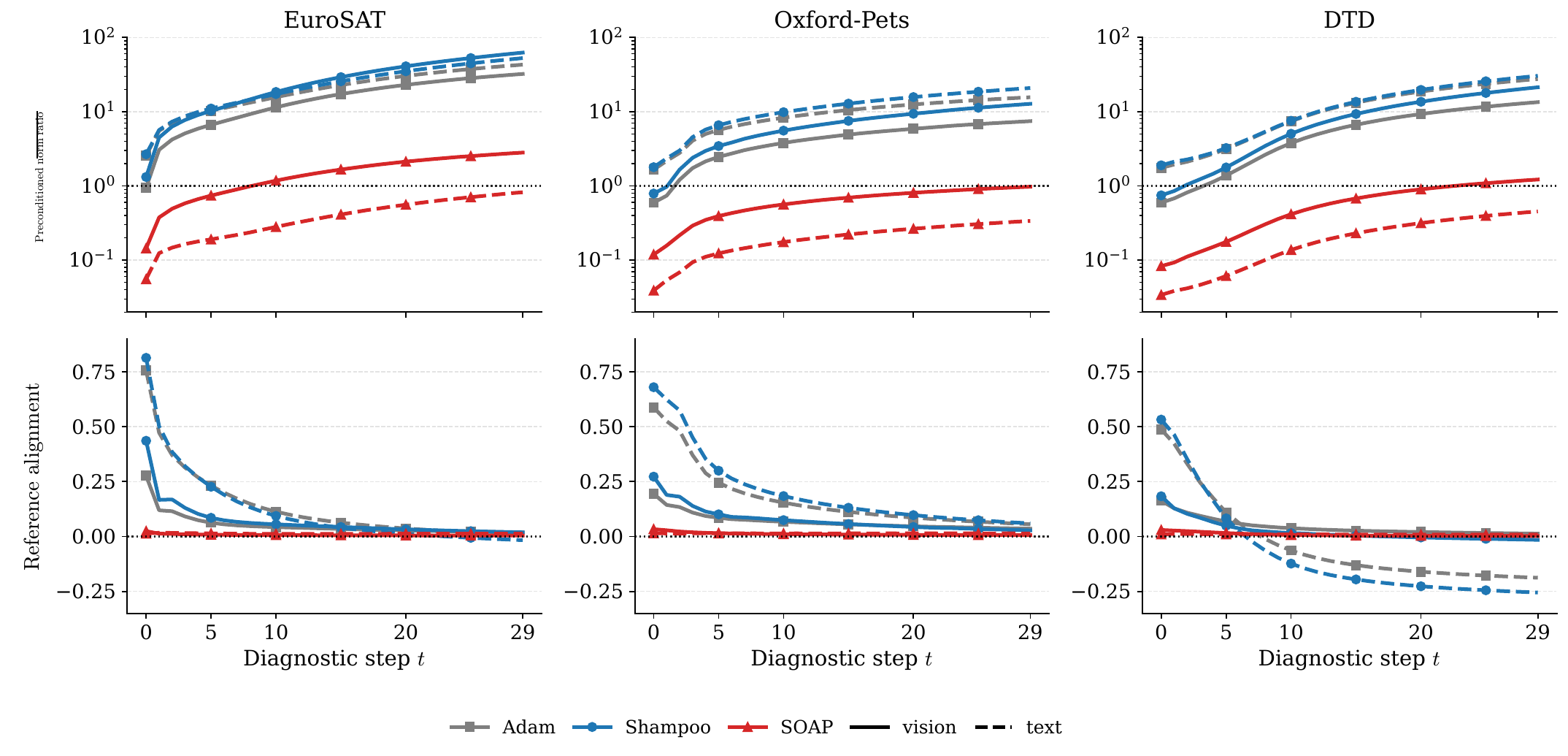}
    \caption{Experiment~9, preconditioner scale and reference alignment over the stable SGD trajectory. The top row reports the scale ratio $S_{\mathcal{O}}^{(e)}(t)$ for Shampoo and SOAP, with the analogous ratio $\lVert u_{ff,t}^{(e)}\rVert_2/\lVert u_{fr,t}^{(e)}\rVert_2$ for Adam. The bottom row reports $A_{\mathcal{O}}^{(e)}(t)$ and the analogous Adam cosine $\cos(u_{fr,t}^{(e)},u_{rr,t}^{(e)})$, which measures the remaining numerator mismatch under the same full-data diagnostic preconditioner.}
    \label{fig:adam_fail_exp9_preconditioner_scale_alignment}
\end{figure*}

Figure~\ref{fig:adam_fail_exp9_preconditioner_scale_alignment} separates scale error from residual geometric error. Adam and Shampoo show large scale inflation. At the final checkpoint, the Adam scale ratio ranges from $7.45$ to $42.97$, and the Shampoo ratio ranges from $12.76$ to $62.47$ across the three diagnostic datasets and the two encoders. SOAP behaves differently. Its final ratio ranges from $0.34$ to $2.81$. Therefore, SOAP does not exhibit the same simple scale-inflation signature as Adam. Some SOAP directions are even damped relative to their full-data-preconditioned counterpart.

The bottom row of Figure~\ref{fig:adam_fail_exp9_preconditioner_scale_alignment} shows why this scale distinction is not enough. SOAP still has almost no reference alignment at the final checkpoint, with $A_{\mathrm{SOAP}}^{(e)}(t)$ between $0.002$ and $0.014$. Shampoo also loses alignment, with $A_{\mathrm{Shampoo}}^{(e)}(t)$ between $-0.255$ and $0.060$. Hence, the full-data diagnostic preconditioner cannot turn the few-shot numerator into a reliable full-data preconditioned direction. SOAP therefore retains substantial numerator mismatch under the full-data preconditioner, even when it avoids the scale inflation observed for Adam. This cosine alone does not identify an error in the support-estimated preconditioning statistic.

To quantify the statistic itself, let $z_{\mathcal{O},\mathcal{D},t}^{(e)}\in\mathbb{R}_{\ge0}^{p_{\mathcal{O},e}}$ denote the flattened nonnegative statistic used by the diagnostic to construct the preconditioner family $\mathcal{O}$ in encoder $e$. For SOAP, $z_{\mathrm{SOAP},\mathcal{D},t}^{(e)}$ contains the diagonal second moments of gradients projected into the shared full-data eigenbasis. The dimension $p_{\mathcal{O},e}$ is the number of scalar diagnostic entries in that structured statistic. It need not equal $d_e$. We define the coordinate mismatch
\begin{equation}
    M_{\mathcal{O}}^{(e)}(t)
    = \frac{1}{p_{\mathcal{O},e}}
      \sum_{i=1}^{p_{\mathcal{O},e}}
      \left|\log_{10}\!\left(
      \frac{z_{\mathcal{O},\mathcal{D}_{\mathrm{few}},t,i}^{(e)}+\varepsilon}
           {z_{\mathcal{O},\mathcal{D}_{\mathrm{full}},t,i}^{(e)}+\varepsilon}
      \right)\right| .
    \label{eq:exp9_structured_mismatch}
\end{equation}
We also measure top-coordinate overlap using the operator $\operatorname{top}_k$ defined above. Here $k=\lceil\rho_{\mathrm{top}}p_{\mathcal{O},e}\rceil$ for density $\rho_{\mathrm{top}}\in(0,1]$. The overlap is
\begin{equation}
    O_{\mathcal{O}}^{(e)}(\rho_{\mathrm{top}};t)
    = \frac{\left|\operatorname{top}_k\!\left(z_{\mathcal{O},\mathcal{D}_{\mathrm{few}},t}^{(e)}\right)
      \cap
      \operatorname{top}_k\!\left(z_{\mathcal{O},\mathcal{D}_{\mathrm{full}},t}^{(e)}\right)\right|}
      {k}
    \in[0,1].
    \label{eq:exp9_structured_overlap}
\end{equation}
We use $\rho_{\mathrm{top}}=0.01$ in the figure. The random-overlap baseline is $0.01$.

\begin{figure*}[t]
    \centering
    \includegraphics[width=0.98\linewidth]{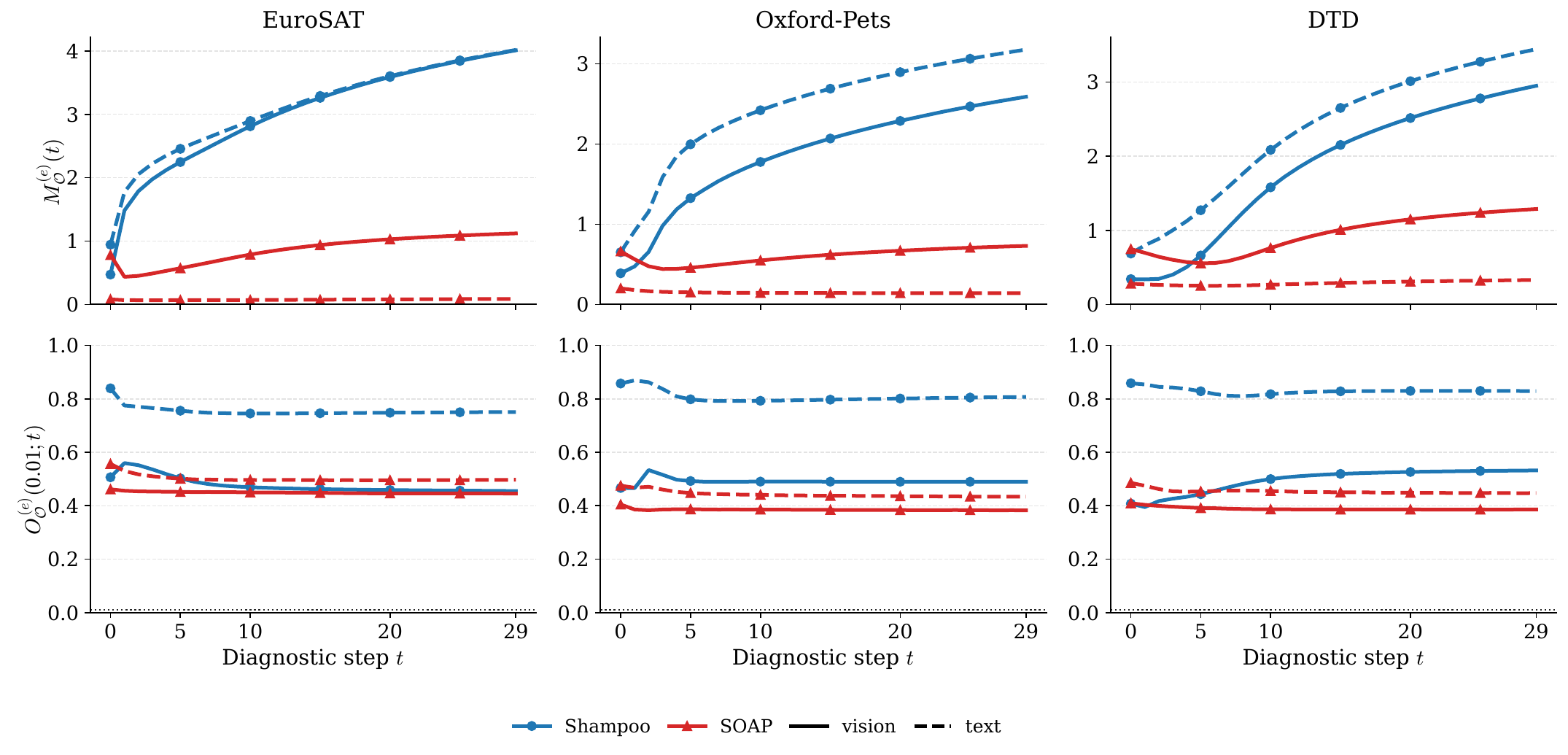}
    \caption{Experiment~9, structured statistic mismatch and top-coordinate overlap for Shampoo and SOAP. The top row reports $M_{\mathcal{O}}^{(e)}(t)$ from Eq.~\eqref{eq:exp9_structured_mismatch}. The bottom row reports $O_{\mathcal{O}}^{(e)}(0.01;t)$ from Eq.~\eqref{eq:exp9_structured_overlap}. The dotted line is the random-overlap baseline. A high overlap means that the support statistic finds some of the same salient coordinates. It does not imply a safe preconditioned update.}
    \label{fig:adam_fail_exp9_structured_mismatch_overlap}
\end{figure*}

Figure~\ref{fig:adam_fail_exp9_structured_mismatch_overlap} shows that the structured statistics are not random. Shampoo has substantial top-coordinate overlap with the full-data statistic. At the final checkpoint, its overlap lies between $0.46$ and $0.83$, which is far above the random baseline. SOAP also remains above random, with overlap between $0.38$ and $0.50$. However, this overlap does not rescue the update. The scale and alignment diagnostics in Figure~\ref{fig:adam_fail_exp9_preconditioner_scale_alignment} still fail. The statistic mismatch and scale ratio measure differences between support and full-data preconditioning statistics. The repaired-direction cosine measures the separate numerator mismatch under a common full-data map.

The depth-wise view confirms that the failure is structured. Let $P_{e,j}:\mathbb{R}^{d_e}\to\mathbb{R}^{d_{e,j}}$ denote the projection onto Transformer block $j$ of encoder $e$, where $d_{e,j}$ is the number of trainable scalar parameters in that block. The restricted directions are
\[
    q_{fr,t}^{(\mathcal{O},e,j)}=P_{e,j}q_{fr,t}^{(\mathcal{O},e)},
    \qquad
    q_{rr,t}^{(\mathcal{O},e,j)}=P_{e,j}q_{rr,t}^{(\mathcal{O},e)}.
\]
Both vectors lie in $\mathbb{R}^{d_{e,j}}$.
We then measure the blockwise repaired-direction alignment $\cos(q_{fr,T}^{(\mathcal{O},e,j)},q_{rr,T}^{(\mathcal{O},e,j)})$ at the final diagnostic checkpoint $T=29$.

\begin{figure*}[t]
    \centering
    \includegraphics[width=0.98\linewidth]{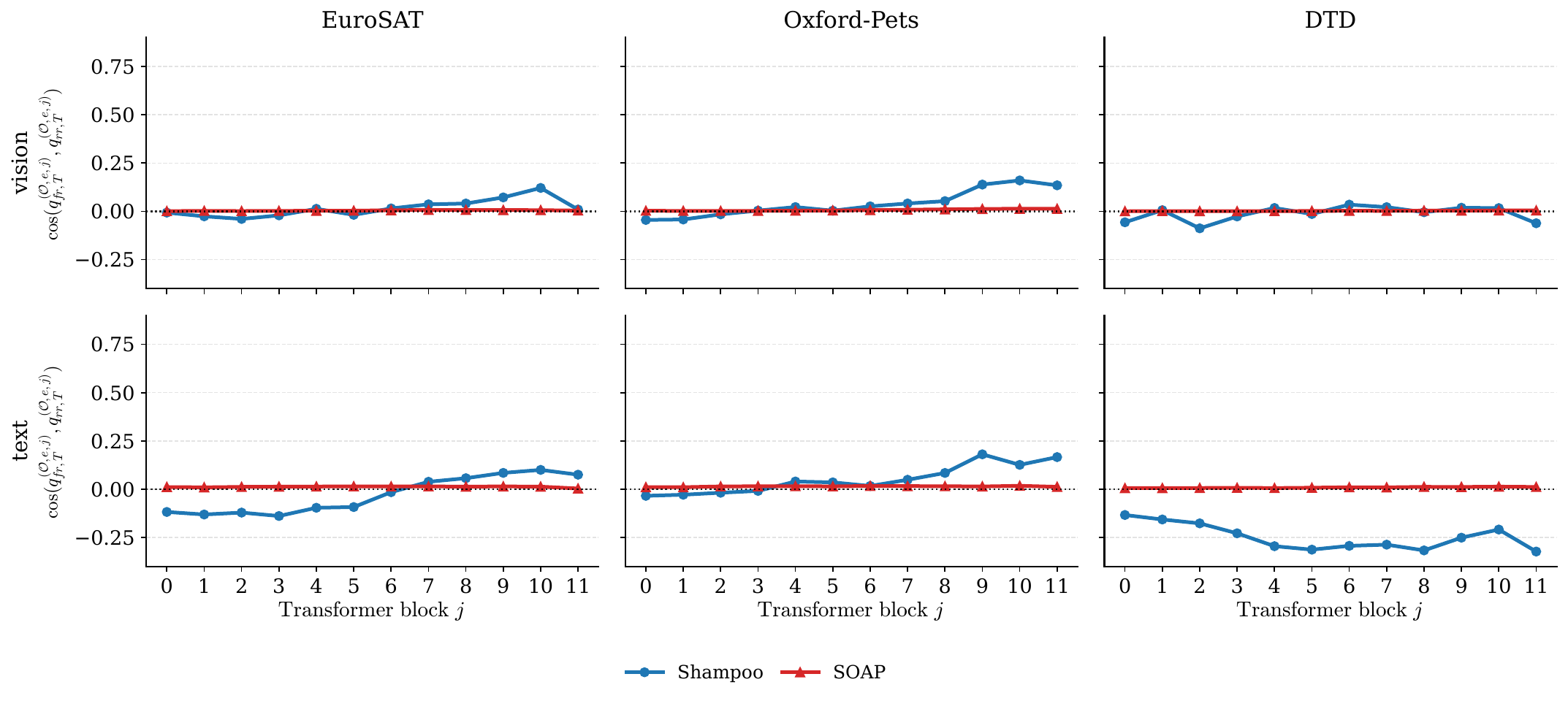}
    \caption{Experiment~9, final blockwise alignment for Shampoo and SOAP. Each curve reports $\cos(q_{fr,T}^{(\mathcal{O},e,j)},q_{rr,T}^{(\mathcal{O},e,j)})$ at the final diagnostic checkpoint $T=29$. The diagnostic uses the full-data preconditioner in both directions. Low values therefore indicate residual few-shot/full-data mismatch that cannot be repaired by replacing only the preconditioner.}
    \label{fig:adam_fail_exp9_block_alignment}
\end{figure*}

Figure~\ref{fig:adam_fail_exp9_block_alignment} shows that the residual error is not uniform across depth. Some blocks maintain weak positive alignment, but many blocks are near zero or negative. DTD text is the clearest case for Shampoo. SOAP is more uniformly close to zero. Both directions use the same full-data preconditioner, so these blockwise cosines measure residual numerator mismatch. They do not identify estimation error in the support preconditioner.

This interpretation is consistent with the experiment in Subsection~\ref{subsec:sgd_support_full_geometry_alignment}. There, the support and full-train objectives become misaligned in gradient direction, curvature scale, and local surface shape along the SGD trajectory. The structured diagnostics distinguish preconditioning-statistic mismatch from numerator mismatch. SOAP retains the latter under the full-data map despite avoiding the largest scale inflation. Shampoo shows both large scale distortion and poor repaired-direction alignment. These observations do not isolate a single cause of the structured optimizers' performance.

\begin{findingbox}[Conclusion]
Richer structured preconditioners do not solve the few-shot CLIP failure. SOAP avoids Adam's diagonal scale-inflation signature in several cases, but its repaired preconditioned direction is nearly orthogonal to the full-data reference. Shampoo preserves substantial saliency structure, yet it develops large scale distortion and poor full-data alignment. The failure is therefore not merely a diagonal calibration artifact. The diagnostics show preconditioning-statistic mismatch and residual numerator mismatch under a common full-data preconditioner.
\end{findingbox}

\section{Theoretical Analysis}
\label{sec:adam_under_sampled_diagonal_preconditioner}

\subsection{Finite-sample error in diagonal second-moment estimation}

We first formalize the statistical object estimated by Adam's denominator. The checkpoint $\Theta_t\in\mathbb{R}^d$ and encoder $e\in\{\mathrm{vision},\mathrm{text}\}$ are fixed. Let $\Theta_t^{(e)} \in \mathbb{R}^{d_e}$ be the trainable parameter vector of encoder $e$, where $d_e$ is the number of trainable scalar parameters in that encoder. Let $\mathcal{D}_{\mathrm{full}}=\{z_1,\ldots,z_N\}$ be the full training split, with $z_i=(x_i,y_i)$ and $N=|\mathcal{D}_{\mathrm{full}}|$. Let $\mathcal{D}_{\mathrm{few}} \subset \mathcal{D}_{\mathrm{full}}$ be the few-shot support set with $n=|\mathcal{D}_{\mathrm{few}}|=KC$ examples, where $K$ is the number of shots and $C$ is the number of
classes. We assume the challenging case, $K=1$. For each example $z_i$, the per-example encoder gradient is
\begin{equation}
    a_{i,t}^{(e)} = \nabla_{\Theta^{(e)}} \ell(z_i;\Theta_t)
    \in \mathbb{R}^{d_e},
\end{equation}
where $\ell(z_i;\Theta_t)$ is the cross-entropy loss of example $z_i$. For coordinate $j \in \{1,\ldots,d_e\}$, the nonnegative squared-gradient value is
\begin{equation}
    X_{i,t,j}^{(e)} = \left(a_{i,t,j}^{(e)}\right)^2 \in \mathbb{R}_{\ge 0}.
\end{equation}
The full-data diagonal second moment in coordinate $j$ is the finite-population mean
\begin{equation}
    v_{\mathrm{full},t,j}^{(e)}
    = \frac{1}{N}\sum_{i=1}^{N} X_{i,t,j}^{(e)} .
    \label{eq:full_diagonal_second_moment_theory}
\end{equation}
The few-shot statistic is the same mean computed on the support indices
$I_{\mathrm{few}} \subset \{1,\ldots,N\}$:
\begin{equation}
    \widehat v_{\mathrm{few},t,j}^{(e)}
    = \frac{1}{n}\sum_{i\in I_{\mathrm{few}}} X_{i,t,j}^{(e)} .
    \label{eq:few_diagonal_second_moment_theory}
\end{equation}
The hat indicates that the support set is viewed as a random sample from the full training split. After the support set is fixed, this quantity is the empirical $v_{\mathrm{few},t,j}^{(e)}$ used by the Adam-like denominator. Thus, Adam must estimate a vector in $\mathbb{R}_{\ge 0}^{d_e}$ from only $n$ examples. This is the source of the under-sampling problem.

The probability calculations hold at a checkpoint $\Theta_t$ fixed independently of the sampled support. The gradient values remain fixed. The support indices $I_{\mathrm{few}}$ are drawn uniformly without replacement from $\{1,\ldots,N\}$. This uniform-subset model provides an idealized analysis and differs from the class-balanced support sampling used in the experiments. Expectations are taken over this draw, with $1\le n\le N$ and $N>1$. For each coordinate, the finite-population variance is
\begin{equation}
    \sigma_{t,j}^{2,(e)}
    = \frac{1}{N-1}\sum_{i=1}^{N}
    \left(X_{i,t,j}^{(e)} - v_{\mathrm{full},t,j}^{(e)}\right)^2 .
    \label{eq:finite_population_variance_theory}
\end{equation}
Under these assumptions, the estimator is unbiased. Its total mean squared error depends on the support size and the effective diagonal variability.

\paragraph{Proposition 1 (finite-population error of the few-shot diagonal estimator).}
For fixed $t$ and $e$, under the sampling and checkpoint assumptions above,
\begin{equation}
    \mathbb{E}\left[\widehat v_{\mathrm{few},t,j}^{(e)}\right]
    = v_{\mathrm{full},t,j}^{(e)}
    \quad \text{for every } j,
    \label{eq:unbiased_second_moment_theory}
\end{equation}
and
\begin{equation}
    \operatorname{Var}\left(\widehat v_{\mathrm{few},t,j}^{(e)}\right)
    = \frac{N-n}{Nn}\,\sigma_{t,j}^{2,(e)} .
    \label{eq:variance_second_moment_theory}
\end{equation}
Consequently,
\begin{equation}
    \mathbb{E}\left[
    \left\|\widehat v_{\mathrm{few},t}^{(e)} - v_{\mathrm{full},t}^{(e)}\right\|_2^2
    \right]
    = \frac{N-n}{Nn}\sum_{j=1}^{d_e} \sigma_{t,j}^{2,(e)} .
    \label{eq:vector_mse_second_moment_theory}
\end{equation}
For Adam's stabilized denominator, the dimensionless effective variability is
\begin{equation}
    \mathcal{U}_t^{(e)}
    = \sum_{j=1}^{d_e}
    \frac{\sigma_{t,j}^{2,(e)}}{\left(v_{\mathrm{full},t,j}^{(e)}+\varepsilon\right)^2},
    \qquad \mathcal{U}_t^{(e)} \in \mathbb{R}_{\ge 0} .
    \label{eq:effective_variability_theory}
\end{equation}
Then
\begin{equation}
    \mathbb{E}\left[
    \sum_{j=1}^{d_e}
    \left(
    \frac{\widehat v_{\mathrm{few},t,j}^{(e)}-v_{\mathrm{full},t,j}^{(e)}}
    {v_{\mathrm{full},t,j}^{(e)}+\varepsilon}
    \right)^2
    \right]
    = \frac{N-n}{Nn}\,\mathcal{U}_t^{(e)} .
    \label{eq:relative_mse_second_moment_theory}
\end{equation}

\emph{Proof.}
Let $S_i$ be the indicator that example $i$ is included in $I_{\mathrm{few}}$. Because $I_{\mathrm{few}}$ is a uniform subset of size $n$, we have
\begin{equation}
    \mathbb{E}[S_i]=\frac{n}{N},
    \qquad
    \operatorname{Var}(S_i)=\frac{n}{N}\left(1-\frac{n}{N}\right),
\end{equation}
and, for $i\ne k$,
\begin{equation}
    \operatorname{Cov}(S_i,S_k)
    = \mathbb{E}[S_iS_k]-\mathbb{E}[S_i]\mathbb{E}[S_k]
    = \frac{n(n-1)}{N(N-1)}-\frac{n^2}{N^2}
    = -\frac{n(N-n)}{N^2(N-1)} .
\end{equation}
The proof abbreviates $X_i=X_{i,t,j}^{(e)}$ and
$v_j=v_{\mathrm{full},t,j}^{(e)}$. The estimator in Eq.~\eqref{eq:few_diagonal_second_moment_theory}
can be written as
\begin{equation}
    \widehat v_j = \frac{1}{n}\sum_{i=1}^{N} S_i X_i .
\end{equation}
Its expectation is
\begin{equation}
    \mathbb{E}[\widehat v_j]
    = \frac{1}{n}\sum_{i=1}^{N} \mathbb{E}[S_i]X_i
    = \frac{1}{n}\cdot \frac{n}{N}\sum_{i=1}^{N} X_i
    = v_j .
\end{equation}
This proves Eq.~\eqref{eq:unbiased_second_moment_theory}. For the variance,
\begin{align}
    \operatorname{Var}(\widehat v_j)
    &= \frac{1}{n^2}\operatorname{Var}\left(\sum_{i=1}^{N} S_iX_i\right) \\
    &= \frac{1}{n^2}
    \left(
    \sum_{i=1}^{N} X_i^2\operatorname{Var}(S_i)
    +2\sum_{1\le i<k\le N} X_iX_k\operatorname{Cov}(S_i,S_k)
    \right) .
\end{align}
Using the expressions above for the variance and covariance of the indicators gives
\begin{align}
    \operatorname{Var}(\widehat v_j)
    &= \frac{1}{n^2}
    \left(
    \frac{n(N-n)}{N^2}\sum_{i=1}^{N}X_i^2
    -\frac{2n(N-n)}{N^2(N-1)}\sum_{1\le i<k\le N}X_iX_k
    \right) \\
    &= \frac{N-n}{nN^2}
    \left(
    \sum_{i=1}^{N}X_i^2
    -\frac{2}{N-1}\sum_{1\le i<k\le N}X_iX_k
    \right) .
\end{align}
Since
\begin{equation}
    \sum_{i=1}^{N}\left(X_i-v_j\right)^2
    = \sum_{i=1}^{N}X_i^2 - Nv_j^2
    = \frac{N-1}{N}\sum_{i=1}^{N}X_i^2
    -\frac{2}{N}\sum_{1\le i<k\le N}X_iX_k,
\end{equation}
we have
\begin{equation}
    \sum_{i=1}^{N}X_i^2
    -\frac{2}{N-1}\sum_{1\le i<k\le N}X_iX_k
    = \frac{N}{N-1}\sum_{i=1}^{N}\left(X_i-v_j\right)^2
    = N\sigma_{t,j}^{2,(e)} .
\end{equation}
Substituting this identity into the variance expression yields
\begin{equation}
    \operatorname{Var}(\widehat v_j)
    = \frac{N-n}{Nn}\,\sigma_{t,j}^{2,(e)} .
\end{equation}
This proves Eq.~\eqref{eq:variance_second_moment_theory}. Because the estimator is unbiased, $\mathbb{E}[(\widehat v_j-v_j)^2]=\operatorname{Var}(\widehat v_j)$. Summing this identity over $j=1,\ldots,d_e$ proves Eq.~\eqref{eq:vector_mse_second_moment_theory}. Dividing each coordinate by $v_{\mathrm{full},t,j}^{(e)}+\varepsilon$ and summing proves
Eq.~\eqref{eq:relative_mse_second_moment_theory}. \hfill $\square$

Proposition~1 shows that an unbiased few-shot diagonal moment can have substantial total mean squared relative error. The expected squared relative error in Eq.~\eqref{eq:relative_mse_second_moment_theory} scales as $\mathcal{U}_t^{(e)}/n$, up to the finite-population factor $(N-n)/N$. Achieving a small total mean squared relative error therefore requires $(N-n)\mathcal{U}_t^{(e)}/(Nn)$ to be small. The identity alone does not establish frequent estimation failure or incompatibility with one-shot learning. It depends on normalized gradient variability, not parameter dimension alone. For CLIP, Adam does not need a single scalar estimate. It needs $d_e$ coordinatewise denominators. Thus, the statistical burden grows with the number of trainable coordinates that have unstable per-example squared gradients.

The same calculation also gives a high-probability diagnostic. For any relative-error tolerance
$\zeta_{\mathrm{err}}>0$,
\begin{equation}
    \mathbb{P}\left(
    \max_{1\le j\le d_e}
    \frac{|\widehat v_{\mathrm{few},t,j}^{(e)}-v_{\mathrm{full},t,j}^{(e)}|}
    {v_{\mathrm{full},t,j}^{(e)}+\varepsilon}
    \ge \zeta_{\mathrm{err}}
    \right)
    \le
    \frac{N-n}{Nn\zeta_{\mathrm{err}}^2}\,\mathcal{U}_t^{(e)} .
    \label{eq:coordinate_uniform_bound_theory}
\end{equation}
Indeed, Eq.~\eqref{eq:coordinate_uniform_bound_theory} follows from Chebyshev's inequality applied to each coordinate and a union bound over $j\in\{1,\ldots,d_e\}$. A small right-hand side guarantees a small probability of a large coordinatewise error. A large right-hand side is inconclusive: an upper bound does not establish a lower bound on the failure probability. In Adam, such coordinate errors are not averaged away after the denominator is formed. They are used directly in the elementwise preconditioner.

A final finite-population example makes the observability problem explicit. In this example, coordinate $j$ has a squared-gradient value $B_j>0$ on exactly $r_j$ examples in $\mathcal{D}_{\mathrm{full}}$, with $1\le r_j\le N$. It is zero on the remaining $N-r_j$ examples. Then
\begin{equation}
    v_{\mathrm{full},t,j}^{(e)} = \frac{r_j}{N}B_j .
\end{equation}
If the support set misses those $r_j$ examples, then
$\widehat v_{\mathrm{few},t,j}^{(e)}=0$. The miss probability is
\begin{equation}
    \mathbb{P}\left(\widehat v_{\mathrm{few},t,j}^{(e)}=0\right)
    = \frac{\binom{N-r_j}{n}}{\binom{N}{n}}
    = \prod_{q=0}^{n-1}\frac{N-r_j-q}{N-q},
    \label{eq:miss_probability_theory}
\end{equation}
for $n\le N-r_j$. The miss probability is zero when $n>N-r_j$. In particular, the probability can be large when $nr_j/N$ is small. Thus, a real full-data denominator coordinate can be invisible to the support statistic. This example is not meant to claim that all Adam errors come from missed coordinates. Our empirical top-$k$ overlaps show that the support statistic often finds many salient coordinates. This example shows that a finite support can miss informative coordinates even under uniform sampling. Experiments~1--3 provide separate evidence of scale miscalibration: $v_{\mathrm{few}}$ is not arbitrary, but its scale is systematically unreliable relative to $v_{\mathrm{full}}$.

\subsection{The inverse square-root preconditioner amplifies estimation error}

The previous proposition shows that the support-set diagonal moment can have a large coordinatewise error even under a favorable sampling model. We now show why this error is especially damaging for Adam. Adam does not use the second moment itself. It uses its inverse square root. The scalar map is
\begin{equation}
\phi_\varepsilon(x) = (x + \varepsilon)^{-1/2},
\qquad x \in \mathbb{R}_{\ge 0},\quad \varepsilon > 0 .
\label{eq:inverse_sqrt_map}
\end{equation}
For fixed encoder $e \in \{\mathrm{vision},\mathrm{text}\}$, checkpoint $t$, and coordinate $j \in \{1,\ldots,d_e\}$, the ideal full-data preconditioner coefficient is $\phi_\varepsilon(v^{(e)}_{\mathrm{full},t,j})$. The few-shot coefficient is $\phi_\varepsilon(\widehat v^{(e)}_{\mathrm{few},t,j})$, where $\widehat v^{(e)}_{\mathrm{few},t,j}$ is the random support estimator defined above. The map $\phi_\varepsilon$ converts an additive denominator error into a multiplicative learning-rate error.

\paragraph{Proposition 2 (inverse-square-root bias of the preconditioner).}
For fixed $e$, $t$, and $j$, let $\widehat v^{(e)}_{\mathrm{few},t,j}$ be the random support estimator defined above, and let $v^{(e)}_{\mathrm{full},t,j}$ be the finite-population mean defined above. Under the uniform sampling model of Proposition~1,
\begin{equation}
\mathbb{E}\!
\left[
\phi_\varepsilon\!
\left(\widehat v^{(e)}_{\mathrm{few},t,j}\right)
\right]
\ge
\phi_\varepsilon\!
\left(v^{(e)}_{\mathrm{full},t,j}\right) .
\label{eq:jensen_inverse_sqrt_bias}
\end{equation}
The inequality is strict whenever $\widehat v^{(e)}_{\mathrm{few},t,j}$ has nonzero variance. The local expansion additionally assumes that, for some $\zeta_{\mathrm{loc}}\in(0,1)$,
\begin{equation}
\left|
\widehat v^{(e)}_{\mathrm{few},t,j}
-
 v^{(e)}_{\mathrm{full},t,j}
\right|
\le
\zeta_{\mathrm{loc}}\bigl(v^{(e)}_{\mathrm{full},t,j}+\varepsilon\bigr)
\quad\text{almost surely.}
\label{eq:local_inverse_sqrt_assumption}
\end{equation}
Then
\begin{equation}
\mathbb{E}\!\left[
\phi_\varepsilon\!\left(\widehat v^{(e)}_{\mathrm{few},t,j}\right)
\right]
=
\phi_\varepsilon\!\left(v^{(e)}_{\mathrm{full},t,j}\right)
+
\frac{3}{8}
\frac{N-n}{Nn}
\frac{\sigma^{2,(e)}_{t,j}}
{\bigl(v^{(e)}_{\mathrm{full},t,j}+\varepsilon\bigr)^{5/2}}
+
R^{(e)}_{t,j},
\label{eq:inverse_sqrt_bias_expansion}
\end{equation}
where the remainder satisfies
\begin{equation}
\left|R^{(e)}_{t,j}\right|
\le
\frac{5}{16}
\frac{
\mathbb{E}\!\left[
\left|
\widehat v^{(e)}_{\mathrm{few},t,j}
-
 v^{(e)}_{\mathrm{full},t,j}
\right|^3
\right]
}
{(1-\zeta_{\mathrm{loc}})^{7/2}
\bigl(v^{(e)}_{\mathrm{full},t,j}+\varepsilon\bigr)^{7/2}} .
\label{eq:inverse_sqrt_remainder_bound}
\end{equation}

\emph{Proof.}
The first two derivatives of $\phi_\varepsilon$ are
\begin{equation}
\phi_\varepsilon'(x)
=
-\frac{1}{2}(x+\varepsilon)^{-3/2},
\qquad
\phi_\varepsilon''(x)
=
\frac{3}{4}(x+\varepsilon)^{-5/2} .
\label{eq:inverse_sqrt_derivatives}
\end{equation}
Since $\phi_\varepsilon''(x)>0$ for every $x\ge 0$, the map $\phi_\varepsilon$ is strictly convex on $\mathbb{R}_{\ge 0}$. Proposition~1 gives
\begin{equation}
\mathbb{E}\!\left[
\widehat v^{(e)}_{\mathrm{few},t,j}
\right]
=
 v^{(e)}_{\mathrm{full},t,j} .
\label{eq:unbiased_for_jensen}
\end{equation}
Jensen's inequality therefore gives Eq.~\eqref{eq:jensen_inverse_sqrt_bias}. Strictness follows from strict convexity when the estimator is not almost surely constant.

It remains to prove the expansion. The following abbreviations are used in the proof:
\begin{equation}
\widehat v_j = \widehat v^{(e)}_{\mathrm{few},t,j},
\qquad
v_j = v^{(e)}_{\mathrm{full},t,j},
\qquad
\sigma_j^2 = \sigma^{2,(e)}_{t,j} .
\end{equation}
Taylor's theorem around $v_j$ gives
\begin{equation}
\phi_\varepsilon(\widehat v_j)
=
\phi_\varepsilon(v_j)
+
\phi_\varepsilon'(v_j)(\widehat v_j-v_j)
+
\frac{1}{2}\phi_\varepsilon''(v_j)(\widehat v_j-v_j)^2
+
\frac{1}{6}\phi_\varepsilon'''(\xi_j)(\widehat v_j-v_j)^3,
\label{eq:taylor_inverse_sqrt}
\end{equation}
where $\xi_j$ lies between $\widehat v_j$ and $v_j$. The third derivative is
\begin{equation}
\phi_\varepsilon'''(x)
=
-\frac{15}{8}(x+\varepsilon)^{-7/2} .
\label{eq:inverse_sqrt_third_derivative}
\end{equation}
Taking expectations in Eq.~\eqref{eq:taylor_inverse_sqrt}, the first-order term vanishes by Eq.~\eqref{eq:unbiased_for_jensen}. Proposition~1 gives
\begin{equation}
\mathbb{E}\!\left[(\widehat v_j-v_j)^2\right]
=
\operatorname{Var}(\widehat v_j)
=
\frac{N-n}{Nn}\sigma_j^2 .
\label{eq:variance_substitution_inverse_sqrt}
\end{equation}
Substituting Eq.~\eqref{eq:inverse_sqrt_derivatives} and Eq.~\eqref{eq:variance_substitution_inverse_sqrt} proves the leading term in Eq.~\eqref{eq:inverse_sqrt_bias_expansion}.

The remainder bound follows from Eq.~\eqref{eq:local_inverse_sqrt_assumption}. For every point $\xi_j$ between $\widehat v_j$ and $v_j$,
\begin{equation}
\xi_j+\varepsilon
\ge
(1-\zeta_{\mathrm{loc}})(v_j+\varepsilon).
\label{eq:xi_lower_bound}
\end{equation}
Therefore,
\begin{equation}
\left|
\frac{1}{6}\phi_\varepsilon'''(\xi_j)(\widehat v_j-v_j)^3
\right|
\le
\frac{5}{16}
\frac{|\widehat v_j-v_j|^3}
{(1-\zeta_{\mathrm{loc}})^{7/2}(v_j+\varepsilon)^{7/2}} .
\label{eq:pointwise_remainder_bound}
\end{equation}
Taking expectations gives Eq.~\eqref{eq:inverse_sqrt_remainder_bound}. This completes the proof. \hfill $\square$

Equation~\eqref{eq:inverse_sqrt_bias_expansion} shows that the expected preconditioner error is positive to second order. Its leading term scales as
\begin{equation}
\frac{1}{n}
\frac{\sigma^{2,(e)}_{t,j}}
{\bigl(v^{(e)}_{\mathrm{full},t,j}+\varepsilon\bigr)^{5/2}},
\label{eq:coordinate_bias_scaling}
\end{equation}
up to the finite-population factor $(N-n)/N$. Thus, coordinates with small full-data denominators and high squared-gradient variability are the most vulnerable. Relative to the ideal coefficient $\phi_\varepsilon(v^{(e)}_{\mathrm{full},t,j})$, the leading inflation is
\begin{equation}
\frac{
\mathbb{E}\!\left[
\phi_\varepsilon\!\left(\widehat v^{(e)}_{\mathrm{few},t,j}\right)
\right]
}
{
\phi_\varepsilon\!\left(v^{(e)}_{\mathrm{full},t,j}\right)
}
-1
=
\frac{3}{8}
\frac{N-n}{Nn}
\frac{\sigma^{2,(e)}_{t,j}}
{\bigl(v^{(e)}_{\mathrm{full},t,j}+\varepsilon\bigr)^2}
+
\widetilde R^{(e)}_{t,j},
\label{eq:relative_inverse_sqrt_bias}
\end{equation}
where $\widetilde R^{(e)}_{t,j}=R^{(e)}_{t,j}/\phi_\varepsilon(v^{(e)}_{\mathrm{full},t,j})$. Summing the leading relative term over $j$ recovers the effective variability $\mathcal{U}^{(e)}_t$ defined above. Hence, the same quantity that makes the diagonal second moment hard to estimate also creates an upward bias in Adam's coordinatewise learning-rate multipliers.

The same amplification can be written without randomness. For a fixed support set and a fixed numerator $m^{(e)}_{\mathrm{few},t}\in\mathbb{R}^{d_e}$, the stabilized relative denominator error is
\begin{equation}
\delta^{(e)}_{t,j}
=
\frac{
 v^{(e)}_{\mathrm{full},t,j}
-
 v^{(e)}_{\mathrm{few},t,j}
}
{v^{(e)}_{\mathrm{full},t,j}+\varepsilon}
\in (-\infty,1) .
\label{eq:stabilized_relative_denominator_error}
\end{equation}
The upper bound $\delta^{(e)}_{t,j}<1$ follows from $v^{(e)}_{\mathrm{few},t,j}\ge 0$ and $\varepsilon>0$. A positive value of $\delta^{(e)}_{t,j}$ means that the support denominator is smaller than the full-data denominator. The coordinatewise preconditioner ratio is exactly
\begin{equation}
\frac{
\phi_\varepsilon\!\left(v^{(e)}_{\mathrm{few},t,j}\right)
}
{
\phi_\varepsilon\!\left(v^{(e)}_{\mathrm{full},t,j}\right)
}
=
\frac{1}{\sqrt{1-\delta^{(e)}_{t,j}}} .
\label{eq:coordinate_multiplier_delta}
\end{equation}
Thus, a denominator underestimate is amplified nonlinearly. If $\delta^{(e)}_{t,j}\in[0,1)$, then
\begin{equation}
\frac{1}{\sqrt{1-\delta^{(e)}_{t,j}}}
\ge
1+\frac{1}{2}\delta^{(e)}_{t,j},
\label{eq:first_order_multiplier_bound}
\end{equation}
because the function $r(\delta)=(1-\delta)^{-1/2}$ is convex, and its tangent at zero is $1+\delta/2$.

Finally, let $u^{(e)}_{\mathrm{ff},t}$ and $u^{(e)}_{\mathrm{fr},t}$ be the two Adam-like directions defined earlier in this section. For nonzero $m^{(e)}_{\mathrm{few},t}$, the active-coordinate weights are
\begin{equation}
 w^{(e)}_{t,j}
 =
 \frac{
 \bigl(m^{(e)}_{\mathrm{few},t,j}\bigr)^2 /
 \bigl(v^{(e)}_{\mathrm{full},t,j}+\varepsilon\bigr)
 }
 {\sum_{k=1}^{d_e}
 \bigl(m^{(e)}_{\mathrm{few},t,k}\bigr)^2 /
 \bigl(v^{(e)}_{\mathrm{full},t,k}+\varepsilon\bigr)} ,
 \qquad
 \sum_{j=1}^{d_e} w^{(e)}_{t,j}=1 .
\label{eq:active_coordinate_weights}
\end{equation}
Then
\begin{equation}
\frac{\|u^{(e)}_{\mathrm{ff},t}\|_2^2}
{\|u^{(e)}_{\mathrm{fr},t}\|_2^2}
=
\sum_{j=1}^{d_e}
\frac{w^{(e)}_{t,j}}{1-\delta^{(e)}_{t,j}} .
\label{eq:norm_ratio_delta_exact}
\end{equation}
This identity follows by substituting Eq.~\eqref{eq:stabilized_relative_denominator_error} into the two squared norms. If the weighted mean error
\begin{equation}
\bar\delta^{(e)}_t
=
\sum_{j=1}^{d_e}w^{(e)}_{t,j}\delta^{(e)}_{t,j}
\label{eq:weighted_mean_denominator_error}
\end{equation}
satisfies $\bar\delta^{(e)}_t\in[0,1)$, then Jensen's inequality, applied to the convex function $h(\delta)=(1-\delta)^{-1}$, gives
\begin{equation}
\frac{\|u^{(e)}_{\mathrm{ff},t}\|_2}
{\|u^{(e)}_{\mathrm{fr},t}\|_2}
\ge
\frac{1}{\sqrt{1-\bar\delta^{(e)}_t}} .
\label{eq:norm_inflation_lower_bound}
\end{equation}
Therefore, denominator underestimation on coordinates that carry numerator mass necessarily inflates the adaptive update norm. This is the theoretical counterpart of the empirical scale diagnostic $\|u^{(e)}_{\mathrm{ff},t}\|_2/\|u^{(e)}_{\mathrm{fr},t}\|_2$ in Experiments 1--5. The estimator error is first created by under-sampling, but Adam's inverse square root turns that error into an oversized coordinatewise learning rate and then into an oversized update budget.

\subsection{Scale inflation can dominate directional alignment}

The previous analysis shows that the few-shot denominator can inflate the norm of an Adam-like direction while changing its angle only moderately. We now show why this scale error can be harmful even when the update direction remains positively aligned with the full-train gradient. The point is
local and deterministic. Directional alignment controls the first-order term of the full-train loss, but the step length controls the second-order term.

Let \(u_{\mathrm{ff},t},u_{\mathrm{fr},t}\in\mathbb{R}^d\) denote the global Adam-like directions obtained by concatenating the encoder-wise vectors \(u^{(e)}_{\mathrm{ff},t}\) and \(u^{(e)}_{\mathrm{fr},t}\) over \(e\in\{\mathrm{vision},\mathrm{text}\}\). The actual Adam-like step has the form
\begin{equation}
    \Theta_{t+1}=\Theta_t-\eta u_{\mathrm{ff},t}.
\end{equation}
For \(u_{\mathrm{ff},t}\ne0\), its unit direction and parameter-space length are
\begin{equation}
    p_{\mathrm{ff},t}
    =
    \frac{u_{\mathrm{ff},t}}{\|u_{\mathrm{ff},t}\|_2}
    \in\mathbb{R}^d,
    \qquad
    \omega_{\mathrm{ff},t}
    =
    \eta\|u_{\mathrm{ff},t}\|_2
    \in\mathbb{R}_{>0}.
    \label{eq:adam_direction_and_length}
\end{equation}
Thus, the same update can be written as
\begin{equation}
    \Theta_{t+1}=\Theta_t-\omega_{\mathrm{ff},t}p_{\mathrm{ff},t}.
\end{equation}
The vector \(p_{\mathrm{ff},t}\) describes geometry. The scalar \(\omega_{\mathrm{ff},t}\) describes scale. A high cosine only constrains the first object. It does not, by itself, make the second object safe.

We first state a local step-length result for the full-train objective. Let
\[
    m_{\mathrm{full},t}
    =
    \nabla L_{\mathrm{full}}(\Theta_t)
    \in\mathbb{R}^d
\]
be the full-train gradient. For a unit vector \(p_t^{\mathrm{step}}\in\mathbb{R}^d\), the full-train directional alignment is
\begin{equation}
    a_{\mathrm{full}}(p_t^{\mathrm{step}};t)
    =
    \frac{\langle m_{\mathrm{full},t},p_t^{\mathrm{step}}\rangle}
    {\|m_{\mathrm{full},t}\|_2}
    \in[-1,1],
    \label{eq:full_directional_alignment}
\end{equation}
whenever \(m_{\mathrm{full},t}\ne 0\). The update \(\Theta_t-\omega p_t^{\mathrm{step}}\) is a first-order descent step for \(L_{\mathrm{full}}\) when \(a_{\mathrm{full}}(p_t^{\mathrm{step}};t)>0\).

\paragraph{Proposition 3 (local step-length threshold).}
The full-train gradient satisfies \(m_{\mathrm{full},t}\ne0\). Let \(p_t^{\mathrm{step}}\in\mathbb{R}^d\) be a unit vector with \(a_{\mathrm{full}}(p_t^{\mathrm{step}};t)>0\), and let \(\omega>0\). The objective \(L_{\mathrm{full}}\) is assumed twice continuously differentiable on an open set containing the segment
\begin{equation}
    \mathcal{S}_{t,\omega}(p_t^{\mathrm{step}})
    =
    \{\Theta_t-sp_t^{\mathrm{step}}:0\le s\le \omega\}.
\end{equation}
For \(s\in[0,\omega]\), the directional curvature along this segment is
\begin{equation}
    \kappa_{\mathrm{full}}(s;p_t^{\mathrm{step}},t)
    =
    (p_t^{\mathrm{step}})^\top
    \nabla^2 L_{\mathrm{full}}(\Theta_t-sp_t^{\mathrm{step}})
    p_t^{\mathrm{step}}
    \in\mathbb{R}.
    \label{eq:full_ray_curvature}
\end{equation}
If \(\kappa_{\mathrm{full}}(s;p_t^{\mathrm{step}},t)\le \lambda_{\mathrm{up}}\) for all \(s\in[0,\omega]\), where
\(\lambda_{\mathrm{up}}>0\), then
\begin{equation}
    L_{\mathrm{full}}(\Theta_t-\omega p_t^{\mathrm{step}})-L_{\mathrm{full}}(\Theta_t)
    \le
    -\omega a_{\mathrm{full}}(p_t^{\mathrm{step}};t)\|m_{\mathrm{full},t}\|_2
    +
    \frac{\lambda_{\mathrm{up}}}{2}\omega^2 .
    \label{eq:full_loss_upper_step_bound}
\end{equation}
Consequently, the step is certified to be non-increasing for \(L_{\mathrm{full}}\) whenever
\begin{equation}
    \omega
    \le
    \frac{2a_{\mathrm{full}}(p_t^{\mathrm{step}};t)\|m_{\mathrm{full},t}\|_2}
    {\lambda_{\mathrm{up}}}.
    \label{eq:certified_safe_length}
\end{equation}
If \(\kappa_{\mathrm{full}}(s;p_t^{\mathrm{step}},t)\ge \lambda_{\mathrm{low}}\) for all \(s\in[0,\omega]\), where
\(\lambda_{\mathrm{low}}>0\), then
\begin{equation}
    L_{\mathrm{full}}(\Theta_t-\omega p_t^{\mathrm{step}})-L_{\mathrm{full}}(\Theta_t)
    \ge
    -\omega a_{\mathrm{full}}(p_t^{\mathrm{step}};t)\|m_{\mathrm{full},t}\|_2
    +
    \frac{\lambda_{\mathrm{low}}}{2}\omega^2 .
    \label{eq:full_loss_lower_step_bound}
\end{equation}
Consequently, the step is certified to increase \(L_{\mathrm{full}}\) whenever
\begin{equation}
    \omega
    >
    \frac{2a_{\mathrm{full}}(p_t^{\mathrm{step}};t)\|m_{\mathrm{full},t}\|_2}
    {\lambda_{\mathrm{low}}}.
    \label{eq:certified_unsafe_length}
\end{equation}

\emph{Proof.}
The scalar restriction is
\begin{equation}
    \psi(s)=L_{\mathrm{full}}(\Theta_t-sp_t^{\mathrm{step}}),
    \qquad s\in[0,\omega].
\end{equation}
By the chain rule,
\begin{equation}
    \psi'(0)
    =
    -\langle m_{\mathrm{full},t},p_t^{\mathrm{step}}\rangle
    =
    -a_{\mathrm{full}}(p_t^{\mathrm{step}};t)\|m_{\mathrm{full},t}\|_2 .
    \label{eq:psi_first_derivative}
\end{equation}
A second application of the chain rule gives
\begin{equation}
    \psi''(s)
    =
    (p_t^{\mathrm{step}})^\top
    \nabla^2L_{\mathrm{full}}(\Theta_t-sp_t^{\mathrm{step}})
    p_t^{\mathrm{step}}
    =
    \kappa_{\mathrm{full}}(s;p_t^{\mathrm{step}},t).
    \label{eq:psi_second_derivative}
\end{equation}
Taylor's formula with integral remainder gives
\begin{equation}
    \psi(\omega)-\psi(0)
    =
    \omega\psi'(0)
    +
    \int_0^\omega(\omega-s)\psi''(s)\,ds .
    \label{eq:integral_remainder_scale_proof}
\end{equation}
If \(\psi''(s)\le\lambda_{\mathrm{up}}\) on \([0,\omega]\), then
\begin{equation}
    \int_0^\omega(\omega-s)\psi''(s)\,ds
    \le
    \lambda_{\mathrm{up}}\int_0^\omega(\omega-s)\,ds
    =
    \frac{\lambda_{\mathrm{up}}}{2}\omega^2 .
\end{equation}
Substituting this inequality and Eq.~\eqref{eq:psi_first_derivative} into Eq.~\eqref{eq:integral_remainder_scale_proof} proves Eq.~\eqref{eq:full_loss_upper_step_bound}. If Eq.~\eqref{eq:certified_safe_length} holds, then the right-hand side of Eq.~\eqref{eq:full_loss_upper_step_bound} is non-positive. This proves the certified non-increase claim.

The lower bound is analogous. If \(\psi''(s)\ge\lambda_{\mathrm{low}}\) on \([0,\omega]\), then
\begin{equation}
    \int_0^\omega(\omega-s)\psi''(s)\,ds
    \ge
    \lambda_{\mathrm{low}}\int_0^\omega(\omega-s)\,ds
    =
    \frac{\lambda_{\mathrm{low}}}{2}\omega^2 .
\end{equation}
Substituting this inequality and Eq.~\eqref{eq:psi_first_derivative} into Eq.~\eqref{eq:integral_remainder_scale_proof} proves Eq.~\eqref{eq:full_loss_lower_step_bound}. If Eq.~\eqref{eq:certified_unsafe_length} holds, then the right-hand side of Eq.~\eqref{eq:full_loss_lower_step_bound} is positive. This proves the certified increase claim.
\hfill \(\square\)

Proposition 3 separates alignment from scale. The scalar \(a_{\mathrm{full}}(p_t^{\mathrm{step}};t)\) measures the first-order usefulness of the direction \(p_t^{\mathrm{step}}\). The admissible step length also depends on the local curvature of \(L_{\mathrm{full}}\). Therefore, positive alignment is not sufficient for safety. The step can still increase the full-train loss when its length exceeds the local threshold.

We next apply this observation to adaptive scale inflation. The statement deliberately uses the same unit direction for two updates. This isolates scale from geometry. It shows that scale alone can cause failure even in the most favorable case where the direction is unchanged.

\paragraph{Proposition 4 (inflated scale can defeat a good direction).}
The full-train gradient satisfies \(m_{\mathrm{full},t}\ne0\). Let \(p_t^{\mathrm{step}}\in\mathbb{R}^d\) be a unit vector with
\(a_{\mathrm{full}}(p_t^{\mathrm{step}};t)>0\). Two updates along this same direction are
\begin{equation}
    \Theta^{\mathrm{ref}}_{t+1}
    =
    \Theta_t-\omega_{\mathrm{ref}}p_t^{\mathrm{step}},
    \qquad
    \Theta^{\mathrm{few}}_{t+1}
    =
    \Theta_t-\omega_{\mathrm{few}}p_t^{\mathrm{step}},
    \label{eq:reference_and_few_updates_same_direction}
\end{equation}
where \(\omega_{\mathrm{ref}}>0\), \(\omega_{\mathrm{few}}>0\), and
\begin{equation}
    \omega_{\mathrm{few}}
    =
    \xi_t\omega_{\mathrm{ref}},
    \qquad
    \xi_t>1 .
    \label{eq:scale_inflation_factor}
\end{equation}
The objective \(L_{\mathrm{full}}\) is assumed twice continuously differentiable on an open set containing
\(\{\Theta_t-sp_t^{\mathrm{step}}:0\le s\le \omega_{\mathrm{few}}\}\). The curvature on this segment satisfies
\begin{equation}
    0<\lambda_{\mathrm{low}}
    \le
    \kappa_{\mathrm{full}}(s;p_t^{\mathrm{step}},t)
    \le
    \lambda_{\mathrm{up}} .
    \label{eq:two_sided_curvature_bound_scale}
\end{equation}
If
\begin{equation}
    \omega_{\mathrm{ref}}
    \le
    \frac{2a_{\mathrm{full}}(p_t^{\mathrm{step}};t)\|m_{\mathrm{full},t}\|_2}
    {\lambda_{\mathrm{up}}}
    \quad\text{and}\quad
    \omega_{\mathrm{few}}
    >
    \frac{2a_{\mathrm{full}}(p_t^{\mathrm{step}};t)\|m_{\mathrm{full},t}\|_2}
    {\lambda_{\mathrm{low}}},
    \label{eq:safe_reference_unsafe_few_scale}
\end{equation}
then
\begin{equation}
    L_{\mathrm{full}}(\Theta^{\mathrm{ref}}_{t+1})
    \le
    L_{\mathrm{full}}(\Theta_t),
    \qquad
    L_{\mathrm{full}}(\Theta^{\mathrm{few}}_{t+1})
    >
    L_{\mathrm{full}}(\Theta_t).
    \label{eq:scale_defeats_alignment_result}
\end{equation}

\emph{Proof.}
The reference update stays inside the longer segment because \(\omega_{\mathrm{ref}}<\omega_{\mathrm{few}}\). The upper curvature bound in Eq.~\eqref{eq:two_sided_curvature_bound_scale} and the first inequality in Eq.~\eqref{eq:safe_reference_unsafe_few_scale} allow us to apply
Proposition~3 with \(\omega=\omega_{\mathrm{ref}}\). This gives
\begin{equation}
    L_{\mathrm{full}}(\Theta^{\mathrm{ref}}_{t+1})
    \le
    L_{\mathrm{full}}(\Theta_t).
\end{equation}
The lower curvature bound in Eq.~\eqref{eq:two_sided_curvature_bound_scale} and the second inequality in Eq.~\eqref{eq:safe_reference_unsafe_few_scale} allow us to apply Proposition~3 with \(\omega=\omega_{\mathrm{few}}\). This gives
\begin{equation}
    L_{\mathrm{full}}(\Theta^{\mathrm{few}}_{t+1})
    >
    L_{\mathrm{full}}(\Theta_t).
\end{equation}
Combining the two inequalities proves the claim.
\hfill \(\square\)

Proposition 4 gives a precise interpretation of the scale-versus-geometry experiments. A cosine diagnostic can remain favorable because it measures the direction \(p_t^{\mathrm{step}}\), or a closely related angular quantity. It does not measure the step length \(\omega\). For Adam, the few-shot denominator can make
\begin{equation}
    \omega_{\mathrm{ff},t}
    =
    \eta\|u_{\mathrm{ff},t}\|_2
\end{equation}
much larger than the corresponding reference length, such as
\begin{equation}
    \omega_{\mathrm{fr},t}
    =
    \eta\|u_{\mathrm{fr},t}\|_2 .
\end{equation}
The norm-inflation bound above shows how this happens when the support denominator is too small in coordinates that carry numerator mass. Thus, Adam can preserve much of the local direction while crossing the safe full-train step-length threshold. This explains why correcting the global step-0 scale removes most of the collapse, while residual encoder-wise and blockwise scale errors can remain after a single global rescaling.

\subsection{Denominator repair under numerator error}

The previous results explain why the few-shot denominator can make Adam unsafe even when the adaptive direction is not badly rotated. We now show why denominator repair is not sufficient on every trajectory. The key point is that replacing the denominator can remove only the denominator error. It cannot repair a stale numerator.

Let $m_{\mathrm{few},t},m_{\mathrm{full},t}\in\mathbb{R}^{d}$ denote the global few-shot and full-train gradients at checkpoint $\Theta_t$. Let $v_{\mathrm{few},t},v_{\mathrm{full},t}\in\mathbb{R}^{d}_{\ge 0}$ denote the corresponding global diagonal second moments. The corresponding diagonal preconditioner matrices are
\begin{equation}
\Pi_{\mathrm{few},t}=\operatorname{diag}\left((v_{\mathrm{few},t,j}+\varepsilon)^{-1/2}\right)_{j=1}^{d},\qquad
\Pi_{\mathrm{full},t}=\operatorname{diag}\left((v_{\mathrm{full},t,j}+\varepsilon)^{-1/2}\right)_{j=1}^{d},
\end{equation}
where $\varepsilon>0$. Both matrices belong to $\mathbb{R}^{d\times d}$ and are positive definite. In this notation,
\begin{equation}
 u_{\mathrm{ff},t}=\Pi_{\mathrm{few},t}m_{\mathrm{few},t},\qquad
 u_{\mathrm{fr},t}=\Pi_{\mathrm{full},t}m_{\mathrm{few},t},\qquad
 u_{\mathrm{rr},t}=\Pi_{\mathrm{full},t}m_{\mathrm{full},t}.
\end{equation}
The direction $u_{\mathrm{ff},t}$ is the live few-shot Adam-like direction. The direction $u_{\mathrm{fr},t}$ is the oracle-denominator repair. The direction $u_{\mathrm{rr},t}$ is the full-train reference.

The denominator and numerator errors are
\begin{equation}
 e_{\mathrm{den},t}=\left(\Pi_{\mathrm{few},t}-\Pi_{\mathrm{full},t}\right)m_{\mathrm{few},t}\in\mathbb{R}^{d},
 \qquad
 e_{\mathrm{num},t}=\Pi_{\mathrm{full},t}\left(m_{\mathrm{few},t}-m_{\mathrm{full},t}\right)\in\mathbb{R}^{d}.
\end{equation}
The first vector is the part of the adaptive-direction error caused by using the few-shot denominator while keeping the few-shot numerator fixed. The second vector is the residual error caused by using the few-shot numerator after the denominator has already been repaired.

\paragraph{Proposition 5 (exact decomposition of denominator repair).}
At every checkpoint $t$,
\begin{equation}
 u_{\mathrm{ff},t}-u_{\mathrm{rr},t}=e_{\mathrm{den},t}+e_{\mathrm{num},t},
 \qquad
 u_{\mathrm{fr},t}-u_{\mathrm{rr},t}=e_{\mathrm{num},t}.
\end{equation}
Consequently, oracle denominator repair decreases the squared distance to the reference direction if and only if
\begin{equation}
 \left\|e_{\mathrm{den},t}\right\|_{2}^{2}
 +2\left\langle e_{\mathrm{den},t},e_{\mathrm{num},t}\right\rangle>0.
\end{equation}
A sufficient condition is
\begin{equation}
 \left\|e_{\mathrm{den},t}\right\|_{2}>2\left\|e_{\mathrm{num},t}\right\|_{2}.
\end{equation}
After the repair, the remaining distance to the reference direction is exactly
\begin{equation}
 \left\|u_{\mathrm{fr},t}-u_{\mathrm{rr},t}\right\|_{2}=\left\|e_{\mathrm{num},t}\right\|_{2}.
\end{equation}

\emph{Proof.}
The first identity follows by adding and subtracting $\Pi_{\mathrm{full},t}m_{\mathrm{few},t}$:
\begin{align}
 u_{\mathrm{ff},t}-u_{\mathrm{rr},t}
 &=\Pi_{\mathrm{few},t}m_{\mathrm{few},t}-\Pi_{\mathrm{full},t}m_{\mathrm{full},t}\\
 &=\left(\Pi_{\mathrm{few},t}-\Pi_{\mathrm{full},t}\right)m_{\mathrm{few},t}
 +\Pi_{\mathrm{full},t}\left(m_{\mathrm{few},t}-m_{\mathrm{full},t}\right)\\
 &=e_{\mathrm{den},t}+e_{\mathrm{num},t}.
\end{align}
The second identity is immediate from
\begin{equation}
 u_{\mathrm{fr},t}-u_{\mathrm{rr},t}
 =\Pi_{\mathrm{full},t}m_{\mathrm{few},t}-\Pi_{\mathrm{full},t}m_{\mathrm{full},t}
 =e_{\mathrm{num},t}.
\end{equation}
Thus, the difference between the squared errors before and after the denominator repair is
\begin{align}
 \left\|u_{\mathrm{ff},t}-u_{\mathrm{rr},t}\right\|_{2}^{2}
 -\left\|u_{\mathrm{fr},t}-u_{\mathrm{rr},t}\right\|_{2}^{2}
 &=\left\|e_{\mathrm{den},t}+e_{\mathrm{num},t}\right\|_{2}^{2}
 -\left\|e_{\mathrm{num},t}\right\|_{2}^{2}\\
 &=\left\|e_{\mathrm{den},t}\right\|_{2}^{2}
 +2\left\langle e_{\mathrm{den},t},e_{\mathrm{num},t}\right\rangle .
\end{align}
This proves the necessary and sufficient condition. By Cauchy's inequality,
\begin{equation}
 \left\|e_{\mathrm{den},t}\right\|_{2}^{2}
 +2\left\langle e_{\mathrm{den},t},e_{\mathrm{num},t}\right\rangle
 \ge
 \left\|e_{\mathrm{den},t}\right\|_{2}^{2}
 -2\left\|e_{\mathrm{den},t}\right\|_{2}\left\|e_{\mathrm{num},t}\right\|_{2}.
\end{equation}
The right-hand side is positive whenever $\left\|e_{\mathrm{den},t}\right\|_{2}>2\left\|e_{\mathrm{num},t}\right\|_{2}$. The residual-distance identity follows from $u_{\mathrm{fr},t}-u_{\mathrm{rr},t}=e_{\mathrm{num},t}$.
\hfill \(\square\)

This proposition formalizes the two-stage empirical picture. Early in training, the denominator term can dominate, so replacing $\Pi_{\mathrm{few},t}$ with $\Pi_{\mathrm{full},t}$ can move the update close to the full-train reference. Later, the residual term $e_{\mathrm{num},t}$ can become the main error. In that regime, denominator repair may still reduce scale damage, but it cannot make $u_{\mathrm{fr},t}$ close to $u_{\mathrm{rr},t}$.

The next result gives a descent version of the same statement. It uses a staleness ratio that is adapted to the repaired denominator. Let
\begin{equation}
 A_{\mathrm{full},t}=\Pi_{\mathrm{full},t}^{1/2}
 =\operatorname{diag}\left((v_{\mathrm{full},t,j}+\varepsilon)^{-1/4}\right)_{j=1}^{d}.
\end{equation}
For $m_{\mathrm{full},t}\ne0$, the repaired-denominator numerator-staleness ratio is
\begin{equation}
 \rho_{\mathrm{num},t}
 =
 \frac{\left\|A_{\mathrm{full},t}\left(m_{\mathrm{few},t}-m_{\mathrm{full},t}\right)\right\|_{2}}
 {\left\|A_{\mathrm{full},t}m_{\mathrm{full},t}\right\|_{2}}.
\end{equation}
The numerator is fresh relative to the full-train objective when $\rho_{\mathrm{num},t}<1$. It is stale at the scale of the repaired denominator when $\rho_{\mathrm{num},t}\ge 1$.

\paragraph{Proposition 6 (descent guarantee before numerator staleness dominates).}
The assumptions are $m_{\mathrm{full},t}=\nabla L_{\mathrm{full}}(\Theta_t)\ne0$ and $\rho_{\mathrm{num},t}<1$. Then $-u_{\mathrm{fr},t}$ is a first-order descent direction for the full-train loss:
\begin{equation}
 \left\langle m_{\mathrm{full},t},u_{\mathrm{fr},t}\right\rangle
 \ge
 \left(1-\rho_{\mathrm{num},t}\right)
 \left\|A_{\mathrm{full},t}m_{\mathrm{full},t}\right\|_{2}^{2}>0.
\end{equation}
In addition, $L_{\mathrm{full}}$ is assumed $L_{\mathrm{sm},t}$-smooth on the segment
\begin{equation}
 \left\{\Theta_t-s u_{\mathrm{fr},t}:0\le s\le \eta\right\}
\end{equation}
for some $L_{\mathrm{sm},t}>0$. If
\begin{equation}
 0<\eta<
 \frac{2\left(1-\rho_{\mathrm{num},t}\right)
 \left\|A_{\mathrm{full},t}m_{\mathrm{full},t}\right\|_{2}^{2}}
 {L_{\mathrm{sm},t}\left\|u_{\mathrm{fr},t}\right\|_{2}^{2}},
\end{equation}
then
\begin{equation}
 L_{\mathrm{full}}\left(\Theta_t-\eta u_{\mathrm{fr},t}\right)<L_{\mathrm{full}}(\Theta_t).
\end{equation}
If $\rho_{\mathrm{num},t}\ge 1$, then the ratio alone gives no uniform descent guarantee for denominator repair. In particular, there exists a few-shot numerator with the same value of $\rho_{\mathrm{num},t}$ for which
\begin{equation}
 \left\langle m_{\mathrm{full},t},u_{\mathrm{fr},t}\right\rangle\le 0.
\end{equation}

\emph{Proof.}
Let
\begin{equation}
 q_t^{\mathrm{sig}}=A_{\mathrm{full},t}m_{\mathrm{full},t},
 \qquad
 q_t^{\mathrm{err}}=A_{\mathrm{full},t}\left(m_{\mathrm{few},t}-m_{\mathrm{full},t}\right).
\end{equation}
Since $\varepsilon>0$, the matrix $A_{\mathrm{full},t}$ is positive definite. Thus, $q_t^{\mathrm{sig}}\ne 0$. By the definition of $\rho_{\mathrm{num},t}$, we have $\left\|q_t^{\mathrm{err}}\right\|_2=\rho_{\mathrm{num},t}\left\|q_t^{\mathrm{sig}}\right\|_2$. The repaired direction satisfies
\begin{equation}
 u_{\mathrm{fr},t}=\Pi_{\mathrm{full},t}m_{\mathrm{few},t}
 =A_{\mathrm{full},t}^{2}m_{\mathrm{few},t}.
\end{equation}
Therefore,
\begin{align}
 \left\langle m_{\mathrm{full},t},u_{\mathrm{fr},t}\right\rangle
 &=\left\langle m_{\mathrm{full},t},A_{\mathrm{full},t}^{2}m_{\mathrm{few},t}\right\rangle\\
 &=\left\langle A_{\mathrm{full},t}m_{\mathrm{full},t},A_{\mathrm{full},t}m_{\mathrm{few},t}\right\rangle\\
 &=\left\langle q_t^{\mathrm{sig}},q_t^{\mathrm{sig}}+q_t^{\mathrm{err}}\right\rangle\\
 &=\left\|q_t^{\mathrm{sig}}\right\|_{2}^{2}+\left\langle q_t^{\mathrm{sig}},q_t^{\mathrm{err}}\right\rangle .
\end{align}
Cauchy's inequality gives
\begin{equation}
 \left\langle q_t^{\mathrm{sig}},q_t^{\mathrm{err}}\right\rangle
 \ge -\left\|q_t^{\mathrm{sig}}\right\|_2\left\|q_t^{\mathrm{err}}\right\|_2
 =-\rho_{\mathrm{num},t}\left\|q_t^{\mathrm{sig}}\right\|_2^2.
\end{equation}
Hence
\begin{equation}
 \left\langle m_{\mathrm{full},t},u_{\mathrm{fr},t}\right\rangle
 \ge \left(1-\rho_{\mathrm{num},t}\right)\left\|q_t^{\mathrm{sig}}\right\|_2^2.
\end{equation}
This proves the first claim.

The smoothness assumption implies the standard descent inequality
\begin{equation}
 L_{\mathrm{full}}\left(\Theta_t-\eta u_{\mathrm{fr},t}\right)
 \le
 L_{\mathrm{full}}(\Theta_t)
 -\eta\left\langle m_{\mathrm{full},t},u_{\mathrm{fr},t}\right\rangle
 +\frac{L_{\mathrm{sm},t}\eta^2}{2}\left\|u_{\mathrm{fr},t}\right\|_2^2.
\end{equation}
Substituting the lower bound on $\left\langle m_{\mathrm{full},t},u_{\mathrm{fr},t}\right\rangle$ shows that the right-hand side is strictly smaller than $L_{\mathrm{full}}(\Theta_t)$ under the stated step-size condition.

It remains to prove the last claim. Let $\rho_{\mathrm{num},t}\ge 1$ be fixed. The construction
\begin{equation}
 q_t^{\mathrm{err}}=-\rho_{\mathrm{num},t} q_t^{\mathrm{sig}},
 \qquad
 m_{\mathrm{few},t}=m_{\mathrm{full},t}+A_{\mathrm{full},t}^{-1}q_t^{\mathrm{err}}
\end{equation}
gives $\left\|q_t^{\mathrm{err}}\right\|_2/\left\|q_t^{\mathrm{sig}}\right\|_2=\rho_{\mathrm{num},t}$. Hence, the staleness ratio equals $\rho_{\mathrm{num},t}$. For this numerator,
\begin{equation}
 \left\langle m_{\mathrm{full},t},u_{\mathrm{fr},t}\right\rangle
 =\left\langle q_t^{\mathrm{sig}},q_t^{\mathrm{sig}}+q_t^{\mathrm{err}}\right\rangle
 =(1-\rho_{\mathrm{num},t})\left\|q_t^{\mathrm{sig}}\right\|_2^2\le 0.
\end{equation}
Thus, once the numerator error is at least as large as the full-train signal in the repaired-denominator metric, denominator repair alone has no uniform first-order descent guarantee.
\hfill \(\square\)

The theory explains why the denominator repair behaves differently across datasets. If $\rho_{\mathrm{num},t}<1$, then the repaired update keeps a positive full-train descent component and has a nonzero safe step interval. The factor $1-\rho_{\mathrm{num},t}$ vanishes as $\rho_{\mathrm{num},t}$ approaches one. The step-size certificate also depends on $\|u_{\mathrm{fr},t}\|_2^2$. If $\rho_{\mathrm{num},t}\ge 1$, a repaired denominator can no longer guarantee descent because the numerator itself may point away from the full-train objective. This matches the empirical pattern: denominator-only repair is strong when $u_{\mathrm{fr},t}$ remains close to $u_{\mathrm{rr},t}$, and it becomes insufficient on harder trajectories where the numerator has become stale.

\subsection{Blockwise scale mismatch remains after global rescaling}
The previous argument controls the global step length. It does not control how that length is distributed across encoders and Transformer blocks. We now formalize this limitation. Let \(\mathcal{B}_{\rm blk}\) be the set of trainable blocks. A block \(b\in\mathcal{B}_{\rm blk}\) denotes one encoder--Transformer-block pair, for example \(b=(e,j)\) with \(e\in\{\text{vision},\text{text}\}\). Let \(d_b\) be the number of trainable scalar parameters in block \(b\). Let \(P_b\in\mathbb{R}^{d\times d}\) be the diagonal projector that keeps the coordinates of block \(b\) and zeros out all other coordinates. The projectors form a partition of the trainable coordinates, so \(P_bP_c=0\) for \(b\neq c\) and \(\sum_{b\in\mathcal{B}_{\rm blk}}P_b=I_d\).

For an update vector \(\Delta_t\in\mathbb{R}^d\), the block relative update budget is
\begin{equation}
    r_b(\Delta_t;t)
    =
    \frac{\|P_b\Delta_t\|_2}{\|P_b\Theta_t\|_2},
    \qquad b\in\mathcal{B}_{\rm blk},
\end{equation}
where we assume \(\|P_b\Theta_t\|_2>0\) for the blocks under comparison. This is the blockwise analogue of the global relative update budget. Let \(\Delta^{\rm ad}_t\in\mathbb{R}^d\) be a raw Adam-like update and let \(\Delta^{\rm ref}_t\in\mathbb{R}^d\) be a reference update. The reference can be the SGD update or the oracle-denominator update. Both updates are assumed nonzero. The active reference blocks are
\begin{equation}
    \mathcal{A}_t
    =
    \{b\in\mathcal{B}_{\rm blk}: \|P_b\Delta^{\rm ref}_t\|_2>0\}.
\end{equation}
For each \(b\in\mathcal{A}_t\), the raw block scale ratio is
\begin{equation}
    a_b(t)
    =
    \frac{r_b(\Delta^{\rm ad}_t;t)}{r_b(\Delta^{\rm ref}_t;t)}
    =
    \frac{\|P_b\Delta^{\rm ad}_t\|_2}{\|P_b\Delta^{\rm ref}_t\|_2}.
\end{equation}
The equality holds because the parameter norm \(\|P_b\Theta_t\|_2\) cancels. A one-time global match chooses
\begin{equation}
    \gamma_t
    =
    \frac{\|\Delta^{\rm ref}_t\|_2}{\|\Delta^{\rm ad}_t\|_2},
    \qquad
    \Delta^{\rm gm}_t
    =
    \gamma_t\Delta^{\rm ad}_t .
\end{equation}
Then \(\|\Delta^{\rm gm}_t\|_2=\|\Delta^{\rm ref}_t\|_2\). Thus, the global relative update budgets are matched.

\paragraph{Proposition 7 (global matching does not imply blockwise matching).}
At checkpoint \(t\), the raw update is assumed to satisfy \(P_b\Delta_t^{\rm ad}=0\) for every \(b\notin\mathcal A_t\). Thus, it has no energy in reference-inactive blocks. For \(b\in\mathcal{A}_t\), the reference energy weight is
\begin{equation}
    \pi_b(t)
    =
    \frac{\|P_b\Delta^{\rm ref}_t\|_2^2}{\|\Delta^{\rm ref}_t\|_2^2},
    \qquad
    \pi_b(t)\in[0,1].
\end{equation}
Then \(\sum_{b\in\mathcal{A}_t}\pi_b(t)=1\), and the globally matched block ratio is
\begin{equation}
    \frac{r_b(\Delta^{\rm gm}_t;t)}{r_b(\Delta^{\rm ref}_t;t)}
    =
    \frac{a_b(t)}{
    \left(\sum_{c\in\mathcal{A}_t}\pi_c(t)a_c(t)^2\right)^{1/2}}
    \qquad
    \text{for every } b\in\mathcal{A}_t .
\end{equation}
Consequently,
\begin{equation}
    \sum_{b\in\mathcal{A}_t}\pi_b(t)
    \left(
    \frac{r_b(\Delta^{\rm gm}_t;t)}{r_b(\Delta^{\rm ref}_t;t)}
    \right)^2
    =1 .
\end{equation}
All active block budgets are matched if and only if the raw ratios \(a_b(t)\) are constant over \(\mathcal{A}_t\). If the raw ratios are not constant, then at least one active block is oversized and at least one active block is undersized after global matching. Moreover, the weighted squared residual after global matching is
\begin{equation}
    \sum_{b\in\mathcal{A}_t}\pi_b(t)
    \left(
    \frac{r_b(\Delta^{\rm gm}_t;t)}{r_b(\Delta^{\rm ref}_t;t)}-1
    \right)^2
    =
    2\left(
    1-
    \frac{\sum_{b\in\mathcal{A}_t}\pi_b(t)a_b(t)}
    {\left(\sum_{b\in\mathcal{A}_t}\pi_b(t)a_b(t)^2\right)^{1/2}}
    \right).
\end{equation}
This residual is zero if and only if \(a_b(t)\) is constant over \(\mathcal{A}_t\).

\emph{Proof.}
Because the block projectors are mutually orthogonal and sum to \(I_d\), we have
\begin{equation}
    \|\Delta^{\rm ref}_t\|_2^2
    =
    \sum_{b\in\mathcal{A}_t}\|P_b\Delta^{\rm ref}_t\|_2^2 .
\end{equation}
Hence \(\sum_{b\in\mathcal{A}_t}\pi_b(t)=1\). By the definition of \(a_b(t)\),
\begin{equation}
    \|P_b\Delta^{\rm ad}_t\|_2
    =
    a_b(t)\|P_b\Delta^{\rm ref}_t\|_2
    \qquad \text{for every } b\in\mathcal{A}_t .
\end{equation}
The inactive-block assumption gives
\begin{align}
    \|\Delta^{\rm ad}_t\|_2^2
    &=
    \sum_{b\in\mathcal{B}_{\rm blk}}\|P_b\Delta^{\rm ad}_t\|_2^2 \\
    &=
    \sum_{b\in\mathcal{A}_t}\|P_b\Delta^{\rm ad}_t\|_2^2 \\
    &=
    \sum_{b\in\mathcal{A}_t}a_b(t)^2\|P_b\Delta^{\rm ref}_t\|_2^2 \\
    &=
    \|\Delta^{\rm ref}_t\|_2^2
    \sum_{b\in\mathcal{A}_t}\pi_b(t)a_b(t)^2 .
\end{align}
It follows that
\begin{equation}
    \gamma_t
    =
    \frac{\|\Delta^{\rm ref}_t\|_2}{\|\Delta^{\rm ad}_t\|_2}
    =
    \left(\sum_{c\in\mathcal{A}_t}\pi_c(t)a_c(t)^2\right)^{-1/2} .
\end{equation}
For each active block,
\begin{align}
    \frac{r_b(\Delta^{\rm gm}_t;t)}{r_b(\Delta^{\rm ref}_t;t)}
    &=
    \frac{\|P_b\gamma_t\Delta^{\rm ad}_t\|_2}{\|P_b\Delta^{\rm ref}_t\|_2} \\
    &=
    \gamma_t a_b(t) \\
    &=
    \frac{a_b(t)}
    {\left(\sum_{c\in\mathcal{A}_t}\pi_c(t)a_c(t)^2\right)^{1/2}} .
\end{align}
This proves the expression for the globally matched block ratio. Squaring the expression and averaging with weights \(\pi_b(t)\) gives
\begin{equation}
    \sum_{b\in\mathcal{A}_t}\pi_b(t)
    \left(
    \frac{r_b(\Delta^{\rm gm}_t;t)}{r_b(\Delta^{\rm ref}_t;t)}
    \right)^2
    =
    \frac{\sum_{b\in\mathcal{A}_t}\pi_b(t)a_b(t)^2}
    {\sum_{c\in\mathcal{A}_t}\pi_c(t)a_c(t)^2}
    =1 .
\end{equation}
If all active block budgets are matched, then \(\gamma_ta_b(t)=1\) for every \(b\in\mathcal{A}_t\). Hence, all \(a_b(t)\) are equal to \(1/\gamma_t\). Conversely, if all \(a_b(t)\) are equal to a constant \(a_t>0\), then \(\gamma_t=1/a_t\), and every active block ratio equals one. This proves the equivalence.

If the raw ratios are not constant, the matched block ratios cannot all equal one. Their weighted squared average is exactly one. Hence, they cannot all be less than or equal to one, and they cannot all be greater than or equal to one. Thus, at least one active block is oversized, and at least one active block is undersized. Finally, expanding the weighted squared residual gives
\begin{align}
    &\sum_{b\in\mathcal{A}_t}\pi_b(t)
    \left(
    \frac{a_b(t)}{\left(\sum_c\pi_c(t)a_c(t)^2\right)^{1/2}}-1
    \right)^2 \\
    &=
    \frac{\sum_b\pi_b(t)a_b(t)^2}{\sum_c\pi_c(t)a_c(t)^2}
    -
    2\frac{\sum_b\pi_b(t)a_b(t)}{\left(\sum_c\pi_c(t)a_c(t)^2\right)^{1/2}}
    +1 \\
    &=
    2\left(
    1-
    \frac{\sum_b\pi_b(t)a_b(t)}
    {\left(\sum_c\pi_c(t)a_c(t)^2\right)^{1/2}}
    \right).
\end{align}
By Cauchy's inequality, \(\sum_b\pi_b(t)a_b(t)\leq(\sum_b\pi_b(t)a_b(t)^2)^{1/2}\). Equality holds if and only if the nonnegative values \(a_b(t)\) are constant on the weighted active set. This completes the proof.
\hfill \(\square\)

The proposition applies to Adam's denominator error with
\begin{equation}
    \Delta^{\rm ad}_t=-\eta u_{ff,t},
    \qquad
    \Delta^{\rm ref}_t=-\eta u_{fr,t},
\end{equation}
where \(u_{ff,t}=\Pi_{\rm few,t}m_{\rm few,t}\) uses the few-shot denominator and \(u_{fr,t}=\Pi_{\rm full,t}m_{\rm few,t}\) uses the full-data denominator with the same numerator. The positive diagonal preconditioners preserve the zero pattern of their common numerator, so the inactive-block assumption holds for this comparison. The raw block ratio becomes
\begin{equation}
    a_b(t)
    =
    \frac{\|P_bu_{ff,t}\|_2}{\|P_bu_{fr,t}\|_2} .
\end{equation}
Thus, a single global correction removes all blockwise denominator-induced scale errors only when this ratio is the same for all active blocks. The empirical blockwise denominator diagnostics show the opposite pattern: the scale error depends on the encoder and depth. Under the stated inactive-block assumption, the proposition explains why global step-0 matching can leave a non-trivial blockwise mismatch.

The remaining mismatch matters because blockwise safe step lengths need not be equal. The following deterministic statement isolates this point along one block line. It is the blockwise analogue of the scale-threshold argument above.

\paragraph{Proposition 8 (blockwise overscale can violate a local safe radius).}
For a checkpoint \(t\) and an active block \(b\), let \(q_{b,t}\in\mathbb{R}^d\) be a unit vector supported only on block \(b\), so \(P_bq_{b,t}=q_{b,t}\) and \(\|q_{b,t}\|_2=1\). The vector \(q_{b,t}\) is assumed to be a descent direction for the full-train loss, with
\begin{equation}
    c_{b,t}
    =
    -\langle \nabla L_{\rm full}(\Theta_t),q_{b,t}\rangle
    >0 .
\end{equation}
Let \(s^{\rm ref}_{b,t}>0\) and \(s^{\rm gm}_{b,t}>0\) be two block step lengths along the same direction. Along the segment \(\{\Theta_t+s q_{b,t}: 0\leq s\leq s^{\rm gm}_{b,t}\}\), the curvature satisfies
\begin{equation}
    0<\lambda_{b,t}
    \leq
    \frac{d^2}{ds^2}L_{\rm full}(\Theta_t+s q_{b,t})
    \leq
    \Lambda_{b,t}
\end{equation}
for finite constants \(\lambda_{b,t}\leq\Lambda_{b,t}\). If
\begin{equation}
    s^{\rm ref}_{b,t}
    \leq
    \frac{2c_{b,t}}{\Lambda_{b,t}},
    \qquad
    s^{\rm gm}_{b,t}
    >
    \frac{2c_{b,t}}{\lambda_{b,t}},
\end{equation}
then
\begin{equation}
    L_{\rm full}(\Theta_t+s^{\rm ref}_{b,t}q_{b,t})
    \leq
    L_{\rm full}(\Theta_t),
\end{equation}
while
\begin{equation}
    L_{\rm full}(\Theta_t+s^{\rm gm}_{b,t}q_{b,t})
    >
    L_{\rm full}(\Theta_t).
\end{equation}

\emph{Proof.}
The scalar restriction is
\begin{equation}
    h_b(s)=L_{\rm full}(\Theta_t+s q_{b,t}),
    \qquad 0\leq s\leq s^{\rm gm}_{b,t}.
\end{equation}
Then \(h_b'(0)=\langle \nabla L_{\rm full}(\Theta_t),q_{b,t}\rangle=-c_{b,t}\). Taylor's theorem with the upper curvature bound gives, for any \(s\in[0,s^{\rm gm}_{b,t}]\),
\begin{equation}
    h_b(s)-h_b(0)
    \leq
    -c_{b,t}s+\frac{\Lambda_{b,t}}{2}s^2 .
\end{equation}
If \(0\leq s\leq 2c_{b,t}/\Lambda_{b,t}\), the right side is non-positive. Therefore, \(s=s^{\rm ref}_{b,t}\) gives
\begin{equation}
    h_b(s^{\rm ref}_{b,t})\leq h_b(0).
\end{equation}
Taylor's theorem with the lower curvature bound gives
\begin{equation}
    h_b(s)-h_b(0)
    \geq
    -c_{b,t}s+\frac{\lambda_{b,t}}{2}s^2 .
\end{equation}
If \(s>2c_{b,t}/\lambda_{b,t}\), the right side is strictly positive. Therefore \(s=s^{\rm gm}_{b,t}\) gives
\begin{equation}
    h_b(s^{\rm gm}_{b,t})>h_b(0).
\end{equation}
Substituting the definition of \(h_b\) proves the two claims.
\hfill \(\square\)

This result shows why global matching can leave a failure mode. The scalar \(\gamma_t\) controls \(\|\Delta_t\|_2\), but it does not control the block lengths \(\|P_b\Delta_t\|_2\) relative to the local safe radii of different blocks. If the denominator underestimation is larger in one block than in another, then the globally matched Adam step must still over-update at least one active block. Blockwise or encoder-wise scale control can correct budget mismatches that a single global factor cannot remove.

\section{Why Does SGD Perform Well?}
\label{sec:why_does_sgd_succeed}

We now explain why plain SGD performs well in CLIP few-shot adaptation. We keep the same protocol as in Section~\ref{sec:why_does_adam_fail}: CLIP (ViT-B/16), $K{=}1$ shot per class, three datasets (EuroSAT, Oxford-Pets, and DTD), full-support updates (i.e., no mini-batch training unless explicitly mentioned), both encoders trainable, and all transformer blocks trainable. Let $\Theta_t \in \mathbb{R}^d$ denote the vector of trainable parameters at epoch $t$, where $d$ is the number of trainable parameters. Let $L_{\text{few}}, L_{\text{full}}, L_{\text{test}} : \mathbb{R}^d \to \mathbb{R}$ denote the support (few-shot subsample of the train-set), full train (full train-set), and test (full test-set) cross-entropy losses. 

\subsection{Does SGD travel into a broad basin or into a narrow support-fitting basin?}

\begin{figure*}[t]
    \centering
    \begin{subfigure}[t]{1.\linewidth}
        \centering
        \includegraphics[width=\linewidth]{figures/sgd_success/expA_eurosat_interpolation.pdf}
        \caption{EuroSAT.}
        \label{fig:sgd_success_expA:a}
    \end{subfigure}\hfill
    \begin{subfigure}[t]{1.\linewidth}
        \centering
        \includegraphics[width=\linewidth]{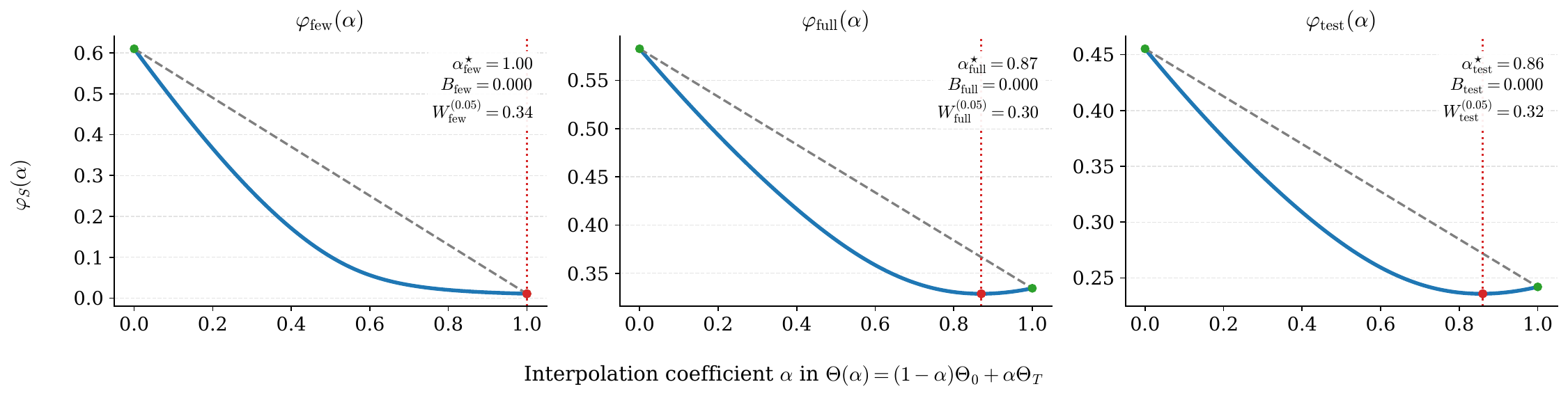}
        \caption{Oxford-Pets.}
        \label{fig:sgd_success_expA:b}
    \end{subfigure}\hfill
    \begin{subfigure}[t]{1.\linewidth}
        \centering
        \includegraphics[width=\linewidth]{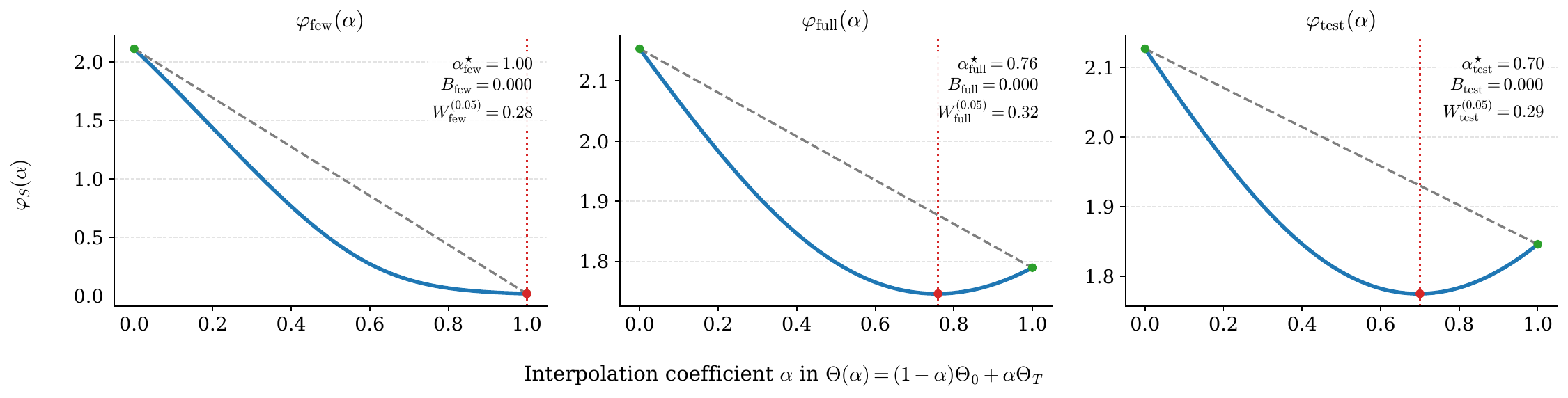}
        \caption{DTD.}
        \label{fig:sgd_success_expA:c}
    \end{subfigure}
    \caption{Experiment~A: one-dimensional interpolation between the pretrained checkpoint $\Theta_0$ and the final SGD solution $\Theta_T$. For each objective, we report the line minimum $\alpha_S^{\star}$, the barrier $B_S$ above the endpoint chord, and the normalized near-minimum width $W_{S}^{(0.05)}$. The support loss always reaches its minimum at $\alpha{=}1$ with zero barrier, and the near-minimum region remains broad on all three datasets. EuroSAT and Oxford-Pets keep the full-train and test minima close to the final SGD point, whereas DTD shows a later support-specialized tail along the interpolation direction.}
    \label{fig:sgd_success_expA}
\end{figure*}

Experiment~A answers this question by probing the one-dimensional loss geometry between the pretrained checkpoint and the final SGD solution. Let $T{=}30$ denote the final epoch of the shared training rule. Let $\Theta_0 \in \mathbb{R}^d$ denote the pretrained trainable state and let $\Theta_T \in \mathbb{R}^d$ denote the final SGD state, where $d$ is the number of trainable scalar parameters. The reported summaries use finite probe grids. The continuous definitions describe the underlying loss curves. We interpolate only the trainable parameters,
\begin{equation}
    \Theta(\alpha) = (1-\alpha)\Theta_0 + \alpha \Theta_T, \qquad \alpha \in [0,1].
\end{equation}
For each objective $L_S$ with $S \in \{\text{few},\text{full},\text{test}\}$, we define the scalar curve
\begin{equation}
    \varphi_S(\alpha) = L_S\!\left(\Theta(\alpha)\right), \qquad \varphi_S : [0,1] \to \mathbb{R},
\end{equation}
and its line minimum
\begin{equation}
    \alpha_S^{\star} \in \arg\min_{\alpha \in [0,1]} \varphi_S(\alpha), \qquad \varphi_S^{\star} = \varphi_S(\alpha_S^{\star}).
\end{equation}
The minimizer $\alpha_S^{\star}$ tells us \emph{where} the best point lies on the init-to-final line. However, it does not tell us whether this point sits inside a broad valley or at the tip of a narrow spike. To resolve this ambiguity, we measure two additional quantities.

First, we define the endpoint chord 
\begin{equation}
    \varphi_{S}^{\mathrm{lin}}(\alpha) = (1-\alpha)\varphi_S(0) + \alpha\varphi_S(1),
\end{equation}
and the interpolation barrier
\begin{equation}
    B_S = \max_{\alpha \in [0,1]} \Big(\varphi_S(\alpha) - \varphi_{S}^{\mathrm{lin}}(\alpha)\Big).
\end{equation}
The barrier measures whether the line from $\Theta_0$ to $\Theta_T$ crosses a ridge. If $B_S{=}0$, the line does not rise above the endpoint chord. This is evidence of a connected valley along that direction. A positive $B_S$ records an excursion above the endpoint chord. It does not alone establish separate basins.

Second, we define the $\delta_{\mathrm{w}}$-near-minimum set on the same line,
\begin{equation}
    \mathcal{S}_{S}^{(\delta_{\mathrm{w}})} = \Big\{\alpha \in [0,1] : \varphi_S(\alpha) \leq \varphi_S^{\star} + \delta_{\mathrm{w}}\big(\varphi_S(0)-\varphi_S^{\star}\big)\Big\},
\end{equation}
and its width
\begin{equation}
    W_{S}^{(\delta_{\mathrm{w}})} = \max \mathcal{S}_{S}^{(\delta_{\mathrm{w}})} - \min \mathcal{S}_{S}^{(\delta_{\mathrm{w}})}.
\end{equation}
We use $\delta_{\mathrm{w}}{=}0.05$ ($5\%$). Since $\alpha \in [0,1]$, the width is normalized by the full init-to-final path length. A small width means that the near-optimal region is concentrated around a very narrow interval. A large width means that the near-minimum set spans a substantial portion of the line. The span alone does not guarantee that the set is connected. Therefore, $W_{S}^{(0.05)}$ is a directional notion of basin width.

These two quantities are needed because they answer two different geometric questions. The barrier tests whether the SGD solution is connected to the pretrained initialization through a low-loss path. The width tests whether the low-loss region around the minimum is broad or fragile along that path. Together with $\alpha_S^{\star}$, they characterize broadness along the sampled interpolation direction.

The support near-minimum set spans a substantial part of the sampled interpolation direction. On all three datasets, the support curve is smooth, has zero barrier, and reaches its minimum at $\alpha_{\text{few}}^{\star}{=}1$. The corresponding widths $W_{\text{few}}^{(0.05)}$ are $0.38$ on EuroSAT, $0.34$ on Oxford-Pets, and $0.28$ on DTD. Thus, the final point already lies in a broad directional basin for the support objective. The full-train and test losses tell a more precise story. On EuroSAT and Oxford-Pets, their minima occur slightly before the final point, with $\alpha_{\text{full}}^{\star}, \alpha_{\text{test}}^{\star} \in [0.85,0.87]$, zero barrier, and widths between $0.30$ and $0.34$. Hence, the full-data near-minimum set spans a substantial part of the sampled line near the final SGD point. DTD is the only dataset where the optimum occurs clearly earlier, with $\alpha_{\text{full}}^{\star}{=}0.76$ and $\alpha_{\text{test}}^{\star}{=}0.70$. Even there, the corresponding widths remain broad, at $0.32$ and $0.29$. Thus, the measured DTD line has a broad near-minimum span, followed by a support-specialized tail. The checkpoint-projection analysis is consistent with this view. The orthogonal deviation from the init-to-final line is moderate in early training and falls below $0.1$ by epoch $20$ on all three datasets. Figure~\ref{fig:sgd_success_expA} summarizes these curves.

\begin{findingbox}[Conclusion]
The init-to-final probes show broad near-minimum spans along the sampled direction on all three datasets. EuroSAT and Oxford-Pets have zero full-train and test barriers on the interpolation grid. DTD shows a support-specialized tail in later epochs.
\end{findingbox}

\subsection{Measuring the local safe step radius of SGD}

\begin{figure*}[t]
    \centering
    \begin{subfigure}[t]{1.\linewidth}
        \centering
        \includegraphics[width=\linewidth]{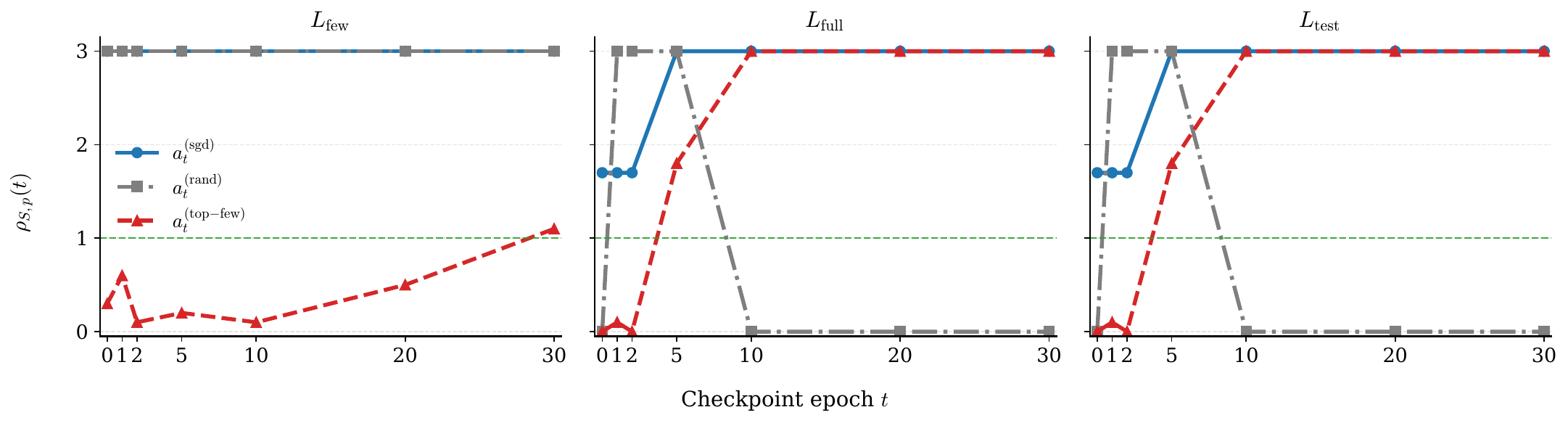}
        \caption{EuroSAT.}
        \label{fig:sgd_success_expB_radius:a}
    \end{subfigure}\hfill
    \begin{subfigure}[t]{1.\linewidth}
        \centering
        \includegraphics[width=\linewidth]{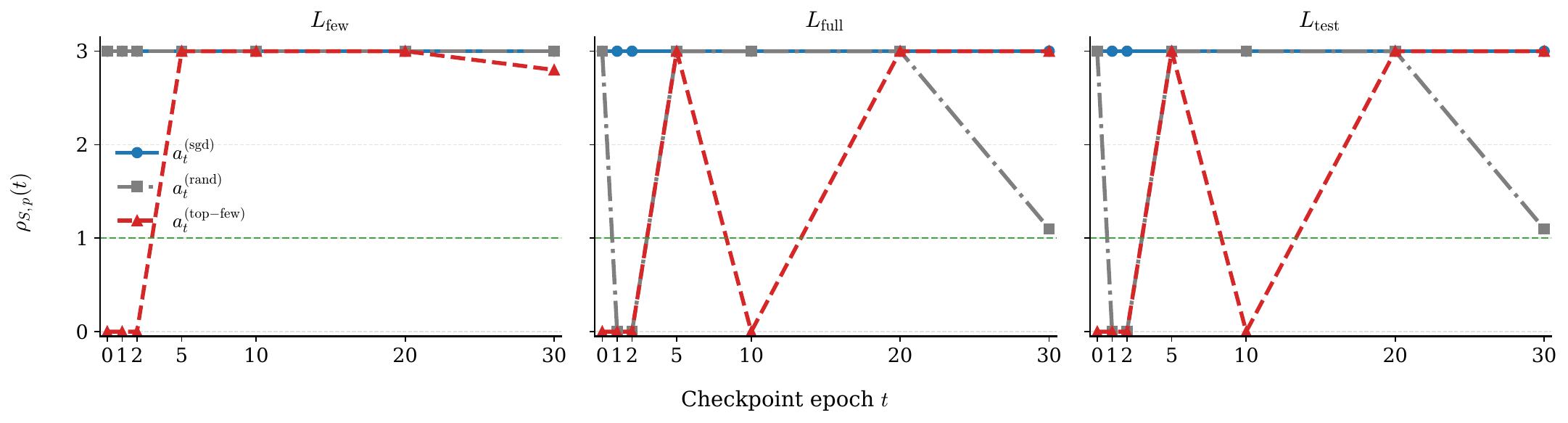}
        \caption{Oxford-Pets.}
        \label{fig:sgd_success_expB_radius:b}
    \end{subfigure}\hfill
    \begin{subfigure}[t]{1.\linewidth}
        \centering
        \includegraphics[width=\linewidth]{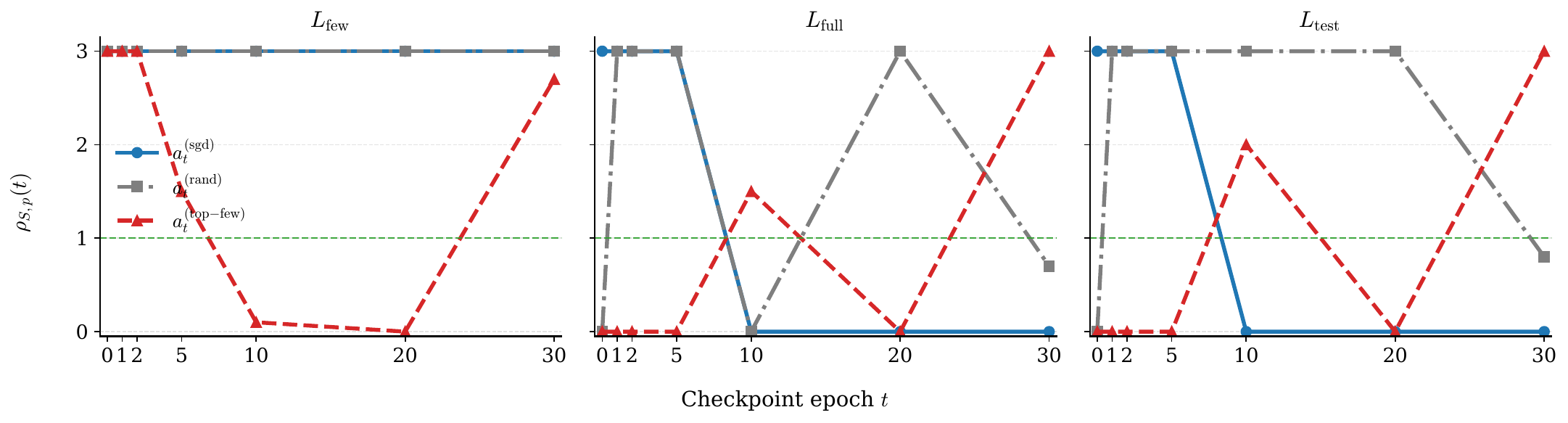}
        \caption{DTD.}
        \label{fig:sgd_success_expB_radius:c}
    \end{subfigure}
    \caption{Experiment~B: strict local safe step radius along three norm-matched directions. For a loss $L_{S}$ and a probe direction $a_t^{(p)}$, the plotted quantity is $\rho_{S,p}(t)=\max\{\alpha \in \mathcal C_B : L_{S}(\Theta_t + \alpha a_t^{(p)}) \le L_{S}(\Theta_t)\}$. The blue curve uses the actual SGD step, the gray curve uses a random direction, and the red curve uses the top eigenvector of the few-shot Hessian. Oxford-Pets is conservative throughout. EuroSAT becomes conservative after the first few epochs. DTD keeps a broad support valley, but the full-train and test safe radii of the SGD direction collapse from epoch $10$, which exposes late support-specific overfitting.}
    \label{fig:sgd_success_expB_radius}
\end{figure*}

\begin{figure*}[t]
    \centering
    \begin{subfigure}[t]{1.\linewidth}
        \centering
        \includegraphics[width=\linewidth]{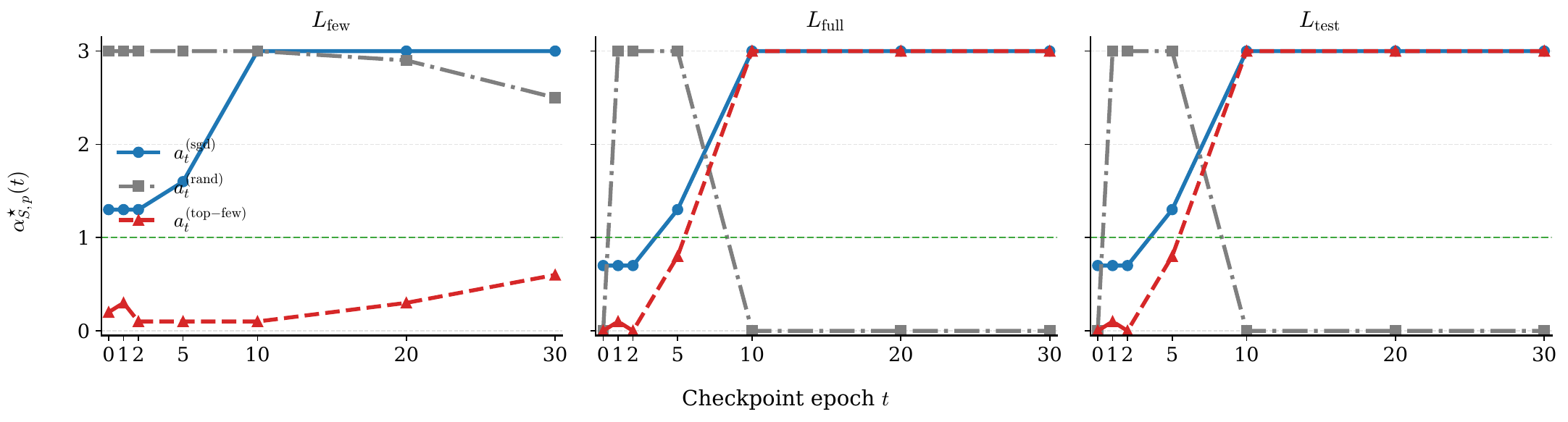}
        \caption{EuroSAT.}
        \label{fig:sgd_success_expB_minimizer:a}
    \end{subfigure}\hfill
    \begin{subfigure}[t]{1.\linewidth}
        \centering
        \includegraphics[width=\linewidth]{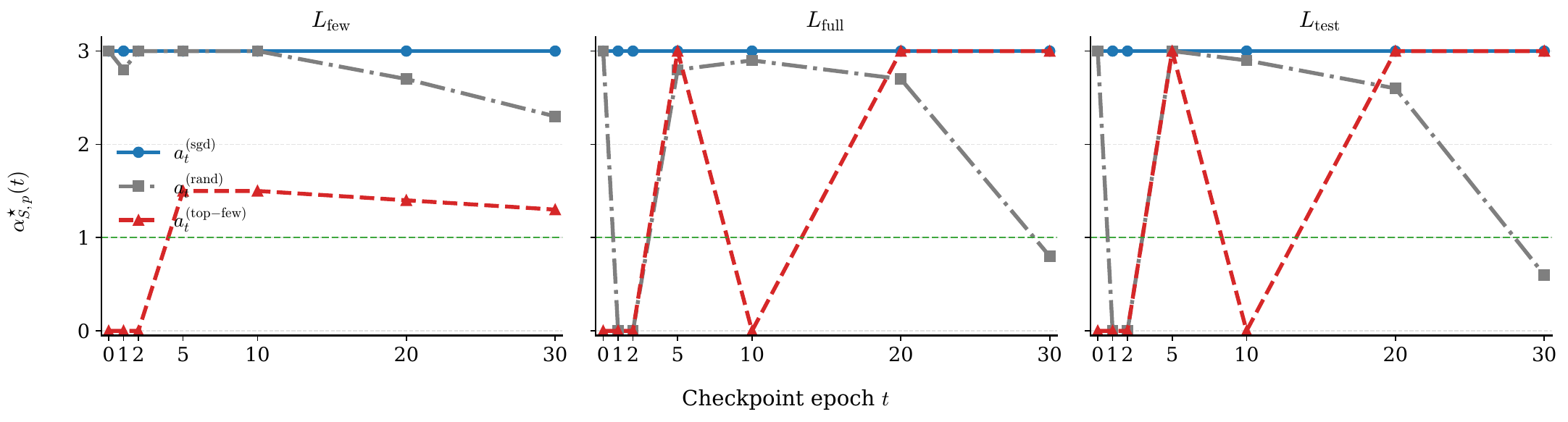}
        \caption{Oxford-Pets.}
        \label{fig:sgd_success_expB_minimizer:b}
    \end{subfigure}\hfill
    \begin{subfigure}[t]{1.\linewidth}
        \centering
        \includegraphics[width=\linewidth]{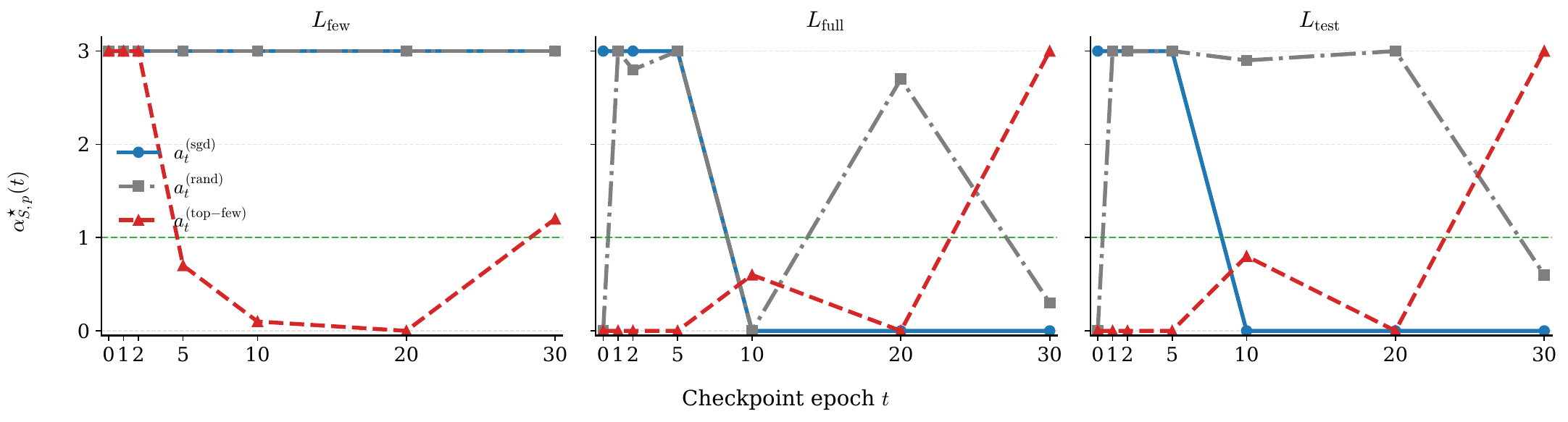}
        \caption{DTD.}
        \label{fig:sgd_success_expB_minimizer:c}
    \end{subfigure}
    \caption{Experiment~B: local minimizer multiplier along the same three norm-matched directions. The plotted quantity is $\alpha_{S,p}^{\star}(t)\in\arg\min_{\alpha\in\mathcal C_B} L_{S}(\Theta_t + \alpha a_t^{(p)})$. The boundary value $\alpha_{S,p}^{\star}(t)=3$ means that a minimum over the sampled nonnegative grid occurs at the right edge. Values below $1$ mean that the actual step is already larger than the locally preferred scale for that loss and direction. The minimizer plot complements the safe-radius plot: $\alpha_{S,p}^{\star}(t)$ identifies the preferred local scale, whereas $\rho_{S,p}(t)$ identifies the largest non-increasing scale.}
    \label{fig:sgd_success_expB_minimizer}
\end{figure*}

Experiment~B studies the local 1D landscape around the actual SGD update and compares it with two norm-matched control directions. Let $\Theta_t \in \mathbb{R}^{d}$ denote the trainable parameter vector at checkpoint $t$, where $d$ is the total number of trainable parameters. Let
\begin{equation}
    g_t = \nabla L_{\text{few}}(\Theta_t) \in \mathbb{R}^{d}, \qquad \Delta_t = -\eta g_t \in \mathbb{R}^{d},
\end{equation}
where $\eta>0$ is the learning rate. For probe family $p\in\{\mathrm{sgd},\mathrm{rand},\mathrm{top\text{-}few}\}$, the displacement $a_t^{(p)} \in \mathbb{R}^{d}$ satisfies $\lVert a_t^{(p)} \rVert_2 = \lVert \Delta_t \rVert_2$. We evaluate the 1D curve
\begin{equation}
    \phi_{S,p}(\alpha;t) = L_{S}(\Theta_t + \alpha a_t^{(p)}),
    \qquad \alpha \in [-1,3], \quad S \in \{\text{few},\text{full},\text{test}\}.
\end{equation}
We use three probe directions. The first is the actual SGD step $a_t^{(\mathrm{sgd})} = \Delta_t$. The second is a random norm-matched direction $a_t^{(\mathrm{rand})}$. It serves as an isotropic baseline. The third is the norm-matched top-Hessian direction $a_t^{(\mathrm{top\text{-}few})} = \lVert \Delta_t \rVert_2 u_t^{\mathrm{top\text{-}few}}$, where $u_t^{\mathrm{top\text{-}few}} \in \mathbb{R}^{d}$ is the dominant eigenvector of the few-shot Hessian $H_{\text{few}}(\Theta_t) = \nabla^2 L_{\text{few}}(\Theta_t) \in \mathbb{R}^{d\times d}$. It serves as a sharp-direction baseline. Because we use full-support SGD without momentum, $a_t^{(\mathrm{sgd})}$ is exactly the negative few-shot gradient direction.

The summaries use the nonnegative sampled grid $\mathcal C_B\subset[0,3]$, although the curves also include negative multipliers. For each loss $L_{S}$ and probe displacement $a_t^{(p)}$, we record a minimizer
\begin{equation}
    \alpha_{S,p}^{\star}(t) \in \arg\min_{\alpha \in \mathcal C_B} \phi_{S,p}(\alpha;t)
\end{equation}
and the largest local safe step radius
\begin{equation}
    \rho_{S,p}(t) = \max\{\alpha \in \mathcal C_B : \phi_{S,p}(\alpha;t) \le \phi_{S,p}(0;t)\}.
\end{equation}
The minimizer $\alpha_{S,p}^{\star}(t)$ measures the locally preferred step scale along a fixed direction. The radius $\rho_{S,p}(t)$ is the largest sampled nonnegative multiplier whose loss does not exceed the checkpoint loss. It does not certify that every intervening multiplier is safe or that the curve is monotone. When $\alpha_{S,p}^{\star}(t)=\rho_{S,p}(t)=3$, the boundary is both a sampled minimum and a safe sampled point. When both are zero, no positive sampled multiplier reduces the loss.

Figures~\ref{fig:sgd_success_expB_radius} and~\ref{fig:sgd_success_expB_minimizer} show a consistent pattern. On Oxford-Pets, the SGD direction is conservative throughout training. For all three losses and all logged checkpoints, the sampled minimum occurs at the right boundary, i.e., $\alpha_{S,\mathrm{sgd}}^{\star}(t)=3$, and the strict safe radius also satisfies $\rho_{S,\mathrm{sgd}}(t)=3$. Thus, the actual step $\alpha=1$ lies well inside a broad descending interval. This is visible already at initialization, where the full-train loss drops from $0.583$ to $0.499$ and the test loss drops from $0.455$ to $0.382$ after one SGD step. EuroSAT is slightly less conservative at the beginning. At $t=0$, the support curve is minimized at $\alpha_{\text{few},\mathrm{sgd}}^{\star}(0)=1.3$, while the full-train and test curves are minimized at $\alpha_{\text{full},\mathrm{sgd}}^{\star}(0)=\alpha_{\text{test},\mathrm{sgd}}^{\star}(0)=0.7$. The corresponding strict safe radii are still $\rho_{\text{full},\mathrm{sgd}}(0)=\rho_{\text{test},\mathrm{sgd}}(0)=1.7$, so the actual SGD step remains safely inside the valley. By $t=5$, the three minimizers move to the range $1.3$--$1.6$, and all three safe radii already reach $3$. From $t \ge 10$, both $\alpha_{S,\mathrm{sgd}}^{\star}(t)$ and $\rho_{S,\mathrm{sgd}}(t)$ saturate at $3$ for all three losses. Hence, SGD becomes clearly conservative once the trajectory enters the valley.

DTD reveals the failure mode that matters for stopping. Up to $t=5$, it behaves like Oxford-Pets and EuroSAT. The support, full-train, and test curves all satisfy $\alpha_{S,\mathrm{sgd}}^{\star}(t)=\rho_{S,\mathrm{sgd}}(t)=3$. At $t=10$, however, the support curve still has $\alpha_{\text{few},\mathrm{sgd}}^{\star}(10)=\rho_{\text{few},\mathrm{sgd}}(10)=3$, but the full-train and test curves collapse to $\alpha_{\text{full},\mathrm{sgd}}^{\star}(10)=\alpha_{\text{test},\mathrm{sgd}}^{\star}(10)=0$ and $\rho_{\text{full},\mathrm{sgd}}(10)=\rho_{\text{test},\mathrm{sgd}}(10)=0$. The actual SGD step still reduces the support loss from $0.237$ to $0.185$, but it slightly increases the full-train loss from $1.687$ to $1.692$ and the test loss from $1.712$ to $1.720$. The same pattern persists at $t=20$ and $t=30$. Therefore, late SGD steps on DTD are not locally unstable for fitting the support set. They are support-specific and no longer safe for the full-data objectives. This explains why SGD can reach a high peak on DTD yet remain highly sensitive to the stopping epoch.

The two control directions clarify why this experiment is needed. The random direction tests whether the local neighborhood is isotropically safe. It is usually almost flat. Its one-step losses stay extremely close to the baseline, and its strict safe radius sometimes flips between $0$ and $3$ because $\rho_{S,p}$ uses a non-increase criterion. The relaxed $1\%$ and $5\%$ radii remain close to $3$ on all three datasets, which confirms that the random line is mostly neutral rather than genuinely useful. The top-Hessian direction plays the opposite role. It probes the sharpest local mode of the few-shot landscape. On EuroSAT at $t=0$, one step with $a_t^{(\mathrm{top\text{-}few})}$ increases the full-train loss from $1.649$ to $2.468$ and the test loss from $1.659$ to $2.473$, with $\rho_{\text{full},\mathrm{top\text{-}few}}(0)=\rho_{\text{test},\mathrm{top\text{-}few}}(0)=0$. On Oxford-Pets, the same direction is also immediately unsafe at the first three checkpoints. These comparisons show that SGD does not perform well in few-shot adaptation because every direction of the same norm is safe. It performs well because the actual SGD direction stays inside a broad descent interval on datasets where the SGD axis remains aligned with the full-train and test objectives, whereas the collapse of that interval on DTD correctly exposes the onset of support-specific overfitting.

\begin{findingbox}[Conclusion]
With a learning rate of $10^{-4}$, SGD is usually locally conservative rather than unstable. Along the SGD step, EuroSAT and Oxford-Pets keep a broad safe descent region, meaning that the step could often be taken farther without increasing the full-train or test loss. DTD is an exception. After the early epochs, the support loss on DTD still favors larger steps, but the full-train and test losses no longer do. Thus, on challenging datasets, SGD may degrade late when the update becomes support-specific, not because the step is too large.
\end{findingbox}

\subsection{2D local surfaces around SGD checkpoints}

Experiment~B probed the SGD trajectory alone. Experiment~C extends this analysis to a two-dimensional neighborhood around an SGD checkpoint. The goal is to determine whether SGD lies inside a broad local corridor, whether there exists a sharp orthogonal ridge, whether the ridge is symmetric, and whether the few-shot SGD direction remains aligned with the full-data objectives.

At checkpoint $t$, let
\begin{equation}
    u_t^{\mathrm{sgd}} = -\frac{\nabla L_{\mathrm{few}}(\Theta_t)}{\lVert \nabla L_{\mathrm{few}}(\Theta_t) \rVert_2}
    \in \mathbb{R}^{d}
\end{equation}
be the normalized SGD direction. Let $p\in\{\mathrm{rand},\mathrm{adam},\mathrm{full},\mathrm{top\text{-}few}\}$ index the probe family. The corresponding raw probe direction is denoted by $q_t^{(p)}\in\mathbb{R}^{d}$ and is defined as follows:
\begin{equation}
    \begin{aligned}
    q_t^{(\mathrm{rand})}&=\xi_t^{\mathrm{rand}},
    &q_t^{(\mathrm{adam})}&=-\frac{m_{\mathrm{few},t}}{\sqrt{v_{\mathrm{few},t}+\varepsilon}},\\
    q_t^{(\mathrm{full})}&=-\nabla L_{\mathrm{full}}(\Theta_t),
    &q_t^{(\mathrm{top\text{-}few})}&=u_t^{\mathrm{top\text{-}few}}.
    \end{aligned}
\end{equation}
where $\xi_t^{\mathrm{rand}}\in\mathbb R^d$ is the raw random vector, $m_{\mathrm{few},t}=\nabla L_{\mathrm{few}}(\Theta_t)$, $v_{\mathrm{few},t}$ is the few-shot second moment used in the virtual Adam probe, $H_{\mathrm{few}}(\Theta_t)$ is the few-shot Hessian, and $u_t^{\mathrm{top\text{-}few}}$ is its dominant eigendirection. We orthogonalize the probe to the SGD axis through
\begin{equation}
    u_{\perp,t}^{(p)} = \frac{q_t^{(p)} - \langle q_t^{(p)},u_t^{\mathrm{sgd}}\rangle u_t^{\mathrm{sgd}}}{\left\lVert q_t^{(p)} - \langle q_t^{(p)},u_t^{\mathrm{sgd}}\rangle u_t^{\mathrm{sgd}} \right\rVert_2}
    \in \mathbb{R}^{d}.
\end{equation}
Let $s_t=\lVert \Delta_t\rVert_2$ be the norm of the actual SGD step at checkpoint $t$. Experiment~C evaluates the local two-dimensional surface
\begin{equation}
    \Psi_{S}^{(p)}(\alpha,\beta;t)
    =
    L_{S}\!\left(\Theta_t + s_t\big(\alpha u_t^{\mathrm{sgd}} + \beta u_{\perp,t}^{(p)}\big)\right),
    \qquad
    S\in\{\mathrm{few},\mathrm{full},\mathrm{test}\},
\end{equation}
with $(\alpha,\beta)\in[-2,2]^2$. Since the slice $\beta{=}0$ does not depend on the probe, we summarize the SGD corridor by
\begin{equation}
    \alpha_{S}^{\star}(t)
    \in
    \arg\min_{\alpha\in[-2,2]} \Psi_{S}^{(p)}(\alpha,0;t).
\end{equation}
We summarize the orthogonal probe by the signed loss changes at the positive and negative boundaries,
\begin{equation}
    \Delta_{S,\beta}^{(p,+)}(t)=\Psi_{S}^{(p)}(0,2;t)-\Psi_{S}^{(p)}(0,0;t),
    \qquad
    \Delta_{S,\beta}^{(p,-)}(t)=\Psi_{S}^{(p)}(0,-2;t)-\Psi_{S}^{(p)}(0,0;t),
\end{equation}
where negative values indicate a beneficial orthogonal move. Finally, we quantify the ridge strength by the local curvature ratio
\begin{equation}
    R_{S}^{(p)}(t)=\frac{\kappa_{S,\beta}^{(p)}(t)}{\kappa_{S,\alpha}(t)},
\end{equation}
where $\kappa_{S,\alpha}(t)$ and $\kappa_{S,\beta}^{(p)}(t)$ are the second finite differences of $\Psi_{S}^{(p)}$ at $(0,0)$ along the SGD axis and the probe axis, respectively. For positive SGD-axis curvature, a ratio above one indicates greater positive curvature along the probe. Ratios with nonpositive or near-zero denominators require separate interpretation.

\begin{figure*}[t]
    \centering
    \includegraphics[width=\linewidth]{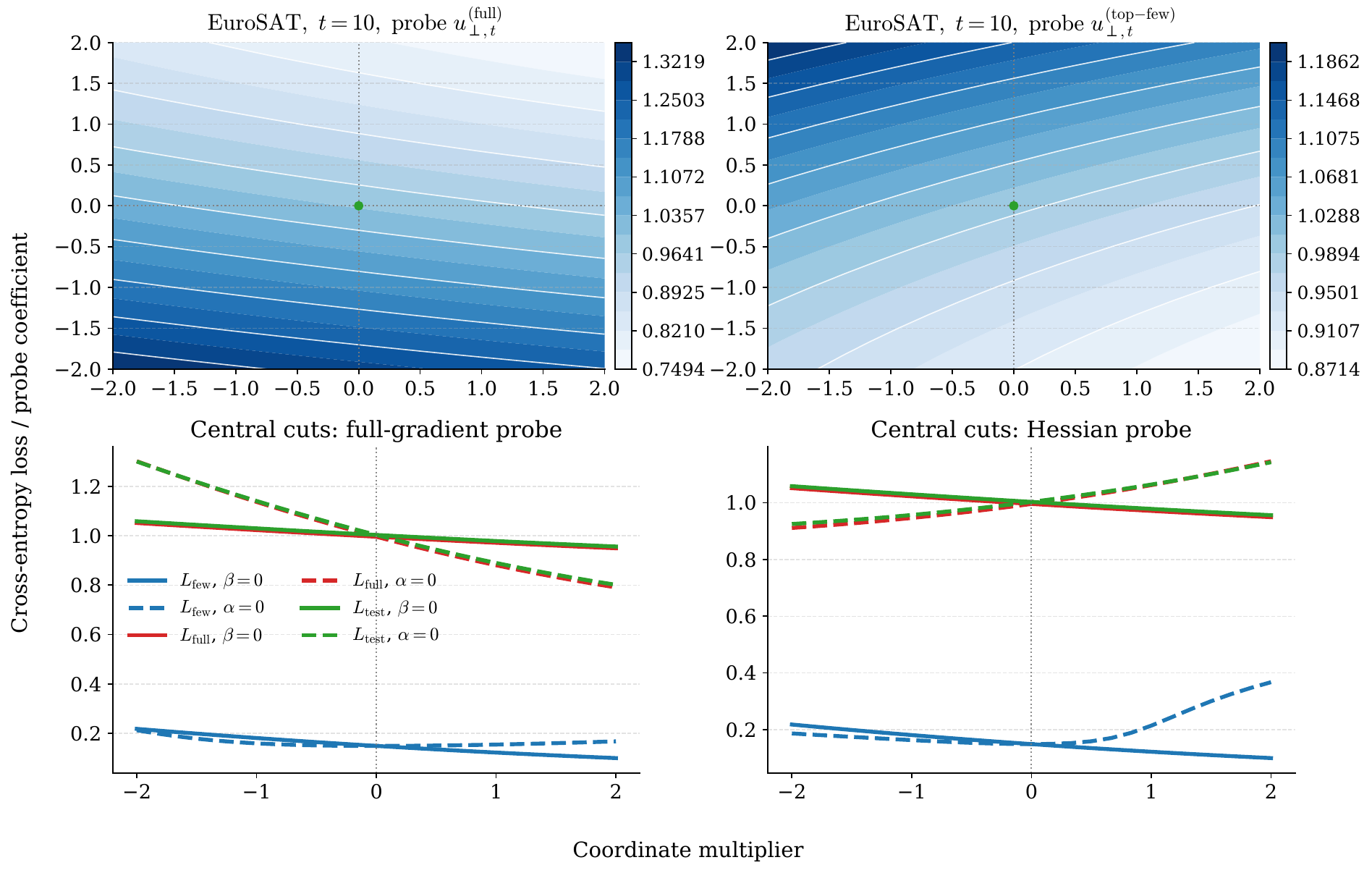}
    \caption{Experiment~C on EuroSAT at checkpoint $t{=}10$. Top: full-train loss contours in the plane spanned by the SGD direction and the orthogonalized full-gradient probe (left) or Hessian probe (right). Bottom: central cuts for $L_{\mathrm{few}}$, $L_{\mathrm{full}}$, and $L_{\mathrm{test}}$ along the SGD axis ($\beta{=}0$) and the probe axis ($\alpha{=}0$). The full-gradient probe exposes a generalization-improving orthogonal direction that the support loss does not favor, whereas the Hessian probe reveals a sharp asymmetric ridge orthogonal to SGD.}
    \label{fig:sgd_success_expC_representative}
\end{figure*}

The local geometry is highly anisotropic. Across the $84$ measured surfaces, the mean raw cosine $\langle q_t^{(p)},u_t^{\mathrm{sgd}}\rangle/\lVert q_t^{(p)}\rVert_2$ is essentially zero for the random probe, $0.55$ for the Adam-like probe, $0.33$ for the full-gradient probe, and $-0.03$ for the Hessian probe. Thus, the virtual Adam direction shares a substantial component with SGD, the full-gradient probe remains only partially aligned, and the Hessian probe is almost orthogonal. However, the orthogonal residuals of the random and Adam probes are nearly flat. On the full loss, the median ridge ratio is approximately $0$ for the random probe and $0.017$ for the Adam-like probe. The corresponding orthogonal loss changes at $\beta{=}+2$ have an average magnitude below $2.5\times 10^{-2}$. Therefore, SGD is not surrounded by a uniformly sharp bowl, and the non-SGD component of the Adam step does not reveal a useful new descent direction. This is consistent with the earlier second-moment analysis: Adam fails mainly by mis-scaling a largely shared descent direction rather than by exploiting a qualitatively different local geometry.

Figure~\ref{fig:sgd_success_expC_representative} shows a representative checkpoint. On EuroSAT at $t{=}10$, the SGD axis itself is a broad descending corridor. Along $\beta{=}0$, both $L_{\mathrm{full}}$ and $L_{\mathrm{test}}$ keep decreasing up to the right boundary, with $\alpha_{\mathrm{full}}^{\star}(10)=\alpha_{\mathrm{test}}^{\star}(10)=2$ and the corresponding loss changes at $\alpha{=}+2$ being $-0.0467$ on $L_{\mathrm{full}}$ and $-0.0468$ on $L_{\mathrm{test}}$. Hence, the checkpoint is not sitting at a narrow spike. It remains inside a stable valley aligned with generalization. Yet, the orthogonalized full-gradient probe reveals that this valley is not perfectly aligned with the support objective. At the same checkpoint, moving along the positive full-gradient probe decreases the full-train loss by $0.205$ and the test loss by $0.202$, while the support loss slightly increases by $0.018$. Thus, there exists a strong generalization-improving direction that is underweighted by the support loss. By contrast, the Hessian probe exposes a genuine orthogonal ridge: its curvature ratio is $R_{\mathrm{full}}^{(\mathrm{top\text{-}few})}(10)=7.54$ on the full loss and $R_{\mathrm{test}}^{(\mathrm{top\text{-}few})}(10)=7.12$ on the test loss. The ridge is asymmetric. On the full loss, the positive Hessian side increases the loss by $0.150$, whereas the opposite side decreases it by $0.085$. The first-order directional term can produce this asymmetry even in a quadratic local model.

\begin{figure*}[t]
    \centering
    \includegraphics[width=\linewidth]{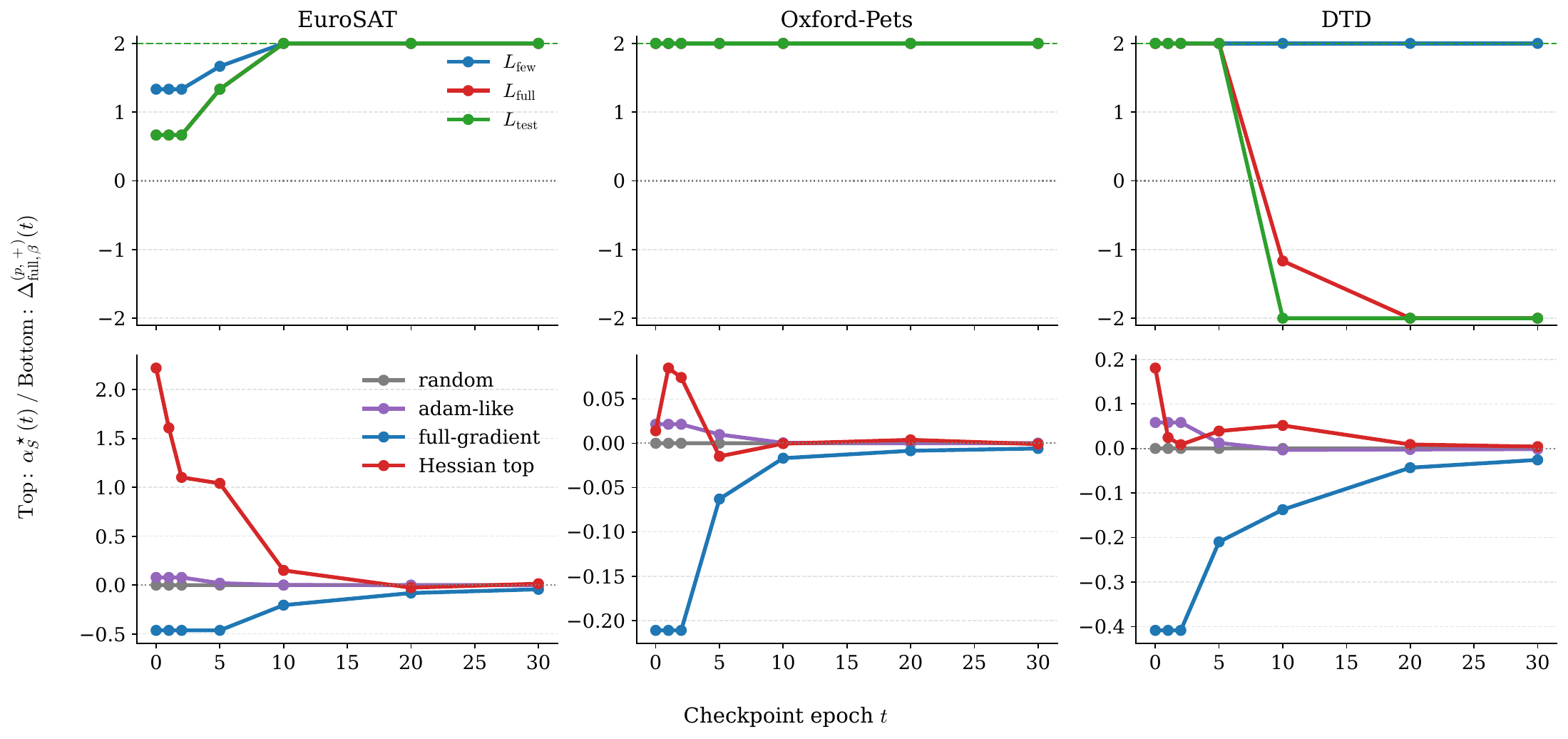}
    \caption{Experiment~C summary over checkpoints. Top: SGD-axis minimizer $\alpha_{S}^{\star}(t)$ for $L_{\mathrm{few}}$, $L_{\mathrm{full}}$, and $L_{\mathrm{test}}$. Bottom: orthogonal full-loss change $\Delta_{\mathrm{full},\beta}^{(p,+)}(t)$ at $\beta{=}+2$ for the four probes. Negative values are beneficial. SGD stays inside a descending corridor on EuroSAT and Oxford-Pets, whereas DTD becomes locally misaligned with the full-data objectives after $t{=}10$. The orthogonal full-gradient direction remains useful, while the orthogonal Adam-like residual stays nearly flat.}
    \label{fig:sgd_success_expC_axisprobe}
\end{figure*}

The full matrix confirms this picture. The local SGD corridor remains aligned with the full-train and test objectives on Oxford-Pets and EuroSAT. The top row of Figure~\ref{fig:sgd_success_expC_axisprobe} shows that $\alpha_{\mathrm{few}}^{\star}(t)$ stays positive on all $21$ checkpoints, while $\alpha_{\mathrm{full}}^{\star}(t)$ and $\alpha_{\mathrm{test}}^{\star}(t)$ stay positive on $18$ of the $21$ checkpoints. Oxford-Pets is the clearest case since all three minima remain at the right boundary, $\alpha_{S}^{\star}(t)=2$, for every checkpoint and every objective. EuroSAT is slightly more aggressive at initialization, with $\alpha_{\mathrm{full}}^{\star}(t)=\alpha_{\mathrm{test}}^{\star}(t)=0.67$ at $t\in\{0,1,2\}$, but these minima move to $2$ by $t{=}10$.  Hence, SGD lies in a broad corridor on both datasets. DTD is the exception. The support minimum remains at $2$ throughout training, but the full-train and test minima flip negative at $t{=}10$ and remain negative afterwards. At $t=30$, the positive SGD axis increases the full-train loss by $0.0047$ and the test loss by $0.0053$, while the negative side still decreases both losses. Therefore, the late SGD direction on DTD becomes support-specific even though it remains beneficial for the support objective.

The orthogonal probes refine the alignment story. The orthogonalized full-gradient direction is consistently informative. At $t{=}10$, moving along its positive side decreases the full-train loss by $0.0168$ on Oxford-Pets and by $0.137$ on DTD. Even at $t{=}30$, when DTD has already become support-specific along the SGD trajectory, the same probe still decreases the full-train and test losses by about $0.026$. Averaged across the full matrix, the positive full-gradient probe yields a mean full-loss change of $-0.217$ and a mean test-loss change of $-0.207$. Therefore, the local landscape still contains useful full-data descent directions even when the support direction becomes suboptimal. What limits SGD is not the absence of good local directions, but the fact that the support objective does not always track them. The Adam-like probe tells a different story. Its raw alignment with SGD is high, with a mean cosine $0.55$, but its orthogonal residual produces a mean full-loss change of only $0.024$ at $\beta{=}+2$. Hence, the part of the Adam direction that is not already captured by SGD is almost flat. This is strong geometric evidence against the view that Adam collapses because it follows a better or more informative local direction than SGD.

\begin{figure*}[t]
    \centering
    \includegraphics[width=\linewidth]{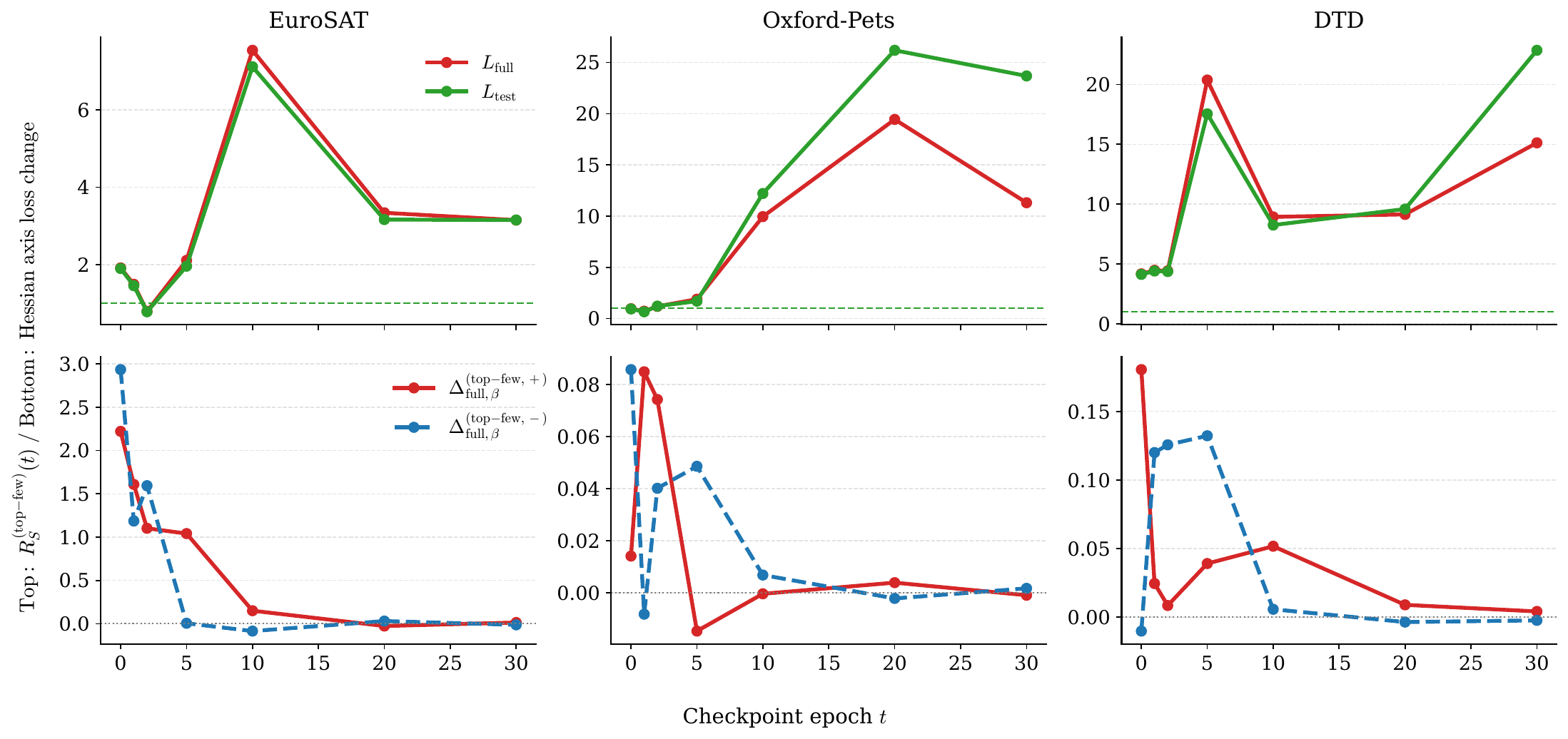}
    \caption{Experiment~C summary of the Hessian probe. Top: ridge ratio $R_{S}^{(\mathrm{top\text{-}few})}(t)$ on the full and test losses. Values above $1$ indicate that the orthogonal Hessian direction is sharper than the SGD axis. Bottom: full-loss changes $\Delta_{\mathrm{full},\beta}^{(\mathrm{top\text{-}few},+)}(t)$ and $\Delta_{\mathrm{full},\beta}^{(\mathrm{top\text{-}few},-)}(t)$ at $\beta{=}\pm 2$. Their separation measures local asymmetry of the ridge. The Hessian direction is typically orthogonal to SGD, much sharper than the SGD axis, and often strongly asymmetric.}
    \label{fig:sgd_success_expC_hessian}
\end{figure*}

The Hessian probe isolates the orthogonal ridge. Across the $21$ Hessian-probe surfaces, the median ridge ratio is $4.17$ on the full loss and $4.13$ on the test loss. The upper quartiles are $9.14$ and $9.59$, respectively. Hence, the Hessian direction is usually much sharper than the SGD corridor. Oxford-Pets and DTD contain the clearest late ridges. At $t{=}10$ on Oxford-Pets, the curvature ratios reach $9.95$ on the full loss and $12.21$ on the test loss. At $t{=}30$ on DTD, they increase further to $15.12$ and $22.85$. These large ratios show that the dangerous geometry is concentrated in a specific orthogonal direction rather than distributed isotropically around the SGD checkpoint. The bottom row of Figure~\ref{fig:sgd_success_expC_hessian} shows that the ridge is also asymmetric. On EuroSAT at $t{=}10$, the positive Hessian side increases the full loss by $0.150$, whereas the negative side decreases it by $0.085$. On DTD at $t{=}10$, the positive side increases the full loss by $0.052$, but the negative side is almost neutral, with a change of only $0.006$. The finite slice is tilted, with different loss changes on its two sides.

Experiment~C reveals why SGD performs well in few-shot adaptation from a geometric perspective. On EuroSAT and Oxford-Pets, the local SGD corridor remains aligned with the full-train and test objectives. The surrounding neighborhood is strongly anisotropic. Random directions are almost flat, and the orthogonal Adam residual is also almost flat. By contrast, the full-gradient probe reveals an orthogonal direction that improves the full-train and test losses more than the support loss, while the Hessian probe reveals a sharp asymmetric ridge. DTD exhibits the failure mode of this geometry: once the support objective becomes misaligned with the full-data objectives, the SGD corridor itself becomes support-specific. Therefore, SGD performs well in few-shot adaptation not because the local landscape is uniformly favorable, but because its update direction usually follows the correct local corridor while avoiding the sharp orthogonal ridge.

\begin{findingbox}[Conclusion]
The measured local slices around SGD contain broad anisotropic corridors. On EuroSAT and Oxford-Pets, SGD stays inside a valley that remains aligned with the full-train and test objectives. On DTD, this alignment breaks late in training and the SGD corridor becomes support-specific. The orthogonal full-gradient probe reveals a generalization-improving direction that the support loss underweights, while the orthogonal Hessian probe reveals a sharp asymmetric ridge. The orthogonal Adam-like residual is nearly flat. Hence, SGD performs well because it follows the right local corridor and avoids the orthogonal ridge, not because the landscape is uniformly flat.
\end{findingbox}
\subsection{How does curvature vary across directions?}

\begin{figure*}[t]
    \centering
    \begin{subfigure}[t]{1.\linewidth}
        \centering
        \includegraphics[width=\linewidth]{figures/sgd_success/expE_eurosat_directional.pdf}
        \caption{EuroSAT: directional curvature over time.}
        \label{fig:sgd_success_expE:a}
    \end{subfigure}\hfill
    \begin{subfigure}[t]{1.\linewidth}
        \centering
        \includegraphics[width=\linewidth]{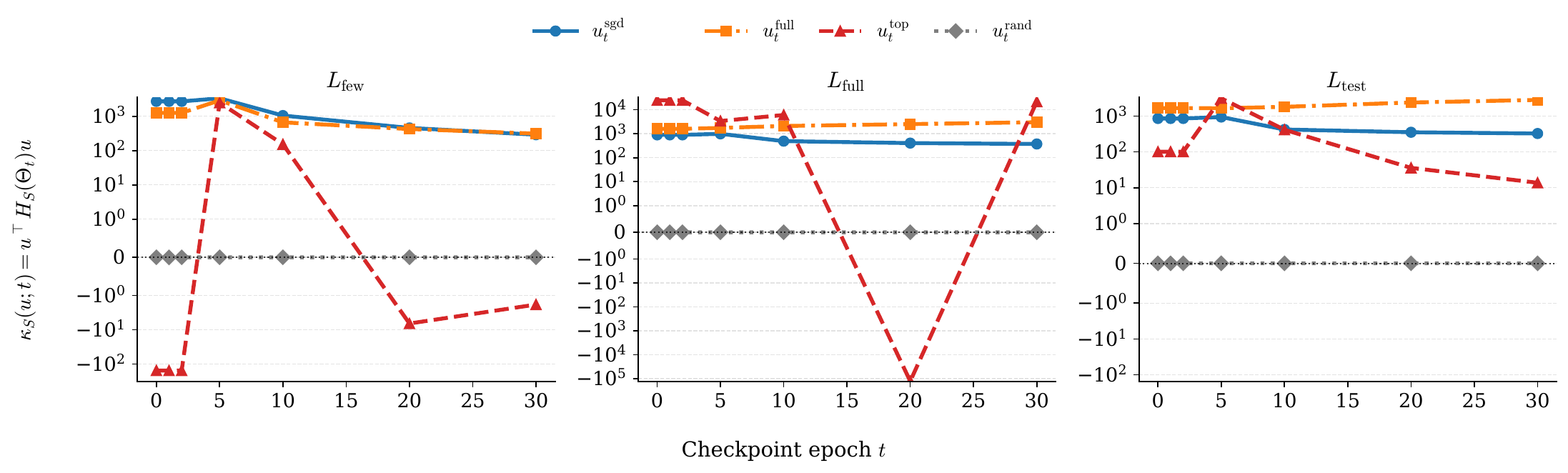}
        \caption{Oxford-Pets: directional curvature over time.}
        \label{fig:sgd_success_expE:b}
    \end{subfigure}\hfill
    \begin{subfigure}[t]{1.\linewidth}
        \centering
        \includegraphics[width=\linewidth]{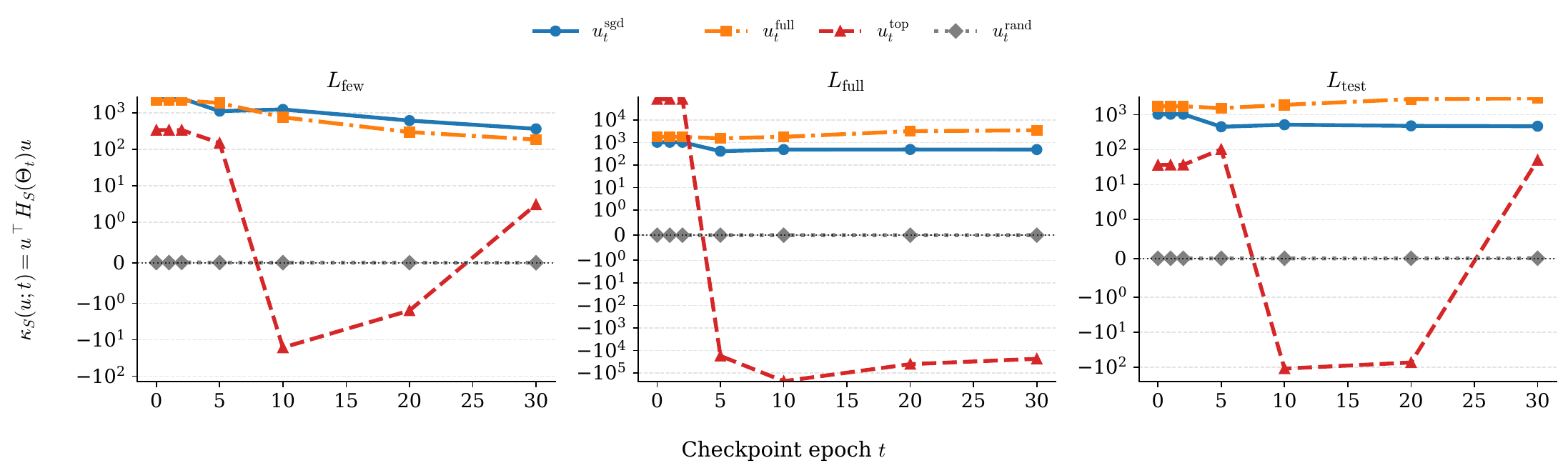}
        \caption{DTD: directional curvature over time.}
        \label{fig:sgd_success_expE:c}
    \end{subfigure}
    \caption{Experiment~D: directional curvature $\kappa_{S}(u;t)=u^{\top}H_{S}(\Theta_t)u$ along four unit directions, namely $u_t^{\mathrm{sgd}}$, $u_t^{\mathrm{full}}$, $u_t^{\mathrm{top}}$, and $u_t^{\mathrm{rand}}$. The vertical axis uses a symmetric logarithmic scale to visualize both large positive curvature and large negative curvature. Across datasets, the SGD direction has lower positive curvature than the full-train gradient direction in the reported comparisons. The dominant full-loss Hessian mode can have either positive or negative curvature.}
    \label{fig:sgd_success_expE_1}
\end{figure*}

\begin{figure*}
    \centering
    \begin{subfigure}[t]{1.\linewidth}
        \centering
        \includegraphics[width=\linewidth]{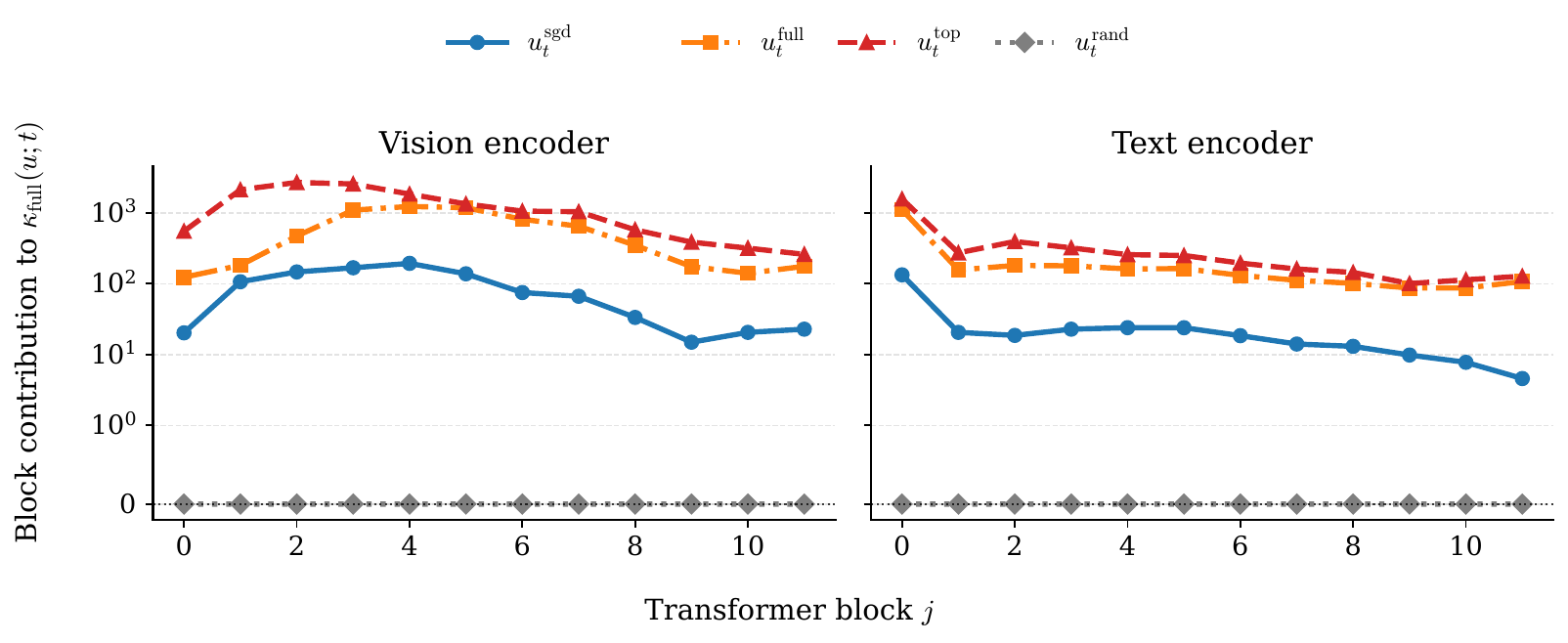}
        \caption{EuroSAT: full-loss block decomposition at $t{=}30$.}
        \label{fig:sgd_success_expE:d}
    \end{subfigure}\hfill
    \begin{subfigure}[t]{1.\linewidth}
        \centering
        \includegraphics[width=\linewidth]{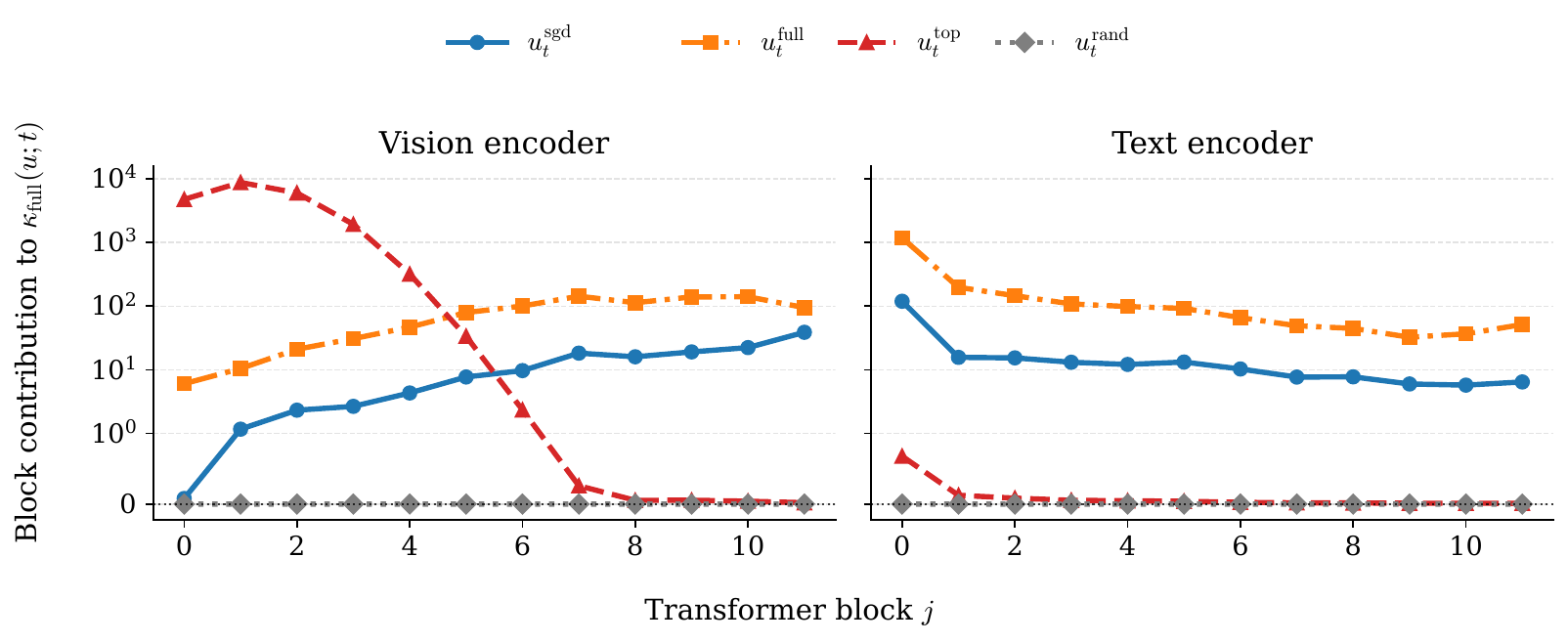}
        \caption{Oxford-Pets: full-loss block decomposition at $t{=}30$.}
        \label{fig:sgd_success_expE:e}
    \end{subfigure}\hfill
    \begin{subfigure}[t]{1.\linewidth}
        \centering
        \includegraphics[width=\linewidth]{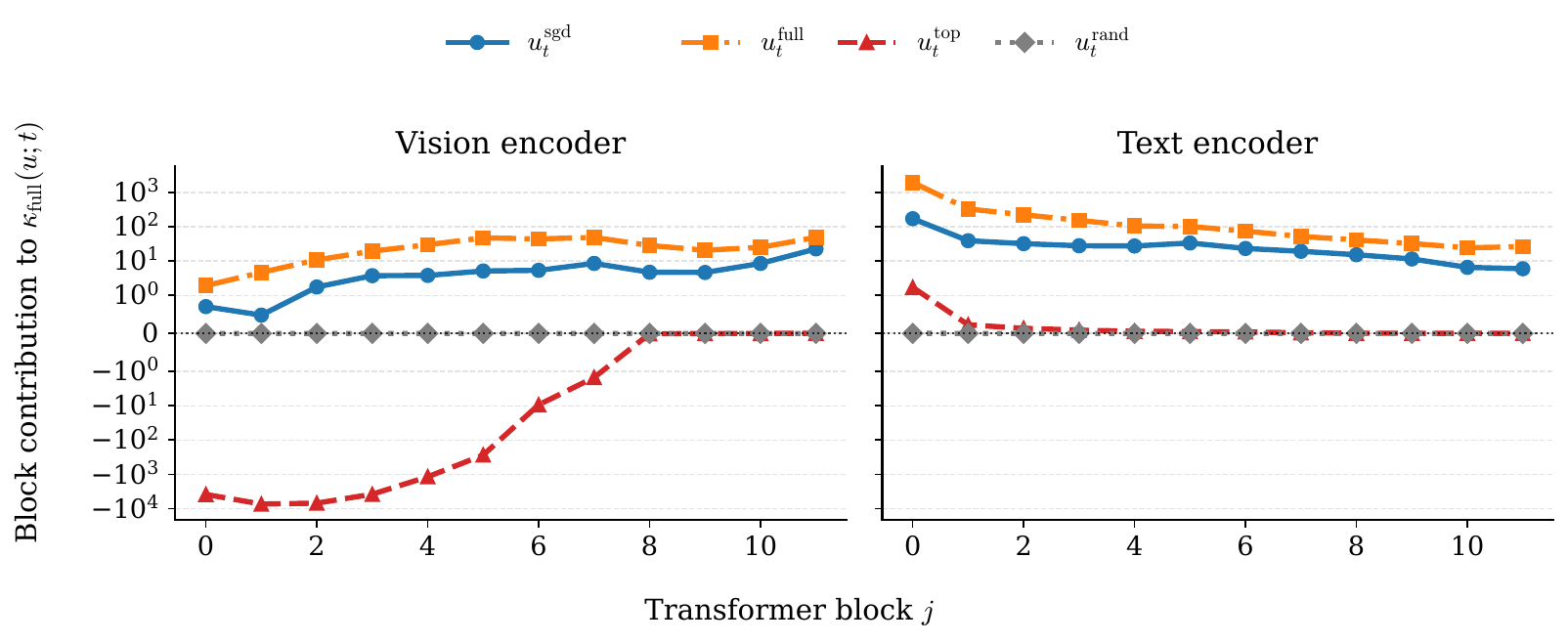}
        \caption{DTD: full-loss block decomposition at $t{=}30$.}
        \label{fig:sgd_success_expE:f}
    \end{subfigure}
    \caption{Experiment~D: additive block decomposition of $\kappa_{\mathrm{full}}(u;t)$ at the final checkpoint. For a direction $u$, the plotted quantity is the block contribution $\kappa_{\mathrm{full}}^{(b)}(u;t)$ of Transformer block $b$. The curvature contributions vary across blocks. Their distribution depends on the dataset and on the chosen direction family.}
    \label{fig:sgd_success_expE_2}
\end{figure*}

Experiment~D uses the Hessian to measure local directional curvature. The gradient only tells us the first-order descent direction. It does not tell us how sensitive the loss is to the step scale. This second-order sensitivity is encoded by the Hessian. For each objective $L_{S}:\mathbb{R}^{d}\to\mathbb{R}$ with $S\in\{\text{few},\text{full},\text{test}\}$, let
\begin{equation}
    H_{S}(\Theta_t)=\nabla^2L_{S}(\Theta_t)\in\mathbb{R}^{d\times d}
\end{equation}
denote the Hessian at checkpoint $\Theta_t\in\mathbb{R}^{d}$, where $d$ is the number of trainable parameters. For any unit direction $u\in\mathbb{R}^{d}$ with $\lVert u\rVert_2=1$, the second-order Taylor expansion along the line $\Theta_t+\alpha u$ is
\begin{equation}
    L_{S}(\Theta_t+\alpha u)
    =
    L_{S}(\Theta_t)
    + \alpha\,\langle \nabla L_{S}(\Theta_t),u\rangle
    + \frac{\alpha^2}{2}\,\kappa_{S}(u;t)
    + o(\alpha^2),
\end{equation}
where
\begin{equation}
    \kappa_{S}(u;t)=u^{\top}H_{S}(\Theta_t)u.
\end{equation}
The quantity $\kappa_{S}(u;t)$ is the directional curvature of $L_{S}$ along $u$. Large positive curvature increases the quadratic contribution to the loss change. Large negative curvature indicates local nonconvexity, but does not establish a saddle point without a stationary-point condition. Finite-step loss changes also depend on the linear term and step size. We use the Hessian of $L_{\mathrm{full}}$ to define the sharp-direction baseline and assess curvature on the full-data objective.

We compare four unit directions. The first is the actual SGD direction
\begin{equation}
    u_t^{\mathrm{sgd}} = \frac{\Delta_t}{\lVert \Delta_t\rVert_2},
    \qquad
    \Delta_t=-\eta\,\nabla L_{\mathrm{few}}(\Theta_t)\in\mathbb{R}^{d},
\end{equation}
where $\eta>0$ is the learning rate. Because we use full-support SGD without momentum, $u_t^{\mathrm{sgd}}$ is the negative normalized few-shot gradient direction; this direction is denoted by $u_t^{\mathrm{sgd}}$. Its sign does not affect directional curvature. The second direction is the normalized full-train gradient
\begin{equation}
    u_t^{\mathrm{full}} = \frac{\nabla L_{\mathrm{full}}(\Theta_t)}{\lVert \nabla L_{\mathrm{full}}(\Theta_t)\rVert_2}.
\end{equation}
The third direction is the dominant eigendirection $u_t^{\mathrm{top}}$ returned by power iteration on $H_{\mathrm{full}}(\Theta_t)$. Since the loss is non-convex, this direction captures the largest-magnitude full-loss curvature mode and can be associated with either positive or negative curvature. The fourth direction $u_t^{\mathrm{rand}}$ is a random unit vector. It serves as an isotropic baseline.

To localize the curvature contributions, we also decompose the full-loss curvature by the Transformer block. Let $\mathcal{I}_b\subset\{1,\dots,d\}$ denote the coordinates of block $b$. We define the additive block contribution
\begin{equation}
    \kappa_{\mathrm{full}}^{(b)}(u;t)
    =
    \sum_{i\in\mathcal{I}_b} u_i\,[H_{\mathrm{full}}(\Theta_t)u]_i,
    \qquad
    \kappa_{\mathrm{full}}(u;t)=\sum_b \kappa_{\mathrm{full}}^{(b)}(u;t).
\end{equation}
These coordinate contributions include cross-block Hessian terms through $H_{\mathrm{full}}u$. They are not the curvatures of isolated block updates.

Figures~\ref{fig:sgd_success_expE_1} and~\ref{fig:sgd_success_expE_2} compare the directional curvatures of the full and test objectives. Random directions stay near zero on all three datasets. Hence, curvature varies strongly with direction. On EuroSAT, the separation is already present at initialization and remains large at the end of training. At $t{=}30$, the full-loss curvature is $1317$ along $u_t^{\mathrm{sgd}}$, $9158$ along $u_t^{\mathrm{full}}$, and $18634$ along $u_t^{\mathrm{top}}$. The test-loss curvature has the same ordering, with values $1365$, $9267$, and $18230$. Thus, the SGD direction has lower positive curvature than both the full-gradient direction and the dominant full-Hessian mode at this checkpoint. The block decomposition shows that the largest full-loss curvature contributions are concentrated in middle vision blocks, especially blocks $3$--$5$, with an additional contribution from text block $0$. The SGD direction allocates much smaller curvature to these same blocks.

Oxford-Pets reveals a subtler phenomenon. At $t{=}30$, the full-loss curvature is $375$ along $u_t^{\mathrm{sgd}}$, $3028$ along $u_t^{\mathrm{full}}$, and $21842$ along $u_t^{\mathrm{top}}$. Hence, the SGD direction again has lower positive curvature on the full-data objective. However, the test-loss curvature tells a different story. The same checkpoint gives $327$ along $u_t^{\mathrm{sgd}}$, $2834$ along $u_t^{\mathrm{full}}$, but only $14$ along $u_t^{\mathrm{top}}$. Therefore, the dominant full-loss Hessian mode has low test-loss curvature at this checkpoint on Oxford-Pets. The block decomposition makes this separation explicit. The curvature contributions along $u_t^{\mathrm{full}}$ are concentrated in text block $0$ and in late vision blocks, whereas those along $u_t^{\mathrm{top}}$ are dominated by early vision blocks. Thus, the largest-magnitude full-loss curvature mode and the direction with the largest positive test curvature among these controls differ.

DTD is the clearest support-overfitting case. At $t{=}30$, the full-loss curvature is $486$ along $u_t^{\mathrm{sgd}}$, $3474$ along $u_t^{\mathrm{full}}$, and $-23437$ along $u_t^{\mathrm{top}}$. The large negative curvature along $u_t^{\mathrm{top}}$ reveals local nonconvexity of the full-loss landscape. This mode has much smaller curvature on the test loss. On the test loss, the same checkpoint yields $465$ along $u_t^{\mathrm{sgd}}$, $2918$ along $u_t^{\mathrm{full}}$, and only $50$ along $u_t^{\mathrm{top}}$. Hence, the full-gradient direction again has larger positive test curvature than the dominant full-Hessian mode on DTD. The block decomposition shows that $u_t^{\mathrm{full}}$ is concentrated in the early text blocks, whereas $u_t^{\mathrm{top}}$ is concentrated in the early vision blocks. This distinction is important. The larger positive curvature along the full-gradient direction is concentrated in text blocks, rather than in the blocks dominating the largest-magnitude Hessian mode.

Experiment~D, therefore, refines the landscape explanation of SGD. SGD does not perform well because the loss is uniformly flat. Its measured update direction has lower positive curvature than the full-gradient direction on the full and test objectives. The Hessian separates large full-loss curvature magnitude from positive test curvature. The dominant full-Hessian mode need not dominate the test Hessian. This directional separation also explains why later improvements should be structured and block-aware rather than isotropic.

\begin{findingbox}[Conclusion]
The SGD direction has lower positive curvature than the full-gradient direction in the reported comparisons. The dominant full-Hessian mode has large full-loss curvature magnitude, but can have smaller test curvature than SGD, as observed on Oxford-Pets and DTD. These diagnostics describe directional sensitivity rather than proving finite-step safety. Curvature contributions vary across blocks and datasets. %
\end{findingbox}

\subsection{Which blocks are safe to update?}

Experiment~E asks whether the useful progress of SGD is uniformly distributed across blocks or concentrated in a smaller subset. Let $g_t=\nabla L_{\mathrm{few}}(\Theta_t)\in\mathbb{R}^{d}$ be the full-support few-shot gradient at checkpoint $t$, where $d$ is the number of trainable parameters. Let $P_{b}\in\mathbb{R}^{d\times d}$ denote the diagonal projector onto block $b$. We define the block-restricted SGD direction
\begin{equation}
    a_t^{(b)} = -\eta \, P_{b} g_t \in \mathbb{R}^{d}.
\end{equation}
For a scalar multiplier $\alpha\in\mathbb{R}$ and an objective index $S\in\{\mathrm{few},\mathrm{full},\mathrm{test}\}$, we probe the one-dimensional block line
\begin{equation}
    \psi_{S}^{(b)}(\alpha;t)
    =
    L_{S}(\Theta_t + \alpha a_t^{(b)}).
\end{equation}
This is a blockwise version of the one-dimensional loss probes used in directional loss-landscape studies~\cite{goodfellow2015qualitatively,li2018visualizing}. The scalar $\alpha$ is dimensionless. The value $\alpha=1$ means \emph{one exact blockwise SGD step}: block $b$ is moved by the step that SGD would take at checkpoint $t$, while all other blocks are frozen. 

We convert these line probes into local block diagnostics through
\begin{equation}
    \Delta_{S}^{(b)}(t;\alpha)
    =
    \psi_{S}^{(b)}(\alpha;t)-\psi_{S}^{(b)}(0;t).
\end{equation}
The quantity $\Delta_{S}^{(b)}(t;1)$ is the local loss change produced by a one-unit blockwise SGD step on objective $L_S$. Negative values are locally beneficial. Positive values are locally harmful. We also define the blockwise full-loss minimizer
\begin{equation}
    \alpha_{\mathrm{full}}^{\star,(b)}(t)
    \in
    \arg\min_{\alpha\in[-2,2]} \psi_{\mathrm{full}}^{(b)}(\alpha;t),
\end{equation}
which specifies the locally preferred scale of that block direction. Values above $1$ indicate that the unit SGD step is conservative for block $b$. Values between $0$ and $1$ favor a smaller step; negative values favor the reverse direction.

The support--full mismatch from Eq.~\eqref{eq:sgd_short_block_mismatch} quantifies support specificity:
\begin{equation}
    \Gamma^{(b)}(t;\alpha)
    =
    \Delta_{\mathrm{full}}^{(b)}(t;\alpha)
    -
    \Delta_{\mathrm{few}}^{(b)}(t;\alpha).
\end{equation}
Large positive values mean that block $b$ helps the support objective more than the full-data objective. Unlike the line probe itself, $\Gamma^{(b)}$ is not a standard named metric in prior work. It is a diagnostic that we introduce for the few-shot setting. Using the checkpoint set $\mathcal{T}$ from the main text, the summaries in Eq.~\eqref{eq:sgd_short_block_averages} are
\begin{equation}
    \overline{\Delta}_{\mathrm{full}}^{(b)}(1)
    =
    \frac{1}{|\mathcal{T}|}\sum_{t\in\mathcal{T}} \Delta_{\mathrm{full}}^{(b)}(t;1),
    \qquad
    \overline{\Gamma}^{(b)}(1)
    =
    \frac{1}{|\mathcal{T}|}\sum_{t\in\mathcal{T}} \Gamma^{(b)}(t;1).
\end{equation}
The first quantity measures whether a unit blockwise SGD step is locally safe on the full-data objective. The second measures how strongly that same block step is biased toward the support objective.

\begin{figure*}[t]
    \centering
    \includegraphics[width=0.98\linewidth]{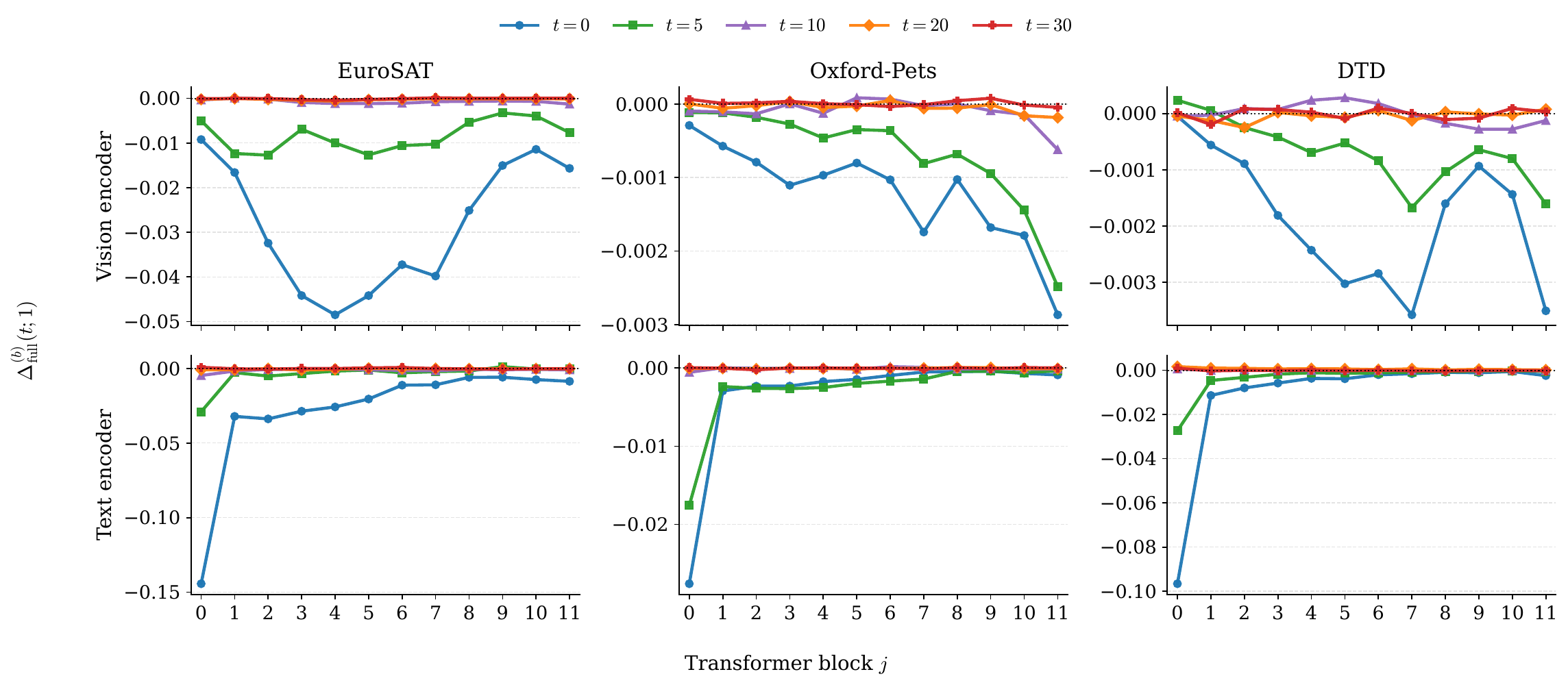}
    \caption{Experiment~E, epoch-resolved block safety. Each curve reports $\Delta_{\mathrm{full}}^{(b)}(t;1)$, the full-loss change after one exact blockwise SGD step at checkpoint $t$. Negative values indicate locally safe blocks. Early checkpoints contain many safe block directions. Later checkpoints shrink toward zero and become harmful on specific blocks.}
    \label{fig:sgd_success_expF_epoch_safety}
\end{figure*}

Figure~\ref{fig:sgd_success_expF_epoch_safety} shows the epoch-resolved values of $\Delta_{\mathrm{full}}^{(b)}(t;1)$. Early in training, most blockwise directions are safe. At $t=0$, many blocks have a negative full-loss change across datasets and encoders. Later in training, these improvements shrink. Some blocks cross zero and become harmful. This shows that block safety is not fixed during adaptation.

The temporal pattern is dataset dependent. On EuroSAT, middle vision blocks remain the most persistent safe region, especially blocks $3$--$5$. On Oxford-Pets, the early improvement is broad, but the landscape quickly becomes nearly flat. At late checkpoints, only a few blocks remain clearly beneficial. On DTD, the phase change is stronger. Some early text and mid/late vision updates are useful at the start, but several text directions become fragile at later checkpoints.

\begin{figure*}[t]
    \centering
    \includegraphics[width=0.98\linewidth]{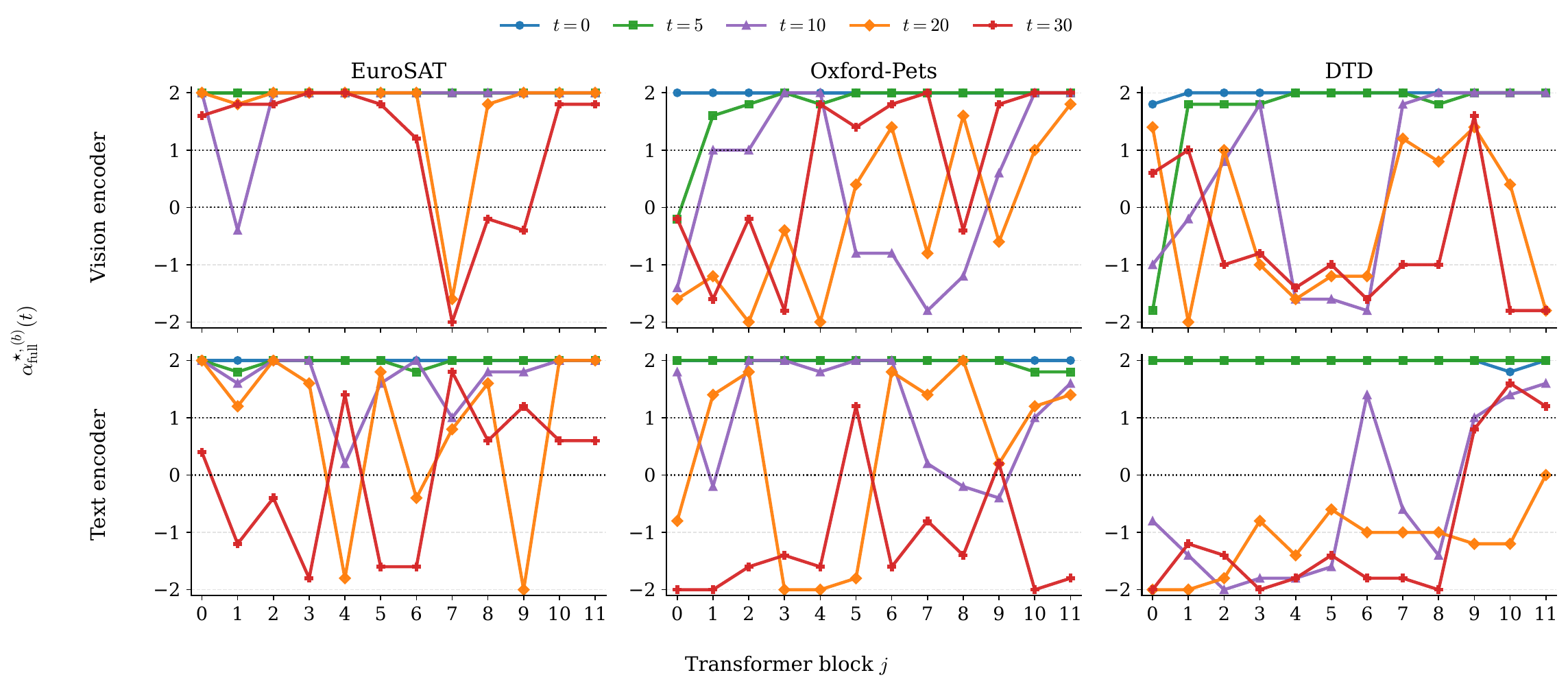}
    \caption{Experiment~E, epoch-resolved preferred block scale. Each curve reports $\alpha_{\mathrm{full}}^{\star,(b)}(t)$, the scale that minimizes the full-loss line probe along block $b$ at checkpoint $t$. The dotted horizontal line marks $\alpha=1$, which is the unit SGD block step. Values above $1$ indicate conservative steps. Values between $0$ and $1$ favor damping. Negative values favor the reverse direction on the sampled interval.}
    \label{fig:sgd_success_expF_epoch_scale}
\end{figure*}

Figure~\ref{fig:sgd_success_expF_epoch_scale} shows the preferred scale $\alpha_{\mathrm{full}}^{\star,(b)}(t)$. Early in training, many blocks have $\alpha_{\mathrm{full}}^{\star,(b)}(t)\approx 2$. The unit SGD step is, therefore, conservative for these blocks. Later in training, several blocks move below $\alpha_{\mathrm{full}}^{\star,(b)}(t)=1$. Some even obtain a negative preferred scale. Hence, safety and scale are distinct. A block can be useful at one checkpoint and require damping at another.

\begin{figure*}[t]
    \centering
    \includegraphics[width=0.98\linewidth]{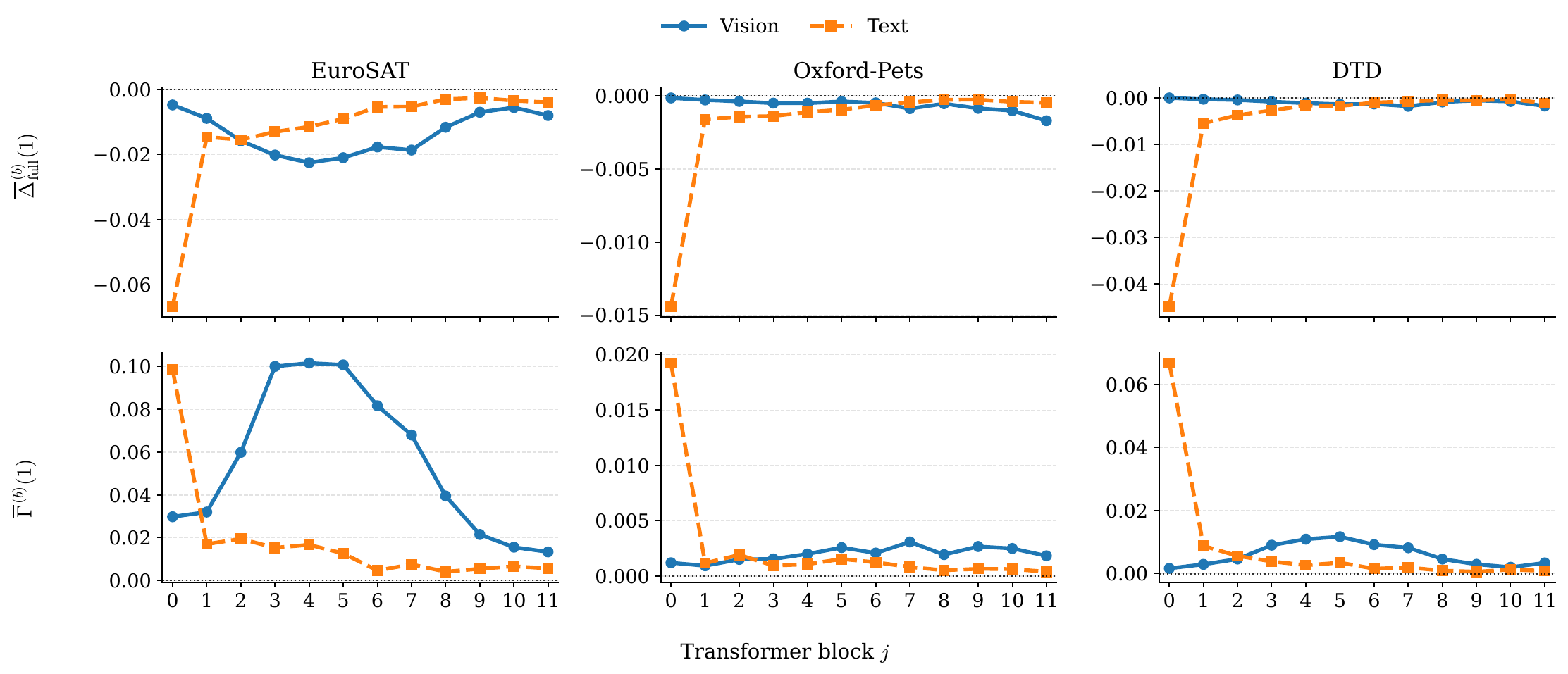}
    \caption{Experiment~E, checkpoint-averaged block diagnostics over $\mathcal{T}=\{0,1,2,5,10,20,30\}$. Top row: $\overline{\Delta}_{\mathrm{full}}^{(b)}(1)$, the mean full-loss change after one exact blockwise SGD step. More negative values indicate safer blocks. Bottom row: $\overline{\Gamma}^{(b)}(1)$, the mean support--full mismatch of the same block step. Larger positive values indicate stronger support bias. The useful SGD budget is structured by dataset, encoder, and depth.}
    \label{fig:sgd_success_expF}
\end{figure*}

After analyzing the temporal behavior, we summarize the persistent block structure by averaging over $\mathcal{T}$. Figure~\ref{fig:sgd_success_expF} reports $\overline{\Delta}_{\mathrm{full}}^{(b)}(1)$ and $\overline{\Gamma}^{(b)}(1)$. On EuroSAT, the safest vision blocks are the middle blocks $3$--$5$, with additional contributions from block $7$. The strongest text gains are concentrated in the early text blocks, especially block $0$. On Oxford-Pets, the safest vision blocks shift toward the late encoder, mainly blocks $10$--$11$. The useful text signal remains concentrated in blocks $0$--$3$. On DTD, the safe set is narrower. Reliable vision progress appears around blocks $5$--$7$ and $11$. Early text blocks provide strong initial gains, but these directions are also among the first to require damping later in training.

The support--full mismatch provides a second view of block reliability. On EuroSAT, several middle vision blocks are both useful and support-biased. They have negative $\overline{\Delta}_{\mathrm{full}}^{(b)}(1)$, but also positive $\overline{\Gamma}^{(b)}(1)$. Thus, these blocks help the full-data objective, but the support loss overstates their benefit. On DTD, text block $0$ is useful early, but later checkpoints often prefer a smaller or negative scale. Therefore, this block should not receive the same update budget throughout training.

The full and test diagnostics are nearly identical in our logs. We, therefore, report the full-loss summaries for compactness. Together, Figures~\ref{fig:sgd_success_expF_epoch_safety}--\ref{fig:sgd_success_expF} show that dense SGD progress is concentrated in a structured subset of blocks. They also show that the preferred update scale changes during training. %

\begin{findingbox}[Conclusion]
Early in training, most blocks are safe, and the unit SGD step is often conservative. Later in training, safety becomes block-dependent. Vision contains robust middle- or late-stage blocks, whereas text contains strong early directions that can become fragile. Safety and scale are distinct. A block may reduce the full-data loss and still require damping because of support bias or a preferred scale below one. %
\end{findingbox}

\subsection{Do SGD solutions across seeds belong to a broad connected basin?}

\begin{figure*}[t]
    \centering
    \begin{subfigure}[t]{1.0\linewidth}
        \centering
        \includegraphics[width=0.98\linewidth]{figures/sgd_success/expG_eurosat_connectivity.pdf}
        \caption{EuroSAT.}
        \label{fig:sgd_success_expG:a}
    \end{subfigure}\hfill
    \begin{subfigure}[t]{1.0\linewidth}
        \centering
        \includegraphics[width=0.98\linewidth]{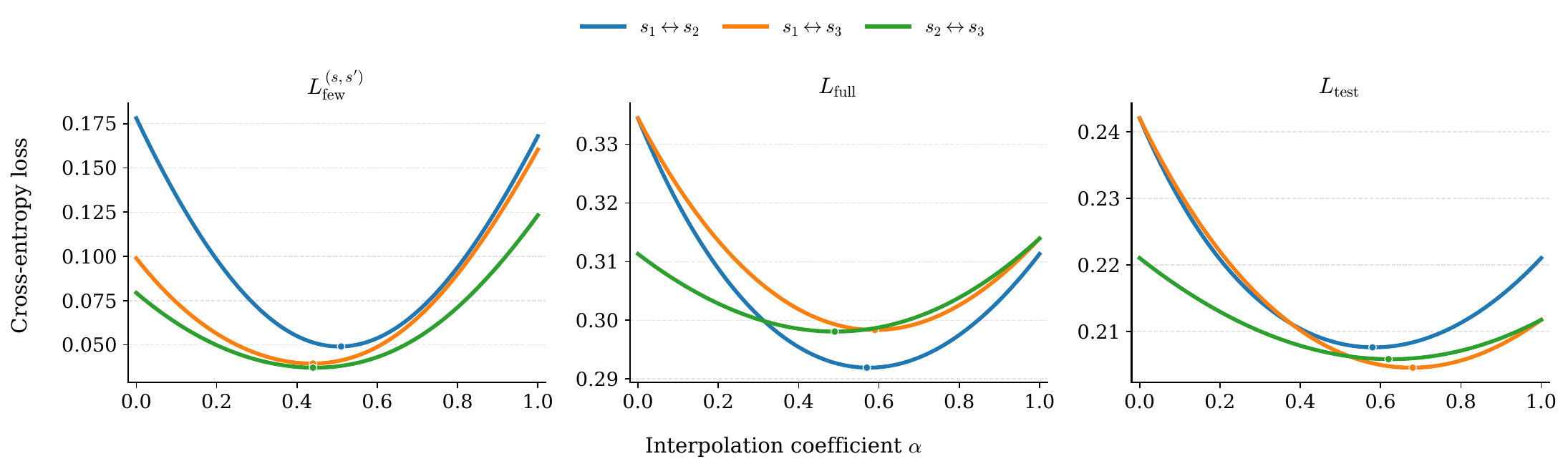}
        \caption{Oxford-Pets.}
        \label{fig:sgd_success_expG:b}
    \end{subfigure}\hfill
    \begin{subfigure}[t]{1.0\linewidth}
        \centering
        \includegraphics[width=0.98\linewidth]{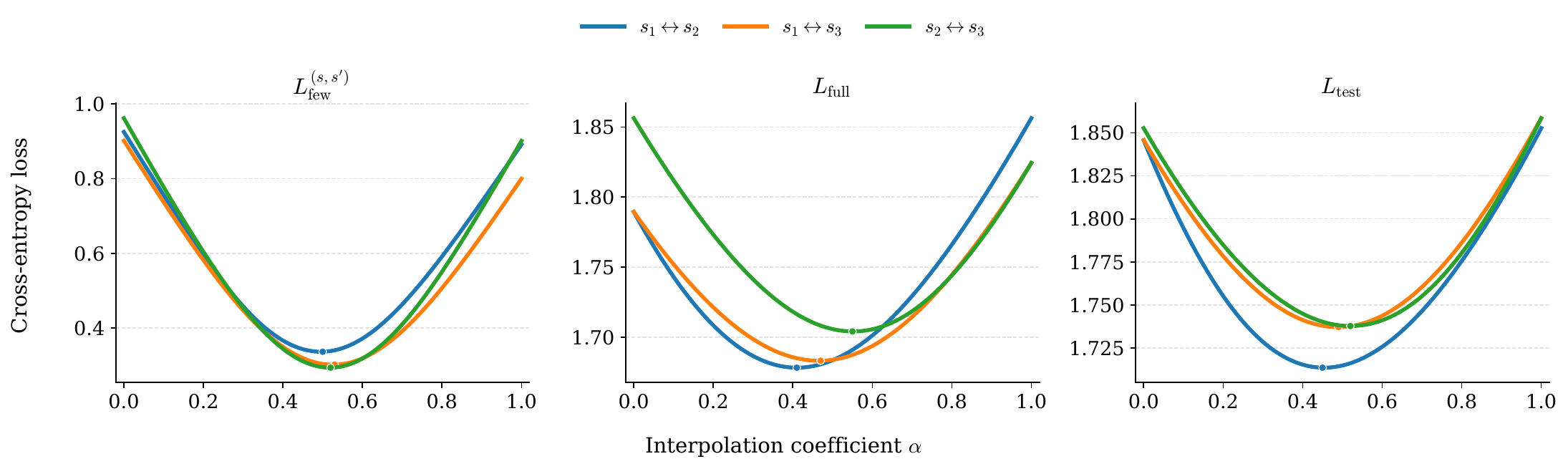}
        \caption{DTD.}
        \label{fig:sgd_success_expG:c}
    \end{subfigure}
    \caption{Experiment~F: pairwise interpolation between final SGD solutions $\Theta_T^{(s)}$ and $\Theta_T^{(s')}$ obtained from different few-shot seeds. The left panel reports the pair-support loss $L_{\mathrm{few}}^{(s,s')}$, while the middle and right panels report $L_{\mathrm{full}}$ and $L_{\mathrm{test}}$. For all tested seed pairs and datasets, the full-train and test minima lie in the interior of the segment, and their measured barriers are zero. These sampled segments are consistent with low-loss connectivity across the tested seeds.}
    \label{fig:sgd_success_expG}
\end{figure*}

Experiment~F tests low-loss connectivity along straight segments between final SGD solutions obtained from different few-shot tasks. Linear interpolation between independently trained solutions is a standard diagnostic for barrier-free connectivity and basin geometry \citep{goodfellow2015qualitatively,li2018visualizing}. Let $\mathcal{S}_{\mathrm{seed}}=\{1,2,3\}$ denote the set of few-shot seeds. For each seed $s\in\mathcal{S}_{\mathrm{seed}}$, let $\Theta_T^{(s)}\in\mathbb{R}^{d}$ be the final trainable state after $T=30$ epochs, where $d$ is the number of trainable parameters. For each seed pair $(s,s')\in\mathcal{S}_{\mathrm{seed}}^2$ with $s<s'$, we interpolate the trainable parameters,
\begin{equation}
    \Theta^{(s,s')}(\alpha)
    =
    (1-\alpha)\Theta_T^{(s)} + \alpha\Theta_T^{(s')},
    \qquad \alpha\in[0,1].
\end{equation}
Because different seeds correspond to different one-shot support sets, the support objective along the segment must be pair-specific. We define the pair-support loss
\begin{equation}
    L_{\mathrm{few}}^{(s,s')}(\Theta)
    =
    \frac{1}{2}\Big(L_{\mathrm{few}}^{(s)}(\Theta)+L_{\mathrm{few}}^{(s')}(\Theta)\Big).
\end{equation}
We then define a unified objective family indexed by $S\in\{\mathrm{few},\mathrm{full},\mathrm{test}\}$,
\begin{equation}
    L_S^{(s,s')}(\Theta)
    =
    \begin{cases}
        L_{\mathrm{few}}^{(s,s')}(\Theta), & S=\mathrm{few},\\
        L_{\mathrm{full}}(\Theta), & S=\mathrm{full},\\
        L_{\mathrm{test}}(\Theta), & S=\mathrm{test},
    \end{cases}
\end{equation}
and its interpolation curve
\begin{equation}
    \varphi_S^{(s,s')}(\alpha)
    =
    L_S^{(s,s')}\!\left(\Theta^{(s,s')}(\alpha)\right),
    \qquad \varphi_S^{(s,s')} : [0,1]\to\mathbb{R}.
\end{equation}
Using the same line-minimum, barrier, and width diagnostics as in Experiment~A, we compute $\alpha_{S}^{\star,(s,s')}$, $B_{S}^{(s,s')}$, and $W_{S}^{(0.05),(s,s')}$ on each pairwise curve. We also record the relative endpoint separation
\begin{equation}
    D_{\mathrm{rel}}^{(s,s')}
    =
    \frac{\lVert\Theta_T^{(s)}-\Theta_T^{(s')}\rVert_2}{\tfrac12\big(\lVert\Theta_T^{(s)}-\Theta_0\rVert_2 + \lVert\Theta_T^{(s')}-\Theta_0\rVert_2\big)}.
\end{equation}
This quantity is important because a zero barrier is only informative when the endpoints are not nearly identical.

Across all three datasets and all three seed pairs, the measured full-train and test barriers are zero on the interpolation grid: $B_{\mathrm{full}}^{(s,s')}=B_{\mathrm{test}}^{(s,s')}=0$. Thus, none of the sampled points exceeds the endpoint chord for either objective. The endpoint separation satisfies $D_{\mathrm{rel}}^{(s,s')}\in[1.13,1.15]$ on EuroSAT, $[1.05,1.08]$ on Oxford-Pets, and $[1.15,1.20]$ on DTD. Thus, the compared endpoints are separated on the same scale as their displacement from initialization.

EuroSAT provides clear evidence of low-loss connectivity along the sampled segments. The full-train minima occur at $\alpha_{\mathrm{full}}^{\star,(s,s')}\in\{0.33,0.33,0.57\}$ and the test minima at $\alpha_{\mathrm{test}}^{\star,(s,s')}\in\{0.34,0.33,0.55\}$. Therefore, the best points lie strictly inside the segment, not at the endpoints. The $5\%$ widths remain broad, with $W_{\mathrm{full}}^{(0.05),(s,s')}$ and $W_{\mathrm{test}}^{(0.05),(s,s')}$ between $0.21$ and $0.29$. Interpolation also improves the best endpoint by about $0.067$ in full-train loss and $0.069$ in test loss on average. Thus, the tested seed pairs have broad near-minimum spans along their interpolation directions.

Oxford-Pets shows the same qualitative pattern, although the gain from interpolation is smaller. The full-train minima occur at $\alpha_{\mathrm{full}}^{\star,(s,s')}\in[0.49,0.59]$ and the test minima at $\alpha_{\mathrm{test}}^{\star,(s,s')}\in[0.58,0.68]$. The $5\%$ widths are again substantial, with $W_{\mathrm{full}}^{(0.05),(s,s')}\in[0.22,0.27]$ and $W_{\mathrm{test}}^{(0.05),(s,s')}\in[0.27,0.32]$. The mean improvement below the best endpoint is about $0.016$ on the full objective and $0.009$ on the test objective. Hence, Oxford-Pets also supports low-loss connectivity along the sampled segments, with smaller differences in endpoint quality.

DTD is the strongest stress test. Even there, all full and test barriers remain zero. The minima stay interior, with $\alpha_{\mathrm{full}}^{\star,(s,s')}\in[0.41,0.55]$ and $\alpha_{\mathrm{test}}^{\star,(s,s')}\in[0.45,0.52]$. The low-loss corridor is somewhat narrower than on EuroSAT and Oxford-Pets, but it is still broad: $W_{\mathrm{full}}^{(0.05),(s,s')}\in[0.20,0.22]$ and $W_{\mathrm{test}}^{(0.05),(s,s')}\in[0.20,0.21]$. The interpolation gain is the largest here, about $0.113$ in full-train loss and $0.119$ in test loss on average. Thus, even on the most overfitting-prone of the three datasets, the sampled segments between independently trained SGD endpoints have zero measured barriers and broad near-minimum spans.

The pair-support objective is useful but secondary here because it averages two different one-shot tasks. The decisive evidence comes from $L_{\mathrm{full}}$ and $L_{\mathrm{test}}$. Figure~\ref{fig:sgd_success_expG} shows the pairwise interpolation curves. The minima of the full and test curves consistently lie in the interior of the segment, while the corresponding barriers are zero. The measured straight segments are consistent with connected low-loss regions across the tested seeds. This complements Experiment~A: broad near-minimum spans occur along both the init-to-final and seed-to-seed interpolation directions. These probes do not establish broadness in unmeasured directions.

\begin{findingbox}[Conclusion]
The measured straight segments are consistent with low-loss connectivity across the tested few-shot seeds. For every tested seed pair and dataset, the full-train and test barriers are zero on the interpolation grid. Their minima lie inside the segment. The relative endpoint separation remains nontrivial. The near-minimum spans are broad along the probed directions, and interpolation often achieves lower loss than the best endpoint. These observations do not establish the geometry in unmeasured directions.
\end{findingbox}

\subsection{Does SGD stay in regions where the support loss and the full-train loss are still geometrically aligned?}
\label{subsec:sgd_support_full_geometry_alignment}
Experiment~G asks whether SGD performs well in few-shot adaptation because the support objective remains a faithful local proxy of the full-train objective. If this hypothesis were true throughout training, then the support and full losses would stay aligned at three levels: gradient direction, dominant local curvature, and local two-dimensional surface shape. We, therefore, measure all three at the same checkpoints $t\in\mathcal{T}=\{0,1,2,5,10,20,30\}$ of the SGD trajectory.

Let $e\in\{\mathrm{vision},\mathrm{text}\}$ index the encoder. Let $d_e$ be the number of trainable parameters in the encoder $e$, with $d_{\mathrm{vision}}+d_{\mathrm{text}}=d$. At checkpoint $\Theta_t\in\mathbb{R}^{d}$, we define the encoder-restricted gradients
\begin{equation}
    m_{\mathrm{few},t}^{(e)},\; m_{\mathrm{full},t}^{(e)} \in \mathbb{R}^{d_e}.
\end{equation}
We quantify first-order agreement by
\begin{equation}
    A_g^{(e)}(t)
    =
    \cos\!\big(m_{\mathrm{few},t}^{(e)}, m_{\mathrm{full},t}^{(e)}\big),
    \qquad
    R_m^{(e)}(t)
    =
    \frac{\lVert m_{\mathrm{few},t}^{(e)}\rVert_2}{\lVert m_{\mathrm{full},t}^{(e)}\rVert_2}.
\end{equation}
A large positive $A_g^{(e)}(t)$ means that the support and full gradients point in a similar direction. Ratios $R_m^{(e)}(t)$ near one mean that the support gradient also has the correct scale.

For second-order geometry, let
\begin{equation}
    H_{\mathrm{few},t}^{(e)}\in\mathbb{R}^{d_e\times d_e},
    \qquad
    H_{\mathrm{full},t}^{(e)}\in\mathbb{R}^{d_e\times d_e}
\end{equation}
be the encoder-restricted Hessians of $L_{\mathrm{few}}$ and $L_{\mathrm{full}}$ at $\Theta_t$. The Hessian controls the quadratic term of the local Taylor expansion. It determines which directions are steep, which directions are flat, and how sensitive a finite update is to curvature mismatch. Let $\big(\lambda_{1,\mathrm{few},t}^{(e)}, u_{1,\mathrm{few},t}^{(e)}\big)$ and $\big(\lambda_{1,\mathrm{full},t}^{(e)}, u_{1,\mathrm{full},t}^{(e)}\big)$ be the dominant eigenpairs, with $u_{1,S,t}^{(e)}\in\mathbb{R}^{d_e}$ and $\lVert u_{1,S,t}^{(e)}\rVert_2=1$. We define
\begin{equation}
    A_H^{(e)}(t)
    =
    \big|\langle u_{1,\mathrm{few},t}^{(e)}, u_{1,\mathrm{full},t}^{(e)}\rangle\big|,
    \qquad
    R_{\lambda}^{(e)}(t)
    =
    \log_{10}\!\left(\frac{|\lambda_{1,\mathrm{full},t}^{(e)}|}{|\lambda_{1,\mathrm{few},t}^{(e)}|}\right).
\end{equation}
The absolute value in $A_H^{(e)}(t)$ is needed because eigenvectors are defined only up to a sign. Positive values of $R_{\lambda}^{(e)}(t)$ indicate that the support loss underestimates the dominant full-train curvature magnitude.

We also measure curvature in a task-relevant direction. We define the normalized gradients
\begin{equation}
    \widehat{u}_{\mathrm{few},t}^{(e)}
    =
    \frac{m_{\mathrm{few},t}^{(e)}}{\lVert m_{\mathrm{few},t}^{(e)}\rVert_2},
    \qquad
    \widehat{u}_{\mathrm{full},t}^{(e)}
    =
    \frac{m_{\mathrm{full},t}^{(e)}}{\lVert m_{\mathrm{full},t}^{(e)}\rVert_2},
    \qquad
    \widehat{u}_{\mathrm{few},t}^{(e)},\widehat{u}_{\mathrm{full},t}^{(e)}\in\mathbb{R}^{d_e}.
\end{equation}
For $S\in\{\mathrm{few},\mathrm{full}\}$ and any unit vector $u\in\mathbb{R}^{d_e}$, we define the directional curvature
\begin{equation}
    \kappa_{S}^{(e)}(u;t)
    =
    u^{\top}H_{S,t}^{(e)}u
    \in\mathbb{R}.
\end{equation}
This is the second derivative of $L_S$ at $\Theta_t$ along direction $u$ while all non-$e$ coordinates are fixed. We compare support and full curvature along the full-train gradient direction by
\begin{equation}
    Q_{\kappa}^{(e)}(t)
    =
    \frac{
        |\kappa_{\mathrm{full}}^{(e)}(\widehat{u}_{\mathrm{full},t}^{(e)};t)|+\varepsilon_{\kappa}
    }{
        |\kappa_{\mathrm{few}}^{(e)}(\widehat{u}_{\mathrm{full},t}^{(e)};t)|+\varepsilon_{\kappa}
    },
    \qquad \varepsilon_{\kappa}=10^{-12}.
\end{equation}
The absolute values compare curvature magnitudes, and $\varepsilon_{\kappa}$ regularizes the denominator. Values $Q_{\kappa}^{(e)}(t)>1$ mean that the support loss underestimates the full-train curvature magnitude along the full-train descent direction. This diagnostic is complementary to the surface correlation below. The Pearson correlation measures the shape agreement of two finite surfaces, but it is invariant to positive rescaling. In contrast, $Q_{\kappa}^{(e)}(t)$ measures curvature-scale agreement along one task-relevant direction.

To compare local surfaces, we construct a two-dimensional probe plane in the spirit of standard loss-landscape visualizations~\citep{goodfellow2015qualitatively,li2018visualizing}. Let $u_{\perp,t}^{(e)}\in\mathbb{R}^{d_e}$ be the normalized component of $\widehat{u}_{\mathrm{full},t}^{(e)}$ orthogonal to $\widehat{u}_{\mathrm{few},t}^{(e)}$. Let $s_t^{(e)}>0$ be the recorded SGD step norm for encoder $e$. Keeping all non-$e$ coordinates fixed, we evaluate
\begin{equation}
    \Psi_S^{(e)}(\alpha,\beta;t)
    =
    L_S\!\left(\Theta_t + \iota_e\!\left(s_t^{(e)}\big(\alpha \widehat{u}_{\mathrm{few},t}^{(e)} + \beta u_{\perp,t}^{(e)}\big)\right)\right),
    \qquad S\in\{\mathrm{few},\mathrm{full}\},
\end{equation}
where $\iota_e:\mathbb{R}^{d_e}\to\mathbb{R}^{d}$ embeds an encoder perturbation into the full trainable parameter vector. We summarize support--full surface agreement by the Pearson correlation
\begin{equation}
    C_{\mathrm{surf}}^{(e)}(t)
    =
    \operatorname{corr}\!\left(
        \{\Psi_{\mathrm{few}}^{(e)}(\alpha,\beta;t)\}_{(\alpha,\beta)\in\mathcal{G}},
        \{\Psi_{\mathrm{full}}^{(e)}(\alpha,\beta;t)\}_{(\alpha,\beta)\in\mathcal{G}}
    \right)
\end{equation}
on the sampled grid $\mathcal{G}\subset[-2,2]^2$. Positive $C_{\mathrm{surf}}^{(e)}(t)$ means that the support and full objectives share a similar local valley in the probed plane.

\begin{figure*}[t]
    \centering
    \begin{subfigure}[t]{1.0\linewidth}
        \centering
        \includegraphics[width=\linewidth]{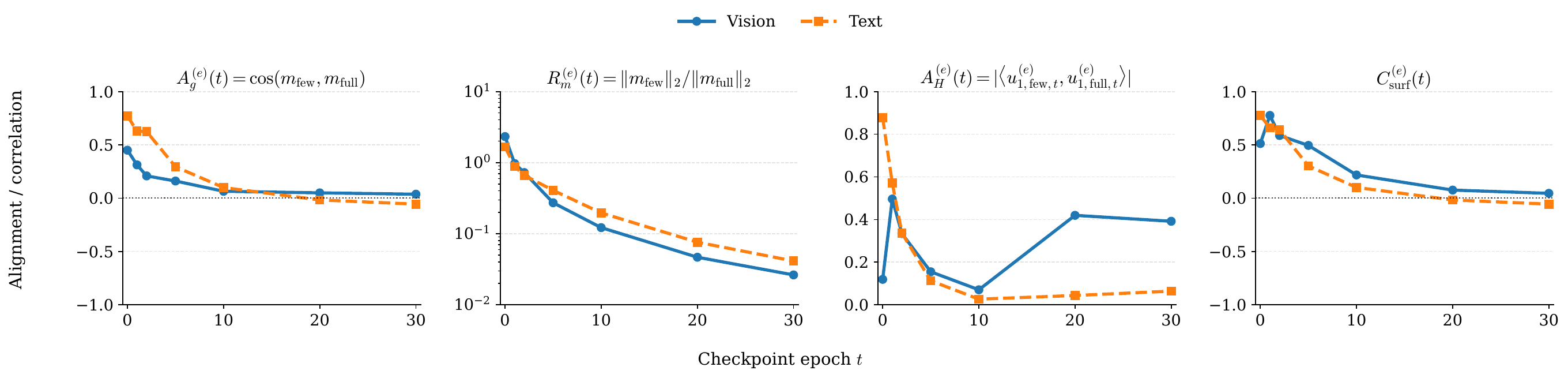}
        \caption{EuroSAT.}
        \label{fig:sgd_success_expH_1:a}
    \end{subfigure}\hfill
    \begin{subfigure}[t]{1.0\linewidth}
        \centering
        \includegraphics[width=\linewidth]{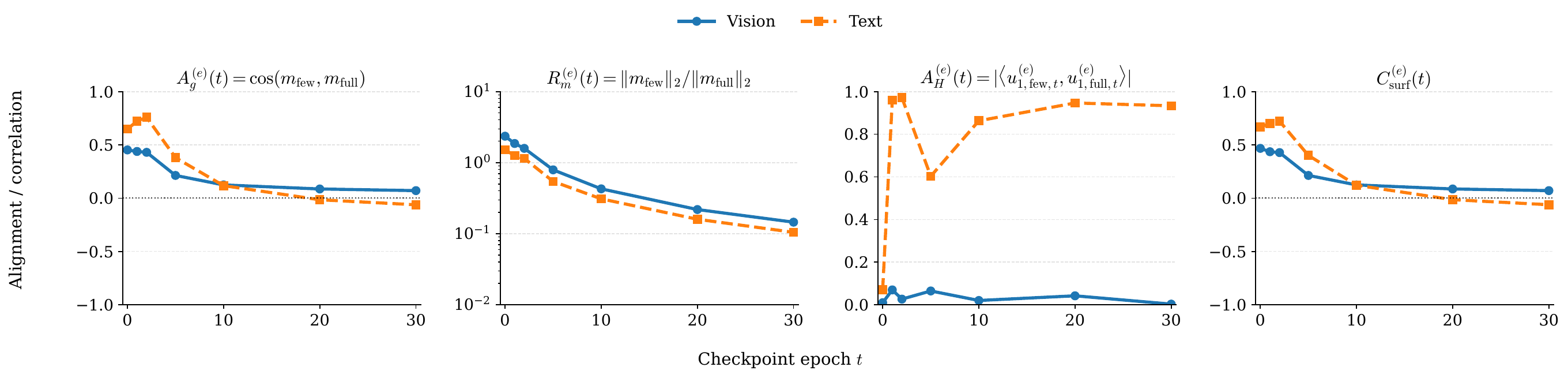}
        \caption{Oxford-Pets.}
        \label{fig:sgd_success_expH_1:b}
    \end{subfigure}\hfill
    \begin{subfigure}[t]{1.0\linewidth}
        \centering
        \includegraphics[width=\linewidth]{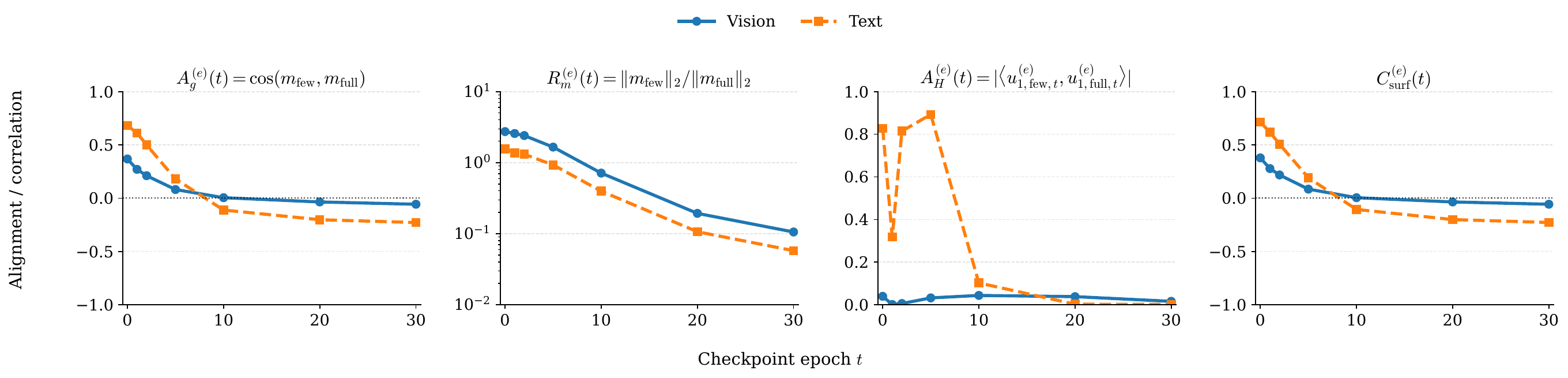}
        \caption{DTD.}
        \label{fig:sgd_success_expH_1:c}
    \end{subfigure}
    \caption{Experiment~G: encoder-level support/full geometric alignment over time. For encoder $e$, the panels report the first-order cosine $A_g^{(e)}(t)$, the gradient-norm ratio $R_m^{(e)}(t)$, the dominant-eigenspace alignment $A_H^{(e)}(t)$, and the surface correlation $C_{\mathrm{surf}}^{(e)}(t)$. Text starts more aligned than vision on all three datasets, but the support/full proxy deteriorates sharply by late training.}
    \label{fig:sgd_success_expH_1}
\end{figure*}

The early regime is aligned enough for SGD to make useful progress. Figure~\ref{fig:sgd_success_expH_1} shows that at $t{=}0$ the gradient cosine lies between $0.37$ and $0.77$ across the six encoder--dataset pairs, and the surface correlation lies between $0.38$ and $0.78$. The text encoder is consistently better aligned than the vision encoder at the start. EuroSAT text is the clearest example, with $A_g^{(\mathrm{text})}(0)=0.77$, $A_H^{(\mathrm{text})}(0)=0.88$, and $C_{\mathrm{surf}}^{(\mathrm{text})}(0)=0.78$. Thus, the support loss is initially a useful local proxy for the full-train loss.

This proxy does not remain stable. By $t{=}30$, the gradient-norm ratio has collapsed to $R_m^{(e)}(30)\in[0.026,0.145]$, so the support gradient is much smaller than the full gradient in all six cases. The gradient cosine also falls to $A_g^{(e)}(30)\in[-0.23,0.07]$, and the surface correlation drops to $C_{\mathrm{surf}}^{(e)}(30)\in[-0.23,0.07]$. Hence, the support objective stops being a faithful first-order and surface-level proxy of the full objective late in training.

The mismatch is structured across depth. At $t{=}0$, the vision block cosine ranges from $0.14$ to $0.69$ on DTD and from $0.33$ to $0.65$ on Oxford-Pets, whereas the text block cosine already lies in the tighter interval $[0.62,0.81]$ across datasets. By $t{=}30$, the DTD text blocks are all negative, with values in $[-0.45,-0.06]$. EuroSAT and Oxford-Pets are milder, but their text blocks also collapse toward zero. Therefore, the support/full mismatch is structured across depth rather than uniformly distributed.

Second-order alignment is even more fragile than first-order alignment. In the vision encoder at initialization, the gradient cosine is already moderately positive on all three datasets, with $A_g^{(\mathrm{vision})}(0)\in[0.37,0.45]$, but the dominant-eigenspace alignment is only $A_H^{(\mathrm{vision})}(0)\in[0.01,0.12]$. Thus, first-order agreement does not imply second-order agreement. SGD can still benefit from the early gradient signal because it is a first-order method with a shrinking step norm, but a curvature-aware method may already face a mismatched local quadratic model at the beginning of training.

The converse also occurs. On Oxford-Pets text, the dominant-eigenspace alignment is high at the final checkpoint, with $A_H^{(\mathrm{text})}(30)=0.93$, even though the gradient cosine and surface correlation have both decayed near zero. Therefore, second-order agreement of one dominant mode does not guarantee first-order or surface-level agreement. The three notions measure different aspects of the local geometry. %

\begin{figure*}[t]
    \centering
    \includegraphics[width=0.98\linewidth]{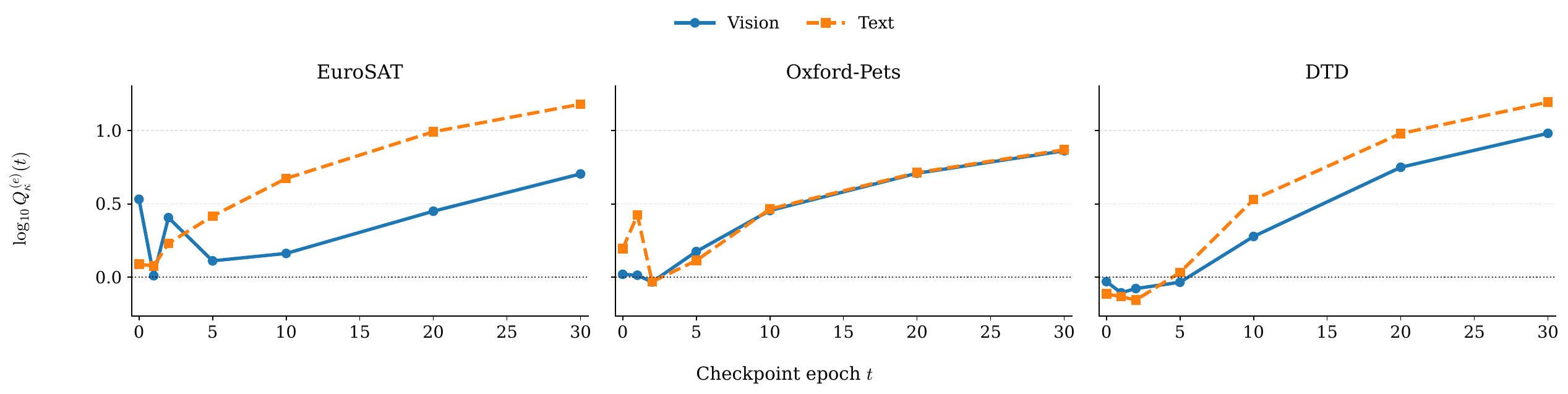}
    \caption{Experiment~G: directional curvature-scale mismatch along the full-train gradient direction. The figure reports $\log_{10} Q_{\kappa}^{(e)}(t)$ for the vision and text encoders. Values above zero indicate that the support loss underestimates the full-train curvature magnitude along $\widehat{u}_{\mathrm{full},t}^{(e)}$. The ratio grows during training on all three datasets. At $t{=}30$, the non-log ratio $Q_{\kappa}^{(e)}(30)$ lies between $5.06$ and $15.62$.}
    \label{fig:sgd_success_expH_kappa}
\end{figure*}

The curvature scale also drifts apart. By $t{=}30$, the dominant-curvature log-ratio $R_{\lambda}^{(e)}(t)$ is positive in all six cases and ranges from $0.39$ to $1.25$. Hence, the support loss underestimates the dominant full curvature by roughly $2.5\times$ to $18\times$, depending on the encoder and dataset. Figure~\ref{fig:sgd_success_expH_kappa} shows that the same conclusion holds along the full-gradient direction. At the final checkpoint, $Q_{\kappa}^{(e)}(30)$ ranges from $5.06$ to $15.62$. Thus, late support overfitting is not only a gradient-direction problem. It is also a curvature-scale problem along a task-relevant descent direction.

\begin{figure*}[t]
    \centering
    \begin{subfigure}[t]{1.0\linewidth}
        \centering
        \includegraphics[width=\linewidth]{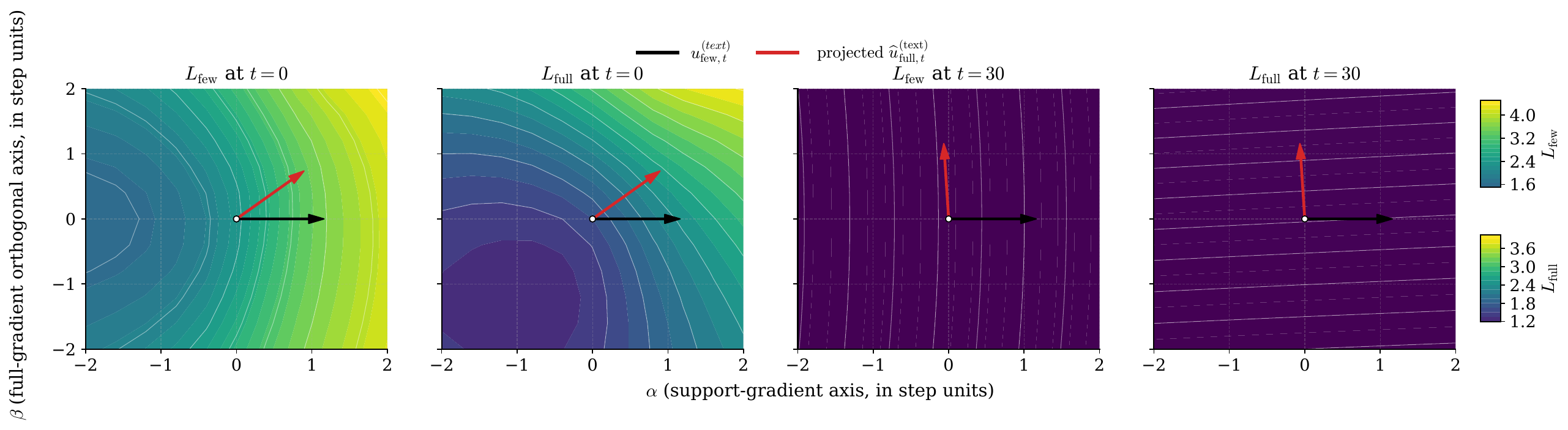}
        \caption{EuroSAT, text encoder.}
        \label{fig:sgd_success_expH_2:a}
    \end{subfigure}\hfill
    \begin{subfigure}[t]{1.0\linewidth}
        \centering
        \includegraphics[width=\linewidth]{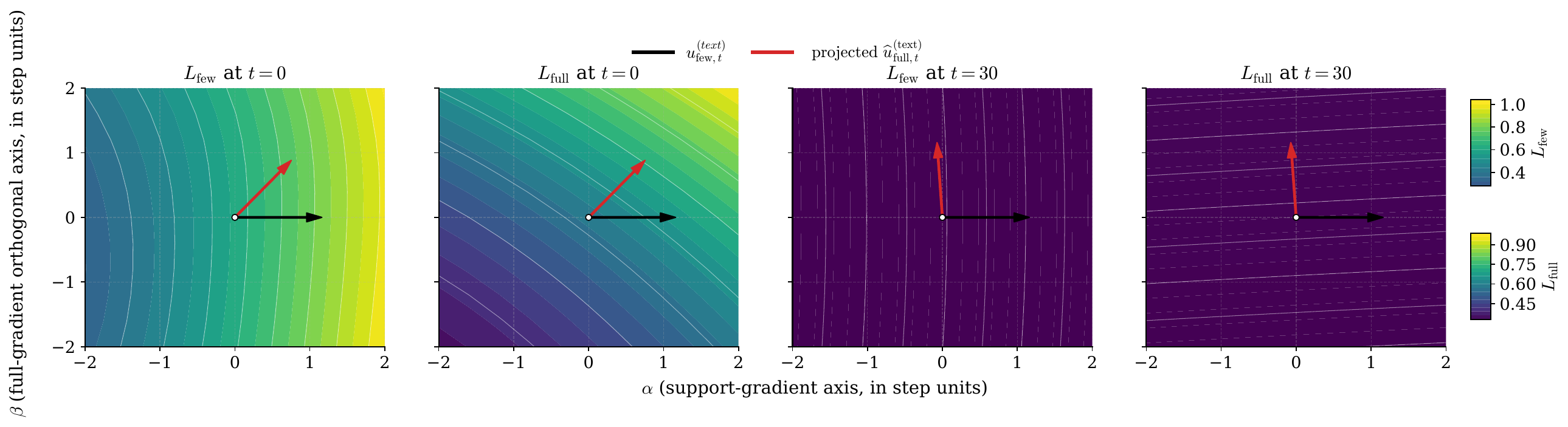}
        \caption{Oxford-Pets, text encoder.}
        \label{fig:sgd_success_expH_2:b}
    \end{subfigure}\hfill
    \begin{subfigure}[t]{1.0\linewidth}
        \centering
        \includegraphics[width=\linewidth]{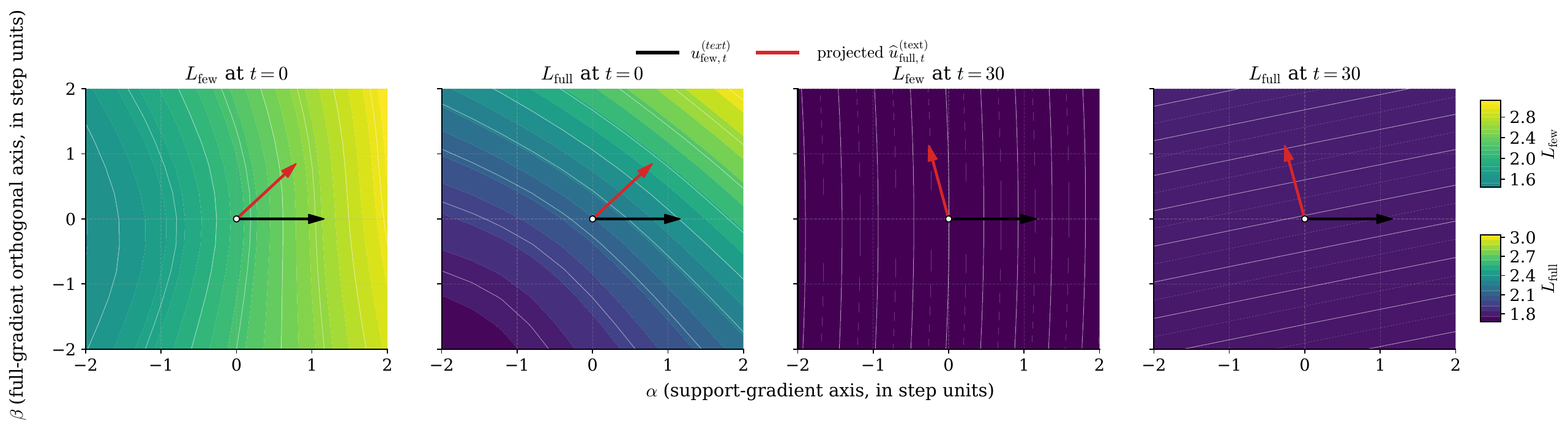}
        \caption{DTD, text encoder.}
        \label{fig:sgd_success_expH_2:c}
    \end{subfigure}
    \caption{Experiment~G: representative local surfaces for the text encoder at $t{=}0$ and $t{=}30$. Each panel reports $\Psi_{\mathrm{few}}^{(\mathrm{text})}$ or $\Psi_{\mathrm{full}}^{(\mathrm{text})}$ on the plane spanned by the support-gradient direction and the orthogonal component of the full gradient. The black arrow is $\widehat{u}_{\mathrm{few},t}^{(\mathrm{text})}$. The red arrow is the projected full-gradient direction. Early surfaces are positively aligned. Late surfaces flatten or rotate away from the full-train objective, especially on EuroSAT and DTD.}
    \label{fig:sgd_success_expH_2}
\end{figure*}

\begin{figure*}[t]
    \centering
    \begin{subfigure}[t]{1.0\linewidth}
        \centering
        \includegraphics[width=\linewidth]{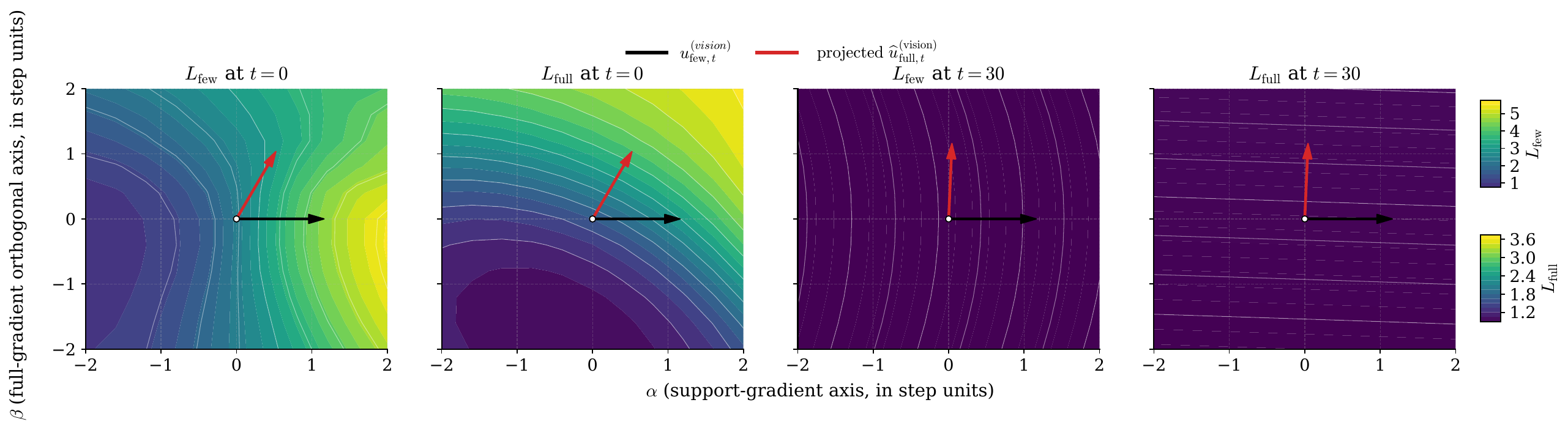}
        \caption{EuroSAT, vision encoder.}
        \label{fig:sgd_success_expH_3:a}
    \end{subfigure}\hfill
    \begin{subfigure}[t]{1.0\linewidth}
        \centering
        \includegraphics[width=\linewidth]{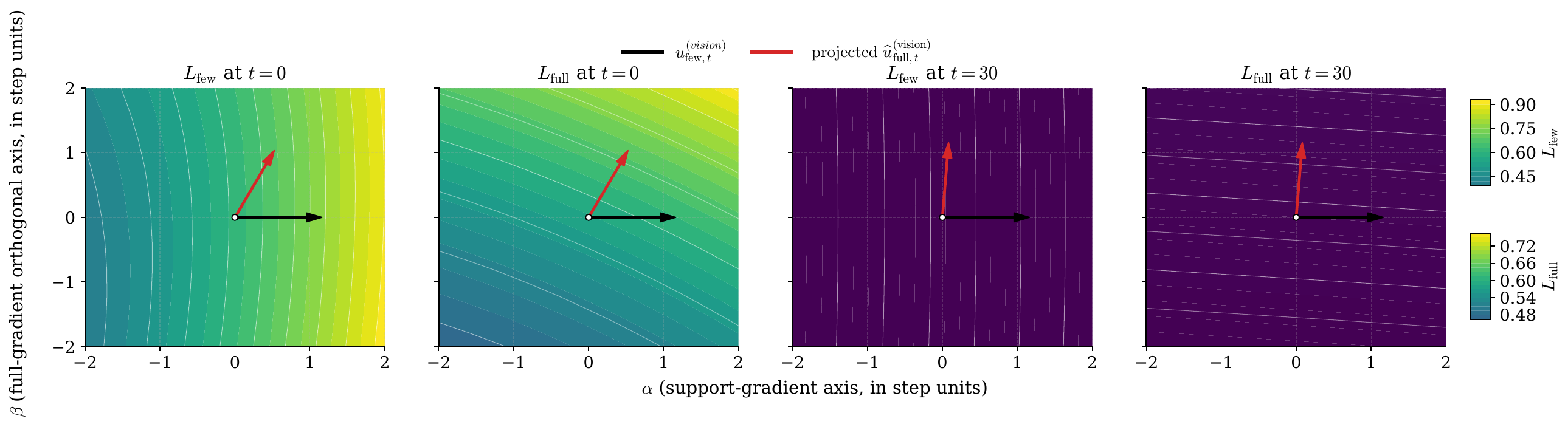}
        \caption{Oxford-Pets, vision encoder.}
        \label{fig:sgd_success_expH_3:b}
    \end{subfigure}\hfill
    \begin{subfigure}[t]{1.0\linewidth}
        \centering
        \includegraphics[width=\linewidth]{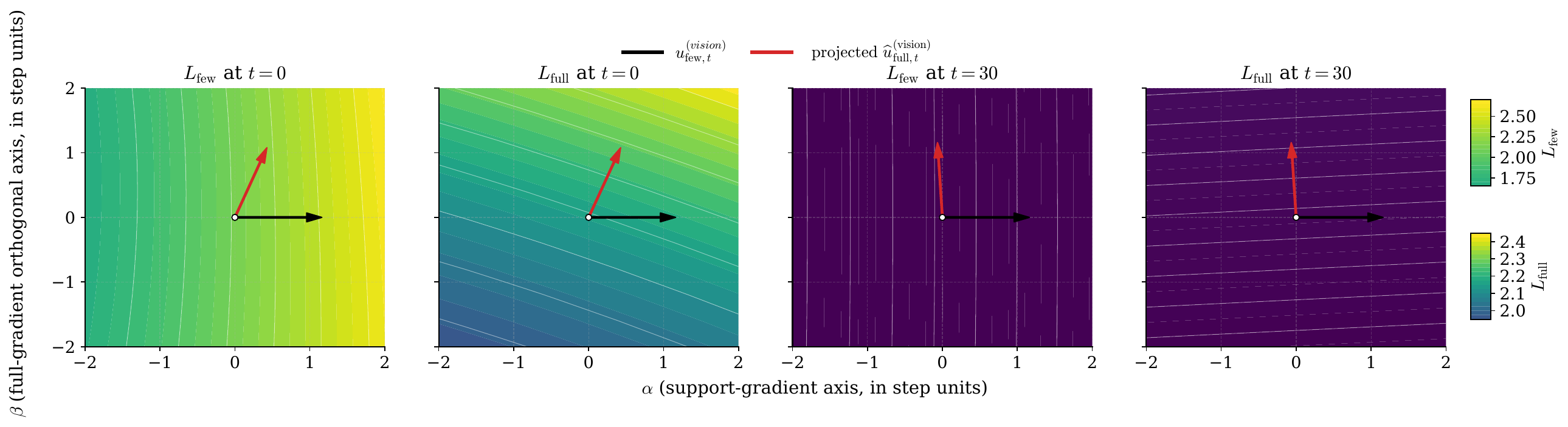}
        \caption{DTD, vision encoder.}
        \label{fig:sgd_success_expH_3:c}
    \end{subfigure}
    \caption{Experiment~G: representative local surfaces for the vision encoder at $t{=}0$ and $t{=}30$. Each panel reports $\Psi_{\mathrm{few}}^{(\mathrm{vision})}$ or $\Psi_{\mathrm{full}}^{(\mathrm{vision})}$ on the same probe plane as in Figure~\ref{fig:sgd_success_expH_2}. Even when the early vision gradient is moderately aligned with the full gradient, the local full-train curvature can already differ substantially. By late training, the support surface becomes flatter and less informative than the full surface on all three datasets.}
    \label{fig:sgd_success_expH_3}
\end{figure*}

Figures~\ref{fig:sgd_success_expH_2} and \ref{fig:sgd_success_expH_3} make the surface mismatch explicit. Figure~\ref{fig:sgd_success_expH_2} focuses on the text encoder. At $t{=}0$, the support and full surfaces share a positively aligned valley on all three datasets. By $t{=}30$, EuroSAT and DTD show a clear rotation and flattening of the support surface relative to the full surface. Oxford-Pets is milder, but its text surface still loses most of its early alignment. Figure~\ref{fig:sgd_success_expH_3} focuses on the vision encoder. Even at $t{=}0$, when the vision gradient cosine is still moderately positive, the full surface can already have a different dominant curvature orientation. By $t{=}30$, the support surface is flatter and less informative than the full surface on EuroSAT, Oxford-Pets, and DTD.

The dataset-level picture is consistent. EuroSAT shows strong early alignment and clear late decoupling. Oxford-Pets is the mildest case: the final vision alignment remains weakly positive, and the surface correlation stays slightly above zero, but the text branch still loses first-order agreement. DTD is the strongest mismatch case. By the final checkpoint, both encoders have near-zero or negative gradient cosine, the text Hessian alignment collapses to zero, and the support and full surfaces become negatively correlated. DTD is, therefore, the clearest support-overfitting regime among the three datasets.

SGD does not perform well in few-shot adaptation because the support and full geometries stay aligned throughout training. It performs well because the early aligned regime is informative enough, and because the SGD step norm shrinks rapidly enough to exploit that regime before severe mismatch sets in. Experiment~G, therefore, refines the explanation of Section~\ref{sec:why_does_sgd_succeed}. Early first-order alignment is useful, but second-order alignment is already fragile, and late support/full mismatch eventually limits dense few-shot adaptation. This complements Experiment~B, which showed that DTD becomes support-specific precisely when the support loss stops being a faithful local proxy of the full loss.

\begin{tcolorbox}[
  enhanced,
  unbreakable,
  colback=findingblueback,
  colframe=findingblueframe,
  colbacktitle=findingblueframe,
  coltitle=white,
  boxrule=0.7pt,
  arc=1.5pt,
  left=6pt,
  right=6pt,
  top=5pt,
  bottom=5pt,
  before skip=7pt,
  after skip=7pt,
  fonttitle=\bfseries\small,
  fontupper=\small,
  title={Conclusion}]
SGD does not remain in a region where the support loss and the full-train loss are geometrically aligned throughout training. The support objective is a useful local first-order proxy for the full-train objective only during the early stage of optimization. Along the SGD trajectory, first-order alignment is initially positive, especially in the text encoder, and local support/full surfaces are positively correlated. However, this proxy degrades during training. The support gradient norm collapses relative to the full gradient. Gradient alignment decays to near-zero or negative values. The support Hessian increasingly underestimates full-train curvature, including along the full-gradient direction. The local support/full surfaces also decorrelate or become anticorrelated. These observations are consistent with SGD exploiting the initially aligned regime through conservative updates. Later, the support loss saturates while the full-train objective retains a substantial descent signal.
\end{tcolorbox}

\subsection{Does SGD stay in a low-drift region of CLIP feature space?}

\begin{figure*}[t]
    \centering
    \begin{subfigure}[t]{1.0\linewidth}
        \centering
        \includegraphics[width=\linewidth]{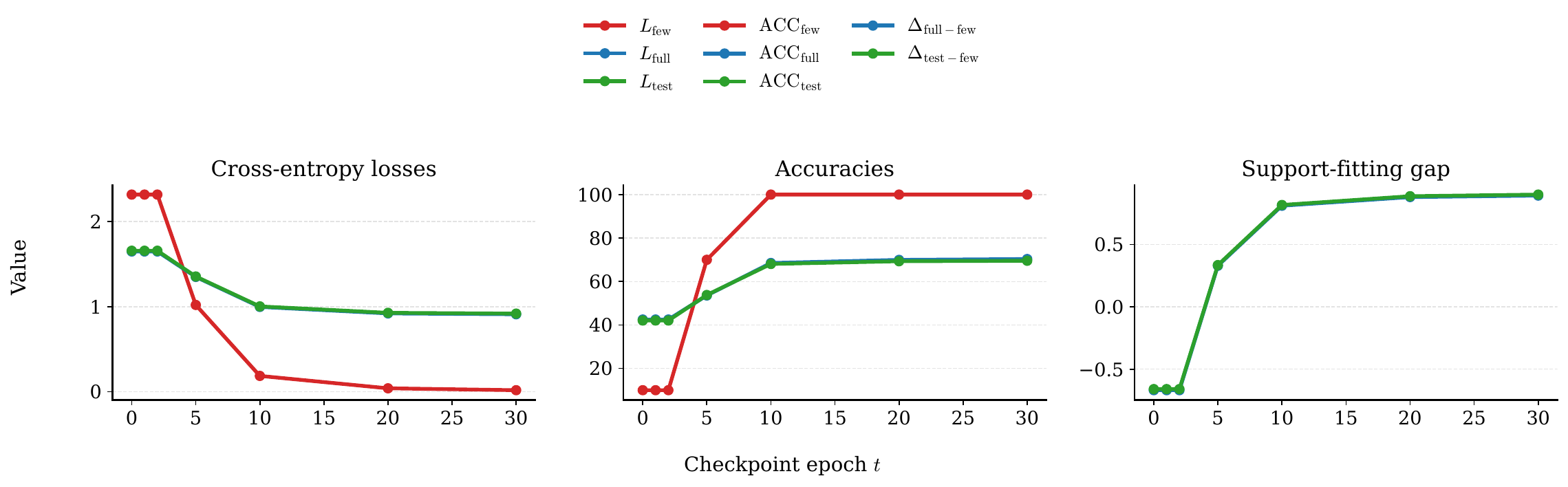}
        \caption{EuroSAT.}
    \end{subfigure}\hfill
    \begin{subfigure}[t]{1.0\linewidth}
        \centering
        \includegraphics[width=\linewidth]{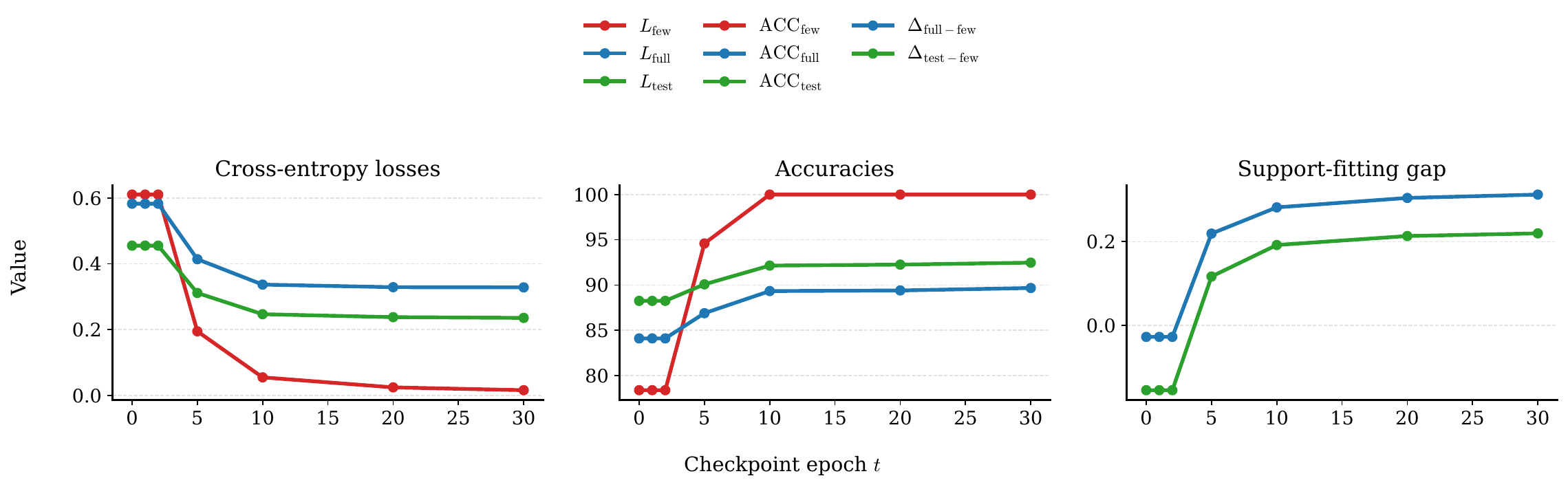}
        \caption{Oxford-Pets.}
    \end{subfigure}\hfill
    \begin{subfigure}[t]{1.0\linewidth}
        \centering
        \includegraphics[width=\linewidth]{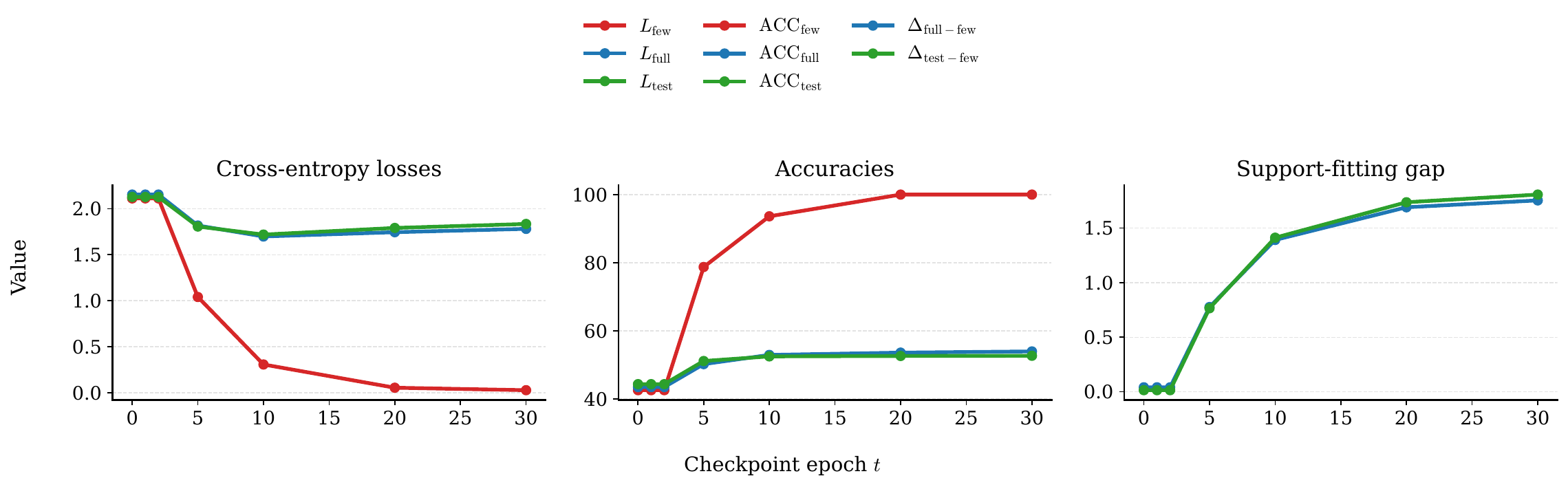}
        \caption{DTD.}
    \end{subfigure}
    \caption{Experiment~H: losses, accuracies, and support-fitting gaps along the SGD trajectory. The support objective improves rapidly on all three datasets. EuroSAT and Oxford-Pets keep improving on the full-train and test objectives until the final checkpoint. DTD is different: after the best full-train checkpoint, the support loss continues to fall while the full-train and test gaps keep widening.}
    \label{fig:sgd_success_expI_1}
\end{figure*}

Experiment~H tests whether the SGD trajectory remains close to the pretrained CLIP in parameter space and in the feature spaces of the vision and text encoders.

We use the parameter and feature notation from Section~\ref{sec:why_does_adam_fail} with trajectory label $o=\mathrm{SGD}$. The trainable state is $\Theta_t$, and $d_{\mathrm{emb}}$ is the CLIP embedding dimension. The parameter, image-feature, text-feature, and class-margin displacements are $R_{\Theta}^{(\mathrm{SGD},G)}(t)$, $R_{\mathrm{img},S}^{(\mathrm{SGD})}(t)$, $R_{\mathrm{text}}^{(\mathrm{SGD})}(t)$, and $R_{\mathrm{mar},S}^{(\mathrm{SGD})}(t)$, respectively. We report parameter displacement for all trainable parameters, each encoder, and each Transformer block.

To quantify support specialization, we also use the loss gaps
\begin{equation}
    \Delta_{\mathrm{full-few}}(t)=L_{\mathrm{full}}(\Theta_t)-L_{\mathrm{few}}(\Theta_t),
    \qquad
    \Delta_{\mathrm{test-few}}(t)=L_{\mathrm{test}}(\Theta_t)-L_{\mathrm{few}}(\Theta_t).
\end{equation}
Large positive gaps indicate that the support objective is being fitted more strongly than the broader task.

\begin{figure*}
    \centering
    \begin{subfigure}[t]{1.0\linewidth}
        \centering
        \includegraphics[width=\linewidth]{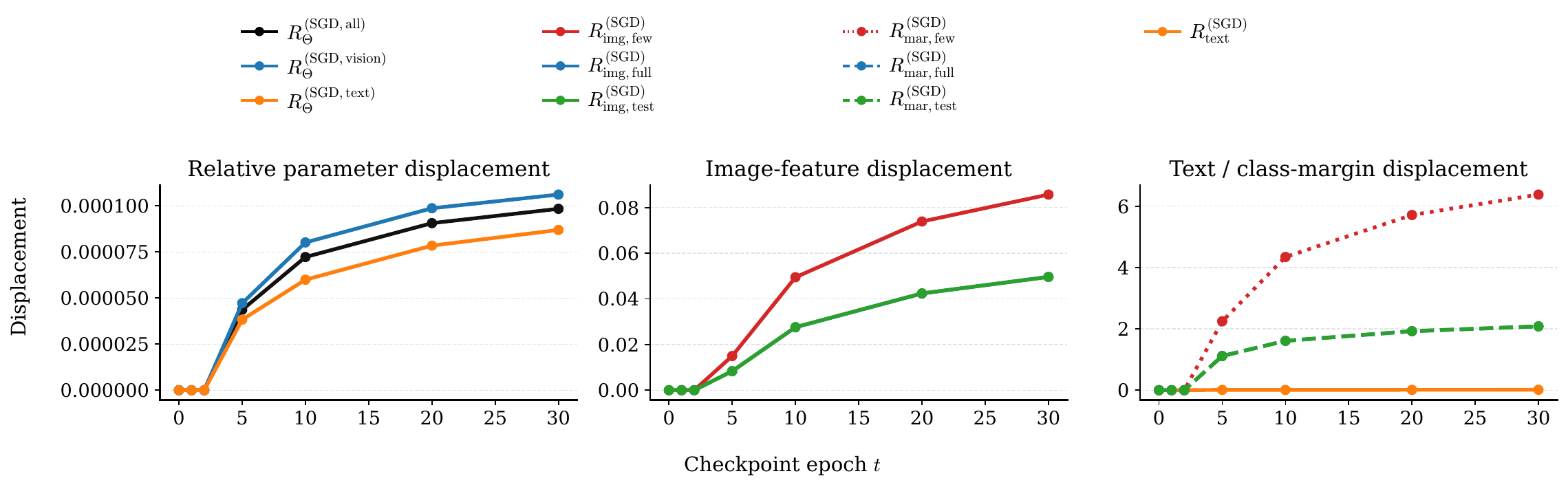}
        \caption{EuroSAT.}
    \end{subfigure}\hfill
    \begin{subfigure}[t]{1.0\linewidth}
        \centering
        \includegraphics[width=\linewidth]{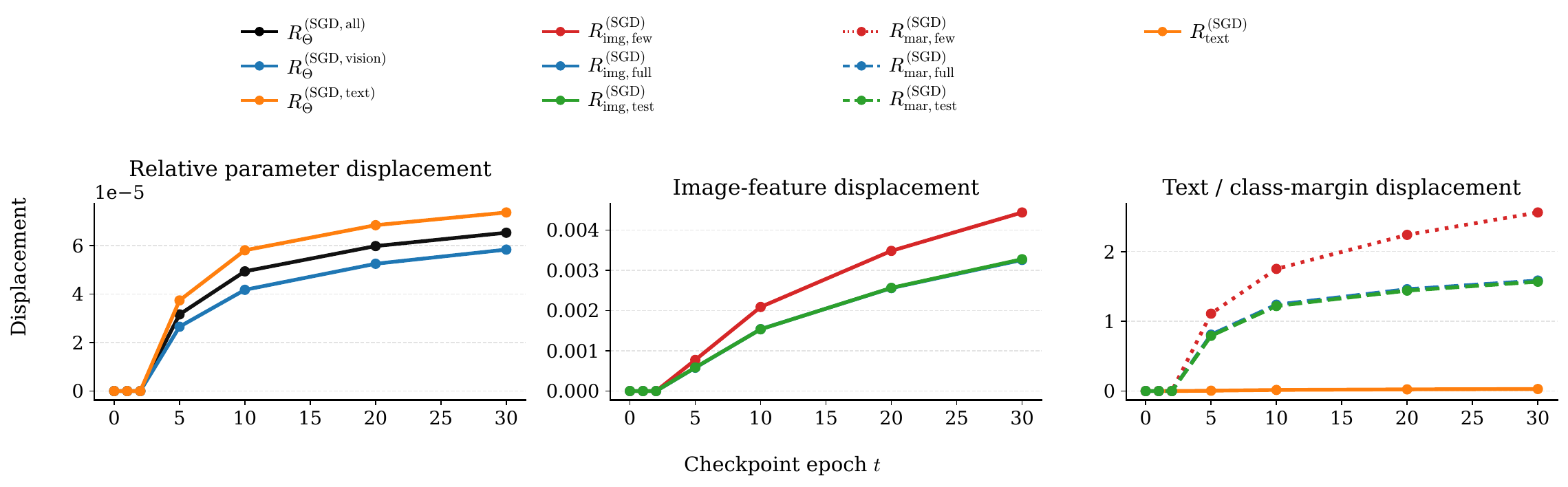}
        \caption{Oxford-Pets.}
    \end{subfigure}\hfill
    \begin{subfigure}[t]{1.0\linewidth}
        \centering
        \includegraphics[width=\linewidth]{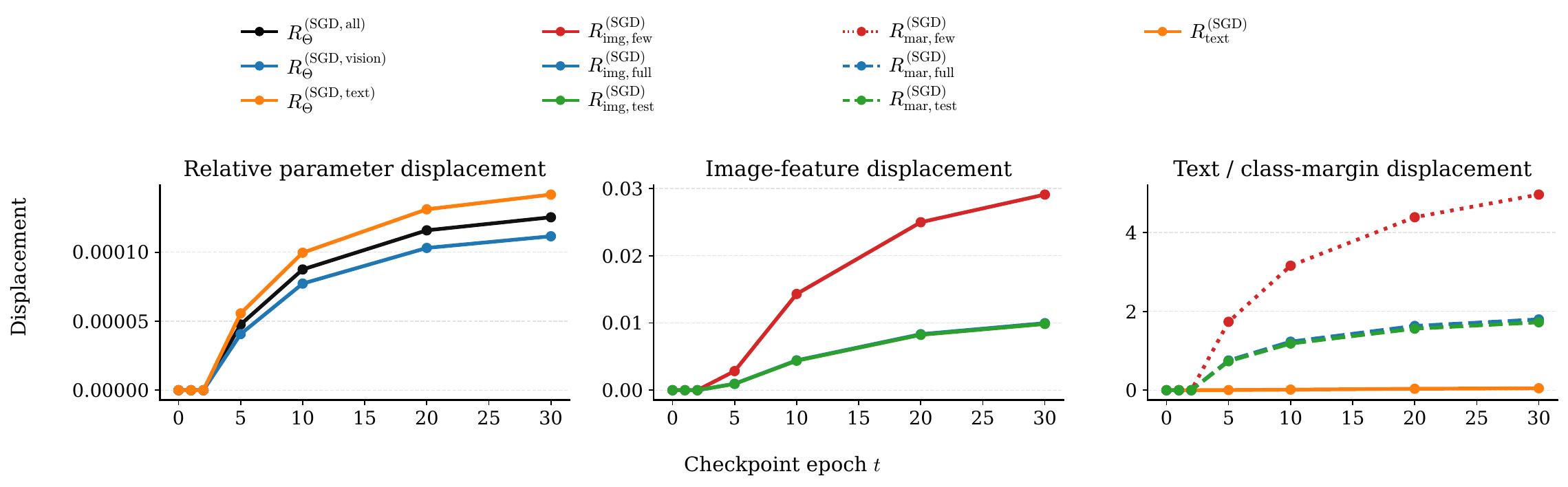}
        \caption{DTD.}
    \end{subfigure}
    \caption{Experiment~H: parameter displacement, feature displacement, and class-margin displacement. The relative parameter displacement stays extremely small on all three datasets. The support image features have larger mean angular displacement than the full-train and test image features. The full-train and test means remain close to each other.}
    \label{fig:sgd_success_expI_2}
\end{figure*}

\begin{figure*}
    \centering
    \begin{subfigure}[t]{1.0\linewidth}
        \centering
        \includegraphics[width=\linewidth]{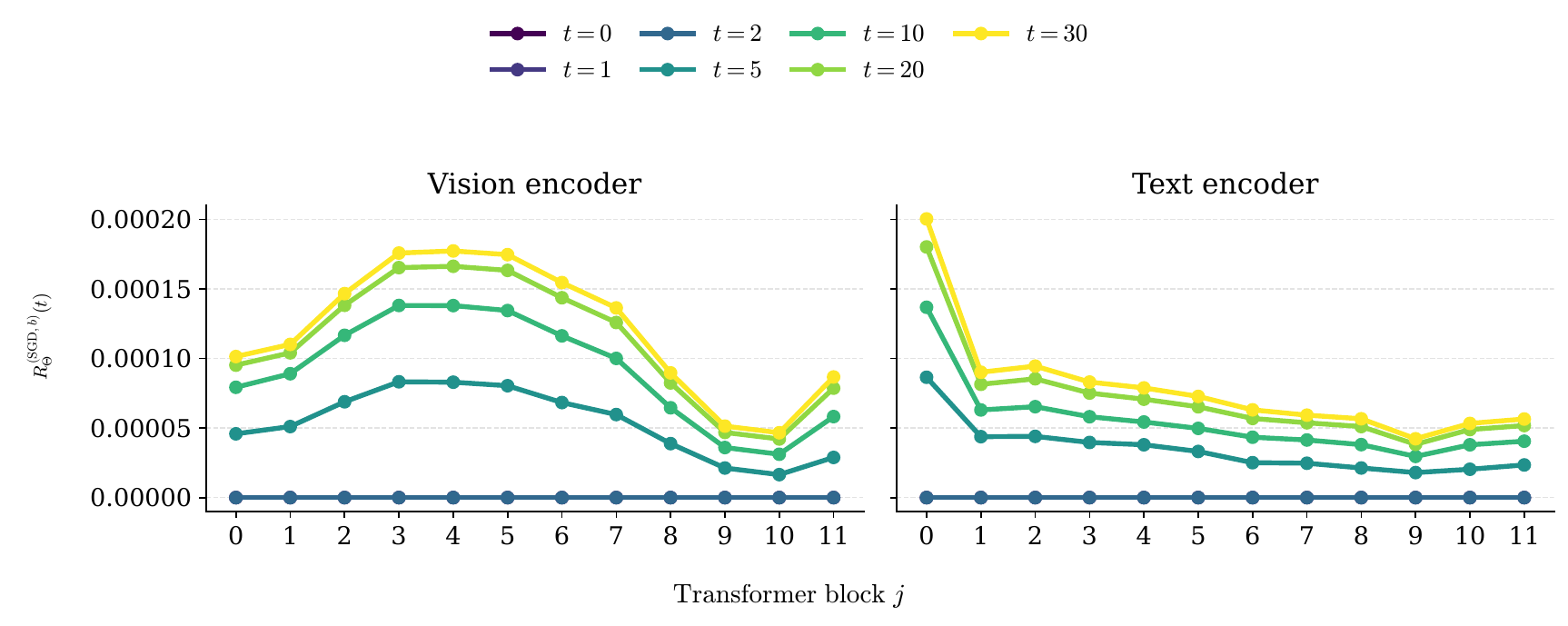}
        \caption{EuroSAT.}
    \end{subfigure}\hfill
    \begin{subfigure}[t]{1.0\linewidth}
        \centering
        \includegraphics[width=\linewidth]{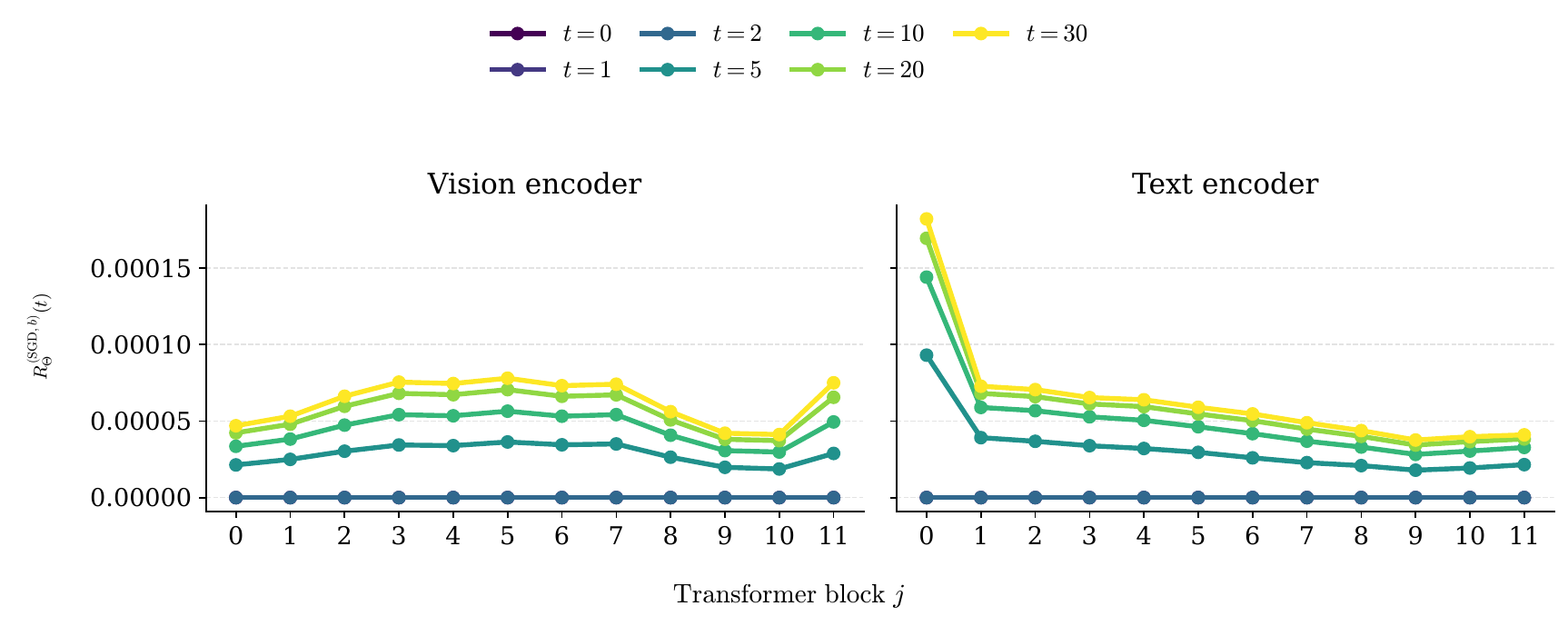}
        \caption{Oxford-Pets.}
    \end{subfigure}\hfill
    \begin{subfigure}[t]{1.0\linewidth}
        \centering
        \includegraphics[width=\linewidth]{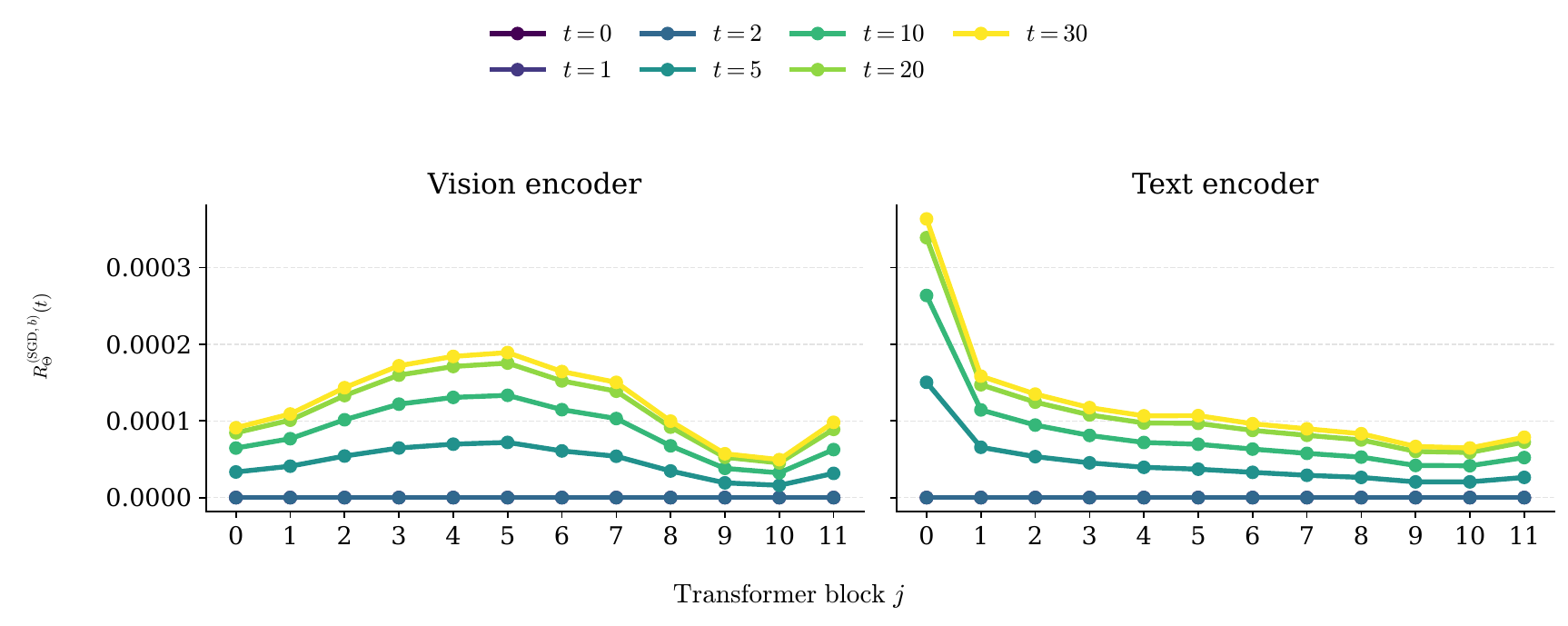}
        \caption{DTD.}
    \end{subfigure}
    \caption{Experiment~H: blockwise parameter displacement. The motion is highly structured rather than uniform. On EuroSAT and DTD, the largest vision-encoder displacement concentrates in middle vision blocks, while the text-encoder displacement is dominated by the earliest text blocks, especially block $0$. Oxford-Pets stays lower-displacement overall and is more diffuse in the vision encoder.}
    \label{fig:sgd_success_expI_3}
\end{figure*}

The first conclusion is that SGD remains in a very low-displacement parameter regime. At the final checkpoint, the total relative displacement is only $6.5\times 10^{-5}$ on Oxford-Pets, $9.8\times 10^{-5}$ on EuroSAT, and $1.25\times 10^{-4}$ on DTD. These values are tiny compared with the norm of the pretrained model. Thus, SGD does not move far away from CLIP. It reduces the few-shot loss while remaining very close to the pretrained solution.

The second conclusion is that mean image-feature displacement is larger on the support set. Figure~\ref{fig:sgd_success_expI_2} shows that the support image displacement is consistently larger than the full-train and test image displacements. By the final checkpoint, the ratio $R_{\mathrm{img},\mathrm{few}}^{(\mathrm{SGD})}(30)/R_{\mathrm{img},\mathrm{full}}^{(\mathrm{SGD})}(30)$ is about $1.36$ on Oxford-Pets, $1.73$ on EuroSAT, and $2.92$ on DTD. The full-train and test splits have similar mean angular displacement on every dataset. Both means remain smaller than the support-set displacement. These scalar averages do not establish preserved pairwise geometry or coupling between the splits.

The third conclusion is that some displacement is beneficial, but only up to the point where support specialization takes over. On EuroSAT and Oxford-Pets, the best full-train and best test checkpoints occur at the final epoch. The full-train loss falls monotonically from $1.650$ to $0.911$ on EuroSAT and from $0.583$ to $0.328$ on Oxford-Pets, while the test accuracy rises to $69.56\%$ and $92.48\%$, respectively. Figure~\ref{fig:sgd_success_expI_1} shows that the support-fitting gaps remain moderate on those two datasets. DTD behaves differently. Its best full-train checkpoint occurs at $t=10$, where $L_{\mathrm{full}}=1.698$ and $R_{\Theta}^{(\mathrm{SGD},\mathrm{all})}(10)=8.75\times 10^{-5}$. After that point, the support loss continues to decrease from $0.307$ to $0.027$, but the full-train loss worsens to $1.781$, and the support-fitting gap grows from $1.39$ to $1.75$. Thus, the late DTD trajectory no longer reflects broad-task progress, even though the total parameter displacement is still very small.

The fourth conclusion is that displacement is highly structured across depth. Figure~\ref{fig:sgd_success_expI_3} shows that the largest vision-encoder displacement concentrates in the middle vision blocks. At $t=30$, the most displaced vision blocks are $j\in\{3,4,5\}$ on EuroSAT, $j\in\{3,5,11\}$ on Oxford-Pets, and $j\in\{4,5,6\}$ on DTD. The text encoder is even more concentrated. On all three datasets, block $0$ has the largest displacement by a wide margin, and the next largest text displacement occurs in the early text blocks. This is consistent with Experiment~E, which showed that the useful SGD motion is concentrated in a structured subset of blocks rather than being spread uniformly across the model.

Experiment~H gives a precise answer to the subsection question. SGD does stay in a low-displacement region of the CLIP feature space, especially on EuroSAT and Oxford-Pets. Improved task performance accompanies small parameter displacement and limited mean feature displacement. DTD shows that small displacement alone does not prevent support specialization. Once the support-fitting gaps keep increasing, the trajectory remains low-displacement in absolute terms but becomes too support-specific. This complements Experiments~B and G. Experiment~B showed that the late DTD steps are no longer locally safe for the full objective, and Experiment~G showed that the support geometry stops being a faithful local proxy of the full geometry. Experiment~H adds that this failure does not require a large departure from pretrained CLIP. A very small but badly placed displacement is already enough to create late support overfitting.

\begin{findingbox}[Conclusion]
SGD combines improved task performance with small parameter displacement and limited mean feature displacement. The full-train and test splits have similar mean angular displacement. 
\end{findingbox}

\subsection{Does SGD’s stochasticity help in finding flatter regions?}

We now study whether SGD benefits from stochasticity in this regime. In Experiments~I1--I3, let $o$ index five variants: full-batch SGD, half-batch SGD, quarter-batch SGD, full-batch SGD with additive isotropic noise, and full-batch SGD with momentum. For each variant $o$ and checkpoint $t$, let $\Theta_t^{(o)}\in\mathbb{R}^{d}$ denote the trainable parameter vector, where $d$ is the total number of trainable scalar parameters.

To quantify stochasticity, let $g_{\mathcal B,t}^{(o)}\in\mathbb{R}^{d}$ be the gradient of micro-batch $\mathcal B$ and let $g_{\mathrm{few},t}^{(o)}\in\mathbb{R}^{d}$ be the full support gradient. Let $\mathfrak B_t$ be the collection of the $B_t$ micro-batches used at checkpoint $t$. For Experiments~I1--I3, the relative batch-noise level is
\begin{equation}
    \nu_{\mathrm{batch}}^{(o)}(t)
    =
    \frac{\left(\sum_{\mathcal B\in\mathfrak B_t}\frac{|\mathcal B|}{\sum_{\mathcal B'\in\mathfrak B_t}|\mathcal B'|}\|g_{\mathcal B,t}^{(o)}-g_{\mathrm{few},t}^{(o)}\|_2^2\right)^{1/2}}
         {\|g_{\mathrm{few},t}^{(o)}\|_2}.
\end{equation}
Each squared deviation is weighted by the fraction of processed examples in its micro-batch. In Experiment~I4, $\nu_{\mathrm{batch}}^{(o)}(t)$ is the root-mean-square deviation of complete accumulated bootstrap-update gradients from the full support gradient, normalized by the same full support gradient norm. The intermediate micro-batch gradients are not used as separate noise samples.
For the additive-noise control, let $\nu_{\mathrm{inj}}^{(o)}\ge 0$ denote the relative injected-noise amplitude. The total stochasticity ratio is
\begin{equation}
    \nu_{\mathrm{tot}}^{(o)}(t)
    =
    \sqrt{\big(\nu_{\mathrm{batch}}^{(o)}(t)\big)^2 + \big(\nu_{\mathrm{inj}}^{(o)}\big)^2}.
\end{equation}
Thus, mini-batch variants carry structured stochasticity from data subsampling, whereas the full-batch noise control adds isotropic perturbations to the same deterministic base step.

To probe task-relevant flatness, we reuse the full-train gradient direction. Let
\begin{equation}
    u_t^{(o,\mathrm{full})}
    =
    \frac{g_{\mathrm{full},t}^{(o)}}{\|g_{\mathrm{full},t}^{(o)}\|_2}
    \in \mathbb{R}^{d}
\end{equation}
be the normalized full-train gradient. For a small finite-difference radius $\delta_{\mathrm{fd}}>0$, we define the directional curvature probe
\begin{equation}
    \chi_{\mathrm{full}}^{(o)}(t)
    =
    \frac{L_{\mathrm{full}}\big(\Theta_t^{(o)}+\delta_{\mathrm{fd}} u_t^{(o,\mathrm{full})}\big)
    +L_{\mathrm{full}}\big(\Theta_t^{(o)}-\delta_{\mathrm{fd}} u_t^{(o,\mathrm{full})}\big)
    -2L_{\mathrm{full}}\big(\Theta_t^{(o)}\big)}{\delta_{\mathrm{fd}}^2}.
    \label{eq:expJ_chi_full}
\end{equation}
The corresponding random-direction control is denoted by $\chi_{\mathrm{rand}}^{(o)}(t)$. For positive curvature, lower $\chi_{\mathrm{full}}^{(o)}(t)$ indicates lower local quadratic sensitivity along the full-train gradient. Negative values instead indicate negative curvature.

Experiment~I1 compares the five variants at matched epochs. Experiment~I2 matches the number of optimizer updates. Let $U_t^{(o)}\in\mathbb{N}$ be the number of updates taken by variant $o$ up to checkpoint $t$, and let $U_\star$ be the full-batch baseline total. The matched update coordinate is
\begin{equation}
    \xi_{\mathrm{upd},t}^{(o)} = \frac{U_t^{(o)}}{U_\star} \in [0,1].
\end{equation}
Experiment~I3 matches the number of seen support examples. Let $E_t^{(o)}\in\mathbb{N}$ be the number of support examples processed by variant $o$ up to checkpoint $t$, and let $E_\star$ be the full-batch baseline total. The matched example coordinate is
\begin{equation}
    \xi_{\mathrm{ex},t}^{(o)} = \frac{E_t^{(o)}}{E_\star} \in [0,1].
\end{equation}
Experiment~I4 replaces the half- and quarter-batch variants with bootstrap-resampled full-support updates. Each resampled step draws as many examples with replacement as there are examples in the support set. The draw is partitioned into half- or quarter-support micro-batches. Each micro-batch mean loss is weighted by its fraction of the draw, and gradients are accumulated before one optimizer update. The two settings use the same bootstrap draws. Their micro-batch sizes affect accumulation, but not the update estimator in exact arithmetic. I4 matches the update count and number of sample draws, including repeated examples, and reports checkpoints satisfying
\begin{equation}
    \xi_{\mathrm{upd},t}^{(o)} = \xi_{\mathrm{ex},t}^{(o)} = \xi.
\end{equation}
Experiments~I1--I3 compare epoch, update, and example budgets. Experiment~I4 also changes the stochastic gradient estimator and therefore does not isolate the effect of ordinary mini-batch noise.

\begin{figure*}[t]
    \centering
    \includegraphics[width=0.985\linewidth]{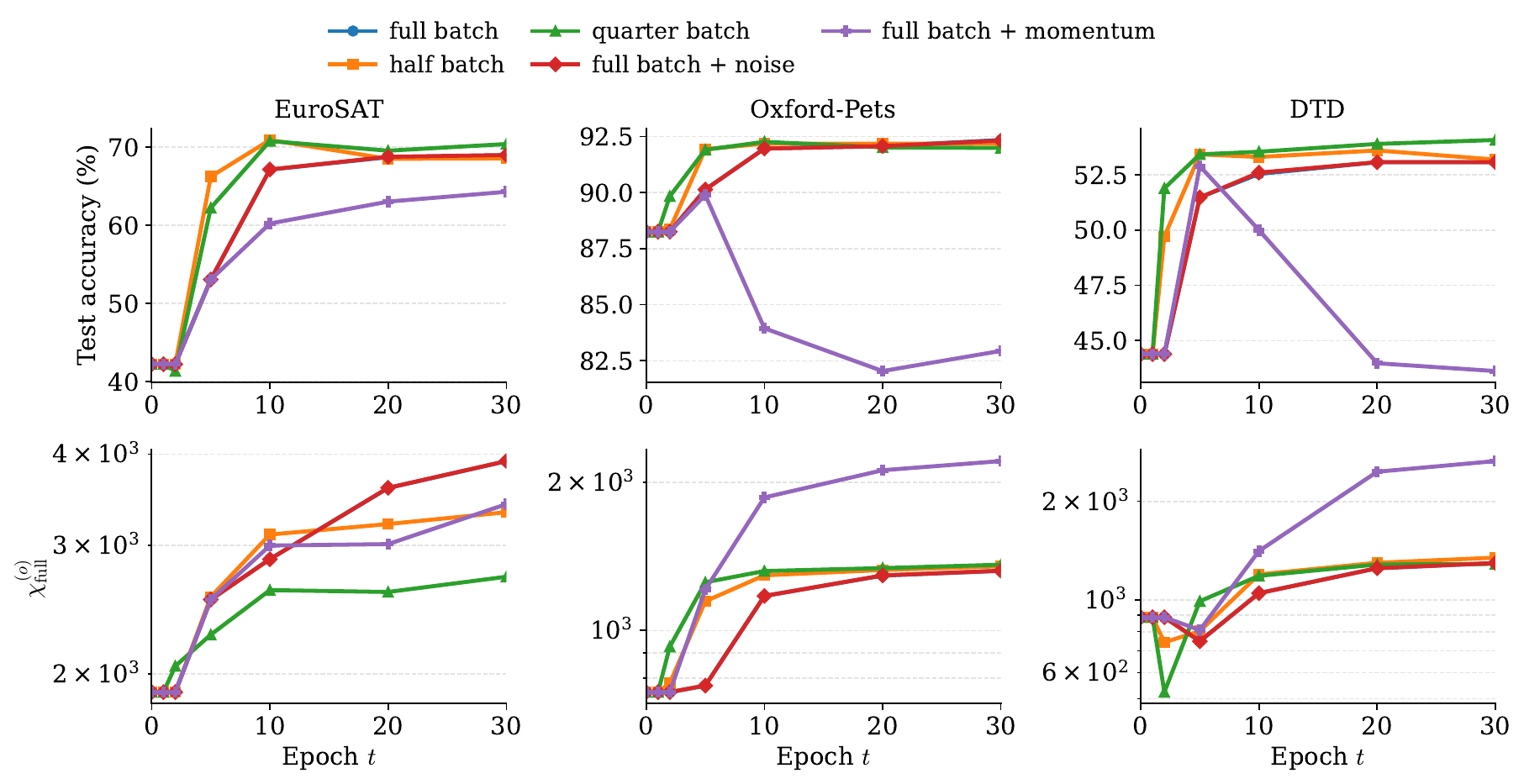}
    \caption{Experiment~I1 (epoch-matched). Top row: test accuracy. Bottom row: directional flatness probe $\chi_{\mathrm{full}}^{(o)}(t)$ from Eq.~\eqref{eq:expJ_chi_full}. The tested mini-batch variants improve accuracy on EuroSAT and DTD, but not on Oxford-Pets. Small isotropic additive noise is nearly indistinguishable from full-batch SGD. Momentum is consistently harmful.}
    \label{fig:sgd_success_expJ_1}
\end{figure*}

Figure~\ref{fig:sgd_success_expJ_1} gives the epoch-matched comparison. The first result is that the local geometry is extremely anisotropic. For the full-batch baseline at the final checkpoint, the ratio $\chi_{\mathrm{full}}/\chi_{\mathrm{rand}}$ is about $2.29\times 10^{6}$ on EuroSAT, $1.38\times 10^{6}$ on Oxford-Pets, and $5.67\times 10^{5}$ on DTD. Thus, the relevant notion of flatness is not isotropic flatness. The landscape is almost flat in generic directions and sharply curved along the full-train gradient direction.

Figure~\ref{fig:sgd_success_expJ_1} also shows that the tested mini-batch variants improve accuracy only on some datasets. On EuroSAT, quarter-batch SGD improves the final test accuracy from $68.98\%$ to $70.41\%$ and reduces the final directional curvature from $3914.6$ to $2717.6$. This is a $1.43$ point gain together with a curvature reduction to about $0.69\times$ the baseline. On DTD, quarter-batch SGD improves the final test accuracy from $53.07\%$ to $54.08\%$, but the curvature change is small, from $1297.1$ to $1289.8$. On Oxford-Pets, full-batch SGD remains the best. Quarter-batch SGD slightly lowers the final test accuracy and slightly increases the curvature. Hence, the tested mini-batch variants do not improve every dataset. Their largest gain occurs on EuroSAT.

A third observation from Figure~\ref{fig:sgd_success_expJ_1} is that isotropic additive noise does almost nothing. On all three datasets, the full-batch noise control is nearly indistinguishable from the full-batch baseline in both test accuracy and directional curvature. By contrast, the mini-batch variants have total stochasticity ratios around $1.0$--$1.9$, whereas the additive-noise control uses only $0.1$. Thus, the helpful effect is not reproduced by a small isotropic perturbation. This comparison does not separate the effects of noise magnitude and covariance.

A fourth observation from Figure~\ref{fig:sgd_success_expJ_1} is that momentum is harmful to all three datasets. The final test accuracy drops by $4.68$ points on EuroSAT, $9.40$ points on Oxford-Pets, and $9.46$ points on DTD. The momentum trajectory also drifts much farther from initialization than the no-momentum baseline. Therefore, the favorable geometry of plain SGD in this regime is not improved by adding a long-horizon first-order accumulator.

\begin{figure*}[t]
    \centering
    \includegraphics[width=0.985\linewidth]{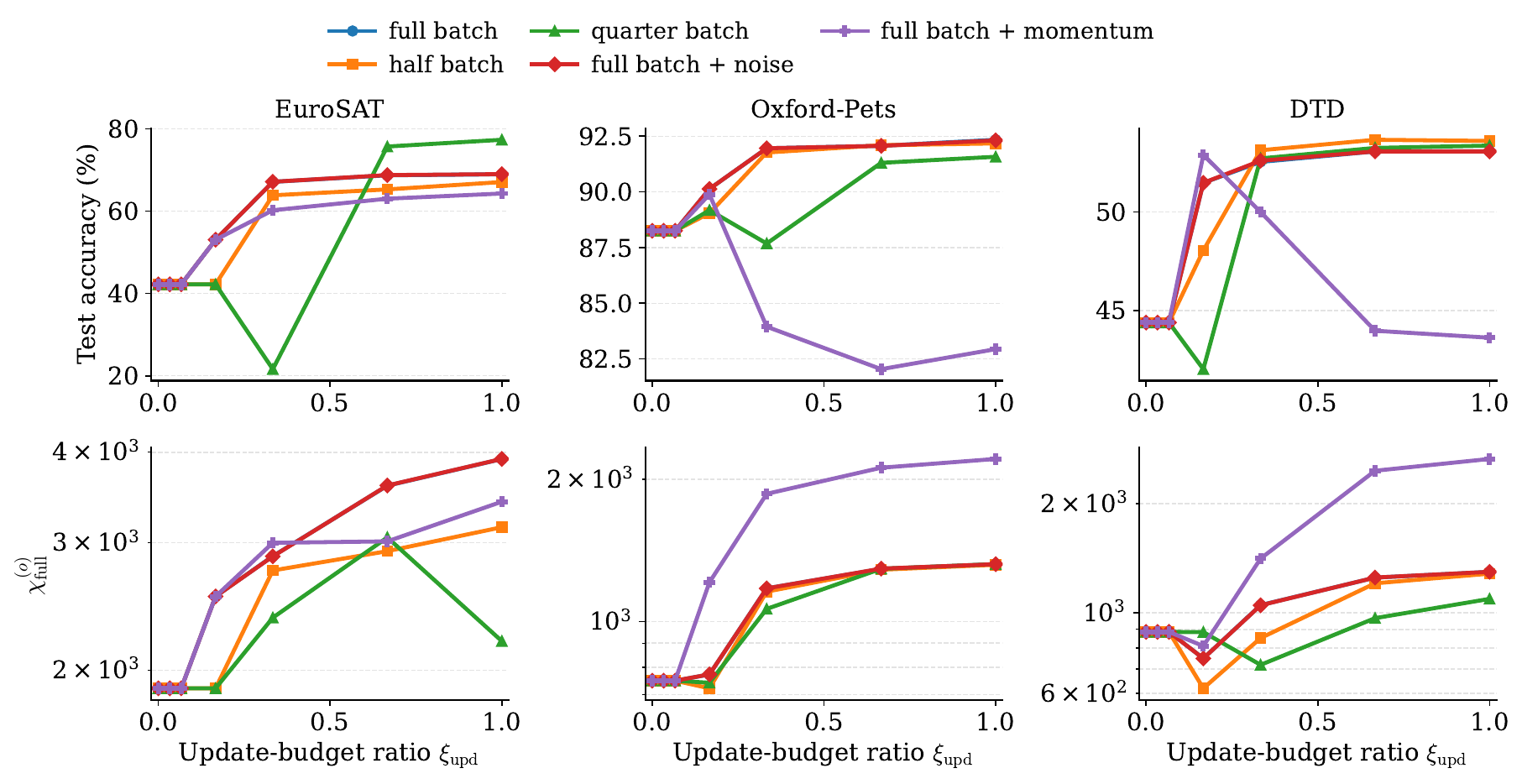}
    \caption{Experiment~I2 (update-matched). The horizontal axis is the update-budget ratio $\xi_{\mathrm{upd}}$. The large EuroSAT gain of quarter-batch SGD survives and even strengthens when the total number of optimizer updates is fixed.}
    \label{fig:sgd_success_expJ_2}
\end{figure*}

Experiment~I2 tests whether the I1 gain is simply caused by taking more updates per epoch. Figure~\ref{fig:sgd_success_expJ_2} shows that the answer is no. At a fixed update budget, quarter-batch SGD on EuroSAT reaches $77.36\%$ test accuracy, compared to $68.98\%$ for full-batch SGD, while reducing the final directional curvature from $3914.6$ to $2193.5$. Thus, the strongest EuroSAT gain survives update matching and becomes even larger. DTD shows a milder version of the same trend. The quarter-batch variant improves the final test accuracy by $0.30$ points and reduces the curvature to about $0.84\times$ of the baseline. Oxford-Pets again shows no benefit. Therefore, the EuroSAT gain cannot be explained as a trivial effect of taking more optimizer steps.

Reusing the barrier diagnostics from Experiment~F, the quarter-batch solution has zero full-loss and test-loss interpolation barriers on all three datasets when updates are matched. On EuroSAT, the interpolation minimum remains close to the quarter-batch endpoint, with $\alpha_{\mathrm{full}}^\star\approx 0.95$ and $\alpha_{\mathrm{test}}^\star\approx 0.93$. Hence, under update matching, the better EuroSAT endpoint is connected to the baseline by a barrier-free straight segment on the measured grid.

\begin{figure*}[t]
    \centering
    \includegraphics[width=0.985\linewidth]{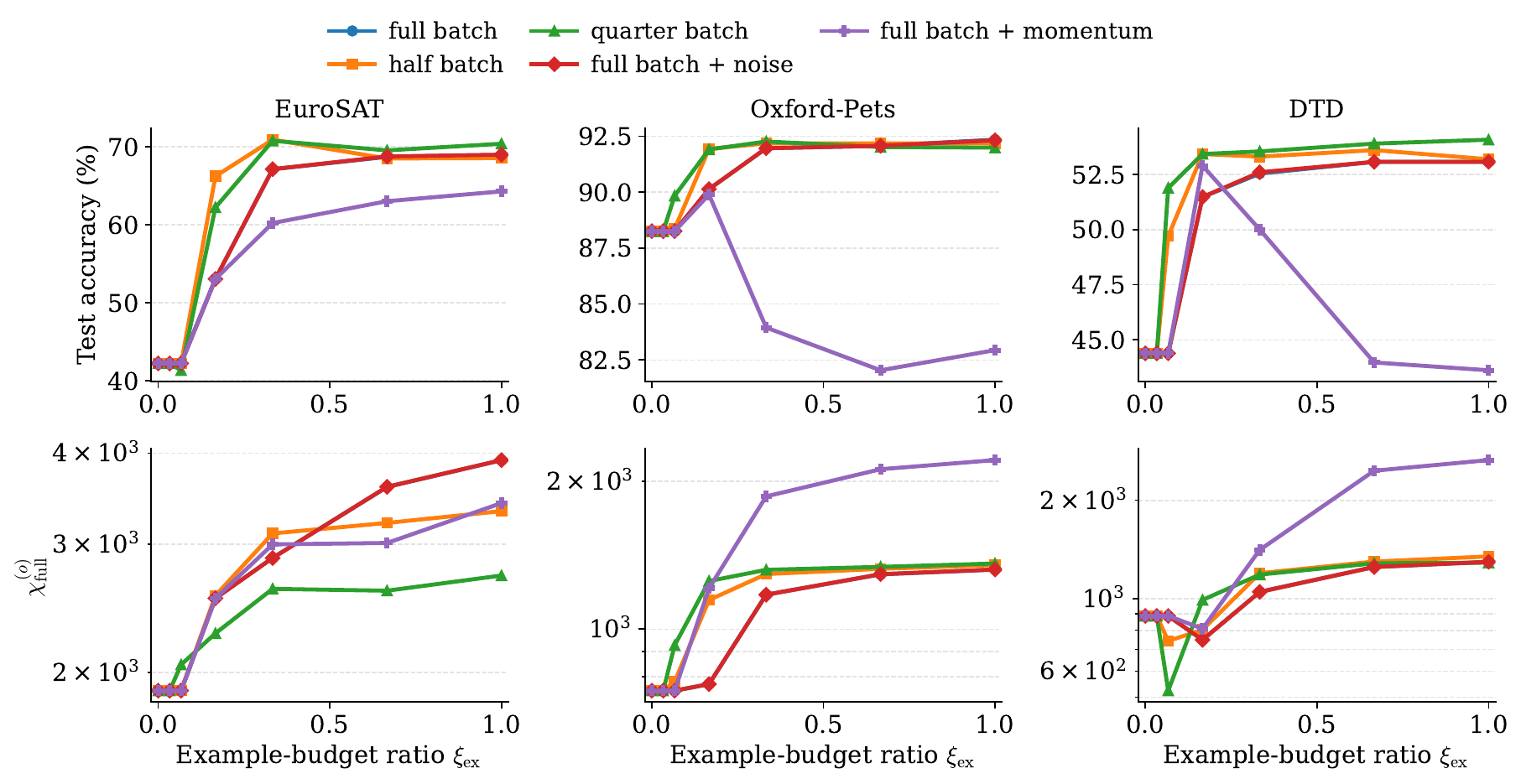}
    \caption{Experiment~I3 (example-matched). The horizontal axis is the example-budget ratio $\xi_{\mathrm{ex}}$. In the full-support regime, every epoch already processes the whole support set once, so Experiment~I3 collapses to Experiment~I1.}
    \label{fig:sgd_success_expJ_3}
\end{figure*}

Experiment~I3 is a sanity control on the number of seen support examples. Figure~\ref{fig:sgd_success_expJ_3} is essentially identical to Figure~\ref{fig:sgd_success_expJ_1}. This is not accidental. In the full-support regime used throughout Section~\ref{sec:why_does_sgd_succeed}, every epoch consumes the entire support set exactly once, regardless of mini-batch size. Matching the number of seen support examples is therefore equivalent to matching epochs. Experiment~I3 rules out the explanation that the I1 gain is caused by showing the model more support data. The relevant difference comes from how those same examples are partitioned into updates.

\begin{figure*}[t]
    \centering
    \includegraphics[width=0.985\linewidth]{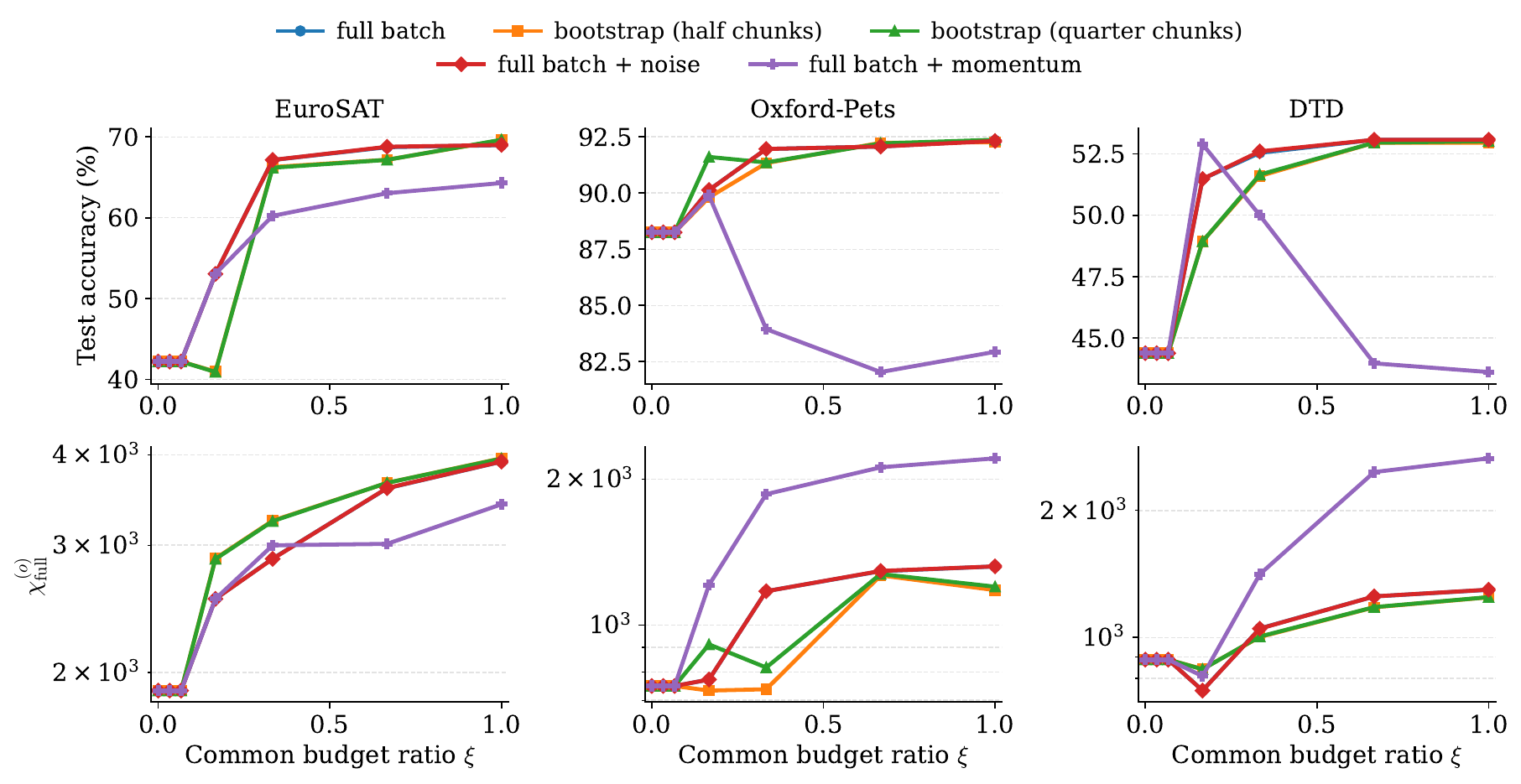}
    \caption{Experiment~I4: bootstrap-resampled full-support updates with matched update counts and sample draws. The legend labels half and quarter chunks refer to half- and quarter-support micro-batches used for gradient accumulation. The horizontal axis is the common budget ratio $\xi$. Accuracy differences from full-batch SGD are small under this estimator.}
    \label{fig:sgd_success_expJ_4}
\end{figure*}

Experiment~I4 compares full-support-sized bootstrap updates with the full-batch baseline at matched update counts and sample draws. Figure~\ref{fig:sgd_success_expJ_4} shows small final accuracy differences under this estimator. The bootstrap variant accumulated in quarter-support micro-batches ends slightly above the baseline on EuroSAT, $69.64\%$ versus $68.98\%$, but its final curvature is slightly \emph{higher}, $3952.8$ versus $3914.6$. On Oxford-Pets, the differences become negligible. The same variant ends at $92.37\%$ versus $92.34\%$, while lowering the final curvature from $1322.0$ to $1200.9$. On DTD, the final differences are essentially zero. These results apply to full-support-sized bootstrap updates. Because I4 changes the stochastic gradient estimator, it does not isolate the contribution of ordinary mini-batch noise to the larger gains in Experiments~I1 and~I2.

The interpolation results further describe the endpoint geometry. On EuroSAT, quarter-batch SGD in Experiments~I1 and~I3 has nonzero interpolation barriers relative to the full-batch baseline, with $B_{\mathrm{full}}\approx 0.079$ and $B_{\mathrm{test}}\approx 0.094$, and the minima lie near the quarter-batch endpoint. These endpoints are not connected by a barrier-free straight segment on the measured grid. This does not exclude other low-loss paths between them. In Experiments~I2 and~I4, these barriers vanish. Thus, their measured straight paths do not rise above the endpoint chords. DTD and Oxford-Pets show zero barriers for the mini-batch variants in Experiments~I1--I3 and the bootstrap variants in Experiment~I4. Momentum also often has a zero barrier, but its endpoint is much worse. Therefore, a zero barrier is not, by itself, sufficient. It only shows that the sampled straight path does not exceed its endpoint chord.

Experiment~I shows that the tested mini-batch configurations can improve endpoint accuracy and reduce task-relevant directional curvature. The strongest gains occur on EuroSAT. The effect is mild on DTD and absent on Oxford-Pets. Small isotropic additive noise does not reproduce these gains, and momentum is consistently harmful. The noise amplitudes differ between the mini-batch and isotropic controls. These comparisons therefore do not identify covariance structure as the cause of the gains.

\begin{findingbox}[Conclusion]
SGD stochasticity is not a universal flatness mechanism in few-shot CLIP adaptation. With full-support-sized bootstrap updates, matching the update count and number of sample draws leaves only small accuracy differences. This control changes the stochastic gradient estimator and therefore does not isolate the effect of ordinary mini-batch noise.
\end{findingbox}

\section{Additional Experiments}
\label{app:additional_results}

These experiments extend the update-scale analysis across support sizes, datasets, and a second vision--language model family. They also test learning-rate reduction, warmup, and repeated budget matching. The fixed-horizon accuracy comparisons use the final checkpoint. The trajectory analysis reports fixed-epoch, peak, and last-observed accuracy separately.

\subsection{Common evaluation and update-budget controls}
\label{app:additional_protocol}

The CLIP experiments use ViT-B/16, seeds $1,2,3$, 30 epochs, and nominal batch size $64$. SGD has no momentum. Adam uses $(\beta_1,\beta_2)=(0.9,0.999)$ and numerical stabilizer $10^{-8}$. Unless varied explicitly, the learning rate is $10^{-4}$ and training uses the same data processing pipeline explained in Appendix~\ref{app:hyperparameters} without weight decay, warmup, a scheduler, or gradient accumulation. The trainable CLIP parameters are the dense attention and MLP weight matrices in all Transformer blocks of both encoders. Biases, normalization parameters, embeddings, external projections, and logit scale remain fixed. The support contains $KC$ examples, so supports with $KC\leq64$ are processed in one batch; larger supports use shuffled mini-batches.

The shot-scaling study uses EuroSAT, Oxford-Pets, DTD, FGVC-Aircraft, Stanford-Cars, and Caltech101 at $K\in\{1,2,4,8,16\}$. Every optimizer receives the same support examples within each dataset--shot--seed setting. The full-training reference is used for measurement, not optimization. The complete augmented comparison contains $270$ runs: six datasets, five shot levels, three seeds, and three optimizers. An additional no-augmentation comparison contains $54$ runs on EuroSAT, Oxford-Pets, and DTD at $K\in\{1,4,16\}$ for Adam and SGD.

\paragraph{First-step and every-step matching.}
Adam-M0 is the first-step control called Adam-M in Section~\ref{sec:why_does_adam_fail_short}. It uses the global relative update budget $r_o(t)$ already defined there. At the common initialization, its learning rate is scaled to satisfy
\begin{equation}
 r_{\text{Adam-M0}}(0)=r_{\mathrm{SGD}}(0),
 \qquad
 \eta_{\mathrm{M0}}
 =\eta\,\frac{r_{\mathrm{SGD}}(0)}{r_{\mathrm{Adam}}(0)}.
 \label{eq:additional_m0}
\end{equation}
Here, $r_{\mathrm{Adam}}(0)$ is measured at the nominal learning rate $\eta$. The resulting rate remains fixed for the remainder of training. The implementation calibrates the applied parameter change to account for finite-precision storage. Adam-Mt repeats global matching at every update, using an SGD proposal computed from the same support gradient at the current checkpoint. Both controls retain Adam's adaptive direction and moment recursion. They change its scalar update scale. Global matching does not impose equal updates within every block or coordinate.

\paragraph{Aggregation.}
Dataset--shot entries report mean $\pm$ sample standard deviation over the three seeds. Aggregate accuracy gives equal weight to every dataset--shot--seed setting. Gap recovery is the ratio of the aggregate Adam-M0 improvement over raw Adam to the aggregate SGD improvement over raw Adam, multiplied by $100$. It is signed and is not clipped at $100\%$. Thus, recovery above $100\%$ means that Adam-M0 exceeds the SGD reference. Reported $95\%$ intervals use $20{,}000$ paired bootstrap resamples of complete dataset clusters, retaining all shots and seeds within each sampled dataset.

\subsection{First-step matching across datasets and shot levels}
\label{app:additional_m0}

Table~\ref{tab:additional_p2_by_shot} gives the matched six-dataset comparison. At the fixed 30-epoch horizon, Adam-M0 reaches $76.03\%$, compared with $74.93\%$ for SGD and $14.51\%$ for raw Adam. The corresponding recovery is $101.82\%$ with a $95\%$ interval of $[100.54,103.43]$. The paired Adam-M0 advantage over SGD is $1.10$ percentage points, with interval $[0.33,1.96]$. First-step matching therefore recovers the performance gap across the full evaluated shot range. Table~\ref{tab:additional_p2_full} reports every dataset--shot result.

\begin{table}[htbp]
\centering
\small
\caption{Fixed-horizon CLIP accuracy averaged over six datasets and three seeds at each shot level. Recovery is relative to the SGD--Adam accuracy gap. Adam-M0 uses only one initial global update-budget correction.}
\label{tab:additional_p2_by_shot}
\begin{tabular}{rrrrr}
\toprule
$K$ & SGD & Adam & Adam-M0 & Recovery (\%) \\
\midrule
1 & 66.37 & 6.30 & 69.28 & 104.84 \\
2 & 72.28 & 7.35 & 72.95 & 101.03 \\
4 & 75.24 & 10.64 & 76.25 & 101.55 \\
8 & 78.53 & 18.25 & 79.18 & 101.07 \\
16 & 82.23 & 29.98 & 82.49 & 100.49 \\
\midrule
Mean & 74.93 & 14.51 & 76.03 & 101.82 \\
\bottomrule
\end{tabular}
\end{table}

The raw Adam/SGD first-update ratio averages between $139.14$ and $150.63$ across the five shot levels. Adam-M0 reduces the applied ratio to approximately one in every paired setting. This intervention changes the update scale without replacing the few-shot gradient or supplying additional labels. It directly supports the importance of initial update scale in the fixed-protocol collapse. The no-augmentation comparison gives a mean SGD--Adam accuracy gap of $60.64$ points across its $27$ paired settings, consistent with the augmented comparison.

\begin{longtable}{lrrrr}
\caption{Final CLIP accuracy (\%, mean $\pm$ sample standard deviation over seeds $1,2,3$) after 30 epochs. Each dataset--shot comparison uses identical support membership across SGD, Adam, and Adam-M0.}\label{tab:additional_p2_full}\\
\toprule
Dataset & $K$ & SGD & Adam & Adam-M0 \\
\midrule
\endfirsthead
\multicolumn{5}{c}{Table~\thetable{} continued}\\
\toprule
Dataset & $K$ & SGD & Adam & Adam-M0 \\
\midrule
\endhead
\bottomrule
\endfoot
EuroSAT & 1 & $70.86\pm9.69$ & $20.69\pm3.26$ & $73.86\pm7.43$ \\
EuroSAT & 2 & $82.28\pm4.29$ & $13.50\pm7.35$ & $81.82\pm4.49$ \\
EuroSAT & 4 & $84.66\pm2.23$ & $23.40\pm3.56$ & $86.50\pm1.35$ \\
EuroSAT & 8 & $88.74\pm2.48$ & $23.76\pm1.93$ & $89.79\pm1.42$ \\
EuroSAT & 16 & $90.85\pm3.45$ & $35.51\pm15.60$ & $91.62\pm1.10$ \\
Oxford-Pets & 1 & $92.48\pm0.26$ & $3.29\pm0.55$ & $92.37\pm0.12$ \\
Oxford-Pets & 2 & $91.82\pm0.41$ & $2.71\pm0.03$ & $93.00\pm0.12$ \\
Oxford-Pets & 4 & $93.07\pm0.17$ & $2.99\pm0.39$ & $93.25\pm0.25$ \\
Oxford-Pets & 8 & $93.34\pm0.18$ & $18.27\pm24.68$ & $93.69\pm0.09$ \\
Oxford-Pets & 16 & $94.29\pm0.52$ & $21.46\pm16.49$ & $94.44\pm0.26$ \\
DTD & 1 & $53.59\pm0.85$ & $10.32\pm6.64$ & $54.29\pm1.35$ \\
DTD & 2 & $59.81\pm1.09$ & $10.84\pm7.55$ & $59.91\pm1.33$ \\
DTD & 4 & $65.01\pm0.75$ & $21.49\pm3.77$ & $65.11\pm0.18$ \\
DTD & 8 & $68.09\pm1.04$ & $22.46\pm9.95$ & $67.65\pm1.77$ \\
DTD & 16 & $72.48\pm0.55$ & $39.64\pm4.08$ & $72.62\pm0.62$ \\
FGVC-Aircraft & 1 & $28.41\pm1.82$ & $0.99\pm0.14$ & $31.01\pm1.06$ \\
FGVC-Aircraft & 2 & $32.14\pm3.03$ & $1.14\pm0.23$ & $34.66\pm1.68$ \\
FGVC-Aircraft & 4 & $35.74\pm2.27$ & $1.18\pm0.21$ & $38.90\pm1.17$ \\
FGVC-Aircraft & 8 & $43.38\pm0.58$ & $1.87\pm0.18$ & $45.38\pm1.36$ \\
FGVC-Aircraft & 16 & $53.35\pm1.10$ & $3.38\pm1.33$ & $53.58\pm0.72$ \\
Stanford-Cars & 1 & $59.28\pm9.02$ & $0.51\pm0.08$ & $70.21\pm0.26$ \\
Stanford-Cars & 2 & $73.27\pm1.35$ & $0.71\pm0.29$ & $73.96\pm0.53$ \\
Stanford-Cars & 4 & $77.42\pm0.48$ & $0.75\pm0.26$ & $78.17\pm0.17$ \\
Stanford-Cars & 8 & $81.85\pm0.39$ & $0.90\pm0.09$ & $82.68\pm0.28$ \\
Stanford-Cars & 16 & $86.10\pm0.53$ & $4.78\pm4.06$ & $86.45\pm0.23$ \\
Caltech101 & 1 & $93.62\pm0.13$ & $2.03\pm2.26$ & $93.94\pm0.06$ \\
Caltech101 & 2 & $94.36\pm0.72$ & $15.23\pm18.16$ & $94.36\pm0.40$ \\
Caltech101 & 4 & $95.56\pm0.35$ & $14.05\pm12.10$ & $95.54\pm0.00$ \\
Caltech101 & 8 & $95.78\pm0.15$ & $42.27\pm26.42$ & $95.86\pm0.19$ \\
Caltech101 & 16 & $96.29\pm0.15$ & $75.15\pm6.94$ & $96.21\pm0.22$ \\
\end{longtable}

\subsection{How calibration changes with the support size}
\label{app:additional_shots}

The extended diagnostics reuse the mean-gradient, second-moment, coordinate-mismatch, and adaptive-direction definitions in Section~\ref{sec:why_does_adam_fail_short} and Appendix~\ref{sec:why_does_adam_fail}. The encoder superscript is omitted when the statistic is computed over the concatenated trainable parameters. The second-moment reference is estimated from $64$ deterministic, class-balanced diagnostic batches of size $16$ for each split. The same batch construction is used for the support and full-training reference. These are squared batch-mean-gradient statistics. They are distinct from Adam's temporal moment state. This matched diagnostic sampling also differs from the original one-shot full-support probes, whose diagnostic batches contain up to $256$ examples.

Table~\ref{tab:additional_shot_diagnostics} reports medians over six datasets and three seeds on the SGD trajectory. It uses the Adam-like directions in Eq.~\eqref{eq:adam_short_directions}, including the same placement of $\varepsilon$ inside the square root. The initial mismatch $M_0$ decreases from $0.244$ to $0.120$ between one and 16 shots. The adaptive norm ratio decreases from $1.651$ to $1.032$, while the shared-denominator cosine $\cos(u_{fr,0},u_{rr,0})$ increases from $0.379$ to $0.755$. Thus, larger support sets improve both scale calibration and directional agreement.

\begin{table}[htbp]
\centering
\small
\setlength{\tabcolsep}{4pt}
\caption{Shot-dependent CLIP diagnostics. Each entry is the median over six datasets and three seeds. $R_v(0)$ is the initial second-moment norm ratio, $M_0$ is the coordinate mismatch, and $R_{u,0}$ is the adaptive-direction norm ratio. The cosine is $\cos(u_{fr,0},u_{rr,0})$, with the full-training denominator in both directions. The final column compares the second-moment and mean-gradient collapse factors along the SGD trajectory.}
\label{tab:additional_shot_diagnostics}
\begin{tabular}{rrrrrr}
\toprule
$K$ & $R_v(0)$ & $M_0$ & $R_{u,0}$ & Reference cosine & $C_v/C_m$ \\
\midrule
1 & 1.256 & 0.244 & 1.651 & 0.379 & 4.207 \\
2 & 1.116 & 0.209 & 1.390 & 0.493 & 4.266 \\
4 & 1.100 & 0.165 & 1.199 & 0.599 & 4.374 \\
8 & 0.995 & 0.140 & 1.118 & 0.687 & 3.447 \\
16 & 1.037 & 0.120 & 1.032 & 0.755 & 2.423 \\
\bottomrule
\end{tabular}
\end{table}

The arithmetic means give the same overall trend: initial coordinate mismatch decreases from $0.236$ to $0.126$ and the relative collapse factor from $7.45$ to $3.67$. These correspond to decreases of $46.5\%$ and $50.7\%$, respectively. Across the $30$ dataset--shot settings, the Spearman correlations with $\log_2 K$ are $-0.621$ for coordinate mismatch, $-0.910$ for adaptive norm inflation, and $0.738$ for the shared-denominator reference cosine. Their dataset-cluster $95\%$ intervals are $[-0.911,-0.521]$, $[-0.975,-0.894]$, and $[0.608,0.949]$.

Under matched batch sampling, the aggregate denominator ratio approaches one from either side. Coordinatewise mismatch remains informative even when this aggregate ratio is close to one. The final accuracy gap is $60.07$, $64.93$, $64.60$, $60.27$, and $52.24$ points for $K=1,2,4,8,16$, respectively. The calibration trend is therefore clearer than a monotone accuracy trend. Aircraft and Cars retain larger accuracy gaps as the support grows. Together with the matched intervention, these results extend the update-scale explanation beyond the original three one-shot diagnostic datasets.

\subsection{First-step versus repeated global matching}
\label{app:additional_mt}

The complete matching study contains 150 paired settings over all ten non-ImageNet datasets, five shot levels, and three seeds. The datasets are EuroSAT, Oxford-Pets, DTD, FGVC-Aircraft, Stanford-Cars, Caltech101, SUN397, Food101, Flowers102, and UCF101. Each pair uses identical support membership and configuration, apart from the matching policy. All 300 runs complete 30 epochs.

\begin{table}[htbp]
\centering
\small
\caption{Complete first-step versus every-step matching comparison. Each shot row averages ten datasets and three seeds. The difference is Adam-Mt minus Adam-M0 in percentage points.}
\label{tab:additional_matching_complete}
\begin{tabular}{rrrr}
\toprule
$K$ & Adam-M0 & Adam-Mt & Difference \\
\midrule
1 & 73.22 & 71.58 & $-1.64$ \\
2 & 76.46 & 74.41 & $-2.05$ \\
4 & 79.29 & 77.86 & $-1.43$ \\
8 & 81.42 & 80.59 & $-0.83$ \\
16 & 84.21 & 84.00 & $-0.20$ \\
\midrule
Mean & 78.92 & 77.69 & $-1.23$ \\
\bottomrule
\end{tabular}
\end{table}

Table~\ref{tab:additional_matching_complete} shows that Adam-M0 reaches $78.92\%$, compared with $77.69\%$ for Adam-Mt. The paired Adam-Mt minus Adam-M0 difference is $-1.23$ points, with a dataset-cluster $95\%$ interval of $[-2.02,-0.47]$. Adam-M0 has higher mean accuracy at every shot level. Its margin decreases from $1.64$ points at one shot to $0.20$ at 16 shots. Table~\ref{tab:additional_matching_complete_full} gives the complete dataset--shot results. These comparisons show that repeated global matching is not needed to obtain the first-step intervention's fixed-horizon accuracy.

\begin{longtable}{lrrr}
\caption{Complete Adam-M0 and Adam-Mt study: final accuracy (\%, mean $\pm$ sample standard deviation over three seeds) on ten datasets and five shot levels. Both policies use the same support examples and 30-epoch protocol.}\label{tab:additional_matching_complete_full}\\
\toprule
Dataset & $K$ & Adam-M0 & Adam-Mt \\
\midrule
\endfirsthead
\multicolumn{4}{c}{Table~\thetable{} continued}\\
\toprule
Dataset & $K$ & Adam-M0 & Adam-Mt \\
\midrule
\endhead
\bottomrule
\endfoot
Caltech101 & 1 & $93.87\pm0.07$ & $93.39\pm0.52$ \\
Caltech101 & 2 & $94.44\pm0.49$ & $94.01\pm0.69$ \\
Caltech101 & 4 & $95.43\pm0.20$ & $94.98\pm0.37$ \\
Caltech101 & 8 & $95.88\pm0.43$ & $95.67\pm0.26$ \\
Caltech101 & 16 & $96.25\pm0.27$ & $96.21\pm0.10$ \\
DTD & 1 & $53.70\pm1.33$ & $53.82\pm1.31$ \\
DTD & 2 & $60.32\pm1.39$ & $60.07\pm1.52$ \\
DTD & 4 & $64.81\pm0.44$ & $64.83\pm0.47$ \\
DTD & 8 & $67.59\pm1.13$ & $67.28\pm1.21$ \\
DTD & 16 & $72.68\pm1.04$ & $73.25\pm0.28$ \\
EuroSAT & 1 & $74.21\pm4.79$ & $73.32\pm6.01$ \\
EuroSAT & 2 & $82.79\pm0.52$ & $81.84\pm0.74$ \\
EuroSAT & 4 & $87.12\pm1.00$ & $86.53\pm0.96$ \\
EuroSAT & 8 & $87.57\pm2.09$ & $88.89\pm0.52$ \\
EuroSAT & 16 & $92.15\pm0.72$ & $92.60\pm1.39$ \\
FGVC-Aircraft & 1 & $31.27\pm0.88$ & $27.51\pm1.23$ \\
FGVC-Aircraft & 2 & $34.25\pm2.45$ & $28.35\pm1.64$ \\
FGVC-Aircraft & 4 & $38.95\pm0.80$ & $35.47\pm0.37$ \\
FGVC-Aircraft & 8 & $45.66\pm1.07$ & $43.59\pm1.44$ \\
FGVC-Aircraft & 16 & $53.64\pm1.68$ & $55.02\pm0.95$ \\
Food101 & 1 & $85.57\pm0.51$ & $83.91\pm0.42$ \\
Food101 & 2 & $85.85\pm0.24$ & $82.66\pm0.59$ \\
Food101 & 4 & $86.30\pm0.08$ & $83.30\pm0.46$ \\
Food101 & 8 & $86.07\pm0.16$ & $83.62\pm0.52$ \\
Food101 & 16 & $86.76\pm0.19$ & $84.41\pm0.39$ \\
Flowers102 & 1 & $84.61\pm0.51$ & $86.37\pm0.34$ \\
Flowers102 & 2 & $89.86\pm0.47$ & $89.89\pm0.67$ \\
Flowers102 & 4 & $93.38\pm0.46$ & $93.41\pm0.26$ \\
Flowers102 & 8 & $95.55\pm0.29$ & $95.29\pm1.20$ \\
Flowers102 & 16 & $97.51\pm0.10$ & $97.90\pm0.19$ \\
Oxford-Pets & 1 & $92.12\pm0.14$ & $91.88\pm0.10$ \\
Oxford-Pets & 2 & $92.48\pm0.36$ & $90.00\pm0.68$ \\
Oxford-Pets & 4 & $93.28\pm0.24$ & $92.17\pm0.34$ \\
Oxford-Pets & 8 & $93.75\pm0.21$ & $93.23\pm0.22$ \\
Oxford-Pets & 16 & $94.23\pm0.21$ & $93.39\pm0.72$ \\
Stanford-Cars & 1 & $70.51\pm0.49$ & $62.52\pm0.98$ \\
Stanford-Cars & 2 & $74.09\pm0.15$ & $69.64\pm0.35$ \\
Stanford-Cars & 4 & $78.27\pm0.43$ & $75.38\pm1.17$ \\
Stanford-Cars & 8 & $82.59\pm0.28$ & $81.36\pm0.20$ \\
Stanford-Cars & 16 & $86.03\pm0.20$ & $86.36\pm0.41$ \\
SUN397 & 1 & $70.26\pm0.28$ & $67.62\pm0.69$ \\
SUN397 & 2 & $71.39\pm0.20$ & $68.77\pm0.65$ \\
SUN397 & 4 & $73.16\pm0.12$ & $70.83\pm0.47$ \\
SUN397 & 8 & $74.95\pm0.22$ & $72.75\pm0.43$ \\
SUN397 & 16 & $76.12\pm0.27$ & $74.31\pm0.28$ \\
UCF101 & 1 & $76.09\pm0.70$ & $75.44\pm1.33$ \\
UCF101 & 2 & $79.13\pm1.20$ & $78.85\pm1.28$ \\
UCF101 & 4 & $82.17\pm1.04$ & $81.71\pm1.69$ \\
UCF101 & 8 & $84.57\pm0.13$ & $84.18\pm0.30$ \\
UCF101 & 16 & $86.69\pm0.19$ & $86.59\pm0.29$ \\
\end{longtable}

\paragraph{Common comparison with raw Adam and SGD.}
A separate comparison evaluates raw Adam, Adam-M0, and global Adam-Mt on their common set of completed runs. This intersection contains Aircraft, Oxford-Pets, and Stanford-Cars at $K\in\{1,4,16\}$ with three seeds, giving $27$ paired settings. The SGD reference is taken from the shot-scaling study, matched by dataset, shot count, seed, and support examples.

\begin{table}[htbp]
\centering
\small
\setlength{\tabcolsep}{4pt}
\caption{First-step and every-step global budget matching on the three-dataset common comparison. Accuracies are averaged over three seeds and datasets. The last two columns report signed recovery of the SGD--Adam gap.}
\label{tab:additional_matching}
\begin{tabular}{rrrrrrr}
\toprule
$K$ & SGD & Adam & Adam-M0 & Adam-Mt & M0 recovery & Mt recovery \\
 & & & & & (\%) & (\%) \\
\midrule
1 & 60.06 & 1.75 & 64.64 & 60.64 & 107.85 & 101.00 \\
4 & 68.74 & 1.62 & 70.17 & 67.68 & 102.12 & 98.41 \\
16 & 77.91 & 8.86 & 77.97 & 78.25 & 100.07 & 100.49 \\
\midrule
Mean & 68.91 & 4.07 & 70.92 & 68.86 & 103.11 & 99.92 \\
\bottomrule
\end{tabular}
\end{table}

Adam-M0 reaches $70.92\%$, and Adam-Mt reaches $68.86\%$, compared with $68.91\%$ for the SGD reference and $4.07\%$ for raw Adam. Their signed recoveries are $103.11\%$ and $99.92\%$, with intervals $[99.92,105.70]$ and $[99.05,100.67]$. Repeated global matching therefore provides no average accuracy gain beyond the initial correction. Adam-Mt exceeds Adam-M0 in five settings and is lower in 22 settings. The mean difference is $-2.07$ points, with interval $[-3.52,-0.73]$.

Across the complete 150-run Adam-M0 collection, the median applied first-step Adam-M0/SGD ratio is $0.99999945$. The maximum relative matching error is below $9.90\times10^{-6}$. All 150 Adam-Mt runs keep the global matching error below $10^{-5}$ over their recorded updates. These measurements verify the applied global scale constraint while preserving the distinction between global and blockwise matching. The broad paired comparison and the common-reference comparison both favor the simpler one-time correction at this horizon.

\subsection{Learning-rate reduction and warmup}
\label{app:additional_lr_warmup}

The shared learning rate in the main benchmark tests optimizer behavior under a common recipe. The following controls examine whether direct scale adjustments restore Adam's accuracy. We compare seven fixed Adam learning rates and four warmup schedules on the common six-dataset grid at $K\in\{1,4,16\}$ with three seeds. All variants have $54$ observations. The SGD reference is the matching subset of the shot-scaling study, with identical support membership. No candidate rate or checkpoint is selected using the test trajectory.

\begin{table}[htbp]
\centering
\small
\setlength{\tabcolsep}{4pt}
\caption{Learning-rate and warmup controls after 30 epochs. Accuracy is averaged over the common six-dataset, three-shot, three-seed grid. The difference and its paired dataset-cluster $95\%$ interval are relative to SGD at $10^{-4}$. Warmup variants use base learning rate $10^{-4}$; the listed fraction is the fraction of training updates used for warmup, followed by a constant or cosine schedule.}
\label{tab:additional_lr_warmup}
\begin{tabular}{lrrr}
\toprule
Optimizer and schedule & Accuracy & Difference & $95\%$ interval \\
\midrule
SGD, $10^{-4}$ & 74.61 & $0.00$ & --- \\
Adam, $10^{-4}$ & 15.04 & $-59.57$ & $[-72.80, -46.34]$ \\
Adam, $3\!\times\!10^{-5}$ & 55.34 & $-19.28$ & $[-24.45, -13.85]$ \\
Adam, $10^{-5}$ & 68.39 & $-6.23$ & $[-8.59, -4.10]$ \\
Adam, $3\!\times\!10^{-6}$ & 74.25 & $-0.36$ & $[-1.45, +0.57]$ \\
Adam, $10^{-6}$ & 76.10 & $+1.48$ & $[+0.32, +2.89]$ \\
Adam, $3\!\times\!10^{-7}$ & 73.58 & $-1.03$ & $[-3.40, +0.91]$ \\
Adam, $10^{-7}$ & 69.32 & $-5.29$ & $[-10.28, -1.62]$ \\
Adam, 5\% warmup + constant & 74.36 & $-0.26$ & $[-1.29, +0.62]$ \\
Adam, 5\% warmup + cosine & 76.28 & $+1.67$ & $[+0.22, +3.21]$ \\
Adam, 20\% warmup + constant & 74.70 & $+0.08$ & $[-1.26, +1.34]$ \\
Adam, 20\% warmup + cosine & 76.26 & $+1.65$ & $[+0.09, +3.23]$ \\
\bottomrule
\end{tabular}
\end{table}

Table~\ref{tab:additional_lr_warmup} shows that reducing Adam's rate from $10^{-4}$ to $10^{-6}$ raises final accuracy from $15.04\%$ to $76.10\%$. The SGD reference reaches $74.61\%$. Five-percent warmup followed by cosine decay reaches $76.28\%$, while 20\% warmup followed by cosine decay reaches $76.26\%$. Constant post-warmup schedules reach $74.36\%$ and $74.70\%$, respectively. These controls show that update-scale correction can be achieved through conventional scheduling as well as explicit budget matching.

The fixed-horizon recovery supports the optimizer-centered interpretation. An unsuccessful dense Adam run at a shared nominal rate does not establish that the dense parameterization is unsuitable. A smaller rate or warmup reduces early displacement, while Adam-M0 directly calibrates that displacement against the SGD budget. The next subsection separates this 30-epoch recovery from behavior over longer training horizons.

\begin{figure}[p]
\centering
\includegraphics[width=\linewidth,height=0.83\textheight,keepaspectratio]{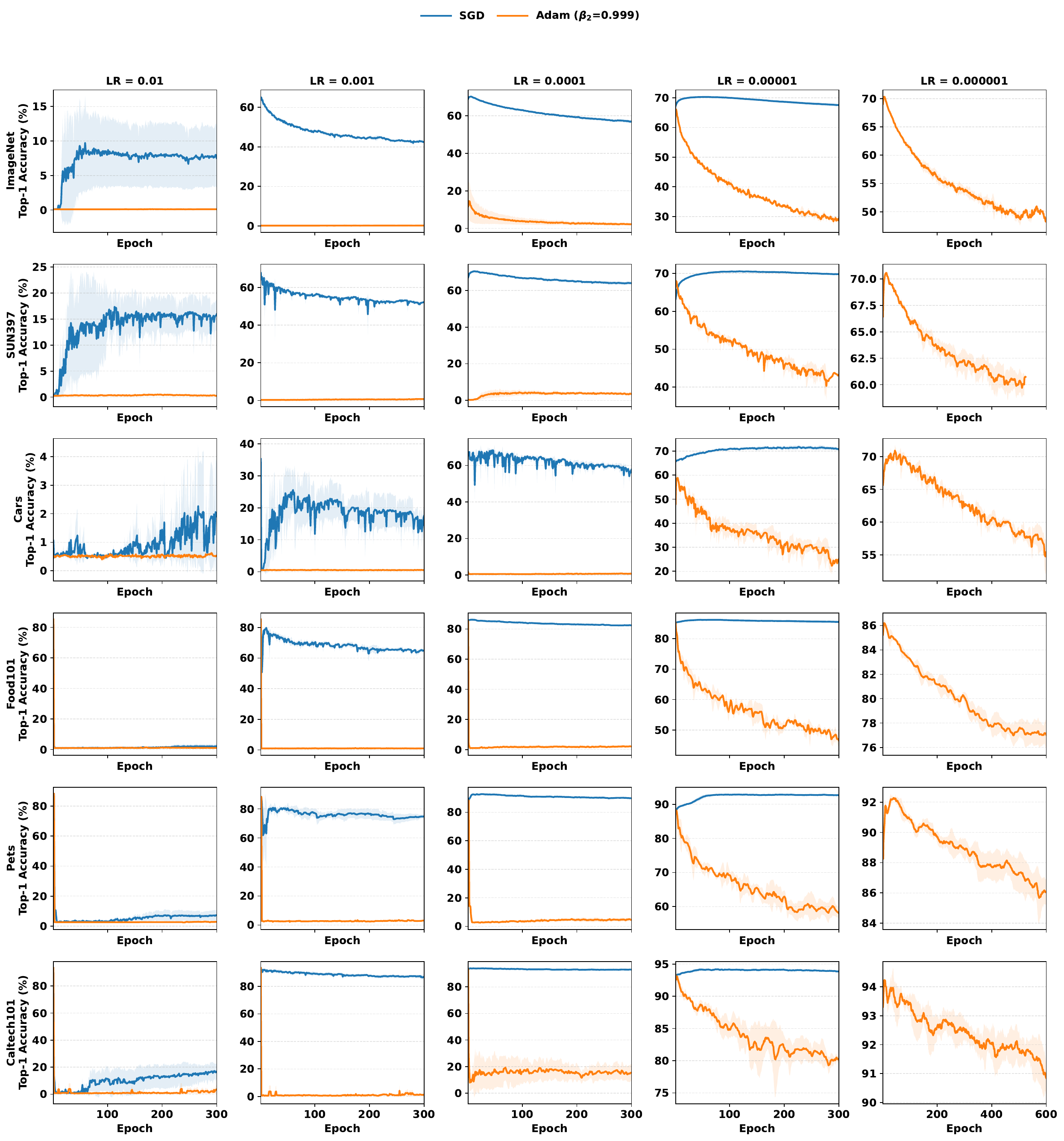}
\caption{Additional one-shot SGD and Adam trajectories on six datasets, using ViT-B/16 and three seeds. Columns correspond to the five listed learning rates. The $10^{-6}$ column contains Adam only. This extension complements the four-rate analysis in Appendix~\ref{app:stability_analysis}. Fixed-epoch and last-observed summaries are distinguished in Table~\ref{tab:additional_long_horizon}.}
\label{fig:additional_long_horizon}
\end{figure}

\subsection{Learning-rate sensitivity over longer horizons}
\label{app:additional_long_horizon}

The additional one-shot trajectories cover ImageNet, SUN397, Stanford-Cars, Food101, Oxford-Pets, and Caltech101 with three seeds. Adam is evaluated at $10^{-2},10^{-3},10^{-4},10^{-5},10^{-6}$. SGD is evaluated at the first four rates. The first four rate settings target 300 epochs. Adam at $10^{-6}$ targets 600 epochs. Table~\ref{tab:additional_long_horizon} distinguishes fixed-epoch accuracy, the trajectory peak, and the last observed checkpoint.

\begin{table}[htbp]
\centering
\scriptsize
\setlength{\tabcolsep}{3pt}
\caption{Long-horizon one-shot accuracy over six datasets. Each optimizer--rate setting contains 18 trajectories. The second run count reports the number observed at epoch 300. Fixed-epoch means use these available observations. The drop is the mean within-run peak-to-last decrease in percentage points. Adam at $10^{-6}$ targets 600 epochs, with last-observed epochs ranging from 338 to 600.}
\label{tab:additional_long_horizon}
\begin{tabular}{llrrrrrrr}
\toprule
Optimizer & LR & Runs & Epoch 30 & Epoch 300 & \makecell{Runs at\\epoch 300} & Peak & Last & Drop \\
\midrule
Adam & $10^{-2}$ & 18 & 1.39 & 1.45 & 17 & 44.71 & 1.39 & 43.32 \\
Adam & $10^{-3}$ & 18 & 0.89 & 1.18 & 17 & 44.78 & 1.15 & 43.62 \\
Adam & $10^{-4}$ & 18 & 4.70 & 4.78 & 17 & 48.14 & 4.73 & 43.41 \\
Adam & $10^{-5}$ & 18 & 64.71 & 47.30 & 17 & 77.17 & 46.96 & 30.21 \\
Adam & $10^{-6}$ & 18 & 79.70 & 73.29 & 18 & 80.88 & 69.95 & 10.93 \\
SGD & $10^{-2}$ & 18 & 4.15 & 8.51 & 18 & 50.00 & 8.51 & 41.49 \\
SGD & $10^{-3}$ & 18 & 61.55 & 56.44 & 18 & 72.98 & 56.44 & 16.54 \\
SGD & $10^{-4}$ & 18 & 78.76 & 73.95 & 18 & 80.60 & 73.95 & 6.65 \\
SGD & $10^{-5}$ & 18 & 79.66 & 80.05 & 18 & 81.07 & 80.05 & 1.02 \\
\bottomrule
\end{tabular}
\end{table}

At $10^{-6}$, Adam reaches $79.70\%$ at epoch 30 and $73.29\%$ at epoch 300. Its mean peak-to-last decrease is $10.93$ points. All 18 trajectories decrease by more than one point from their peaks, and 15 decrease by more than five points. At the common rate $10^{-5}$, Adam's mean peak-to-last decrease is $30.21$ points, compared with $1.02$ for SGD. The Adam summary includes one trajectory ending at epoch 281. All SGD trajectories reach epoch 300. %

The smaller Adam rate substantially improves early accuracy and delays degradation. The longer trajectories show why fixed-horizon recovery and stability across training time are separate properties. Figure~\ref{fig:additional_long_horizon} provides the complete additional rate sweep. Its trajectory peaks describe sensitivity to training duration and do not define a model-selection procedure.

\subsection{Replication with SigLIP}
\label{app:additional_siglip}

We repeat the fixed-horizon comparison with SigLIP~\cite{zhai2023sigmoid} (\texttt{google/siglip-base-patch16-224}), using revision \texttt{7fd15f0689c79d79e38b1c2e2e2370a7bf2761ed}. The four datasets are EuroSAT, FGVC-Aircraft, DTD, and Oxford-Pets. Each uses $K\in\{1,4,16\}$ and seeds $1,2,3$. Training uses 30 epochs, nominal batch size $64$, base learning rate $10^{-4}$, and the optimizer settings in Appendix~\ref{app:additional_protocol}. No scheduler, warmup, weight decay, clipping, gradient accumulation, or early stopping is used.

The native $224$-pixel preprocessing uses mean and standard deviation $0.5$. Training augmentation adds random resized crops and horizontal flips. The text prompt is \texttt{a photo of a \{\}.} The Transformer blocks of both towers are trainable, totaling $170{,}108{,}928$ parameters. Embeddings, projection and pooling modules, post-tower modules, and the native logit scale and bias remain fixed. This experiment preserves SigLIP's native preprocessing and logit calibration.

\begin{table}[htbp]
\centering
\small
\caption{SigLIP final accuracy (\%, mean $\pm$ sample standard deviation over three seeds) after 30 epochs. All three optimizers use the same support examples within each setting.}
\label{tab:additional_siglip_full}
\begin{tabular}{lrrrr}
\toprule
Dataset & $K$ & SGD & Adam & Adam-M0 \\
\midrule
EuroSAT & 1 & $72.56\pm5.97$ & $13.94\pm3.92$ & $72.88\pm4.39$ \\
EuroSAT & 4 & $79.41\pm1.07$ & $28.40\pm2.36$ & $82.29\pm0.92$ \\
EuroSAT & 16 & $89.20\pm0.08$ & $61.49\pm5.99$ & $91.38\pm1.14$ \\
Oxford-Pets & 1 & $94.43\pm0.54$ & $34.30\pm2.50$ & $94.47\pm0.36$ \\
Oxford-Pets & 4 & $94.95\pm0.23$ & $58.03\pm5.84$ & $95.05\pm0.09$ \\
Oxford-Pets & 16 & $94.69\pm0.28$ & $76.45\pm2.63$ & $94.75\pm0.26$ \\
DTD & 1 & $67.97\pm0.63$ & $29.98\pm2.33$ & $67.77\pm2.37$ \\
DTD & 4 & $75.53\pm0.47$ & $48.29\pm0.92$ & $75.20\pm0.90$ \\
DTD & 16 & $79.55\pm0.31$ & $56.78\pm1.41$ & $79.12\pm0.24$ \\
FGVC-Aircraft & 1 & $48.67\pm1.45$ & $5.04\pm1.97$ & $47.78\pm0.56$ \\
FGVC-Aircraft & 4 & $55.48\pm0.27$ & $10.77\pm7.95$ & $54.67\pm0.80$ \\
FGVC-Aircraft & 16 & $62.89\pm0.51$ & $43.14\pm5.11$ & $63.86\pm0.35$ \\
\midrule
Mean & & 76.28 & 38.88 & 76.60 \\
\bottomrule
\end{tabular}
\end{table}

\begin{table}[htbp]
\centering
\small
\caption{Global SigLIP mechanism diagnostics on EuroSAT and DTD. Entries are medians over two datasets and three seeds at each shot level. The collapse factor is measured along SGD trajectories. These diagnostics use the same meanings of $R_m$, $R_v$, $M_0$, $C_m$, and $C_v$ as the CLIP study; their reference batches use SigLIP gradients.}
\label{tab:additional_siglip_mechanism}
\begin{tabular}{rrrrr}
\toprule
$K$ & $R_m(0)$ & $R_v(0)$ & $M_0$ & $C_v/C_m$ \\
\midrule
1 & 1.410 & 0.919 & 0.319 & 3.492 \\
16 & 1.033 & 1.004 & 0.137 & 1.531 \\
\bottomrule
\end{tabular}
\end{table}

\noindent\begin{minipage}{\linewidth}
\setlength{\parskip}{5.5pt}
Table~\ref{tab:additional_siglip_full} shows that SGD exceeds raw Adam in all 12 dataset--shot settings. Their mean difference is $37.39$ points, with interval $[31.61,43.34]$. Adam-M0 reaches $76.60\%$, recovering $100.87\%$ of the signed SGD--Adam gap, with interval $[99.14,103.04]$. Its difference from SGD is $0.32$ points, with interval $[-0.28,1.29]$. The paired scale correction therefore yields mean accuracy comparable to SGD on the evaluated SigLIP settings.

The raw first-update Adam/SGD ratio ranges from $70.25$ to $400.77$ across the 12 dataset--shot means. Mechanistic trajectories are additionally measured on EuroSAT and DTD at $K\in\{1,16\}$ with three seeds. Table~\ref{tab:additional_siglip_mechanism} reports global statistics on this subset. Initial coordinate mismatch decreases from $0.319$ to $0.137$, and the second-moment-to-gradient collapse-factor ratio decreases from $3.492$ to $1.531$. The SigLIP result thus combines a second-model accuracy replication with a direct update-budget intervention.
\end{minipage}

\end{document}